%% file: deep-neural-networks-marine-debris-detection-oneside.tex
\documentclass[a4paper, notoc, oneside, openany]{tufte-book} 

\usepackage{mparhack}

\usepackage[utf8]{inputenc}
\usepackage[caption=false]{subfig}

\usepackage{ragged2e}

\usepackage{fancyhdr}
\fancypagestyle{mainmatter}{    
    \fancyfoot{}
    \fancyfoot[C]{ \centering \thepage }
}

\fancypagestyle{plain}{%
    \fancyhf{}%
    \fancyfoot{}
    \fancyfoot[C]{ \centering \thepage }%
}

\titleformat{\chapter}[block]
{\huge\rmfamily\color{darkgray}}
{\colorbox{black}{\parbox{1.5cm}{\hfill\huge\color{white}\thechapter}}}
{5pt}
{}
[]

\titleformat{\section}%
{\Large\rmfamily\color{darkgray}}
{\colorbox{darkgray}{\parbox{1.5cm}{\hfill\Large\color{white}\thesection}}}
{5pt}
{}
[]

\titleformat{\subsection}%
{\Large\rmfamily\color{darkgray}}
{\colorbox{gray}{\parbox{2.0cm}{\hfill\Large{\color{white}\thesubsection}}}}
{5pt}
{}
[]

\usepackage{pdfpages}

\usepackage{amsmath}
\usepackage{amssymb}

\DeclareMathOperator*{\argmax}{arg\,max}

\usepackage{longtable}

\usepackage{microtype} 
\usepackage{lipsum} 
\usepackage{booktabs} 

\usepackage{multirow}

\usepackage{graphicx} 
\setkeys{Gin}{width=\linewidth,totalheight=\textheight,keepaspectratio} 

\usepackage{fancyvrb} 
\fvset{fontsize=\normalsize} 

\usepackage{tikz, pgfplots, pgfplotstable}
\usetikzlibrary{calc}
\usetikzlibrary{positioning}
\usepgflibrary{fpu}

\usepgfplotslibrary{external}
\usepgfplotslibrary{fillbetween}
\usepgfplotslibrary{colorbrewer}

\usetikzlibrary{external}
\pgfplotsset{compat=newest}

\newcommand{\errorband}[6]{
	\pgfplotstableread{#1}\datatable
	\addplot [name path=pluserror,draw=none,no markers,forget plot]
	table [x={#2},y expr=\thisrow{#3}+\thisrow{#4} / 2] {\datatable};
	
	\addplot [name path=minuserror,draw=none,no markers,forget plot]
	table [x={#2},y expr=\thisrow{#3}-\thisrow{#4} / 2] {\datatable};
	
	\addplot [forget plot,fill=#5,opacity=#6]
	fill between[on layer={},of=pluserror and minuserror];
	
	\addplot [#5,thick,no markers]
	table [x={#2},y={#3}] {\datatable};
}

\newenvironment{customlegend}[1][]{%
	\begingroup
	\pgfplots@init@cleared@structures
	\pgfplotsset{#1}%
}{%
	\pgfplots@createlegend
	\endgroup
}%

\def\addlegendimage{\pgfplots@addlegendimage}

\usepackage{booktabs}

\usepackage{xspace} 

\newcommand{\monthyear}{\ifcase\month\or January\or February\or March\or April\or May\or June\or July\or August\or September\or October\or November\or December\fi\space\number\year} 

\usepackage{units} 

\newcommand{\hlred}[1]{\textcolor{Maroon}{#1}} 
\newcommand{\hangleft}[1]{\makebox[0pt][r]{#1}} 
\newcommand{\tuftebs}{\symbol{'134}} 

\providecommand{\XeLaTeX}{X\lower.5ex\hbox{\kern-0.15em\reflectbox{E}}\kern-0.1em\LaTeX}

\newcommand{\doccmddef}[2][]{\hlred{\texttt{\tuftebs#2}}\label{cmd:#2}\ifthenelse{\isempty{#1}} 
{ 
\index{#2 command@\protect\hangleft{\texttt{\tuftebs}}\texttt{#2}}
}
{ 
\index{#2 command@\protect\hangleft{\texttt{\tuftebs}}\texttt{#2} (\texttt{#1} package)}
\index{#1 package@\texttt{#1} package}\index{packages!#1@\texttt{#1}}
}}

\newcommand{\doccmd}[2][]{
\texttt{\tuftebs#2}%
\ifthenelse{\isempty{#1}}
{%
\index{#2 command@\protect\hangleft{\texttt{\tuftebs}}\texttt{#2}}
}
{%
\index{#2 command@\protect\hangleft{\texttt{\tuftebs}}\texttt{#2} (\texttt{#1} package)}
\index{#1 package@\texttt{#1} package}\index{packages!#1@\texttt{#1}}
}}

\usepackage{makeidx} 
\makeindex 

\graphicspath{{chapters/images/}}

\usepackage{stackengine}

\g@addto@macro\normalsize{%
    \setlength\abovedisplayskip{-3.5pt}
    \setlength\belowdisplayskip{2.5pt}
    \setlength\abovedisplayshortskip{-3.5pt}
    \setlength\belowdisplayshortskip{2.5pt}
}

\makeatletter
\newcommand{\emptypage}[1]{%
    \cleardoublepage
    \begingroup
    \let\ps@plain\ps@empty
    \pagestyle{empty}
    #1
    \cleardoublepage}
\makeatletter

\usepackage{appendix}

\renewcommand*\l@figure{\@dottedtocline{1}{1.5em}{2.3em}}
\renewcommand*\l@table{\@dottedtocline{1}{1.5em}{2.3em}}

\title[DNNs for Marine Debris Detection in Sonar Images]{Deep Neural Networks for Marine Debris Detection in Sonar Images}

\author{Matias A. Valdenegro Toro} 

\publisher{Thesis submitted for the degree of\\Doctor of Philosophy in Electrical Engineering} 

\begin{document}

\frontmatter





\cleardoublepage%
{%
    \begin{fullwidth}%
        \vspace*{3cm}%
        \fontsize{30}{32}\selectfont\textcolor{darkgray}{
            \centering
            \flushleft
            \hspace*{0.15\textwidth}Deep Neural Networks\\
            \hspace*{0.15\textwidth}For Marine Debris Detection\\
            \hspace*{0.15\textwidth}In Sonar Images\\            
        }%
        \sffamily%
        \vspace*{2cm}%
        \fontsize{18}{20}\selectfont\par\noindent\textcolor{darkgray}{
            \flushleft
            \hspace*{0.3\textwidth}Matias Alejandro\\
            \hspace*{0.3\textwidth}Valdenegro Toro, M.Sc., B.Sc.\\
            \vspace*{1cm}
            \hspace*{0.3\textwidth}April 2019.\\
        }%
        \vspace*{4.5cm}
        \fontsize{18}{20}\selectfont\par\noindent{
            \flushleft
            Dissertation submitted for the degree of\\ Doctor of Philosophy in Electrical Engineering.\\
            \vspace*{0.5cm}
            Heriot-Watt University, Edinburgh, Scotland.
        }
        \vfill
        \fontsize{14}{16}\selectfont\par\noindent{
            \flushleft
            The copyright in this thesis is owned by the author. Any quotation from the thesis 
            or use of any of the information contained in it must acknowledge this thesis as the source of the quotation or information.\\
        }%
    \end{fullwidth}%
}
\thispagestyle{empty}%
\clearpage%


\newpage
\begin{fullwidth}
~\vfill
\thispagestyle{empty}
\setlength{\parindent}{0pt}
\setlength{\parskip}{\baselineskip}
Copyright \copyright\ 2018 - 2019 Matias Alejandro Valdenegro Toro.

\par\smallcaps{Latex Template based on \url{https://github.com/Tufte-LaTeX/tufte-latex}, which is Licensed under the Apache License, Version 2.0.}

\par\textit{Final version, arXiv release, April 2019. Compiled on \today.}
\end{fullwidth}

\justifying
\cleardoublepage
\chapter{Abstract}
\thispagestyle{empty}
Garbage and waste disposal is one of the biggest challenges currently faced by mankind. Proper waste disposal and recycling is a must in any sustainable community, and in many coastal areas there is significant water pollution in the form of floating or submerged garbage. This is called marine debris.

It is estimated that 6.4 million tonnes of marine debris enter water environments every year \textit{[McIlgorm et al. 2008, APEC Marine Resource Conservation WG]}, with 8 million items entering each day. An unknown fraction of this sinks to the bottom of water bodies. Submerged marine debris threatens marine life, and for shallow coastal areas, it can also threaten fishing vessels \textit{[I{\~n}iguez et al. 2016, Renewable and Sustainable Energy Reviews]}.

Submerged marine debris typically stays in the environment for a long time (20+ years), and consists of materials that can be recycled, such as metals, plastics, glass, etc. Many of these items should not be disposed in water bodies as this has a negative effect in the environment and human health. 

Encouraged by the advances in Computer Vision from the use Deep Learning, we propose the use of Deep Neural Networks (DNNs) to survey and detect marine debris in the bottom of water bodies (seafloor, lake and river beds) from Forward-Looking Sonar (FLS) images.

This thesis performs a comprehensive evaluation on the use of DNNs for the problem of marine debris detection in FLS images, as well as related problems such as image classification, matching, and detection proposals. We do this in a dataset of 2069 FLS images that we captured with an ARIS Explorer 3000 sensor on marine debris objects lying in the floor of a small water tank. We had issues with the sensor in a real world underwater environment that motivated the use of a water tank.

The objects we used to produce this dataset contain typical household marine debris and distractor marine objects (tires, hooks, valves, etc), divided in 10 classes plus a background class.

Our results show that for the evaluated tasks, DNNs are a superior technique than the corresponding state of the art. There are large gains particularly for the matching and detection proposal tasks. We also study the effect of sample complexity and object size in many tasks, which is valuable information for practitioners.

We expect that our results will advance the objective of using Autonomous Underwater Vehicles to automatically survey, detect and collect marine debris from underwater environments. 


\newpage
\begin{flushright}
\noindent\fontsize{14}{18}\selectfont
Do not be sorry. Be better...\\ And you must be better than me...\\
\smallcaps{Kratos - God of War (SIEE, 2018)}\\
\vspace*{1cm}
Hope is what makes us strong.\\ It is why we are here.\\ It is what we fight with when all else is lost...\\
\smallcaps{Pandora - God of War 3 (SCEE, 2010)}
\end{flushright}
~\vfill
\begin{doublespace}
\thispagestyle{empty}
\noindent\fontsize{18}{22}\selectfont\itshape
\nohyphenation
Dedicada a mis queridos Abuelos: Paulina, Rubén, y Sergio.
\end{doublespace}
\vfill
\vfill

\chapter{Acknowledgements}
\thispagestyle{empty}
Many people have contributed or helped in the development of my Doctoral research. First I would like to thank the European Commission for funding my research through the FP7-PEOPLE-2013-ITN project ROBOCADEMY (Ref 608096). The project also supported me during my research visit to the German Research Center for Artificial Intelligence in Bremen, where I currently work. For this I acknowledge the support of Prof. David Lane, Lynn Smith, and Prof. Yvan Petillot during my PhD.

Acknowledging the support of the European Union is a must in the uncertain times during which this thesis was written. People have to know that the European Union is a project worth pursuing.\\

This vital funding through the Marie-Skłodowskca Curie Actions (MSCA) allowed me to freely develop my scientific ideas and steer the project into a direction I believe is positive for the environment and common good. Science must always be performed for the benefit of humankind and our planet. 

Leonard McLean at the Ocean Systems Lab provided key technical support in experimental setup and data capture. My Best Friend Polett Escanilla supported me during good and bad times. The Hochschule Bonn-Rhein-Sieg (Sankt Augustin, Germany) allowed me to advise their students in R\&D/Master Thesis projects for the MAS program, which gave me vital experience in advising and mentoring people. Prof. Dr. Paul Plöger, Iman Awaad, Anastassia Küstenmacher, Maria do Carmo Massoni, Kira Wazinsky, Miriam Lüdtke-Handjery, and Barbara Wieners-Horst from the previously mentioned institution provided key support after my time in Sankt Augustin and during my PhD when visiting the University.\\

The Hochschule Bonn-Rhein-Sieg was created as compensation to Bonn for the move of Germany's capital to Berlin after reunification, which makes me feel a special bond and debt to Germany. It was in this city that life-long friendships with Octavio Arriaga, Maryam Matin, and Nour Soufi, began.\\

Elizabeth Vargas provided invaluable support and physically delivered this thesis to the University's Academic Registry, and for this I am deeply indebted to her.

The Robotarium Cluster provided the key high-performance computing resources I needed to train over 1000 neural networks during my Doctoral Research. I acknowledge and thank Hans-Nikolai Viessmann for his support administrating the Cluster.\\

Finally, this thesis uses computational implementation of algorithms from third party libraries. For this I acknowledge the authors of \textit{Keras}, \textit{Theano}, and \textit{scikit-learn}. Without these open source libraries, the development of science would be much slower than currently is.


\pagestyle{mainmatter}
\setcounter{tocdepth}{2}
\tableofcontents 


\listoftables 


\listoffigures 


\chapter{Symbols and Abbreviations}

\setstretch{1.5} 

\begin{longtable}[c]{ll}
	
	$x_i$			& i-th element of vector $\textbf{x}$\\
	
	$\alpha$   		& Learning Rate\\
	$B$				& Batch Size\\
	$M$				& Number of Training Epochs or GD iterations\\
	
	$\hat{y}$		& Predicted Value\\
	$y$				& Ground Truth Scalar Label\\
	$x$				& Input Scalar\\
	$\textbf{x}$	& Input Vector\\
	$C$				& Number of Classes\\
					&   \\
	$Tr$			& Training Set\\
	$Vl$			& Validation Set\\
	$Ts$			& Test Set\\
					&   \\
	$\Theta$		& Parameter set of a neural network\\
	$\textbf{w}$	& Weight vector of a layer of a neural network\\
					&	\\
	\textbf{ML}		& Machine Learning\\			
	\textbf{CNN}	& Convolutional Neural Network(s)\\
	\textbf{DNN}	& Deep Neural Network(s)\\
	
	\textbf{GD}		& Gradient Descent\\
    \textbf{MGD}	& Mini-Batch Gradient Descent\\
	\textbf{SGD}	& Stochastic Gradient Descent\\
	
	\textbf{GPU}	& Graphics Processing Unit\\
	
	\textbf{FLS}	& Forward-Looking Sonar\\
	
\end{longtable}

\chapter{Neural Network Notation}

Through this thesis we use the following notation for neural network layers:

\begin{description}
	\item[Conv($f$, $w \times h$)] \hfill \\
		Two Dimensional Convolutional layer with $f$ filters of $w \times h$ spatial size.
	\item[MaxPool($w \times h$)] \hfill \\
		Two Dimensional Max-Pooling layer with spatial sub-sampling size $w \times h$.
	\item[AvgPool()] \hfill \\
		Two Dimensional Global Average Pooling layer.
	\item[FC($n$)] \hfill \\
		Fully Connected layer with $n$ neurons.
	
\end{description}

\mainmatter
\pagestyle{mainmatter}

\input{chapters/introduction.tex}
\input{chapters/marine-debris.tex}
\input{chapters/background.tex}
\input{chapters/sonar-image-classification.tex}
\input{chapters/limits-neural-networks.tex}
\input{chapters/matching-patches-sonar.tex}
\input{chapters/detection-proposals-sonar.tex}
\input{chapters/applications.tex}
\input{chapters/conclusions.tex}


\appendix
\input{chapters/appendices.tex}

\backmatter

\bibliography{thesis-biblio}
\bibliographystyle{plain}



\end{document}

%% file: chapters/introduction.tex
\chapter{Introduction}
\label{chapter:introduction}

Starting with the Industrial Revolution, human populations in every country have continuously been polluting the environment. Such contamination varies from biodegradable human waste to materials that take hundreds of years to degrade, such as plastics, heavy metals, and other processed materials. Virtually all aspects of development in modern times pollute the environment in one way or another.

Human-made pollution is problematic due to the negative effect it has on the environment, as well as on human health. Air and Water pollution are of special interest due to its particular impact in humans and animals. 

Massive efforts are on the way to reduce the impact of modern human life in the environment, like reducing the amount of produced waste, recycling materials that can be reused after some processing, and reusing products that have multiple uses or can have alternate uses. One aspect that has been left out is the collection and capture of existing pollutants in the environment. For example, collecting stray garbage in cities and/or beaches, or recovering spilled oil after a marine disaster. Even after the human race evolves into a state where no contamination is produced from its way of life, cleanup will be a must \cite{mcilgorm2008understanding}.

Robots and Autonomous Vehicles are a natural choice for this task, as it has been depicted in movies (Pixar's Wall-E) and novels. This concept has led to consumer robots like iRobot's Roomba and others, that can regularly clean small apartments with mixed results. We believe this is a worthy application of Robotics that can have massive impact in our lives, if it is implemented in a robust way.

A particular kind of pollution that is not covered by consumer robots and typical policy is marine pollution, specially marine debris. This kind of debris consists of human-made garbage that has found its way into bodies of water. This kind of pollution is usually ignored because it is not easily "seen". Discarding an object into water usually implies that the object sinks and it could be forgotten in the bottom of the water body.

There is an extensive scientific literature\cite{li2016plastic} about describing, locating, and quantifying the amount of marine debris found in the environment. There are reports of human-made discarded objects at up to 4000 meters deep at the coasts of California \cite[1em]{schlining2013debris}, and at more than 10K meters in the Mariana Trench \cite[1em]{chiba2018human}.

During trials at Loch Earn (Scotland, UK) We particularly saw the amount of submerged marine debris in the bottom of this lake. This experience was the initial motivation for this Doctoral research. For an Autonomous Underwater Vehicle, it would be a big challenge to detect and map objects with a large intra and inter-class variability such as marine debris.

This thesis proposes the use of Autonomous Underwater Vehicles to survey and recover/pick-up submerged marine debris from the bottom of a water body. This is an important problem, as we believe that contaminating our natural water sources is not a sustainable way of life, and there is evidence \cite[1em]{iniguez2016marine} that debris is made of materials that pollute and have a negative effect on marine environments \cite[1em]{sheavly2007marine}.

Most research in underwater object detection and classification deals with mine-like objects (MLOs). This bias is also affected by large streams of funding from different military sources around the world. We believe that a much more interesting and challenging problem for underwater perception is to find and map submerged marine debris.

There has been increasing and renewed interest in Neural Networks in the last 5 years\footnote{Since 2012}, producing breakthroughs in different fields related to Artificial Intelligence, such as image recognition, generative modeling, machine translation, etc. This drives us to also evaluate how state of the art neural networks can help us with the problem of detecting marine debris.

This Doctoral Thesis deals with the problem of detecting marine debris in sonar images. This is a challenging problem because of the complex shape of marine debris, and the set of possible objects that we want to detect is open and possibly unbounded.

For the purpose of object detection we use neural networks. This is not a novel application as neural networks have been used for object detection, but they have not been commonly applied in Forward-Looking Sonar images. This poses a different challenge due to the increased noise in the image (compared to color images captured with CCD/CMOS sensors) and the small sample size of the datasets that are typically used in underwater robotics.

We deal with several sub-problems related to the task of marine debris detection in Forward-Looking sonar images, divided into several research lines. This includes image classification, patch matching, detection proposals, end-to-end object detection, and tracking. We also perform fundamental experiments on how these methods behave as we vary the size of the training sets, including generalization on different objects.

While we propose that AUVs can be the solution, we do not deal with the full problem. This thesis only focuses on the perception problem of marine debris, and leaves the manipulation and grasping problem for future work.

\section{Research Questions}

This thesis investigates the following research questions, in the context of submerged marine debris detection in Forward-Looking sonar images.

\begin{itemize}
	\item \textbf{How can small objects (like marine debris) be detected in underwater environments?}\\
	Marine debris poses a particular challenge due to viewpoint dependence, complex shape, physical size, and large shape variability. It is not clear a priori which computer vision and machine learning techniques are the best in order to localize such objects in a sonar image.
	
	\item \textbf{How can objects be detected in a sonar image with the minimum number of assumptions?}
	Many object detection techniques make many assumptions on object shape. For example the use of Haar cascades works well for mine-like objects due to the size of their acoustic shadows, but fail when used in shadowless objects like marine debris. A similar argument can be constructed for template matching, which is typically used in sonar images.
	Reducing the number of implicit or explicit assumptions made by the algorithm will improve its performance on complex objects like marine debris.
	
	\item \textbf{How much data is needed to train a large neural network for sonar applications?}
	Deep Learning has only been made possible due to the availability of large labeled datasets of color images. Predictions cannot be made on how it will perform on sonar images (due to fundamental differences in the image formation process) and on radically smaller datasets that are more common in robotics.
	End-to-end learning is also problematic in small datasets as the "right" features can only be learned in massive datasets.
	
	\item \textbf{Is end-to-end learning possible in sonar images?}
	End-to-end learning consists of using a neural network to learn a mapping between an input and target labels without performing feature engineering or doing task-specific tuning. As mentioned before, this is problematic on datasets with low sample count and low object variability. We have not previously seen results and/or analysis of end-to-end task learning in sonar images. This will be useful as methods that are developed for one kind of sonar sensor could potentially work without effort on images produced by different sonar devices.
\end{itemize}

\section{Feasibility Analysis}

This thesis proposes the use of an Autonomous Underwater Vehicles for the purpose of surveying and capturing marine debris lying in the seafloor of a water body. In this subsection we make a short discussion of the operational concept that underpins the research proposal, from a practical point of view.

Our long term vision is that an AUV can be used to both survey and capture marine debris in any kind of underwater environment, like open and deep sea, lakes, river estuaries, swamps, etc.

The amount of marine debris varies with the type of water body. Deep sea usually has sparse samples of marine debris (due to its large size) measured at an average of 0.46 items per km$^2$ (median 0.14, range $[0.0019, 2.34]$, all items per km$^2$)\cite{katsanevakis2008marine}. For shallow coastal areas including river estuaries the average is higher, at 153 items per km$^2$ (median 139, range $[13.7, 320]$, all items per km$^2$) \footnotesize{$^6$}.
\normalsize
Land-based marine debris at beaches is highly likely to transfer to the sea. Debris density on beaches is average 2110 items per km$^2$ (median 840, range $[210, 4900]$, all items per km$^2$) \footnotesize{$^6$} \normalsize.

Two use cases for our proposed technology then arise:

\begin{description}
    \item[Capture of Marine Debris at Deep Sea] Marine debris in this case is sparse, implying that from an economic perspective it might not make much sense to survey and capture marine debris, as most of the seafloor is empty, and little marine debris can be recovered. If previous information (for example, plane wrecks, tsunamis, etc) that marine debris is present in a specific area of the deep sea, then it would be appropriate to use an AUV to recover these pieces of debris. This can be an important application in surveys related to accidents at sea.
    
    \item[Capture of Marine Debris in Shallow Coastal Areas] In this case marine debris is quite dense, depending on the specific area to be surveyed. It seems appropriate to recommend an automated solution using an AUV, as there is plenty of marine debris to be found.
\end{description}

As a simplified model for survey time, if we assume an AUV can move at velocity $V$ (in meters per second) with a sensor that can "see" $S$ meters at a time (related to the swath or range of the sensor), then in order to survey a area $A$ in squaree meters, the AUV will take approximately $T$ seconds, given by:

\begin{equation}
    T = \frac{A}{V \times S}
\end{equation}

Note that $S$ might be limited not only by the AUV's maximum speed, but also by the maximum operational speed that the sonar sensor requires (towing speed). For example, some sonar sensors collect information across time (multiple pings) and they might not operate correctly if the vehicle is moving too fast. High speeds might also produce motion blur in the produced images. Maximum towing speeds are also limited by depth range.

A typical value is $S = 2$ meters per second, which is limited by drag forces and energy efficiency \cite{fossen2011handbook}. Assuming $A = 1000000$ m$^2$ (one squared kilometer), then we can evaluate the required survey time as a function of the sensor range/swath $S$, shown in Figure \ref{intro:rangeVsTime}. For $S = 10$ m, 13.8 hrs are required, and this value drops to 42 minutes with $S = 200$ m.

\begin{marginfigure}
    \centering
    \begin{tikzpicture}
    \begin{axis}[domain=10:200, width=1.1\textwidth, xmin = 10, xmax = 200, tick label style={font=\scriptsize}, xlabel = {S [M]}, ylabel={T [Mins]}, ytick={900,700,500,300,100}, ymajorgrids=true, grid style=dashed,
    xtick={10,50,100,150,200}]
    \addplot+[mark=none, color=darkgray] {1000000/(2 * x * 60)};    
    \end{axis}
    \end{tikzpicture}
    \caption[Survey time in minutes as function of sensor range]{Survey time in minutes as function of sensor range for a 1 km$^2$ patch and $V=2$ m/s}
    \label{intro:rangeVsTime}
\end{marginfigure}

Assuming that there are $N$ marine debris elements in the seafloor per square kilometer, then every $\frac{T}{N}$ seconds a piece of marine debris will be found.

As mentioned before, for a shallow coastal area $N \in [13.7, 320]$, and taking $T = 42 \times 60$ s, this implies that $\frac{T}{N}$ is in the range $[7.8, 184.0]$ seconds. The lower bound implies that for the most dense debris distribution, one piece will be found every $8$ seconds. This motivates a real-time computational implementation, where processing one frame should at most take approximately one second. This requirement also depends on the sensor, as Forward-Looking sonars can provide data at up to 15 frames per second, also requiring a real-time perception implementation, up to $\frac{1}{15} = \sim 66.6$ milliseconds per frame.

For the marine debris use case, there are three important parameters to select an appropriate sonar sensor: the per-pixel spatial resolution, the maximum range, and the power usage. Marine debris objects can be quite small (less than 10 cm), so the highest spatial resolution is required. The maximum range ($S$) defines how many passes have to be performed to fully cover the desired environment. Power usage constrains the maximum battery for the AUV before needing to recharge.

To select these parameters, we made a short survey if different sonar sensing systems, available in Table \ref{intro:sonar-survey}. We considered three major kinds of sonars: Forward-Looking, Sidescan, and Synthetic Aperture. We only considered sensors that are appropriate as AUV payload, where the manufacturer provided all three previously mentioned parameters that we evaluate.

The longest AUVs can operate up to 20 hrs while performing seafloor mapping with a 2000 Watt-Hour battery \cite{kirkwood2007development}. Assuming this kind of battery, an AUV has enough power for several hours of endurance with an active sensor, but this varies as more power hungry sensors will deplete battery power faster. This relationship is shown in Figure \ref{intro:powerVsBatteryLife}. With the ARIS Explorer 3000 we can expect at most $33-125$ Hours of life, while with a more power consuming Kraken Aquapix SAS we can expect $14-15$ Hours. This calculation does not include power consumption by other subsystems of the AUV, such as propulsion, control, perception processing, and autonomy, so we can take them as strict maximums in the best case.

\begin{marginfigure}
    \centering
    \begin{tikzpicture}
    \begin{axis}[domain=6:145, restrict y to domain=1:350, width=1.1\textwidth, xmin = 6, xmax = 145, tick label style={font=\scriptsize}, xlabel = {Sensor Power [W]}, ylabel={Battery Life [Hrs]}, xtick={6,25,50,100,145},ytick={10,50,100,200,300}, ymajorgrids=true, grid style=dashed]
    \addplot+[mark=none, color=darkgray] {2000/x};    
    \end{axis}
    \end{tikzpicture}
    \caption[Battery life as function of sensor power requirement]{Battery life as function of sensor power requirement for a 2000 Watt-Hour battery}
    \label{intro:powerVsBatteryLife}
\end{marginfigure}

Since the ARIS Explorer 3000 has an approximate value $S = 10$ meters, surveying one km$^2$ will take 13.8 Hours. In the best possible case, $2.4-9$ km$^2$ of surface can be surveyed with a single battery charge. For the Kraken Aquapix SAS, which has $S = 200$ meters (approximately), surveying the one km$^2$ will take $0.7$ Hours, so with the available battery life, up to $20-21.4$ km$^2$ of surface can be surveyed with a single battery charge. This value is considerable better than the one for the ARIS.

To detect marine debris, we wish for the largest sensor resolution, ideally less than one centimeter per pixel, as marine debris elements are physically small ($< 10$ centimeters). Using a sensor with low resolution risks missing marine debris targets as they will be represented by less pixels in the output image, making classification and detection difficult. A higher resolution might also imply a high power consumption, as an active sensor will need a powerful signal to distinguish object from seafloor noise.

\begin{table*}[t]
    \centering
    \forcerectofloat
    \begin{tabular}{lllll}
        \hline 
        Type 		& Brand/Model Example	& Spatial Resolution 	& Max Range & Power \\ 
        \hline 
        FLS 	& ARIS Explorer		 	& $\sim 0.3$ cm/pix 		& 5-15 M 	& 16-60 W \\ 
                & Blueview M900-2250	& $\sim 0.6-1.3$ cm/pix		& 100 M 	& 20-26 W\\	
                & Blueview P900 Series	& $\sim 2.54$ cm/pix		& 100 M 	& 9-23 W\\
        \hline
        SS		& Tritech Starfish 		& $\sim 2.5-5$ cm/pix 		& 35-100 M 	& 6-12 W \\ 
                & Klein UUV-3500		& $\sim 2.4-4.8$ cm/pix		& 75-150 M 	& 18-30 W\\
                & Sonardyne Solstice	& $\sim 2.5-5$ cm/pix		& 200M 		& 18 W\\
        \hline
        SAS		& Kraken Aquapix		& $\sim 1.5-3.0$ cm/pixel	& 120-220 M & 130-145 W\\
                & Kongsberg HISAS 1030	& $< 5.0$ cm/pixel			& 200-260 M	& 100 W\\
                & Atlas Elektronik Vision 600 & $\sim 2.5$ cm/pixel	& 100 M 	& 100 W\\
        \hline
    \end{tabular}
    \caption[Survey of AUV Sonar Sensors across different manufacturers]{Survey of AUV Sonar Sensors across different manufacturers. We show Forward-Looking Sonars (FLS), Sidescan Sonars (SS), and Synthetic Aperture Sonars (SAS). Important parameters for our use case are the per-pixel spatial resolution (in centimeters), the maximum range (in meters), and power requirements (in watts).}
    \label{intro:sonar-survey}
    \vspace*{0.5cm}
\end{table*}

It is also important to mention that there is a trade-off between the frequency of the sound waves and the maximum range allowable for the sensor \cite[5em]{hansen2009introduction}. The amount of attenuation in water increases with the frequency of the sound wave, while a high frequency signal allows for more detail to be sensed from the environment. This effectively means that in order to sense marine debris, a high frequency sonar is required (such as the ARIS Explorer 300 at 1.8 MHz), but this limits the range that the sensor can see at a time, increasing the amount of time required to survey an area, as computed before.

The ARIS Explorer 3000 has a resolution of 0.3 cm per pixel, allowing it to see small objects easily. This value is $5-10$ times better than the best Sidescan or Synthetic Aperture sonar in our survey, indicating that the ARIS might be the best choice to detect marine debris in underwater environments.

We believe that Sidescan and Synthetic Aperture sonars are still useful for marine debris detection. For example, they can be used to quickly survey a large area of the seafloor, and to identify areas where marine debris might concentrate, and then use this information to direct an AUV using a high resolution sensor (like the ARIS) for a detailed survey of this area.

There are other possible use cases. In the case of surveying, a robot can more quickly detect areas of possible marine debris contamination at a large scale, and then direct human intervention, for example, with divers that can recover marine debris, or just by providing additional information over time, to locate sources of marine pollution. This would be useful for local authorities in coastal areas that wish to provide close monitoring of their shores.

We believe that possible users for this proposed technology are: governments and agencies interested in marine environments, specially in coastal areas, rivers, and lakes. Private companies that need marine debris mitigation, as well as ports and maritime commercial facilities, as these usually have large environmental impacts that include marine debris.

We note that the most feasible use of this proposed technology is on coastal areas close to shore, as the density of marine debris is higher. Open seas usually has a lower density of marine debris, and it is much larger in area, making debris location much more sparse, which reduces the technical feasibility of this solution, as an AUV would have to cooperate with a mother ship to provide energy recharges.

The marine debris task does not have widely known performance metrics \cite{sheavly2007marine}. Generally when using machine learning techniques, we wish to predict performance in unseen data, which is not trivial. For the object detection task, we need to predict how many debris objects can be successfully detected by our algorithms, but we also wish to produce an algorithm that maximizes the number of detected debris objects. This motivates the use of recall, as a false positive is not as troubling as missing one piece of marine debris in the seafloor. For classification tasks, we use accuracy as a proxy metric.

Our simplified analysis shows that the proposed technique is feasible, at least from an operational point of view, mostly for coastal areas, and not for open seas.

\section{Scope}

The thesis scope is limited by the following:

\begin{itemize}
    \item We only deal with the research problem of perception of marine debris in Forward-Looking Sonar images, leaving manipulation (picking) of marine debris as future work. This is mostly a pragmatic decision as manipulation in underwater environments is a harder problem \cite{ridao2014intervention}.
    \item The dataset used for training and evaluation of various techniques was captured in a water tank with a selection of marine debris and distractor objects. For marine debris we used household objects (bottles, cans, etc), and for distractors we selected a hook, an underwater valve, a tire, etc. The dataset is not intended to absolutely represent a real world underwater scenario. This selection is motivated by issues we had with the sonar system while trialing at lakes and rivers.
    \item We only consider submerged marine debris that is resting in the water body floor.
    \item Regarding sensing systems, we only use a high resolution Forward-Looking Sonar (Soundmetrics' ARIS Explorer 3000), and do not consider other kinds such as synthetic aperture or sidescan sonars.
\end{itemize}

\section{Thesis Structure}

Chapter \ref{chapter:motivation} describes the problem of submerged and floating marine debris in ocean and rivers in detail. From this we make the scientific proposal that Autonomous Underwater Vehicles can help to clean the environment.

We provide a technical introduction to basic Machine Learning and Neural Network techniques in Chapter \ref{chapter:background}, with a focus on practical issues that most textbooks do not cover.

Chapter \ref{chapter:sonar-classification} is the first technical chapter of this thesis. We deal with the sonar image classification problem, and show that neural networks can perform better than state of the art approaches, and even be usable in low power platforms such as AUVs.

In Chapter \ref{chapter:limits} we use an experimental approach to explore the practical limits of Convolutional Neural Networks, focusing on how the image size and training set size affect prediction performance. We also evaluate the use of transfer learning in sonar images.

Chapter \ref{chapter:matching} uses CNNs to match pairs of sonar image patches. We find this is a difficult problem, and state of the art solutions from Computer Vision perform poorly on sonar images, but Neural Networks can provide a good solution that generalizes well to different objects.

Chapter \ref{chapter:proposals} is the core of this thesis, where we use detection proposals to detect any kind of object in a sonar image, without considering the object class. We propose a Neural Network that outputs an objectness score from a single scale image, which can be used to generate detections on any kind of object and generalizes well outside its training set.

In Chapter \ref{chapter:applications} we show two applications of detection proposals. The first is end-to-end object detection of marine debris in sonar images, and the second being object tracking using proposals and a matching CNN. We show that in both cases our approaches work better than template matching ones, typically used in sonar images.

Finally in Chapter \ref{chapter:conclusions} this thesis is closed with a set of conclusions and future work.

\section{Software Implementations}

We used many third party open source implementation of common algorithms. For general neural network modeling we used Keras \footnote{Available at \url{https://github.com/keras-team/keras}} (version 1.2.2), with the Theano \footnote[][1em]{Available at \url{https://github.com/Theano/Theano}} backend (version 0.8.2).

Many common machine learning algorithms (SVM, Random Forest, Gradient Boosting), evaluation metrics (Area under the ROC Curve), and dimensionality reduction techniques (t-SNE and Multi-Dimensional Scaling) come from the scikit-learn library\footnote{Available from \url{https://github.com/scikit-learn/scikit-learn}} (version 0.16.1).

We used the SIFT, SURF, AKAZE, and ORB implementations from the OpenCV library \footnote{Available at \url{https://github.com/opencv/opencv}} (version 3.2.0).

In terms of hardware, many neural networks are small enough to train on a CPU, but in order to speedup evaluation of multiple neural networks, we used a NVIDIA GTX 1060 with 6 GB of VRAM. GPU acceleration was implemented through CUDA (version 8.0.61, used by Theano) and the CuDNN library (version 6.0.21) for GPU acceleration of neural network primitives.

\section{Contributions}

This Doctoral Thesis has the following contributions:

\begin{itemize}
    \item We captured a dataset of $\sim 2000$ marine debris sonar images with bounding box annotations using an ARIS Explorer 300 Forward-Looking Sonar. The images contain a selected a subset of household and typical marine objects as representative targets of marine debris. We use this dataset to evaluate different techniques in the contest of the Marine Debris detection task. We plan to release this dataset to the community through a journal paper.
    
	\item We show that in our FLS Marine Debris dataset, a deep neural network can outperform cross-correlation\cite{hurtos2013automatic} and sum of squared differences template matching algorithms in Forward-Looking sonar image classification of Marine Debris objects.
	\item We propose the use of sum of squared differences similarity for sonar image template matching, which outperforms the cross-correlation similarity in our dataset of FLS Marine Debris images.
	\item We shown that a deep neural network trained for FLS classification can generalize better than other techniques with less number of data points and using less trainable parameters, as evaluated on our FLS Marine Debris images.
    
    \item We show that specially designed deep neural network based on the Fire module\cite{iandola2016squeezenet} allows for a model with low number of parameters and similar classification accuracy than a bigger model, losing only $0.5 \%$ accuracy on the FLS Marine Debris dataset.
    
    \item We demonstrate that our small FLS classification model can efficiently run on a Raspberry Pi 2 at 25 frames per second due to the reduction in parameters and required computation, indicating that it is a good candidate to be used in real-time applications under resource constrained platforms.
    
	\item We evaluate our neural network models with respect to the training set size, measured as the number of samples per class, on the FLS Marine Debris dataset. We find out that these models do not need a large dataset to generalize well.
	\item We evaluate the effect of varying the training and transfer set sizes on feature learning classification performance, on the FLS Marine Debris dataset. We find out that convolutional feature learning on FLS images works well even with small dataset sizes, and this holds even when both datasets do not have object classes in common. These results suggest that learning FLS classifiers from small datasets with high accuracy is possible if features are first learned on a different FLS dataset.
    
	\item We propose the use of a two image input neural network to match FLS image patches. On the FLS Marine Debris dataset, this technique is superior to state of the art keypoint detection methods\cite{rublee2011orb} and shallow machine learning models, as measured by the Area under the ROC Curve. This result also holds when training and testing sets do not share objects, indicating good generalization.
    
	\item We propose a simple algorithm to automatically produce objectness labels from bounding box data, and we use a neural network to predict objectness from FLS image patches. We use this model to produce detection proposals on a FLS image that generalize well both in unlabeled objects in our FLS Marine Debris dataset and in out of sample test data.
    \item We show that our detection proposal techniques using objectness from FLS images can obtain a higher recall than state of the art algorithms (EdgeBoxes\cite[-8em]{zitnick2014edge} and Selective Search\cite[-2em]{uijlings2013selective}) on the FLS Marine Debris dataset, while requiring less detections per image, which indicates that is more useful in practice.
	\item We show that our detection proposals can be combined with a neural network classifier to build an end-to-end object detector with good performance on the FLS Marine Debris dataset. Our detection proposals can also be combined with our patch matching network to built a simple tracking algorithm that works well in the same dataset.
    
	\item Finally we show that end-to-end learning works well to learn tasks like object detection and patch matching, on the FLS Marine Debris dataset, improving the state of the art and potentially being useful for other sonar sensor devices (like sidescan or synthetic aperture sonar), as we do not make modeling assumptions on sensor characteristics.
\end{itemize}

\section{Related Publications}

This thesis is based in the following papers published by the author:

\begin{itemize}
	\item \textit{Object Recognition in Forward-Looking Sonar Images with Convolutional Neural Networks}
	Presented at Oceans'16 Monterey. These results are extended in Chapter \ref{chapter:sonar-classification}.
	
	\item \textit{End-to-end Object Detection and Recognition in Forward-Looking Sonar Images with Convolutional Neural Networks} Presented at the IEEE Workshop on Autonomous Underwater Vehicles 2016 in Tokyo. We show these results in Chapter \ref{chapter:applications}.
	
	\item \textit{Objectness Scoring and Detection Proposals in Forward-Looking Sonar Images with Convolutional Neural Networks}
	Presented at the IAPR Workshop on Neural Networks for Pattern Recogntion 2016 in Ulm. This paper is the base for Chapter \ref{chapter:proposals} and we massively extend our results. This is also used in Chapter \ref{chapter:applications}.
	
	\item \textit{Submerged Marine Debris Detection with Autonomous Underwater Vehicles}
	Presented in the International Conference on Robotics and Automation for Humanitarian Applications 2016 in Kerala. This paper was our original inspiration for Chapter \ref{chapter:applications} as well as the motivation of detecting marine debris.
	
	\item \textit{Real-time convolutional networks for sonar image classification in 
	low-power embedded systems}
	Presented in the European Symposium on Artificial Neural Networks (ESANN) 2017 in Bruges. These results are extended in Chapter \ref{chapter:sonar-classification}.
	
	\item \textit{Best Practices in {C}onvolutional {N}etworks for {F}orward-{L}ooking {S}onar {I}mage {R}ecognition}
	Presented in Oceans'17 Aberdeen. These results are the base for Chapter \ref{chapter:limits} where we also present extended versions.
	
	\item \textit{{I}mproving {S}onar {I}mage {P}atch {M}atching via {D}eep {L}earning}
	Presented at the European Conference on Mobile Robotics 2016 in Paris. This is the base for Chapter \ref{chapter:matching}, where we extends those results.
\end{itemize}

%% file: chapters/marine-debris.tex
\chapter[Marine Debris As Motivation For Object Detection]{Marine Debris As \newline Motivation For Object Detection}
\label{chapter:motivation}

\begin{marginfigure}
    \centering
    \includegraphics[width = 0.95\textwidth]{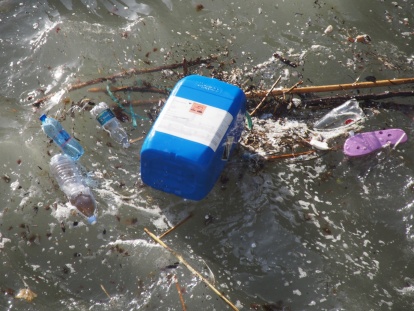}
    \caption[Surface Marine Debris at Manarola, Italy.]{Surface Marine Debris captured by the author at Manarola, Italy.}
    \label{md:manarola-surface}
\end{marginfigure}

\begin{marginfigure}
    \centering
    \includegraphics[width = 0.95\textwidth]{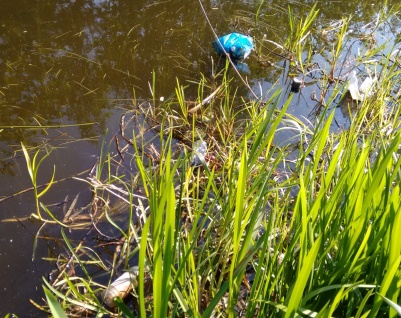}
    \caption[Surface Marine Debris at the Union Canal in Edinburgh, Scotland]{Surface Marine Debris captured by the author at the Union Canal in Edinburgh, Scotland.}
    \label{md:edinburgh-surface}
\end{marginfigure}

This chapter describes the "full picture" that motivates this thesis. While the introduction provides a summarized version of that motivation, this chapter will take a deeper dive into the problem of polluting our natural environment, with a specific look into pollution of water bodies.

After reading this chapter, the reader will have a general idea of how our daily lives are affected by marine debris, and how the use of Autonomous Underwater Vehicles can help us to reduce this problem.

This chapter is structured as follows. First we define what is marine debris, how it is composed and where it can be found in the seafloor. Then we describe the effect of marine debris in the environment as pollutant and its ecological consequences. We then make the scientific argument that submerged marine debris can be recovered by AUVs. Finally we close the chapter by describing a small datasets of marine debris in sonar images, which we use in the technical chapters of this thesis.

\begin{figure*}[tb]
    \centering
    \setfloatalignment{t}
    \forceversofloat
    \subfloat[Refrigerator]{
        \includegraphics[height = 0.14\textheight]{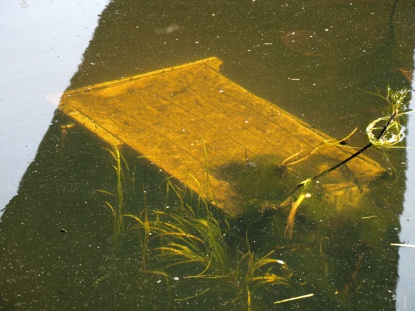}
    }
    \subfloat[Traffic Cone and Pram]{
        \includegraphics[height = 0.14\textheight]{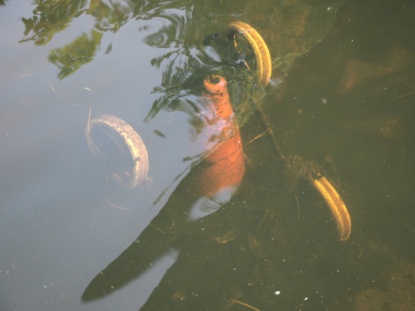}    
    }
    \subfloat[Shopping Cart]{
        \includegraphics[height = 0.14\textheight]{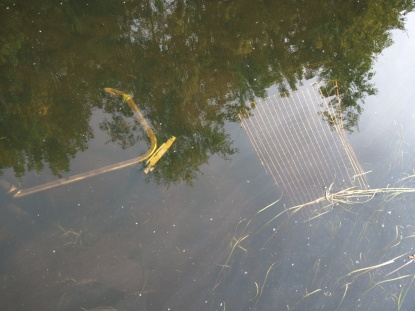}
    }
    \caption[Photos of Large Submerged Marine Debris at the Union Canal in Edinburgh, Scotland]{Photos of Large Submerged Marine Debris captured by the author at the Union Canal in Edinburgh, Scotland.}
    \label{md:edinburgh-large}
\end{figure*}

\begin{figure*}[tb]
    \centering
    \forcerectofloat
    \subfloat[Chips Bag]{
        \includegraphics[height = 0.13\textheight]{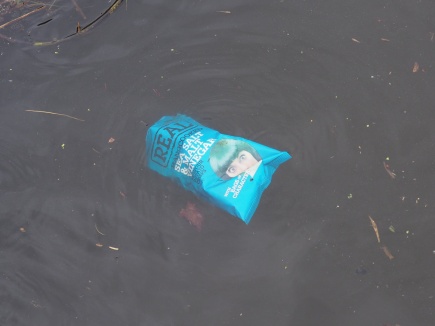}
    }
    \subfloat[Glass Bottle]{
        \includegraphics[height = 0.13\textheight]{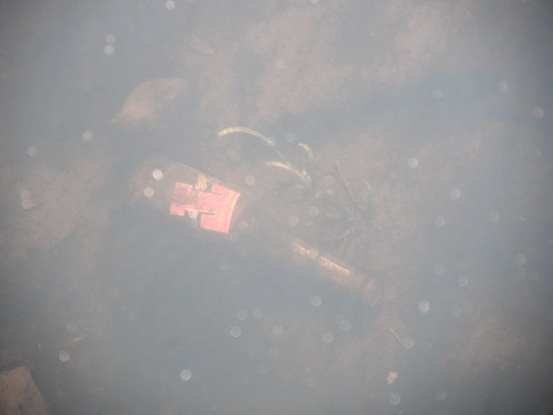}    
    }
    \subfloat[Can]{
        \includegraphics[height = 0.13\textheight]{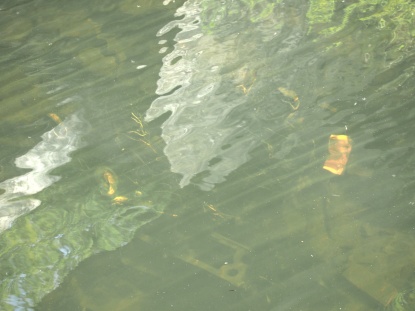}
    }
    
    \subfloat[Cans]{
        \includegraphics[height = 0.13\textheight]{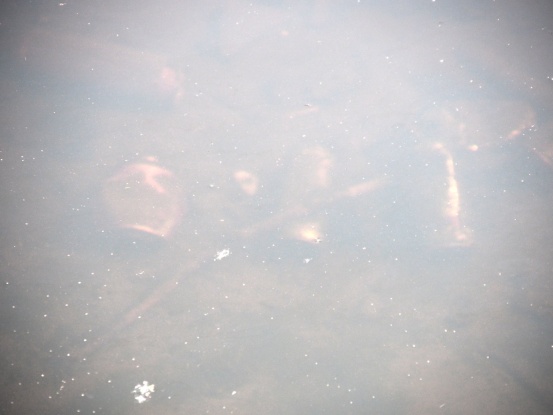}
    }
    \subfloat[Traffic Cone]{
        \includegraphics[height = 0.13\textheight]{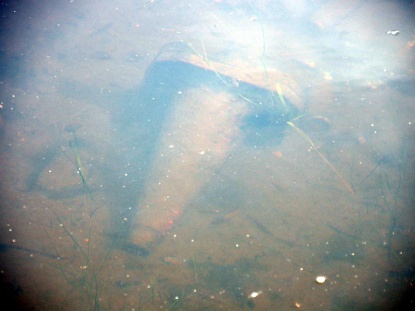}
    }
    \caption[Photos of Small Submerged Marine Debris at the Union Canal in Edinburgh, Scotland]{Photos of Small Submerged Marine Debris captured by the author at the Union Canal in Edinburgh, Scotland.}
    \label{md:edinburgh}
\end{figure*}

The author got his initial motivation about Marine Debris by experimental observation of the environment. During a couple of excursions to Loch Earn (Scotland) we observed submerged marine debris (beer and soft drink cans, tires) in the bottom of the Loch, and during daily commute to Heriot-Watt University through the Union Canal in Edinburgh, we also observed both submerged and floating marine debris. Figures \ref{md:edinburgh} shows small submerged objects in the Union Canal, while Figure \ref{md:edinburgh-large} shows large objects that were discarded in the same place.

\FloatBarrier
\section{What is Marine Debris?}

\marginnote {Marine Debris}Marine Debris encompasses a very large category of human-made objects that have been discarded, and are either floating in the ocean, partially submerged in the water column, or fully submerged and lying in the floor of a water body . 

Marine Debris is mostly composed of processed materials \cite[-3em]{weis2015marine} that are useful to human populations, such as plastics, metals, wood, glass, styrofoam, rubber, and synthetic materials derived from the previously mentioned such as nylon. 

The original use of many items discarded as marine debris is: food and drink packaging, fishing gear, plastic bags, bottles, maritime use, and others. Many of these categories are intended to be single-use items, and it can be expected that many objects used by maritime tasks end up in the ocean.

There are several ways Marine Debris ends up in the environment:

\begin{description}
    \item[Fishing] Activities that capture marine animals for human consumption are a major source of marine debris \cite{dayton1995environmental}. Fishing vessels regularly use nets that might get stuck on the bottom of the water body, and most of the time fishermen just decide to cut the nets, leaving them in the environment and becoming marine debris. Fishing line used by more amateur fishermen can also be considered marine debris, and any cut line usually floats in water. Most fishing nets are submerged and not usually visible to the naked eye. Fishing nets and lines are mostly made out of different kinds of plastics.
    
    \item[Accidental Release] Human-made debris can make it out to rivers or oceans by accident. One example is catastrophical events such as Tsunamis \cite[-10em]{mori2011survey}, Hurricanes, Storms, etc. But these events are rare and not controlled by humans. A more common kind of accidental release is garbage left at beaches \cite[-6em]{smith1997marine}, as high tide can carry left debris and make it into the ocean. Similar mechanisms work in rivers and estuaries, but it is more common that garbage is directly dropped into a river. Scientific research can also sometimes accidentally leave marine debris in the environment.
    
    \item[Intentional Release] There are of course people and entities that will directly dump garbage into the ocean (usually from ships) or in rivers and lakes, without any consideration about damage to the environment. Legality of these releases varies with country, but generally it is very hard to enforce any kind of regulation, specially in developing countries mostly due to economical needs.
    
    \item[Improper Waste Management] Corresponds to waste that escapes containment\cite{willis2017differentiating} due to improper management, such as overflow from filled garbage cans, or failures in landfill containment. These can reach water bodies by a variety of means, such as the wind carrying lightweight plastic bags, or cans from a garbage can in a sidewalk falling and reaching storm drains. This is the most controllable factor for governments and NGOs in order to prevent debris from reaching water bodies, as more infrastructure can always be built.
\end{description}

The dropping of emissions and garbage in water bodies is regulated by the \textit{International Convention for the Prevention of Pollution from Ships} (MARPOL 73/78) \cite[-3em]{canyon1978international}, which was signed in 1973 but not ratified until 1978. It prohibits all release of non-food solid waste, while regulating liquid waste.

\begin{figure}[tb]
    \centering
    \subfloat[Plastic Bottle]{
        \includegraphics[height = 0.11\textheight]{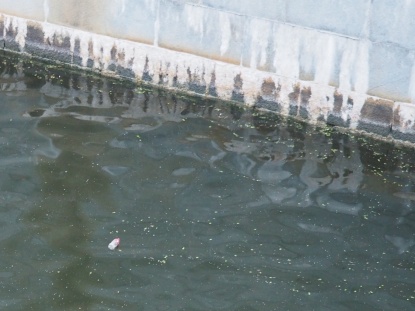}
    }
    \subfloat[Plastic Objects and Drink Carton][Plastic Objects and \newline Drink Carton]{
        \includegraphics[height = 0.11\textheight]{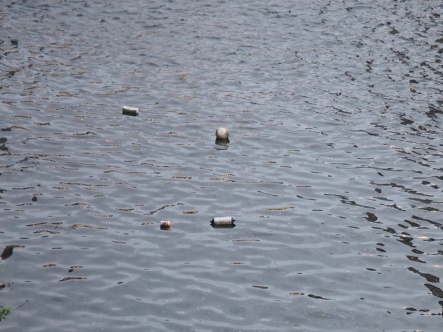}    
    }
    \subfloat[Submerged Plastic Bags][Submerged \newline Plastic Bags]{
        \includegraphics[height = 0.11\textheight]{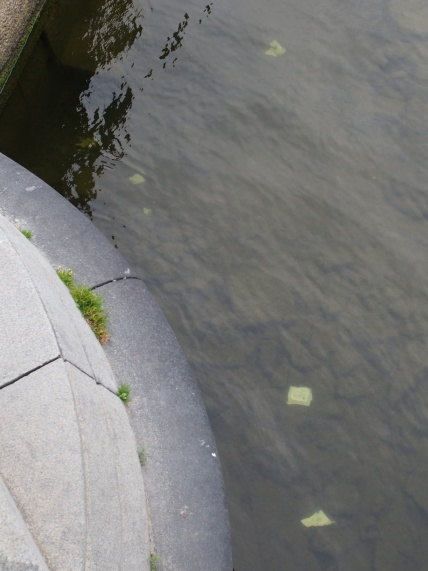}
    }
    \vspace*{0.5cm}
    \caption[Photos of Surface Marine Debris in Stockholm, Sweden]{Photos of Surface Marine Debris captured by the author in Stockholm, Sweden.}
    \label{md:stockholm}
\end{figure}

\newpage
Spengler et al. \cite{spengler2008methods} surveys the methods for marine debris in the seafloor, indicating that the most common method used to study submerged marine debris is a bottom trawl net, followed by diving-related techniques. Sonar is also used to survey seafloor areas. This research paper also describes 26 other studies that survey seafloor marine debris, noting the large variation of sampled area (from 0.008 km$^2$ to 4016 km$^2$), and the lack of a unified methodology to categorize and measure marine debris.

As mentioned before, Marine debris can be found along different parts of the water column. Ryan et al. \cite[-3em]{ryan2014litter} surveys the South Atlantic garbage patch, and finds out that Polystyrene and foamed plastics mostly are protruding the surface, while ropes and nets float in the surface, and most of bags and food packaging sinks to the sea bottom. These results are consistent with our experience but miss that bottle material determines if it sinks or floats. We have found out that most light plastics float, while heavy plastics and glass immediately sinks to the bottom. Whether the bottle is open or closed (with a cap) also influences, as a closed bottle will not sink because it is filled by air, and it will freely float instead.

Debris materials are not equally distributed. The most common material is plastics by far, which is consistent with production and use of single-use plastic items. Li et al. \cite[-4em]{li2016plastic} surveys the literature about plastic marine debris, finding that field surveys have found up to $98 \%$ of plastic in their marine debris samples, but this ratio varies wildly between sites. The highest ratio of plastic debris is found in the North Pacific Ocean, while the lowest ratio is found in the Eastern China Sea and South Korea Sea.

A large scale survey of debris in the Monterey bay was presented by Schlining et al.\cite[-9em]{schlining2013debris}, where the authors analyzed thousands video hours captured by ROVs\sidenote[][2em]{Obtained for a period of 22 years by MBARI, between 1989 and 2011}, finding 1537 pieces of submerged marine debris, which were classified as $33 \%$ plastics, $23 \%$ metal, $14 \%$ rope, $7 \%$ unknown debris, $6 \%$ glass, $5 \%$ fishing debris, $4 \%$ paper, and others with less than $3 \%$ like rubber, fabrics, clothing, wood, etc. Majority of the items were found along the Monterey Canyon system, which makes sense as debris could be dragged by underwater currents and accumulated there.

Another study in the Western Gulfs of Greece by Stefatos et al.\cite{stefatos1999marine}, where the debris recovered by nets from fishing boats was counted. In the Gulf of Patras 240 items per km$^2$ were found, while in the Gulf of Echinadhes 89 items per km$^2$ were recovered. Debris distributions in both sites are similar, with over $80 \%$ of recovered debris being plastics, $10$ \% were metals, and less than $5 \%$ were glass, wood, nylon, and synthetics. The authors attribute a high percentage of drink packaging in Echinadhes to shipping traffic, while general packaging was mostly found in Patras, suggesting that the primary source for this site is carried by land through rivers.

Chiba et al. \cite{chiba2018human} built and examined a database of almost 30 years (1989 - 2018) of video and photographic evidence of marine debris at deep-sea locations around the world. They found 3425 man-made debris pieces in their footage, where more than one third was plastics, and $89 \%$ were single use products. This dataset covers deep-sea parts of the ocean, showing that marine debris has reached depths of 6-10 thousand meters, at distances of up to 1000 km from the closest coast.  At the North-Western Pacific Ocean, up to 17-335 debris pieces per km$^2$ were found at depths of 1000-6000 meters. The deepest piece of debris was found in the Mariana Trench at 10898 meters deep. Their survey shows that plastic debris accumulates in the deepest parts of the ocean from land-based sources, and they suggest that a way to monitor debris is needed.

Jambeck et al.\cite{jambeck2015plastic} studies the transfer of plastic debris from land to sea. The authors built a model that estimates up to an order of magnitude the contribution of each coastal country to global plastic marine debris. The model considers plastic resin production, population growth, mass of plastic waste per capita, and the ratio of plastic waste that it is mismanaged and has the potential to become marine debris. According to this mode, the top five contributors of plastic marine debris per year are the countries that hold the biggest coastal populations, namely China ($[1.3 - 3.5]$ MMT)\footnote{Millions of Metric Tons (MMT)}, Indonesia ($[0.5 - 1.3]$ MMT), The Philippines ($[0.3 - 0.8]$ MMT), Vietnam ($[0.3 - 0.7]$ MMT), an Sri Lanka ($[0.2 - 0.6]$ MMT). If the coastal European Union countries would be considered as a single source, it would be ranked 18th with $[0.05 - 0.12]$ MMT.

Note that this model does not consider marine debris generated by cruise ships, which are rumored to be a major contributor to marine debris in the Mediterranean sea. These results are consistent with other measurements \cite{eriksen2014plastic}, and point that the biggest contributor to marine debris from land is just the level of development. Less developed countries usually lack good waste management facilities, while developed countries manage their waste properly. For example, Recycling in the European Union is overall of good quality and therefore much less waste goes into the ocean as marine debris in Europe. It is also important that waste is included in development policies and plans, in order for waste management to scale in relation to a country's development status.

Overall studies about the composition of marine debris are usually spatially localized, but overall there is a high consistency between studies about plastics being the most common marine debris element, with metal and glass being second. Plastics are explained by the ubiquity of single use items made out of this material, mostly food and drink packaging. Fishing materials and ropes are also commonly found in areas where fishermen work.

\section{Ecological Issues with Marine Debris}

In the previous section we have described what is marine debris, what materials is it composed of, and how it reaches bodies of water in the environment. In this section we will describe how marine debris pollutes the environment and what kind of ecological issues it produces.

Marine debris poses serious dangers to native marine life. One of the biggest issues is that marine animals can get entangled in discarded fishing nets, six pack rings, or derelict ropes \cite{laist1997impacts}. This could cause the animal to asphyxiate (for example, sea turtles that need air), or severely limiting its movement which could cause to starve (for fish). There are multiple reports of this happening in the scientific literature \cite[1em ]{gall2015impact} and public media.

Fishing gear is overall quite deadly for marine life, with effects on reducing biodiversity and thus affecting the ecosystem. For other kinds of debris, marine animals can accidentally ingest them, depending on its size and buoyancy.

Plastic debris is the most dangerous material, as it usually floats, and ultraviolet radiation from sunlight slowly breaks it down. After many iterations of this process, plastic transform into small (less than 5mm) particles that are called micro-plastics. Marine animals can easily ingest these particles without noticing\cite[-5em]{laist1987overview}, which leads to poisoning or blocking of airways or digestive tracts, killing the animal. Marine fauna that survives might be captured for human consumption, from where plastics might enter the human food chain.

There are also increasing effects due to the food chain. As plastic breaks up into large number of smaller pieces, there is a higher chance that small marine animals and filtering feeders will ingest it, from where bigger animals can consume them, propagating up into the food chain, eventually reaching humans.

Not only marine animals that live in the water itself can be affected by marine debris. There are many reports\cite{wilcox2015threat} of coastal and sea birds eating plastic that floats in the water, or capturing fish that has previously consumed micro-plastics. This has a more profound effect in this kind of animal, as they have a higher chance of accidentally ingesting a medium sized piece of plastic, and for debris to block their airways or digestive systems. There are studies that show a decreasing trend in this kind of birds \cite{paleczny2015population}, which could be in part caused by the increasing amounts of plastic debris in our coasts and oceans.

Plastics also collects other kinds of chemical pollutants present in the water (such as pesticides, flame retardants, industrial waste, oil, etc), making the polluted plastic particles highly toxic\cite{engler2012complex}. This increases the potential of marine debris to be harmful, specially for human populations, as they might be consuming seafood and fish that could be contaminated by chemicals transferred by debris. Preventing this kind of contamination of food produce will likely increase the costs associated to food.

Other kinds of debris such as rubber and wood can also be problematic as marine animals can ingest them, but in significantly lower quantities than plastics.

A more direct effect of debris in human population can happen as it can washes up in beaches, hurting children and unbeknown people. Accumulation of debris in beaches can also reduce the economic value of tourism in coastal areas \cite{engler2012complex}, indirectly hurting human populations. Land animals that live close to coasts and shores can also be affected by floating or accumulated debris, in similar ways as marine animals.

Discarding plastic that could be recycled makes no economic sense, as the source material for plastic is oil, which is not infinite and we will run out of it at some point. The same could be said of metals and rubbers.

Overall, we should not pollute the environment because it can have unexpected consequences and it is not economically viable. Sustainability is an important topic to reach a development level where our lifestyle does not require the sacrifice of the planet's ecosystem. 

\section{Proposal - Marine Debris Recovery with AUVs}

This section we makes a proposal that AUVs can be used to recover marine debris, which could have more scientific and civilian applications than other current uses.

We believe that the Underwater Robotics community is too dominated by military and non-civilian applications. This is clearly seen by the large literature in \textit{mine countermeasures} and related topics into detecting military mines in underwater environments. We believe this is more pronounced inside the object detection and automatic target identification sub-field, as most of the literature covers methods specifically designed to perform recognition of marine mines.

While detecting and removing marine mines could have a clear humanitarian value, this research is mostly performed by military research centers and defense companies. This is generally fine but it is problematic from a scientific point of view, as data is usually classified and not available for the scientific community, which reduces public interest and the introduction of new ideas into the field. We see this reflected in the fact that there is no public dataset that contains this kind of objects on sonar images.

Overall there are other current non-military research directions in Marine Robotics that are also worth pursuing, such as Underwater Archeology, Mapping, Pipe Search and Following, etc. The Oil and Gas Industries have also interesting applications that require further research. In general humanitarian applications of Marine Robotics have not been explored.

There are some efforts to capture or remove marine debris, but overall these are weak and not at large scale. Volunteers, non-governmental organizations, and local governments do efforts\cite[-10em]{nelms2017marine} to remove debris from beaches, coastal areas, and rivers, but these are localized and usually small scale.

Boats and nets can be used to capture surface debris at sea \cite[-5em]{kader2015design}, while debris traps can be installed in rivers and affluents to capture debris\cite[-1em]{beduhn1966removal} before it reaches the sea and is lost. In general these are quite low technology solutions, that work at smaller scales and localized areas, but cannot be used to perform large scale removal of debris in the seafloor.

Rochman et al. \cite[-1em]{rochman2016strategies} suggests that marine debris removal strategies should be diverse, given the large diversity of debris materials and types, and the ubiquity of its location.

The Rozalia Project \footnote{\url{https://rozaliaproject.org/}} seems to be an effort to develop robots for debris removal, but we could not find project status or additional information. Recently a startup company called \textit{The Ocean Cleanup} \footnote{\url{https://www.theoceancleanup.com/}} is developing technologies to remove surface marine debris. In both cases it seems that the efforts concentrate in surface debris, while we propose technologies for submerged debris.

For these reasons, this thesis proposes that Autonomous Underwater Vehicles can be used to survey, map, and recover submerged marine debris from the seafloor. The basic idea is that we can equip an AUV with both a perception sensor and a manipulator, with advanced object detection algorithms using sensor data to locate and identify marine debris, and the manipulator with an appropriate end effector to capture the piece of debris and collect it inside a special area or basket inside the AUV.

While this proposal sounds simple, it entails many difficult research questions, for example:

\begin{enumerate}
    \item What is the most appropriate perception sensor to sense submerged debris in any kind of marine environment?
    \item How to perceive and detect submerged debris in a perception sensor without making assumptions in object shape or environment? What are the most appropriate algorithms?
    \item What is the most appropriate manipulator arm and end effector configuration to capture the largest variety of submerged debris?
    \item How can perception and manipulation be combined for successful capture of submerged debris, while making as little assumptions on object shape or environment?
\end{enumerate}

This thesis only develops questions related to perception of submerged marine debris, more precisely, Question 2. We believe that the most appropriate sensor to detect submerged debris is a high spatial resolution sonar sensor, and for this purpose we use an ARIS Explorer 3000. \cite[-4em]{arisExplorer3K}, which has a per-pixel spatial resolution of up to 3.0 millimeters. This sensor is ideal to "see" small objects (less than 10 cm) and provides a high frame-rate of up to 15 Hz, which is ideal for manipulation. In comparison other sonar sensors such as sidescan have multiple centimeters per pixel spatial resolution, for which objects like bottles would be only represented as a few pixels, not enough for successful recognition.

Synthetic Aperture Sonar (SAS) could also be an option, but only for tasks like surveying an area before using a high frame-rate sensor for precision debris capture, as it is not suitable to combine SAS with a manipulator due to the low frame-rate of the sensor.

It is well known that performing manipulation in underwater environments is challenging \cite{ridao2014intervention}, specially when combining this with a sonar sensor, due to reflections, noise, and unexpected interactions between the sonar sensor, the environment, and a moving manipulator arm.

One very important point that must be made now is that we are not proposing a "silver bullet" that can fully solve the problem of marine debris, or the general issues with waste management. The first and most obvious solution is not to pollute our environment. Recycling and properly disposal of waste is a key element of any sustainable policy. Our proposal only deals with the submerged marine debris that is currently lying on the seafloor, and does not prevent further debris being discarded into the ocean and other water bodies. Some debris also floats and does not sink, and these would require a whole different set of techniques to be collected.

\section{Datasets used in this Thesis}

This section describes the data used to train deep neural networks and to produce all the results presented in further chapters. We describe both the data capture setup and the data itself.

\begin{marginfigure}
    \centering
    \includegraphics[width = 0.95\textwidth]{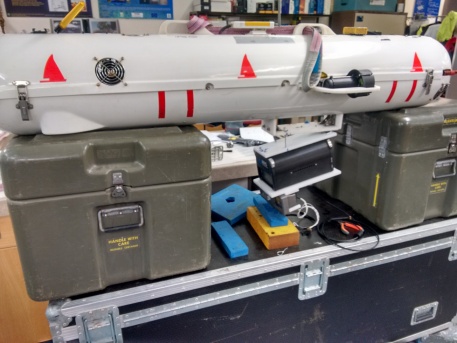}
    \caption{Nessie AUV with ARIS Sonar attached in the underside.}
    \label{md:nessie-aris}
\end{marginfigure}

There are no public datasets that contain marine debris in sonar images, and for the purposes of this thesis, a dataset of sonar images with bounding box annotations. We captured such dataset on the Water Tank of the Ocean Systems Lab, Heriot-Watt University, Edinburgh, Scotland. The Water Tank measures approximately $(W, H, D) = 3 \times 2 \times 4$ meters, and on it we submerged the Nessie AUV with a sonar sensor attached to the underside, as shown in Figure \ref{md:nessie-aris}.

\subsection{Sonar Sensing}

The sonar sensor used to capture data was the ARIS Explorer 3000\cite[-6em]{arisExplorer3K} built by SoundMetrics, which is a Forward-Looking Sonar, but can also be considered to as an \textit{acoustic camera} due to its high data frequency (up to 15 Hz). A big advantage of this sensor is its high spatial resolution, as one pixel can represent up to 2.3 millimeters of the environment.

This sonar has 128 acoustic beams over a $30^\circ \times 15^\circ$ field of view, with a $0.25^\circ$ spacing between beams. The minimum distance range of the sonar is around 70 centimeters, and the maximum distance depends on the sampling frequency. at 1.8 MHz the range is up to 5 meters, while at 3.0 MHz the range is only 5 meters. Depending on distance this Sonar has 2.3 millimeters per pixel spatial resolution in the close range, and up to 10 centimeters per pixel at the far range.

We only have limited information on how the ARIS Explorer sonar works, as we believe most of the information is a trade secret of SoundMetrics and/or classified by the US Navy. A related sonar sensor is the DIDSON (Dual-Frequency Identification Sonar) \cite[-2em]{belcher2002dual}, which is an earlier iteration of what later would become the ARIS sonar. Development of the DIDSON sonar was funded by the US Navy through the Space and Naval Warfare Systems Center Bayside (San Diego, California).

The basic working of the ARIS sonar is that it uses an \textit{acoustic lens} \cite[-2em]{wu2010beam} to perform beamforming and focus beams to insonify an area of the seafloor with a higher resolution than normally possible with classic sonars.  This works in a similar way that of a optical camera, where a lens allows to \textit{zoom} into a part of the scene. This also means that the acoustic lens moves inside the sonar in order to focus different parts of the scene, and focusing distance can be set by the user.

There is some publicly available information about the use of acoustic lenses with the DIDSON sonar. Belcher et al in 1999 \cite[-6em]{belcher1999beamforming} showed three high-frequency sonar prototypes using acoustic lenses, and in 2001 \cite[-1em]{belcher2001object} they showcased how the DIDSON sonar can be used for object identification. A Master Thesis by Kevin Fink \cite[2em]{fink1994computer} produced computer simulations of sound waves through an acoustic lens beamformer. Kamgar-Parsi et al \cite[1em]{kamgar1998underwater} describes how to perform sensor fusion from multiple views of the scene obtained by a moving acoustic lens.

\subsection{Sonar Image Capture}

\begin{figure}
    \centering
    \subfloat[]{
        \includegraphics[width=0.45\textwidth]{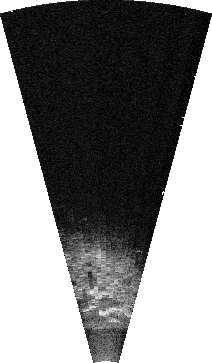}
    }
    \subfloat[]{
        \includegraphics[width=0.45\textwidth]{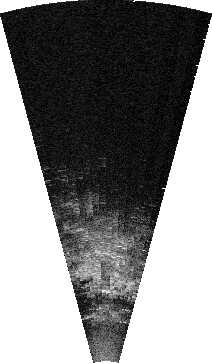}
    }

    \subfloat[]{
        \includegraphics[width=0.45\textwidth]{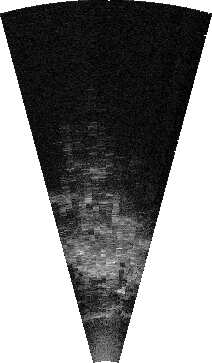}
    }
    \subfloat[]{
        \includegraphics[width=0.45\textwidth]{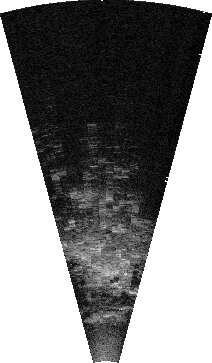}
    }
    \caption[Selected images showing beam interference producing wrong Sonar images]{Selected images captured in a lake using the ARIS attached to a Sonobot, showing the beam interference that produces wrong Sonar images. The beam pattern can be clearly seen in the image, altering the estimated range of some pixels along the beam, producing a distorted image}
    \label{md:sonar-beam-interference}
\end{figure}

We were motivated to use a water tank to capture our dataset due to the difficulties in using a real world underwater environment. We initially mounted the ARIS sonar on a surface vehicle (An Evologics Sonobot) and tested it at some locations around Edinburgh, Scotland\footnote{The Union Canal and Tally Ho Flooded Quarry}, since it would allows us to precisely move the sonar underwater and to have line of sight with the vehicle, but surface waves produced by the wind provoked rotation of the Sonobot on the roll axis, moving the sonar sideways, and producing heavy interference between the sound beams. This way the produced images that were not usable. Some examples of these images with beam interfence are shown in Figure \ref{md:sonar-beam-interference}.

\begin{marginfigure}[1em]
    \centering
    \includegraphics[width = 0.85\textwidth]{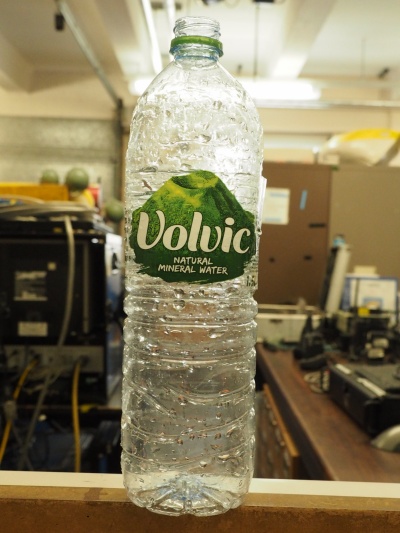}
    \includegraphics[width = 0.85\textwidth]{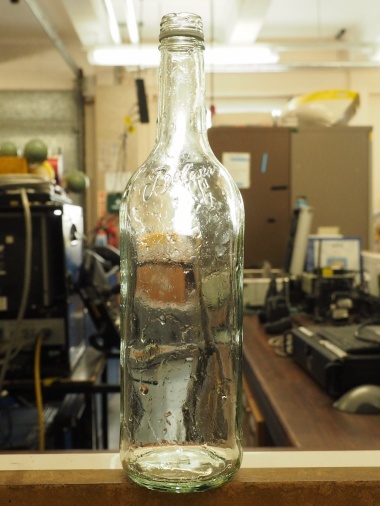}
    \caption{Samples of Bottle Class}
    \label{md:bottles}
\end{marginfigure}

\begin{marginfigure}
    \centering
    \includegraphics[width = 0.85\textwidth]{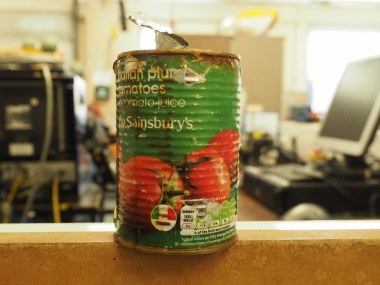}
    \caption{Sample of Can Class}
    \label{md:can}
\end{marginfigure}

\begin{marginfigure}
    \centering
    \includegraphics[width = 0.85\textwidth]{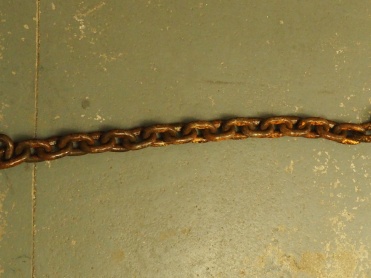}
    \caption{Sample of Chain Class}
    \label{md:chain}
\end{marginfigure}

\begin{marginfigure}
    \centering
    \includegraphics[width = 0.85\textwidth]{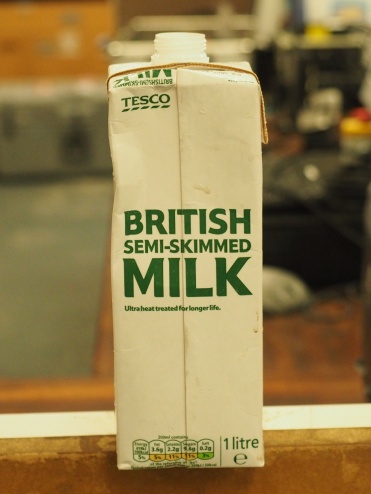}
    \caption{Sample of Drink Carton Class}
    \label{md:drink-carton}
\end{marginfigure}

An alternative would have been to use the Nessie AUV, but tele-operating it underwater is difficult since the water at many sites is cloudy, preventing us to see the vehicle and the objects through the water, and we would need to rely on sonar navigation only. Placing the objects in a real environment is also not easy for the same reasons, and recovering the marine debris that we placed would be problematic, depending on the depth of the bottom floor. We do not wish to pollute the environment this way.

Using a water tank is a controlled environment, where the objects are clearly visible (fundamental for labeling), there is no environmental interference, and placing the objects is considerably easier, with line of sight possible due to the clear water and shallow bottom.

The structure used to mount the sonar under the AUV allows for variable orientation of the Sonar sensor in the pitch axis, approximately in the range $[0^\circ, 75^\circ]$. We manually set the pitch angle to a fixed value between $15^\circ$ and $30^\circ$ in order for the acoustic beam to insonify the bottom of the water tank and we can see the objects in the output sonar image.

Objects were placed in the floor of the water tank, and were laid out in a pseudo-random way, and non-overlapping with each other, with the general idea that multiple objects can be in the sonar's field of view at the same time. Positioning of each object was performed by dropping them into the water, and manually adjusting position using a pole with a hook.

We tele-operated the AUV in order to capture different views of objects in the scene, but we had limited range due to the size of the water tank. This means that not all perspectives of the objects were observed, but we did get many perspectives of several objects. The vehicle was operated on an approximate quarter of circle path, at an approximate speed of 0.1 meters per second. Both the path and speed were constrained by the size of our water tank and the minimum range of the sonar sensor.

We used the ARIScope application provided by Sound Metrics to capture data at a frequency of 3.0 MHz, with focusing, gain setting, and initial image processing done by the application, while we manually selected the minimum and maximum ranges in order to contain our scene. We captured images at 15 Hz with the AUV moving, which produces a low framerate video.

\begin{figure}[t]
    \centering
    \includegraphics[width = 0.85\textwidth]{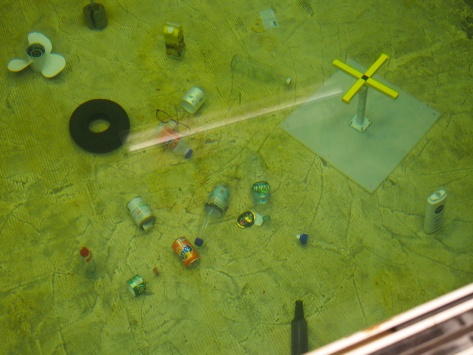}
    \caption{Example of one Scene Setup for Data Capture}
    \label{md:scene}
\end{figure}

Figure \ref{md:scene} shows one view of the objects placed in the scene. We moved the objects around before capturing each scene in order to introduce variability and make it possible to capture many views of an object.

Captured sonar data was extracted from \texttt{.aris} files and projected into a polar field of view and saved as \texttt{PNG} files.

To produce the dataset in this thesis, we selected a small group of household and typical marine objects that we placed in the water tank. This is motivated by the fact that most marine debris is composed of discarded household objects \cite{chiba2018human}.

Household objects used to build the dataset correspond to: bottles of different materials, tin cans, different drink cartons, and plastic shampoo bottles. For typical marine objects we used a chain, a hook, a propeller, a rubber tire, and a mock-up valve. While Marine Debris covers far more objects than we did, we believe this set of objects is appropriate for the scope of this thesis. 

We also included a set of marine objects as distractors (or counter examples), namely a chain, a hook, a propeller, a rubber tire, and a mock-up valve. These objects can be expected to be present in a marine environment \cite{dayton1995environmental} due to fishing and marine operations, and are not necessarily debris.

The object set was pragmatically limited by the objects we could easily get and were readily available at our lab. There is a slight imbalance between the marine debris and distractor objects, with approximately 10 object instances for marine debris, and 5 instances of distractors.

A summary of the object classes is shown in Table \ref{md:classes}. Bottles of different materials that lie horizontally in the tank floor are grouped in a single class, but a beer bottle that was standing on the bottom was split into its own class, as it looks completely different in a sonar image. A shampoo bottle was also found standing in the tank bottom and assigned its own class. The rest of the objects map directly to classes in our dataset. We also have one additional class called \textit{background} that represents anything that is not an object in our dataset, which is typically the tank bottom. Due to the high spatial resolution of the ARIS Explorer sonar, it is possible to see the concrete marks in the tank bottom clearly in a sonar image.

\begin{table*}[t]
    \begin{tabular}{llp{8cm}l}
        \hline 
        Class ID 	& Name 				& Description & Figure\\ 
        \hline 
        0			& Bottle			& Plastic and Glass bottles, lying horizontally & \ref{md:bottles}\\
        1			& Can				& Several metal cans originally containing food & \ref{md:can} \\
        2			& Chain				& A one meter chain with small chain links & \ref{md:chain}\\
        3			& Drink Carton		& Several milk/juice drink cartons lying horizontally & \ref{md:drink-carton} \\
        4			& Hook				& A small metal hook & \ref{md:hook} \\
        5			& Propeller			& A small ship propeller made out of metal & \ref{md:propeller} \\
        6			& Shampoo Bottle	& A standing plastic shampoo bottle & \ref{md:shampoo-bottle} \\
        7			& Standing Bottle	& A standing beer bottle made out of glass & \ref{md:standing-bottle}\\
        8			& Tire				& A small rubber tire lying horizontally & \ref{md:tire}\\
        9			& Valve				& A mock-up metal valve originally designed for the euRathlon 2015 competition & \ref{md:valve}\\
        10			& Background		& Anything that is not an object, usually the bottom of our water tank. & N/A\\
        \hline 
    \end{tabular} 
    \vspace*{0.5cm}
    \caption{Classes Available in the Marine Debris Dataset.}
    \label{md:classes}
\end{table*}

\begin{marginfigure}[1cm]
    \centering
    \includegraphics[width = 0.85\textwidth]{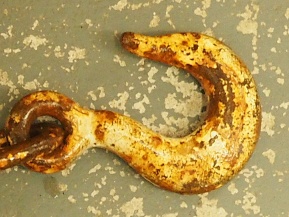}
    \caption{Sample of Hook Class}
    \label{md:hook}
\end{marginfigure}

From the captured raw sonar data, we selected 2069 images, and labeled the objects with bounding boxes and class information for each bounding box. The selection was made in a way to maximize the number of images that we kept, but with the following conditions:

\begin{itemize}
    \item Objects of interest (marine debris) were present in the image. Images without any objects were discarded.
    \item At least one marine debris object is clearly recognizable for labeling.
    \item At least five frames away in time from another previously selected image. This is done to reduce temporal correlation between selected images.
\end{itemize}

\begin{marginfigure}[-4cm]
    \centering
    \includegraphics[width = 0.85\textwidth]{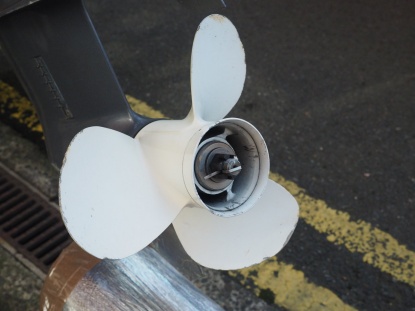}
    \caption{Sample of Propeller Class}
    \label{md:propeller}
\end{marginfigure}

We were not able to label all the objects in each image, as some objects looked blurry and we could not determine its class, but we labeled all the objects where its class can be easy and clearly recognized by the human annotator. In total 2364 objects are annotated in our dataset.

\begin{marginfigure}
    \centering
    \includegraphics[width = 0.85\textwidth]{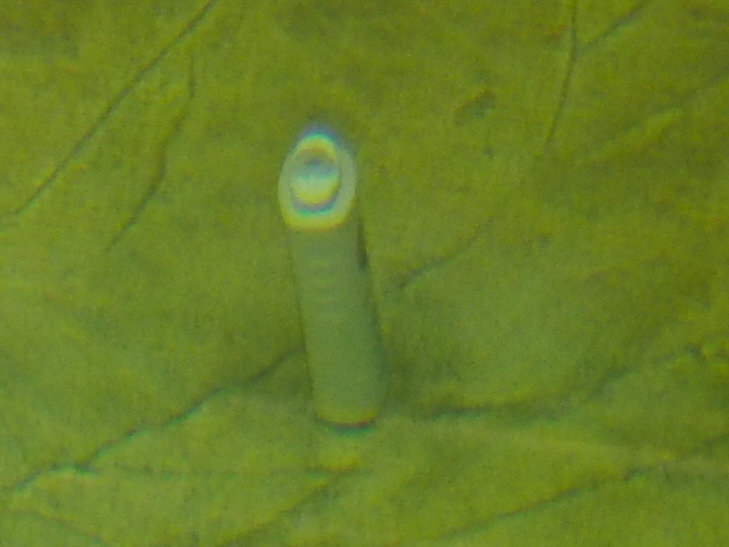}
    \caption{Sample of Shampoo Bottle Class}
    \label{md:shampoo-bottle}
\end{marginfigure}

Most of the objects we used for this dataset did not produce a shadow, depending on the perspective and size of the object. For the purpose of bounding box labeling, we decided to always include the highlight of the object, as it is the most salient feature, but we decided not to include the shadow of most objects in the bounding box, as there is a large variability of the shadow in small objects (sometimes present and sometimes completely absent).

\begin{marginfigure}
    \centering
    \includegraphics[width = 0.85\textwidth]{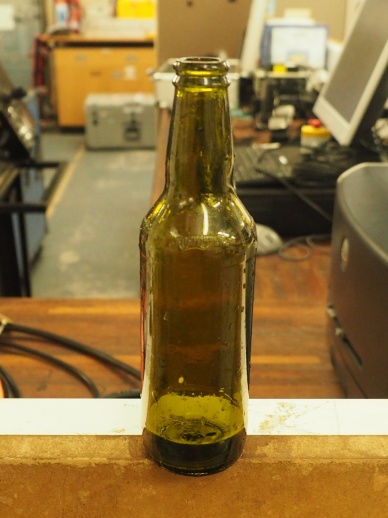}
    \caption{Sample of Standing Bottle Class}
    \label{md:standing-bottle}
\end{marginfigure}

For two object classes we always included the shadow, namely the standing bottle and the shampoo bottle, as sometimes the highlight of the object is easily confused with the background, but the shadow should allow for discrimination of this kind of object. The drink carton object is one that many times it had a shadow that we did not label. More detailed image crops of these objects are presented in Appenxix Figures \ref{appendix:standingBottle}, and \ref{appendix:shampooBottle}, and \ref{appendix:drinkCarton}.

We note that this kind of labeling is not the best as there is bias by not always labeling shadows. We made the practical decision of not always including shadows as it made labeling easier, and in the future new labels can be made for this dataset, specially if crowd-sourced labels from several sonar experts can be obtained, like how ImageNet\cite[-5em]{russakovsky2015imagenet} was labeled.

\begin{figure}[t]
    \centering
    \forcerectofloat
    \subfloat[Bottles]{
        \includegraphics[height = 0.2\textheight]{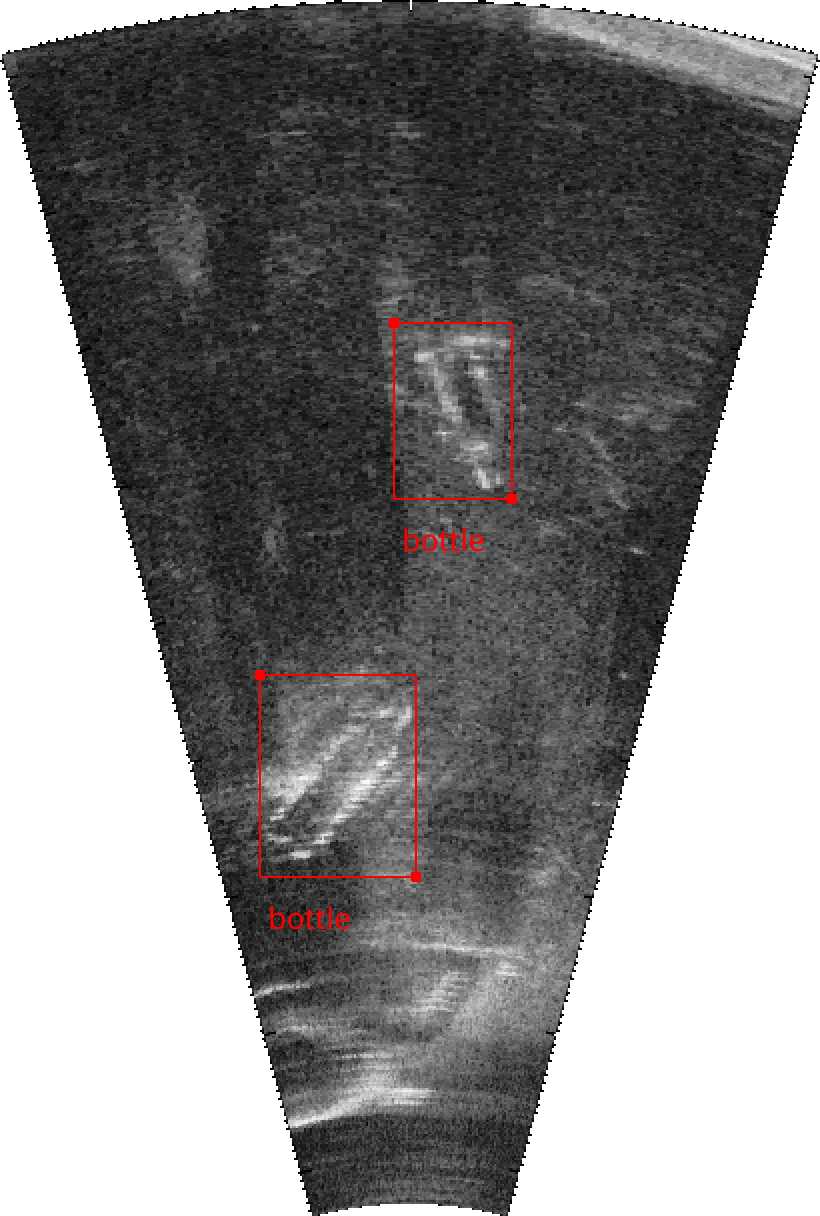}
    }
    \subfloat[Can]{
        \includegraphics[height = 0.2\textheight]{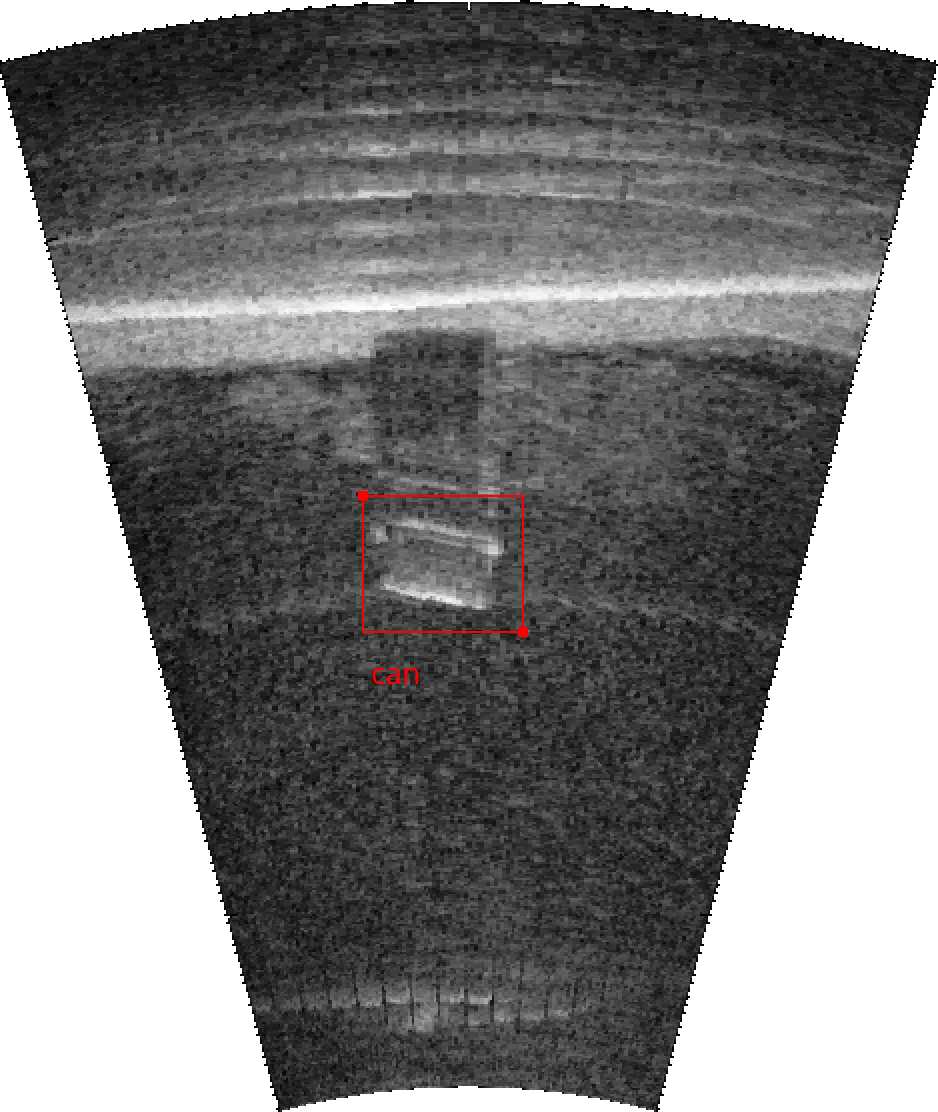}    
    }
    \subfloat[Chain]{
        \includegraphics[height = 0.2\textheight]{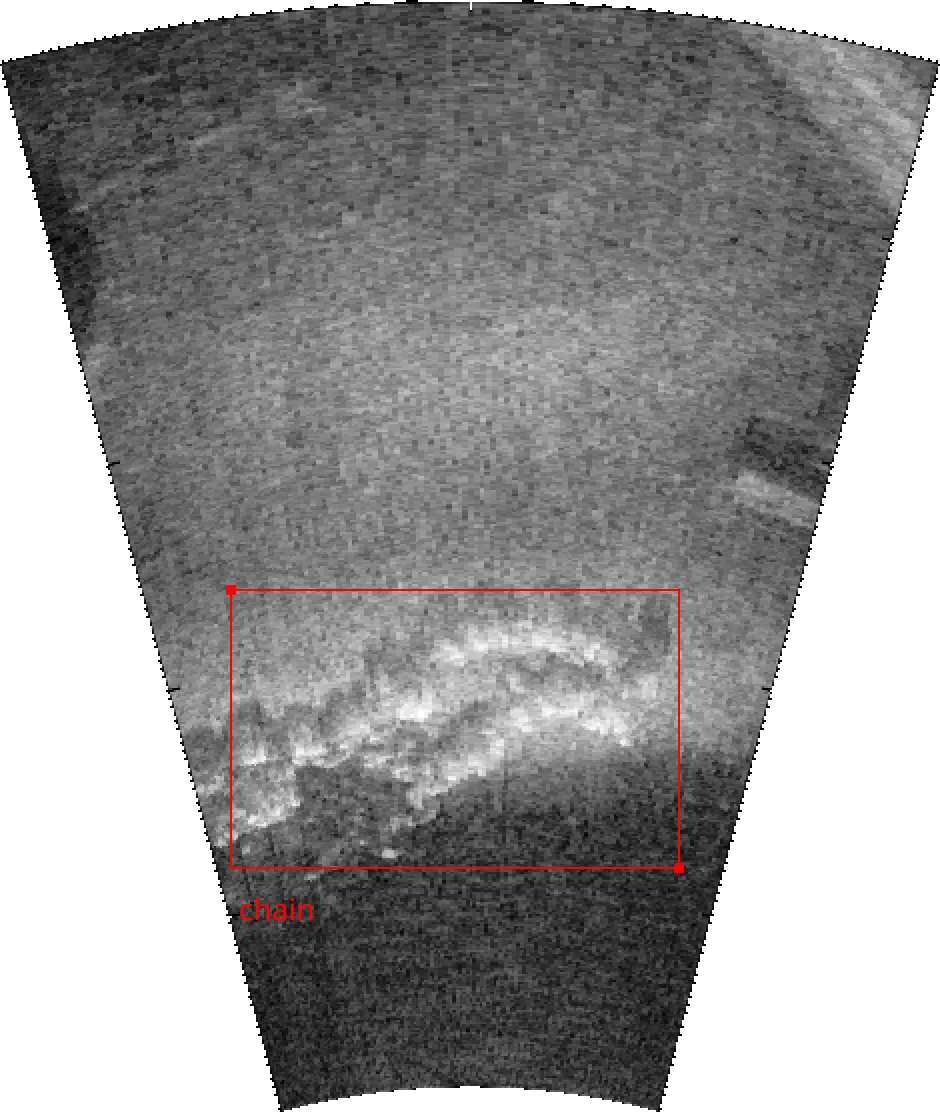}
    }

    \subfloat[Drink Carton]{
        \includegraphics[height = 0.2\textheight]{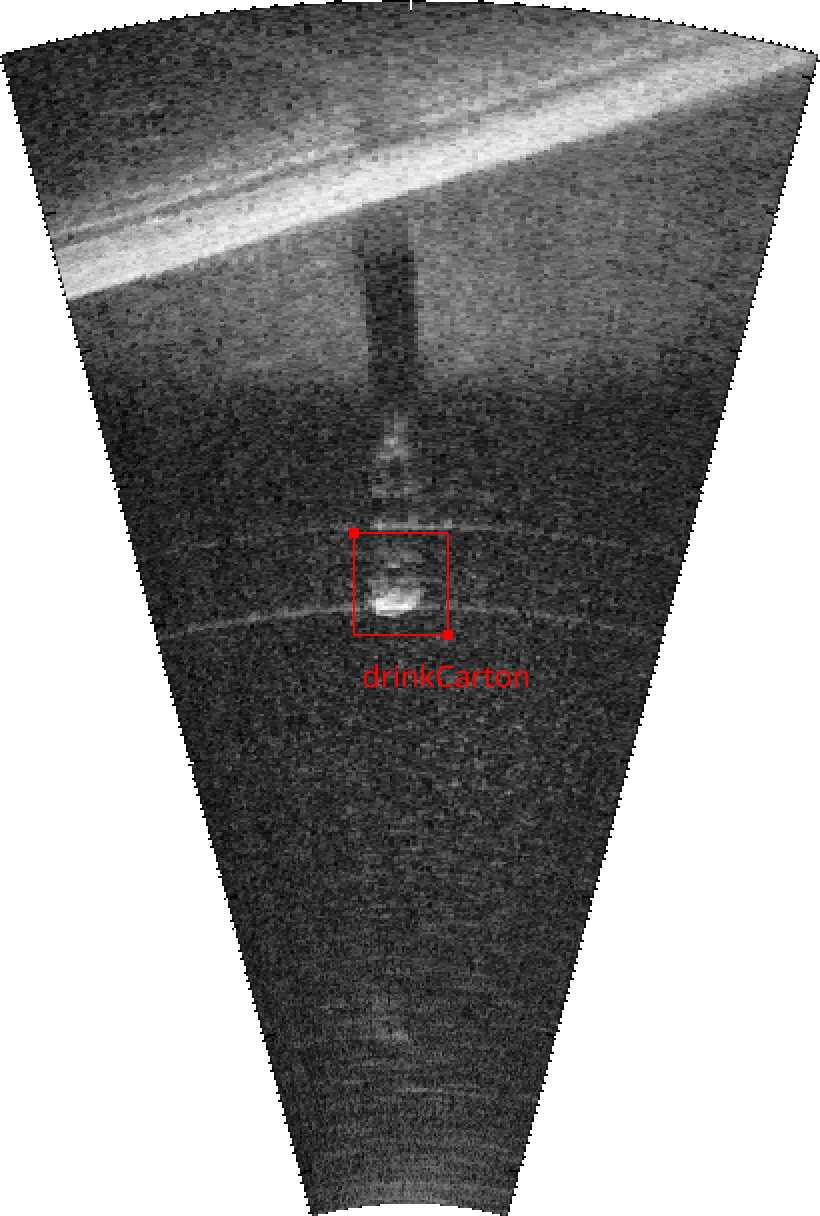}
    }
    \subfloat[Hook and Propeller]{
        \includegraphics[height = 0.2\textheight]{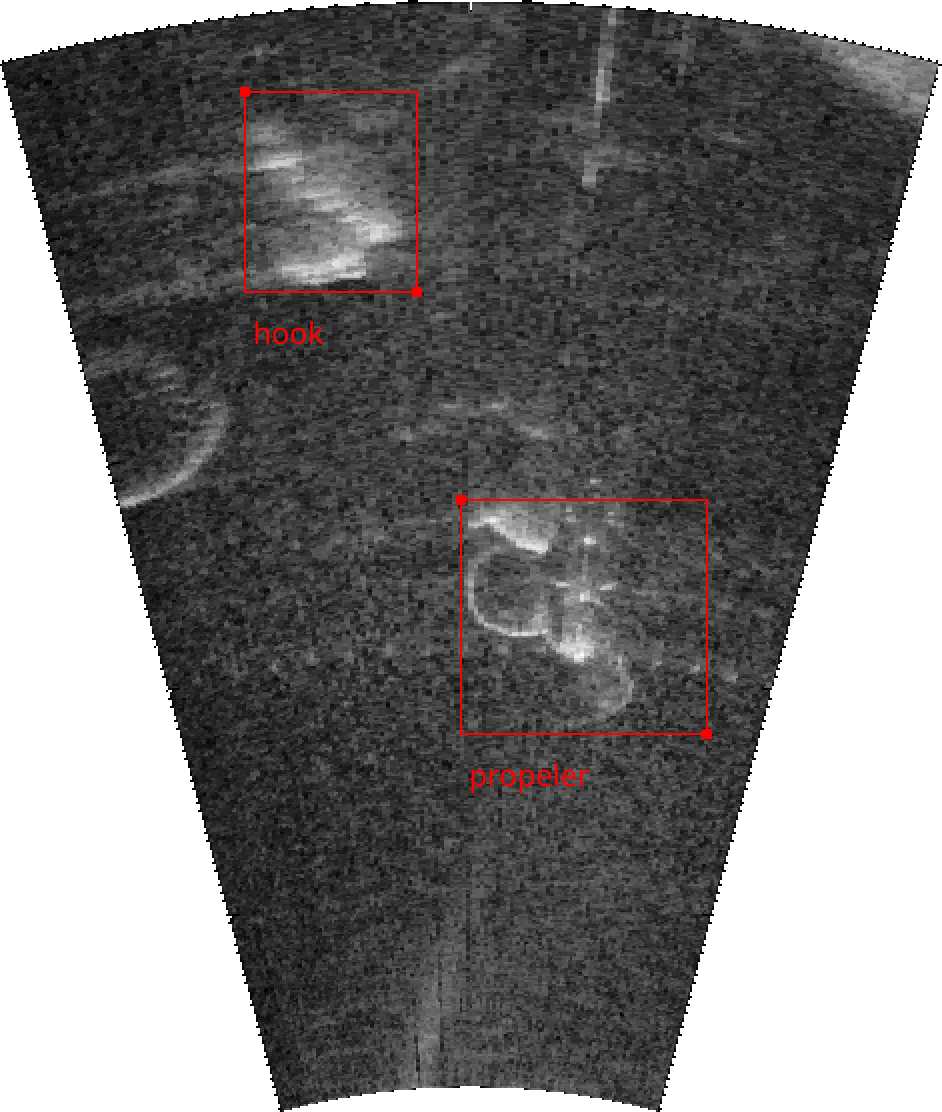}    
    }
    \subfloat[Shampoo Bottle]{
        \includegraphics[height = 0.2\textheight]{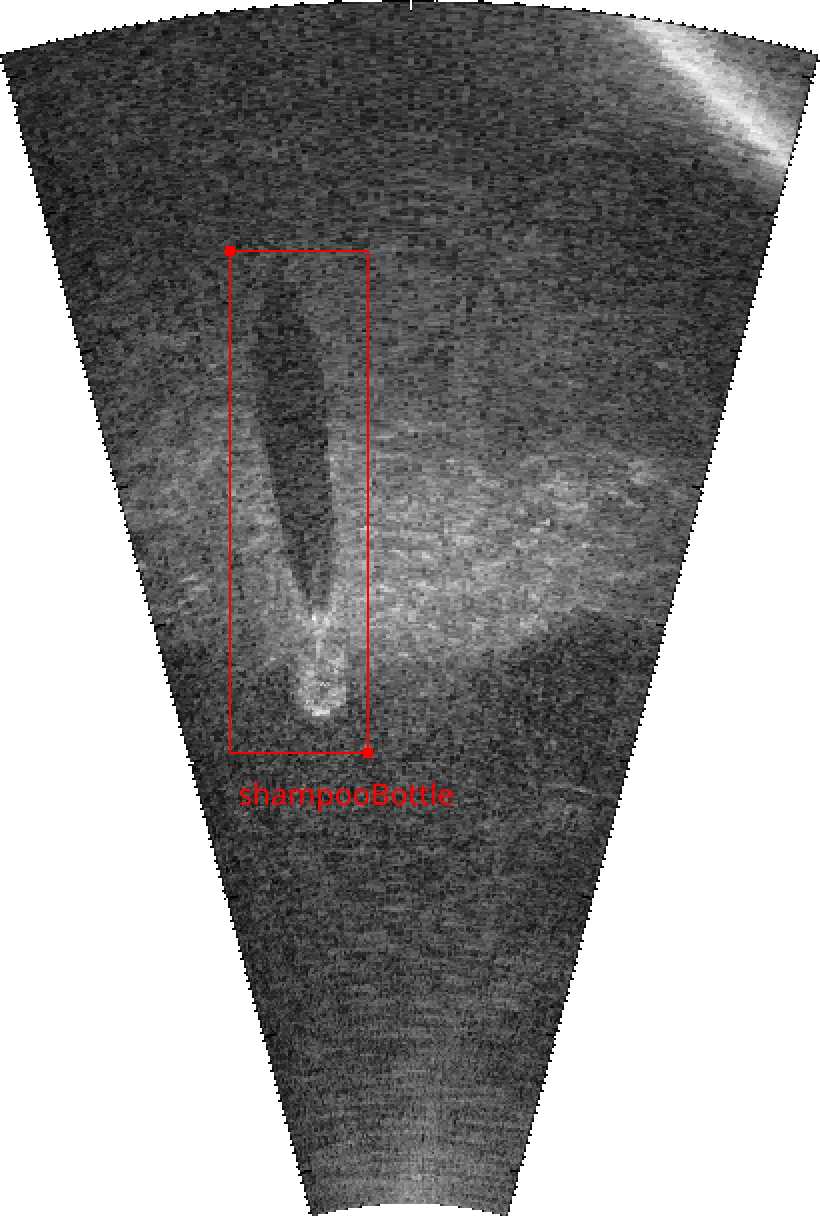}
    }

    \subfloat[Standing Bottle]{
        \includegraphics[height = 0.2\textheight]{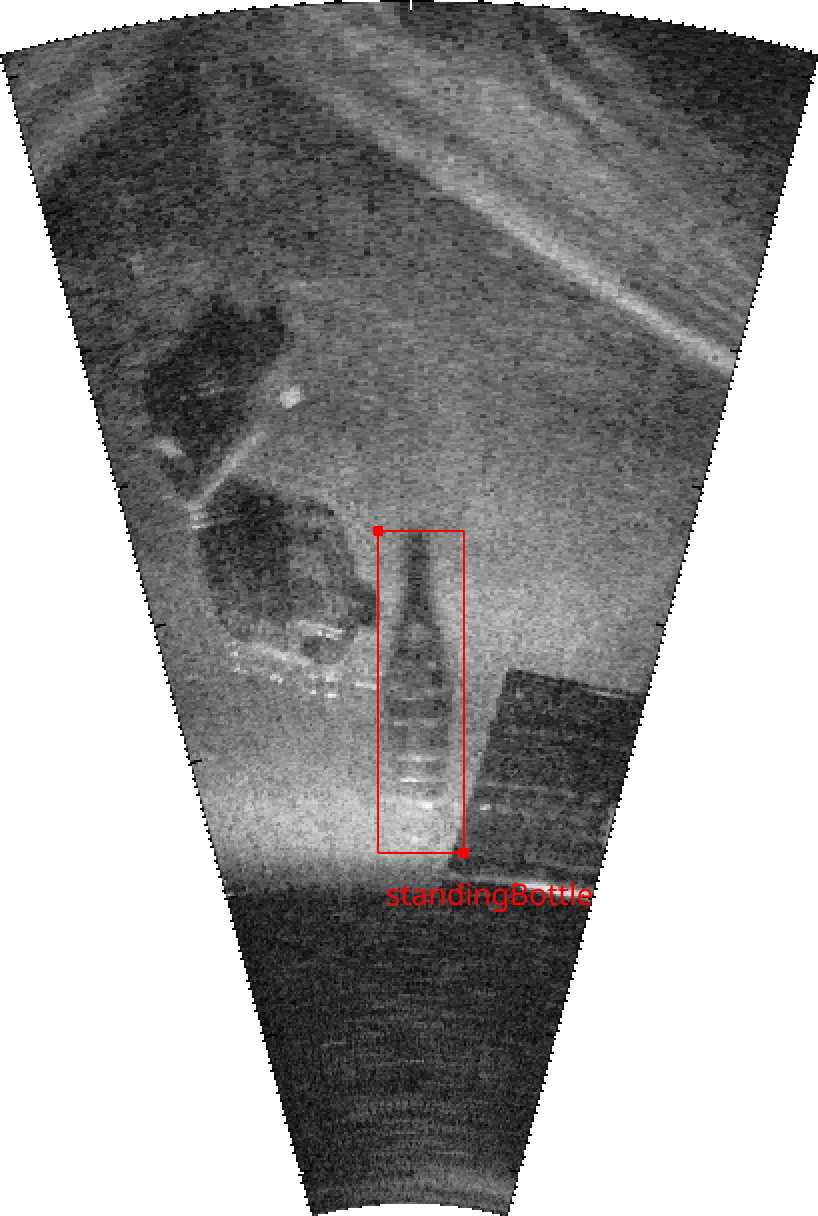}    
    }
    \subfloat[Tire, Bottle, and Valve]{
        \includegraphics[height = 0.2\textheight]{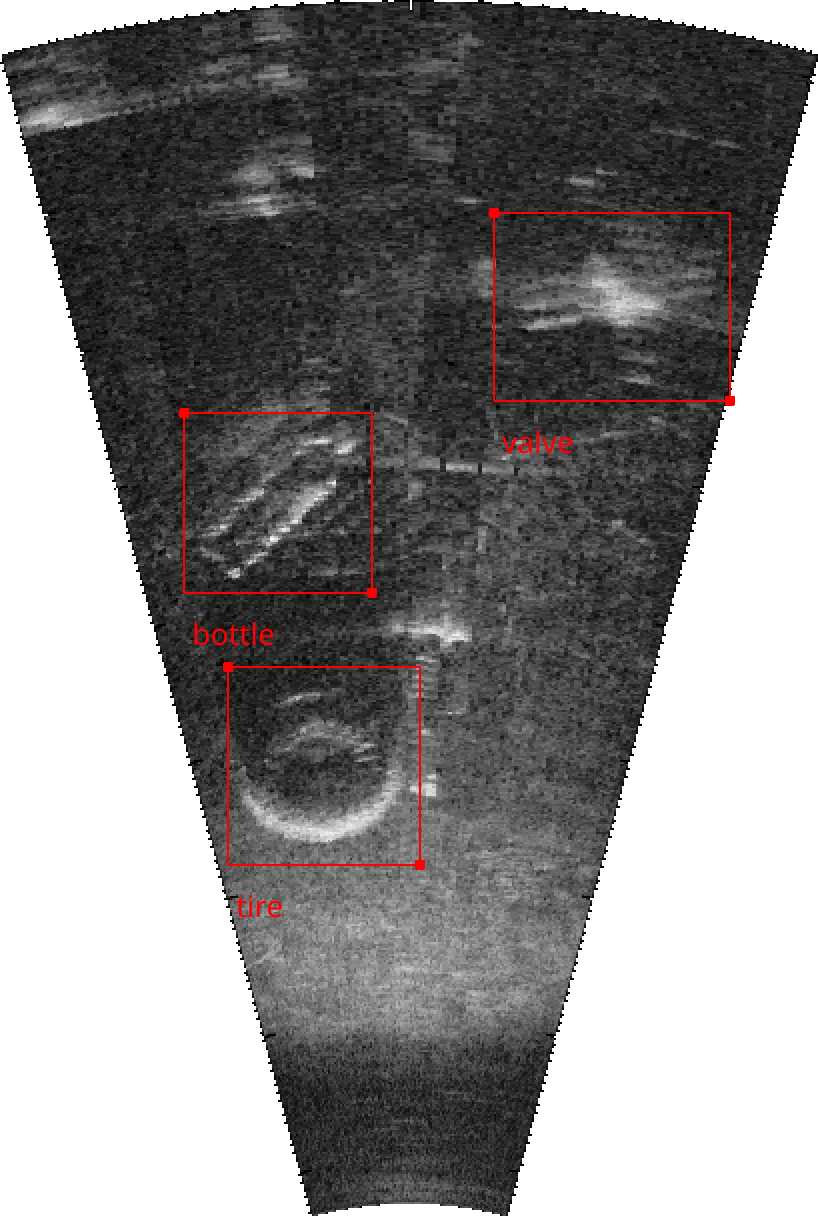}
    }
    \caption{Sample Sonar Views of Scenes with Bounding Box Annotations}
    \label{md:sonar-scenes}
    \setfloatalignment{b}
\end{figure}

\begin{figure}[t]
    \centering
    \forcerectofloat
    \subfloat[Bottle]{
        \includegraphics[height = 0.06\textheight]{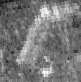}
        \includegraphics[height = 0.06\textheight]{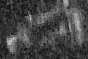}
        \includegraphics[height = 0.06\textheight]{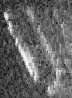}
    }\hspace*{0.5cm}
    \subfloat[Can]{
        \includegraphics[height = 0.06\textheight]{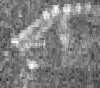}
        \includegraphics[height = 0.06\textheight]{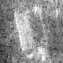}
        \includegraphics[height = 0.06\textheight]{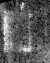}
    }
    
    \subfloat[Chain]{
        \includegraphics[height = 0.06\textheight]{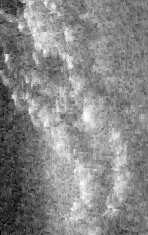}
        \includegraphics[height = 0.06\textheight]{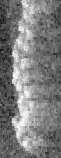}
        \includegraphics[height = 0.06\textheight]{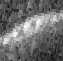}
    }\hspace*{1.9cm}
    \subfloat[Drink Carton]{
        \includegraphics[height = 0.06\textheight]{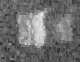}
        \includegraphics[height = 0.06\textheight]{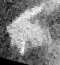}
        \includegraphics[height = 0.06\textheight]{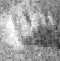}
    }

    \subfloat[Hook]{
        \includegraphics[height = 0.06\textheight]{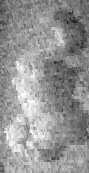}
        \includegraphics[height = 0.06\textheight]{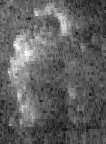}
        \includegraphics[height = 0.06\textheight]{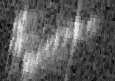}
    }\hspace*{1.4cm}
    \subfloat[Propeller]{
        \includegraphics[height = 0.06\textheight]{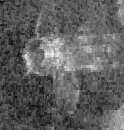}
        \includegraphics[height = 0.06\textheight]{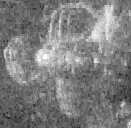}
        \includegraphics[height = 0.06\textheight]{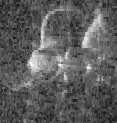}
    }

    \subfloat[Shampoo Bottle]{
        \includegraphics[height = 0.044\textheight]{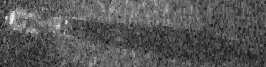}
        \includegraphics[height = 0.044\textheight]{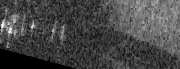}
        \includegraphics[height = 0.044\textheight]{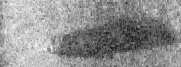}
    }

    \subfloat[Standing Bottle]{
        \includegraphics[height = 0.0375\textheight]{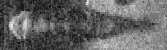}
        \includegraphics[height = 0.0375\textheight]{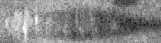}
        \includegraphics[height = 0.0375\textheight]{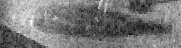}
    }

    \subfloat[Tire]{
        \includegraphics[height = 0.06\textheight]{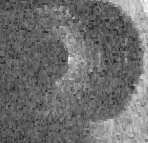}
        \includegraphics[height = 0.06\textheight]{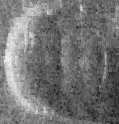}
        \includegraphics[height = 0.06\textheight]{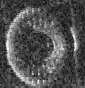}
    }\hspace*{1.1cm}
    \subfloat[Valve]{
        \includegraphics[height = 0.06\textheight]{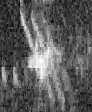}
        \includegraphics[height = 0.06\textheight]{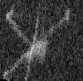}
        \includegraphics[height = 0.06\textheight]{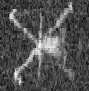}
    }
    \vspace*{0.5cm}
    \caption{Sample Sonar Image Crops from our Dataset, by Class}
    \label{md:sonar-object-crops}
    \setfloatalignment{b}
\end{figure}

Figure \ref{md:sonar-scenes} shows a selection of sonar images that we captured, with their corresponding bounding box and class annotations. In Figure \ref{md:sonar-object-crops} we show a selection of crops from full-size sonar images, separated per class.  Most labeled sonar images contain only a single object, and some contain multiple objects. Our images show that marine debris is not easy to visually recognize in sonar data, for example, we can only typically see the borders of a bottle, from which the shape is clear but not its contents. The standing bottle can only be recognized by its long shadow that replicated the bottle shape, but the highlight is camouflaged with the background, depending on how the sonar beams hit the object. 

\begin{marginfigure}
    \centering
    \includegraphics[width = 0.85\textwidth]{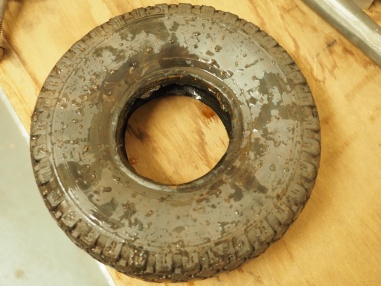}
    \caption{Sample of Tire Class}
    \label{md:tire}
\end{marginfigure}

Figure \ref{md:sonar-scenes} also shows some objects that we did not label, as they are not visually recognizable and look at blurry blobs in the image. This happens because the sonar is not pointing at the "right" angle to see such objects, or because the sonar is slightly higher than needed, producing blurry objects that are not in focus. We can also see the walls of the water tank as very strong reflections of a linear structure.

\begin{figure}
    \centering
    \begin{tikzpicture} 
        \begin{axis}[xbar, enlarge y limits=0.05, xlabel={Count}, symbolic y coords={ 
            Valve, Tire, Standing Bottle, Shampoo Bottle, Propeller, Hook, Drink Carton, Chain, Can, Bottle},
            ytick=data, nodes near coords, nodes near coords align={horizontal}, height = 0.25\textheight]
            \addplot coordinates {(449,Bottle) (367,Can) (226,Chain) (349,Drink Carton) (133,Hook)
                                  (137,Propeller) (99,Shampoo Bottle) (65,Standing Bottle) (331,Tire)
                                  (208,Valve)};
         \end{axis}
     \end{tikzpicture}
     \caption{Histogram of Samples per Class of the Marine Debris Dataset}
     \label{md:dataset-histogram}
\end{figure}
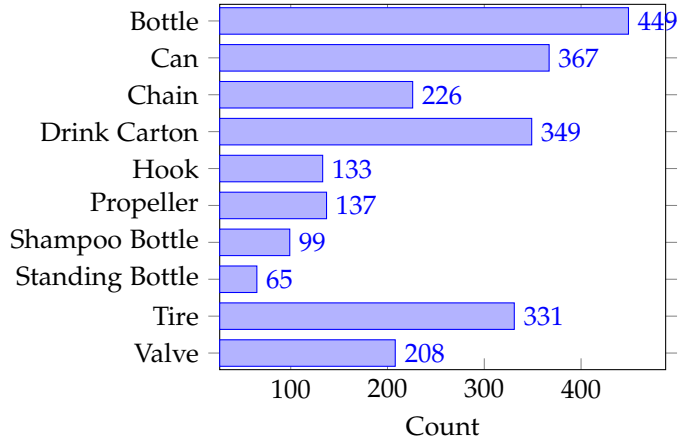

Figure \ref{md:dataset-histogram} shows the count distribution of the labels in our dataset. It is clearly unbalanced, as we made no effort to make a balanced dataset, and this does not prove to be an issue during learning, as there is no dominant class that would make learning fail.

\begin{marginfigure}
    \centering
    \includegraphics[width = 0.85\textwidth]{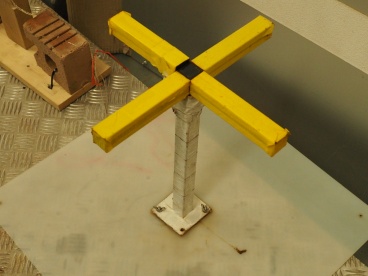}
    \caption{Sample of Valve Class}
    \label{md:valve}
\end{marginfigure}

In the Appendix, Figures \ref{appendix:can} to \ref{appendix:background} show randomly selected image crops of each class, with a variable number depending on the size of the object in order to fill one page. This shows the intra and inter-class variability of our dataset, which is not high, specially for the intra-class variability, due to the low number of object instances that we used.

%% file: chapters/background.tex
\chapter{Machine Learning Background}
\label{chapter:background}

This chapter is intended to be a gentle introduction to the use of machine learning techniques and neural networks for supervised classification and regression. This is required for two reasons: most researchers are aware of neural networks, as it is a well known topic, but Convolutional Neural Networks are less covered in the teaching literature \footnote{For example, there is currently only one book for Deep Learning by Goodfellow et al. \cite{Goodfellow2016deep}}. The topic of Deep Learning is understood by many as just "more layers" in a network, without considering the recent advances in the field starting from 2012. Small contributions like the Rectified Linear Unit (ReLU) or the ADAM optimizer, or bigger ones like Dropout and Batch Normalization, have allowed to push the limits of neural networks.

We aim to fill the gap in that common knowledge, and to provide a self-contained thesis that can be read by specialists that do not know neural networks in detail. We also made the effort to cover in detail many practical issues when training neural networks, such as how to tune hyper-parameters, and the proper machine learning model iteration loop from the point of view of the neural network designer, as well as the best practices when using machine learning models.

We also summarize my experience training neural networks, as it is a process that contains equal parts of science and art. The artistic part is quite of a problem, as it introduces "researcher degrees of freedom" that could skew the results. One must always be aware of this.

Training a neural network is not an easy task, as it requires a minimum the following steps:

\begin{description}
	\item[Preprocessing] Training set must be carefully prepared. Input and output data must be normalized, typically to the $[0,1]$ or $[-1, 1]$ ranges, else the network might not converge.
	The initial training set must also be split into at least two sets: Training and Testing. Typically a third Validation set is also included in the split, in order to monitor network performance during training, and to detect overfitting.
	\item[Architectural Design] An appropriate network architecture has to be designed, or reused from previous designs. The learning capacity of a network must match or surpass the capacity required by the task to be learned. A big problem is that the task capacity is usually unknown and is difficult to estimate, which means network designers must use a trial-and-error approach. The learning capacity is related to the number of parameters (weights) in the network, but this relationship is not simple \footnote{A clear example is the general decreasing trend in the number of parameters in networks for ImageNet classification.}.
	\item[Training] The network must be trained using gradient descent with respect to a loss function. A key parameter controlling the convergence of this algorithm is the learning rate $\alpha$. If $\alpha$ is too big, then training might diverge \footnote{Divergence is signalled by a loss value going to infinity, or becoming NaN (Not a Number).}, while if $\alpha$ is too small, then convergence may be slow. The optimal value of the learning rate has to be found by experimentation on a validation set.
	\item[Testing] After the network has been trained and the loss indicates convergence, then the network must be tested. Many computational frameworks include the testing step explicitly during training, as it helps to debug issues. The testing step requires a validation set and metrics on this set are reported continuously during training.
	\item[Production Use] After many iterations of neural network development, once performance on various test sets is satisfactory, the network can be used in a production setting to perform predictions as needed.
\end{description}

It requires careful preparation of training data, then an appropriate network architecture has to be designed or reused from a previous design. 

\section{Classical Neural Networks}

The basic unit of a neural network is the neuron, which computes the function:
\marginnote{Activation Notation}

\begin{equation}
	\begin{split}
		z(\textbf{x}) &= \sum_i \textbf{w}_i \textbf{x}_i + b = \textbf{w} \cdot \textbf{x} + b  \\
		a(\textbf{x}) &= g(z(\textbf{x}))
		\label{background:neuron}	
	\end{split}
\end{equation}

Where $\textbf{x}$ is a input vector of $n$ elements, $\textbf{w}$ is a learned weight vector, and $b$ is a scalar bias that completes an affine transformation of the input. $g(x)$ is a scalar activation function, which is intended to introduce non-linear behavior into the network. A neural network with no activation function can just compute a linear combination of its input, and therefore is very limited on what tasks can be learned.

$a(\textbf{x})$ is called the activation\footnote{Do not confuse activation and activation function} of a neuron, while $z(\textbf{x})$ is usually called the pre-activation of a neuron, or just the output value before the activation is applied. This notation will be useful later.

\marginnote{Multilayer Perceptron} A neural network is then made by "stacking" a certain number of "layers", where each layer contains a predefined number of neurons. This is also called a Multilayer Perceptron (MLP). 
It is common to then represent inputs and outputs to each layer as vectors (or tensors), as this allows for explicit vector representation that can be easily accelerated in CPUs and GPUs. As an example, a three layer network would compute:

\begin{align*}
	z_1 &= \Theta_1 \cdot \textbf{x} + B_1\\
	a_1 &= g(z_1)\\
	z_2 &= \Theta_2 \cdot a_1 + B_2\\
	a_2 &= g(z_2)\\
	z_3 &= \Theta_3 \cdot a_2 + B_3\\
	a_3 &= g(z_3)
\end{align*}

Where now the $\Theta_i = \{ \textbf{w}_j \}_j$ is a matrix that contains the weights (row-wise) for the $i$-th layer, the input $\textbf{x}$ is a column vector and biases are stored in a row vector $\textbf{B}_i$. It should be noted that the number of neurons in each layer does not have to be the same, and the number of neurons in the last layer defines the output dimensionality of the network. Note than an alternate notation of $\Theta$ can include Biases $\textbf{B}$ as well, which is useful for a more direct implementation.

Training most machine learning models consists of minimizing a loss function $L$, which changes the model parameters in a way that the predictive performance of the model increases. In a sense, the loss function measures how well the model is doing with the current value of the parameter set. As neural networks usually have a large\footnote{AlexNet has 60 million trainable parameters (weights), ResNet has 20 million parameters, and VGG-16 has 138 million parameters.} number of parameters (the number of weights), unconstrained optimization algorithms are used to minimize the loss. The most common optimization algorithm is gradient descent.

Gradient descent uses the theoretical implications that the gradient of a scalar function points in the direction of maximum increase rate of that function. Then it is intuitive to think the rate of maximum decrease is exactly the opposite direction that of the gradient. Then gradient descent is an iterative algorithm that updates the parameters with the following relation:
\vspace*{1em}
\begin{equation}
	\Theta_{n+1} = \Theta_{n} - \alpha \nabla L(\hat{y}, y)
	\label{background:simpleGD}
\end{equation}

Where $\hat{y} = h_{\Theta_{n}}(\textbf{x})$ is the neural network output computed over inputs $\textbf{x}$ from a given dataset, with parameters $\Theta_n$, and loss function $L(\hat{y}, y)$. A key parameter in gradient descent is the \marginnote{Learning Rate} learning rate $\alpha$, which controls the "speed" at which the network learns. Intuitively the learning rate is a step size, as the gradient only provides a direction on which the parameters can be moved to decrease the loss, but not a magnitude on how much to move. This parameter has to be tuned in a validation set. If the learning rate is larger than necessary, then the loss value could oscillate or diverge. If the learning rate is too small, then convergence will be slow. The proper value of the learning rate will make convergence at an appropriate rate.

\subsection{Training as Optimization}

Training any machine learning model is typically formulated as an optimization problem. The most common formulation is the minimization of an objective function. This is typically called the \textit{loss function}, and it is designed in such a way that the model performance improves when the loss function decreases.

As previously mentioned, once a ML model is constructed, a loss function must be defined so the model can be trained. Gradient descent \marginnote{Gradient Descent} is the most common algorithm to train DNNs:
\vspace*{1em}
\begin{equation}
\Theta_{n+1} = \Theta_{n} - \alpha \nabla L(\hat{y}, y)
\label{background:fullGD}
\end{equation}

Where $\hat{y} = h_{\Theta_{n}}(\textbf{x})$ is the neural network output computed over inputs $\textbf{x}$ from a given dataset, with parameters $\Theta_n$, and loss function $L(h_{\Theta_{n}}(\textbf{x}), y)$. Gradient descent is an iterative algorithm and Eq \ref{background:fullGD} is executed for a predefined number of steps $n$. The value of the loss function at each step must be monitored as a simple check that the loss function is being minimized and is decreasing after each step.

The quality of the solutions obtained by gradient descent depends on several factors:

\begin{description}
	\item[Gradient Quality] Gradients must be of "high quality", which typically means that they must have many non-zero values. A related common problem is the vanishing gradient problem, specially in deeper networks, where the function composition nature of a DNN makes the gradients very small, slowing down or preventing training. Low quality gradients have many zero elements that prevent many parameters from converging to their optimal values.
	
	\item[Learning Rate] Setting the right value of the learning rate is a key factor to using gradient descent successfully. Typical values of the learning rate are $ \alpha \in [0, 1]$, and common starting values are $10^{-1}$ or $10^{-2}$.
	It is important that the learning rate is set to the right value before training, as a larger than necessary learning rate can make the optimization process fail (by overshooting the optimum or making the process unstable), a lower learning rate can converge slowly to the optimum, and the "right" learning rate will make the process converge at an appropriate speed.
	The learning rate can also be changed during training, and this is called  Learning Rate Schedule\marginnote{Learning Rate Schedule}. Common approaches are to decrease the learning rate by a factor after a certain number of iterations, or to decay the learning rate by a factor after each step.
	
	\item[Loss Surface] The geometry and smoothness of the loss surface is also key to good training, as it defines the quality of the gradients. The ideal loss function should be convex on the network parameters, but typically this is not the case for outputs produced by DNNs. Non-convexity leads to multiple local optima where gradient descent can become "stuck". In practice a non-convex loss function is not a big problem, and many theoretical results show that the local optima in deep neural networks are very similar and close in terms of loss value.
\end{description}

Gradient descent makes several implicit assumptions: the dataset fits into the computer's RAM, and that computing the loss function for the whole dataset is not an expensive operation. One forward pass of the complete network is required for each data point in the training set, and with complex neural networks and datasets of considerable size, these assumptions do not hold true.

\marginnote{Gradient Descent Variations} A simple way to overcome this problem is to only use part of the training set at a time during the training process. Assuming that the training set can be split into equal sized and non-intersecting sets, called batches, then we can iterate over batches, compute the loss function for a batch, compute the gradients, and apply one step of gradient descent. This is called Mini-Batch Gradient Descent (MGD):
\vspace*{1em}
\begin{equation}
\Theta_{n+1} = \Theta_{n} - \alpha \nabla L(h_{\Theta_{n}}(\textbf{x}_{i:j}), y)
\label{background:SGD}
\end{equation}

Where $\textbf{x}_{i:j}$ denotes that the neural network $h_{\Theta_{n}}(\textbf{x})$ and loss function $L$ are only evaluated for inputs $\textbf{x}_k$ where $k = [i, i + 1, i + 2, \dots, j]$. Then each batch is made by setting different values if $i$ and $j$, constrained to $i > j$ and $B = j - i$. The hyper-parameter $B$ is denoted the Batch Size \marginnote{Batch Size}. Typically batches are made by setting $B$ to a given value and using values $i, j = \{ (0, B), (B, 2B), (2B, 3B), \dots, (cB, n) \}$. Note that not all batches have the same size due that $B$ might not divide $|Tr|$ exactly. That means the last batch can be smaller. A variation of MGD is Stochastic Gradient Descent (SGD), where simply $B$ is set to one.
After approximately $\frac{|Tr|}{B}$ iterations of gradient descent, the learning process will have "seen" the whole dataset. This is called an Epoch \marginnote{Epochs}, and corresponds to one single pass over the complete dataset. Typically training length is controlled by another hyper-parameter, the Number of Epochs $M$.

Setting the value of the Batch Size $B$ is one hyper-parameter that controls the tradeoff between more RAM use during training, or more computation. It must be pointed out that MGD and SGD all introduce noise into the learning process, due to the approximation of the true gradient with the per-batch gradient. Larger values of $B$ use more RAM, but require less number of iterations, and additionally it reduces noise in the gradient approximation. Smaller values of $B$ require less RAM but more iterations and provide a more stable gradient approximation. It is common that $B$ is set such as the training process fills the maximum amount of RAM available\footnote{While training on a GPU, it is typical that the amount of GPU RAM is the only limitation to set $B$. The same applies while training on CPU but with system RAM.}.

The exact equations that compute gradients of the loss $\nabla L$ depend on network architecture. The back-propagation \cite[1em]{bishop2006pattern} algorithm is commonly referred to as a way to hierarchically compute gradients in multi-layer neural networks. In practice this algorithm is rarely used, as modern neural network frameworks such as TensorFlow and Theano use automatic differentiation \cite{baydin2017automatic} (AD) to compute gradients of the loss function automatically, making the developer's life much easier, as exotic architectures can be easily experimented.

\subsection{Loss Functions}

We now describe the most common loss functions used to train Neural Networks. A loss function is a scalar function $\mathbb{R}^n \times \mathbb{O} \rightarrow R^{+} \cup 0$ that gives a score to a set of predictions from a learning algorithm (typically a classifier or a regressor). Set $\mathbb{O}$ defines the ground truth labels, and in the case of regression it is typically $\mathbf{R}$, $[0, 1]$ or $[-1, 1]$. For classification then $\mathbb{O}$ is the set of classes, converted to a numerical form.

The most basic loss function is the mean squared error (MSE), typically used for regression:

\begin{equation}
	MSE(\hat{y}, y) = n^{-1} \sum_{i=0}^{n} (\hat{y}_i - y_i)^2
\end{equation}

The MSE loss penalizes the predicted values $\hat{y}$ that diverge from the ground truth values $y$. The error is defined just as the difference between $\hat{y}$ and $y$, and squaring is done to get a smooth positive value. One problem with the MSE is that due to the square term, large errors are penalized more heavily than smaller ones. This produces a practical problem where using the MSE loss might lead the convergence of the output to a mean of the ground truth values instead of predicting values close to them.
This issue could be reduced by using the Mean Absolute Error (MAE), which is just the mean of absolute values of errors:

\begin{equation}
    MAE(\hat{y}, y) = n^{-1} \sum_{i=0}^{n} |\hat{y}_i - y_i|
\end{equation}

The MSE is also called the $L_2$ loss, while the MAE is named as $L_1$ loss, both defined as the order of the norm applied to the errors. Note that the MAE/$L_1$ loss is not differentiable at the origin, but generally this is not a big issue. The $L_1$ loss can recover the median of the targets, in contrast to the mean recovered by the $L_2$ loss.

For classification, the cross-entropy loss function is preferred, as it produces a much smoother loss surface, and it does not have the outlier weighting problems of the MSE. Given a classifier that outputs a probability value $\hat{y}^c$ for each class $c$, then the categorical cross-entropy loss function is defined as:

\begin{equation}
	CE(\hat{y}, y) = -\sum_{i=0}^n \sum_{c=0}^C y_i^c \log \hat{y}_i^c
\end{equation}

Minimizing the cross-entropy between ground truth probability distribution and the predicted distribution is the equivalent to minimizing the Kullback-Leibler divergence \cite{mackay2003information}. For the case of binary classification, then there is a simplification usually called binary cross-entropy:

\begin{equation}
BCE(\hat{y}, y) = -\sum_{i=0}^n \left[ y_i  \log \hat{y}_i + (1 - y_i) \log (1 - \hat{y}_i) \right]
\end{equation}

In this case $\hat{y}$ is the probability of the positive class.

\subsection{Activation Functions}

There is a large selection of activation functions that can be used. A small summary is shown in Table \ref{background:activations}. The most common "classic" activation functions are the sigmoid and the hyperbolic tangent (TanH). These activation functions dominated the neural networks literature before 2010, as they produce the well known problem of vanishing gradient.

The vanishing gradient problem \marginnote{Vanishing Gradient Problem} happens when the gradient of the activation function becomes zero, and this is problematic because the network stops training. Looking at Figure \ref{background:saturatingActivations}, it can be seen that the activation function "saturates" when the input is small or large, making the gradient effectively zero. Stacking multiple layers that use sigmoid or TanH activation functions amplifies this effect and prevent the use of a large number of layers.

For this reason, sigmoid and TanH are called saturating activation functions \marginnote{Saturating Activation Functions}. This sparked the development of non-saturating activation functions, and the most well known of such functions is the Rectified Non-Linear Unit (ReLU) \cite{glorot2011deep}. The ReLu is a very simple function given by:

\begin{equation}
    g(x) = \max(0, x)
    \label{background:relu}
\end{equation}

This \marginnote{Rectified Linear Unit (ReLU)} activation function has constant output of zero for negative input, and a linear output for positive inputs. It should be noted that the ReLU is not differentiable at $x = 0$, as the slope at each side of the origin is different, but this usually poses no practical problem.

Use of the ReLU as activation function is one reason why Deep Learning is possible now. The breakthrough paper by Krizhevsky et al. \cite[-8em]{krizhevsky2012imagenet} mentions that using the ReLU as an activation function requires 4 times less iterations for GD to converge at the same loss value when compared to a sigmoid activation. There are multiple reports that using ReLU activations leads to loss surfaces that are easier to optimize and are less prone to local minima. One reason for this behavior is that the ReLU makes a network prefer sparse outputs.

\begin{table}[t]
	\centering
	\begin{tabular}{@{}lll@{}}
		\hline
		Name					& Range 			& Function\\
		\hline
		Linear					& $[-\infty, \infty]$	& $g(x) = x$\\
		Sigmoid					& $[0, 1]$			& $g(x) = (1 + e^{-x})^{-1}$\\
		Hyperbolic Tangent		& $[-1, 1]$			& $g(x) = (e^{2x} -1)(e^{-2x} + 1)^{-1}$\\
		\hline
		ReLU					& $[0, \infty]$		& $g(x) = \max(0, x)$\\
		SoftPlus				& $[0, \infty]$		& $g(x) = \ln(1 + e^x)$\\
		SoftMax					& $[0, 1]^n$		& $g(\textbf{x}) = (e^{x_i}) (\sum_k e^{x_k})^{-1}$\\
		\hline
	\end{tabular}
	\caption{Summary of commonly used activation functions.}
	\label{background:activations}
\end{table}

\begin{marginfigure}[-8em]
	\begin{tikzpicture}
	 	\begin{axis}[width = 0.9\textwidth, xlabel=, ylabel=Activation, ymin = -1.1, ymax = 1.1, legend style={at={(0.5, -0.3)},anchor=north}]
	 		\addplot+[mark=none] {1.0 / (1.0 + exp(-x))};
	 		\addlegendentry{Sigmoid}
	 		\addplot+[mark=none] {tanh(x)};
	 		\addlegendentry{TanH}

	 	\end{axis}
	 \end{tikzpicture}
	 \caption{Saturating Activation Functions}
	 \label{background:saturatingActivations}
\end{marginfigure}

Another commonly used activation function is the softmax \cite{bishop2006pattern}, and unlike the previously seen activations, it is not a scalar function. Instead the softmax takes a vector and transforms the input into a discrete probability distribution. This is very useful for multi-class classification, as a neural network can then output a probability distribution over class labels in $[0, 1, 2, \cdots, C]$. Then recovering the discrete class can be performed as taking the class with maximum probability. The vector definition of the softmax activation function is:

\begin{equation}
	g(\textbf{x}) = \left[ \frac{e^{x_i}}{\sum_j e^{x_j} } \right]_i
	\label{background:softmax}
\end{equation}

Given a softmax output $\textbf{a}$, the class decision can be obtained by:
\vspace*{1em}
\begin{equation}
	c = \argmax_i a_i
\end{equation}

\begin{marginfigure}
    \begin{tikzpicture}
    \begin{axis}[width = 0.9\textwidth, xlabel=, ylabel=Activation, ymin = -0.1, ymax = 4.0, xmin = -4.0, xmax = 4.0, legend style={at={(0.5, -0.3)},anchor=north}]
    \addplot+[mark=none] {max(0, x)};
    \addlegendentry{ReLU}
    \addplot+[mark=none] {ln(1 + exp(x))};
    \addlegendentry{Softplus}
    \end{axis}
    \end{tikzpicture}
    \caption{Non-Saturating Activation Functions}
    \label{background:nonSaturatingActivations}
\end{marginfigure}

Looking at Equation \ref{background:softmax} one can see that softmax outputs are then "tied" by the normalization value in the denominator. This produces a comparison operatiog between the inputs, and the biggest softmax output will always be located at the largest input relative to the other inputs. Inputs to a softmax  are typically called logits. As the softmax operation is differentiable, its use as an activation function then produces a loss surface that is easier to optimize.

Softmax combined with a categorical cross-entropy loss function is the base building block to construct DNN classifiers.

As the ReLU is not differentiable at $x = 0$ and it has constant zero output for negative inputs, this could produce a new kind of problem called "dying ReLU", where neurons that use ReLU can stop learning completely if they output negative values. As the activations and gradients become zero, the neuron can "get stuck" and not learn anymore. While this problem does not happen very often in practice, it can be prevented by using other kinds of activation functions like the Softplus function, which can be seen as a "softer" version of the ReLU that only has a zero gradient as the limit when $x \rightarrow -\infty$. Figure \ref{background:nonSaturatingActivations} shows the Softplus versus the ReLU activation functions. The Softplus function is given by:

\begin{equation}
	g(x) = \ln(1 + exp(x))
\end{equation}

Another similar activation function is the Exponential Linear Unit:
\vspace*{-0.1cm}
\begin{equation}
	g(x) = 
	\begin{cases}
		x       			& \quad \text{if } x \geq 0\\
		\gamma (e^x - 1) 	& \quad \text{if } x < 0\\
	\end{cases}
\end{equation}

There is a clear trend in recent literature about the use of learnable activations, where the activation function has a parameter that can be tuned during learning. Examples of this are the PReLU, Leaky ReLU and the MaxOut.

As a general rule, most deep neural networks use exclusively the ReLU as activation function, and when designing new networks, it should be preferred as it completely avoids the vanishing gradient problem.

\subsection{Weight Initialization}

SGD gives a way to iteratively improve the weight matrix to reduce some loss function that controls how and what the model is learning. But it does not specify the initial values of the weights. These are typically initialized by setting them to a random value drawn from some probability distribution. Weights cannot be initialized to zero, since this would lead to all neurons producing a zero value, and the network outputting constant zero, producing a failed learning process. Randomizing weights breaks the "symmetry" of initializing them to a particular value. If initial weights are too large, it could produce chaotic behavior (exploding gradients) and make the training process fail. There are many distributions that are used to draw initial weights:

\begin{description}
	\item[Uniform] Draw the weights from a Uniform distribution with a fixed range, parametrized by a scale parameter $s$. Popular values for $s$ are $s \in [0.1, 0.01, 0.05]$, and a common heuristic is to set $s = n^{-0.5}$, where $n$ is the dimensionality of the input to the layer.
        \vspace*{1em}
		\begin{equation}
			w \sim U(-s, s)
		\end{equation}
	\item[Gaussian] Draw the weights from a Gaussian distribution with a fixed standard deviation. Popular values are $\sigma \in [0.1, 0.01, 0.05]$
        \vspace*{1em}
		\begin{equation}
			w \sim N(0, \sigma)
		\end{equation}
	\item[Glorot or Xavier] Named after Xavier Glorot \cite{glorot2010understanding}. Draw the weights from:
		\begin{equation}
			w \sim U(-s, s) \qquad s^2 = \frac{6}{F_{in} + F_{out}}
		\end{equation}
         Where $F_{in}$ is the number of input elements to the neuron, and $F_{out}$ is the number of output elements. This initialization scheme is based on the intuition that variance of the input and output should be approximately equal for a stable behavior at the beginning of training, which implies that initial weights are scaled differently depending on the dimensionality of the inputs and outputs.
	\item[Orthogonal] Draw weights from a random orthogonal matrix, after applying gain scaling \cite[-6em]{saxe2013exact}. This method is theoretically grounded and guarantees that convergence will be achieved in a number of iterations that is independent of network depth. The weight matrix is generated by first generating a random matrix with elements $w_{ij} \sim N(0, 1)$, then performing Singular Value Decomposition on $w$, and then picking either the $U$ or $V$ matrices depending on the required output dimensionality. Finally this matrix is scaled by a gain factor $g$ that depends on the activation function, in order to ensure stable learning dynamics.
\end{description}

Biases $b$ can also be initialized with the same scheme as weights, but there is no problem if bias are initialized to zero, and some authors prefer this. We have shown a small subset of weight initialization schemes available in the literature, and overall there is no consensus if one is superior to another, and in general it does not matter much which one is chosen, as other learning tools\footnote{Like Batch Normalization, Dropout, and Non-Saturating Activation Functions.} can be used to provide superior training stability. Before these techniques were known, the neural network designed had to carefully adjust the weight initialization scheme in order for learning to converge to a useful solution.

\subsection{Data Normalization}

One key technique that always must be used \footnote{My experience from answering Stack Overflow questions is that too many people do not normalize their data, making training their networks much more difficult.} is data normalization. As the input and output data typically comes from real sources, it is contaminated by non-ideal behaviors, like different scales for each feature, or simply are in a range that typical neural networks have issues modeling.

The scale of input features is an important issue as if inputs have different ranges, then the weights associated to those features will be in different scales. Since we usually use fixed learning rates, this leads to the problem that some parts of a neuron learn at different speeds than others, and this issue propagates through the network, making learning harder as the network becomes deeper.

The scale of outputs poses a different but easier problem. The designer has to make sure that the range of the activation of the output layer matches the range of the desired targets. If these do not match, then learning will be poor or not possible. Matching the ranges will make sure that learning happens smoothly.

\begin{description}
	\item[Min-Max Normalization] Normalize each component of the input vector by subtracting the minimum value of that component, and divide by the range. This produces values in the $[0, 1]$ range.
    
		\begin{equation}
			\hat{x} = \frac{x - \min_i x_i}{\max_i x_i - \min_i x_i}
		\end{equation}
		
	\item[Z-Score Normalization] Subtract the sample mean $\mu_x$ and divide by the sample standard deviation $\sigma_x$. This produces values that are approximately in the $[-1, 1]$ range.
    
		\begin{equation}
			\hat{x} = \frac{x - \mu_x}{\sigma_x}
		\end{equation}
        \begin{equation}
            \mu_x = n^{-1} \sum x_i \qquad \sigma_x = \sqrt{(n-1)^{-1} \sum (x_i - \mu_x)^2}
        \end{equation}
        
    \item[Mean Substraction] Typically used to train models that take images as inputs \cite{krizhevsky2012imagenet}. Images are represented either as tensors $(W, H, C)$ with values in the $[0, 255]$ or $[0, 1]$ range, and they are normalized by computing the mean image over the training set, or the individual per-channel means over the training set, and subtracting this mean from each image, which overall will produce values in the $[-128, 128]$ or $[-0.5, 0.5]$ range.
\end{description}

\subsection{Regularization}

Regularization is a way to control the learning capacity of a machine learning model, by imposing constraints or prior information to bias the model to a preferred configuration. There are many ways to regularize neural networks, and in this section we will describe two modern regularization techniques: Dropout and Batch Normalization.

A common view of regularization from statistical learning theory is that it controls the number of trainable parameters, which is related to overfitting \cite[-5em]{bishop2006pattern}, but a more recent view \cite[-2em]{luo2018understanding} is that the number of parameters does not completely explain overfitting and the expected predictive performance at inference time.

Dropout \marginnote{Dropout} is a technique pioneered by Srivastava et al \cite[1em]{srivastava2014dropout}. The authors of Dropout noticed that when a neural network overfits, the neurons inside it \textit{co-adapt} or their outputs become correlated. This reduces model efficiency and generalization greatly. They proposed that this co-adaptation can be broken by introducing noise in the activations, and they choose a model that produces a mask $\mathbf{m}$ of length $n$ where each element is Bernoulli distributed with probability $p$: $m_i \sim \text{Bernoulli}(p)$.

Then this mask is multiplied with the activations of a particular layer, which has the effect of \textit{turning off} some activations of a layer, and letting others pass unchanged. This is called the dropout mechanism. During training the masks at each Dropout layer are randomly sampled at each iteration, meaning that these masks change during training and mask different activations at an output. This breaks any correlations between activations in one layer and the one before it (where the Dropout layer is placed), meaning that more strong features can be learned and co-adaptation of neurons is prevented.

At inference or test time, Dropout layers do not perform any stochastic dropping of neurons, and instead they just multiply any incoming activation by $p$, which accounts for all activations being present during inference, unlike at training time. This also prevents any kind of stochastic effect during inference. It should also be noted that Dropout can also be used with its stochastic behavior at inference time, which produces very powerful model uncertainty estimates, as it was proven by Gal et al 2015. \cite{gal2015dropout}.

Using Dropout layers in a neural network, typically before fully connected ones, has the effect of reducing overfitting and improving generalization significantly.

Dropout can also be seen as a way of Bayesian model averaging \cite{gal2016uncertainty} as when dropping neurons at training time, new architectures with less neurons are produced, which are then averaged at inference time, which is a theoretical explanation of why Dropout works and improves generalization. Note that while the number of effective parameters during training is reduced by Dropout (by a factor of $p$) due to the dropping of activations, during inference/testing the number of parameters does not change, and all parameters are used to make a prediction.

Another technique that can be used for regularization is Batch Normalization \cite[1em]{ioffe2015batch}. This technique was not specifically designed to reduce overfitting in neural networks, but to increase convergence speed while training. The authors of Batch Normalization also noticed that it has a powerful regularization effect, and most current deep neural networks are trained with it, as it improves generalization almost "for free".

Batch Normalization \marginnote{Batch Normalization} is based on the concept of internal covariate shift reduction. Given a set of layers, the covariate is the empirical probability distribution of their outputs. Covariate shift is the change of that distribution as training progresses. Internal covariate shift is the covariate shift of the hidden layers of a neural network. The intuition for Batch Normalization is that drastic internal covariate shift is prejudicial for neural network training, as it is more likely to saturate non-linearities, and unstable activation distributions will need more training iterations to converge successfully. Both situations causes slow training convergence.

Reducing the internal covariate shift can be achieved by normalizing the activations (before applying a non-linearity). As typical decorrelation methods like PCA or ZCA whitening are too expensive to use during neural network training (specially with high dimensional data), the Batch Normalization authors proposed two simplifications.

The first is to perform Component-Wise Normalization along the features. Given a vector of activations $\textbf{x}$, as a middle step for layer output computation, then each component of that vector should be normalized independently, by using a simple mean and standard deviation normalization: 
\vspace*{1em}
\begin{equation}
    \mu_{B} = |B|^{-1} \sum x_i \qquad \sigma^2_{B} = |B|^{-1} \sum (x_i - \bar{x})
\end{equation}

The normalization happens along the features dimension of the a mini-batch of activations. This allow for an efficient implementation using SGD. Normalizing a mini-batch $\textbf{x} \in B$ of size $|B|$ is performed as:
\vspace*{1em}
\begin{equation}
    \hat{x}_i = \frac{x_i - \mu_{B}}{\sqrt{\sigma^2_{B} + \epsilon}}
\end{equation}

Where $\epsilon = 0.001$ is a small constant to prevent division by zero. Normalizing activations has the unintended effect of destroying the representation capability of the network, but it can be easily restored with a linear transformation:
\vspace*{1em}
\begin{equation}
    y_i = \gamma_i \hat{x}_i + \beta_i
\end{equation}

Where $\gamma_i$ and $\beta_i$ are scalar parameters that are learned using gradient descent\footnote{These parameters are added to the set of learnable parameters in a layer}. It should be noted that this transformation does not correspond to a fully connected layer, as these are per-feature scaling and bias coefficients for the activation instead.

At inference time, mini-batch statistics are not available, as many inference calls use a single test sample. Fixed normalization coefficients can then be estimated from the training set and used during inference. This is performed as part of the learning process as a exponentially averaged running mean and variance of the mini-batch activation statistics, but with an unbiased variance estimate $\frac{n}{n-1} E[\sigma^2_{B}]$ used instead.

Recently more advanced versions of activation normalization schemes have appeared in the literature, such as Layer Normalization \cite[-7em]{ba2016layer}, Instance Normalization, and Group Normalization \cite[-2em]{wu2018group}. These methods expand Batch Normalization to specific use cases and make it less dependent on mini-batch sizes.

The inclusion of Batch Normalization in neural network architectures is considered to be a good practice, and most modern neural network architectures (like ResNet, GoogleNet, DenseNets, etc) use it. Using batch normalization has a regularizing effect, generally improving performance. There is strong evidence that this is due to a smoother loss surface \cite[-6em]{santurkar2018does} as the result of activation normalization, and not to a reduction in the covariate shift, as the original paper argues.

Luo et al. \cite{luo2018understanding} provide a theoretical framework for the understanding of the regularization effect of Batch Normalization. They find that Batch Normalization is an implicit regularized that can be decomposed into population normalization and gamma decay, the latter being an explicit regularizer. These results show that Batch Normalization of CNNs shares many properties of regularization.

Classical regularization techniques for machine learning models can also be used for neural networks, and the most common ones are $L_1$ and $L_2$ regularization \marginnote{$L_1$ and $L_2$ Regularization or Weight Decay} (also called Weight Decay). Both of them add a term $\lambda \sum_i ||w_i||^p$ to the loss function, which penalizes large weights that are not supported by evidence from the data. $p$ is the order of the norm that is being computed over the weights.

\subsection{Optimizers}

Optimizers are algorithms that perform minimization of the loss function through gradient descent, as mentioned previously. The standard formulation of SGD uses a constant learning rate, but in practice this does not have to be the case, and a rich literature\cite{ruder2016overview} exists on optimization methods that scale the learning rate with gradient information, so an individual learning rate is used for each trainable parameter in $\Theta$.

The most straightforward way to incorporate this idea is to use the square root of the sum of past squared gradients as a scaling factor for the learning rate. An optimizer using this idea is AdaGrad \cite{duchi2011adaptive}, with the update rule:
\vspace*{1em}
\begin{align}
    g_n &= \nabla L(h_{\Theta_{n}}(\textbf{x}_{i:j}), y)\nonumber \\
    r_n &= r_{n-1} + g_n \odot g_n\nonumber\\
    \Theta_{n+1} &= \Theta_{n} - \alpha \frac{g_n}{\epsilon + \sqrt{r_n}}
    \label{background:AdaGrad}
\end{align}

Where $\odot$ represents component-wise multiplication, and $\epsilon$ is a small constant for numerical stability and to prevent division by zero. The gradient is divided scaled component-wise by the square root of sum of gradients as well, in order to scale the learning rate for each parameter independently. AdaGrad works most of the time, but the use of a complete history of gradients makes it unstable, since any gradient disturbance at the beginning of training would over-reduce the gradients and prevent the model from reaching its full potential.

An alternative to AdaGrad is RMSProp \cite{tieleman2012lecture}, where instead of keeping a sum of the full history of squared gradients, a moving exponential weighted average is kept, so older squared gradient are given less importance than more recent ones. The update rule is:
\vspace*{1em}
\begin{align}
    r_n &= \rho r_{n-1} + (1 - \rho) g_n \odot g_n\nonumber\\
    \Theta_{n+1} &= \Theta_{n} - \alpha \frac{g_n}{\sqrt{\epsilon + r_n}}
    \label{background:RMSProp}
\end{align}

Where $\rho$ is a decay parameter that controls the weight of past squared gradients through the moving average, usually it is set to $0.9$ or $0.99$. RMSProp is more stable than AdaGrad due to better control of the history of squared gradients, and allows a model to reach a better optima, which improves generalization. It is regularly used by practitioners as one of the first methods to start training a model.

Another advanced Optimizer algorithm is Adam \cite[-1cm]{kingma2014adam}, which stands for \textit{adaptive moments}. Adam combines improvements of RMSProp with direct application of momentum in the gradient update. The authors of Adam found that the exponentially weighted averages are biased, and correct this bias using a term that depends on the decay parameter of the exponential weight averages. Additionally, Adam applies an exponential weighted average to the gradient itself, which is equivalent to performing momentum on the gradient update. The update rule is:
\vspace*{1em}
\begin{align}
    s_n &= \rho_1 s_{n-1} + (1 - \rho_1) g_n\nonumber\\
    r_n &= \rho_2 r_{n-1} + (1 - \rho_2) g_n \odot g_n\nonumber\\
    \Theta_{n+1} &= \Theta_{n} - \alpha \frac{s_n}{\epsilon + \sqrt{r_n}} \frac{1 - \rho_2^n}{1 - \rho_1^n}
    \label{background:Adam}
\end{align}

Where $s_n$ is the biased estimate of the gradient and $r_n$ is the biased estimate of the squared gradients, both obtained with an exponential moving average with different decay rates $\rho_1$ and $\rho_2$. The factors $1 - \rho_1^n$ and $1 - \rho_2^n$ are used to correct bias in the exponential moving averages. These computations are done component-wise.

Overall Adam performs considerably better than RMSProp and AdaGrad, training models that converge faster and sometimes obtain slightly better predictive performance, but this is not always the case. Recent advances have shown that Adam can be improved as there are some theoretical issues \cite{reddi2018convergence} that have been fixed.
Adam is generally preferred by practitioners when training a newly designed model, as it requires less tuning of the learning rate.

The use of advanced Optimizer algorithms makes tuning the learning rate an easier task, and in some cases it allows the use of larger learning rates, which translates into faster convergence and lower training times. The only disadvantage of using Optimizers is that sometimes the learning process does not converge to the \textit{best} optimum\footnote{For example, it is well known in the community that using Adam on a VGG-like network fails to converge, and we have experimentally confirmed this, the loss just does not decrease, plain SGD works perfectly.}, and there are increased memory usage to store the exponentially weighted averages, usually by $100 \%$ for RMSProp and AdaGrad, and by $200 \%$ for Adam, which can constraint the kind of models that can be trained on a GPU.

\subsection{Performance Evaluation}

The idea of using machine learning models is to learn to generalize, that is, learn a model from a limited set of data that is useful for samples that the model has never seen during training. A useless model only performs well in the training set.

An \marginnote{Cross Validation} important part of designing a machine learning model is related to its desired performance. This is measured by the loss function during training, but just minimizing it in the training set does not guarantee any kind of performance in new and unseen samples. For this an additional set is needed, called the test set. A model is trained on the training set, and its performance evaluated on the test set, which provides an unbiased estimate of the generalization performance of the model. This is typically called Cross Validation.

Typically \marginnote{Train, Validation, and Test Splits} the available data is randomly split into three datasets: the Training set, the Validation set, and the Test set. The fractions for each split vary, but it is ideal to make the biggest split for the training set, at least 50 \% of the available data, and use the rest in equal splits for the validation and test sets.

The validation set is used to evaluate performance during hyper-parameter selection, and only after fixing these values, a final evaluation on the test set can be performed. This prevents any kind of bias in samples in the training or validation set from affecting conclusions about model performance.

Overfitting \marginnote{Overfitting} is the problem where the model learns unwanted patterns and/or noise from the training data, and fails to generalize outside of its training set. Detecting overfitting is key during training any machine learning model, and is the reason why validation or test sets are used, as it is the only known way to detect overfitting.

During the training process, the loss and metrics on the training set is typically tracked and displayed to the designer, and after each epoch, loss and associated metrics can be computed on the validation set. The overall pattern of both training and validation loss tells a picture about that is happening, we cover three cases:

\begin{description}
    \item[Training Loss Decreasing, Validation Loss Decreasing] Indicates that the model is learning and also generalizing well. This is the ideal case.
    \item[Training Loss Decreasing, Validation Loss Increasing] Indicates that the model is overfitting, as more noise is learned from the training set, which does not generalize to the validation set.
    \item[Training Loss Not Decreasing, Validation Loss Not Decreasing] The model is not overfitting, but it indicates that the model does not fit the data. A model with more learning capacity might be needed, as the current model cannot really predict the data given the input features. For example, if fitting a linear model to data with a quadratic shape. This case might also indicate that the input features might not be well correlated to the desired output or that the learning problem is ill-defined.
\end{description}

For classification problems, the loss is typically the cross-entropy and its variations, but humans prefer to evaluate performance with the accuracy \marginnote{Accuracy} metric, as it is directly interpretable:
\vspace*{1em}
\begin{equation}
    \text{ACC}(\hat{y}, y) = n^{-1} \sum \mathbb{1}[y = \hat{y}]
\end{equation}

Accuracy is just the fraction of samples that are correctly classified, that is, the predicted label $\hat{y}$ is the same as the ground truth label $y$. Note that the accuracy metric is completely meaningless for regression problems, as a continuous output equality is ill-defined. One way to define accuracy for continuous outputs is to consider equality if prediction differs from ground truth by not more than a given $\epsilon$, or to use pearson's correlation coefficient $R$, or the $R^2$ coefficient of determination.

\subsection{Hyper-parameter Tuning}

Hyper-parameters are all parameters in a model that need to be decided before starting the learning process, and that cannot be directly learned (as in by a learning algorithm) from data. The values of these parameters must be decided by the human designer. This also includes the neural network architecture.

A general way to tune these parameters is to use Grid Search \marginnote{Grid Search}. For this process, a range of values is defined for each parameter $P_i$, and each range is discretized to a finite set of values that will be tested. Then a grid is built by all possible combinations of parameter values $S = P_0 \times P_1 \times P_2 \times \cdots \times P_n$. For each value in the parameter space $S$, a model is trained and evaluated on the validation set. The set of parameters that produces the lowest validation loss is used to build the full model, but this is not the only possible criteria.
It is common that many sets of parameters provide the same or similar performance than the best model, so other criteria could be used, such as minimizing the number of total weights in the model, or maximizing computational performance subject to a given learning performance.

Grid Search is computationally expensive, as making the grid will exponentially explode the number of parameter sets that have to be tried, and training a neural network for each of these parameter sets is also computationally expensive. A common observation after performing grid search is that some parameters are not really important, having only a small or zero effect on learning performance.

For\marginnote[-2em]{Random Search} this reason, Random Search \cite{bergstra2012random} was proposed, where instead of using discrete values for the parameters, probability distributions are used to model each parameter, and during the search process, a parameter set is drawn by sampling each parameter distribution, and a model trained and evaluated.
Random Search has the advantage of minimizing the use of computational budget on uninteresting parameters, and it has been experimentally proven to obtain the same or slightly better models than Grid Search, with less computational budget. Note that since the exploration of the parameter space is random, the bias from parameter values set by the designer can potentially be reduced, and even work in other datasets for which random search is being performed.

Two parameters deserve special consideration due to their effect on the learning process: the learning rate (LR) and the number of training epochs.

\begin{description}
    \item[Learning Rate] This parameter controls the "speed" over which learning is performed, as it scales the gradient, which effectively makes it a kind of step size in the parameter space. Valid learning rate values are typically in the $[0, 1]$ range, but small values are mostly used in practice. If a large LR is used, then learning could diverge (producing infinite or NaN loss values), if a too small LR is used, learning happens but very slowly, taking a large number of epochs to converge. The \textit{right} LR value will produce fast learning, with a loss curve that is similar to exponential decay. Figure \ref{background:effectLR} shows typical loss curves with different learning rates. The case of a high LR shows that in the case where learning does not fail, but the loss decreases and stays approximately constant after a certain number of epochs. Note that the LR does not have to be a constant, and it can be varied during training. The typical method is to decrease the LR by a factor after a \textit{plateau} of the loss curve has been detected, which potentially could allow the loss to decrease further.
    
    \begin{marginfigure}
        \begin{tikzpicture}
            \begin{axis}[width = \textwidth, xlabel={Epochs}, ylabel={Loss},
            xmin = 0.0, xmax = 100.0, ymin = 0.0, ymax = 10.0,
            legend style={at={(0.5, 2.0)},anchor=north}]
                \addplot+[mark=none, domain=0:100] {10 - 0.09 * x + 0.5 * rnd};
                \addlegendentry{Low LR}
                \addplot+[mark=none, domain=0:100] {exp(2.5 - 0.1 * x) + 0.4 * rnd};
                \addlegendentry{Correct LR}
                \addplot+[mark=none, domain=0:100] {exp(2.0 - 0.05 * x) + 2 + 0.6 * rnd};
                \addlegendentry{High LR}
            \end{axis}
        \end{tikzpicture}
        \caption{Effect of Learning Rate on the Loss Curve during Training}
        \label{background:effectLR}
    \end{marginfigure}
    
    Learning rate can be tuned using grid or random search, but a faster way is to guess an initial LR, train different models and vary the learning rate manually, decreasing or increasing it accordingly to the previously mentioned rules. A common heuristic \cite{Goodfellow2016deep} is that if learning fails, decrease the LR by a factor of ten until the loss starts to decrease consistently, and adjust further by small steps to produce the ideal loss curve. Typical learning rates used in the literature are negative power of 10, like $\alpha = [0.1, 0.01, 0.001]$.
    
    \item[Number of Epochs] This parameter controls the length of the training process. If a model is trained for a short number of epochs, then it might not have converged, meaning that the loss could have continued to decrease if trained for longer, while training for more epochs than necessary risks overfitting, assuming no regularization was used.
    
    The number of training epochs can be tuned by experimenting manually in a similar way as the learning rate. The designer could set an initial number of epochs (say, 10) and then increase or decrease it accordingly if the loss shows no signs of convergence \footnote{A model has converged when the training process shows that the loss function can no longer decrease, and the loss shows a wiggling behavior, staying approximately constant.}, or if the model overfits.
    
    Related to this parameter is the early stopping criterion, \marginnote[0.5cm]{Early Stopping} where validation loss is monitored after each epoch, and training stopped if the validation loss starts to increase consistently after a tunable number of epochs. This prevents overfitting and would only require to tune a reasonable minimum number of epochs. The designer should also make sure to tune the learning rate appropriately, as the constant loss that indicates convergence could also be caused by a learning rate that is too high.
\end{description}

Note that both the value of learning rate and number of epochs depend on the actual loss function that is being minimized, any change to the loss implies retuning both parameters, as the actual loss surface or landscape is what defines the learning rate and length of training.

\section{Convolutional Neural Networks}

Convolutional Neural Networks (usually abbreviated CNNs or ConvNets) were introduced by Yann LeCun \cite{lecun1998gradient} in the 90's, initially for the task of handwritten object recognition, but they have been successfully applied to other vision tasks, like Object Detection and Localization, Semantic and Instance Segmentation, and Image Super-Resolution, etc. CNNs have revolutionized computer vision, as now many visual tasks can be learned using neural networks that are specifically designed for image processing.

This kind of networks are designed to process images as inputs and they have an optimized network structure to exploit three common properties of images:

\begin{description}
	\item[Local Statistics] In images the correlation between neighboring pixels in a region is higher than the correlation of far away pixels. This observation also has a biological counterpart as the receptive field in the visual cortex, where cells are excited by patterns in regions of the visual field.
	
	This property is implemented by a CNN by using neurons that are connected to a neighboring region in the input image. This is represented by a convolution filter or kernel, which can be square or rectangular. After convolution of the input image with a certain number of filters (one for each neuron), such layer produces the same number of output images called feature maps.
	
	\item[Translation Invariance] Generally the filters in that \textit{look} into a neighboring region of the input image do not depend on a spatial position in the image. Filters are generic and should be useful in any part of the image. This is one kind of translation invariance, since instead of connecting a neuron with all pixels of the input image, which increases the number of parameters, we can only learn the weights associated to the filter, and run it over the whole input image in a sliding window manner, which is effectively the convolution operation.
	
	Another kind of translation invariance is downsampling. When the network \textit{looks} for relevant features to detect or recognize objects, the convolution filter might produce high responses at several neighboring positions. One way to filter these high responses is to perform pooling on a feature map, which will only keep the high responses and discard irrelevant information, as well as reducing the dimensions of a feature map.
	This allows the network to concentrate on important features, and since pooling is invariant to small translations (inside the pooling region), this introduces a small degree of translation invariance into the network design.
	
	\item[Feature Extraction] In most computer vision tasks, features are used to discriminate relevant parts of an image. These features are manually engineered by researchers, which is a labor intensive task. Instead, a CNN can learn relevant features to the problem by means of learning the weights of convolution filters.
	This is a very important property of CNNs, features can automatically be learned directly from the training data, since the feature extraction and classifier modules are both part of the same neural network, which can be trained end-to-end. This means that relevant features for each vision problem can be learned directly from the data, without any manual engineering and with minimal preprocessing \footnote{In general this preprocessing consists of dataset augmentation and input/output normalization.}.
\end{description}

The basic design of CNNs introduces two new types of layers: Convolution and Pooling layers.

Images are usually represented as multi-dimensional arrays or tensors, \marginnote{Image Representation} with shapes $(W, H, C)$, where $W$ is the width, $H$ is the height, and $C$ is the channels dimension, \footnote{This is also referred as depth dimension in some papers.} which is one for a grayscale image, and three for a RGB color image. This representation also allows for arbitrary numbers of channels, and it will be useful later on.

\subsection{Convolutional Layers}

For a convolutional layer \marginnote{Convolutional Layer}, the output $y$ is given by:
\vspace*{1em}
\begin{equation}
    y = f(\mathbf{x} \ast \mathbf{F} + b)
\end{equation}

Where $\ast$ is the convolution operation, $x$ is the input image, $\mathbf{F}$ is the convolution filter (weights), $b$ is a bias and $f$ is a non-linear activation function that is applied element-wise to the output. In other words, a convolution layer takes an image as input, convolves it with a filter, adds a bias, and then passes the convolved image through a non-linear activation function. The output of a this kind of layer is called a \textit{convolutional feature map} \marginnote{Convolutional Feature Map}, as it represents visual features of an image (a map).

In practice, convolutional layers use more than one filter per layer, as this allows to learn different kinds of features at the same time, and later it allows the learning of feature hierarchies \cite{zeiler2014visualizing}. This is represented as a layer taking inputs of shape $(W, H, C)$, and the output having shape $(W, H, K)$, where $K$ is the number of filters in the convolution layer. Each filter is convolved individually with the input image, and the result after applying bias and activation function is stored in the channels dimension of the output, stacking all feature maps into a 3D volume. Convolutional layers can also take feature maps as inputs, which forms the feature hierarchy previously mentioned.

Another important detail of a convolutional layer is that both bias and weights on the filter are learned using gradient descent. They are not hand tuned as previously was done for image processing, like to make edge detection or sharpness filters. The filter in a convolutional layer is not necessarily a two dimensional matrix, as when the input image or feature map has $K > 1$ channels, then the filter must have a matching shape $(W, H, K)$, so convolution can be possible. When the filter has multiple channels, convolution is performed individually for each channel using the classical convolution operation from image processing \cite[-1em]{gonzalezDIP2006}.

The filter size (width and height) in a convolutional layer is a hyper-parameter that must be tuned for specific applications, and generally it must be a odd integer, typical values are $3 \times 3$ or $5 \times 5$, but some networks such as AlexNet \cite{krizhevsky2012imagenet} used filter sizes up to $11 \times 11$. The width and height of a filter do not have to be the same, but generally square filters are used.

Convolution is normally performed with a stride of 1 pixel \marginnote[1em]{Stride}, meaning that the convolution sliding window is moved by one pixel at a time, but different strides can also be used which is a kind of sub-sampling of the feature map.

The output dimensions of a convolutional layer are defined by the filter sizes, as convolution is only typically performed for pixels that lie inside the image region, and out of bound pixels are not considered (at the edges of the image). Padding \marginnote{Padding} can be added to the input image or feature map in order to output the same spatial dimensions as the input.

For a $N \times N$ filter and a $W \times W$ input image or feature map, with padding of $P$ pixels and stride $S$, the output has dimensions $O$:
\vspace*{1em}
\begin{equation}
    O = \frac{W - N + 2P}{S} + 1
\end{equation}

\subsection{Pooling Layers}

A pooling layer is used to introduce a small degree of translation invariance to the network design, and to control the flow of information through the network, by performing down-sampling on feature maps. This works by forcing the network during learning to produce meaningful features that will pass through the pooling operation. 

Pooling works by partitioning \marginnote{Down-sampling} an input feature map of size $W \times H$ in non-overlapping regions of the same size $D \times D$, and then performing a aggregation operation on each region, producing a scalar value, and then replacing each region with this aggregated scalar value, effectively performing down-sampling, as the output of the pooling operation has size $\frac{W}{D} \times \frac{H}{D}$. $W$ and $H$ must be divisible by $D$ or else padding is required. Pooling can be performed in overlapping regions, as well as with a stride $S > 1$, depending on the designer's needs.

Two types of aggregation operations are used in the CNN literature:

\begin{description}
    \item[Max-Pooling] The maximum value in each region $R$ is kept and used as output:
        \vspace*{1em}
        \begin{equation}
            y = \max_{x \in R} x 
        \end{equation}
        
        The concept of Max-Pooling is that only the maximum activation in each region of the feature map will pass, suppressing the non-maximums. While it works well in practice, there are issues since small disturbances to the activations in a feature map will lead to big changes after Max-Pooling. Passing gradients through this operation is not simple, as the position of the maximum has to be tracked for correct gradient propagation.
    \item[Average Pooling] Output the average value in each region $R$:
        \vspace*{1em}
        \begin{equation}
            y = D^{-2} \sum_{x \in R} x
        \end{equation}
        Taking the average of each region could be preferable to the maximum, as it has a less chaotic effect on the output when there are small disturbances, unlike max-pooling. This operation also has the effect of passing important information through, and it is more friendly to gradient propagation, as the average operation is differentiable.
\end{description}

Note that the pooling operation operates independently for each channel in a feature map, and the channels dimension is unaffected, as only spatial dimensions ($W$ and $H$) are down-sampled. Pooling defines the receptive field size of the network, as using more down-sampling operations will increase the receptive field exponentially.

Depending on the size of the input image (usually fixed at design time), there is a limited number of pooling operations with down-sampling that can be placed in a network. If a model contains pooling operations with down-sampling of $D \times D$ for an input of $W \times H$, then the maximum number of pooling operations is $\log_D \min \{ W, H\}$. Any down-sampling layer inserted above this limit will be operating on a $1 \times 1$ feature map, making the operation useless. This computation does not consider padding or slight down-sampling performed by convolution operations without padding.

Note that pooling operations have no trainable parameters, they are effectively computations that do not learn anything, but they influence what other trainable layers learn.

\subsection{Convolutional Network Architectures}

Now that we have defined the basic building blocks of convolution and pooling, we can define a full convolutional neural network.

A convolutional neural network (CNN) is any neural network that uses convolutional layers in its design, typically as the first layers in the architecture. The effect of using these layers is that the learn to extract relevant features from the input image. The combination of convolutional and pooling layers forms a natural feature hierarchy \cite{zeiler2014visualizing}, where low level features (edges, lines, etc) are extracted in the convolutional layers closest to the input, and more complicated features (object parts, ) are extracted in subsequent layers.

This feature hierarchy is a natural result of applying convolution and pooling over feature maps, as convolution over the input image can only extract very simple features, while convolution on a feature map that contains these simple features can then do further processing to extract more complex ones. As the network becomes deeper in terms of the number of convolutional layers, the complexity of features that can be modeled increases.

\begin{figure}
    \centering
    \includegraphics[width = 0.95 \textwidth]{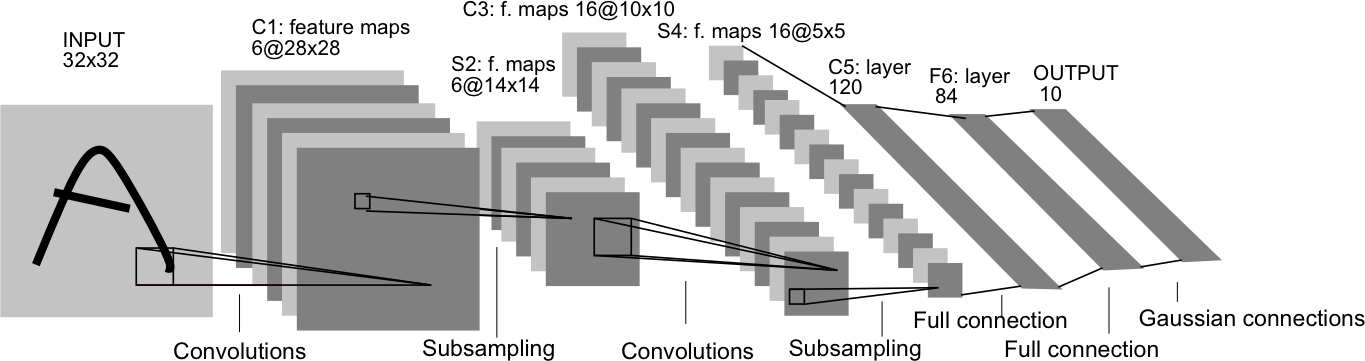}
    \caption[Architecture of LeNet-5]{Architecture of LeNet-5, Figure extracted from LeCun et al. 1998}
    \label{background:lenet5}
\end{figure}

Figure \ref{background:lenet5} shows LeNet-5 \cite{lecun1998gradient}, one of the first CNNs successfully used to recognize digits from the MNIST dataset. This network contains a first convolutional layer of six $5 \times 5$ filters, connected to a $2 \times 2$ pooling layer that subsamples by a weighted average and passes the output through a sigmoid activation function. Then another convolutional layer of sixteen $5 \times 5$ filters, also connected to a $2 \times 2$ max-pooling layer. The output of the last layer is then flattened\footnote{Array is reshaped to become one-dimensional.} and output to two fully connected layers (an MLP), that outputs to a softmax activation function.

LeNet was a big innovation for the time, since it obtains a $0.95 \%$ error rate on the MNIST dataset (corresponding to $99.05 \%$ accuracy), which is very close to human performance. Other kinds of classifiers such as K-Nearest-Neighbors with euclidean distance obtained $5 \%$ error rates, which shows the advantages of a CNNs.

LeNet set that initial standard for CNN design, starting with convolution and max-pooling blocks that are repeated a certain number of times, and perform feature extraction, followed by a couple of fully connected layers that perform classification or regression of those learned features. The network can be trained end-to-end using gradient descent.

A second milestone in CNNs is AlexNet \cite[-5em]{krizhevsky2012imagenet}, which is one of the first real deep neural networks trained on a large scale dataset. This network was designed to compete in the ImageNet Large Scale Visual Recognition Challenge \cite[-2em]{russakovsky2015imagenet}, where the task is to classify variable-sized images over 1000 different classes, with a training set containing 1.2 million images. It is a very difficult task due to the large training set, large number of classes, and many visual confusion between classes. \footnote[][2em]{For example, ImageNet contains many different races of dogs and cats, which a human cannot always visually distinguish.}

AlexNet unexpectedly won the ILSVRC competition in 2012, where most contenders were using classical computer vision methods and manually engineered features, but Kryzhevsky et al. showed that neural networks are competitive for this problem, and this is proven by the margin with the second place, of around $10 \%$ top-5 accuracy less than AlexNet.

The architecture of AlexNet is shown in Figure \ref{background:alexnet}, the network has 15 layers and approximately 60 million trainable parameters. AlexNet obtains $83.6$ \% top-5 accuracy on the ImageNet 2012 dataset, while the second place winner of the same competition obtains $73.8$ \% top-5 accuracy, showing the superior performance and capability of a deep neural network.

\begin{table}[t]
    \centering
    \begin{tabular}{@{}lll@{}}
        \hline
        Name					& Operation 							& Output Shape\\
        \hline
        Input					& Input image							& $(224, 224, 3)$ \\
        \hline
        Conv-1					& Conv($96$, $11 \times 11$, $S = 2$)	& $(55, 55, 96)$ \\
        MP-1					& Max-Pool($2 \times 2$)				& $(27, 27, 96)$ \\
        Conv-2					& Conv($256$, $5 \times 5$, $S = 1$)	& $(27, 27, 256)$ \\
        MP-2					& Max-Pool($2 \times 2$,)				& $(13, 13, 256)$ \\
        Conv-3					& Conv($384$, $3 \times 3$, $S = 1$)	& $(13, 13, 384)$ \\
        Conv-4					& Conv($384$, $3 \times 3$, $S = 1$)	& $(13, 13, 384)$ \\
        Conv-5					& Conv($256$, $3 \times 3$, $S = 1$)	& $(13, 13, 256)$ \\
        MP-3					& Max-Pool($2 \times 2$,)				& $(7, 7, 256)$ \\
        Flatten					& Flatten()								& $(7 \times 7 \times 256)$\\
        \hline
        FC-1					& FC($4096$, RELU)						& $(4096)$\\
        Dropout-1				& Dropout($0.5$)						& $(4096)$\\
        FC-2					& FC($4096$, ReLU)						& $(4096)$\\
        Dropout-2				& Dropout($0.5$)						& $(4096)$\\
        \hline
        FC-3					& FC($1000$, Softmax)					& $(1000)$\\
        \hline
    \end{tabular}
    \caption[Architecture of AlexNet as defined in Krizhevsky et al 2012]{Architecture of AlexNet as defined in Krizhevsky et al 2012. ReLU activations are used in each Convolutional Layer.}
    \label{background:alexnet}
\end{table}

Progress in the ImageNet competition has been constant over the years, producing advances in CNN architecture engineering. Pretty much all of the contenders after 2012 were using CNNs. In 2013 the VGG group at Oxford made a deeper version of AlexNet, which is typically just called VGG \cite{simonyan2014very}, with over 144 million parameters and obtaining $92 \%$ top-5 accuracy. The VGG networks use a simpler structure, with only $3 \times 3$ filters, and combining two consecutive convolutions both with $3 \times 3$ filters to simulate a bigger $5 \times 5$ filter.

In 2014 Google entered the competition with their GoogleNet \cite{szegedy2015going} architecture, which uses what they called the Inception module, that contains convolutions of multiple filter sizes combined with max pooling in a parallel structure, including $1 \times 1$ convolutions to learn features across channels/depth and $3 \times 3$ and $5 \times 5$ convolutions to learn spatial structure. GoogleNet obtains $93.3$ \% top-5 accuracy with 22 layers.

In 2015 Microsoft Research proposed Deep residual networks\cite{he2016deep}, which use a slightly different architecture that allows the network to be much more deep than possible before. A residual function is modeled as $F(x) + x$, where $x$ is the input to a set of layers, and $F(x)$ is the output of those layers. This addition operation is implemented as a skip connection that outputs the sum of the input and the output of a set of layers. The authors hypothesize that optimizing such structure is easier than the optimization process for a normal CNN, and they show this by building a much deeper (over 150) layer network that obtains $95.5$ \% top-5 accuracy.

\subsection{Discussion}

Different patterns in CNN architectures have emerged over the years, and overall there are some design choices that can be learned. In general a deeper network performs better, because it can learn higher level features which is only possible with a deep feature hierarchy. Seems that filters sizes do not have to be big (as in AlexNet), as most state of the art ImageNet CNNs use $3 \times 3$ and sometimes $5 \times 5$ filters. Even $1 \times 1$ filters are useful in order to influence information in the channels dimension.

A common pattern in the CNNs presented in this Section is that the number of filters increases with depth, and this makes sense, as deeper networks have smaller filter sizes and this has to be compensated by having more filters, which can represent a more rich feature hierarchy. There could be a one-to-one match between some features and specific filters, so in order to have more complex features, more filters are needed.

Overall, designing a neural network architecture for a specific task always requires a degree of experimentation. Designers should start with a small network, and expand it as needed, but only doing this kind of experimentation by evaluating on a evaluation set, and obtaining a final performance measure in a test set.

Neural network architectures can also be automatically designed by an algorithm. Popular choices are genetic algorithms \cite[-4em]{stanley2002evolving}, and newer techniques like Neural Architecture Search \cite[1em]{zoph2018learning} and Differentiable Architecture Search \cite[1em]{liu2018darts}.

In general automatic architecture design techniques have trouble achieving results that outperform the state of the art, as measured by classification performance in many common datasets (like ImageNet, CIFAR-10/100), only recent techniques are able to automatically produce an architecture that outperforms manually crafted architectures, so it can be expected that neural networks will be increasingly designed by algorithms and not by humans.

These kind of techniques are contributing to the long term goal of \textit{automatic machine learning}\cite[1em]{quanming2018taking}, where only labeled data is provided and the hyper-parameters of the architecture are automatically tuned in a validation subset, which has the potential of expanding the use of machine learning techniques to non-technical users.

%% file: chapters/sonar-image-classification.tex
\chapter[Forward-Looking Sonar Image Classification]{Forward-Looking Sonar \newline Image Classification}
\label{chapter:sonar-classification}

Image classification is one of the fundamental problems in Computer Vision and Robot Perception. It is also one of the most basic problems, as many techniques have image classification sub-tasks.

The Image Classification is the task given an image, classify it as one of $C$ predefined classes. This task is commonly implemented with ML algorithms, as it directly maps to the training of a classifier on image inputs.

In this chapter we perform a comprehensive evaluation of image classification algorithms on FLS images, including classic techniques like template matching as well as advanced classification algorithms. We wish to investigate the following research questions:

\begin{itemize}
	\item Can we use feature learning methods on FLS images?
	\item Can object recognition be performed with high accuracy and low number of assumptions on object shape, shadow or highlight?
    \item Can we use deep neural networks to classify FLS images on embedded systems?
\end{itemize}

\section{Related Work}

Image classification methods are typically embedded into a bigger system, such as object detection or automatic target recognition systems. In this literature review we focus on classification as a separate problem, as the bottleneck for a good detection or recognition system is usually classifier performance. Without a good classification algorithm, high performance on object detection or target recognition is not possible.

The most basic classification method for Sonar images is to use template matching, which is a specific way of computing similarity between two images. If a labeled set of template images\footnote{This is just a set of same-size images where each is labeled with a corresponding class.} is available, then a given test image can be classified by computing the similarity with each template image and finding the "most similar" image, and outputting the class label of that image. This method is similar to a k-nearest neighbours (KNN) classifier with $k = 1$ and using the similarity as a feature space.

Two similarity computation techniques are popular \cite{gonzalezDIP2006}, namely normalized cross-correlation and sum of squared differences.

Cross-correlation, computes the correlation $\sum I \star T$ between an image $I$ and template image $T$, where $\star$ is component-wise multiplication. As the correlation is unbounded, a normalized version is preferred. $I$ and $T$ are normalized by subtracting the mean and dividing by the standard deviation, which leads to the operation in Eq \ref{sic:ccSimilarityEq}:
\vspace*{1em}
\begin{equation}
	S_{CC}(T, I) = \frac{\sum (I - \bar{I}) \sum (T - \bar{T})}{\sqrt{\sum (I - \bar{I})^2 \sum (T - \bar{T})^2}}
	\label{sic:ccSimilarityEq}
\end{equation}

The normalized cross-correlation produces a number in the $[-1, 1]$ range. Note that this version of cross-correlation is slightly different from the one used to perform object detection, as the template is not slid over the image, as the test and template images are assumed to be the same size. In order to perform classification, the template with highest normalized cross-correlation is selected.

Sum of squared differences similarity (SQD) just computes the average square of the difference between the test image $I$ and the template image $T$ (Eq \ref{sic:sqdSimilarityEq}). The idea of SQD is to find the template that is most similar to the test image in terms of raw euclidean distance between pixels.
\vspace*{1em}
\begin{equation}
	S_{SQD}(T, I) = n^{-1} \sum (I - T)^2
	\label{sic:sqdSimilarityEq}
\end{equation}

Reed et al. \cite{reed2004automated} propose a shape matching approach for image classification. A training dataset is generated by using CAD models to generate synthetic sidescan sonar images for matching. A test image is classified by first detecting mine-like objects (Cylinder, Sphere and Cone) with a Markov Random Field model and the shadow and highlight are segmented. The shadow shape is then matched with one in the training set by means of the minimum Hausdorff distance.

As one match is produced per class, the authors combine a match from each of the three classes using Fuzzy Logic. This method is quite complex as multiple manually engineered equations are required, with little theory to support them.

In order to improve the robustness of their results, Reed et al. \cite[-7em]{reed2004automated} use Dempster-Shafer information theory to fuse predictions from multiple views. In single-view classification, their method obtains $90 \%$ correct classifications with $50 \%$ false alarms produced by the detection/segmentation step of their method.
In multi-view classification, a Cylinder can be classified with $93.9 \%$ accuracy, while a Cone produces $67.5 \%$ accuracy, and a Sphere with $98.3 \%$ accuracy.
The results are not impressive for such simple shaped objects, as multiple views are required to obtain a high accuracy classifier, but still the Cone object is poorly classified.

Fawcett et al. \cite{fawcett2007computer} use engineered features based on shadow and highlight shape to classify mine-like objects (Cylinder, Cone, Truncated Cone) in SAS images. Shadow/highlight segmentation is required. 38 different features are used for shadows, while 57 features are used for highlights. The authors only provide vague information about the features they used, namely "profiles", perimeter, area, pixel statistics of the regions. A  Radial Basis Function classifier with a Gaussian Kernel

Additional features are computed with Principal Component Analysis (PCA), by projecting each image into the first 5 principal components, which produces an additional 5-dimensional feature vector. Different combination of features are tested. $90 \%$ classification accuracy is obtained with shadow, highlight, and PCA features. Other combinations of features performed slightly worse, but notably shadow-only features obtained considerably worse accuracy (around $40 \%$). An interesting observation made in this work is that using the normalized image pixels as a feature vector obtains almost the same classification performance as the more complex set of features.

Myers and Fawcett \cite{myers2010template} proposed the Normalized Shadow-Echo Matching (NSEM) method for object detection in SAS images. NSEM first requires to segment the sonar image into bright echo, echo, background, shadow and dark shadow. Fixed values are assigned to each segmentation class (in the range $[-1, 1]$).
Then similarity is computed with a custom correlation function shown in Eq. \ref{sic:myersFawcettSimilarity}, where $I$ and $T$ are the segmented and post-processed test and template images, $I \otimes T = \sum T \star I$ is the standard cross-correlation operator, and $I_E$/$T_E$ are the highlight components of the corresponding images, and $I_S$/$T_S$ are the shadow components. The bar operation inverts the image, setting any non-zero element to zero, and any zero value to one.
\vspace*{1em}
\begin{equation}
	f(T, I) = \frac{I \otimes T_E}{1 + I_E \otimes \bar{T}_E} + \frac{I \otimes T_S}{1 + I_S \otimes \bar{T}_S}
	\label{sic:myersFawcettSimilarity}
\end{equation}

The final classification is performed by outputting the class of the template with the highest similarity score as given by Eq. \ref{sic:myersFawcettSimilarity}. This method can also be used as an object detector by setting a minimum similarity threshold to declare a detection.

This method was tested on a 3-class dataset of MLOs, namely Cylinder, Cone, Truncated Cone, and Wedge shapes. Target templates were generated using a SAS simulator, adding multiple views by rotating the objects. Objects are correctly classified with accuracies in the range $97-92 \%$, which is slightly higher than the reported baseline using normalized cross-correlation with the raw image ($92-62 \%$) and segmented images ($95-81 \%$). This method performs quite well as reported, but it requires a segmentation of the input image, which limits its applicability to marine debris.

Sawas et al. \cite{sawas2010cascade} \cite[1em]{sawas2012cascade} propose boosted cascades of classifiers for object detection in SAS images. This method can also be used for classification, as its core technique (Adaboost) is a well-known ML classification framework . Haar features are quite similar to the shadow-highlight segmentation present in typical sonar images, which is that motivates their use. Haar features have the additional advantage that they can be efficiently computed using summed-area tables.

A boosted cascade of classifiers \cite[1em]{bishop2006pattern} is a set of weak classifiers \footnote[][1em]{Classifiers with low accuracy, but computationally inexpensive} that are stacked in a cascade fashion. The basic idea originally introduced by Viola-Jones \cite[2em]{viola2001rapid} for face detection is that weak classifiers at the beginning of the cascade can be used to quickly reject non-face windows, while classifiers close to the end of the cascade can concentrate on more specific features for face detection. AdaBoost is then used to jointly train these classifiers.
This structure produces a very efficient algorithm, as the amount of computation varies with each stage and depth in the cascade (deep classifiers can use more features, while shallow classifiers can use less), but fast to compute features are required, as feature selection is performed during training.

Sawas et al. proposes an extension to the classic Haar feature, where a long shadow area with a small highlight is used as a Haar feature, matching the signature from the MLOs that are used as targets. On a synthetic dataset with Manta, Rockan and Cylinder objects, the authors obtain close to $100 \%$ accuracy for the Manta, but Rockan and Cylinder only obtain around $80-90 \%$ accuracy.
On a semi-synthetic dataset with the same objects, results are slightly different, with Manta saturating at $97 \%$ accuracy, while the other objects obtain $93-92 \%$ accuracy. When used as an object detector, this method is very fast but it suffers from a large amount of false positives. The use of Haar features also makes it quite specific to MLOs, as their unique shadow-highlight pattern is not produced by marine debris.

Dura et al. \cite{dura2011image} surveys techniques for detection and classification of MLOs in sidescan sonar images. She describes two principal types of features for MLO classification: shape and gray-level.

Shape features consist of geometric characteristics from shadow or highlight regions. It is mentioned that Man-made objects project regular and predictable shadow and highlight regions. Shape features include: area, elongation, circularity, orientation, eccentricity, rectangularity, number of zero crossings of the curvature function at multiple scales. Many of these features are computed as functions of the central moments of order $p + q$:
\vspace*{1em}
\begin{equation}
	\mu_{pq} = \sum_x \sum_y (x - x_g)^p (y - y_g)^q I(x, y)
\end{equation}

Where $(x_g, y_g)$ is the centre of mass of the region and $I(x, y)$ is the region pixels. 

Gray-level features are statistics of the shadow and highlight regions, and are useful for low resolution images. Such features consider: shadow/highlight mean and variance, ratios between shadow and highlight, shadow vs background, and highlight vs background. These ratios are typically computed from means across each kind of region. While only a qualitative comparison is provided, the only quantitative results correspond to the ones published in each work.

All of these features require segmentation between shadow, highlight and background. This is a considerable disadvantage, as marine debris does not always possess a shadow, due to the small object size.
This work also briefly compares single and multi-view classification. Using multiple views usually reduces classification uncertainty, but this process introduces additional complexity due to the choice of fusion algorithm and more complex features.

Fandos and Zoubir \cite[-2em]{fandos2011optimal} provides a numerical comparison of shape and statistical shadow/highlight features on SAS images. Cylindrical, Spherical and Background objects were considered. Normalized central moments obtain $90-80 \%$ accuracy, while PCA features with three principal components obtains close to $80 \%$ accuracy.

Statistical features on shadow and highlight saturate to $70 \%$ accuracy across many combinations. Brute force combination of all features produces overfitting due to the curse of dimensionality. An empirical obtained optimal feature vector formed by shadow and highlight shape features, two normalized central moments, the second principal component, the shadow-background/shadow-highlight/highlight-background scale Weybull parameters, and the segmentation quality parameter.
This feature vector produces $95 \%$ classification accuracy, which is the best result in this work. But such a complicated feature vector does not necessarily generalize to other objects, specially when considering the variability of marine debris shape.

Fandos et al. \cite{fandos2012sparse} compare the use of sparse representations versus a Linear Discriminant Analysis (LDA) classifier. Features consist of the raw SAS image, a segmented SAS image, the fourier coefficients of the shadow, 50 principal components obtained from PCA applied to the shadow, and 10-th order normalized central moments of the segmented shadow. Feature selection was performed with sequential forward floating search and a optimal feature set was obtained, it contains a subset of the previously mentioned features.

$95 \%$ classification accuracy can be obtained on the optimal feature set, but the same features after a sparse representation has been obtained perform considerably worse, with differences up to $25 \%$ less absolute accuracy. As mentioned in the similar previous work, it is not clear how these features would generalize to different objects, and these features seem too specific to MLOs.

Hurtos et al. \cite{hurtos2013automatic} uses normalized cross-correlation template matching to detect the four corners of a chain link in a FLS image (from an ARIS Explorer 3000). This application is the one that can be considered closest to this work, as detecting and classifying chain link corners is a considerably more hard problem than MLO classification.
Templates are selected by cropping images at each of the four corners of the chain link, and rotations are introduced in order to generate a bigger template set. Results over three datasets shows classification performance in the range $92-84 \%$ accuracy. While this result is appropriate for the chain following task, it is quite weak from a classification point of view, as template matching does not seem to produce robust results. Authors of this work had to use mosaicing by averaging three frames in order to reduce noise in the image, which shows that using the raw FLS image with template matching could potentially reduce classification performance.

In the same line as the previous work, Ferreira et al. \cite{ferreira2014improving} also use mosaicing to improve classification performance in FLS images (obtained from a BlueView P450-130). A classification algorithm using morphological and echo/background ratios is used. Using this classifier with raw data produces $40 \%$ correct classifications, while using mosaiced images produces $38.4 \$$ accuracy, but the number of misclassifications is reduced.

Barngrover et al. \cite{barngrover2015semisynthetic} evaluate the effect of using semi-synthetic data versus real data. Their motivation is to use ML classification algorithms that require several thousands of data points in order to generalize well. They generate semi-synthetic data by segmenting the object out of a real image and placing it in new different backgrounds.

The authors use a boosted cascade of weak classifier (same as Sawas et al.) with Haar and Local Binary Pattern (LBP) features. For Haar features, classifiers that were trained on real data have an accuracy advantage of around $3 \%$ over classifiers that were trained on semi-synthetic data, but the classifiers on real data saturate at $93-92 \%$ correct classification.
For LBP features, the difference is more drastic, as using a classifier trained on real data has a $20 \%$ accuracy advantage over using synthetic data. A classifier trained on real data obtains close to $90 \%$ accuracy, while synthetic classifiers obtain $70 \%$. Some specific configurations of semi-synthetic data generation can improve accuracy to the point of a $6 \%$ difference versus the real classifier.

David Williams \cite{williams2016underwater} performs target classification in SAS images using a CNN with sigmoid activations. His dataset contains MLO-like objects for the positive class (cylinders, wedges, and truncated cones), while the negative class contain distractor objects like rocks, a washing machine, a diving bottle, and a weighted duffel bag, with a training set of over 65K images. Three binary classification experiments are performed, namely fully positive vs negative class, truncated cones vs rocks, and MLO mantas vs truncated cones as they are visually similar. The ROC curve and the area under the curve (AUC) are used as  evaluation metrics. In all experiments a 10-layer convolutional neural network obtained better results than the baseline (a relevance vector machine).

David Williams \cite[-1em]{williams2018underwater} has also explored transfer learning for SAS images. Using the previously mentioned networks trained on SAS data, each network was fine-tuned on new data of cylindrical objects versus clutter, and evaluated on a previously unseen dataset, showing that the fine-tuning process improves classification performance in terms of the AUC. A more interesting results is for transfer learning across different sonar sensors, where a CNN is trained on one SAS sensor, and fine-tuned in another. The fine-tuning process indeed shows improvements in AUC, even as less samples per class are available.
This work does not use a robust method to decide the network architecture, as a model with less parameters could perform better. No baselines are provided, which makes interpreting the results difficult.

Zhu et al. \cite{zhu2017deeplearning} uses AlexNet pre-trained on the ImageNet dataset to extract features from sidescan sonar images, and use them for object classification with an SVM. CNN features obtain up to $95 \%$ accuracy, outperforming Local Binary Patterns and Histogram of Oriented Gradient features when used with an SVM classifier. The dataset contains 35 sonar images, containing objects of interest over binary classes (targets/no targets), but no more detail about object classes is provided.

K\"ohntopp et al. \cite{kohntopp2017seafloor} provide results for seafloor classification in SAS images. Their dataset contains three seafloor classes: flat, rocky and ripples. They compare different handcrafted features, namely fractal dimension, wavelets, anisotropy and complexity, texture features, lacunarity, and a combination of these features, versus a Convolutional Neural Network that they designed. Two classic classifiers are evaluated, including a SVM and Naive Bayes.

The CNN outperforms all other handcrafted features with both classifiers, obtaining an accuracy of $98.7 \%$. Handcrafted features' performance varies, with the best being the fractal dimension with an SVM classifier ($97.7 \%$) and wavelets features ($94.9 \%$). A CNN outperforms an SVM by $1 \%$.
There is still many ways to improve these results, like including additional regularization (Batch Normalization) and fine-tuning the network architecture, but these results show that using CNNs on sonar data is promising.

Buss et al. \cite{buss2018hand} compare hand-crafted features from a feed-forward neural network versus feature learning by a convolutional neural network on data from a Cerberus active diver detection sonar developed by Atlas Electronics. Their dataset contains 490 samples of divers, and 114K non-diver or background samples. Surprisingly, the feed-forward neural network using hand-crafted features outperforms both a shallow CNN and VGG, obtaining a higher area under the ROC curve ($0.99 - 0.93$) and reducing false positives by $89 \%$. This result can be explained by the large imbalance between targets and non-targets, and additional domain knowledge introduced by the feature engineering process.

\subsection{Discussion}

Most applications of object recognition/classification in underwater domains correspond to mine-like objects. While it is questionable that large parts of the literature in this topic are devoted to pseudo-military research, there are better arguments about why this is a big problem.

Mine-like objects have simple geometrical shapes, namely Spheres, Cylinders, Cones, Truncated Cones, etc. This simplifies the development of recognition algorithms but these simplifications are not free, they usually make implicit assumptions. For example, using template matching can be done with templates that match certain views of the object, and an assumption is made that other views are not important for generalization. Enumerating all views of an object is not always possible.

Datasets used for MLO detection and classification are typically collected by military organizations and such data is most of the time classified\footnote{Information that is restricted from public access because of confidentiality, typically regulared by law.}. This hinders the development of the field, as only people in military research centers or with security clearance access can effectively perform experiments and advance the field. This is in contrast with the "open" policy of the Machine Learning community, where Datasets and Source Code are routinely shared and used as standard benchmarks. Machine Learning moves at a faster pace precisely because of their openness.

Moving into specific issues with classification algorithms for sonar images. All methods based on template matching suffer from severe disadvantages.

Using templates make implicit assumptions on object shape. Given a fixed set of templates, only a finite number of variations of the object can be modeled. Deciding which templates to use in order to detect/classify a set of objects is an open problem. For mine-like objects, typical choices are several views of each object that are synthetically generated, which alleviates the problem.

Each template's pixels could be considered as parameters in a ML classification algorithm. Using a large number of templates to classify an object could potentially be subject to overfitting, as model capacity increases and there is a higher chance than one of the templates produces a misclassified match.

Many methods require a segmentation of the image/template into shadow/highlight/background. From a ML point of view, this can be considered as regularization, as the number of free parameters is reduced by constraining them. For example, setting background pixels in segmented templates to zero reduces the effect of background. Segmenting sonar images is not trivial and it is a cumbersome part of the process, as it adds additional complexity that might not be required.

If the template and the object in a test image do not align, then the similarity will be low and classification will fail. To alleviate this problem it is typical to perform data augmentation by translating templates, introducing a small degree of translation invariance. But as mentioned before, increasing the number of templates could potentially lead to overfitting.

As a general overview of template matching accuracy, it is notable that most reviewed methods do not obtain accuracies higher than $95 \%$, with only some selected method obtaining higher accuracies, but never approaching $99 \%$. This pattern also supports our argument of mild overfitting, as the methods do not seem to generalize well and their accuracies quickly saturate.

About engineered features, they also suffer from some drawbacks. Many handcrafted features have no interpretation. For example, the fractal dimension, while having a physical meaning, does not necessarily have an interpretable meaning as a feature for image classification. Same can be said about Fourier or Wavelet features, as they are high dimensional and a learning algorithm can learn to overfit them.

Engineered features also suffer from generalization problems. While shadow and highlight geometric features seem to work for simple shaped objects like MLOs, they might fail to generalize over different objects, specially if their shapes are more complex. There is no reason that features that work well for MLOs will work well for marine debris.

It is well known that feature learning generally outperforms most kinds of feature engineering \cite{sharif2014cnn}. While it is not in question if learning features will outperform most engineered features, it is unknown by how much and how applicable is feature learning for the purpose of marine debris classification.

In the context of marine debris, we have observed that in our datasets, most marine debris-like objects have considerable differences with underwater mines that are relevant for classification:

\begin{itemize}
	\item In some cases, shape is not predictable. For example a plastic bottle might be crushed or deformed and his shape would be completely different than expected. In contrast, underwater mines are not typically deformed and they are designed to sustain water column pressure.
	\item MLOs have mostly convex shapes, while marine debris can have shapes with concave parts. This has the effect of producing strong reflections in the sonar image, and consequentially a much stronger viewpoint dependence. Simply, objects look quite different in the sonar image if you rotate them.
	\item Since the shape of MLOs is predictable and with a low intra/inter-class variation, the acoustic shadows that they produce are also predictable, and many methods exploit this property in order to do classification and detection. Marine debris does not have a predictable shadow shape as the object can be in any pose.
    \item Marine debris is usually much more small in physical size than MLOs, and in some cases marine debris does not produce acoustic shadows, which implies that only highlight information has to be used for classification.
\end{itemize}

\section{Classification using Convolutional Neural Networks}

In this section we describe our approach to classify FLS images with a Convolutional Neural Network. As most neural network architectures are designed for color images, we are forced to design our own architectures to fit the complexity and requirements of sonar data.

There are multiple hyper-parameters that must be tuned in order to obtain a proper neural network architecture. We take a simplified approach. We design several basic neural network modules that are "stacked" in order to build a more complex network. We let all modules inside a network to share the same set of hyper-parameters, and introduce a depth hyper-parameter, which is the the number of stacked modules on the network.

\begin{figure*}[t]
	\centering
	\subfloat[Classic Module]{
		\begin{tikzpicture}[style={align=center, minimum width=2.2cm}]
		\node (E) {\scriptsize Input};
		\node[draw, above=1em of E] (D) {\scriptsize Conv2D($f$, $5 \times 5$)};
		\node[draw, above=1em of D] (C) {\scriptsize Max-Pool($2 \times 2$)};
		\node[draw, above=1em of C] (B) {\scriptsize Batch Norm};
		\node[above=1em of B] (A) {\scriptsize Output};
		\draw[-latex] (B) -- (A);
		\draw[-latex] (C) -- (B);
		\draw[-latex] (D) -- (C);
		\draw[-latex] (E) -- (D);
		\end{tikzpicture}
	}
	\subfloat[Fire Module]{
		\begin{tikzpicture}[style={align=center, minimum height=0.4cm, minimum width = 1.5cm}]
		
		\node[] (dummy) {};
		\node[draw, below=1em of dummy](B) {{\scriptsize Conv2D($s_{11}$, $1 \times 1$)}};
		\node[draw, left=0.25em of dummy] (C) {{\scriptsize Conv2D($e_{33}$, $3 \times 3$)}};
		\node[draw, right=0.25em of dummy] (D) {{\scriptsize Conv2D($s_{33}$, $1 \times 1$)}};
		\node[draw, above=1em of dummy](CONC) {{\scriptsize Merge}};
		\node[draw, above=1em of CONC](BN) {{\scriptsize Batch Norm}};
		\node[above=1em of BN] (O) {\scriptsize Output};
		\node[below=1em of B] (I) {\scriptsize Input};
		\draw[-latex] (B) -- (C);
		\draw[-latex] (C) -- (CONC);
		\draw[-latex] (B) -- (D);
		\draw[-latex] (D) -- (CONC);
		\draw[-latex] (I) -- (B);
		\draw[-latex] (BN) -- (O);
		\draw[-latex] (CONC) -- (BN);
		\end{tikzpicture}
	}	
	\subfloat[Tiny Module]{
		\begin{tikzpicture}[style={align=center, minimum height=0.4cm, minimum width = 2.3cm}]
		\node (I) {\scriptsize Input};
		\node[draw, above=1em of I] (A) {{\scriptsize Conv2D($f$, $3 \times 3$)}};
		\node[draw, above=1em of A] (B) {{\scriptsize Conv2D($f$, $1 \times 1$)}};
		\node[draw, above=1em of B](BN) {{\scriptsize Batch Norm}};
		\node[draw, above=1em of BN] (C) {{\scriptsize Max-Pool($2 \times 2$)}};
		\node[above=1em of C] (O) {\scriptsize Output};
		\draw[-latex] (A) -- (B);
		\draw[-latex] (C) -- (O);
		\draw[-latex] (I) -- (A);
		\draw[-latex] (B) -- (BN);
		\draw[-latex] (BN) -- (C);
		\end{tikzpicture}
	}
	\subfloat[MaxFire Module]{
		\begin{tikzpicture}[style={align=center, minimum height=0.4cm, minimum width = 2.0cm}]
		\node (I) {\scriptsize Input};
		\node[draw, above=1em of I] (A) {{\scriptsize Fire($s, e, e$)}};
		\node[draw, above=1em of A] (B) {{\scriptsize Fire($s, e, e$)}};
		\node[draw, above=1em of B](BN) {{\scriptsize Batch Norm}};
		\node[draw, above=1em of BN] (C) {{\scriptsize Max-Pool($2 \times 2$)}};
		\node[above=1em of C] (O) {\scriptsize Output};
		\draw[-latex] (A) -- (B);
		\draw[-latex] (C) -- (O);
		\draw[-latex] (I) -- (A);
		\draw[-latex] (B) -- (BN);
		\draw[-latex] (BN) -- (C);
		\end{tikzpicture}
	}
	\vspace*{0.2cm}
	\caption[Basic Convolutional Modules]{Basic Convolutional Modules that are used in this Chapter}
	\label{sic:basicModules}
\end{figure*}
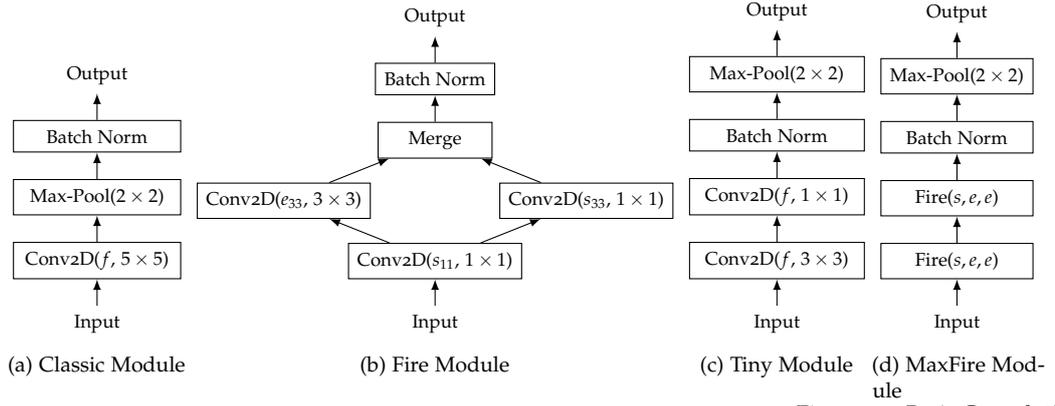

Modules that we use are shown in Fig. \ref{sic:basicModules}. We now describe the modules and their hyper-parameters:

\begin{description}
	\item[Classic Module] \hfill \\
		This is the most common module use by CNNs, starting from LeNet \cite[-4em]{lecun1998gradient}. It consists of one convolution layer followed by a max-pooling layer. Hyper-parameters are the number of filters $f$ and the size of the convolutional filters $s$. In this work we set $s = 5 \times 5$. This module is shown in Figure \ref{sic:basicModules}a.
		
	\item[Fire Module] \hfill \\
		The Fire module was introduced by Iandola et al. \cite{iandola2016squeezenet} as part of SqueezeNet.The basic idea of the Fire module is to use $1 \times 1$ convolutions to reduce the number of channels and $3 \times 3$ convolutions to capture spatial features. This module is shown in Figure \ref{sic:basicModules}b. The initial $1 \times 1$ convolution is used to "squeeze" the number of input channels, while the following $1 \times 1$ and $3 \times 3$ convolutions "expand" the number of channels.
		A Fire module has three hyper-parameters, the number of squeeze filters $s_{11}$, the number of expand $1 \times 1$ filters $e_{11}$, and the number of expand $3 \times 3$ filters $e_{33}$.
		
	\item[Tiny Module] \hfill \\
		The Tiny module was designed as part of this thesis. It is a modification of the Fire module, removing the expand $1 \times 1$ convolution and adding $2 \times 2$ Max-Pooling into the module itself. The basic idea of these modifications is that by aggressively using Max-Pooling in a network, smaller feature maps require less computation, making a network that is more computationally efficient. This module is shown in Figure \ref{sic:basicModules}c.
		The Tiny module has one hyper-parameter, the number of convolutional filters $f$, which is shared for both $1 \times 1$ and $3 \times 3$ convolutions.
		
	\item[MaxFire Module] \hfill \\
	This is a variation of the Fire module that includes two Fire modules with the same hyper-parameters and one $2 \times 2$ Max-Pooling inside the module. It is shown in Figure \ref{sic:basicModules}d and has the same hyper-parameters as a Fire module.
\end{description}
\vspace*{-1em}

All modules in Figure \ref{sic:basicModules} use ReLU as activation. We designed four kinds of neural networks, each matching a kind of module. Networks are denominated as ClassicNet, TinyNet and FireNet. While FireNet is quite similar to SqueezeNet, we did not want to use that name as it refers to a specific network architecture that uses the Fire module. Our FireNet uses the MaxFire module instead.

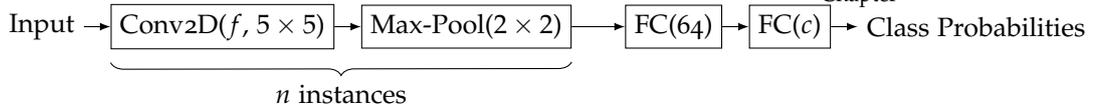
\begin{figure*}[tb]
    \vspace*{0.5cm}
	\centering
	\begin{tikzpicture}[style={align=center, minimum width=1cm}]
	\node (I) {Input};
	\node[draw, right=1em of I] (conv2d1) {Conv2D($f$, $5 \times 5$)};
	\node[draw, right=1em of conv2d1] (mp1) {Max-Pool($2 \times 2$)};
	\node[draw, right=2em of mp1] (fc1) {FC(64)};
	\node[draw, right=1em of fc1] (fc2) {FC($c$)};
	\node[right = 1 em of fc2] (O) {Class Probabilities};
	\draw[decoration = {brace, mirror, raise=0.5em, amplitude=0.5em}, decorate]	(conv2d1.south west) -- node[below=1em of mp1] {$n$ instances} (mp1.south east);
	\draw[-latex] (I) -- (conv2d1);
	\draw[-latex] (conv2d1) -- (mp1);
	\draw[-latex] (mp1) -- (fc1);
	\draw[-latex] (fc1) -- (fc2);
	\draw[-latex] (fc2) -- (O);
	\end{tikzpicture}
	\caption{ClassicNet Network Architecture}
	\label{sic:classicNet}
\end{figure*}

To build ClassicNet, we stack $N$ Classic modules and add two fully connected layers as classifiers. This corresponds to a configuration FC(64)-FC(C), where $C$ is the number of classes. The first fully connected layer uses a ReLU activation, while the second uses a softmax in order to produce class probabilities. This architecture can be seen in Figure \ref{sic:classicNet}.

\begin{figure}[t]    
	\centering
	\begin{tikzpicture}[style={align=center, minimum width=1cm}]
	\node (I) {Input};
	\node[draw, right=1em of I] (fire) {Tiny($f$)};
	\node[draw, right=1em of fire] (conv2d2) {Conv2D($c$, $1 \times 1$)};
	\node[draw, right=1em of conv2d2] (gap) {AvgPool()};
	\node[draw, right=1em of gap] (out) {Softmax()};
	\node[below = 1 em of out] (O) {Class Probabilities};
	\draw[-latex] (I) -- (fire);
	\draw[-latex] (fire) -- (conv2d2);
	\draw[-latex] (conv2d2) -- (gap);
	\draw[-latex] (gap) -- (out);
	\draw[-latex] (out) -- (O);
	\draw[decoration = {brace, mirror, raise=0.5em, amplitude=0.5em}, decorate]	(fire.south west) -- node[below=5mm] {$n$ instances} (fire.south east);
	\end{tikzpicture}
	\caption[TinyNet Network Architecture, based on the Tiny module]{TinyNet Network Architecture, based on the Tiny module. All layers use ReLU activation}
	\label{sic:tinyNet}
\end{figure}
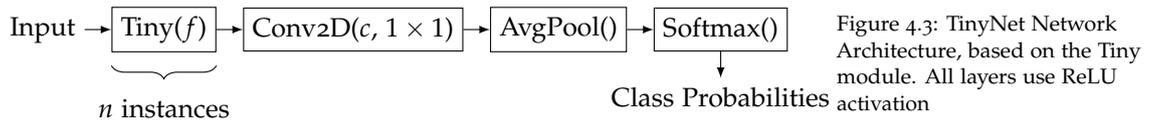

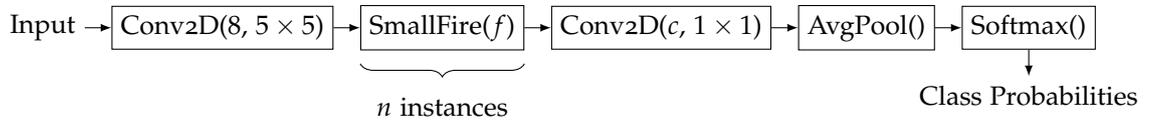
\begin{figure*}[t]
	\centering
	\begin{tikzpicture}[style={align=center, minimum width=1cm}]
	\node (I) {Input};
	\node[draw, right=1em of I] (startConv) {Conv2D($8$, $5 \times 5$)};
	\node[draw, right=1em of startConv] (fire) {SmallFire($f$)};
	\node[draw, right=1em of fire] (endConv) {Conv2D($c$, $1 \times 1$)};
	\node[draw, right=1em of endConv] (gap) {AvgPool()};
	\node[draw, right=1em of gap] (out) {Softmax()};
	\node[below = 1 em of out] (O) {Class Probabilities};
	\draw[-latex] (I) -- (startConv);
	\draw[-latex] (startConv) -- (fire);
	\draw[-latex] (fire) -- (endConv);
	\draw[-latex] (endConv) -- (gap);
	\draw[-latex] (gap) -- (out);
	\draw[-latex] (out) -- (O);
	\draw[decoration = {brace, mirror, raise=0.5em, amplitude=0.5em}, decorate]	(fire.south west) -- node[below=5mm] {$n$ instances} (fire.south east);
	\end{tikzpicture}
    \forceversofloat
	\caption[FireNet Network Architecture, based on the SmallFire module]{FireNet Network Architecture, based on the SmallFire module. All layers use ReLU activation}
	\label{sic:fireNet}
\end{figure*}

FireNet is built in a similar way, but differently from ClassicCNN. This network contains an initial convolution to "expand" the number of available channels, as sonar images are single channel. Then $N$ MaxFire modules are stacked. Then a final convolution is used, in order to change the number of channels to $C$. Then global average pooling \cite{lin2013network} is applied to reduce feature maps from any size to $1 \times 1 \times C$. FireNet is shown in Figure \ref{sic:fireNet}.
TinyNet is similarly constructed, but it does not have a initial convolution. It contains a stack of $n$ Tiny modules with a final $1 \times 1$ convolution to produce $C$ output channels. Global average pooling is applied and a softmax activation is used to produce output class probabilities. TinyNet is shown in Figure \ref{sic:tinyNet}.

Both FireNet and TinyNet do not use fully connected layers for classification, and instead such layers are replaced by global average pooling and a softmax activation. This is a very different approach, but it is useful as it reduces the number of learning parameters, reducing the chance of overfitting and increasing computational performance.

Each network is trained using the same algorithm, namely gradient descent with the ADAM optimizer \cite[-1em]{kingma2014adam}, using an initial learning rate of $\alpha = 0.01$ and a batch size $B = 64$.

\section{Experimental Evaluation}

\subsection{Evaluation Metrics}

For multi-class classification problems, the most common evaluation metric is accuracy, as it can be directly interpreted by a human, and has intuitive interpretations. Accuracy is computed as the fraction of samples that are correctly classified, with $0 \%$ as the worst value, and $100 \%$ as the best. A related metric commonly used in the ML literature is the error \cite[2em]{Goodfellow2016deep}, which is the inverse ($100 - \text{acc}$) of accuracy. There is no advantage of choosing accuracy versus error, so we pragmatically choose accuracy.

Note that accuracy can be biased by class imbalance and improper classifiers. While our dataset is not class balanced, the classifiers we evaluated do not collapse to predicting a single class, and all classes are well represented in predictions, indicating that using accuracy

Other metrics are less appropriate, such as precision and recall (well defined only for binary classification), and the area under the ROC curve (not directly interpretable by humans, and well defined for binary classification).

\subsection{Convolutional Network Design}
\label{sic:cnnDesignSection}
               
In this section we explore some design choices that must be made. We parameterized our networks primarily with two parameters: the number of modules (which affects depth) and the number of convolution filters (affecting width). We evaluate the importance of these parameters by training multiple networks over a defined grid of parameter values and we test each network in a validation set. We then make decision of which network configurations we will use later.

\begin{description}
	\item[ClassicNet] \hfill \\
		For this network architecture we vary depth through the number of modules, in the range $[1, 6]$. As the input images are $96 \times 96$, $6 = \log_2{96}$ is the biggest number of modules that we can try before max-pooling reduces feature maps to zero width and height. We evaluate $8$, $16$ and $32$ filters per module, as we have previously used networks with 32 filters that performed adequately \cite{valdenegro2016object}.
		We also evaluate the effect of regularization as an additional parameter. We use Batch Normalization, Dropout and No Regularization, which corresponds to the removal of any regularization layer from the model. Batch Normalization is used after each layer, except the last. Dropout is only used after the first fully connected layer.
	\item[TinyNet] \hfill \\
		For this architecture we also evaluate up to 6 modules, but only 4 and 8 filters. The main reason driving the number of filters is to minimize the total number of parameters, as these networks were designed for fast executing in embedded devices \cite[-1.5em]{valdenegro2017rtcnns}.
	\item[FireNet] \hfill \\
		This network was evaluated up to 6 modules, as accuracy saturated at the maximum value with more modules. We only evaluate 4 filters per module, corresponding to $s_{11} = e_{11} = e_{33} = 4$ filters in each Fire module inside the MaxFire one.
\end{description}

Each network is trained in the same way, using the ADAM optimizer \cite{kingma2014adam} with an initial learning rate $\alpha = 0.01$. ClassicNet is trained for 20 epochs, while TinyNet and FireNet are trained for 30 epochs. We train 10 instances of each network architecture for each parameter set. We do this because of random initialization, as training a single network can produce biased or "lucky" results. For each parameter set we report the mean and standard deviation of accuracy evaluated on the validation set.

\begin{figure*}[t]
	\centering
	\subfloat[]{
		\begin{tikzpicture}
			\begin{customlegend}[legend columns = 8,legend style = {column sep=1ex}, legend cell align = left,
			legend entries={Batch Normalization, Dropout, No Regularization}]
			\addlegendimage{red}
			\addlegendimage{green}
			\addlegendimage{blue}
			\end{customlegend}
		\end{tikzpicture}
	}

	\subfloat[8 filters] {
		\begin{tikzpicture}
		\begin{axis}[
		height = 0.3 \textheight,
		xtick = {1,2,3,4,5,6},
		ytick = {92, 94, 96, 97, 98, 99, 100},
		xlabel={\# of Modules},
		ylabel={Val Accuracy (\%)},
		ymin=90, ymax=100,
		legend pos=south east,
		ymajorgrids=true,
		grid style=dashed,
		width = 0.32\textwidth]
		
		\errorband{chapters/data/classicNet/classicCNN-nf8-fs5x5-BN-DepthVsAcc.csv}{numConvModules}{meanTestAcc}{stdTestAcc}{red}{0.4}
		\errorband{chapters/data/classicNet/classicCNN-nf8-fs5x5-Dropout-DepthVsAcc.csv}{numConvModules}{meanTestAcc}{stdTestAcc}{green}{0.4}
		\errorband{chapters/data/classicNet/classicCNN-nf8-fs5x5-NoReg-DepthVsAcc.csv}{numConvModules}{meanTestAcc}{stdTestAcc}{blue}{0.4}		
		\end{axis}
		\end{tikzpicture}
	}
	\subfloat[16 filters] {
		\begin{tikzpicture}
		\begin{axis}[
		height = 0.3 \textheight,
		xtick = {1,2,3,4,5,6},
		ytick = {92, 94, 96, 97, 98, 99, 100},
		xlabel={\# of Modules},
		ymin=90, ymax=100,
		legend pos=south east,
		ymajorgrids=true,
		grid style=dashed,
		width = 0.32\textwidth]
		
		\errorband{chapters/data/classicNet/classicCNN-nf16-fs5x5-BN-DepthVsAcc.csv}{numConvModules}{meanTestAcc}{stdTestAcc}{red}{0.4}
		\errorband{chapters/data/classicNet/classicCNN-nf16-fs5x5-Dropout-DepthVsAcc.csv}{numConvModules}{meanTestAcc}{stdTestAcc}{green}{0.4}
		\errorband{chapters/data/classicNet/classicCNN-nf16-fs5x5-NoReg-DepthVsAcc.csv}{numConvModules}{meanTestAcc}{stdTestAcc}{blue}{0.4}
		\end{axis}
		\end{tikzpicture}
	}
	\subfloat[32 filters] {
		\begin{tikzpicture}
			\begin{axis}[
			height = 0.3 \textheight,
			xtick = {1,2,3,4,5,6},
			ytick = {92, 94, 96, 97, 98, 99, 100},
			xlabel={\# of Modules},
			ymin=90, ymax=100,
			legend pos=south east,
			ymajorgrids=true,
			grid style=dashed,
			width = 0.32\textwidth]
			
			\errorband{chapters/data/classicNet/classicCNN-nf32-fs5x5-BN-DepthVsAcc.csv}{numConvModules}{meanTestAcc}{stdTestAcc}{red}{0.4}
			\errorband{chapters/data/classicNet/classicCNN-nf32-fs5x5-Dropout-DepthVsAcc.csv}{numConvModules}{meanTestAcc}{stdTestAcc}{green}{0.4}
			\errorband{chapters/data/classicNet/classicCNN-nf32-fs5x5-NoReg-DepthVsAcc.csv}{numConvModules}{meanTestAcc}{stdTestAcc}{blue}{0.4}
			
			\end{axis}
		\end{tikzpicture}
	}
	\vspace*{0.5cm}
    \forceversofloat
	\caption[ClassicNet Network Depth versus Validation Accuracy]{ClassicNet Network Depth versus Validation Accuracy. The shaded areas represent one-$\sigma$ standard deviations.}
	\label{sic:classicNetTuning}
    \vspace*{0.5cm}
\end{figure*}
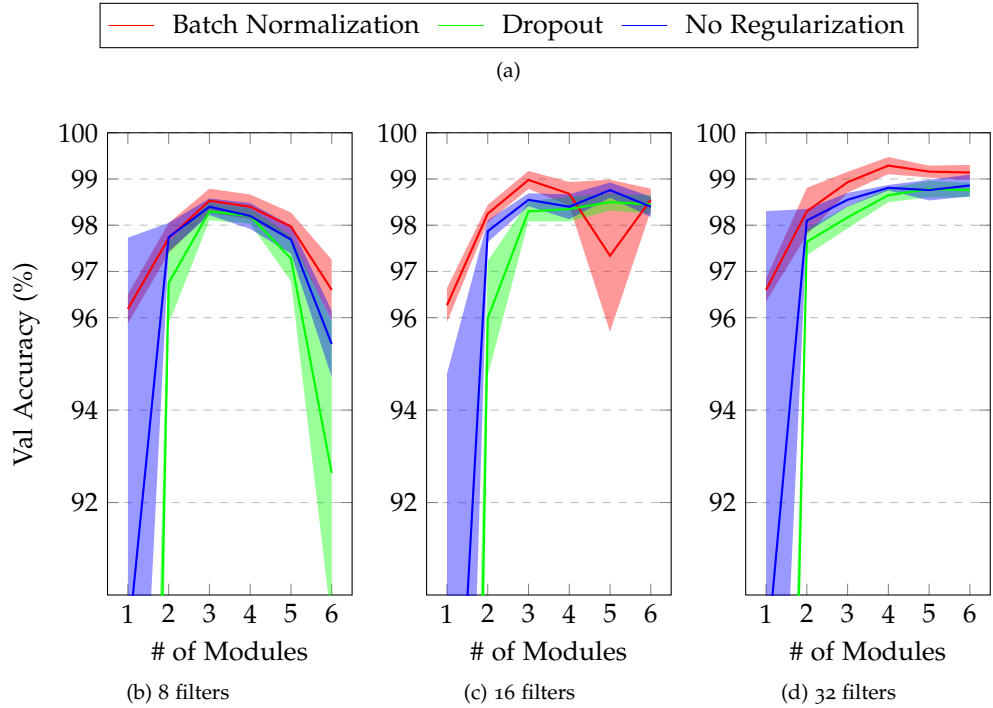

ClassicNet results are shown in Figure \ref{sic:classicNetTuning}. We see that a choice of 32 filters seems to be the best, as it produces the biggest accuracy in the validation set and learning seems to be more stable. Configurations with less filters seem to be less stable, as shown in the 8-filter configuration with decreasing accuracy after adding more than 3 modules, and the 16-module configuration showing large variations in accuracy.
In general it is expected that a deeper network should have better accuracy, but tuning the right number of layers/modules is not easy, as these results show.

We compared three other design choices, whether to use regularization (Dropout or Batch Normalization), or not use it. Our results clearly show that Batch Normalization outperforms using Dropout by a large margin, but this only holds with a large enough learning capacity, represented as number of modules.
An interesting effect is that removing regularization from the network increases the variability in the results, specially when the model does not fit well the data as seen in the case of using a single module per network.

\begin{figure}[t]
    \forcerectofloat
	\centering
	
	\subfloat[TinyNet] {
		\centering
		\begin{tikzpicture}
		\begin{axis}[
		height = 0.3 \textheight,
		xtick = {1,2,3,4,5,6},
		ytick = {92, 94, 96, 97, 98, 99, 100},
		xlabel={\# of Modules},
		ylabel={Val Accuracy (\%)},
		ymin=90, ymax=100,
		legend pos=south east,
		ymajorgrids=true,
		grid style=dashed,
		width = 0.48\textwidth]
		
		\errorband{chapters/data/tinyNet/tinyNet-nf4-DepthVsAcc.csv}{numConvModules}{meanTestAcc}{stdTestAcc}{red}{0.4}
		\errorband{chapters/data/tinyNet/tinyNet-nf8-DepthVsAcc.csv}{numConvModules}{meanTestAcc}{stdTestAcc}{green}{0.4}
		\legend{4 Filters, 8 Filters}
		\end{axis}
		\end{tikzpicture}
	}
	\subfloat[FireNet] {
		\centering
		\begin{tikzpicture}
		\begin{axis}[
		height = 0.3 \textheight,
		xtick = {1,2,3,4,5,6},
		ytick = {92, 94, 96, 97, 98, 99, 100},
		xlabel={\# of Modules},
		ymin=90, ymax=100,
		legend pos=south east,
		ymajorgrids=true,
		grid style=dashed,
		width = 0.48\textwidth]
		
		\errorband{chapters/data/fireNet/fireNet-nf4-DepthVsAcc.csv}{numConvModules}{meanTestAcc}{stdTestAcc}{red}{0.4}
		\legend{4 Filters}
		\end{axis}
		\end{tikzpicture}
	}
	\vspace*{0.5cm}
	\caption[TinyNet and FireNet Network Depth versus Validation Accuracy]{TinyNet and FireNet Network Depth versus Validation Accuracy. The shaded areas represent one-$\sigma$ standard deviations.}
	\label{sic:tinyFireTuning}
\end{figure}
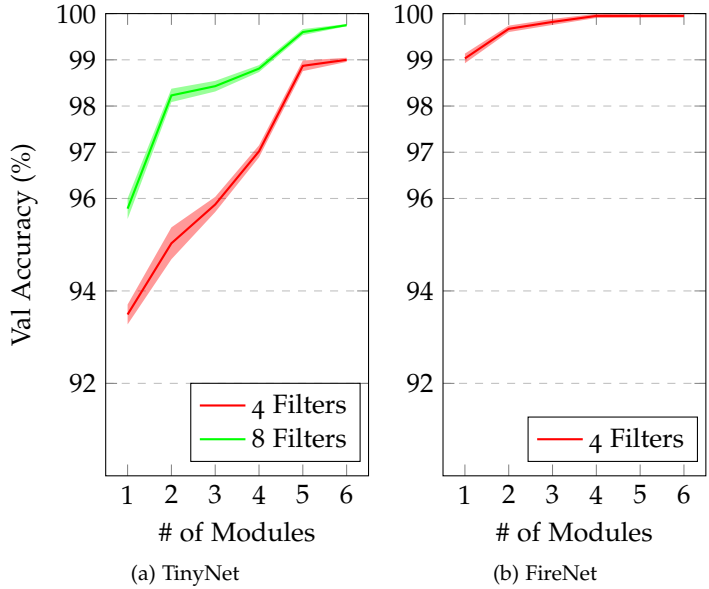

TinyNet and FireNet results are shown in Figure \ref{sic:tinyFireTuning}. TinyNet was designed in order to minimize the number of parameters, even if that requires some sacrifices in accuracy. Our results show that with 5 modules we can obtain high accuracy, which is comparable to what ClassicNet can obtain. As it can be expected, TinyNet with four filters is slightly less accurate than using eight filters, but the difference gets smaller as more modules are added.

In contrast, FireNet gets state of the art accuracy with only 1 module, and perfectly fits the dataset with two or more modules. Only four filters seem to be necessary for this task.

A numerical summary of these results is presented in Table \ref{sic:comparisonCNTNFN}. The best result for ClassicNet is with Batch Normalization, 32 filters and four modules at $99.29 \%$ accuracy, while TinyNet has best performance with 8 filters and 5 modules at $99.60 \%$ accuracy. A slightly less accurate version of TinyNet is available with also 5 modules and 4 filters, at $98.87 \%$ accuracy, which is a $0.7 \%$ absolute difference. 

\begin{table}[t]
    \forcerectofloat
	\begin{tabular}{lllll}
		\hline
		& Fs	& 1 Module 			  & 2 Modules 			& 3 Modules \\
		\hline
		\multirow{3}{*}{\rotatebox{90}{CN}}& 8 	& $96.19 \pm 0.63$ \% & $97.72 \pm 0.66$ \% & $98.53 \pm 0.52$ \% \\
		& 16 & $96.27 \pm 0.74$ \% & $98.25 \pm 0.38$ \% & $98.98 \pm 0.39$ \% \\
		& 32 & $96.60 \pm 0.50$ \% & $98.30 \pm 1.02$ \% & $98.93 \pm 0.45$ \% \\
		\hline
		\multirow{2}{*}{\rotatebox{90}{TN}} & 4 & $93.49 \pm 0.43$ \% & $95.03 \pm 0.69$ \% & $95.87 \pm 0.33$ \% \\
		& 8 & $95.78 \pm 0.45$ \% & $98.23 \pm 0.29$ \% & $98.43 \pm 0.23$ \% \\
		\hline
		\multirow{2}{*}{\rotatebox{90}{FN}} & 4 & $99.03 \pm 0.21$ \% & $99.67 \pm 0.14$ \% & $99.82 \pm 0.13$ \% \\
		\\
		\hline
	\end{tabular}

    \begin{tabular}{lllll}
        \hline
        & Fs	& 4 Modules 			& 5 Modules 		  & 6 Modules\\
        \hline
        \multirow{3}{*}{\rotatebox{90}{CN}}& 8 	& $98.40 \pm 0.53$ \% & $97.97 \pm 0.61$ \% & $96.60 \pm 1.31$ \%\\
        & 16 & $98.68 \pm 0.51$ \% & $97.34 \pm 3.29$ \% & $98.55 \pm 0.38$ \%\\
        & 32 & $99.29 \pm 0.37$ \% & $99.19 \pm 0.26$ \% & $99.14 \pm 0.33$ \%\\
        \hline
        \multirow{2}{*}{\rotatebox{90}{TN}} & 4 & $97.02 \pm 0.26$ \% & $98.87 \pm 0.22$ \% & $98.95 \pm 0.09$ \%\\
        & 8 & $98.81 \pm 0.15$ \% & $99.60 \pm 0.13$ \% &  $99.75 \pm 0.05$ \%\\
        \hline
        \multirow{2}{*}{\rotatebox{90}{FN}} & 4 & $99.95 \pm 0.09$ \% & $99.95 \pm 0.08$ & $99.95 \pm 0.07$\%\\
        \\
        \hline
    \end{tabular}
    \vspace*{0.5cm}
	\caption[Numerical comparison of validation accuracy of different network configurations as function of number of modules]{Numerical comparison of validation accuracy of different network configurations, as function of number of modules. ClassicNet with BN (CN), TinyNet (TN), and FireNet (FN) are evaluated. Varying filter (Fs) configurations are shown.}
	\label{sic:comparisonCNTNFN}
\end{table}

\subsection{Template Matching Baselines}

We have also evaluated a simple template matching classifier on our marine debris dataset. Our motivation for this experiment is to validate our theoretical prediction that a template matching classifier is more likely to produce overfitting to objects in the training set.

To construct a template matching classifier, two key parameters must be tuned: the number of templates and the contents of each template. As the whole training set of a template matching classifier is required at inference time, its design is key to good performance.

In order to provide a unbiased evaluation, we sample $T$ templates for each class and build a template set. We evaluate CC and SQD matching with these templates. As the number of samples per each class is the same, there is no class balance problems, even as the test set is not balanced. Selecting random images from the training set prevents one template dominating and being able to classify a large set of images. This effect would show up as a large variation in test accuracy.

To measure the effect of the number of templates, we vary $T \in [1, 150]$. As our dataset has 11 classes, this range produces template sets from 11 to 1650 templates in total. We sample $N = 100$ different template sets for each value of $T$, and evaluate accuracy on the test set for each template set. We report the mean and standard deviation of accuracy.

We also evaluate the number of free parameters in the template matching classifier. We compute the number of parameters $P$ in a template set $D$ as:
\vspace*{1em}
\begin{equation}
	P = M \times W \times H
\end{equation}

Where $M = |D|$ is the number of templates in the template set, $W$ is the width, and $H$ is the height. For the images in our dataset $W = H = 96$. We also evaluate computation time as measured on a AMD Ryzen 7-1700 CPU.

\begin{figure}
	\centering
	\begin{tikzpicture}
	\begin{axis}[
	height = 0.3 \textheight,
	xtick = {1, 10, 20, 30, 40, 50, 60, 70, 80, 90, 100, 110, 120, 130, 140, 150},
	ytick = {40, 50, 60, 70, 80, 85, 90, 95, 100},
	xlabel={\# of Templates per Class},
	ylabel={Test Accuracy (\%)},
	ymin=40, ymax=100,
	legend pos=south east,
	ymajorgrids=true,
	x tick label style={font=\tiny, rotate=90}]
	
	\errorband{chapters/data/templateMatching-CC-accuracyVsTemplates.csv}{tspc}{meanAcc}{stdAcc}{red}{0.4}
	\errorband{chapters/data/templateMatching-SQD-accuracyVsTemplates.csv}{tspc}{meanAcc}{stdAcc}{green}{0.4}
	
	\legend{CC, SQD}
	\end{axis}
	\end{tikzpicture}
	\caption[Template Matching Test Accuracy as a function of the TPC with two similarity functions]{Template Matching Test Accuracy as a function of the Number of Templates per Class with two similarity functions. The shaded areas represent one $\sigma$ error bounds.}
	\label{sic:tmAccVsTSPC}
\end{figure}
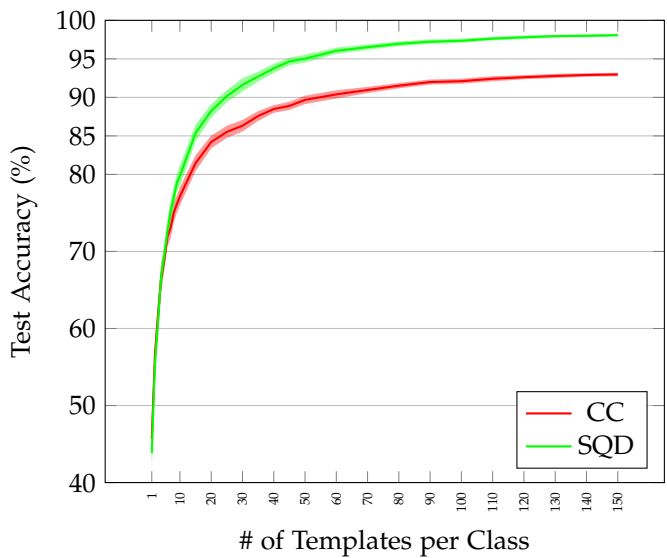

Our principal results are presented in Fig. \ref{sic:tmAccVsTSPC} and Table \ref{sic:tmResultsData}. Classification accuracy results show that using cross-correlation similarity saturates at $93.5 \%$, while sum of squared differences saturates closer to $98.4 \%$. Sum of squared differences is clearly superior for generalization over the commonly used cross-correlation similarity.

It must be pointed out that SQD performs better in our tests, but theoretically it should not generalize well, as due to the use of a pixel-wise distance there is less possible transformations that maintain such distances. SQD is not used in the literature, and cross-correlation is mostly preferred, so more research is required.

In order to obtain such generalization over a test set, both methods require a considerably high number of templates. CC crosses the $80$ \% threshold with 20 templates per class, while SQD obtains at least the same accuracy with only 10 templates per class. $90$ \% accuracy is only reached 70 (CC) and 30 (SQD) templates per class. After reaching the point of $90$ \% accuracy, both methods produce diminishing returns, with only small increments in accuracy even after using a large (more than 100) number of templates. These observations have two possible explanations:

\begin{itemize}
	\item The inter-class variability of our marine debris images is high, and this is reflected in the fact that many templates are required to model such variability.
	\item Template matching methods require a large number of parameters to generalize well, and one template can only model a limited number of testing samples due to limitations in the matching process.
\end{itemize}

Analyzing the number of parameters (available in Table \ref{sic:tmResultsData}) shows that to model our data both template matching methods require at least 10 million parameters. This is a considerable number of parameters, as it contains almost the complete training set (at $\text{TPC} = 150$).

\begin{table*}[t]
    \forceversofloat
	\begin{tabular}{llllll}
		\hline
				&              & \multicolumn{2}{c}{CC}		& \multicolumn{2}{c}{SQD}\\
		TPC     & \# of Params & Accuracy (\%) & Time (ms)  & Accuracy (\%) & Time (ms)\\
		\hline
		1.0     & 0.10 M       & $ 45.67 \pm 5.99$ \%    &       $ 0.7 \pm 0.1$ ms & $ 43.85 \pm 4.94$ \%    &       $ 0.2 \pm 0.0$ ms \\
		5.0     & 0.50 M       & $ 68.9 \pm 3.01$ \%     &       $ 3.2 \pm 0.1$ ms & $ 69.47 \pm 3.18$ \%    &       $ 1.1 \pm 0.0$ ms\\
		10.0    & 1.01 M       & $ 77.17 \pm 2.35$ \%    &       $ 6.3 \pm 0.1$ ms & $ 79.83 \pm 2.67$ \%    &       $ 2.3 \pm 0.1$ ms\\
		20.0    & 2.02 M       & $ 84.21 \pm 1.65$ \%    &       $ 12.6 \pm 0.1$ ms & $ 88.25 \pm 1.75$ \%    &       $ 4.6 \pm 0.1$ ms\\
		30.0    & 3.04 M       & $ 86.32 \pm 1.52$ \%    &       $ 18.9 \pm 0.2$ ms & $ 91.62 \pm 1.75$ \%    &       $ 7.0 \pm 0.1$ ms\\
		40.0    & 4.05 M       & $ 88.49 \pm 1.0$ \%     &       $ 25.2 \pm 0.7$ ms & $ 93.76 \pm 1.2$ \%     &       $ 9.2 \pm 0.1$ ms\\
		50.0    & 5.07 M       & $ 89.67 \pm 1.09$ \%    &       $ 31.4 \pm 0.3$ ms & $ 95.03 \pm 1.02$ \%    &       $ 11.6 \pm 0.1$ ms\\
		60.0    & 6.09 M       & $ 90.39 \pm 1.08$ \%    &       $ 37.6 \pm 0.3$ ms & $ 96.05 \pm 0.81$ \%    &       $ 13.9 \pm 0.2$ ms\\
		70.0    & 7.09 M       & $ 90.96 \pm 0.81$ \%    &       $ 43.9 \pm 0.4$ ms & $ 96.52 \pm 0.71$ \%    &       $ 16.2 \pm 0.2$ ms\\
		80.0    & 8.11 M       & $ 91.52 \pm 0.7$ \%     &       $ 50.1 \pm 0.4$ ms & $ 96.96 \pm 0.63$ \%    &       $ 18.6 \pm 0.2$ ms\\
		90.0    & 9.12 M       & $ 91.99 \pm 0.67$ \%    &       $ 56.5 \pm 0.4$ ms & $ 97.23 \pm 0.55$ \%    &       $ 20.7 \pm 0.2$ ms\\
		100.0   & 10.13 M      & $ 92.1 \pm 0.65$ \%     &       $ 62.7 \pm 0.5$ ms & $ 97.35 \pm 0.54$ \%    &       $ 23.0 \pm 0.2$ ms\\
		110.0   & 11.15 M      & $ 92.42 \pm 0.67$ \%    &       $ 68.9 \pm 0.5$ ms & $ 97.63 \pm 0.46$ \%    &       $ 25.2 \pm 0.3$ ms\\
		120.0   & 12.16 M      & $ 92.62 \pm 0.54$ \%    &       $ 75.1 \pm 0.5$ ms & $ 97.8 \pm 0.46$ \%     &       $ 27.5 \pm 0.3$ ms\\
		130.0   & 13.17 M      & $ 92.78 \pm 0.56$ \%    &       $ 81.3 \pm 0.6$ ms & $ 97.95 \pm 0.34$ \%    &       $ 29.8 \pm 0.3$ ms\\
		140.0   & 14.19 M      & $ 92.91 \pm 0.46$ \%    &       $ 87.7 \pm 0.6$ ms & $ 97.99 \pm 0.39$ \%    &       $ 32.1 \pm 0.3$ ms\\
		150.0   & 15.20 M      & $ 92.97 \pm 0.47$ \%    &       $ 93.8 \pm 0.7$ ms & $ 98.08 \pm 0.34$ \%    &       $ 34.6 \pm 0.3$ ms\\
		\hline
	\end{tabular}
	\vspace*{0.5cm}
	\caption[Template Matching with Cross-Correlation and Sum of Squared Differences]{Template Matching with Cross-Correlation (CC) and Sum of Squared Differences (SQD). Accuracy, Number of Parameters and Computation versus Number of Templates per Class is presented. Number of parameters is expressed in Millions.}
	\label{sic:tmResultsData}
\end{table*}

Computation time is shown in Table \ref{sic:tmResultsData}. SQD is considerably faster, almost 3 times faster than CC. Both similarity functions can run on real-time on a normal computer, but the required large number of templates could be an issue when using these classifiers on embedded platforms.

\subsection{Comparison with State of the Art}

In this section we compare multiple classification algorithms in sonar images, in order to put our classification results with CNNs into context. We have selected algorithms in the following categories:

\begin{description}
	\item[Template Matching] We include the best results produced by a CC and SQD template matching classifiers as described in the previous section.
	\item[Classic Machine Learning] We evaluate multiple classic classification algorithms, namely a SVM with Linear and RBF kernels, Gradient Boosting and a Random Forest. All of these classifiers are trained on normalized image pixels.
	\item[Neural Networks] We also include our best results produced by CNN classifiers, as described in Section \ref{sic:cnnDesignSection}.
\end{description}

All classifiers\footnote{I used the scikit-learn 0.18.1 implementation of these algorithms} are trained on the same training set, and evaluated on the same testing set. Each classifier was tuned independently to produce maximum classification performance on the validation set, using grid search with a predefined set of parameters.

For a Random Forest, we tuned the maximum number of features (either $log_2(n)$ or $sqrt(n)$, where $n$ is the number of input features) and the number of trees (in the range $[100, 200, 300, ..., 1000]$ ). The best parameters reported by grid search are $log_2$ number of features and 600 trees, producing $7901$ parameters.

A Gradient Boosting \cite{murphy2012machine} classifier was tuned over three parameters, namely number of weak classifiers (in range $[50, 100, 150, ..., 1000]$), the learning rate $[0.1, 0.01, 0.001]$, and the maximum depth of the regression tree with values $[3, 5, 7, 9, 11, 15]$. The best parameters were $300$ weak classifiers, learning rate $0.1$, and maximum depth of $3$. This produces $9900$ parameters.

For the SVM classifier we only tuned the regularization coefficient $C$ in the range $[10^{-3}, 10^{-2}, 10^{-1}, ..., 10^6]$ and used two types of kernels: a gaussian radial basis function (RBF) and a linear kernel. As an SVM is only a binary classifier, we use one-versus-one decision function which consists of training $\frac{C (C - 1)}{2}$ SVMs and evaluate them at test time and get the majority decision as class output. The linear kernel gets best performance with $C = 0.1$, while the RBF uses $C = 100.0$. Both classifiers have 506880 parameters, considering 110 trained SVMs. The ratio of parameters to number of training data points ranges from $\frac{1200}{2069} \sim 0.58$ for TinyNet(5, 4) to $\frac{15.2 M}{2069} \sim 7346.5$ for template matching classifiers. ClassicCNN with Dropout has a parameter to data point ratio of $\frac{930000}{2069} \sim 449.5$, which is reduced to $224.8$ during the training phase due to the use of Dropout.

Our comparison results are shown in Table \ref{sic:comparisonMLDLTM}. Machine learning methods perform poorly, as gradient boosting and random forests obtain accuracies that are lower than simpler classifiers like a Linear SVM. We expected both algorithms to perform better, due to their popularity in competitions like Kaggle, which suggests that they might generalize well with low amounts of data.

The best classic ML classifier according to our results is a Linear SVM, which is surprising, but still it does not perform better than the state of the art template matching methods using sum of square differences.

The best performing method is a convolutional neural network, either TinyNet with 5 modules and 8 filters per layer or FireNet with 3 layers. There is a small difference in accuracy ($0.1$ \%) between both networks. These results show that a CNN can be successfully trained with a small quantity of data (approx 2000 images) and that even in such case, it can outperform other methods, specially template matching with cross-correlation and ensemble classifiers like random forests and gradient boosting.

The second best performing method is also a CNN, namely the ClassicCNN. It should be noted that there is a large difference in the number of parameters between ClassicCNN and TinyNet/FireNet. Those networks are able to efficiently encode the mapping between image and class. Considering TM-SQD as the baseline, then TinyNet(5, 8) is $1.18 $ \% more accurate, and FireNet-3 outperforms it by $1.28$ \%. A more realistic baseline is TM-CC as it is used in many published works \cite{hurtos2013automatic}, and in such case TinyNet(5, 8) outperforms TM-CC by $6.16$ \% while FireNet-3 is superior by $6.26$ \%.

\begin{table}[t]
	\begin{tabular}{llll}
		\hline 
		& Method 							& Test Accuracy (\%)	& \# of Params\\ 
		\hline 
		\multirow{4}{*}{\rotatebox{90}{ML}} & RBF SVM		& $97.22$ \% 			& 506K\\ 
		& Linear SVM	& $97.46$ \% 			& 506K\\ 
		& Gradient Boosting				& $90.63$ \% 			& 9.9K\\
		& Random Forest					& $93.17$ \%  			& 7.9K\\
		\hline
		\multirow{2}{*}{\rotatebox{90}{TM}} & TM with CC		& $93.44$ \% 			& 15.2M \\
		& TM with Sum of SQD	& $98.42$ \% 			& 15.2M \\
		\hline
		\multirow{5}{*}{\rotatebox{90}{CNN}} & ClassicCNN-BN					& $99.24$ \% 	& 903K\\
		& ClassicCNN-Dropout			& $98.98$ \% 			& 903K\\
		\cline{2-4}
		& TinyNet(5, 4)					& $98.8$ \%				& \textbf{1.2K}\\
		& TinyNet(5, 8)					& $\textbf{99.6}$ \%	& 3.2K\\
		\cline{2-4}
		& FireNet-3						& $\textbf{99.7}$ \%	& 4.0K\\
		\hline 
	\end{tabular} 
	\caption{Comparison of Machine Learning (ML), Template Matching (TM), and Convolutional Neural Networks (CNN) methods for FLS Image Classification.}
	\label{sic:comparisonMLDLTM}
\end{table}

\subsection{Feature Visualization}

In the previous section we have established that a CNN is the best classifier for FLS images. In this section we would like to move away from a quantitative analysis of plain accuracy on a test set and focus into the features that are learned by a CNN. This corresponds to a qualitative approach.

For this purpose we extract features from selected layers in a neural network, reduce the dimensionality of these features, and then visualize the 2D representation of these features as a scatter plot, including class information.

We use two methods for dimensionality reduction:

\begin{description}
	\item[t-distributed Stochastic Neighbor Embedding] t-SNE \cite{maaten2008visualizing} is a well known technique for dimensionality reduction with emphasis on data visualization. t-SNE works by first estimating pair-wise similarities of high-dimensional input points $x_i$:
    
	\begin{equation}
		p_{j|i} = \frac{\text{exp}(-||x_i - x_j||^2 / 2\sigma_i^2)}{\sum _{k \neq i}\text{exp}(-||x_i - x_k||^2 / 2\sigma_i^2)}
	\end{equation}
	
	Which corresponds to placing a gaussian distribution at $x_i$ with variance $\sigma_i^2$. This variance is automatically selected by binary search given a used defined perplexity. The basic idea of t-SNE is to model these probabilistic similarities with another low-dimensional distribution:
	\vspace*{1em}
	\begin{equation}
		q_{ij} = \frac{(1 + ||y_i - y_j||^2)^{-1}}{\sum_{k \neq l} (1 + ||y_k - y_l||^2)^{-1} }
	\end{equation}
	
	The $y_i$ values are low-dimensional representation of the input data $x_i$. The $q_{ij}$ equation is selected to mimic a student's t-distribution, which is more robust to outliers and to variations in feature space. Then t-SNE tries to make the $Q$ distribution similar to the $P$ distribution, by moving the $y_i$ values to minimize a distance metric for probability distributions: the KL-divergence:
	\vspace*{1em}
	\begin{equation}
		KL(P||Q) = \sum_i \sum_j p_{ij} \log \frac{p_{ij}}{q_{ij}}
	\end{equation}
	
	Where $p_{ij} = \frac{p{j|i} + p_{i|j}}{2n}$. Then stochastic gradient descent is used to minimize the KL divergence, which produces the low-dimensional representation.
	t-SNE is commonly used to visualize high-dimensional data, as it has desirable properties, like the ability to model structure in the input space at several scales, and to let local structure influence the final result.
	
	\item[Multi-Dimensional Scaling] MDS is a another non-linear dimensionality reduction method. For our purposes we use Metric MDS \cite{de2011multidimensional}, by first estimating a matrix of distances from high-dimensional input $x_i$:
	\vspace*{1em}
	\begin{equation}
		d_{ij} = ||x_i - x_j||
	\end{equation}
	
	Then in a similar idea that t-SNE, as MDS also wants to find a low-dimensional representation that approximates the distances $d_{ij}$, by minimizing the following loss function:
	\vspace*{1em}
	\begin{equation}
		S = \sum_{i \neq j} (d_{ij} - ||y_i - y_j||)^2
	\end{equation}
	
	$S$ is denominated the stress. By minimizing the stress, the locations of each low-dimensional point $y_i$ can be found. It should be noted that MDS preserves real distances more closely than t-SNE, as the real distance is approximated, instead of a proxy probabilistic similarity. The stress is minimized by using SMACOF (Scaling by Majorizing a Complicated Function).
	
\end{description}

Both methods are commonly used to visualize high-dimensional data into a 2D/3D manifold. We use two methods in order not to obtain general conclusions from a single method, as performing dimensionality reduction for visualization does include a degree of bias (from selected parameters) and uncertainty (due to stochastic sampling).

We selected these two methods as t-SNE is one of the best dimensionality reducers with strong theoretical support, while MDS is a very simple method with no tunable parameters other than a distance metric. As previously mentioned, MDS also preserves real distances in the high-dimensional space better than t-SNE.

We selected three previously mentioned neural networks for evaluation: TinyNet with 5 modules and 8 filters per module, and ClassicNet with 5 modules with Batch Normalization or Dropout. We extract high-dimensional features from each module, and additionally from the FC1 layer in ClassicNet. The feature dimensionality for each module in each network is shown in Table \ref{sic:featureDimensions}.

Before feature extraction we subsample the test set to 50 samples per class, in order to normalize the number of data points in each plot. This produces 550 points in each scatter plot. To extract features for a given layer we perform the following procedure: Compute the features for each data point in the sub-sampled test set\footnote[][7em]{The same sub-sampled test set is used to produce all plots, in order to enable comparison}, then they are L2 normalized. Then dimensionality reduction is applied and 2D reduced features are displayed in a scatter plot. Class labels are not used while performing dimensionality reduction, but only used for display purposes while constructing the scatter plot.

\begin{figure*}[p]
	\centering
	\subfloat[Module 1]{
		\begin{tikzpicture}
		\begin{axis}[width = 0.32 \textwidth,
		scatter/classes={
			0={mark=square*,green},
			1={mark=triangle*,red},
			2={mark=o,blue},
			3={mark=diamond*,yellow},
			4={mark=pentagon*,black},
			5={mark=square*,magenta},
			6={mark=triangle*,cyan},
			7={mark=o,gray},
			8={mark=diamond*,brown},
			9={mark=square*,darkgray},
			10={mark=pentagon*,teal}
		}]
		\addplot[scatter,only marks, scatter src=explicit symbolic] table[x  = x, y  = y, meta = classId, col sep = space] {chapters/data/mds-test-embedding-tinyNet-module0.csv};		
		\end{axis}
		\end{tikzpicture}
	}
	\subfloat[Module 2]{
		\begin{tikzpicture}
		\begin{axis}[width = 0.32 \textwidth,
		scatter/classes={
			0={mark=square*,green},
			1={mark=triangle*,red},
			2={mark=o,blue},
			3={mark=diamond*,yellow},
			4={mark=pentagon*,black},
			5={mark=square*,magenta},
			6={mark=triangle*,cyan},
			7={mark=o,gray},
			8={mark=diamond*,brown},
			9={mark=square*,darkgray},
			10={mark=pentagon*,teal}
		}]
		\addplot[scatter,only marks, scatter src=explicit symbolic] table[x  = x, y  = y, meta = classId, col sep = space] {chapters/data/mds-test-embedding-tinyNet-module1.csv};		
		\end{axis}
		\end{tikzpicture}
	}
	\subfloat[Module 3]{
		\begin{tikzpicture}
		\begin{axis}[width = 0.32 \textwidth,
		scatter/classes={
			0={mark=square*,green},
			1={mark=triangle*,red},
			2={mark=o,blue},
			3={mark=diamond*,yellow},
			4={mark=pentagon*,black},
			5={mark=square*,magenta},
			6={mark=triangle*,cyan},
			7={mark=o,gray},
			8={mark=diamond*,brown},
			9={mark=square*,darkgray},
			10={mark=pentagon*,teal}
		}]
		\addplot[scatter,only marks, scatter src=explicit symbolic] table[x  = x, y  = y, meta = classId, col sep = space] {chapters/data/mds-test-embedding-tinyNet-module2.csv};		
		\end{axis}
		\end{tikzpicture}
	}
	
	\subfloat[Module 4]{
		\begin{tikzpicture}
		\begin{axis}[width = 0.32 \textwidth,
		scatter/classes={
			0={mark=square*,green},
			1={mark=triangle*,red},
			2={mark=o,blue},
			3={mark=diamond*,yellow},
			4={mark=pentagon*,black},
			5={mark=square*,magenta},
			6={mark=triangle*,cyan},
			7={mark=o,gray},
			8={mark=diamond*,brown},
			9={mark=square*,darkgray},
			10={mark=pentagon*,teal}
		}]
		\addplot[scatter,only marks, scatter src=explicit symbolic] table[x  = x, y  = y, meta = classId, col sep = space] {chapters/data/mds-test-embedding-tinyNet-module3.csv};		
		\end{axis}
		\end{tikzpicture}
	}
	\subfloat[Module 5]{
		\begin{tikzpicture}
		\begin{axis}[width = 0.32 \textwidth,
		scatter/classes={
			0={mark=square*,green},
			1={mark=triangle*,red},
			2={mark=o,blue},
			3={mark=diamond*,yellow},
			4={mark=pentagon*,black},
			5={mark=square*,magenta},
			6={mark=triangle*,cyan},
			7={mark=o,gray},
			8={mark=diamond*,brown},
			9={mark=square*,darkgray},
			10={mark=pentagon*,teal}
		}]
		\addplot[scatter,only marks, scatter src=explicit symbolic] table[x  = x, y  = y, meta = classId, col sep = space] {chapters/data/mds-test-embedding-tinyNet-module4.csv};		
		\end{axis}
		\end{tikzpicture}
	}
	\vspace*{0.5cm}
	\caption{TinyNet5-8 MDS Feature visualization for the output of each Tiny module.}
	\label{sic:tinyNet58MDSVisualization}
\end{figure*}
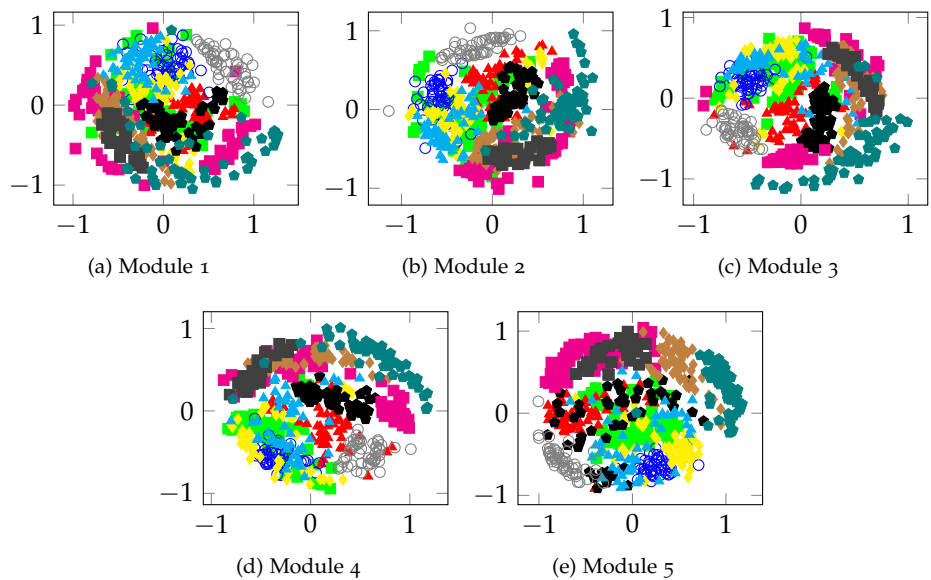

Feature visualization of TinyNet-5-8 is show in Figures \ref{sic:tinyNet58tSNEVisualization} and \ref{sic:tinyNet58MDSVisualization}. A clear pattern in shown in the t-SNE visualization, as the features from each module can cluster each class progressively better. For example features for the Background class in the first module (Figure \ref{sic:tinyNet58tSNEVisualization}a) are spread over the visualization map, which can be easily confused with other classes, like Bottle or Tire.

But features in the fifth module can cluster the Background class and separate them from other classes, which is represented in the visualization as Background having less overlap with other classes. The same pattern is visible for Propeller and Can classes.
The MDS visualization in Figure \ref{sic:tinyNet58MDSVisualization} shows a very similar pattern, showing that our conclusions from the t-SNE are not biased by the use of a single dimensionality reduction method. For example, in the first module Tire and Background overlap considerably, while in the third module the overlap is considerably reduced, and in the fifth module there is no overlap between those classes.

While using a single module produces a decent accuracy of $93.5 \%$ (from Table \ref{sic:comparisonCNTNFN}), it is possible to see in Figure \ref{sic:tinyNet58MDSVisualization}a that there is no good discrimination between classes. A considerably improved class discrimination is shown in the fourth module, but the best separation is given by the fifth module.

\begin{table*}[t]
	\begin{tabular}{lllllll}
		\hline
		Architecture	& Module 1 		& Module 2 		& Module 3 		& Module 4 	& Module 5   & Module 6\\
		\hline
		TinyNet-5-8				& (8, 48, 48) 	& (8, 24, 24)	& (8, 12, 12)	& (8, 6, 6)	& (8, 3, 3)	  & N/A\\		
		\hline
		ClassicNet-BN-5			& (32, 48, 48)  & (32, 24, 24)  & (32, 12, 12)  & (32, 6, 6) & (32, 3, 3) & (64)\\
		ClassicNet-Do-5	& (32, 48, 48)  & (32, 24, 24)  & (32, 12, 12)  & (32, 6, 6) & (32, 3, 3) & (64)\\
		\hline
	\end{tabular}
	\vspace*{0.5cm}
	\caption[Feature dimensionality of evaluated layer/module for each network]{Feature dimensionality of evaluated layer/module for each network. For TinyNet we evaluate features produced by the output of each Tiny module. For ClassicNet-BN we evaluate the output features from layers BN1-5 and FC1, while for ClassicNet-Do (Dropout) we evaluate layers MP1-5 and FC1. Shapes in this table are in format (channels, width, height).}
	\label{sic:featureDimensions}
\end{table*}

\begin{figure*}[p]
	\centering
	\begin{tikzpicture}
		\begin{customlegend}[legend columns = 7,legend style = {column sep=1ex}, legend cell align = left,
	       legend entries={Can, Bottle, Drink Carton, Chain, Propeller, Tire, Hook, Valve, Shampoo Bottle, Standing Bottle, Background}]
	       \addlegendimage{only marks, mark=square*,green}
	       \addlegendimage{only marks, mark=triangle*,red}
	       \addlegendimage{only marks, mark=o,blue}
	       \addlegendimage{only marks, mark=diamond*,yellow}
	       \addlegendimage{only marks, mark=pentagon*,black}
	       \addlegendimage{only marks, mark=square*,magenta}
	       \addlegendimage{only marks, mark=triangle*,cyan}
	       \addlegendimage{only marks, mark=o,gray}
	       \addlegendimage{only marks, mark=diamond*,brown}
	       \addlegendimage{only marks, mark=square*,darkgray}
	       \addlegendimage{only marks, mark=pentagon*,teal}
		\end{customlegend}
	\end{tikzpicture}
	
	\subfloat[Module 1]{
		\begin{tikzpicture}
			\begin{axis}[width = 0.32 \textwidth,
			    scatter/classes={
				0={mark=square*,green},
				1={mark=triangle*,red},
				2={mark=o,blue},
				3={mark=diamond*,yellow},
				4={mark=pentagon*,black},
				5={mark=square*,magenta},
				6={mark=triangle*,cyan},
				7={mark=o,gray},
				8={mark=diamond*,brown},
				9={mark=square*,darkgray},
				10={mark=pentagon*,teal}
			}]
				\addplot[scatter,only marks, scatter src=explicit symbolic] table[x  = x, y  = y, meta = classId, col sep = space] {chapters/data/tsne-test-embedding-tinyNet-module0.csv};		
			\end{axis}
		\end{tikzpicture}
	}
	\subfloat[Module 2]{
		\begin{tikzpicture}
		\begin{axis}[width = 0.32 \textwidth,
		scatter/classes={
			0={mark=square*,green},
			1={mark=triangle*,red},
			2={mark=o,blue},
			3={mark=diamond*,yellow},
			4={mark=pentagon*,black},
			5={mark=square*,magenta},
			6={mark=triangle*,cyan},
			7={mark=o,gray},
			8={mark=diamond*,brown},
			9={mark=square*,darkgray},
			10={mark=pentagon*,teal}
		}]
		\addplot[scatter,only marks, scatter src=explicit symbolic] table[x  = x, y  = y, meta = classId, col sep = space] {chapters/data/tsne-test-embedding-tinyNet-module1.csv};		
		\end{axis}
		\end{tikzpicture}
	}
	\subfloat[Module 3]{
		\begin{tikzpicture}
		\begin{axis}[width = 0.32 \textwidth,
		scatter/classes={
			0={mark=square*,green},
			1={mark=triangle*,red},
			2={mark=o,blue},
			3={mark=diamond*,yellow},
			4={mark=pentagon*,black},
			5={mark=square*,magenta},
			6={mark=triangle*,cyan},
			7={mark=o,gray},
			8={mark=diamond*,brown},
			9={mark=square*,darkgray},
			10={mark=pentagon*,teal}
		}]
		\addplot[scatter,only marks, scatter src=explicit symbolic] table[x  = x, y  = y, meta = classId, col sep = space] {chapters/data/tsne-test-embedding-tinyNet-module2.csv};		
		\end{axis}
		\end{tikzpicture}
	}
	
	\subfloat[Module 4]{
		\begin{tikzpicture}
		\begin{axis}[width = 0.32 \textwidth,
		scatter/classes={
			0={mark=square*,green},
			1={mark=triangle*,red},
			2={mark=o,blue},
			3={mark=diamond*,yellow},
			4={mark=pentagon*,black},
			5={mark=square*,magenta},
			6={mark=triangle*,cyan},
			7={mark=o,gray},
			8={mark=diamond*,brown},
			9={mark=square*,darkgray},
			10={mark=pentagon*,teal}
		}]
		\addplot[scatter,only marks, scatter src=explicit symbolic] table[x  = x, y  = y, meta = classId, col sep = space] {chapters/data/tsne-test-embedding-tinyNet-module3.csv};		
		\end{axis}
		\end{tikzpicture}
	}
	\subfloat[Module 5]{
		\begin{tikzpicture}
		\begin{axis}[width = 0.32 \textwidth,
		scatter/classes={
			0={mark=square*,green},
			1={mark=triangle*,red},
			2={mark=o,blue},
			3={mark=diamond*,yellow},
			4={mark=pentagon*,black},
			5={mark=square*,magenta},
			6={mark=triangle*,cyan},
			7={mark=o,gray},
			8={mark=diamond*,brown},
			9={mark=square*,darkgray},
			10={mark=pentagon*,teal}
		}]
		\addplot[scatter,only marks, scatter src=explicit symbolic] table[x  = x, y  = y, meta = classId, col sep = space] {chapters/data/tsne-test-embedding-tinyNet-module4.csv};		
		\end{axis}
		\end{tikzpicture}
	}
	\vspace*{0.5cm}
	\caption{TinyNet5-8 t-SNE Feature visualization for the output of each Tiny module.}
	\label{sic:tinyNet58tSNEVisualization}
\end{figure*}

\begin{figure*}[p]
	\centering
	\begin{tikzpicture}
	\begin{customlegend}[legend columns = 7,legend style = {column sep=1ex}, legend cell align = left,
	legend entries={Can, Bottle, Drink Carton, Chain, Propeller, Tire, Hook, Valve, Shampoo Bottle, Standing Bottle, Background}]
	\addlegendimage{only marks, mark=square*,green}
	\addlegendimage{only marks, mark=triangle*,red}
	\addlegendimage{only marks, mark=o,blue}
	\addlegendimage{only marks, mark=diamond*,yellow}
	\addlegendimage{only marks, mark=pentagon*,black}
	\addlegendimage{only marks, mark=square*,magenta}
	\addlegendimage{only marks, mark=triangle*,cyan}
	\addlegendimage{only marks, mark=o,gray}
	\addlegendimage{only marks, mark=diamond*,brown}
	\addlegendimage{only marks, mark=square*,darkgray}
	\addlegendimage{only marks, mark=pentagon*,teal}
	\end{customlegend}
	\end{tikzpicture}
	
	\subfloat[BN1]{
		\begin{tikzpicture}
		\begin{axis}[width = 0.32 \textwidth,
		scatter/classes={
			0={mark=square*,green},
			1={mark=triangle*,red},
			2={mark=o,blue},
			3={mark=diamond*,yellow},
			4={mark=pentagon*,black},
			5={mark=square*,magenta},
			6={mark=triangle*,cyan},
			7={mark=o,gray},
			8={mark=diamond*,brown},
			9={mark=square*,darkgray},
			10={mark=pentagon*,teal}
		}]
		\addplot[scatter,only marks, scatter src=explicit symbolic] table[x  = x, y  = y, meta = classId, col sep = space] {chapters/data/tsne-test-embedding-classicNet5BN-bn1.csv};		
		\end{axis}
		\end{tikzpicture}
	}
	\subfloat[BN2]{
		\begin{tikzpicture}
		\begin{axis}[width = 0.32 \textwidth,
		scatter/classes={
			0={mark=square*,green},
			1={mark=triangle*,red},
			2={mark=o,blue},
			3={mark=diamond*,yellow},
			4={mark=pentagon*,black},
			5={mark=square*,magenta},
			6={mark=triangle*,cyan},
			7={mark=o,gray},
			8={mark=diamond*,brown},
			9={mark=square*,darkgray},
			10={mark=pentagon*,teal}
		}]
		\addplot[scatter,only marks, scatter src=explicit symbolic] table[x  = x, y  = y, meta = classId, col sep = space] {chapters/data/tsne-test-embedding-classicNet5BN-bn2.csv};		
		\end{axis}
		\end{tikzpicture}
	}
	\subfloat[BN3]{
		\begin{tikzpicture}
		\begin{axis}[width = 0.32 \textwidth,
		scatter/classes={
			0={mark=square*,green},
			1={mark=triangle*,red},
			2={mark=o,blue},
			3={mark=diamond*,yellow},
			4={mark=pentagon*,black},
			5={mark=square*,magenta},
			6={mark=triangle*,cyan},
			7={mark=o,gray},
			8={mark=diamond*,brown},
			9={mark=square*,darkgray},
			10={mark=pentagon*,teal}
		}]
		\addplot[scatter,only marks, scatter src=explicit symbolic] table[x  = x, y  = y, meta = classId, col sep = space] {chapters/data/tsne-test-embedding-classicNet5BN-bn3.csv};		
		\end{axis}
		\end{tikzpicture}
	}

	\subfloat[BN4]{
		\begin{tikzpicture}
		\begin{axis}[width = 0.32 \textwidth,
		scatter/classes={
			0={mark=square*,green},
			1={mark=triangle*,red},
			2={mark=o,blue},
			3={mark=diamond*,yellow},
			4={mark=pentagon*,black},
			5={mark=square*,magenta},
			6={mark=triangle*,cyan},
			7={mark=o,gray},
			8={mark=diamond*,brown},
			9={mark=square*,darkgray},
			10={mark=pentagon*,teal}
		}]
		\addplot[scatter,only marks, scatter src=explicit symbolic] table[x  = x, y  = y, meta = classId, col sep = space] {chapters/data/tsne-test-embedding-classicNet5BN-bn4.csv};		
		\end{axis}
		\end{tikzpicture}
	}
	\subfloat[BN5]{
		\begin{tikzpicture}
		\begin{axis}[width = 0.32 \textwidth,
		scatter/classes={
			0={mark=square*,green},
			1={mark=triangle*,red},
			2={mark=o,blue},
			3={mark=diamond*,yellow},
			4={mark=pentagon*,black},
			5={mark=square*,magenta},
			6={mark=triangle*,cyan},
			7={mark=o,gray},
			8={mark=diamond*,brown},
			9={mark=square*,darkgray},
			10={mark=pentagon*,teal}
		}]
		\addplot[scatter,only marks, scatter src=explicit symbolic] table[x  = x, y  = y, meta = classId, col sep = space] {chapters/data/tsne-test-embedding-classicNet5BN-bn5.csv};		
		\end{axis}
		\end{tikzpicture}
	}
	\subfloat[FC1]{
		\begin{tikzpicture}
		\begin{axis}[width = 0.32 \textwidth,
		scatter/classes={
			0={mark=square*,green},
			1={mark=triangle*,red},
			2={mark=o,blue},
			3={mark=diamond*,yellow},
			4={mark=pentagon*,black},
			5={mark=square*,magenta},
			6={mark=triangle*,cyan},
			7={mark=o,gray},
			8={mark=diamond*,brown},
			9={mark=square*,darkgray},
			10={mark=pentagon*,teal}
		}]
		\addplot[scatter,only marks, scatter src=explicit symbolic] table[x  = x, y  = y, meta = classId, col sep = space] {chapters/data/tsne-test-embedding-classicNet5BN-fc1.csv};		
		\end{axis}
		\end{tikzpicture}
	}
	\vspace*{0.5cm}
	\caption[ClassicNet-BN-5 t-SNE Feature visualization]{ClassicNet-BN-5 t-SNE Feature visualization for each layer.}
	\label{sic:classicNetBN5tSNEVisualization}
\end{figure*}
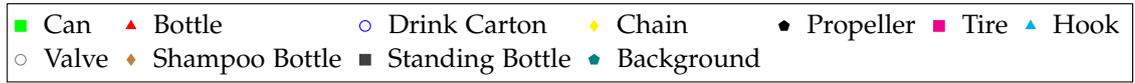
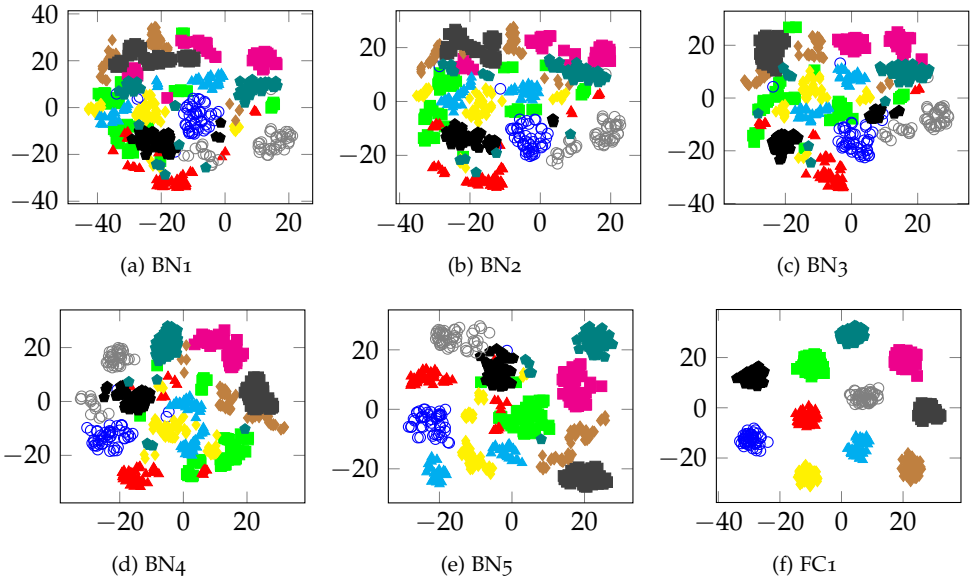

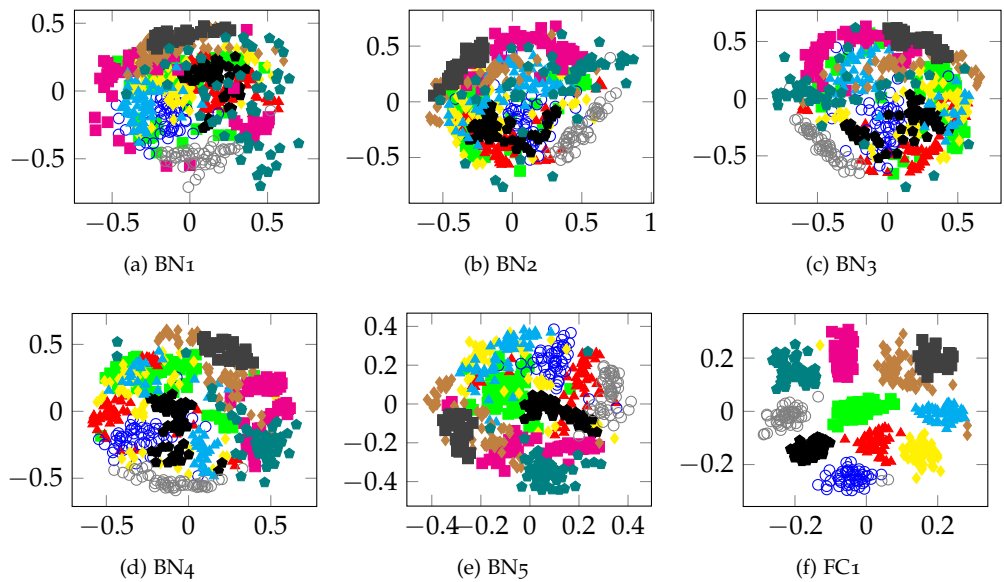
\begin{figure*}[p]
	\centering
	\subfloat[BN1]{
		\begin{tikzpicture}
		\begin{axis}[width = 0.32 \textwidth,
		scatter/classes={
			0={mark=square*,green},
			1={mark=triangle*,red},
			2={mark=o,blue},
			3={mark=diamond*,yellow},
			4={mark=pentagon*,black},
			5={mark=square*,magenta},
			6={mark=triangle*,cyan},
			7={mark=o,gray},
			8={mark=diamond*,brown},
			9={mark=square*,darkgray},
			10={mark=pentagon*,teal}
		}]
		\addplot[scatter,only marks, scatter src=explicit symbolic] table[x  = x, y  = y, meta = classId, col sep = space] {chapters/data/mds-test-embedding-classicNet5BN-bn1.csv};		
		\end{axis}
		\end{tikzpicture}
	}
	\subfloat[BN2]{
		\begin{tikzpicture}
		\begin{axis}[width = 0.32 \textwidth,
		scatter/classes={
			0={mark=square*,green},
			1={mark=triangle*,red},
			2={mark=o,blue},
			3={mark=diamond*,yellow},
			4={mark=pentagon*,black},
			5={mark=square*,magenta},
			6={mark=triangle*,cyan},
			7={mark=o,gray},
			8={mark=diamond*,brown},
			9={mark=square*,darkgray},
			10={mark=pentagon*,teal}
		}]
		\addplot[scatter,only marks, scatter src=explicit symbolic] table[x  = x, y  = y, meta = classId, col sep = space] {chapters/data/mds-test-embedding-classicNet5BN-bn2.csv};		
		\end{axis}
		\end{tikzpicture}
	}
	\subfloat[BN3]{
		\begin{tikzpicture}
		\begin{axis}[width = 0.32 \textwidth,
		scatter/classes={
			0={mark=square*,green},
			1={mark=triangle*,red},
			2={mark=o,blue},
			3={mark=diamond*,yellow},
			4={mark=pentagon*,black},
			5={mark=square*,magenta},
			6={mark=triangle*,cyan},
			7={mark=o,gray},
			8={mark=diamond*,brown},
			9={mark=square*,darkgray},
			10={mark=pentagon*,teal}
		}]
		\addplot[scatter,only marks, scatter src=explicit symbolic] table[x  = x, y  = y, meta = classId, col sep = space] {chapters/data/mds-test-embedding-classicNet5BN-bn3.csv};		
		\end{axis}
		\end{tikzpicture}
	}
	
	\subfloat[BN4]{
		\begin{tikzpicture}
		\begin{axis}[width = 0.32 \textwidth,
		scatter/classes={
			0={mark=square*,green},
			1={mark=triangle*,red},
			2={mark=o,blue},
			3={mark=diamond*,yellow},
			4={mark=pentagon*,black},
			5={mark=square*,magenta},
			6={mark=triangle*,cyan},
			7={mark=o,gray},
			8={mark=diamond*,brown},
			9={mark=square*,darkgray},
			10={mark=pentagon*,teal}
		}]
		\addplot[scatter,only marks, scatter src=explicit symbolic] table[x  = x, y  = y, meta = classId, col sep = space] {chapters/data/mds-test-embedding-classicNet5BN-bn4.csv};		
		\end{axis}
		\end{tikzpicture}
	}
	\subfloat[BN5]{
		\begin{tikzpicture}
		\begin{axis}[width = 0.32 \textwidth,
		scatter/classes={
			0={mark=square*,green},
			1={mark=triangle*,red},
			2={mark=o,blue},
			3={mark=diamond*,yellow},
			4={mark=pentagon*,black},
			5={mark=square*,magenta},
			6={mark=triangle*,cyan},
			7={mark=o,gray},
			8={mark=diamond*,brown},
			9={mark=square*,darkgray},
			10={mark=pentagon*,teal}
		}]
		\addplot[scatter,only marks, scatter src=explicit symbolic] table[x  = x, y  = y, meta = classId, col sep = space] {chapters/data/mds-test-embedding-classicNet5BN-bn5.csv};		
		\end{axis}
		\end{tikzpicture}
	}
	\subfloat[FC1]{
		\begin{tikzpicture}
		\begin{axis}[width = 0.32 \textwidth,
		scatter/classes={
			0={mark=square*,green},
			1={mark=triangle*,red},
			2={mark=o,blue},
			3={mark=diamond*,yellow},
			4={mark=pentagon*,black},
			5={mark=square*,magenta},
			6={mark=triangle*,cyan},
			7={mark=o,gray},
			8={mark=diamond*,brown},
			9={mark=square*,darkgray},
			10={mark=pentagon*,teal}
		}]
		\addplot[scatter,only marks, scatter src=explicit symbolic] table[x  = x, y  = y, meta = classId, col sep = space] {chapters/data/mds-test-embedding-classicNet5BN-fc1.csv};		
		\end{axis}
		\end{tikzpicture}
	}
	\vspace*{0.5cm}
	\caption[ClassicNet-BN-5 MDS Feature visualization]{ClassicNet-BN-5 MDS Feature visualization for each layer.}
	\label{sic:classicNetBN5MDSVisualization}
\end{figure*}

Visualization of ClassicNet-BN features is shown in Figure \ref{sic:classicNetBN5tSNEVisualization} and Figure \ref{sic:classicNetBN5MDSVisualization}. The t-SNE visualization shows again a progressive improvement of the feature discrimination capabilities. At the first layer (BN1) there is little separation between classes, with many of them overlapping. But at the fourth layer (BN4) discrimination has been considerably improved. An interesting effect can be seen from the transition between BN5 and FC1, as the discriminatory power of the features is considerably increased, shown by the almost perfect separation of the classes. As this is right before applying the last fully connected layer with a softmax activation, it should be of no surprise that this network can achieve such high accuracy (over $99 \%$).

The MDS visualizations \ref{sic:classicNetBN5MDSVisualization} show the same pattern of increasing improvements of feature discrimination, with a considerably jump between BN5 and FC1 layers. The MDS visualization shows a better class separation in layer FC1, as many classes are almost perfectly clustered (Drink Carton, Bottle, Valve, etc). One class that shows a considerable difference between t-SNE and MDS visualizations is Background on the FC1 layer. t-SNE shows a good clustering on these features, but MDS shows overlap between Background and other classes like Tire and Chain. We do not know if this is a visualization artifact or a underlying property of the features, as the Background class is "special" as it is implicitly present in all classes as the background of each object.

Visualization of ClassicNet-Dropout features is shown in Figure \ref{sic:classicNetDropout5tSNEVisualization} and \ref{sic:classicNetDropout5MDSVisualization}. Both visualizations shows the power of using Dropout, as the learned features are considerably different. For example, a good initial discrimination is shown in the MP1 layer, but many classes still considerably overlap. Discrimination is only improved from the fifth layer (MP5).
Dropout features in the FC1 layer show a good separation, which is evident from the MDS visualization.

\begin{figure*}[p]
	\centering
	
	\begin{tikzpicture}
	\begin{customlegend}[legend columns = 7,legend style = {column sep=1ex}, legend cell align = left,
	legend entries={Can, Bottle, Drink Carton, Chain, Propeller, Tire, Hook, Valve, Shampoo Bottle, Standing Bottle, Background}]
	\addlegendimage{only marks, mark=square*,green}
	\addlegendimage{only marks, mark=triangle*,red}
	\addlegendimage{only marks, mark=o,blue}
	\addlegendimage{only marks, mark=diamond*,yellow}
	\addlegendimage{only marks, mark=pentagon*,black}
	\addlegendimage{only marks, mark=square*,magenta}
	\addlegendimage{only marks, mark=triangle*,cyan}
	\addlegendimage{only marks, mark=o,gray}
	\addlegendimage{only marks, mark=diamond*,brown}
	\addlegendimage{only marks, mark=square*,darkgray}
	\addlegendimage{only marks, mark=pentagon*,teal}
	\end{customlegend}
	\end{tikzpicture}
	
	\subfloat[MP1]{
		\begin{tikzpicture}
		\begin{axis}[width = 0.32 \textwidth,
		scatter/classes={
			0={mark=square*,green},
			1={mark=triangle*,red},
			2={mark=o,blue},
			3={mark=diamond*,yellow},
			4={mark=pentagon*,black},
			5={mark=square*,magenta},
			6={mark=triangle*,cyan},
			7={mark=o,gray},
			8={mark=diamond*,brown},
			9={mark=square*,darkgray},
			10={mark=pentagon*,teal}
		}]
		\addplot[scatter,only marks, scatter src=explicit symbolic] table[x  = x, y  = y, meta = classId, col sep = space] {chapters/data/tsne-test-embedding-classicNet5DO-mp1.csv};		
		\end{axis}
		\end{tikzpicture}
	}
	\subfloat[MP2]{
		\begin{tikzpicture}
		\begin{axis}[width = 0.32 \textwidth,
		scatter/classes={
			0={mark=square*,green},
			1={mark=triangle*,red},
			2={mark=o,blue},
			3={mark=diamond*,yellow},
			4={mark=pentagon*,black},
			5={mark=square*,magenta},
			6={mark=triangle*,cyan},
			7={mark=o,gray},
			8={mark=diamond*,brown},
			9={mark=square*,darkgray},
			10={mark=pentagon*,teal}
		}]
		\addplot[scatter,only marks, scatter src=explicit symbolic] table[x  = x, y  = y, meta = classId, col sep = space] {chapters/data/tsne-test-embedding-classicNet5DO-mp2.csv};		
		\end{axis}
		\end{tikzpicture}
	}
	\subfloat[MP3]{
		\begin{tikzpicture}
		\begin{axis}[width = 0.32 \textwidth,
		scatter/classes={
			0={mark=square*,green},
			1={mark=triangle*,red},
			2={mark=o,blue},
			3={mark=diamond*,yellow},
			4={mark=pentagon*,black},
			5={mark=square*,magenta},
			6={mark=triangle*,cyan},
			7={mark=o,gray},
			8={mark=diamond*,brown},
			9={mark=square*,darkgray},
			10={mark=pentagon*,teal}
		}]
		\addplot[scatter,only marks, scatter src=explicit symbolic] table[x  = x, y  = y, meta = classId, col sep = space] {chapters/data/tsne-test-embedding-classicNet5DO-mp3.csv};		
		\end{axis}
		\end{tikzpicture}
	}
	
	\subfloat[MP4]{
		\begin{tikzpicture}
		\begin{axis}[width = 0.32 \textwidth,
		scatter/classes={
			0={mark=square*,green},
			1={mark=triangle*,red},
			2={mark=o,blue},
			3={mark=diamond*,yellow},
			4={mark=pentagon*,black},
			5={mark=square*,magenta},
			6={mark=triangle*,cyan},
			7={mark=o,gray},
			8={mark=diamond*,brown},
			9={mark=square*,darkgray},
			10={mark=pentagon*,teal}
		}]
		\addplot[scatter,only marks, scatter src=explicit symbolic] table[x  = x, y  = y, meta = classId, col sep = space] {chapters/data/tsne-test-embedding-classicNet5DO-mp4.csv};		
		\end{axis}
		\end{tikzpicture}
	}
	\subfloat[MP5]{
		\begin{tikzpicture}
		\begin{axis}[width = 0.32 \textwidth,
		scatter/classes={
			0={mark=square*,green},
			1={mark=triangle*,red},
			2={mark=o,blue},
			3={mark=diamond*,yellow},
			4={mark=pentagon*,black},
			5={mark=square*,magenta},
			6={mark=triangle*,cyan},
			7={mark=o,gray},
			8={mark=diamond*,brown},
			9={mark=square*,darkgray},
			10={mark=pentagon*,teal}
		}]
		\addplot[scatter,only marks, scatter src=explicit symbolic] table[x  = x, y  = y, meta = classId, col sep = space] {chapters/data/tsne-test-embedding-classicNet5DO-mp5.csv};		
		\end{axis}
		\end{tikzpicture}
	}
	\subfloat[FC1]{
		\begin{tikzpicture}
		\begin{axis}[width = 0.32 \textwidth,
		scatter/classes={
			0={mark=square*,green},
			1={mark=triangle*,red},
			2={mark=o,blue},
			3={mark=diamond*,yellow},
			4={mark=pentagon*,black},
			5={mark=square*,magenta},
			6={mark=triangle*,cyan},
			7={mark=o,gray},
			8={mark=diamond*,brown},
			9={mark=square*,darkgray},
			10={mark=pentagon*,teal}
		}]
		\addplot[scatter,only marks, scatter src=explicit symbolic] table[x  = x, y  = y, meta = classId, col sep = space] {chapters/data/tsne-test-embedding-classicNet5DO-fc1.csv};		
		\end{axis}
		\end{tikzpicture}
	}
	\vspace*{0.5cm}
	\caption[ClassicNet-Dropout-5 t-SNE Feature visualization]{ClassicNet-Dropout-5 t-SNE Feature visualization for each layer.}
	\label{sic:classicNetDropout5tSNEVisualization}
\end{figure*}

\begin{figure*}[p]
	\centering
	\subfloat[MP1]{
		\begin{tikzpicture}
		\begin{axis}[width = 0.32 \textwidth,
		scatter/classes={
			0={mark=square*,green},
			1={mark=triangle*,red},
			2={mark=o,blue},
			3={mark=diamond*,yellow},
			4={mark=pentagon*,black},
			5={mark=square*,magenta},
			6={mark=triangle*,cyan},
			7={mark=o,gray},
			8={mark=diamond*,brown},
			9={mark=square*,darkgray},
			10={mark=pentagon*,teal}
		}]
		\addplot[scatter,only marks, scatter src=explicit symbolic] table[x  = x, y  = y, meta = classId, col sep = space] {chapters/data/mds-test-embedding-classicNet5DO-mp1.csv};		
		\end{axis}
		\end{tikzpicture}
	}
	\subfloat[MP2]{
		\begin{tikzpicture}
		\begin{axis}[width = 0.32 \textwidth,
		scatter/classes={
			0={mark=square*,green},
			1={mark=triangle*,red},
			2={mark=o,blue},
			3={mark=diamond*,yellow},
			4={mark=pentagon*,black},
			5={mark=square*,magenta},
			6={mark=triangle*,cyan},
			7={mark=o,gray},
			8={mark=diamond*,brown},
			9={mark=square*,darkgray},
			10={mark=pentagon*,teal}
		}]
		\addplot[scatter,only marks, scatter src=explicit symbolic] table[x  = x, y  = y, meta = classId, col sep = space] {chapters/data/mds-test-embedding-classicNet5DO-mp2.csv};		
		\end{axis}
		\end{tikzpicture}
	}
	\subfloat[MP3]{
		\begin{tikzpicture}
		\begin{axis}[width = 0.32 \textwidth,
		scatter/classes={
			0={mark=square*,green},
			1={mark=triangle*,red},
			2={mark=o,blue},
			3={mark=diamond*,yellow},
			4={mark=pentagon*,black},
			5={mark=square*,magenta},
			6={mark=triangle*,cyan},
			7={mark=o,gray},
			8={mark=diamond*,brown},
			9={mark=square*,darkgray},
			10={mark=pentagon*,teal}
		}]
		\addplot[scatter,only marks, scatter src=explicit symbolic] table[x  = x, y  = y, meta = classId, col sep = space] {chapters/data/mds-test-embedding-classicNet5DO-mp3.csv};		
		\end{axis}
		\end{tikzpicture}
	}
	
	\subfloat[MP4]{
		\begin{tikzpicture}
		\begin{axis}[width = 0.32 \textwidth,
		scatter/classes={
			0={mark=square*,green},
			1={mark=triangle*,red},
			2={mark=o,blue},
			3={mark=diamond*,yellow},
			4={mark=pentagon*,black},
			5={mark=square*,magenta},
			6={mark=triangle*,cyan},
			7={mark=o,gray},
			8={mark=diamond*,brown},
			9={mark=square*,darkgray},
			10={mark=pentagon*,teal}
		}]
		\addplot[scatter,only marks, scatter src=explicit symbolic] table[x  = x, y  = y, meta = classId, col sep = space] {chapters/data/mds-test-embedding-classicNet5DO-mp4.csv};		
		\end{axis}
		\end{tikzpicture}
	}
	\subfloat[MP5]{
		\begin{tikzpicture}
		\begin{axis}[width = 0.32 \textwidth,
		scatter/classes={
			0={mark=square*,green},
			1={mark=triangle*,red},
			2={mark=o,blue},
			3={mark=diamond*,yellow},
			4={mark=pentagon*,black},
			5={mark=square*,magenta},
			6={mark=triangle*,cyan},
			7={mark=o,gray},
			8={mark=diamond*,brown},
			9={mark=square*,darkgray},
			10={mark=pentagon*,teal}
		}]
		\addplot[scatter,only marks, scatter src=explicit symbolic] table[x  = x, y  = y, meta = classId, col sep = space] {chapters/data/mds-test-embedding-classicNet5DO-mp5.csv};		
		\end{axis}
		\end{tikzpicture}
	}
	\subfloat[FC1]{
		\begin{tikzpicture}
		\begin{axis}[width = 0.32 \textwidth,
		scatter/classes={
			0={mark=square*,green},
			1={mark=triangle*,red},
			2={mark=o,blue},
			3={mark=diamond*,yellow},
			4={mark=pentagon*,black},
			5={mark=square*,magenta},
			6={mark=triangle*,cyan},
			7={mark=o,gray},
			8={mark=diamond*,brown},
			9={mark=square*,darkgray},
			10={mark=pentagon*,teal}
		}]
		\addplot[scatter,only marks, scatter src=explicit symbolic] table[x  = x, y  = y, meta = classId, col sep = space] {chapters/data/mds-test-embedding-classicNet5DO-fc1.csv};		
		\end{axis}
		\end{tikzpicture}
	}
	\vspace*{0.5cm}
	\caption[ClassicNet-Dropout-5 MDS Feature visualization]{ClassicNet-Dropout-5 MDS Feature visualization for each layer.}
	\label{sic:classicNetDropout5MDSVisualization}
\end{figure*}

\subsection{Computational Performance}
\label{sic:compPerformanceSection}

In this section we evaluate the computational performance of our models. While most research papers only consider classification performance as a metric, computational performance is also an important component of any real system, specially when using these kind of algorithms in platforms that must be used in everyday life, like robots.

We use two platforms for evaluation. One is a x86\_64 platform consisting of an AMD Ryzen 7 1700 processor (3.0 GHz) with 16 GB of RAM. This represents the most common use platforms using Intel and AMD processors without hard limits on power or heat. This platform is named "High-Power" (HP). The second platform is a Raspberry Pi 2, containing an 900 MHz ARM Cortex A7 processor with one GB of RAM. This represents a low-power embedded system that is more typical to what is used inside a robot. This platform is denominated "Low-Power" (LP).

Both platforms are radically different. The Thermal Design Power of the Ryzen processor is 65 Watts, while the Raspberry Pi 2 only consumes up to 5 Watts. Components other than the CPU will also affect performance and power consumption, like the use of DDR/DDR4 memory.

Performance is measured using the python \textit{timeit} module after running the prediction code 100 times, and computing mean and standard deviation of computation time. We used Theano 0.9.0 with Keras 1.0.0 on python 3.6 in both platforms with GCC 7.1.1 as compiler.

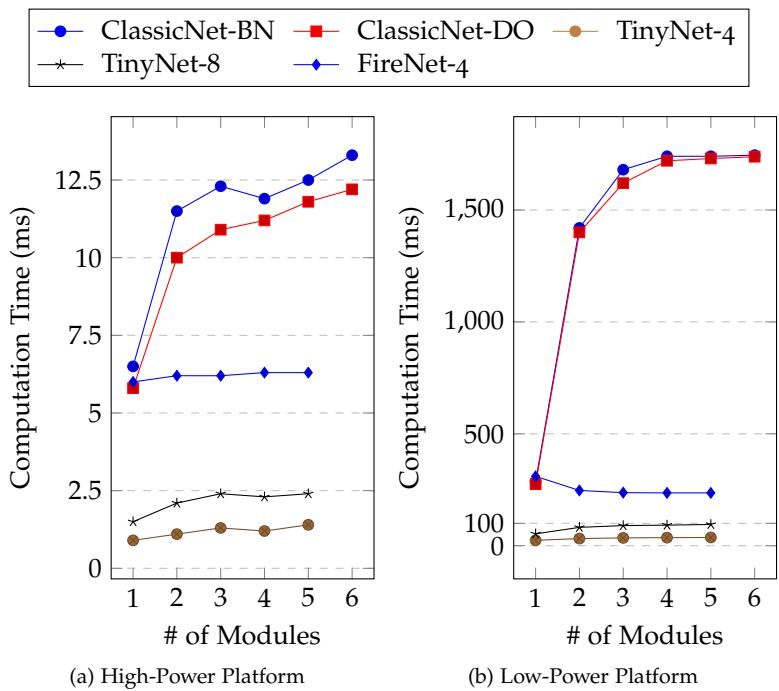
\begin{figure}[h]
	\centering
	\begin{tikzpicture}
		\begin{customlegend}[legend columns = 3,legend style = {column sep=1ex}, legend cell align = left,
			legend entries={ClassicNet-BN, ClassicNet-DO, TinyNet-4, TinyNet-8, FireNet-4}]
			\addlegendimage{mark=*,blue}
			\addlegendimage{mark=square*,red}
			\addlegendimage{mark=otimes*,brown}
			\addlegendimage{mark=star,black}
			\addlegendimage{mark=diamond*,blue}
		\end{customlegend}
	\end{tikzpicture}
	\subfloat[High-Power Platform] {
		\centering
		\begin{tikzpicture}
		\begin{axis}[
		height = 0.3 \textheight,
		xtick = {1,2,3,4,5,6},
		ytick = {0,2.5,5,7.5,10,12.5},
		xlabel={\# of Modules},
		ylabel={Computation Time (ms)},
		ymajorgrids=true,
		grid style=dashed,
		width = 0.48\textwidth]
		\addplot coordinates { (1, 6.5) (2, 11.5) (3, 12.3) (4, 11.9) (5, 12.5) (6, 13.3) };
		\addplot coordinates { (1, 5.8) (2, 10.0) (3, 10.9) (4, 11.2) (5, 11.8) (6, 12.2) };
		\addplot coordinates { (1, 0.9) (2, 1.1) (3, 1.3) (4, 1.2) (5, 1.4) };
		\addplot coordinates { (1, 1.5) (2, 2.1) (3, 2.4) (4, 2.3) (5, 2.4) };
		\addplot coordinates { (1, 6.0) (2, 6.2) (3, 6.2) (4, 6.3) (5, 6.3) };
		
		\end{axis}
		\end{tikzpicture}
	}
	\subfloat[Low-Power Platform] {
		\centering
		\begin{tikzpicture}
		\begin{axis}[
		height = 0.3 \textheight,
		xtick = {1,2,3,4,5,6},
		ytick = {0,100,500, 1000, 1500},
		xlabel={\# of Modules},
		ylabel={Computation Time (ms)},
		ymajorgrids=true,
		grid style=dashed,
		width = 0.48\textwidth]
		
		\addplot coordinates { (1, 286.0) (2, 1420.0) (3, 1680.0) (4, 1740.0) (5, 1740.0) (6, 1745.0) };
		\addplot coordinates { (1, 275.0) (2, 1400.0) (3, 1620.0) (4, 1720.0) (5, 1730.0) (6, 1738.0) };
		\addplot coordinates { (1, 24.0) (2, 32.0) (3, 35.0) (4, 36.0) (5, 37.0) };
		\addplot coordinates { (1, 53.0) (2, 82.0) (3, 90.0) (4, 92.0) (5, 95.0) };
		\addplot coordinates { (1, 310.0) (2, 247.0) (3, 237.0) (4, 236.0) (5, 236.0) };
		
		\end{axis}
		\end{tikzpicture}
	}
	\vspace*{0.5cm}
	\caption{Computation Time comparison for the HP and LP platforms on our tested networks.}
	\label{sic:computationTimePlatformComparison}
\end{figure}

Results are shown in Table \ref{sic:classicNetPerformanceVsModules} for ClassicNet and Table \ref{sic:tinyFireNetPerformanceVsModules} for TinyNet and FireNet. A visual comparison of these results is available in Figure \ref{sic:computationTimePlatformComparison}.

High-Power platform results show that all networks are quite fast, with a maximum computation time of $12$ milliseconds per image, which is able to run in real-time at 83 frames per second. Still TinyNet is considerably faster than the other networks, at less than $2.5$ milliseconds per frame (over 400 frames per second), which is 5 times faster. FireNet as well is at 2-6 times slower than TinyNet, depending on the number of modules.

Low-Power platform results are more interesting, as ClassicNet is quite slow in this platform, peaking at $1740$ milliseconds per frame, while TinyNet-4 and TinyNet-8 only taking less than $40$ and $100$ milliseconds per frame, correspondingly. TinyNet-4 is 43 times faster than ClassicNet while TinyNet-8 is $17$ times faster. As seen previously, using less filters in ClassicNet considerably degrades classification performance, specially when using a small number of modules. FireNet is also more accuracy than ClassicNet, and considerably faster at $310$ milliseconds per frame (5-times speedup).

TinyNet and FireNet are quite similar, with only FireNet containing extra $3 \times 3$ filters

Selecting a network to run on a low-power embedded device is now quite easy, as TinyNet is only $0.5 \%$ less accurate than ClassicNet and $1.0 \%$ less than FireNet, but many times faster. TinyNet can run at $25$ frames per second on a Raspberry Pi 2.

There is a measurable difference between using Batch Normalization and Dropout, specially in the Low-Power platform

It must also be pointed out that comparing results in this section with Table \ref{sic:tmAccVsTSPC}, which shows computation time on the HP platform, it can be seen that template matching is not an option either, as it is slightly less accurate than TinyNet, but considerably slower. TinyNet is 24 times faster than template matching with SQD, and 67 times faster than CC. In the Low-Power platform with 150 templates per class, a CC template matcher takes $3150$ milliseconds to classify one image, a SQD template matcher takes $1200$ milliseconds per image.

These results form the core argument that for sonar image classification, template matching or classic machine learning should not be used, and convolutional neural networks should be preferred. A CNN can be more accurate and perform faster, both in high and low power hardware, which is appropriate for use in autonomous underwater vehicles.

\begin{table}[t]
	\centering
	\begin{tabular}{lllllll}
		\hline
		& \multicolumn{3}{c}{ClassicNet-BN} & \multicolumn{3}{c}{ClassicNet-Dropout}\\
		& \multicolumn{3}{c}{32 Filters} & \multicolumn{3}{c}{32 Filters}\\
		\#   	& P & LP Time & HP Time & P & LP Time & HP Time\\
		\hline 
		1 		& 4.7 M  & 286 ms  & 6.5 ms  & 4.7 M  & 275 ms  &  5.8 ms\\ 
		2  		& 1.2 M  & 1420 ms & 11.5 ms & 1.2 M  & 1400 ms & 10.0 ms\\ 
		3 		& 348 K  & 1680 ms & 12.3 ms & 348 K  & 1620 ms & 10.9 ms\\
		4 		& 152 K	 & 1740 ms & 11.9 ms & 152 K  & 1720 ms & 11.2 ms\\
		5 		& 123 K  & 1740 ms & 12.5 ms & 122 K  & 1730 ms & 11.8 ms\\
		6		& 132 K	 & 1745 ms & 13.3 ms & 131 K  & 1738 ms	& 12.2 ms\\
		\hline 
	\end{tabular}
	\vspace*{0.5cm}
	\caption[ClassicNet performance as function of number of modules and convolution filters]{ClassicNet performance as function of number of modules (\#) and convolution filters. Mean computation time is shown, both in the High-Power platform (HP) and Low-Power platform (LP). Standard deviation is not shown as it is less than 0.4 ms for HP and 4 ms for LP. The number of parameters (P) in each model is also shown.}
	\label{sic:classicNetPerformanceVsModules}. 
\end{table}

\begin{table*}[t]
    \forcerectofloat
	\centering
	\begin{tabular}{llllllllll}
		\hline
		& \multicolumn{6}{c}{TinyNet} & \multicolumn{3}{c}{FireNet}\\
		& \multicolumn{3}{c}{4 Filters} & \multicolumn{3}{c}{8 Filters} & \multicolumn{3}{c}{4 Filters}\\
		\#   	& P & LP Time & HP Time & P & LP Time & HP Time & P & LP Time & HP Time\\
		\hline 
		1 		& 307 	 & 24 ms & 0.9 ms & 443  & 53 ms  & 1.5 ms & 3499   & 310 ms & 6.0 ms\\ 
		2  		& 859    & 32 ms & 1.1 ms & 1483 & 82 ms  & 2.1 ms & 8539   & 247 ms & 6.2 ms\\ 
		3 		& 1123   & 35 ms & 1.3 ms & 2235 & 90 ms  & 2.4 ms & 10195  & 237 ms & 6.2 ms\\
		4 		& 1339	 & 36 ms & 1.2 ms & 2939 & 92 ms  & 2.3 ms & 11815  & 236 ms & 6.3 ms\\
		5 		& 1531   & 37 ms & 1.4 ms & 3619 & 95 ms  & 2.4 ms & 13415  & 236 ms & 6.3 ms\\
		6 		& 1711   & 38 ms & 1.4 ms & 4287 & 96 ms  & 2.4 ms & 15003  & 237 ms & 6.5 ms\\
		\hline 
	\end{tabular}
	\vspace*{0.5cm}
	\caption[TinyNet and FireNet performance as function of number of modules and convolution filters]{TinyNet and FireNet performance as function of number of modules (\#) and convolution filters. Mean computation time is shown, both in the High-Power platform (HP) and Low-Power platform (LP). Standard deviation is not shown as it is less than 0.2 ms for HP and 1 ms for LP. The number of parameters (P) in each model is also shown.}
	\label{sic:tinyFireNetPerformanceVsModules}. 
\end{table*}

\newpage
\section{Summary of Results}

This chapter has performed a in-depth evaluation of image classification algorithms for sonar image classification. While the state of the art typically uses template matching with cross-correlation \cite[4em]{hurtos2013automatic}, we have shown that a convolutional neural network can outperform the use of template matching, both in classification and computational performance.

We have shown that in our dataset of marine debris objects, template matching requires a large number of templates per class, which indicates that it has trouble modelling the variability in our dataset. Almost all the training set needs to be memorized in order to generalize to the validation and test sets. This indicates a high chance of overfitting.

In contrast, a CNN with only 930 K parameters is able to provide better generalization than a template matching classifier. We have also shown that other classic classification algorithms, like a support vector machine, gradient boosting and random forests, do not perform at the same level than a CNN.

We evaluated the features learned by three different convolutional networks, and we studied the learned features by using dimensionality reduction through t-SNE and Multi-Dimensional Scaling. Our results show that the network is not performing randomly and a pattern in each latter appears, where the features allow for increasing separation of classes. It also seems that the use of a fully connected layer is fundamental for class separability, as features produced by this layer cluster very well into the different classes.

We have also shown that a carefully designed CNN, based on designs similar to SqueezeNet \cite[-4em]{iandola2016squeezenet}, can perform at real-time performace on a low-power embedded system (a Raspberry Pi 2), while only sacrificing $0.5-1.0 \%$ accuracy with respect to the baseline.

We expect that in the future CNNs will be used for different tasks involving sonar image classification, as their predictive and computational performance is on par to what is required in the marine robotics domain.

%% file: chapters/limits-neural-networks.tex
\chapter[Limits of Convolutional Neural Networks on FLS Images]{Limits of Convolutional \newline Neural Networks on FLS Images}
\label{chapter:limits}

This chapter deals with a less applied problem than the rest of this thesis. While the use of Deep Neural Networks for different tasks has exploded during the last 5 years, many questions of practical importance remain unanswered \cite{zhang2016understanding}.

A general rule of thumb in machine learning is that more data always improves the generalization ability of a classifier or regressor. For Deep Learning it has been largely assumed that a large dataset is required. But experiments done in this thesis, specially in the previous chapter, show that CNNs can be trained with smaller datasets.

A big problem is then defining what is a "large" or a "small" dataset. In the computer vision community, a large dataset might be ImageNet \cite{russakovsky2015imagenet} with 1.2 million images and 1000 labeled classes, while for the marine robotics community 2000 images might be large, and 200 images with 2-4 classes might be small.

A very common requirement by practitioners is an estimate of how many data points are required to solve a given learning task. Typically as data needs to be gathered, an estimate of how much data is needed would be quite useful. Other effects are also of interest, such as what is the optimal object size for recognition, and what are the best ways to perform transfer learning, and how does that affect the training set size that is required for a given performance target.

In this chapter we would like to explore the following research questions:

\begin{itemize}
	\item How does object size affect classification performance? Can small objects be recognized similarly to bigger ones?
	\item How much data is required for FLS classification?
	\item How can networks be made to require less training data given a classification performance target.
	\item How effective is transfer learning in features learned from sonar images?
\end{itemize}

We try to answer these questions from an experimental point of view, by performing different synthetic experiments on our dataset of marine debris objects.

\section{Related Work}

Surprisingly, there is little literature about these research questions, even as they are quite important for practitioners.

Sharif et al. \cite[-3em]{sharif2014cnn} was one of the first to evaluate features learned by a CNN. They used a pre-trained OverFeat \cite[1em]{sermanet2013overfeat} as a feature extractor network, and computed features from the fc1 \footnote[][1em]{First fully connected layer} layer, which corresponds to a $4096$ long vector. This vector is then normalized with the L2 norm and used to train a multi-class SVM. An additional combination using data augmentation is also explored by the authors.

On the PASCAL VOC 2007 dataset, this method obtains better accuracy for 14 out of 21 classes, with a slightly worse accuracy on 7 classes. Note that the state of the art compared in this work is mostly composed of engineered features, like bag of visual words, clustering, and dictionary learning from HoG, SIFT and LBP features. Even as OverFeat is not trained on the same dataset, its features generalize outside of this set quite well.

On the MIT 67 indoor scenes dataset, the authors obtain $69.0$ \% mean accuracy with data augmentation, which is $5$ \% better than the state of the art. This dataset is considerably different from the ImageNet dataset used to train OverFeat.

In order to evaluate a more complex task, the authors used the Caltech-UCSD Birds dataset, where the task is to classify images of 200 different species of birds, where many birds "look alike" and are hard to recognize. Again this simple method outperforms the state of the art by $6$ \%, producing $61.8$ accuracy. This result shows how CNN features outperform engineered ones, even when the task is considerably different from the training set.
This work also shows the importance of data augmentation for computer vision tasks.

Pailhas, Petillot and Capus \cite[-3em]{pailhas2010high} have explored the relationship between sonar resolution and target recognition accuracy. While this is not the same question as we are exploring, it is similar enough to warrant inclusion in this state of the art. This work concentrates on the sonar resolution as a physical property of the device itself, while we want to explore this relation from the image processing point of view.

The authors use a sidescan sonar simulator that produces synthetic sonar images. The background that were considered are a flat seabed, sand ripples, a rocky seabed, and a cluttered environment (rocks). The target are mine-like objects, including six classes (manta, rockan, cuboid, hemisphere, a lying cylinder on the side and a standing one).

The classifier used in this work is based on a Principal Component Analysis representation, that is matched with templates in a training set by means of minimizing a distance in feature space. The authors analyse the use of shadow or highlight features.

For classification using highlight, $95$ \% accuracy is obtained with 5 cm pixel resolution, which is considerably fine grained for a sidescan sonar. In contrast, classification using shadow requires less than 20 cm pixel resolution to obtain close to $100$ \% accuracy, but highlight classification at 20 cm pixel resolution is close to $50$ \%.
This work shows that using the shadow of an object is fundamental for good classification performance, but we believe these results are skewed due to the use of a PCA-based classifier. Other classifiers might perform differently. There is also the issue of the objects used in this work, as marine debris is considerably different in shape variation and lack of shadow information.

Mishkin et al. \cite{mishkin2016systematic} do a systematic evaluation of many network parameters for the ImageNet dataset in the context of image classification. This work consists of a large number of ablation studies, varying activation functions, different kinds of pooling, learning rate policies, pre-processing and normalization, batch size, etc.

Two results from these work are of interest for this chapter. The authors evaluated the effect of varying the input image size, which shows that decreasing the input image has the effect of reducing accuracy from $50$ \% at $224 \times 224$ input size to $30$ \% at $66 \times 66$ pixels. The relationship between input image size and accuracy is almost linear.
One way to offset this loss is to vary the network architecture as a function of the input size, as the authors tried to vary the strides and filter sizes to produce a constant size pooling output, reducing the effect of image size as accuracy only varies from $40$ \% to $45$ \%.

The second result is the variation of the training set size. The authors down-sample the ImageNet dataset to $0.2$ M, $0.4$ M, $0.6$ M, $0.8$ M and 1.2 million images (the original size). Accuracy decreases from $45$ \% at $1.2$ M to $30$ \% at $0.2$ M. The relationship between training set size and accuracy is quite close to linear, as it slowly decreases linearly from $1.2$ M to $0.4$ M, but then decreases more sharply.
While both results are quite interesting, these authors have not controlled for the random weight initialization, and variations of accuracy should be computed. Due to the large size of the ImageNet dataset, it can expected that these kind of evaluation protocol is not available due to the large computational resources required.

\section{Transfer Learning}
\label{lim:secTransferLearning}

In this experiment we evaluate the transfer learning capabilities of three networks we designed: ClassicNet, TinyNet and FireNet.

Our general procedure to evaluate transfer learning in a network is to first define a set of layers $L$ that we wish to evaluate. The features produced as output from these layers are then used to train a multi-class SVM \cite{sharif2014cnn}. In order to produce a fair evaluation of the features, we decided to split the training and testing sets according to the classes they contain, in order to learn features on one set of objects and test them in a different set of objects.
This should aid to verify the generalization capabilities of the network architecture.

We first split the dataset $D$ into datasets $F$ and $T$ by selecting a random subset of $\lfloor \frac{C}{2} \rfloor$ classes and assigning all samples from those classes to $F$, while the remaining classes and their samples are assigned to $T$. As our Marine Debris dataset contains 11 samples, 6 classes are assigned to $F$ and 5 are assigned to $T$. We split both the training and testing splits of our dataset separately, producing $F_{tr}$, $F_{ts}$ and $T_{tr}$, $T_{ts}$.

Dataset $F$ is to learn features by training a network model for classification, while $T$ is used to evaluate features. A given network model is trained on $F_{tr}$ and then for each layer in $L$, features are extracted at that layer from the network model by passing each sample in $T_{tr}$. Then a multi-class linear SVM with regularization coefficient $C = 1$  and decision surface "one-versus-one" is trained on those features \footnote{$C$ was obtained by cross-validation on a small part of the dataset}. Using the same network, features are again extracted using $T_{ts}$ and the SVM is tested on this dataset, producing an accuracy score. We repeat this process $N = 20$ times to account for random initialization of the feature extraction network and compute mean and standard deviation of test accuracy. Note that $F_{ts}$ is not used by this procedure, but it could be used to evaluate test accuracy of the feature extractor.

ClassicNet with 5 modules was tested with four different configurations: 8 or 32 filters, and Batch Normalization or Dropout as regularization. Features are extracted from the batch normalized outputs in each module (layers bn1-5), or from the Max-Pooling outputs in the case of the Dropout configurations (layers mp1-5), and we also include the output from the first fully connected layer (fc1). We used TinyNet with 5 modules and 8 filters per module, and FireNet with 3 modules and 4 filters. For TinyNet, features are extracted at the output of each of the five modules, while for FireNet features are the outputs of each module (three in total) and we also consider the output of the initial convolution (called convS in the figures). Each feature extraction network is trained for 15 epochs with a batch size $B = 64$ using the ADAM optimizer \cite{kingma2014adam} with a learning rate $\alpha = 0.01$. The data is randomly shuffled after each epoch in order to prevent spurious patterns in the data ordering to influence the network.

\begin{figure*}
	\subfloat[ClassicNet]{
		\label{lim:transferLearningNetworks:classic}
		\begin{tikzpicture}
		\begin{axis}[ybar,
		xlabel={Layers},
		ylabel={Test Accuracy (\%)},        
		symbolic x coords={bn1, bn2, bn3, bn4, bn5, fc1},
		xtick=data,
		legend style={at={(0.5,1.15)},anchor=north,legend columns=-1},
		ymajorgrids=true,
		grid style=dashed,
		height = 0.3\textheight,
		width = 0.9\textwidth]
				
		\addplot+[error bars/.cd, y dir=both,y explicit] table[x = layerIdx, y = meanTransferAcc, y error = stdTransferAcc, col sep = space] {chapters/data/limits/transferLearning-disjoint-ClassicNet-BN-5modules-8filters.csv};
		\addplot+[error bars/.cd, y dir=both,y explicit] table[x = layerIdx, y = meanTransferAcc, y error = stdTransferAcc, col sep = space] {chapters/data/limits/transferLearning-disjoint-ClassicNet-BN-5modules-32filters.csv};
		
		\addplot+[error bars/.cd, y dir=both,y explicit] table[x = layerIdx, y = meanTransferAcc, y error = stdTransferAcc, col sep = space] {chapters/data/limits/transferLearning-disjoint-ClassicNet-DO-5modules-8filters.csv};
		\addplot+[error bars/.cd, y dir=both,y explicit] table[x = layerIdx, y = meanTransferAcc, y error = stdTransferAcc, col sep = space] {chapters/data/limits/transferLearning-disjoint-ClassicNet-DO-5modules-32filters.csv};
		
		\legend{8 filters with BN, 32 filters with BN, 8 filters with Dropout, 32 filters with Dropout}
		
		\end{axis}
		\end{tikzpicture}		
	}
	
	\subfloat[TinyNet]{
		\label{lim:transferLearningNetworks:tiny}
		\begin{tikzpicture}
		\begin{axis}[ybar,
		xlabel={Modules},
		ylabel={Test Accuracy (\%)},        
		xtick=data,
		legend style={at={(0.5,1.15)},anchor=north,legend columns=-1},
		ymajorgrids=true,
		grid style=dashed,
		height = 0.3\textheight,
		width = 0.45\textwidth]
		
		\addplot+[error bars/.cd, y dir=both,y explicit] table[x = layerIdx, y = meanTransferAcc, y error = stdTransferAcc, col sep = space] {chapters/data/limits/transferLearning-disjoint-tinyNet5-8.csv};
		
		\end{axis}
		\end{tikzpicture}
	}
	\subfloat[FireNet]{
		\label{lim:transferLearningNetworks:fire}
		\begin{tikzpicture}
		\begin{axis}[ybar,
		xlabel={Layers},
		ylabel={Test Accuracy (\%)},        
		symbolic x coords={convS, mod1, mod2, mod3},
		xtick=data,
		legend style={at={(0.5,1.15)},anchor=north,legend columns=-1},
		ymajorgrids=true,
		grid style=dashed,
		height = 0.3\textheight,
		width = 0.45\textwidth]
		
		\addplot+[error bars/.cd, y dir=both,y explicit] table[x = layerIdx, y = meanTransferAcc, y error = stdTransferAcc, col sep = space] {chapters/data/limits/transferLearning-disjoint-fireNet3.csv};
		
		\end{axis}
		\end{tikzpicture}
	}
	
	\vspace*{0.5cm}
	\caption[Transfer Learning on Sonar Images]{Transfer Learning on Sonar Images. Mean test accuracy produced by an SVM trained on features output by different layers. Three networks are shown.}
	\label{lim:transferLearningNetworks}
\end{figure*}
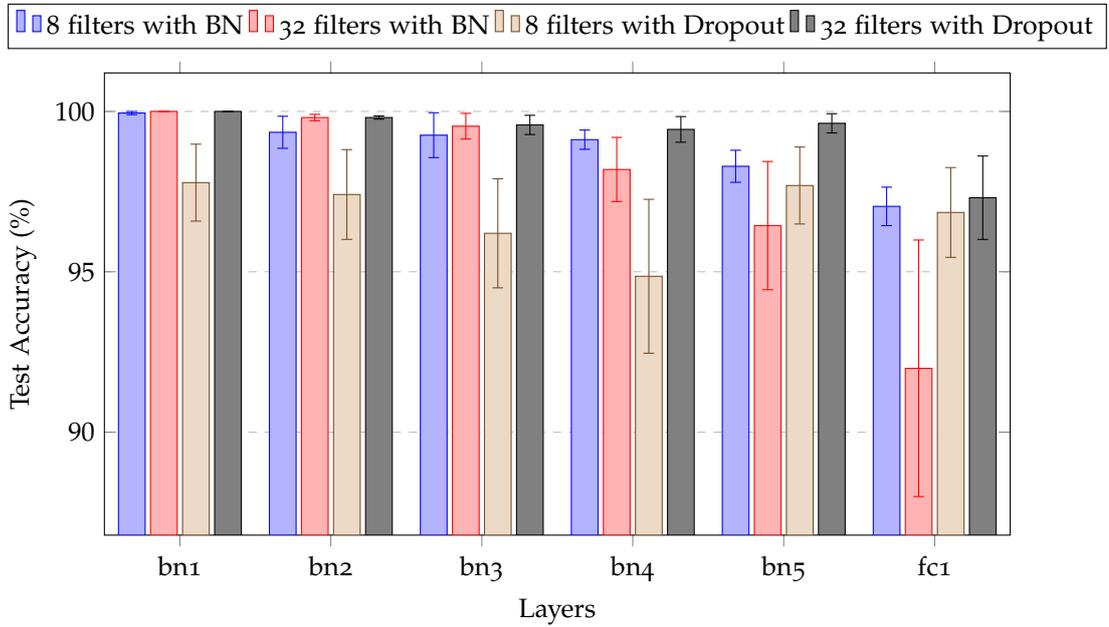
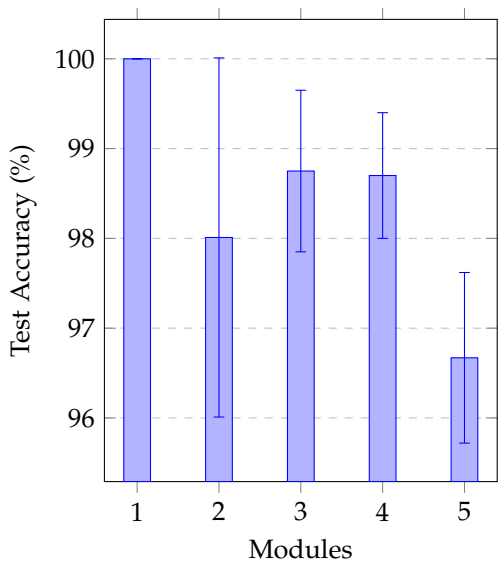
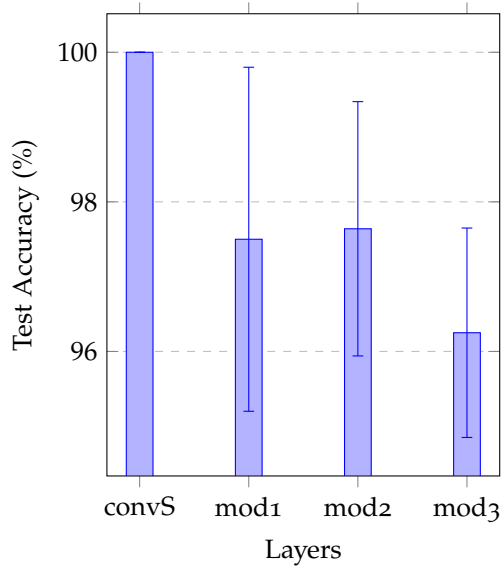

Our results are shown in Figure \ref{lim:transferLearningNetworks}. A general pattern that appears in all three experimental plots is that testing accuracy decreases with deeper layers/modules in the network. For example, in ClassicNet, features in the fc1 layer have lower accuracy than bn1 and bn2. The same effect can be seen in TinyNet and FireNet, specially as the last module/layer has the lowest testing accuracy. It is also notable that the features in the first layers have $100$ \% accuracy, with zero variation. This can be explained that as shallow features are typically very high dimensional, a linear SVM has a high chance of finding a separating hyperplane and perfectly classifying test data.

ClassicNet feature results are shown in Figure \ref{lim:transferLearningNetworks}\subref*{lim:transferLearningNetworks:classic}. Generalization varies considerably with different layers. 8 filters with Batch Normalization produces quite good generalization, but 32 filters with Dropout has almost the same accuracy, with it being superior for bn5 and fc1. Dropout with 8 filters has a considerable drop in accuracy compared with the other configurations. 32 filters with Dropout seems to be the best option for good generalization, which is consistent with the use of Dropout to both de-correlate neurons and increase their generalization power \cite{srivastava2014dropout}.

Results for TinyNet are shown in Figure \ref{lim:transferLearningNetworks}\subref*{lim:transferLearningNetworks:tiny}, and for FireNet in Figure  \ref{lim:transferLearningNetworks}\subref*{lim:transferLearningNetworks:fire}. The shape of both plots is quite similar, with a decrease from the first to the second module, and then a increase, followed by another decrease. Seems using the layer before the last might have the best generalization performance, but using the first layer has by far the best accuracy.

It should be noted that for both TinyNet and FireNet, their generalization capability is very good, with the minimum accuracy being greater than $96$ \%. Our results also seem to indicate that choosing features from the last layer might not always be the best option, as it has consistently been done by many researchers. This could be a peculiarity of sonar images, and we do not believe it applies for larger datasets.

As summary, we expected that transfer learning using CNN features will perform adequately and it did, but it was unexpected that we found negative correlation between layer depth and test accuracy, For this kind of data and architecture, the best choice is to extract a high dimensional feature vector from a layer close to the input.

\begin{marginfigure}
	\centering
	\stackunder[5pt] {
		\includegraphics[width = 3cm]{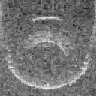}
	}{$96 \times 96$}

	\vspace{5pt}
	
	\stackunder[5pt] {
		\includegraphics[width = 2.5cm]{tire-96x96.jpg}
	}{$80 \times 80$}
	
	\vspace{5pt}
	
	\stackunder[5pt] {
		\includegraphics[width = 2.0cm]{tire-96x96.jpg}
	}{$64 \times 64$}
	
	\vspace{5pt}
	
	\stackunder[5pt] {
		\includegraphics[width = 1.5cm]{tire-96x96.jpg}
	}{$48 \times 48$}
	
	\vspace{5pt}
	
	\stackunder[5pt] {
		\includegraphics[width = 1.0cm]{tire-96x96.jpg}
	}{$32 \times 32$}
	
	\vspace{5pt}
	
	\stackunder[5pt] {
		\includegraphics[width = 0.5cm]{tire-96x96.jpg}
	}{$16 \times 16$}
	
	\caption[Example of object scales used for our experiment]{Example of object scales used for our experiment. These images correspond to a Tire.}
\end{marginfigure}

\FloatBarrier
\section{Effect of Object Size}

In this section we experiment with object size, as we would like to investigate how the object size/scale affects classification accuracy.

For this purpose we take the initial $96 \times 96$ image crops and downscale them using bilinear filtering to a predefined size. We use square pixel sizes $s \times s$ with $s \in [16, 32, 48, 64, 80, 96]$. To obtain the size parameters, we started from the natural size of the bounding boxes in our dataset ($96 \times 96$ pixels) and downscaled them in 16 pixel steps, until we reached the smallest size that can be classified by a 2-module ClassicNet, which is 16 pixels (due to downsampling by the use of Max-Pooling). Both the training and testing sets are resized. We evaluate accuracy on the test set. In order to account for the effect of random initialization, we train $N = 20$ networks and compute the mean and standard deviation of accuracy.

We found experimentally \cite{valdenegro2017limits} that the kind of regularization and optimizer that are used greatly affect the results. We evaluate four combinations, using Batch Normalization or Dropout for regularization, and ADAM or SGD for optimizer. All networks for every configuration are trained for $30$ epochs, using a learning rate $\alpha = 0.01$ with a batch size $B = 128$ samples.

We selected three networks to be evaluated: ClassicNet with 2 modules and 32 filters, TinyNet with 5 modules and 8 filters, and FireNet with 3 modules and 4 filters. We only evaluate ClassicNet with both Batch Normalization and Dropout, as it is not appropriate to use Dropout with a fully convolutional network such as TinyNet and FireNet. In these two networks we only use Batch Normalization as regularizer.  

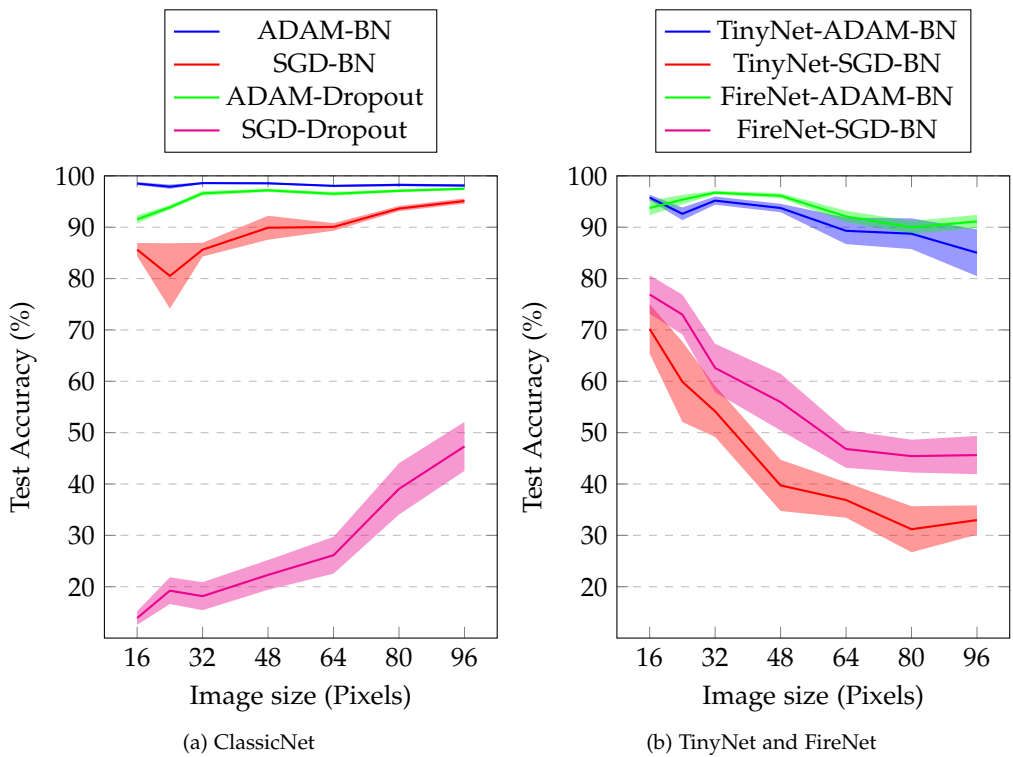
\begin{figure*}
	\subfloat[ClassicNet]{
		\begin{tikzpicture}
		\begin{axis}[
		xlabel={Image size (Pixels)},
		ylabel={Test Accuracy (\%)},        
		ymin=10, ymax=100,
		xtick={16, 32, 48, 64, 80, 96},
		ytick={20,30,40,50,60,70,80,90,100},
		legend style={at={(0.5, 1.05)},anchor=south},
		ymajorgrids=true,
		grid style=dashed,
		height = 0.3\textheight,
		width = 0.45\textwidth]
		
		\errorband{chapters/data/limits/classicCNN-BN-ADAM-AccuracyVsImageSize.csv}{pixelImageSize}{meanAcc}{stdAcc}{blue}{0.4}
		\errorband{chapters/data/limits/classicCNN-BN-SGD-AccuracyVsImageSize.csv}{pixelImageSize}{meanAcc}{stdAcc}{red}{0.4}
		
		\errorband{chapters/data/limits/classicCNN-Dropout-ADAM-AccuracyVsImageSize.csv}{pixelImageSize}{meanAcc}{stdAcc}{green}{0.4}
		\errorband{chapters/data/limits/classicCNN-Dropout-SGD-AccuracyVsImageSize.csv}{pixelImageSize}{meanAcc}{stdAcc}{magenta}{0.4}
		
		\legend{ADAM-BN, SGD-BN, ADAM-Dropout, SGD-Dropout}
		
		\end{axis}
		\end{tikzpicture}
	}
	\subfloat[TinyNet and FireNet]{
		\begin{tikzpicture}
		\begin{axis}[
		xlabel={Image size (Pixels)},
		ylabel={Test Accuracy (\%)},        
		ymin=10, ymax=100,
		xtick={16, 32, 48, 64, 80, 96},
		ytick={20,30,40,50,60,70,80,90,100},
        legend style={at={(0.5, 1.05)},anchor=south},
		ymajorgrids=true,
		grid style=dashed,
		height = 0.3\textheight,
		width = 0.45\textwidth]
		
		\errorband{chapters/data/limits/tinyNetCNN-BN-AccuracyVsImageSize.csv}{pixelImageSize}{meanAcc}{stdAcc}{blue}{0.4}
		\errorband{chapters/data/limits/tinyNetCNN-BN-SGD-AccuracyVsImageSize.csv}{pixelImageSize}{meanAcc}{stdAcc}{red}{0.4}
		
		\errorband{chapters/data/limits/smallFireNetCNN-BN-AccuracyVsImageSize.csv}{pixelImageSize}{meanAcc}{stdAcc}{green}{0.4}
		\errorband{chapters/data/limits/smallFireNetCNN-BN-SGD-AccuracyVsImageSize.csv}{pixelImageSize}{meanAcc}{stdAcc}{magenta}{0.4}
		
		\legend{TinyNet-ADAM-BN, TinyNet-SGD-BN, FireNet-ADAM-BN, FireNet-SGD-BN}
		
		\end{axis}
		\end{tikzpicture}
	}
	
	\vspace*{0.5cm}
    \forceversofloat
	\caption[Graphical summary of the effect of object size/scale for different CNN models]{Graphical summary of the effect of object size/scale for different CNN models. The shaded areas represent one $\sigma$ error bars.}
	\label{lim:graphicalObjectSize}
\end{figure*}

We present our results as a plot in Figure \ref{lim:graphicalObjectSize} and as numerical values in Table \ref{lim:numericalObjectSize}. For ClassicNet, we can see that a high accuracy classifier is produced by using both ADAM and Batch Normalization. ADAM with Dropout also produces quite high accuracy but lower than using Batch Normalization. Using SGD produces considerably lower accuracy classifiers, specially when combined with Dropout.

One of the results we expected is that accuracy should decrease with smaller object size, as less information is available for the classifier and its typical that smaller objects are harder to classify. Our results show that this happens when using SGD on ClassicNet, as accuracy monotonically increases as object size also increases. This is more noticeable with the SGD-Dropout configuration.

But unexpectedly, the ADAM combinations produce high accuracy that seems to be invariant to object size. The ADAM with Batch Normalization combination consistently produces results that are very accurate (only $1.5$ \% from perfect classification) with little variation.

\begin{table*}[!ht]
    \centering
	\forceversofloat
	\begin{tabular}{llll}
		\hline 
		Model / Pixel Size 			& 16 				   & 32 				    & 48\\ 
		\hline
		ClassicNet-ADAM-BN  & $98.5 \pm 0.5$ \%    & $98.6 \pm 0.3$ \%  & $98.5 \pm 0.3$ \%\\
		ClassicNet-SGD-BN 	& $85.7 \pm 2.6$ \%    & $85.6 \pm 2.7$ \%  & $89.9 \pm 4.7$ \%\\
		ClassicNet-ADAM-DO  & $91.5 \pm 1.5$ \%    & $96.6 \pm 0.9$ \%  & $97.2 \pm 0.6$ \%\\
		ClassicNet-SGD-DO 	& $13.9 \pm 2.6$ \%    & $18.2 \pm 5.5$ \%  & $22.3 \pm 5.8$ \%\\
		\hline
		TinyNet-ADAM-BN 	& $95.8 \pm 1.1$ \%    & $95.2 \pm 1.6$ \%  & $93.7 \pm 1.6$ \%\\
		TinyNet-SGD-BN 	    & $70.2 \pm 9.7$ \%    & $54.2 \pm 10.0$ \% & $39.7 \pm 10.0$ \%\\
		\hline
		FireNet-ADAM-BN  	& $93.7 \pm 2.9$ \%    & $96.7 \pm 0.7$ \%  & $96.1 \pm 1.0$ \%\\
		FireNet-SGD-BN 		& $76.9 \pm 7.5$ \%    & $62.6 \pm 9.5$ \%  & $56.0 \pm 11.1$ \%\\
		\hline 
	\end{tabular}
    \begin{tabular}{llll}
        \hline 
        Model / Pixel Size 			& 64 				  & 80 				& 96\\ 
        \hline
        ClassicNet-ADAM-BN  & $98.1 \pm 0.3$ \% & $98.2 \pm 0.5$ \% & $98.1 \pm 0.5$ \%\\
        ClassicNet-SGD-BN 	& $90.1 \pm 1.5$ \% & $93.6 \pm 1.0$ \% & $95.1 \pm 1.0$ \%\\
        ClassicNet-ADAM-DO  & $96.5 \pm 0.7$ \% & $97.1 \pm 0.6$ \% & $97.5 \pm 0.5$ \%\\
        ClassicNet-SGD-DO 	& $26.1 \pm 7.2$ \% & $39.1 \pm 10.0$ \% & $47.3 \pm 9.5$ \%\\
        \hline
        TinyNet-ADAM-BN 	& $89.3 \pm 5.2$ \% & $88.7 \pm 6.0$ \% & $85.0 \pm 9.1$ \%\\
        TinyNet-SGD-BN 	    & $36.9 \pm 6.9$ \% & $31.2 \pm 9.0$ \% & $33.0 \pm 5.7$ \%\\
        \hline
        FireNet-ADAM-BN  	& $92.1 \pm 2.2$ \% & $90.0 \pm 2.5$ \% & $91.1 \pm 2.6$ \%\\
        FireNet-SGD-BN 		& $46.8 \pm 7.3$ \% & $45.4 \pm 6.4$ \% & $45.6 \pm 7.5$ \%\\
        \hline 
    \end{tabular}
    \vspace*{0.5cm}
	\caption{Numerical summary of the effect of object size/scale for different CNN models.}
	\label{lim:numericalObjectSize}
\end{table*}

TinyNet and FireNet results are not as good as ClassicNet. Both networks seem to have a negative correlation with object size, starting from high accuracy for small objects, and decreasing the precision of their predictions as objects gets bigger. This was quite unexpected. We believe this result can be explained by the fact that as these networks have a considerably lower number of parameters, the number of "acceptable" or "right" values for the weights is smaller, and thus these networks require more data in order to generalize properly.
Comparing these results with Chapter \ref{chapter:sonar-classification}, where we used data augmentation, we can see that not using data augmentation as we do here considerably reduces classifier generalization.
Using ADAM produces acceptable accuracy, but it still decreases slightly with bigger objects. These results also show that FireNet can be considerably more accurate than TinyNet, probably owning to the larger number of parameters.

Our combined results show that the combination of both ADAM and Batch Normalization produce a very good classifier that seems to be invariant to object size. This can be explained as both ADAM and Batch Normalization are adaptive algorithms. ADAM adapts the learning rate with the exponentially running mean of the gradients, so when the optimization process is close to a high-accuracy minima, it can adapt the learning rate in order to consistently reach that minima. SGD alone cannot do this, even if fixed learning rate schedules are used. Gradient information as part of learning rate calculation is a key for this process to succeed.

As a general summary of these results, it is possible to say that a convolutional neural network can be designed and trained in a way that it is approximately invariant to object size. This requires the use of an adaptive learning rate (ADAM) and an appropriate regularization and control of the co-variate shift throught the use of Batch Normalization. Dropout combined with ADAM also produces a size invariant classifier but it is less accurate than other configurations.

\FloatBarrier
\section{Effect of Number of Training Samples}
\label{lim:secNumTrainingSamples}

In this section we investigate how many training samples are required for a given generalization target. We do this by a simple but powerful experiment.

The basic idea of this experiment is to take a given training set $T$ and produce sub-sampled version of that dataset. As we are approaching a classification problem, we decided to normalize the number of image samples in each class (abbreviated SPC). We decide a set of SPC values and produce several sub-sampled versions of $T$, where for $T_{i}$ the number of samples per class in that dataset is $i$. This allows comparisons using different SPC values. The testing set is not sub-sampled in anyway in order to enable comparisons. Note that as our dataset is not balanced, and using this procedure will produce a balanced training set, so it is expected that the results will not match the ones from Chapter \ref{chapter:sonar-classification}, but as we are using the same testing set, results are comparable.

As it has been previously done, in order to consider the effect of random initialization, we train $N = 10$ instances of the same network model on each dataset $T_{i}$, but also we must consider the variations in the sub-sampled training set, as sampling is performed randomly. Then we also generate $M = 10$ different training sets with the same value of SPC, and train $N$ networks on each of these sets. After both variations are taken into account, we will train $N \times M = 100$ networks for each value of SPC.

We selected $\text{SPC} \in \{ [1, 2, 3, ..., 20] \cup [25, 30, 35, ..., 50] \cup [60, 70, 80, \newline ..., 150] \}$. The first range is designed to show the differences in generalization with small samples of data, while the other ranges show behaviour with large samples. As our dataset is unbalanced, we only evaluate up to 150 samples per class, which is only two times the number of samples of the class with the least samples.

We evaluate three networks, as it has previously been done: ClassicNet with 2 modules and 32 filters, combined and Batch Normalization and Dropout as regularizers. TinyNet with 5 modules and 8 filters, and FireNet with 3 modules and 4 filters. We have also included a linear SVM with $C = 10$ as a comparison baseline.

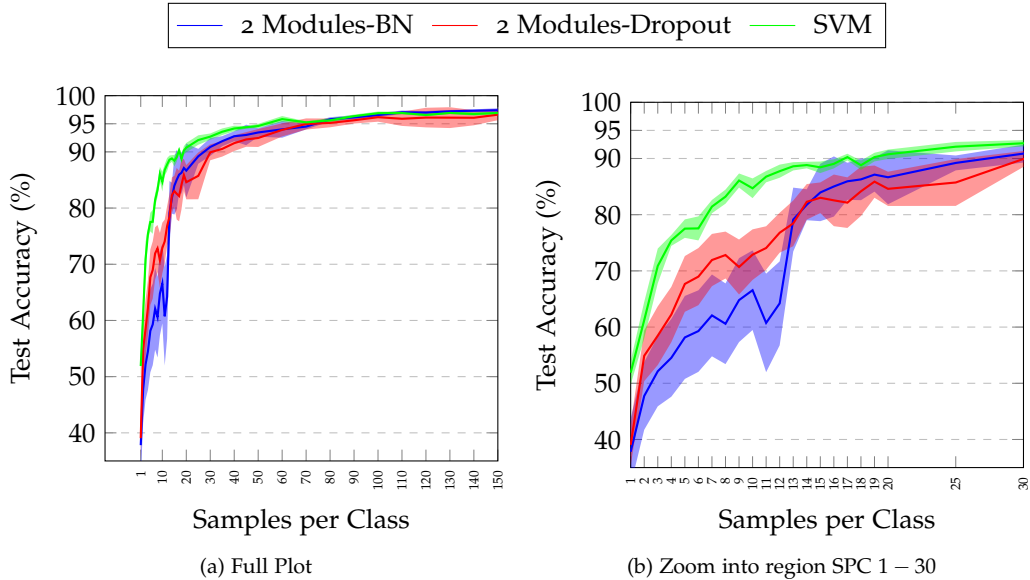
\begin{figure*}[t]
	\forcerectofloat
	\centering
	\begin{tikzpicture}
	\begin{customlegend}[legend columns = 4,legend style = {column sep=1ex}, legend cell align = left,
	legend entries={2 Modules-BN, 2 Modules-Dropout, SVM}]
	\addlegendimage{mark=none,blue}
	\addlegendimage{mark=none,red}
	
	\addlegendimage{mark=none,green}
	\end{customlegend}
	\end{tikzpicture}
	
	\subfloat[Full Plot]{
		\begin{tikzpicture}
		\begin{axis}[
		xlabel={Samples per Class},
		ylabel={Test Accuracy (\%)},        
		xmax = 150,
		ymin=35, ymax=100,
		xtick={1,10,20,30,40,50,60,70,80,90,100,110,120,130,140,150},
		ytick={30,40,50,60,70,80,90,95,100},
		x tick label style={font=\tiny, rotate=90},
		legend pos=south east,
		ymajorgrids=true,
		grid style=dashed,
		height = 0.25\textheight,
		width = 0.45\textwidth]
		
		\errorband{chapters/data/limits/classicCNN2L-BN-AccuracyVsTrainSetSize.csv}{samplesPerClass}{meanAcc}{stdAcc}{blue}{0.4}
		\errorband{chapters/data/limits/classicCNN2L-Dropout-AccuracyVsTrainSetSize.csv}{samplesPerClass}{meanAcc}{stdAcc}{red}{0.4}
		
		\errorband{chapters/data/SVM-C10-AccuracyVsTrainSetSize.csv}{samplesPerClass}{meanAcc}{stdAcc}{green}{0.4}
		
		\end{axis}
		\end{tikzpicture}
	}
	\subfloat[Zoom into region SPC $1-30$]{
		\begin{tikzpicture}
		\begin{axis}[
		xlabel={Samples per Class},
		ylabel={Test Accuracy (\%)},        
		xmin=1, xmax=30,
		ymin=35, ymax=100,
		xtick={1,2,3,4,5,6,7,8,9,10,11,12,13,14,15,16,17,18,19,20,25,30},
		ytick={30,40,50,60,70,80,90,95,100},
		x tick label style={font=\tiny, rotate=90},
		legend pos=south east,
		ymajorgrids=true,
		grid style=dashed,
		height = 0.25\textheight,
		width = 0.45\textwidth]
		
		\errorband{chapters/data/limits/classicCNN2L-BN-AccuracyVsTrainSetSize.csv}{samplesPerClass}{meanAcc}{stdAcc}{blue}{0.4}
		\errorband{chapters/data/limits/classicCNN2L-Dropout-AccuracyVsTrainSetSize.csv}{samplesPerClass}{meanAcc}{stdAcc}{red}{0.4}
		
		\errorband{chapters/data/SVM-C10-AccuracyVsTrainSetSize.csv}{samplesPerClass}{meanAcc}{stdAcc}{green}{0.4}
		
		\end{axis}
		\end{tikzpicture}
	}
	\vspace*{0.5cm}
	\caption[Samples per Class versus Accuracy for ClassicNet with 2 modules]{Samples per Class versus Accuracy for ClassicNet with 2 modules, including error regions.}
	\label{lim:classicNetSPCVSAccuracy}
\end{figure*}

ClassicNet results are shown in Figure \ref{lim:classicNetSPCVSAccuracy}. Our results show that these networks scale quite well with the number of samples per class, and the results are comparable with what is produced by the SVM. But it is clear that the SVM outperforms and obtains slightly better accuracy than ClassicNet (both with Batch Normalization and Dropout).

For small samples, approximately less than 15 samples per class, Dropout produced better results than Batch Normalization. This is unexpected as Batch Normalization is considered to be a better regularizer than Dropout, but seems that when the number of training samples is small, the added noise from Dropout could better regularize the neural network. As the number of samples per class increases, then Batch Normalization dominates and produces slightly better results.

In the large sample case (more than 100 samples per class), ClassicNet outperforms the SVM by a small margin, which is expected and consistent with the results obtained in Chapter \ref{chapter:sonar-classification}. Variations in generalization (accuracy) considerably decrease as more samples are added. The SVM classifier does not seem to have any change in variation of accuracy as a function of the samples per class, unlike the neural networks shown here.

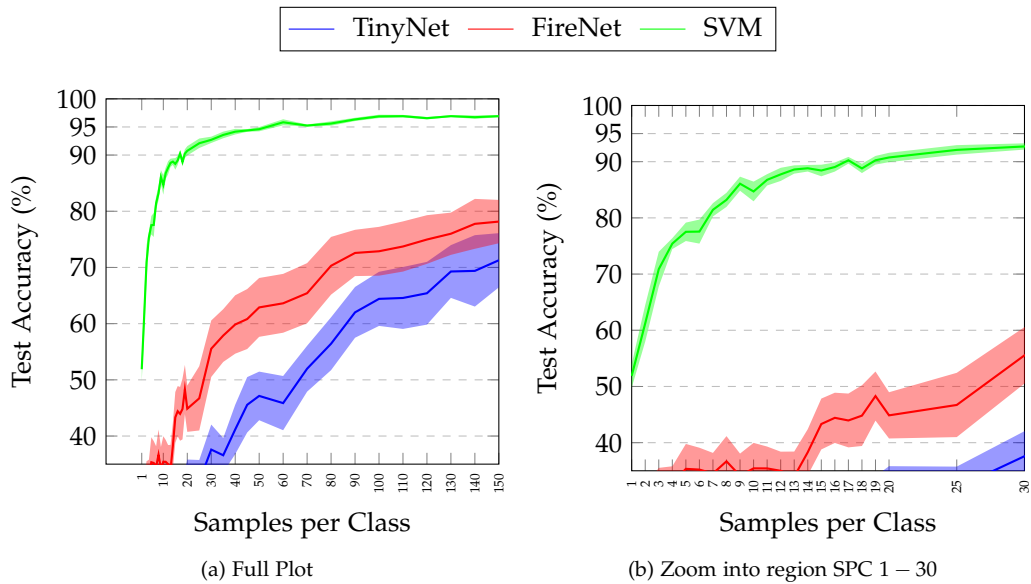
\begin{figure*}[t]
	\centering
	\begin{tikzpicture}
	\begin{customlegend}[legend columns = 4,legend style = {column sep=1ex}, legend cell align = left,
	legend entries={TinyNet, FireNet, SVM}]
	\addlegendimage{mark=none,blue}
	\addlegendimage{mark=none,red}
	
	\addlegendimage{mark=none,green}
	\end{customlegend}
	\end{tikzpicture}
	
	\subfloat[Full Plot]{
		\begin{tikzpicture}
		\begin{axis}[
		xlabel={Samples per Class},
		ylabel={Test Accuracy (\%)},        
		xmax = 150,
		ymin=35, ymax=100,
		xtick={1,10,20,30,40,50,60,70,80,90,100,110,120,130,140,150},
		ytick={30,40,50,60,70,80,90,95,100},
		x tick label style={font=\tiny, rotate=90},
		legend pos=south east,
		ymajorgrids=true,
		grid style=dashed,
		height = 0.25\textheight,
		width = 0.45\textwidth]
		
		\errorband{chapters/data/limits/tinyNetCNN-BN-AccuracyVsTrainSetSize.csv}{samplesPerClass}{meanAcc}{stdAcc}{blue}{0.4}
		\errorband{chapters/data/limits/fireNetCNN-BN-AccuracyVsTrainSetSize.csv}{samplesPerClass}{meanAcc}{stdAcc}{red}{0.4}
		
		\errorband{chapters/data/SVM-C10-AccuracyVsTrainSetSize.csv}{samplesPerClass}{meanAcc}{stdAcc}{green}{0.4}
		
		\end{axis}
		\end{tikzpicture}
	}
	\subfloat[Zoom into region SPC $1-30$]{
		\begin{tikzpicture}
		\begin{axis}[
		xlabel={Samples per Class},
		ylabel={Test Accuracy (\%)},        
		xmin=1, xmax=30,
		ymin=35, ymax=100,
		xtick={1,2,3,4,5,6,7,8,9,10,11,12,13,14,15,16,17,18,19,20,25,30},
		ytick={30,40,50,60,70,80,90,95,100},
		x tick label style={font=\tiny, rotate=90},
		legend pos=south east,
		ymajorgrids=true,
		grid style=dashed,
		height = 0.25\textheight,
		width = 0.45\textwidth]
		
		\errorband{chapters/data/limits/tinyNetCNN-BN-AccuracyVsTrainSetSize.csv}{samplesPerClass}{meanAcc}{stdAcc}{blue}{0.4}
		\errorband{chapters/data/limits/fireNetCNN-BN-AccuracyVsTrainSetSize.csv}{samplesPerClass}{meanAcc}{stdAcc}{red}{0.4}
		
		\errorband{chapters/data/SVM-C10-AccuracyVsTrainSetSize.csv}{samplesPerClass}{meanAcc}{stdAcc}{green}{0.4}
		
		\end{axis}
		\end{tikzpicture}
	}
	\vspace*{0.5cm}
	\caption[Samples per Class versus Accuracy for TinyNet-5 and FireNet-3]{Samples per Class versus Accuracy for TinyNet-5 and FireNet-3, including error regions.}
	\label{lim:tinyFireNetSPCVsAccuracy}
\end{figure*}

Results for TinyNet and FireNet are shown in Figure \ref{lim:tinyFireNetSPCVsAccuracy}. For these networks, results show that they perform poorly with less data, specially when the number of samples per class is low, as it can be seen in Figure \ref{lim:tinyFireNetSPCVsAccuracy}b. This confirms the results obtained in the previous section, where we saw that training a network with varying image sizes decreased accuracy and generalization with this networks when image size was increased, but the number of samples was kept constant.

In all tested samples per class configurations, FireNet outperformed TinyNet by a considerably margin (up to $8$ \%). This can be expected as FireNet has more parameters than TinyNet, but it is unexpected as we know from Chapter \ref{chapter:sonar-classification} that TinyNet can achieve high accuracy close to $99$ \%. Then the only difference is the quantity of data that is required to learn the model with good generalization.

We believe that as these models have considerably less number of parameters, there are less possible combinations of parameters that produce good accuracy and generalization (local or global minima), so it seems more data is required to reach these sets of parameters. The loss function could be quite noisy instead of smooth. This theory is supported by the considerable variation in test accuracy, which stays almost constant as the number of samples is varied. In some cases during the experiment, we say accuracy of up to $90$ \% as maximum values for SPC 100-150, but still this is a rare example and not a consistent pattern as shown by the mean value.

We are aware that we could have used data augmentation in order to obtain a higher accuracy, but this would only correspond to higher SPC values. We did not perform these tests using data augmentation due to the considerably amount of time it takes for them to run on a GPU (several days), as hundreds of neural networks have to be trained. We leave this for future work.

Table \ref{lim:samplesPerClassVsAccuracy} shows a numerical view of our results, for selected values of the number of samples per class (SPC). A more accurate view of our results can be shown. For 150 samples per class, the baseline SVM obtains $96.9 \pm 0.4$ \% accuracy, while ClassicNet with Batch Normalization gets $97.4 \pm 0.7$ \% and the same network with Dropout obtains $96.6 \pm 2.0$ \%. TinyNet at the same samples per class configuration gets $71.3 \pm 9.7$ \%, and FireNet obtains $78.2 \pm 7.7$ \%.

Evaluating these networks at small sample sizes, approximately $40$ \% accuracy can be obtained with a single sample, which is not too bad, as it is better than the random chance limit for 11 classes ($\frac{100}{10} \% \sim 9.1$ \%), but it does not produce an accurate classifier that can be used for practical applications. If at least $90$ \% accuracy is desired, then at least 30-50 samples per class are required, and with no more than 150 samples per class might be required for a high accuracy classifier, as our experiments show.

\begin{table}[t]
	\forcerectofloat
	\begin{tabular}{llll}
		\hline 
		Method/SPC 		& 1 				   & 5 				    & 10 \\ 
		\hline
		ClassicNet-2-BN & $37.8 \pm 12.0$ \%   & $58.2 \pm 14.8$ \% & $66.6 \pm 14.2$ \%\\
		ClassicNet-2-DO & $39.1 \pm 7.4$ \%    & $67.7 \pm 9.9$ \%  & $72.9 \pm 9.0$ \%\\
		\hline
		TinyNet-5-8 	& $19.3 \pm 5.6$ \%    & $23.4 \pm 6.7$ \%  & $23.9 \pm 6.8$ \%\\
		FireNet-3-4 	& $26.5 \pm 5.9$ \%    & $35.4 \pm 8.9$ \%  & $35.4 \pm 9.1$ \%\\
		\hline
		SVM				& $51.9 \pm 4.2$ \%    & $77.5 \pm 3.3$ \%  & $84.7 \pm 3.5$ \%\\
		\hline 
	\end{tabular}
    \begin{tabular}{llll}
        \hline 
        Method/SPC 		& 30 				  & 50 				& 100\\ 
        \hline
        ClassicNet-2-BN & $90.9 \pm 3.2$ \% & $93.5 \pm 1.5$ \% & $96.6 \pm 0.7$ \%\\
        ClassicNet-2-DO & $89.9 \pm 2.8$ \% & $92.5 \pm 3.2$ \% & $96.2 \pm 1.6$ \%\\
        \hline
        TinyNet-5-8 	& $37.6 \pm 8.9$ \% & $47.2 \pm 8.7$ \% & $64.4 \pm 9.6$ \%\\
        FireNet-3-4 	& $55.5 \pm 10.1$ \% & $62.9 \pm 10.5$ \% & $72.9 \pm 8.7$ \%\\
        \hline
        SVM				& $92.7 \pm 1.1$ \% & $94.6 \pm 0.7$ \% & $96.9 \pm 0.3$ \%\\
        \hline 
    \end{tabular}
	\caption[Mean and standard deviation of test accuracy as the number of samples per class is varied]{Mean and standard deviation of test accuracy as the number of samples per class is varied, for a selected values of SPC.}
	\label{lim:samplesPerClassVsAccuracy}
\end{table}

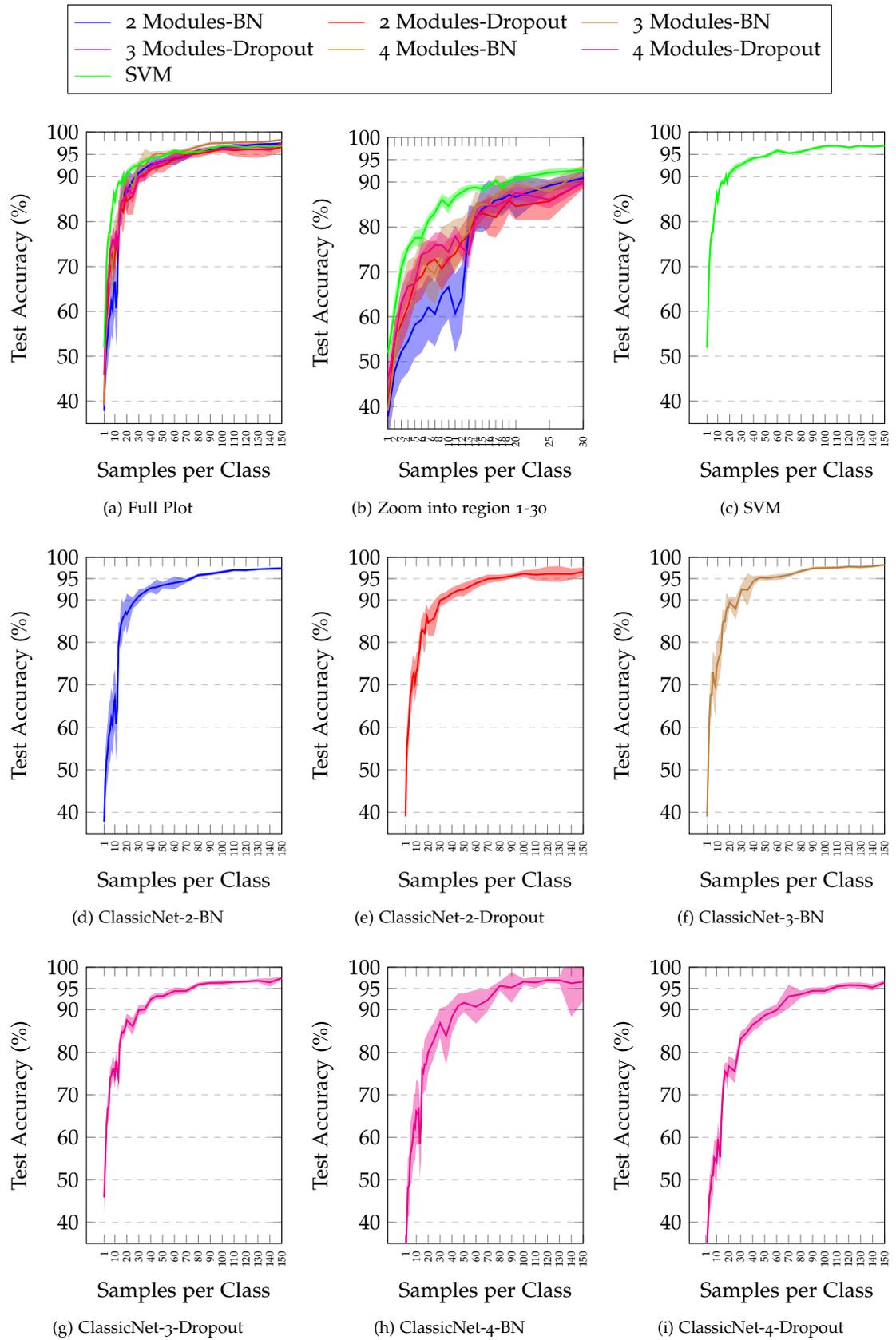
\begin{figure*}[t]
	\centering
    \vspace*{-2cm}
	\begin{tikzpicture}
	\begin{customlegend}[legend columns = 3,legend style = {column sep=1ex}, legend cell align = left,
	legend entries={2 Modules-BN, 2 Modules-Dropout, 3 Modules-BN, 3 Modules-Dropout, 4 Modules-BN, 4 Modules-Dropout, SVM}]
	\addlegendimage{mark=none,blue}
	\addlegendimage{mark=none,red}
	
	\addlegendimage{mark=none,brown}
	\addlegendimage{mark=none,magenta}
	
	\addlegendimage{mark=none,orange}
	\addlegendimage{mark=none,purple}
	
	\addlegendimage{mark=none,green}
	\end{customlegend}
	\end{tikzpicture}
	
	\subfloat[Full Plot]{
	\begin{tikzpicture}
	\begin{axis}[
	xlabel={Samples per Class},
	ylabel={Test Accuracy (\%)},        
	xmax = 150,
	ymin=35, ymax=100,
	xtick={1,10,20,30,40,50,60,70,80,90,100,110,120,130,140,150},
	ytick={30,40,50,60,70,80,90,95,100},
	x tick label style={font=\tiny, rotate=90},
	legend pos=south east,
	ymajorgrids=true,
	grid style=dashed,
	height = 0.25\textheight,
	width = 0.32\textwidth]
	
	\errorband{chapters/data/limits/classicCNN2L-BN-AccuracyVsTrainSetSize.csv}{samplesPerClass}{meanAcc}{stdAcc}{blue}{0.4}
	\errorband{chapters/data/limits/classicCNN2L-Dropout-AccuracyVsTrainSetSize.csv}{samplesPerClass}{meanAcc}{stdAcc}{red}{0.4}
	
	\errorband{chapters/data/limits/classicCNN3L-BN-AccuracyVsTrainSetSize.csv}{samplesPerClass}{meanAcc}{stdAcc}{brown}{0.4}
	\errorband{chapters/data/limits/classicCNN3L-Dropout-AccuracyVsTrainSetSize.csv}{samplesPerClass}{meanAcc}{stdAcc}{magenta}{0.4}
	
	\errorband{chapters/data/SVM-C10-AccuracyVsTrainSetSize.csv}{samplesPerClass}{meanAcc}{stdAcc}{green}{0.4}
	
	\end{axis}
	\end{tikzpicture}
	}
	\subfloat[Zoom into region 1-30]{
		\begin{tikzpicture}
		\begin{axis}[
		xlabel={Samples per Class},
		ylabel={Test Accuracy (\%)},        
		xmin=1, xmax=30,
		ymin=35, ymax=100,
		xtick={1,2,3,4,5,6,7,8,9,10,11,12,13,14,15,16,17,18,19,20,25,30},
		ytick={30,40,50,60,70,80,90,95,100},
		x tick label style={font=\tiny, rotate=90},
		legend pos=south east,
		ymajorgrids=true,
		grid style=dashed,
		height = 0.25\textheight,
		width = 0.32\textwidth]
		
		\errorband{chapters/data/limits/classicCNN2L-BN-AccuracyVsTrainSetSize.csv}{samplesPerClass}{meanAcc}{stdAcc}{blue}{0.4}
		\errorband{chapters/data/limits/classicCNN2L-Dropout-AccuracyVsTrainSetSize.csv}{samplesPerClass}{meanAcc}{stdAcc}{red}{0.4}
		
		\errorband{chapters/data/limits/classicCNN3L-BN-AccuracyVsTrainSetSize.csv}{samplesPerClass}{meanAcc}{stdAcc}{brown}{0.4}
		\errorband{chapters/data/limits/classicCNN3L-Dropout-AccuracyVsTrainSetSize.csv}{samplesPerClass}{meanAcc}{stdAcc}{magenta}{0.4}
		
		\errorband{chapters/data/SVM-C10-AccuracyVsTrainSetSize.csv}{samplesPerClass}{meanAcc}{stdAcc}{green}{0.4}
		
		\end{axis}
		\end{tikzpicture}
	}
	\subfloat[SVM]{
		\begin{tikzpicture}
		\begin{axis}[
		xlabel={Samples per Class},
		ylabel={Test Accuracy (\%)},        
		xmax = 150,
		ymin=35, ymax=100,
		xtick={1,10,20,30,40,50,60,70,80,90,100,110,120,130,140,150},
		ytick={30,40,50,60,70,80,90,95,100},
		x tick label style={font=\tiny, rotate=90},
		legend pos=south east,
		ymajorgrids=true,
		grid style=dashed,
		height = 0.25\textheight,
		width = 0.32\textwidth]
		
		\errorband{chapters/data/SVM-C10-AccuracyVsTrainSetSize.csv}{samplesPerClass}{meanAcc}{stdAcc}{green}{0.4}
		
		\end{axis}
		\end{tikzpicture}
	}

	\subfloat[ClassicNet-2-BN]{
		\begin{tikzpicture}
		\begin{axis}[
		xlabel={Samples per Class},
		ylabel={Test Accuracy (\%)},        
		xmax = 150,
		ymin=35, ymax=100,
		xtick={1,10,20,30,40,50,60,70,80,90,100,110,120,130,140,150},
		ytick={30,40,50,60,70,80,90,95,100},
		x tick label style={font=\tiny, rotate=90},
		legend pos=south east,
		ymajorgrids=true,
		grid style=dashed,
		height = 0.24\textheight,
		width = 0.32\textwidth]
		
		\errorband{chapters/data/limits/classicCNN2L-BN-AccuracyVsTrainSetSize.csv}{samplesPerClass}{meanAcc}{stdAcc}{blue}{0.4}				
		\end{axis}
		\end{tikzpicture}
	}
	\subfloat[ClassicNet-2-Dropout]{
		\begin{tikzpicture}
		\begin{axis}[
		xlabel={Samples per Class},
		ylabel={Test Accuracy (\%)},        
		xmax = 150,
		ymin=35, ymax=100,
		xtick={1,10,20,30,40,50,60,70,80,90,100,110,120,130,140,150},
		ytick={30,40,50,60,70,80,90,95,100},
		x tick label style={font=\tiny, rotate=90},
		legend pos=south east,
		ymajorgrids=true,
		grid style=dashed,
		height = 0.24\textheight,
		width = 0.32\textwidth]
		
		\errorband{chapters/data/limits/classicCNN2L-Dropout-AccuracyVsTrainSetSize.csv}{samplesPerClass}{meanAcc}{stdAcc}{red}{0.4}				
		\end{axis}
		\end{tikzpicture}
	}
	\subfloat[ClassicNet-3-BN]{
		\begin{tikzpicture}
		\begin{axis}[
		xlabel={Samples per Class},
		ylabel={Test Accuracy (\%)},        
		xmax = 150,
		ymin=35, ymax=100,
		xtick={1,10,20,30,40,50,60,70,80,90,100,110,120,130,140,150},
		ytick={30,40,50,60,70,80,90,95,100},
		x tick label style={font=\tiny, rotate=90},
		legend pos=south east,
		ymajorgrids=true,
		grid style=dashed,
		height = 0.24\textheight,
		width = 0.32\textwidth]
		
		\errorband{chapters/data/limits/classicCNN3L-BN-AccuracyVsTrainSetSize.csv}{samplesPerClass}{meanAcc}{stdAcc}{brown}{0.4}
		
		\end{axis}
		\end{tikzpicture}
	}

	\subfloat[ClassicNet-3-Dropout]{
		\begin{tikzpicture}
		\begin{axis}[
		xlabel={Samples per Class},
		ylabel={Test Accuracy (\%)},        
		xmax = 150,
		ymin=35, ymax=100,
		xtick={1,10,20,30,40,50,60,70,80,90,100,110,120,130,140,150},
		ytick={30,40,50,60,70,80,90,95,100},
		x tick label style={font=\tiny, rotate=90},
		legend pos=south east,
		ymajorgrids=true,
		grid style=dashed,
		height = 0.24\textheight,
		width = 0.32\textwidth]
		
		\errorband{chapters/data/limits/classicCNN3L-Dropout-AccuracyVsTrainSetSize.csv}{samplesPerClass}{meanAcc}{stdAcc}{magenta}{0.4}
		
		\end{axis}
		\end{tikzpicture}
	}
	\subfloat[ClassicNet-4-BN]{
		\begin{tikzpicture}
		\begin{axis}[
		xlabel={Samples per Class},
		ylabel={Test Accuracy (\%)},        
		xmax = 150,
		ymin=35, ymax=100,
		xtick={1,10,20,30,40,50,60,70,80,90,100,110,120,130,140,150},
		ytick={30,40,50,60,70,80,90,95,100},
		x tick label style={font=\tiny, rotate=90},
		legend pos=south east,
		ymajorgrids=true,
		grid style=dashed,
		height = 0.24\textheight,
		width = 0.32\textwidth]
		
		\errorband{chapters/data/limits/classicCNN4L-BN-AccuracyVsTrainSetSize.csv}{samplesPerClass}{meanAcc}{stdAcc}{magenta}{0.4}
		
		\end{axis}
		\end{tikzpicture}
	}
	\subfloat[ClassicNet-4-Dropout]{
		\begin{tikzpicture}
		\begin{axis}[
		xlabel={Samples per Class},
		ylabel={Test Accuracy (\%)},        
		xmax = 150,
		ymin=35, ymax=100,
		xtick={1,10,20,30,40,50,60,70,80,90,100,110,120,130,140,150},
		ytick={30,40,50,60,70,80,90,95,100},
		x tick label style={font=\tiny, rotate=90},
		legend pos=south east,
		ymajorgrids=true,
		grid style=dashed,
		height = 0.24\textheight,
		width = 0.32\textwidth]
		
		\errorband{chapters/data/limits/classicCNN4L-Dropout-AccuracyVsTrainSetSize.csv}{samplesPerClass}{meanAcc}{stdAcc}{magenta}{0.4}
		
		\end{axis}
		\end{tikzpicture}
	}	
	\vspace*{0.5cm}
	\caption[Samples per Class versus Accuracy for different ClassicNet configurations]{Samples per Class versus Accuracy for different ClassicNet configurations, varying the number of modules from two to four. Error regions are also displayed.}
	\label{lim:classicNetSPCVsAccuracyMultipleModules}
\end{figure*}

We also obtained experimental results using different module configurations of ClassicNet. We varied the number of modules from two to four, and these results can be seen in Figure \ref{lim:classicNetSPCVsAccuracyMultipleModules}. This figure shows that our results have little variation even as different number of modules are used. Some configurations, like using 3 modules with Batch Normalization, seem to generalize slightly better, which can be seen as accuracy closes up to $98$ \%.

As a summary, we can say that training a convolutional neural network (like ClassicNet) does not require the use of very large datasets, and good results can be obtained with only 30-50 samples per class. Testing accuracy will increase as one adds more samples, but the gains diminish as samples are added, which it can be expected as a natural phenomena. If very high accuracy (over $99$ \%) is desired, then large datasets are needed, and this falls out of the scope of our experimental results.

\FloatBarrier
\section{Combining Transfer Learning with Variations of the Number of Training Samples}

In this section we combine the ideas of Section \ref{lim:secTransferLearning} and Section \ref{lim:secNumTrainingSamples}, into evaluating how transfer learning can be used to make a CNN that can produce good generalization with small number of samples. 

\subsection{Varying the Training Set Size}

In this section we perform the first experiment, which consists of simply splitting the dataset as Section \ref{lim:secTransferLearning} recommended, but we use the splits differently. Our basic idea is that we will vary the number of samples per class in $T_{tr}$, while the rest of the procedure is kept the same. We use $\text{SPC} \in [1,10, 20, 30, ..., 150]$.

Then the idea is to train a CNN model in $F_{tr}$, and then subsample $T_{tr}$ to a given number of samples per class, and then train a multi-class linear SVM on $T_{tr}$ with $C = 1$ and decision surface "one-versus-one" and test this trained SVM on $T_{ts}$, after extracting features again. Motivated by the results produced by an SVM in the previous section, we believe this can show that less samples can be required by a combination of feature learning and an SVM classifier than just using a CNN to do both feature extraction and classification.

We also evaluate the effect of using the same set of objects in $F$ and $T$, or selecting a disjoint set of objects between $F$ and $T$. This could potentially show how learned features generalize outside their training set. We extract features from the \textit{fc1} layer of ClassicNet, as it is the usual approach when performing transfer learning in CNNs \cite{sharif2014cnn}.

\begin{figure*}[t]
	\subfloat[Different Objects]{
		\begin{tikzpicture}
		\begin{axis}[xlabel={SVM Samples per Class},
		ylabel={Test Accuracy (\%)},        
		xmax = 150,
		ymin=40, ymax=100,
		xtick={1,10,20,30,40,50,60,70,80,90,100,110, 120, 130, 140,150},
		ytick={1,10,20,30,40,50,60,70,80,90,95,100},
		x tick label style={font=\tiny, rotate=90},
		legend pos=south east,
		ymajorgrids=true,
		grid style=dashed,
		height = 0.3\textheight,
		width = 0.45 \textwidth,
		legend style={font=\tiny}]
		
		\errorband{chapters/data/limits/classicCNN-BN-TransferLearningVsTrainSetSize-disjointClasses.csv}{spc}{meanAcc}{stdAcc}{blue}{0.4}
		\errorband{chapters/data/limits/classicCNN-Dropout-TransferLearningVsTrainSetSize-disjointClasses.csv}{spc}{meanAcc}{stdAcc}{red}{0.4}
		
		\errorband{chapters/data/limits/classicCNN2L-BN-noSmall-AccuracyVsTrainSetSize.csv}{samplesPerClass}{meanAcc}{stdAcc}{cyan}{0.4}
		\errorband{chapters/data/limits/classicCNN2L-Dropout-noSmall-AccuracyVsTrainSetSize.csv}{samplesPerClass}{meanAcc}{stdAcc}{magenta}{0.4}
		
		\errorband{chapters/data/limits/SVM-C10-noSmall-AccuracyVsTrainSetSize.csv}{samplesPerClass}{meanAcc}{stdAcc}{green}{0.4}
		
		\legend{ClassicNet-BN-TL, ClassicNet-Dropout-TL, ClassicNet-BN, ClassicNet-Dropout, SVM}
		
		\end{axis}
		\end{tikzpicture}
	}
	\subfloat[Same Objects]{
		\begin{tikzpicture}
		\begin{axis}[xlabel={SVM Samples per Class},
		ylabel={Test Accuracy (\%)},        
		xmax = 150,
		ymin=40, ymax=100,
		xtick={1,10,20,30,40,50,60,70,80,90,100,110, 120, 130, 140,150},
		ytick={1,10,20,30,40,50,60,70,80,90,95,100},
		x tick label style={font=\tiny, rotate=90},
		legend pos=south east,
		ymajorgrids=true,
		grid style=dashed,
		height = 0.3\textheight,
		width = 0.45 \textwidth,
		legend style={font=\tiny}]
		
		\errorband{chapters/data/limits/classicCNN-BN-TransferLearningVsTrainSetSize-sameClasses.csv}{spc}{meanAcc}{stdAcc}{blue}{0.4}
		\errorband{chapters/data/limits/classicCNN-Dropout-TransferLearningVsTrainSetSize-sameClasses.csv}{spc}{meanAcc}{stdAcc}{red}{0.4}
		
		\errorband{chapters/data/limits/classicCNN2L-BN-noSmall-AccuracyVsTrainSetSize.csv}{samplesPerClass}{meanAcc}{stdAcc}{cyan}{0.4}
		\errorband{chapters/data/limits/classicCNN2L-Dropout-noSmall-AccuracyVsTrainSetSize.csv}{samplesPerClass}{meanAcc}{stdAcc}{magenta}{0.4}
		
		\errorband{chapters/data/limits/SVM-C10-noSmall-AccuracyVsTrainSetSize.csv}{samplesPerClass}{meanAcc}{stdAcc}{green}{0.4}
		
		\legend{ClassicNet-BN-TL, ClassicNet-Dropout-TL, ClassicNet-BN, ClassicNet-Dropout, SVM}
		
		\end{axis}
		\end{tikzpicture}
	}
	\vspace*{0.5cm}
    \forcerectofloat
	\caption[Samples per Class versus Accuracy for Transfer Learning using an SVM]{Samples per Class versus Accuracy for Transfer Learning using an SVM. In this figure we only vary the number of samples per class used to train a SVM on features learned by ClassicNet.}
	\label{lim:transferLearningSPCVsAccuracy}
\end{figure*}
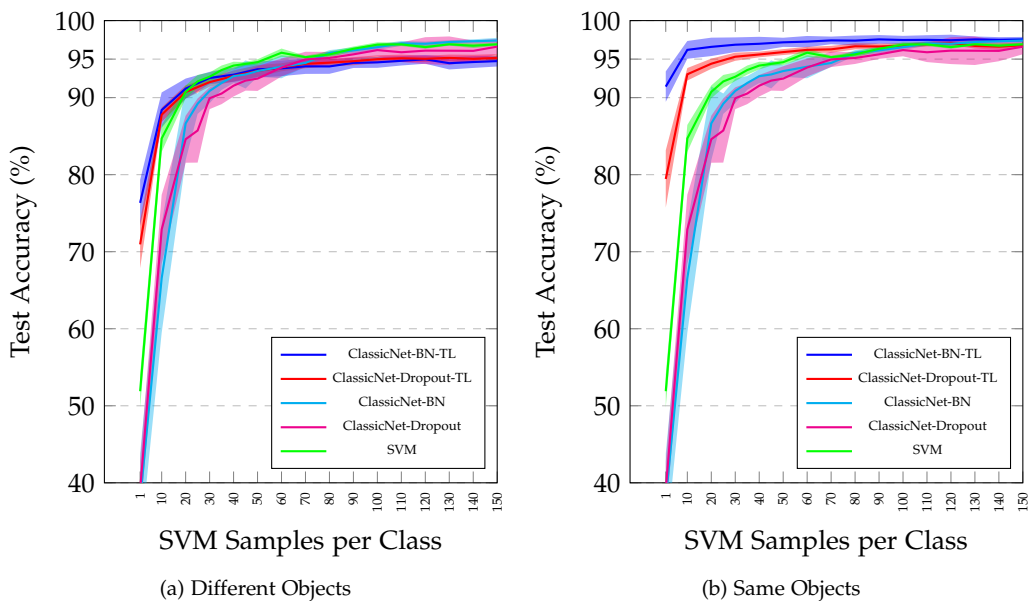

Our results are shown in Figure \ref{lim:transferLearningSPCVsAccuracy}. In this figure we include the results from the previous section as a comparison. For different objects, we can see that learned features both outperform a SVM and the baseline networks by a considerably margin, specially when the number of samples is low. For a single sample per class, ClassicNet-BN-TL produces approximately $76$ \% accuracy, while ClassicNet-Dropout-TL produces $71$ \%. This is a considerably improvement over training a CNN, which produces accuracy no better than $40$ \%.

In the same case, but sharing objects between $F$ and $T$, produces $80$ \% accuracy for the Dropout network, and $91$ \% for the Batch Normalized network. This shows that learning features with a CNN is key to obtaining good generalization, even when the number of samples is small. We believe that these results show that feature learning introduces some additional information that produces invariances into the learned features, which can then be exploited by the SVM trained on those features, producing a better generalization result.

Considering now ten samples per class. In the case of different objects, both networks produce generalization that is very close to $90$ \% accuracy, while for the same objects Dropout produces $93$ \% accuracy, and Batch Normalization $96$ \%. Both are results that can be considered usable for practical applications.

Now considering large sample sizes (more than 30 samples per class), the performance of the learned features is not considerably different from learning a classifier network from the data directly. This means the only advantage of learning features is when one has a small number of samples to train. Only in the case of using the same objects the generalization of feature learning is slightly better than the baselines from the previous section.

\FloatBarrier
\subsection{Varying the Training and Transfer Sets Sizes}

Motivated by the results in the previous section, we now repeat the last experiment, but we vary both the sizes of $F$ and $T$ by means of sub-sampling them to a fixed number of samples per class. We again use $\text{SPC} \in [1,10, 20, 30, ..., 150]$ for sub-sampling both sets.

We perform this experiment in order to know how many samples are actually needed, as we split the original training set into $F$ and $T$, we would like to know how many samples are needed for feature learning ($F$) and how many could potentially be used to train an SVM on those learned features ($T$).

\begin{figure*}[p]
    \vspace*{-3cm}
	\subfloat[1-50]{
		\begin{tikzpicture}
		\begin{axis}[xlabel={SVM Samples per Class},
		ylabel={Test Accuracy (\%)},        
		xmax = 150,
		ymin=40, ymax=100,
		xtick={1,10,20,30,40,50,60,70,80,90,100,110, 120, 130, 140,150},
		ytick={30,40,50,60,70,80,90,95,100},
		x tick label style={font=\tiny, rotate=90},
		legend pos=south east,
		ymajorgrids=true,
		grid style=dashed,
		height = 0.3\textheight,
		width = 0.33 \textwidth,
		legend style={font=\tiny}]
		
		\errorband{chapters/data/classicCNN-BN-TransferLearningVsTrainAndTransferSetSize-disjointClasses-transferSPC1.csv}{spc}{meanAcc}{stdAcc}{blue}{0.4}
		\errorband{chapters/data/classicCNN-BN-TransferLearningVsTrainAndTransferSetSize-disjointClasses-transferSPC10.csv}{spc}{meanAcc}{stdAcc}{red}{0.4}
		\errorband{chapters/data/classicCNN-BN-TransferLearningVsTrainAndTransferSetSize-disjointClasses-transferSPC20.csv}{spc}{meanAcc}{stdAcc}{green}{0.4}
		\errorband{chapters/data/classicCNN-BN-TransferLearningVsTrainAndTransferSetSize-disjointClasses-transferSPC30.csv}{spc}{meanAcc}{stdAcc}{cyan}{0.4}
		\errorband{chapters/data/classicCNN-BN-TransferLearningVsTrainAndTransferSetSize-disjointClasses-transferSPC40.csv}{spc}{meanAcc}{stdAcc}{brown}{0.4}
		
		\legend{Feature 1, Feature 10, Feature 20, Feature 30, Feature 40}
		
		\end{axis}
		\end{tikzpicture}
	}
	\subfloat[60-100]{
		\begin{tikzpicture}
		\begin{axis}[xlabel={SVM Samples per Class},
		xmax = 150,
		ymin=70, ymax=100,
		xtick={1,10,20,30,40,50,60,70,80,90,100,110, 120, 130, 140,150},
		ytick={70,75,80,85,90,95,100},
		x tick label style={font=\tiny, rotate=90},
		legend pos=south east,
		ymajorgrids=true,
		grid style=dashed,
		height = 0.3\textheight,
		width = 0.33 \textwidth,
		legend style={font=\tiny}]
		
		\errorband{chapters/data/classicCNN-BN-TransferLearningVsTrainAndTransferSetSize-disjointClasses-transferSPC60.csv}{spc}{meanAcc}{stdAcc}{blue}{0.4}
		\errorband{chapters/data/classicCNN-BN-TransferLearningVsTrainAndTransferSetSize-disjointClasses-transferSPC70.csv}{spc}{meanAcc}{stdAcc}{red}{0.4}
		\errorband{chapters/data/classicCNN-BN-TransferLearningVsTrainAndTransferSetSize-disjointClasses-transferSPC80.csv}{spc}{meanAcc}{stdAcc}{green}{0.4}
		\errorband{chapters/data/classicCNN-BN-TransferLearningVsTrainAndTransferSetSize-disjointClasses-transferSPC90.csv}{spc}{meanAcc}{stdAcc}{cyan}{0.4}
		\errorband{chapters/data/classicCNN-BN-TransferLearningVsTrainAndTransferSetSize-disjointClasses-transferSPC100.csv}{spc}{meanAcc}{stdAcc}{brown}{0.4}
		
		\legend{Feature 60, Feature 70, Feature 80, Feature 90, Feature 100}
		
		\end{axis}
		\end{tikzpicture}
	}
	\subfloat[110-150]{
		\begin{tikzpicture}
		\begin{axis}[xlabel={SVM Samples per Class},
		xmax = 150,
		ymin=70, ymax=100,
		xtick={1,10,20,30,40,50,60,70,80,90,100,110, 120, 130, 140,150},
		ytick={70,75,80,85,90,95,100},
		x tick label style={font=\tiny, rotate=90},
		legend pos=south east,
		ymajorgrids=true,
		grid style=dashed,
		height = 0.3\textheight,
		width = 0.33 \textwidth,
		legend style={font=\tiny}]
		
		\errorband{chapters/data/classicCNN-BN-TransferLearningVsTrainAndTransferSetSize-disjointClasses-transferSPC110.csv}{spc}{meanAcc}{stdAcc}{blue}{0.4}
		\errorband{chapters/data/classicCNN-BN-TransferLearningVsTrainAndTransferSetSize-disjointClasses-transferSPC120.csv}{spc}{meanAcc}{stdAcc}{red}{0.4}
		\errorband{chapters/data/classicCNN-BN-TransferLearningVsTrainAndTransferSetSize-disjointClasses-transferSPC130.csv}{spc}{meanAcc}{stdAcc}{green}{0.4}
		\errorband{chapters/data/classicCNN-BN-TransferLearningVsTrainAndTransferSetSize-disjointClasses-transferSPC140.csv}{spc}{meanAcc}{stdAcc}{cyan}{0.4}
		\errorband{chapters/data/classicCNN-BN-TransferLearningVsTrainAndTransferSetSize-disjointClasses-transferSPC150.csv}{spc}{meanAcc}{stdAcc}{brown}{0.4}
		
		\legend{Feature 110, Feature 120, Feature 130, Feature 140, Feature 150}
		
		\end{axis}
		\end{tikzpicture}
	}
	\vspace*{0.5cm}
	\caption[Samples per Class versus Accuracy for ClassicCNN-BN Transfer Learning with different objects]{Samples per Class versus Accuracy for ClassicCNN-BN Transfer Learning with different objects. In this figure we vary both the samples per class to train the feature extractor (as different plots) and the samples for training the SVM for the target classes. Note that the scale of each figure is different.}
	\label{lim:tlBothSPCVsAccuracyClassicNetBNDifferent}
\end{figure*}
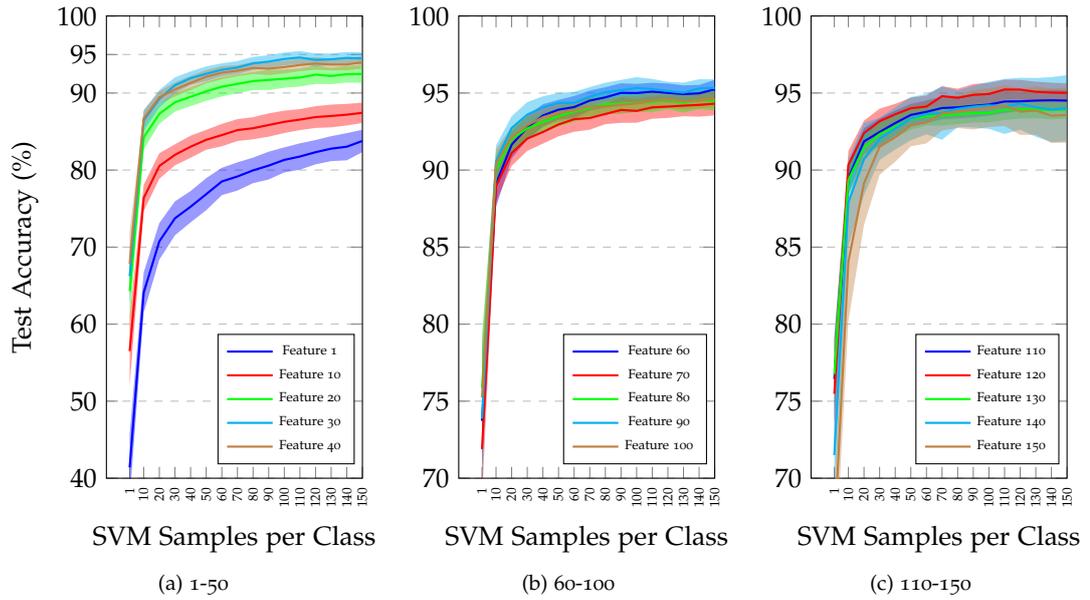

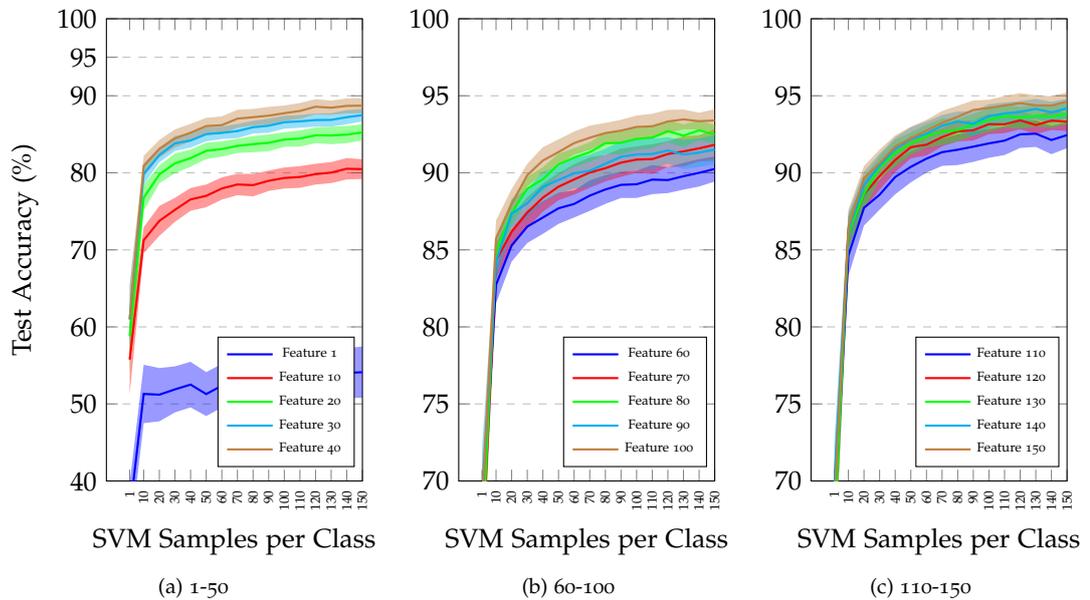
\begin{figure*}[p]
    \vspace*{-3cm}
	\subfloat[1-50]{
		\begin{tikzpicture}
		\begin{axis}[xlabel={SVM Samples per Class},
		ylabel={Test Accuracy (\%)},        
		xmax = 150,
		ymin=40, ymax=100,
		xtick={1,10,20,30,40,50,60,70,80,90,100,110, 120, 130, 140,150},
		ytick={30,40,50,60,70,80,90,95,100},
		x tick label style={font=\tiny, rotate=90},
		legend pos=south east,
		ymajorgrids=true,
		grid style=dashed,
		height = 0.3\textheight,
		width = 0.33 \textwidth,
		legend style={font=\tiny}]
		
		\errorband{chapters/data/classicCNN-Dropout-TransferLearningVsTrainAndTransferSetSize-disjointClasses-transferSPC1.csv}{spc}{meanAcc}{stdAcc}{blue}{0.4}
		\errorband{chapters/data/classicCNN-Dropout-TransferLearningVsTrainAndTransferSetSize-disjointClasses-transferSPC10.csv}{spc}{meanAcc}{stdAcc}{red}{0.4}
		\errorband{chapters/data/classicCNN-Dropout-TransferLearningVsTrainAndTransferSetSize-disjointClasses-transferSPC20.csv}{spc}{meanAcc}{stdAcc}{green}{0.4}
		\errorband{chapters/data/classicCNN-Dropout-TransferLearningVsTrainAndTransferSetSize-disjointClasses-transferSPC30.csv}{spc}{meanAcc}{stdAcc}{cyan}{0.4}
		\errorband{chapters/data/classicCNN-Dropout-TransferLearningVsTrainAndTransferSetSize-disjointClasses-transferSPC40.csv}{spc}{meanAcc}{stdAcc}{brown}{0.4}
		
		\legend{Feature 1, Feature 10, Feature 20, Feature 30, Feature 40}
		
		\end{axis}
		\end{tikzpicture}
	}
	\subfloat[60-100]{
		\begin{tikzpicture}
		\begin{axis}[xlabel={SVM Samples per Class},
		xmax = 150,
		ymin=70, ymax=100,
		xtick={1,10,20,30,40,50,60,70,80,90,100,110, 120, 130, 140,150},
		ytick={70,75,80,85,90,95,100},
		x tick label style={font=\tiny, rotate=90},
		legend pos=south east,
		ymajorgrids=true,
		grid style=dashed,
		height = 0.3\textheight,
		width = 0.33 \textwidth,
		legend style={font=\tiny}]
		
		\errorband{chapters/data/classicCNN-Dropout-TransferLearningVsTrainAndTransferSetSize-disjointClasses-transferSPC60.csv}{spc}{meanAcc}{stdAcc}{blue}{0.4}
		\errorband{chapters/data/classicCNN-Dropout-TransferLearningVsTrainAndTransferSetSize-disjointClasses-transferSPC70.csv}{spc}{meanAcc}{stdAcc}{red}{0.4}
		\errorband{chapters/data/classicCNN-Dropout-TransferLearningVsTrainAndTransferSetSize-disjointClasses-transferSPC80.csv}{spc}{meanAcc}{stdAcc}{green}{0.4}
		\errorband{chapters/data/classicCNN-Dropout-TransferLearningVsTrainAndTransferSetSize-disjointClasses-transferSPC90.csv}{spc}{meanAcc}{stdAcc}{cyan}{0.4}
		\errorband{chapters/data/classicCNN-Dropout-TransferLearningVsTrainAndTransferSetSize-disjointClasses-transferSPC100.csv}{spc}{meanAcc}{stdAcc}{brown}{0.4}
		
		\legend{Feature 60, Feature 70, Feature 80, Feature 90, Feature 100}
		
		\end{axis}
		\end{tikzpicture}
	}
	\subfloat[110-150]{
		\begin{tikzpicture}
		\begin{axis}[xlabel={SVM Samples per Class},
		xmax = 150,
		ymin=70, ymax=100,
		xtick={1,10,20,30,40,50,60,70,80,90,100,110, 120, 130, 140,150},
		ytick={70,75,80,85,90,95,100},
		x tick label style={font=\tiny, rotate=90},
		legend pos=south east,
		ymajorgrids=true,
		grid style=dashed,
		height = 0.3\textheight,
		width = 0.33 \textwidth,
		legend style={font=\tiny}]
		
		\errorband{chapters/data/classicCNN-Dropout-TransferLearningVsTrainAndTransferSetSize-disjointClasses-transferSPC110.csv}{spc}{meanAcc}{stdAcc}{blue}{0.4}
		\errorband{chapters/data/classicCNN-Dropout-TransferLearningVsTrainAndTransferSetSize-disjointClasses-transferSPC120.csv}{spc}{meanAcc}{stdAcc}{red}{0.4}
		\errorband{chapters/data/classicCNN-Dropout-TransferLearningVsTrainAndTransferSetSize-disjointClasses-transferSPC130.csv}{spc}{meanAcc}{stdAcc}{green}{0.4}
		\errorband{chapters/data/classicCNN-Dropout-TransferLearningVsTrainAndTransferSetSize-disjointClasses-transferSPC140.csv}{spc}{meanAcc}{stdAcc}{cyan}{0.4}
		\errorband{chapters/data/classicCNN-Dropout-TransferLearningVsTrainAndTransferSetSize-disjointClasses-transferSPC150.csv}{spc}{meanAcc}{stdAcc}{brown}{0.4}
		
		\legend{Feature 110, Feature 120, Feature 130, Feature 140, Feature 150}
		
		\end{axis}
		\end{tikzpicture}
	}
	\vspace*{0.5cm}
	\caption[Samples per Class versus Accuracy for ClassicCNN-Dropout Transfer Learning with different objects]{Samples per Class versus Accuracy for ClassicCNN-Dropout Transfer Learning with different objects. In this figure we vary both the samples per class to train the feature extractor (as different plots) and the samples for training the SVM for the target classes. Note that the scale of each figure is different.}
	\label{lim:tlBothSPCVsAccuracyClassicNetDropoutDifferent}
\end{figure*}

For this experiment we do not perform any comparison with previous results, as we are pursuing a different question. Results are presented in Figures \ref{lim:tlBothSPCVsAccuracyClassicNetBNDifferent} and \ref{lim:tlBothSPCVsAccuracyClassicNetBNSame} for the Batch Normalized networks, and \ref{lim:tlBothSPCVsAccuracyClassicNetDropoutDifferent} and \ref{lim:tlBothSPCVsAccuracyClassicNetDropoutSame} for the networks using Dropout. In order to facilitate comparison in these figures, we split the variations of sub-sampling $F$ into different plots, aggregated as three sub-figures.

Results with different objects show that a single sample for feature learning performs poorly (as it could be expected), but this effect is much more noticeable with Dropout than with features learned by Batch Normalization. Using Dropout in this case produces generalization that quickly saturates to $50$ \% accuracy, which is far from ideal. The Batch Normalized features perform considerably better, overcoming the $80$ \% barrier without any problem.
Adding more samples to train the feature extractor improves transfer learning performance, which can be seen as features learned over ten samples per class have an improvement of $10$ \% in the Batch Normalization case, and more than $25$ \% in the case of Dropout. It can be seen that adding more samples per class for feature learning quickly saturated and performance increments diminish, starting from 40 samples per class in the Batch Normalization case, and 30 samples per class for Dropout features.

Performance of using a single sample to train the SVM ($T$) over learned features is the one most affected by the number of samples used to learn those features ($F$), as accuracy starts at $40$ \% and increases to $70$ \% with 60-70 samples per class in $F$, but also saturates and stops improving after using over 100 samples.

As the results from the previous section showed, in all cases generalization saturates at $95$ \% accuracy and it does not improve further than this point. In order to reliably obtain such generalization, 150 or more samples per class are needed.

\FloatBarrier
Results using the same objects for feature learning show improved generalization over using different objects. This is acceptable, as the learned features have a natural bias to well represent the learned objects. We believe that this invariance can be considerably improved with more data and variation among object classes.

In this case, achieving $95$ \% accuracy reliably requires only 40 samples per class for feature learning ($F$). Performance at a single sample per class for $T$ also improves considerably with more feature learning samples, starting at $40$ \% and increasing to $80$ \% for 40 samples per class, and it further increases up to $90$ \% when more samples are used for feature learning.

The same case of a single sample for training $T$ shows that Batch Normalization features are superior, as BN produces $50$ \% accuracy versus less than $40$ \% for Dropout. When more samples are added to $F$, single sample $T$ performance improves considerably, reaching more than $80$ \% with BN features and $70$ \% with Dropout. As more samples are used to $F$, performance continues to slowly improve, eventually achieving $98$ \% accuracy reliably with 100 samples per class in $F$. In the case of a large number of samples in $F$, Batch Normalization is still superior, reaching the $98$ \% barrier more consistently than Dropout.

\begin{figure*}[p]
    \vspace*{-3cm}
	\subfloat[1-50]{
		\begin{tikzpicture}
		\begin{axis}[xlabel={SVM Samples per Class},
		ylabel={Test Accuracy (\%)},        
		xmax = 150,
		ymin=40, ymax=100,
		xtick={1,10,20,30,40,50,60,70,80,90,100,110, 120, 130, 140,150},
		ytick={30,40,50,60,70,80,90,95,100},
		x tick label style={font=\tiny, rotate=90},
		legend pos=south east,
		ymajorgrids=true,
		grid style=dashed,
		height = 0.3\textheight,
		width = 0.33 \textwidth,
		legend style={font=\tiny}]
		
		\errorband{chapters/data/classicCNN-BN-TransferLearningVsTrainAndTransferSetSize-sameClasses-transferSPC1.csv}{spc}{meanAcc}{stdAcc}{blue}{0.4}
		\errorband{chapters/data/classicCNN-BN-TransferLearningVsTrainAndTransferSetSize-sameClasses-transferSPC10.csv}{spc}{meanAcc}{stdAcc}{red}{0.4}
		\errorband{chapters/data/classicCNN-BN-TransferLearningVsTrainAndTransferSetSize-sameClasses-transferSPC20.csv}{spc}{meanAcc}{stdAcc}{green}{0.4}
		\errorband{chapters/data/classicCNN-BN-TransferLearningVsTrainAndTransferSetSize-sameClasses-transferSPC30.csv}{spc}{meanAcc}{stdAcc}{cyan}{0.4}
		\errorband{chapters/data/classicCNN-BN-TransferLearningVsTrainAndTransferSetSize-sameClasses-transferSPC40.csv}{spc}{meanAcc}{stdAcc}{brown}{0.4}
		
		\legend{Feature 1, Feature 10, Feature 20, Feature 30, Feature 40}
		
		\end{axis}
		\end{tikzpicture}
	}
	\subfloat[60-100]{
		\begin{tikzpicture}
		\begin{axis}[xlabel={SVM Samples per Class},
		xmax = 150,
		ymin=70, ymax=100,
		xtick={1,10,20,30,40,50,60,70,80,90,100,110, 120, 130, 140,150},
		ytick={70,75,80,85,90,95,98,100},
		x tick label style={font=\tiny, rotate=90},
		legend pos=south east,
		ymajorgrids=true,
		grid style=dashed,
		height = 0.3\textheight,
		width = 0.33 \textwidth,
		legend style={font=\tiny}]
		
		\errorband{chapters/data/classicCNN-BN-TransferLearningVsTrainAndTransferSetSize-sameClasses-transferSPC60.csv}{spc}{meanAcc}{stdAcc}{blue}{0.4}
		\errorband{chapters/data/classicCNN-BN-TransferLearningVsTrainAndTransferSetSize-sameClasses-transferSPC70.csv}{spc}{meanAcc}{stdAcc}{red}{0.4}
		\errorband{chapters/data/classicCNN-BN-TransferLearningVsTrainAndTransferSetSize-sameClasses-transferSPC80.csv}{spc}{meanAcc}{stdAcc}{green}{0.4}
		\errorband{chapters/data/classicCNN-BN-TransferLearningVsTrainAndTransferSetSize-sameClasses-transferSPC90.csv}{spc}{meanAcc}{stdAcc}{cyan}{0.4}
		\errorband{chapters/data/classicCNN-BN-TransferLearningVsTrainAndTransferSetSize-sameClasses-transferSPC100.csv}{spc}{meanAcc}{stdAcc}{brown}{0.4}
		
		\legend{Feature 60, Feature 70, Feature 80, Feature 90, Feature 100}
		
		\end{axis}
		\end{tikzpicture}
	}
	\subfloat[110-150]{
		\begin{tikzpicture}
		\begin{axis}[xlabel={SVM Samples per Class},
		xmax = 150,
		ymin=80, ymax=100,
		xtick={1,10,20,30,40,50,60,70,80,90,100,110, 120, 130, 140,150},
		ytick={70,75,80,85,90,95,97,98,100},
		x tick label style={font=\tiny, rotate=90},
		legend pos=south east,
		ymajorgrids=true,
		grid style=dashed,
		height = 0.3\textheight,
		width = 0.33 \textwidth,
		legend style={font=\tiny}]
		
		\errorband{chapters/data/classicCNN-BN-TransferLearningVsTrainAndTransferSetSize-sameClasses-transferSPC110.csv}{spc}{meanAcc}{stdAcc}{blue}{0.4}
		\errorband{chapters/data/classicCNN-BN-TransferLearningVsTrainAndTransferSetSize-sameClasses-transferSPC120.csv}{spc}{meanAcc}{stdAcc}{red}{0.4}
		\errorband{chapters/data/classicCNN-BN-TransferLearningVsTrainAndTransferSetSize-sameClasses-transferSPC130.csv}{spc}{meanAcc}{stdAcc}{green}{0.4}
		\errorband{chapters/data/classicCNN-BN-TransferLearningVsTrainAndTransferSetSize-sameClasses-transferSPC140.csv}{spc}{meanAcc}{stdAcc}{cyan}{0.4}
		\errorband{chapters/data/classicCNN-BN-TransferLearningVsTrainAndTransferSetSize-sameClasses-transferSPC150.csv}{spc}{meanAcc}{stdAcc}{brown}{0.4}
		
		\legend{Feature 110, Feature 120, Feature 130, Feature 140, Feature 150}
		
		\end{axis}
		\end{tikzpicture}
	}
	\vspace*{0.5cm}
	\caption[Samples per Class versus Accuracy for ClassicCNN-BN Transfer Learning with same objects]{Samples per Class versus Accuracy for ClassicCNN-BN Transfer Learning with same objects. In this figure we vary both the samples per class to train the feature extractor (as different plots) and the samples for training the SVM for the target classes. Note that the scale of each figure is different.}
	\label{lim:tlBothSPCVsAccuracyClassicNetBNSame}
\end{figure*}

\begin{figure*}[p]
    \vspace*{-3cm}
	\subfloat[1-50]{
		\begin{tikzpicture}
		\begin{axis}[xlabel={SVM Samples per Class},
		ylabel={Test Accuracy (\%)},        
		xmax = 150,
		ymin=40, ymax=100,
		xtick={1,10,20,30,40,50,60,70,80,90,100,110, 120, 130, 140,150},
		ytick={30,40,50,60,70,80,90,95,100},
		x tick label style={font=\tiny, rotate=90},
		legend pos=south east,
		ymajorgrids=true,
		grid style=dashed,
		height = 0.3\textheight,
		width = 0.33 \textwidth,
		legend style={font=\tiny}]
		
		\errorband{chapters/data/classicCNN-Dropout-TransferLearningVsTrainAndTransferSetSize-sameClasses-transferSPC1.csv}{spc}{meanAcc}{stdAcc}{blue}{0.4}
		\errorband{chapters/data/classicCNN-Dropout-TransferLearningVsTrainAndTransferSetSize-sameClasses-transferSPC10.csv}{spc}{meanAcc}{stdAcc}{red}{0.4}
		\errorband{chapters/data/classicCNN-Dropout-TransferLearningVsTrainAndTransferSetSize-sameClasses-transferSPC20.csv}{spc}{meanAcc}{stdAcc}{green}{0.4}
		\errorband{chapters/data/classicCNN-Dropout-TransferLearningVsTrainAndTransferSetSize-sameClasses-transferSPC30.csv}{spc}{meanAcc}{stdAcc}{cyan}{0.4}
		\errorband{chapters/data/classicCNN-Dropout-TransferLearningVsTrainAndTransferSetSize-sameClasses-transferSPC40.csv}{spc}{meanAcc}{stdAcc}{brown}{0.4}
		
		\legend{Feature 1, Feature 10, Feature 20, Feature 30, Feature 40}
		
		\end{axis}
		\end{tikzpicture}
	}
	\subfloat[60-100]{
		\begin{tikzpicture}
		\begin{axis}[xlabel={SVM Samples per Class},
		xmax = 150,
		ymin=70, ymax=100,
		xtick={1,10,20,30,40,50,60,70,80,90,100,110, 120, 130, 140,150},
		ytick={70,75,80,85,90,95,98,100},
		x tick label style={font=\tiny, rotate=90},
		legend pos=south east,
		ymajorgrids=true,
		grid style=dashed,
		height = 0.3\textheight,
		width = 0.33 \textwidth,
		legend style={font=\tiny}]
		
		\errorband{chapters/data/classicCNN-Dropout-TransferLearningVsTrainAndTransferSetSize-sameClasses-transferSPC60.csv}{spc}{meanAcc}{stdAcc}{blue}{0.4}
		\errorband{chapters/data/classicCNN-Dropout-TransferLearningVsTrainAndTransferSetSize-sameClasses-transferSPC70.csv}{spc}{meanAcc}{stdAcc}{red}{0.4}
		\errorband{chapters/data/classicCNN-Dropout-TransferLearningVsTrainAndTransferSetSize-sameClasses-transferSPC80.csv}{spc}{meanAcc}{stdAcc}{green}{0.4}
		\errorband{chapters/data/classicCNN-Dropout-TransferLearningVsTrainAndTransferSetSize-sameClasses-transferSPC90.csv}{spc}{meanAcc}{stdAcc}{cyan}{0.4}
		\errorband{chapters/data/classicCNN-Dropout-TransferLearningVsTrainAndTransferSetSize-sameClasses-transferSPC100.csv}{spc}{meanAcc}{stdAcc}{brown}{0.4}
		
		\legend{Feature 60, Feature 70, Feature 80, Feature 90, Feature 100}
		
		\end{axis}
		\end{tikzpicture}
	}
	\subfloat[110-150]{
		\begin{tikzpicture}
		\begin{axis}[xlabel={SVM Samples per Class},
		xmax = 150,
		ymin=80, ymax=100,
		xtick={1,10,20,30,40,50,60,70,80,90,100,110, 120, 130, 140,150},
		ytick={70,75,80,85,90,95,97,98,100},
		x tick label style={font=\tiny, rotate=90},
		legend pos=south east,
		ymajorgrids=true,
		grid style=dashed,
		height = 0.3\textheight,
		width = 0.33 \textwidth,
		legend style={font=\tiny}]
		
		\errorband{chapters/data/classicCNN-Dropout-TransferLearningVsTrainAndTransferSetSize-sameClasses-transferSPC110.csv}{spc}{meanAcc}{stdAcc}{blue}{0.4}
		\errorband{chapters/data/classicCNN-Dropout-TransferLearningVsTrainAndTransferSetSize-sameClasses-transferSPC120.csv}{spc}{meanAcc}{stdAcc}{red}{0.4}
		\errorband{chapters/data/classicCNN-Dropout-TransferLearningVsTrainAndTransferSetSize-sameClasses-transferSPC130.csv}{spc}{meanAcc}{stdAcc}{green}{0.4}
		\errorband{chapters/data/classicCNN-Dropout-TransferLearningVsTrainAndTransferSetSize-sameClasses-transferSPC140.csv}{spc}{meanAcc}{stdAcc}{cyan}{0.4}
		\errorband{chapters/data/classicCNN-Dropout-TransferLearningVsTrainAndTransferSetSize-sameClasses-transferSPC150.csv}{spc}{meanAcc}{stdAcc}{brown}{0.4}
		
		\legend{Feature 110, Feature 120, Feature 130, Feature 140, Feature 150}
		
		\end{axis}
		\end{tikzpicture}
	}
	\vspace*{0.5cm}
	\caption[Samples per Class versus Accuracy for ClassicCNN-Dropout Transfer Learning with same objects]{Samples per Class versus Accuracy for ClassicCNN-Dropout Transfer Learning with same objects. In this figure we vary both the samples per class to train the feature extractor (as different plots) and the samples for training the SVM for the target classes. Note that the scale of each figure is different.}
	\label{lim:tlBothSPCVsAccuracyClassicNetDropoutSame}
\end{figure*}

Two clear conclusions can be obtained from these experiments: High generalization ($95$ \% accuracy) can be achieved with small samples (10-30 samples per class with for both $T$ and $F$) but only if the same objects are used for both sets. This implies that generalization outside of the training set will probably be reduced. The second conclusion is that if $T$ and $F$ do not share objects, there will be a performance hit compared to sharing objects, but this case still learning features will improve generalization when compared to training a CNN over the same data.

It has to be mentioned that our results show that by using the same data, but changing the training procedure, a considerable improvement in generalization can be obtained, even when using low samples to learn features ($F$) and to train a SVM on those features ($T$).

\section{Summary of Results}

In this chapter we have explored different limitations in the use of convolutional neural networks with forward-looking sonar data.

First we evaluated how transfer learning performs in these images with varying neural networks and layer configurations. We found out that all layers produce very good features that can discriminate classes with good accuracy, but as depth increases, features become slightly less discriminative, which was unexpected. The best features are produced by layers that are close to the input.

Then we evaluated how changing the input size affects generalization. We found that ClassicNet can be trained to have the same generalization independent of the object size, but TinyNet and FireNet exhibit decreasing accuracy as objects become bigger. This was unexpected and shows that these networks require more training data than ClassicNet. Our results also indicate that it is possible to also reduce the input image size as a way to reduce the number of parameters and computation required, improving computational performance.

We also have evaluated the relationship between the number of training samples and generalization produced by a CNN. ClassicNet scales quite well with the number of samples per class in the training set, and requires 30-50 samples per class to reach $90$ \% accuracy. Training using Dropout seems to be slightly better than Batch Normalization in the low sample case, but Batch Normalization is better when many samples are available. TinyNet and FireNet scale poorly with the number of samples, producing less generalization than ClassicNet. This confirms our previous results that pointed that these networks require more training data than ClassicNet, even as they have less parameters. In theory, networks with less parameters require less data to be trained, but these models seem to require more data for a given accuracy target.

Finally we evaluated the combination of feature learning and how it affects generalization as a function of the size of the training set. We learn features in one part of the dataset, and use the other part to train a linear SVM that is evaluated on a test set. Our results show that learning features on a dataset that shares objects, accuracy increases to over $90$ \% when using a single sample per class to train an SVM. If feature learning is performed on a different set of objects, then single image per class accuracy can only reach $70-80$ \%, but it is  still a considerable improvement over training the network on the same sub-sampled dataset.

Our last experiment evaluated transfer learning by varying both the samples per class in the feature learning dataset ($F$) and the SVM training dataset ($T$). We found out that high generalization, at $95$ \% accuracy, can be obtained with small datasets in the order of $10-30$ samples per class for $F$ and $T$, but only if the same objects are used in both datasets. In the case of learning features in one set of objects, and training an SVM for a different one, then more data is required to achieve $95$ \% accuracy, in the order of 100 $T$ samples per class and $40-50$ feature learning ($F$) samples.

We expect that our results will contribute to the discussion about how many samples are actually required to use Deep Neural Networks in different kinds of images. For the marine robotics community, we expect that our argument is convincing and more use of neural networks can be seen on the field.

%% file: chapters/matching-patches-sonar.tex
\chapter{Matching Sonar Image Patches}
\label{chapter:matching}

Matching in Computer Vision is the task of deciding whether two images show the same viewpoint of a scene, or contain the same object. The task becomes considerably harder when one has to consider variations in viewpoint, lighting conditions, occlusion, shading, object shape, scale, and background. This problem is similar to Image Registration, where two images are aligned to match their contents. But instead of aligning images, Matching is concerned about a binary decision without resorting to aligning the images. In extreme cases, alignment may not be possible.

Matching is a fundamental problem in Computer Vision \cite{szeliski2010computer}, as it is a vital component of many other tasks, such as stereo vision, structure from motion, and image query/retrieval. Comparison of image patches and prediction of a similarity measure is a common requirement for these

In Robot Perception, Matching is also a fundamental problem with many applications. For example, a Robot might be shown an object and then asked to find it inside a room. This requires the Robot to learn an appropriate representation of the object such as it can be matched to what the Robot is seeing in the scene. Some use cases that requires such functionality are:

\begin{description}
	\item[\textbf{Object Recognition}] \hfill \\
		Given an image, classify its contents into a predefined set of classes. Instead of using a trainable classifier, recognition can be performed by using a database of labeled images and match the input image with one in the labeled database. This approach has the advantage of being dynamic, so new object classes can be easily added to the database, but recognition performance now depends on the quality of the matching functionality.
	\item[\textbf{Object Detection}] \hfill \\
		Similar to Object Recognition, but instead of classifying the complete image, specific objects must be localized in the image. Typically bounding boxes are generated that correspond to the object's localizations in the image.
		Detection can be implemented with Matching by generating candidate locations with a sliding window or a proposal algorithm, and match each of them to a database of labeled images.
	\item[\textbf{Simultaneous Localization and Mapping}] \hfill \\
		SLAM is a fundamental localization problem in Robotics, where a Robot simultaneously builds a map of its environment and localizes itself relative to the map \cite{Cadena16tro-SLAMfuture}. Most SLAM formulations use landmarks in the environment. These landmarks must be tracked and matched to previously seen ones in order for the robot to localize itself, which is a data association problem. This could be done by matching sonar image patches in a underwater SLAM implementation \cite[1em]{hidalgo2015review}.
	\item[\textbf{Tracking}] \hfill \\
		Given a set of images and an object of interest, localize the object in the first image and re-localize the same object in posterior images in the set, even as the object's appearance, background or lighting conditions might change \cite{yilmaz2006object}. Tracking can be implemented by detecting the object of interest in an image, and then detecting candidate objects (through a detection proposal algorithm) and match candidates with the initial tracking target.
\end{description}

While many of the use cases described in this chapter are a field on its own, with mature techniques, using matching to perform such tasks does offer some potential advantages:

\begin{itemize}
	\item Most modern techniques use Machine Learning, and in general, after training an ML model it is difficult to add new data or classes to the model. Normally any model modification requires retraining, which is computationally expensive. Using image matching does not require task-specific training.
	\item Depending on the performance of the matching function, it could better generalize to unseen data, when compared to methods that are trained to specific datasets. One example are keypoint matching methods like SIFT \cite[-3em]{lowe2004distinctive} and SURF \cite[1em]{bay2006surf}, which have been used in many different domains with varying success.
	\item For practical purposes, any useful image matching function can match objects that have never been seen before, which potentially can increase task robustness.
\end{itemize}

For images produced by sonar, matching is currently an open problem. Descriptors and keypoint detectors designed for color images have been previously used for sonar images, but in general they perform poorly. This is expected, as color/optical images are radically different from the acoustic images produced by a sonar.

In this chapter we introduce the use of CNNs to match patches extracted from a sonar image. We formulate the matching problem as binary classification or regression of a similarity score. We show that using CNNs for matching with this setup considerably improves matching performance, moving it from close-to-random-chance to over $90$ \% chance that the correct matching decision will be made. This means that matching as similarity or binary decisions is now usable for real world applications.

\section{Related Work}

Reliably matching sonar images has been an open problem for a long time, but not many publications mention it. It is well known that matching sonar image is considerably more difficult than other kinds of images \cite{negahdaripour2011dynamic}, due to the unique properties of acoustic sensing.

The first approaches to match sonar images use keypoint-based algorithms \cite[1em]{szeliski2010computer}. Many computer vision problems require finding "interesting" or relevant points in an input image, in order to perform tasks such as image alignment and object recognition. The idea of such points is that they are inherent properties of the image, the objects in the image, and/or the environment. These salient features of the image are typically invariant to many transformations, making them very powerful. 

A keypoint is just one of these salient or interesting points in the image, but defining such kind of points in an algorithmic way is not easy. One of the most simple approaches is to detect corners in the image via the Harris corner operator.

Once a keypoint has been detected, in order to be able to find the same keypoint in another image (potentially a different view of the object/scene), a way to match such keypoints must be devised. The most typical approach is to extract a feature vector from the neighbourhood of the keypoint position, and store it in order to perform distance-based matching. Then if many keypoints match across two images, a matching decision can be made by imposing a threshold on this number of matches. This is a very simple but popular approach.

Scale Invariant Feature Transform (SIFT), introduced by David Lowe \cite{lowe2004distinctive}, is one of the first keypoint algorithms that provides reliable matching across different viewpoints of the scene. SIFT uses a scale space ($\sigma$) representation of the image, computed as Difference of Gaussians (DoG), in order to approximate the Laplacian of Gaussians function. The input image $I$ is blurred using a Gaussian filter $G$ at different scales forming a sub-octave. Adjacent sub-octave levels are subtracted to produce the DoG images, and then the input image is down-sampled and the process repeated, producing a Difference of Gaussian Pyramid. Finding keypoints at multiple scales introduces a degree of scale invariance.

\begin{align}
	D(x, y, \sigma) &= L(x, y, \sigma) - L(x, y, \beta \sigma)\\
	L(x, y, \sigma) &= G(x, y, \sigma) * I(x, y)
\end{align}

Then in scale space keypoints are found by comparing the DoG values $D(x, y, \sigma)$, both spatially and in scale (26 neighbours), to the pixel in the centre. If it is an extrema (either a minimum or maximum) then that spatial location is selected as a potential keypoint. Final keypoints are determined by a stability criteria (removing low contrast, noisy and edge keypoints), and interpolated to a more accurate location by using a second-order Taylor expansion.

Then each keypoint is assigned a dominant orientation by estimating the most common gradient orientation around a neighbourhood of the keypoint. This makes sure the descriptor is approximately orientation invariant. Then the image at the detected keypoint scale is rotated around the dominant orientation and a histogram descriptor is computed, producing a 128-element normalized vector.

While SIFT is very popular for color images, its performance on sonar images is quite poor. Some publications \cite[-2em]{vandrish2011side} claim that it works reliably on their data and task, but our own results show that it works quite poorly on our forward-looking sonar data.

SURF \cite[1em]{bay2006surf} is an approximate algorithm based on SIFT, with the aims of a faster implementation. SURF uses averaging filters as an approximation to the Gaussian filters, as it can be implemented with integral images, massively improving performance. The feature vector is computed as the sum of Haar wavelets, which also can be efficiently computed using integral images.

Many publications related to matching sonar images concentrate on slightly different problems, such as mosaicing and image registration. Mosaicing is a popular application of interest point matching, specially in sonar images, that requires registering two or more images accurately in order to blend image pixels, as the image overlap can be used to obtain average pixel values, which decrease sonar image noise and can reveal underlying structure hidden by noise \cite{hurtos2014real}. This operation does require binary matching decisions, but most research work concentrates on the application rather than the technique.

Kim et al. \cite[1em]{kim2005mosaicing} uses the Harris Corner detector to detect feature points on a side-scan sonar image, for the task of matching interest points in two images in order to register them to build a mosaic. After detecting interest points, these are matched between images using a cross-correlation operation (Eq \ref{sic:ccSimilarityEq}) and a minimum feature matching threshold. This approach is generally based on Zhang et al. \cite{zhang1995robust}. The difference between the two techniques is that Kim et al. uses an adaptive threshold based on the $k$-th percentile, while Zhang et al. uses a direct threshold of the correlation score. Both techniques seem to work appropriately to build mosaics, specially as Kim et al. uses RANSAC to estimate a transformation between the images. Note that patch matching for registration in general is not hard, as the transformations between the two images are usually small, as they typically correspond to consecutive image frames in a stream of images produced by the sensor.

Negahdaripour et al. \cite{negahdaripour2011dynamic} also performed matching for image registration with the purpose of mosaicing forward-looking sonar images. Their technique is not fully explained in their paper, and it is described as performing shadow analysis to classify between seafloor and water-column objects, followed by identification of stationary and moving objects. Stationary objects are clustered and their features are used to compute inliers for image registration. Keypoints and used features are based on SIFT. This work good looking mosaics, but as the authors mention, it requires a lot of domain specific knowledge to produce good results, and many parts of the pipeline are hard to train and implement, specially finding moving objects and determining which parts of the image correspond to the seafloor.

Vandrish et al. \cite[-2em]{vandrish2011side} has compared different kinds of registration methods, broadly categorized into global methods, feature-based, and hybrid approaches. Two global methods were evaluated mutual information and phase correlation. The mutual information between two random variables $X$ and $Y$ is:

\begin{align}
	I(X, Y) &= H(X) + H(Y) - H(X, Y)\\		
	H(X) 	&= -\sum_{x \in X} p(x) \log(p(x))\\
	H(X, Y) &= -\sum_{x, y \in X, Y} p(x, y) \log(p(x, y))
\end{align}

Where $H(X)$ is the entropy of $X$ and $H(X, Y)$ is the joint entropy of $X$ and $Y$. This method then tries to find a transformation $T$ that maximizes $I(X, T(Y))$. The authors claim that when the images are maximally aligned, the mutual information is also maximized. The authors mention that this method is very slow, and that it fails if the image pairs do not share enough intensity variation, specially in areas where the background is homogeneous.

Phase Correlation registration works by computing the cross-power spectrum of two images $X$ and $Y$:

\begin{equation}
	e^{j2\pi(\beta u_o + \nu v_0)} = \frac{F_X(\beta, \nu)F^{*}_Y(\beta,\nu)}{||F_X(\beta, \nu)F_Y(\beta,\nu)||}
	\label{mat:cpSpectra}
\end{equation}

Where $F_s$ is the Fourier transform of $s$. This equation comes from the Fourier shift theorem, associating the Fourier transforms of two images that are related by a spatial shift $(u_o, v_o)$ with a constant multiplicative factor. The phase correlation function is computed as the inverse Fourier transform of Eq \ref{mat:cpSpectra}. This function will have a peak at $(u_o, v_o)$, and by finding the maxima it is possible to recover this shift. For the case of a rotation and a translation, the same principle applies, as the spectra is also rotated. To recover the rotation shift, the same method can be applied to the log-polar transformation of the Fourier transform. The authors mention that this method is the most accurate across all evaluated methods. This result is consistent with other publications on the topic \cite{hurtos2014real}.

Only one feature-based registration method was evaluated, corresponding to SIFT. The authors detect SIFT keypoints in both images and try to match their feature descriptors using a nearest neighbour search. The authors experiments show that many false matches are produced in sonar images. RANSAC is applied to discard outliers while fitting a transformation to align both images. The results are good but only when a large number of keypoints are detected. The method fails when keypoints are sparse or when they are mismatched, requiring methods that iteratively refine the learned transformation parameters.

Two hybrid methods were evaluated. The first is a combination of the log-polar transformation:

\begin{align}
	r 		&= \sqrt{(x - x_c)^2 + (y - y_c)^2}\\
	\theta 	&= \tan^{-1}\left ( \frac{y - y_c}{x - x_c} \right )
\end{align}

Where $(x_c, y_c)$ represents the polar center. This transforms the input images in order to provide some kind of normalization, as image rotations are represented as linear shifts in polar coordinates. Then normalized cross-correlation (as presented previously in Eq \ref{sic:ccSimilarityEq}) is applied to obtain the different shifts by using a sliding window and keeping the transformation parameters that maximize the cross-correlation. The second method is Region-of-Interest detection with a custom operator that detects regions of interest through a variance saliency map. Both methods perform poorly, as the correlation between features in multiple observations is quite low, leading to false matches.

Pham and Gueriot \cite{pham2013guided} propose the use of guided block-matching for sonar image registration. The method we propose in this chapter could be considered similar to the block matching step, as this step just needs to make a binary decision whether two pairs of blocks (image patches) match or not. The authors use a pipeline that first extracts dense features from the image, performs unsupervised segmentation of the image through a Self-Organizing map (SOM) applied on the computer features.
The unsupervised segmentation of the image then is used to aid the block matching process, as only blocks that are similar in feature space according to the SOM will be compared, but this process is unsupervised, which implies that different comparison functions will be learned for each image. Then the final step is to estimate a motion vector from the matched blocks, from where a geometrical transformation for registration can be estimated.

The results from this paper show that it performs considerably faster than standard block matching, but only visual results are provided. The displayed mosaics make sense, showing that standard and guided block matching can recover the correct translation vector, but a more through numerical comparison is needed.

Moving into modern methods for matching, Zagoryuko and Komodakis \cite{zagoruyko2015learning} were one of the first to propose the use of CNNs to learn an image matching function, but they instead learn how to compare image patches, corresponding to predicting a similarity score instead of a plain binary classification. We base our own matching networks on this work.

This work defined several CNN models that can be used for patch matching:

\begin{itemize}
	\item \textbf{Siamese Network}: This is a neural network with two branches, where each branch shares weights \cite{bromley1994signature}, as the idea is to compute a relevant feature vector from the image. The computed feature vectors from each branch are then merged and processed into a decision network that outputs the binary matching decision.
	\item \textbf{Pseudo-Siamese Network}: This is a siamese network but in this case each branch does not share weights. This increases the number of learnable parameters, but the authors do not mention any other advantage or intuitive foundation.
	\item \textbf{Two-Channel Network}: This architecture does not have an intrinsic concept of feature vector, as the two input images are combined channel-wise and this two-channel image is input to a single branch CNN that includes a decision network.
	\item \textbf{Central-Surround Two Stream Network}: This architecture is more complex, as it considers two different spatial resolutions from the input image. This is a combination of the two-channel and siamese architectures. The central stream consists of a central crop of each input image, given to one branch, while the surround stream takes two down-sampled (2x) patches as input. The idea of this architecture is to process multiple scale information at the same time.
	\item \textbf{SPP-based}: This network is based on Spatial Pyramid Pooling \cite{he2014spatial}, which is a pooling technique that can make a CNN accept variable-sized inputs. The basic idea of SPP is that instead of performing pooling with fixed-size regions, the regions are variable sized, but the number of output regions is fixed. This produces a fixed-length pooling output, even as the input size is variable. The authors applied SPP to their networks in order to match variable-size patches.
\end{itemize}

These networks are trained on a dataset presented by Brown et al. \cite{brown2011discriminative}. This dataset contains approximate half million $64 \times 64$ labeled image patches, evenly balanced between positives and negatives, that were captured using real 3D correspondences obtained from multi-view stereo depth. The authors evaluate the ROC curves from binary classification, as well as the false positive ratio at $95$ \% true positive rate. The networks are trained with a L2 regularized hinge loss:

\begin{equation}
	L = \frac{\lambda}{2} ||w|| + \sum_i \max(0, 1 - y_i \hat{y}_i)
\end{equation}

Their results show that the best performing network configuration is the two-channel central-surround one, with a considerably margin over all other configurations. The plain two-channel networks perform slightly worse. As two-channel networks are considerably simpler to implement, this is one choice that we made for our own matching networks in this chapter. These results are quite good, but they seem to only be possible due to the large labeled dataset that is available.

The authors also test their matching networks for wide-baseline stereo on a different dataset, showing superior performance when compared with DAISY \cite{tola2008fast}. These results show that using a CNN for matching is quite powerful and can be used for other tasks. Generalization outside of the training set is quite good.

Zbonar and LeCun \cite[1em]{zbontar2016stereo} also propose the use of a CNN to compare image patches. Their method is specifically designed for wide-baseline stereo matching. Given a stereo disparity map, the authors construct a binary classification dataset by extracting one positive and one negative example where true disparity is known a priori.

The authors evaluate two kinds of CNN architectures. The fast architecture is a siamese network with a dot product (cosine) similarity computed between features produced by each branch, while the accurate architecture uses several fully connected layers where the input is a concatenation of the feature vectors produced by each branch. The networks are trained with a hinge loss.

The raw decisions from the CNN models are not good to produce accurate disparity maps, so additional post-processing is used to produce good results. This method was evaluated on the KITTI dataset, and the leaderboard as by October 2015 showed that it was the best performing method, with an error rate of $2.43$ \% with the accurate architecture, and $2.78$ \% with the fast one.

Finally, there is also an application of CNNs to learn keypoint detection, as the Learned Invariant Feature Transform (LIFT) by Yi et al. \cite[-5em]{yi2016lift}. This method train a CNN with four different stages that detect keypoints by producing a dense score map. Keypoints are selected by using a \textit{softargmax} function that finds the maxima of the score map. A spatial transformer network \cite[-3em]{jaderberg2015spatial} is then used to crop a patch from the input image, in the position given by the maxima in the score map.
Then a orientation estimation network estimates the orientation of the image patch, in order to normalize and introduce a degree of orientation invariance. The estimated orientation is then used by a second spatial transformer network to produce a normalized patch that is input to a description network, which outputs a feature vector.

This method mimics the standard SIFT-like feature detection and description pipelines, but the big difference is that all operations are fully differentiable and the whole pipeline can be trained end-to-end. The authors do mention that given the data they had, end-to-end training failed, but training different parts in a hierarchical way. First the descriptor is trained, then the orientation estimator given the descriptor, then the detector given the two previous parts. Different loss functions are used to train the various components in the pipeline.

LIFT was evaluated against classic keypoint detectors, namely SIFT, SURF, ORB, etc. In matching scores, LIFT outperforms all methods by a considerably margin. LIFT obtains matching scores $0.317-0.374$, while SIFT obtains $0.272-0.283$, SURF gets $0.208-0.244$ and ORB performs poorly at $0.127-0.157$. When considering repeatability LIFT obtains $0.446$ while SIFT gets $0.428$. The nearest neighbours area under the Precision-Recall curve is $0.686$ (SIFT obtains $0.517$).

While we do not use LIFT, we believe that it is a major milestone in keypoint detection and feature matching, as this method can learn domain and sensor specific features that could significantly improve the performance of SIFT-like algorithms in sonar images. Training such a method probably requires large quantities of data, and special labels, but it is a clear future direction.

\subsection{Discussion}

While there is a rich literature on mosaicing and registration for sonar images, these approaches typically use simplistic method for image patch matching. The most complex technique that is commonly used is SIFT, which was not designed specifically for sonar images.

Many techniques use extensive domain knowledge in order to obtain good registration performance, but in general evaluation is quite simplistic. Only visual confirmation that the method works in a tiny dataset (less than 10 images) and no further numerical comparison is performed. This is common in the marine robotics community, but that does not mean it is right.

We have not found any technique working on sonar that is similar to the work presented in this chapter. Most research is concentrated around mosaicing and registration, and no other applications that could profit from comparing sonar image patches have been explored. This can easily be explained by the lack of techniques that can reliably match sonar image patches. SIFT and other keypoint matching methods perform poorly according to our own experiments (presented later in Table \ref{mat:matchingResultsTable}). Use of such techniques will produce results that are not publishable.

Our evaluation of CNN-based methods for matching shows that this approach is promising. While it has been applied for tasks that are not common in underwater domains (stereo vision for example), we believe that it can trigger the development of new applications. As we have previously mentioned, object detection and recognition without training sets, SLAM, and Tracking can all benefit from such technique.

\section{Dataset Construction and Preprocessing}

As we do not possess a dataset that is specifically collected and built to train an patch matching algorithm, we decided to build our own from the classification dataset that we possess.

Exploiting the fact that we captured a small set of objects from multiple views, we use this fact to produce synthetic matching pairs where the matching label (positive or negative) can be deduced from the classification class labels. Note that this is not the best way to produce such a dataset. For example, using an additional distance sensor can be used to obtain precise matching labels to produce a much larger dataset. This approach has been used by Aanaes et al. \cite{aanaes2012interesting} to evaluate keypoints detectors, but we have not taken such approach as it is costly and requires a special experimental setup.

We generate three kinds of matching pairs:

\begin{description}
	\item[\textbf{Positive Object-Object}] A positive match pair is constructed from two randomly sampled objects that share a class label. Each object crops can typically correspond to different perspectives of the same object or different insonification levels from the sonar sensor.
	\item[\textbf{Negative Object-Object}] A negative match pair is constructed from two randomly sampled objects that have different class labels.
	\item[\textbf{Negative Object-Background}] A negative match pair is constructed from one randomly sampled object and another randomly sampled background patch with IoU score less than 0.1 with ground truth. We assume that all objects in the dataset are labeled and there is only a small chance that a random background patch might contain an object.
\end{description}

The number of negative match pairs is unbounded, specially for object-background pairs. As a balanced dataset is preferred, we equalize the number of positive and negative matching pairs. To build our dataset, for each labeled object we samples 10 positive object matches, 5 negative object matches, and 5 negative background matches. We use $96 \times 96$ image crops to construct our pairs.

In order to evaluate the generalization ability of our models, we built two datasets, one that shares objects between all splits (train, test and validation), and one that does not share objects. Dataset D (different objects) contains 47280 image pairs, while Dataset S (same objects) contains 54720 image pairs. We split these datasets as $70$ \% training, $15 $ \% validation and $15$ \% testing.

\begin{figure*}
	\centering
	\subfloat[Object-Object Positive Matches]{
		\parbox{0.98\textwidth}{
			\includegraphics[width=0.11 \textwidth]{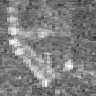}
			\includegraphics[width=0.11 \textwidth]{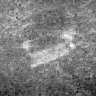} \;
			\includegraphics[width=0.11 \textwidth]{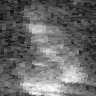}
			\includegraphics[width=0.11 \textwidth]{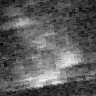} \;
			\includegraphics[width=0.11 \textwidth]{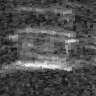}
			\includegraphics[width=0.11 \textwidth]{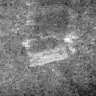}
            
			\vspace*{0.3cm}
            
			\includegraphics[width=0.11 \textwidth]{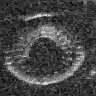}
			\includegraphics[width=0.11 \textwidth]{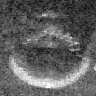} $\;$
			\includegraphics[width=0.11 \textwidth]{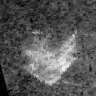}
			\includegraphics[width=0.11 \textwidth]{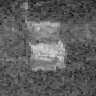} $\;$
			\includegraphics[width=0.11 \textwidth]{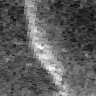}
			\includegraphics[width=0.11 \textwidth]{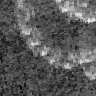}
		}	
	}

	\subfloat[Object-Object Negative Matches]{
		\parbox{0.98\textwidth}{
			\includegraphics[width=0.11 \textwidth]{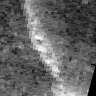}
			\includegraphics[width=0.11 \textwidth]{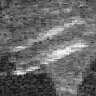} \;
			\includegraphics[width=0.11 \textwidth]{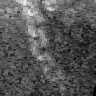}
			\includegraphics[width=0.11 \textwidth]{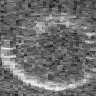} \;
			\includegraphics[width=0.11 \textwidth]{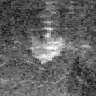}
			\includegraphics[width=0.11 \textwidth]{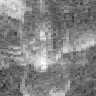}
            
			\vspace*{0.3cm}
            
			\includegraphics[width=0.11 \textwidth]{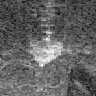}
			\includegraphics[width=0.11 \textwidth]{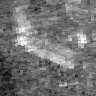} \;
			\includegraphics[width=0.11 \textwidth]{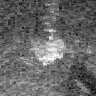}
			\includegraphics[width=0.11 \textwidth]{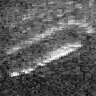} \;
			\includegraphics[width=0.11 \textwidth]{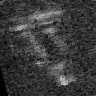}
			\includegraphics[width=0.11 \textwidth]{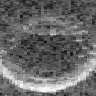}
		}
	}

	\subfloat[Object-Background Negative Matches]{
		\parbox{0.98\textwidth}{
			\includegraphics[width=0.11 \textwidth]{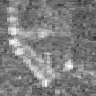}
			\includegraphics[width=0.11 \textwidth]{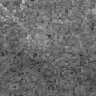} \;
			\includegraphics[width=0.11 \textwidth]{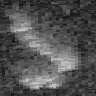}
			\includegraphics[width=0.11 \textwidth]{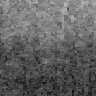} \;
			\includegraphics[width=0.11 \textwidth]{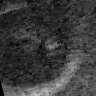}
			\includegraphics[width=0.11 \textwidth]{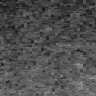}
            
			\vspace*{0.3cm}
            
			\includegraphics[width=0.11 \textwidth]{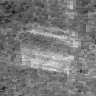}
			\includegraphics[width=0.11 \textwidth]{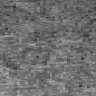} \;
			\includegraphics[width=0.11 \textwidth]{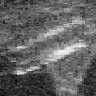}
			\includegraphics[width=0.11 \textwidth]{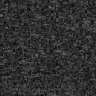} \;
			\includegraphics[width=0.11 \textwidth]{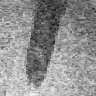}
			\includegraphics[width=0.11 \textwidth]{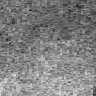}
		}
	}
	\vspace*{0.5cm}
	\caption[Small sample of sonar image patch pairs generated by our methodology]{Small sample of sonar image patch pairs generated by our methodology. One positive match class and two negative match classes are displayed.}
	\label{mat:datasetPatchSamples}
\end{figure*}

\FloatBarrier
\section{Matching with CNNs}

In this chapter we propose the use of Convolutional Neural Networks to match two image patches of the same size. Most of our work is inspired by the work of Zagoryuko and Komodakis \cite{zagoruyko2015learning}, as they introduced the use of CNNs for patch comparison. 

We framed the matching problem as both a classification and a regression problem. The idea is that a classification problem will fit a model to the true class labels, but additionally this model could produce probability scores (through a softmax activation function) that could aid in the matching decision process. A natural extension would be to regress these scores directly from the class labels. These scores then could be thresholded to produce binary matching decisions.

We evaluated two different CNN architectures. One is a siamese CNN with two branches, and the other is a simple feed-forward CNN that receives a two-channel input image.

The two-channel CNN was originally introduced by Zagoryuko and Komodakis. Our incarnation of that model is shown in Figure \ref{mat:twoChanCNN}. As matching uses two images as inputs, this network takes a two-channel input image. This is constructed by just concatenating the two $96 \times 96$ sonar image patches along the channels dimension into $2 \times 96 \times 96$ tensor.

All layers use the ReLU activation, except for the output layer. Dropout \cite{srivastava2014dropout} with $p = 0.5$ is used in the two first fully connected layers in order to reduce overfitting and improve generalization performance.

For matching as classification, we use $c = 2$ as it corresponds to a binary classification problem. A softmax activation function is used at the output fully connected layer. In this configuration the network has 126513 parameters (with a parameter to data points ratio of $\frac{126513}{33096} \sim 3.8$, reduced to $1.9$ with Dropout at training time). This model is trained with a categorical cross-entropy loss function.
For matching as regression, we use $c = 1$ and a sigmoid activation at the output fully connected layer. This network configuration has 125K parameters (with a parameter to data points ratio of $\frac{125000}{33096} \sim 3.7$, reduced to $1.8$ with Dropout at training time). This model is trained with a binary cross-entropy layer (Equation \ref{mat:binaryCELoss}) which is just the scalar output version of the categorical cross-entropy. We also tried other losses such as mean average error (L1 loss), mean squared error (L2 loss) and smooth L1 loss, but they were not any enhancement over the binary cross-entropy.

\begin{equation}
	L(y, \hat{y}) = - \sum_i y_i \log(\hat{y}_i) = -y_0 \log(\hat{y}_0) - (1 - y_0) \log(1 - \hat{y}_0)
	\label{mat:binaryCELoss}
\end{equation}

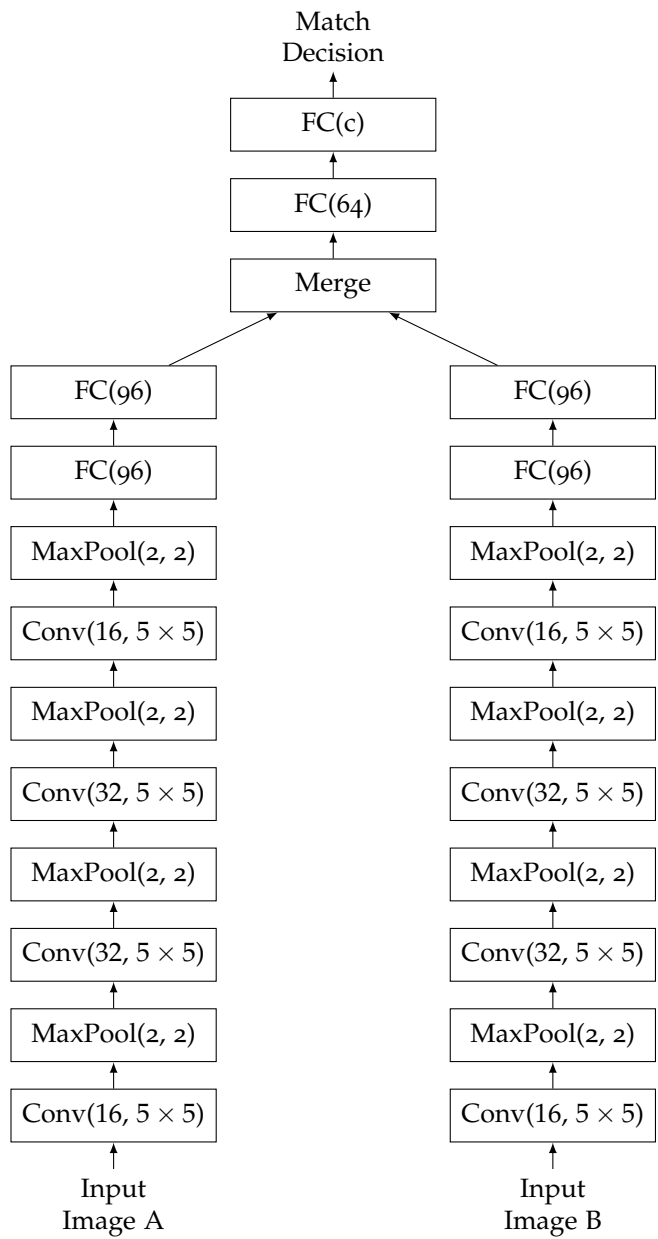
\begin{figure}[p]
	\vspace*{1.0cm}
	\centering
	\begin{tikzpicture}[style={align=center, minimum height=0.7cm, minimum width=2.7cm}]
	
	\node[] (dummy) {};
	
	\node[draw, above=2em of dummy] (merge) {Merge};
	\node[draw, above=1em of merge] (fc1) {FC(64)};
	\node[draw, above=1em of fc1] 	(fc2) {FC(c)};
	
	\node[above=1em of fc2] 	(output) {Match\\Decision};
	
	\node[draw, left=0.5em of dummy] (upB) {FC(96)};
	\node[draw, below=1em of upB] (upC) {FC(96)};
	
	\node[draw, below=1em of upC] (upD) {MaxPool(2, 2)};
	\node[draw, below=1em of upD] (upE) {Conv($16$, $5 \times 5$)};
	
	\node[draw, below=1em of upE] (upF) {MaxPool(2, 2)};
	\node[draw, below=1em of upF] (upG) {Conv($32$, $5 \times 5$)};
	
	\node[draw, below=1em of upG] (upH) {MaxPool(2, 2)};
	\node[draw, below=1em of upH] (upI) {Conv($32$, $5 \times 5$)};
	
	\node[draw, below=1em of upI] (upJ) {MaxPool(2, 2)};
	\node[draw, below=1em of upJ] (upK) {Conv($16$, $5 \times 5$)};	
	
	\node[draw, right=0.5em of dummy] (downB) {FC(96)};
	\node[draw, below=1em of downB] (downC) {FC(96)};
	
	\node[draw, below=1em of downC] (downD) {MaxPool(2, 2)};
	\node[draw, below=1em of downD] (downE) {Conv($16$, $5 \times 5$)};
	
	\node[draw, below=1em of downE] (downF) {MaxPool(2, 2)};
	\node[draw, below=1em of downF] (downG) {Conv($32$, $5 \times 5$)};
	
	\node[draw, below=1em of downG] (downH) {MaxPool(2, 2)};
	\node[draw, below=1em of downH] (downI) {Conv($32$, $5 \times 5$)};
	
	\node[draw, below=1em of downI] (downJ) {MaxPool(2, 2)};
	\node[draw, below=1em of downJ] (downK) {Conv($16$, $5 \times 5$)};	
	
	\node[below=1em of upK] (inputA) {Input\\Image A};
	\node[below=1em of downK] (inputB) {Input\\Image B};
	
	\draw[-latex] (upC) -- (upB);
	\draw[-latex] (upD) -- (upC);
	\draw[-latex] (upE) -- (upD);
	\draw[-latex] (upF) -- (upE);
	\draw[-latex] (upG) -- (upF);
	\draw[-latex] (upH) -- (upG);
	\draw[-latex] (upI) -- (upH);
	\draw[-latex] (upJ) -- (upI);
	\draw[-latex] (upK) -- (upJ);
	
	\draw[-latex] (downC) -- (downB);
	\draw[-latex] (downD) -- (downC);
	\draw[-latex] (downE) -- (downD);
	\draw[-latex] (downF) -- (downE);
	\draw[-latex] (downG) -- (downF);
	\draw[-latex] (downH) -- (downG);
	\draw[-latex] (downI) -- (downH);
	\draw[-latex] (downJ) -- (downI);
	\draw[-latex] (downK) -- (downJ);
	
	\draw[-latex] (inputA) -- (upK);
	\draw[-latex] (inputB) -- (downK);
	
	\draw[-latex] (upB) -- (merge);
	\draw[-latex] (downB) -- (merge);
	\draw[-latex] (merge) -- (fc1);
	\draw[-latex] (fc1) -- (fc2);
	
	\draw[-latex] (fc2) -- (output);
	\end{tikzpicture}
	\vspace*{0.5cm}
	\caption{Siamese CNN architecture for matching using a two one-channel input images}
	\label{mat:siameseCNN}
\end{figure}

The Siamese CNN architecture is shown in Figure \ref{mat:siameseCNN}. Unlike the previous network, this model takes two $96 \times 96$ input images separately. Each branch of the siamese network shares weights. The basic idea of this model is that each branch extracts relevant features from each image, and weight sharing allows for a reduction in the number of parameters, as well as making sure that invariant and equivalent features are extracted from each image.

Both networks are trained in the same way, using ADAM as optimizer with initial learning rate $\alpha = 0.01$ and batch size $B = 128$ elements. The two-channel CNN model is trained for $M = 5$ epochs, while the Siamese CNN model is trained for $M = 15$ epochs.

\begin{marginfigure}
    \centering
    \begin{tikzpicture}[style={align=center, minimum height=0.5cm, minimum width=2.5cm}]		
    \node[draw] (A) {FC(c)};	
    
    \node[above=1em of A] (output) {Match\\Decision};
    
    \node[above=1 em of output] (dummy) {};
    \node[draw, below=1em of A] (B) {FC(32)};
    \node[draw, below=1em of B] (C) {FC(64)};
    
    \node[draw, below=1em of C] (D) {MaxPool(2, 2)};f
    \node[draw, below=1em of D] (E) {Conv($16$, $5 \times 5$)};
    
    \node[draw, below=1em of E] (F) {MaxPool(2, 2)};
    \node[draw, below=1em of F] (G) {Conv($32$, $5 \times 5$)};
    
    \node[draw, below=1em of G] (H) {MaxPool(2, 2)};
    \node[draw, below=1em of H] (I) {Conv($32$, $5 \times 5$)};
    
    \node[draw, below=1em of I] (J) {MaxPool(2, 2)};
    \node[draw, below=1em of J] (K) {Conv($16$, $5 \times 5$)};	
    
    \node[below=1em of K] (Input) {Two-Channel\\Input Image};	
    
    \draw[-latex] (B) -- (A);
    \draw[-latex] (C) -- (B);
    \draw[-latex] (D) -- (C);
    \draw[-latex] (E) -- (D);
    \draw[-latex] (F) -- (E);
    \draw[-latex] (G) -- (F);
    \draw[-latex] (H) -- (G);
    \draw[-latex] (I) -- (H);
    \draw[-latex] (J) -- (I);
    \draw[-latex] (K) -- (J);
    \draw[-latex] (Input) -- (K);
    \draw[-latex] (A) -- (output);
    \end{tikzpicture}
    \caption{CNN architecture for matching using a two-channel input image}
    \label{mat:twoChanCNN}
\end{marginfigure}

\section{Experimental Evaluation}

\subsection{Evaluation Metrics}

Matching is a fundamentally different problem from multi-class image classification, as we have modeled matching as a binary classification problem. This defines a different set of candidate metrics: false and true positive rates, precision and recall, ROC curves, and accuracy \cite{bishop2006pattern}.

From the Marine Debris task point of view, it is desirable that the matching algorithm produces a quantifiable measure of similarity or confidence, so it can be interpreted by a human. Metrics that prefer binary results like \textit{match} or \textit{no match} without an explanation by similarity or confidence score are not as useful as ones that evaluate confidence in the decision that was made.

For this reason we chose the area under the ROC curve (AUC) as primary evaluation metric. This measures the confidence score quality produced by the classifier \cite{murphy2012machine}. Classifiers that receive higher AUC values produce confidence scores that are better separate classes, so a simple threshold on the score can be used to make class decisions. The ROC curve, and consequently the AUC, incorporate the precision and recall metrics in its calculation.

\subsection{Model Comparison}

In this section we evaluate our CNNs for matching with state of the art keypoint matching algorithms, and we also evaluate some baselines based on classic machine learning algorithms trained on our datasets. For keypoint matching we evaluate SIFT, SURF, ORB and AKAZE.

These keypoint detectors are evaluated through a common protocol. Keypoints are detected in each input image and matched using a k-nearest neighbour search with $k = 2$ over the computed descriptors in each keypoint. For continuous features (AKAZE, SIFT, and SURF) we compute the number of "good" matches using the ratio test \cite{lowe2004distinctive}. This test requires at least three matches, and it filters descriptor matches according to the distance ratio between one match and the second-best match. If this ratio is too large, the match is discarded. Then we use the number of "good" matches produced with a ratio threshold of $0.75$ and set a minimum threshold to declare a match between two images.

For binary features (ORB) we only use the number of matches and put a minimum threshold to declare a match between two images. This threshold is used to construct the ROC curve later on.

As machine learning baselines we evaluate:

\begin{description}
	\item[\textbf{Random Forest}] A random forest \cite{murphy2012machine} seems to be a good choice as it is both resistant to overfitting due to the use of an ensemble of decision trees, and it is a learning algorithm with a non-linear decision boundary. We trained a random forest classifier and a random forest regressor. Both share the same hyper-parameters. We used a forest with 30 trees and a maximum depth of 40 levels. Each classification decision tree is trained by minimizing the gini impurity. For regression decision trees, we used the mean squared error. The random forest classifier can provide class probabilities for positive and negative matches as the voting ratio of each leaf.
	
	\item[\textbf{Support Vector Machine}] A support vector machine represents a lower bound on performance, as this learning algorithm is reproducible due to the use of an optimal separating hyperplane. We tried both a support vector machine for classification, and a support vector regressor for regression of the target scores. The SVM is trained with regularization coefficient $C = 1$, while the SVR regressor is trained with $C = 10$ and $\epsilon = 0.05$. The trained SVM includes probability information through the use of internal cross-validation.
\end{description}

Both kinds of ML classifiers were tuned using grid search over a predefined grid of parameters. Keypoint method have no hyper-parameters to be tuned. We now describe our evaluation metrics.

As we model matching as a binary classification problem, we use the Area under the ROC Curve (AUC) as our primary metric. The ROC curve is obtained by computing the true positive (TPR) and false positive rates (FPR):

\begin{equation}
	\text{TPR} = \frac{TP}{P} \qquad \text{FPR} = \frac{FP}{N}
\end{equation}

Assuming a classifier that provides a score for each class, then the TPR and FPR rates vary as a threshold is set on the output classification score. Then the ROC curve is built as points in the $(\text{FPR}, \text{TPR})$ space as the threshold is varied. This curve indicates the different operating points that a given classifier outputs can produce, and it is useful in order to tune a specific threshold while using the classifier in production.

The AUC is the just the area under the ROC curve, which is a number in the $[0, 1]$ range. The AUC is a metric that is not simple to interpret \cite{sammut2011encyclopedia}. One interpretation is that the AUC is the probability that the classifier will produce a higher score for a randomly chosen positive example than a randomly chosen negative example. A classifier with a higher AUC is then preferable.

We also evaluated accuracy, but as we have labels for the three components that were used to generate each dataset, we also evaluated accuracy on each component. This includes examples that represent a object-object positive match, a object-object negative match, and a object-background negative match. We also compute and present mean accuracy. As we are also evaluating regressors to predict a score, we compute accuracy from class predictions:

\begin{equation}
	c = \argmax\{1 - p, p\}
\end{equation}

Where $p$ is the output prediction from the regressor or probability from a classifier. For keypoint detectors we compute accuracies using a threshold equal to zero for the minimum number of matches. This value produces the maximum possible accuracy for these methods.

Numerical results are presented in Table \ref{mat:matchingResultsTable}, while ROC curve plots are shown in Figures \ref{mat:rocSamePlot} and \ref{mat:rocDiffPlot}.

We only evaluated keypoint detection methods on one of the datasets. Their performance is virtually the same and it is not affected by objects in the training set, as these methods do not use learning. Keypoint matching algorithms perform quite poorly, with AUC that is slightly larger than random chance ($0.5$). Their mean accuracy is also quite close to random chance, which is product of very low accuracies in the object-object negative match case. These results show that keypoint methods work at an acceptable performance when matching the same object under a different view (a positive match) but fail to declare a mismatch for different objects (negative matches). Seems that keypoint detection is overconfident and produces too many positive matches, which translates to lower accuracies in the two negative cases.

\begin{table*}[t]
	\forcerectofloat
	\centering
	
	\begin{tabular}{lllllll}
		\hline 
		& Method 	& AUC	 	& Mean Acc & Obj-Obj $+$ Acc 	& Obj-Obj $-$ Acc 	& Obj-Bg $-$ Acc\\ 
		\hline 
		& SIFT	& $0.610$ 	& $54.0 $ \% & $74.5 $ \% 		&  $43.6 $ \%			&  $44.0 $ \% \\
		& SURF	& $0.679$	& $48.1 $ \% & $89.9 $ \% 		&  $18.6 $ \%			&  $35.9 $ \% \\
		& ORB	& $0.682$	& $54.9 $ \% & $72.3 $ \% 		&  $41.9 $ \%			&  $60.5 $ \% \\
		& AKAZE	& $0.634$	& $52.2 $ \% & $95.1 $ \% 		&  $4.8 $ \%				&  $56.8 $ \% \\
		\hline
		
		\multirow{6}{*}{\rotatebox{90}{Different}}& RF-Score	& $0.741$	& $57.6 $ \% & $22.5 $ \% 		&  $88.2 $ \%				&  $97.2 $ \% \\
		& RF-Class	& $0.795$	& $69.9 $ \% & $12.5 $ \% 		&  $\textbf{97.7} $ \%				&  $\textbf{99.7} $ \% \\
		\cline{2-7}
		& SVR-Score	& $0.663$	& $70.5 $ \% & $57.2 $ \% 		&  $66.6 $ \%				&  $87.5 $ \% \\
		& SVM-Class	& $0.652$	& $67.1 $ \% & $54.4 $ \% 		&  $69.1 $ \%				&  $90.5 $ \% \\
		\cline{2-7}		
		& 2-Chan CNN Scr & $0.894$	& $82.9 $ \% & $\textbf{68.0} $ \% &  $96.1 $ \%				&  $84.5 $ \% \\
		& 2-Chan CNN Cls & $\textbf{0.910}$	& $86.2 $ \% & $67.3 $ \% &  $95.2 $ \%				&  $96.1 $ \% \\
		\cline{2-7}
		& Siam CNN Scr & $0.826$	& $77.0 $ \% & $49.2 $ \% &  $84.7 $ \%				&  $97.0 $ \% \\
		& Siam CNN Cls & $0.855$	& $82.9 $ \% & $62.9 $ \% &  $89.9 $ \%				&  $96.0 $ \% \\
		\hline
		
		\multirow{6}{*}{\rotatebox{90}{Same}}& RF-Score	& $0.972$	& $85.2 $ \% & $\textbf{98.7} $ \% 		&  $58.8 $ \%				&  $98.1 $ \% \\
		& RF-Class	& $\textbf{0.982}$	& $90.9 $ \% & $97.3 $ \% 		&  $75.8 $ \%				&  $\textbf{99.6} $ \% \\
		\cline{2-7}
		& SVR-Score	& $0.767$	& $66.2 $ \% & $86.3 $ \% 		&  $17.6 $ \%				&  $94.7 $ \% \\
		& SVM-Class	& $0.742$	& $64.8 $ \% & $83.7 $ \% 		&  $18.3 $ \%				&  $92.4 $ \% \\
		\cline{2-7}		
		& 2-Chan CNN Scr & $0.934$	& $85.4 $ \% & $85.0 $ \% &  $\textbf{77.5} $ \%				&  $93.7 $ \% \\
		& 2-Chan CNN Cls & $0.944$	& $86.7 $ \% & $86.6 $ \% &  $75.7 $ \%				&  $97.8 $ \% \\
		\cline{2-7}		
		& Siam CNN Scr & $0.895$	& $80.6 $ \% & $89.1 $ \% &  $55.3 $ \%				&  $97.3 $ \% \\
		& Siam CNN Cls & $0.864$	& $75.8 $ \% & $92.2 $ \% &  $39.4 $ \%				&  $95.8 $ \% \\
		\hline
	\end{tabular}
	\vspace*{0.5cm}
	\caption[Comparison of different models for matching]{Comparison of different models for matching. Area Under the ROC Curve (AUC), Accuracy at match threshold zero, and Accuracy for each match type is reported for both datasets (Same and Different).}
	\label{mat:matchingResultsTable}
\end{table*}

Machine learning methods perform considerably better. Considering different objects (Dataset D), a Random Forest performs adequately, obtaining AUC in the range $[0.74, 0.79]$, and this method produces very good accuracy for the negative cases, but poor accuracy for the positive match case. Seems a random forest has the opposite behaviour than the keypoint matching algorithms, as it is overconfident on negative matches but performs poorly for the positive ones.

An SVM and SVR both perform quite poorly, with similar AUC close to $0.65$, but a more balanced performance across match cases, but still biased performance towards object-background negative cases.

Our proposed method using a CNN outperforms all other methods in the different objects case. This means that a CNN can learn appropriate features that generalize well outside of the training set, even generalizing to different objects. The best performing network is a two-channel one producing class probabilities, closely followed by also a two-channel network that produces a single regression score that is used for matching. There is a $2$ \% difference in the AUC of these networks.

Siamese networks also perform quite well but below the performance of the two-channel networks. All CNN methods have a common performance pitfall, as negative cases have very good accuracy, but the positive case has lower accuracy, which is approximately $20$ \% over the random chance minimum.

Considering the same objects in the training and testing sets (Dataset S), results change radically. All method perform better than in Dataset D. For this case, the best performing method is a random forest classifier, with $0.982$ AUC. A regression random forest obtains $1$ \% less AUC. Comparing RF performance with that of Dataset D shows that the RF classifier suffers from mild overfitting. It is not memorizing the training set but the learned classifier favours objects in the training set, performing very well on them, but decreasing performance considerably with unseen objects.

Other methods also perform considerably better. An SVM/SVR increases AUC by approximately $10$ \%, but a two-channel CNN only improves AUC by $3$ \%. This suggests that CNN methods are not overfitting and the loss in performance is acceptable for unseen objects.

It seems that as a general pattern, matching as a classification problem produces better results than using a regression problem. Only some exceptions to this pattern occur, such as a Siamese network in Dataset S, and SVR on both datasets.

Our results show that using a CNN is a very good option for the problem of matching pairs of sonar image patches.

\pgfplotsset{cycle list/Set1-6}

\pgfplotscreateplotcyclelist{matiasList}{%
	{color=red,solid},
	{color=blue,solid},
	{color=green,solid},
	{color=black,solid},
	{color=red,densely dashed},
	{color=blue,densely dashed},
	{color=green,densely dashed},
	{color=black,densely dashed},
	{color=red,densely dotted},
	{color=blue,densely dotted},
	{color=green,densely dotted},
	{color=black,densely dotted}}

\begin{figure}[h]
	\begin{tikzpicture}
	\begin{axis}[height = 0.35\textheight, width = 0.7\textwidth, xlabel={False Positive Rate}, ylabel={True Positive Rate}, xmin=0, xmax=1.0, ymin=0, ymax=1.0, ymajorgrids=true, xmajorgrids=true, grid style=dashed, legend pos = outer north east, legend style={font=\scriptsize, at = {(1.1, 1.0)}}, cycle multiindex* list={matiasList}, legend columns = 1]
	\addplot +[no markers, raw gnuplot] gnuplot {
		plot 'chapters/data/matching/different/MatchingCNNDropoutClassNoAugment-DiffObjects-ROCCurve.csv' using 2:3 smooth sbezier;
	};
	\addlegendentry{CNN 2-Chan Class}
	
	\addplot +[no markers, raw gnuplot] gnuplot {
		plot 'chapters/data/matching/different/MatchingCNNDropoutScoreBCENoAugment-DiffObjects-ROCCurve.csv' using 2:3 smooth sbezier;
	};
	\addlegendentry{CNN 2-Chan Score}
	
	\addplot +[no markers, raw gnuplot] gnuplot {
		plot 'chapters/data/matching/different/svmProbsROCCurve.csv' using 2:3 smooth sbezier;
	};
	\addlegendentry{SVM-Class}
	
	\addplot +[no markers, raw gnuplot] gnuplot {
		plot 'chapters/data/matching/different/svrROCCurve.csv' using 2:3 smooth sbezier;
	};
	\addlegendentry{SVR-Score}
	
	\addplot +[no markers, raw gnuplot] gnuplot {
		plot 'chapters/data/matching/different/randomForestClassifierROCCurve.csv' using 2:3 smooth sbezier;
	};
	\addlegendentry{RF-Class}
	
	\addplot +[no markers, raw gnuplot] gnuplot {
		plot 'chapters/data/matching/different/randomForestROCCurve.csv' using 2:3 smooth sbezier;
	};
	\addlegendentry{RF-Score}
	
	\addplot+[mark = none] table[x  = fpr, y  = tpr, col sep = space] {chapters/data/matching/siftMatcherPRCurve.csv};
	\addlegendentry{SIFT}
	\addplot+[mark = none] table[x  = fpr, y  = tpr, col sep = space] {chapters/data/matching/surfMatcherPRCurve.csv};
	\addlegendentry{SURF}
	\addplot+[mark = none] table[x  = fpr, y  = tpr, col sep = space] {chapters/data/matching/orbMatcherPRCurve.csv};
	\addlegendentry{ORB}	
	\addplot+[mark = none] table[x  = fpr, y  = tpr, col sep = space] {chapters/data/matching/akazeMatcherPRCurve.csv};
	\addlegendentry{AKAZE}
	\draw[gray] (axis cs:0,0) -- (axis cs:1,1);
	\end{axis}        
	\end{tikzpicture}
	\caption[ROC curve comparing different methods for sonar image patch matching on Dataset D]{ROC curve comparing different methods for sonar image patch matching on Dataset D, meaning different objects were used to produce the training and testing sets. The grey line represents random chance limit.}
	\label{mat:rocDiffPlot}
\end{figure}

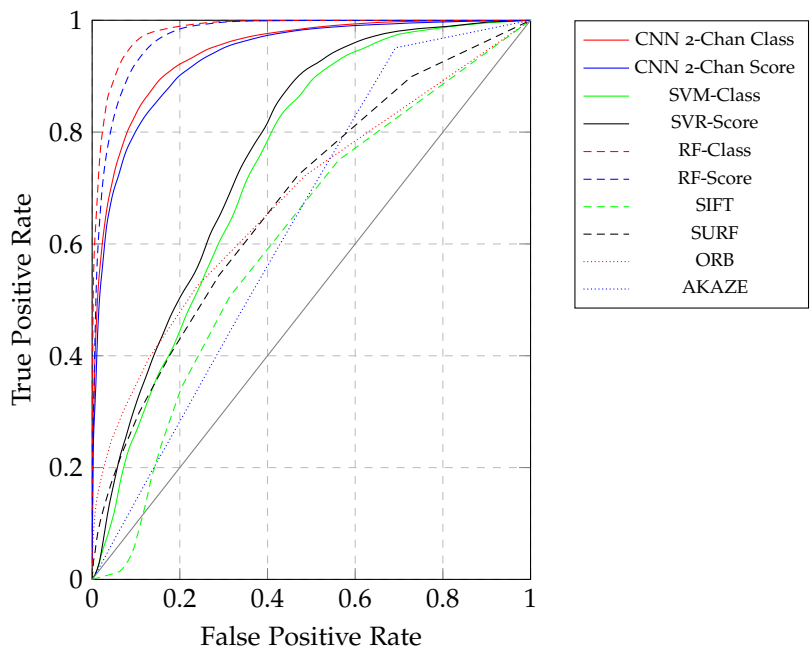
\begin{figure}[h]
	\begin{tikzpicture}
	\begin{axis}[height = 0.35\textheight, width = 0.7\textwidth, xlabel={False Positive Rate}, ylabel={True Positive Rate}, xmin=0, xmax=1.0, ymin=0, ymax=1.0, ymajorgrids=true, xmajorgrids=true, grid style=dashed, legend pos = outer north east, legend style={font=\scriptsize, at = {(1.1, 1.0)}}, cycle multiindex* list={matiasList}, legend columns = 1]
	\addplot +[no markers, raw gnuplot] gnuplot {
		plot 'chapters/data/matching/same/MatchingCNNDropoutClassNoAugment-SameObjects-ROCCurve.csv' using 2:3 smooth sbezier;
	};
	\addlegendentry{CNN 2-Chan Class}
	
	\addplot +[no markers, raw gnuplot] gnuplot {
		plot 'chapters/data/matching/same/MatchingCNNDropoutScoreBCENoAugment-SameObjects-ROCCurve.csv' using 2:3 smooth sbezier;
	};
	\addlegendentry{CNN 2-Chan Score}
	
	\addplot +[no markers, raw gnuplot] gnuplot {
		plot 'chapters/data/matching/same/svmProbsROCCurve.csv' using 2:3 smooth sbezier;
	};
	\addlegendentry{SVM-Class}
	
	\addplot +[no markers, raw gnuplot] gnuplot {
		plot 'chapters/data/matching/same/svrROCCurve.csv' using 2:3 smooth sbezier;
	};
	\addlegendentry{SVR-Score}
	
	\addplot +[no markers, raw gnuplot] gnuplot {
		plot 'chapters/data/matching/same/randomForestClassifierROCCurve.csv' using 2:3 smooth sbezier;
	};
	\addlegendentry{RF-Class}
	
	\addplot +[no markers, raw gnuplot] gnuplot {
		plot 'chapters/data/matching/same/randomForestROCCurve.csv' using 2:3 smooth sbezier;
	};
	\addlegendentry{RF-Score}
	
	\addplot+[mark = none] table[x  = fpr, y  = tpr, col sep = space] {chapters/data/matching/siftMatcherPRCurve.csv};
	\addlegendentry{SIFT}
	\addplot+[mark = none] table[x  = fpr, y  = tpr, col sep = space] {chapters/data/matching/surfMatcherPRCurve.csv};
	\addlegendentry{SURF}
	\addplot+[mark = none] table[x  = fpr, y  = tpr, col sep = space] {chapters/data/matching/orbMatcherPRCurve.csv};
	\addlegendentry{ORB}	
	\addplot+[mark = none] table[x  = fpr, y  = tpr, col sep = space] {chapters/data/matching/akazeMatcherPRCurve.csv};
	\addlegendentry{AKAZE}
	\draw[gray] (axis cs:0,0) -- (axis cs:1,1);
	\end{axis}        
	\end{tikzpicture}
	\caption[ROC curve comparing different methods for sonar image patch matching on Dataset S]{ROC curve comparing different methods for sonar image patch matching on Dataset S, meaning the same objects were used to produce the training and testing sets. The grey line represents random chance limit.}
	\label{mat:rocSamePlot}
\end{figure}

\FloatBarrier
\subsection{How Much Data is Needed?}

In this section we explore the question of how generalization varies with training set size. We use the same basic methodology as previously mentioned in Section \ref{lim:secNumTrainingSamples}. We only evaluated the two-channel networks, as they performed the best.

We vary the number of samples per class (SPC) from 1 to 5000. The original training set contains approximately 40000 samples which corresponds to a SPC of 20000. Due to computational constraints we only evaluate up to SPC 5000. As like our previous evaluations of matching algorithms, we evaluate the area under the curve (AUC) for each data point.

\begin{figure*}[!ht]
	\forcerectofloat
	\centering
	\begin{tikzpicture}
	\begin{customlegend}[legend columns = 4,legend style = {column sep=1ex}, legend cell align = left,
	legend entries={CNN 2-Chan Class, CNN 2-Chan Score}]
	\addlegendimage{mark=none,blue}
	\addlegendimage{mark=none,red}
	\end{customlegend}
	\end{tikzpicture}
	
	\subfloat[Sub-Region $1-500$]{
		\begin{tikzpicture}
		\begin{axis}[
		xlabel={Samples per Class},
		ylabel={Test AUC (\%)},        
		xmax = 500,
		ymin=35, ymax=100,
		xtick={1,25,50,100,150,200,300,400,500},
		ytick={30,40,50,60,70,80,90,95,100},
		x tick label style={font=\tiny, rotate=90},
		legend pos=south east,
		ymajorgrids=true,
		grid style=dashed,
		height = 0.25\textheight,
		width = 0.45\textwidth]
		
		\errorband{chapters/data/matching/different/matching-classifierDropout-AUCVsTrainSetSize.csv}{samplesPerClass}{meanAUC}{stdAUC}{blue}{0.4}
		
		\end{axis}
		\end{tikzpicture}
	}	
	\subfloat[Full Plot]{
		\begin{tikzpicture}
		\begin{axis}[
		xlabel={Samples per Class},
		ylabel={Test AUC (\%)},        
		xmax = 5000,
		ymin=35, ymax=100,
		xtick={100,250,500,750,1000, 2000, 3000, 4000, 5000},
		ytick={30,40,50,60,70,80,90,95,100},
		x tick label style={font=\tiny, rotate=90},
		legend pos=south east,
		ymajorgrids=true,
		grid style=dashed,
		height = 0.25\textheight,
		width = 0.45\textwidth]
		
		\errorband{chapters/data/matching/different/matching-classifierDropout-AUCVsTrainSetSize.csv}{samplesPerClass}{meanAUC}{stdAUC}{blue}{0.4}
		
		\end{axis}
		\end{tikzpicture}
	}

	\subfloat[Sub-Region $1-500$]{
		\begin{tikzpicture}
		\begin{axis}[
		xlabel={Samples per Class},
		ylabel={Test AUC (\%)},        
		xmax = 500,
		ymin=35, ymax=100,
		xtick={1,25,50,100,150,200,300,400,500},
		ytick={30,40,50,60,70,80,90,95,100},
		x tick label style={font=\tiny, rotate=90},
		legend pos=south east,
		ymajorgrids=true,
		grid style=dashed,
		height = 0.25\textheight,
		width = 0.45\textwidth]
		
		\errorband{chapters/data/matching/different/matching-scorerDropout-AUCVsTrainSetSize.csv}{samplesPerClass}{meanAUC}{stdAUC}{red}{0.4}
		
		\end{axis}
		\end{tikzpicture}
	}
	\subfloat[Full Plot]{
		\begin{tikzpicture}
		\begin{axis}[
		xlabel={Samples per Class},
		ylabel={Test AUC (\%)},        
		xmax = 5000,
		ymin=35, ymax=100,
		xtick={100,250,500,750,1000, 2000, 3000, 4000, 5000},
		ytick={30,40,50,60,70,80,90,95,100},
		x tick label style={font=\tiny, rotate=90},
		legend pos=south east,
		ymajorgrids=true,
		grid style=dashed,
		height = 0.25\textheight,
		width = 0.45\textwidth]
		
		\errorband{chapters/data/matching/different/matching-scorerDropout-AUCVsTrainSetSize.csv}{samplesPerClass}{meanAUC}{stdAUC}{red}{0.4}
		
		\end{axis}
		\end{tikzpicture}
	}
	\vspace*{0.5cm}
	\caption[Samples per Class versus Accuracy for ClassicNet with 2 modules]{Samples per Class versus Accuracy for ClassicNet with 2 modules, including error regions.}
	\label{mat:spcVsAUC}
\end{figure*}
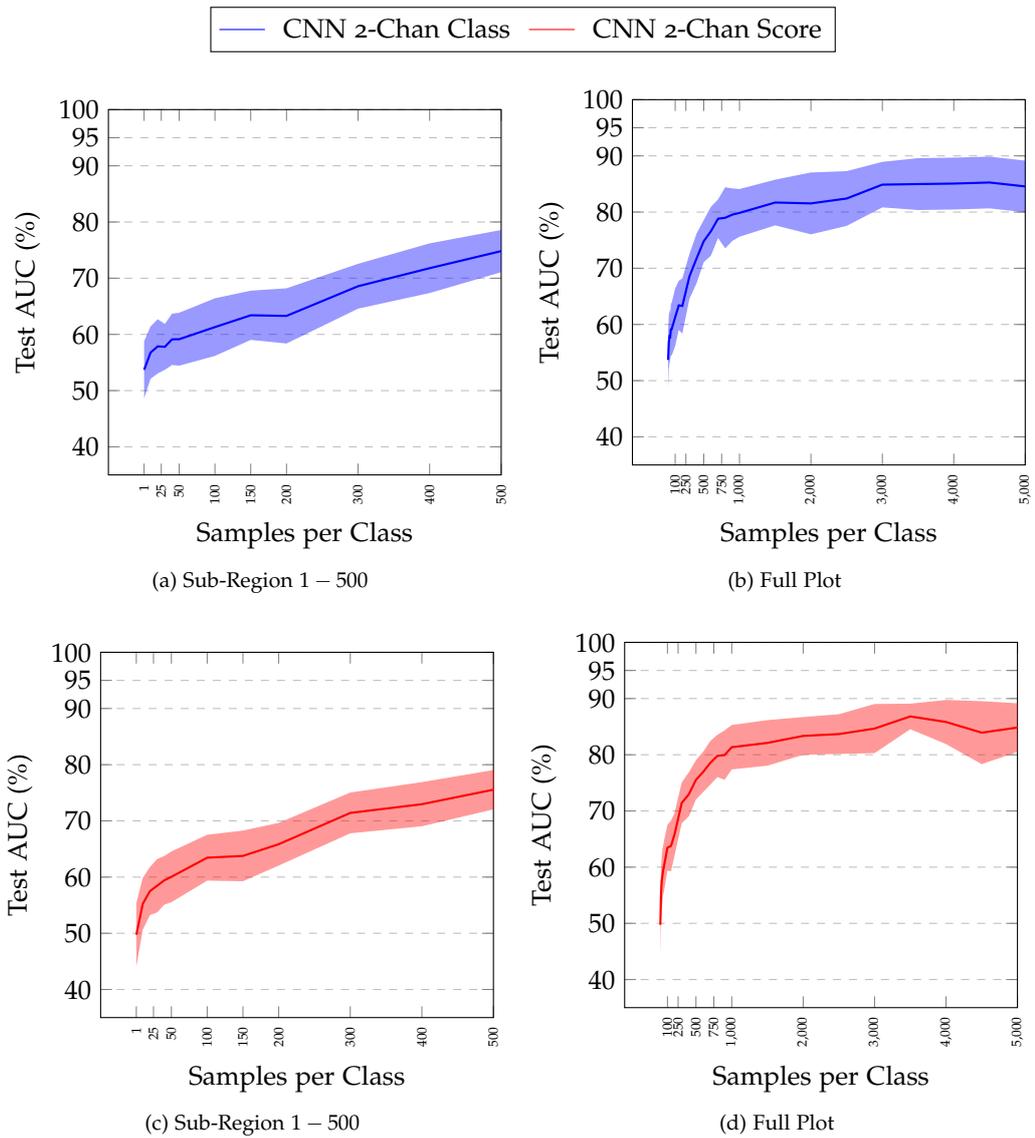

Results are presented in Figure \ref{mat:spcVsAUC}. Both scoring and classification networks perform quite similarly as the training set size is varied. But the classification network gets slightly better AUC performance with low sample quantity. This can be seen as the classifier obtaining performance that is slightly better than random chance ($50$ \%) at one sample per class, while the scorer obtains worse than random chance performance at the same SPC value.

As SPC increases, performance increases rapidly for SPC less than 750, and after that the increases are smaller (diminishing returns). Both models converge after SPC 5000 close to $90$ \% AUC. This shows that the full dataset is required to produce good generalization performance, and these networks do not perform well if only a smaller dataset is available.

These results are consistent with the difficulty of matching sonar image patches. It seems the only way to improve these models is to increase the size of the training set, as with many other ML algorithms. But we also believe that other approaches could work better, specially the ones from fields like one-shot learning. One natural choice would be a network with a contrastive or triplet loss \cite{Schroff_2015_CVPR}. This kind of networks is trained in a different way, where an embedding is learned in such a way to maximize embedded distances between negative samples, and minimize the distances between positive examples. This makes a much more discriminative model.

\FloatBarrier

\newpage
\section{Summary of Results}

In this chapter we have propose the use of Convolutional Neural Networks for matching sonar image patches. This problem has been open for a long time, mostly due to the specific details of a sonar sensor: viewpoint dependence, noise, non-uniform insonification, and hard to model objects.

We transformed one of our classification datasets into a matching one, by generating positive and negative image pairs, corresponding to object-object positive pair (same object class), object-object negative (different object class), and object-background negative pairs. This generated over 39K $96 \times 96$ image pairs for training, with 7.5K pairs for testing. We made sure that objects used to generate the training set were different from the testing set, meaning that our performance figures represent true generalization outside the training set.

We evaluated classic keypoint matching algorithms, namely SIFT, SURF, ORB, and AKAZE. We show that these techniques do not perform well in sonar images, with area under the ROC curve in the range 0.61-0.63, which is slightly better than a random classifier.

We also evaluated the use of classic ML classifiers for this problem, including a Random Forest and a Support Vector Machine. In this case we model matching as binary classification given two $96 \times 96$ image patches. These methods work better than keypoint detectors at AUC 0.65-0.80. Based on previous work by Zagoryuko et al. \cite{zagoruyko2015learning}, we decided to implement and compare a two-channel and a siamese network.

The two-channel network obtains the best matching performance at 0.91 AUC, performing binary classification. It is closely followed by a regression two-channel network with 0.89 AUC. In comparison, siamese networks perform poorly at AUC 0.82-0.86.

Our results show that a CNN outperforms other methods and sets a new state of the art results for matching sonar image patches. Other machine learning techniques perform well, but are not competitive versus a CNN, with just one exception. We also evaluated using the same objects for training and testing, and in this specific case a Random Forest can outperform a CNN.

Still there is considerable research to be done. Our method was trained only with 40K training samples, and more data would improve classification performance. In comparison, there are datasets with half million labeled color image patch pairs, where CNNs obtain much higher accuracy. Ideally the implicit distance information in a sonar image could be used to automatically label patches belonging to multiple views of an object, and a much bigger dataset could be constructed.

The data that we used was captured in the OSL water tank, which does not have a complex background, which could bias our results. We expect that with more realistic data the results we presented would degrade, but still with increased amounts of training data this can be easily offset. More variation in objects, including natural degradation such as bio-fouling, and more object classes, would also be needed for good generalization.

We expect that our techniques will be adopted by the underwater robotics community and used to build other systems, such as improved SLAM, trackers and automatic target recognition.

%% file: chapters/detection-proposals-sonar.tex
\chapter[Detection Proposals In Forward-Looking Sonar]{Detection Proposals In \newline Forward-Looking Sonar}
\label{chapter:proposals}

This chapter deals with the core problem of this thesis, namely the detection of marine debris in sonar images. But here we focus on a slightly different problem.

Most literature work on object detection deals with the problem of designing and testing a class-specific object detector, which just detects objects of a certain class or classes. But our object of interest, marine debris, has a very large intra-class and inter-class variability. This motivates to construct a generic or class-agnostic object detector, which in the computer vision literature is called detection proposals.

For example, in order for a robot to detect novel objects \cite{endres2010category}, a class-independent object detector must be available. A novel object could be placed in front of the robot's sensors, and the robot would be able to say that there is an object in front of him, but it does not match any class that was previously trained, and it could ask the operator about information in order to label the new object.

Detection proposals are connected to the concept of objectness, which is a basic measurement of how likely an image window or patch contains an object of interest. This is also related to the concept of an object itself, which is hard to define.

While there are many computer vision algorithms that produce detection proposals, the concept of an class-agnostic object detector has not been applied to sonar images. Detection proposals in general are used in order to construct class-specific object detectors and improve their performance, but in our case we would like to design and build a class-agnostic detector as a purpose in itself, as we want an AUV to have have the capabilities to detect novel and new objects that were not considered during training time.

For our specific objective of detecting marine debris, as we cannot possibly model or collect training data for all kinds of marine debris, we wish to construct a detection proposal algorithm for sonar that can tell the robot of novel objects that could then be reported back to the operator. Then he or she would decide if the object is debris and should be collected by the robot, or ignored. Then the robot could perform a "human in the loop" cycle to train itself to find new kinds of marine debris that are present in the environment.

This objective can be summarized in the following research questions:

\begin{itemize}
	\item How can objects be detected on sonar images while making the minimum number of assumptions?
	\item How can a class-agnostic sonar detection proposal algorithm be trained? 
	\item Can we make sure that it generalizes well outside its training set?
	\item Can real-time performance and low number of proposals with high recall be achieved within these constraints?
	\item How much training data is actually required to generalize?
\end{itemize}

We propose the use of a patch-based convolutional neural network that predicts an objectness score, which can be either thresholded or ranked in order to produce detection proposals.

\section{Related Work}

There is a rich literature in this field, but most of it comes from computer vision applied to color images. We first discuss the basic evaluation metrics and then we describe the literature.

Object detection proposals are evaluated by computing the recall metric $R$:
\vspace*{1em}
\begin{equation}
	R = \frac{TP}{TP + FN}
\end{equation}

Where $TP$ are the number of true positives (correctly detected objects) and $FN$ is the number of false negatives (missed detections). Recall is used because it only considers how many of the true objects in the test set can be recovered by the proposals algorithm, but any extra object that is not labeled in the test set does not affect recall, but it affects precision, which is only used when evaluating class-specific object detectors. It must be noted that there are reports \cite[-4em]{chavali2016object} that using recall as part of an evaluation protocol can be "gamed" and a higher recall does not mean generalization outside of the training set.

Given two bounding boxes, their match is considered a correct detection if and only if their intersection-over-union score (also called IoU) is larger than a given overlap threshold $O_t$:

\begin{equation}
	\text{IoU}(A, B) = \frac{\text{area}(A \cap B)}{\text{area}(A \cup B)}
\end{equation}

The most common value \cite{everingham2010pascal} is $O_t = 0.5$, but higher values are possible. The IoU score measures how well two bounding boxes match, and it is used because ground truth bounding boxes are human-generated and could be considered arbitrary for a computer algorithm. Using IoU introduces a degree of slack into the proposals that can be generated by the algorithm and still considered correct.

The concept of object proposals were originally introduced by Endres and Hoiem \cite{endres2010category}. The authors were motivated by the human ability to localize objects without needing to recognize them. Their work basically is a category-independent object detector that is able to detect unknown objects, not previously seem them during training or considered by the underlying features. Their long term application is a vision system that can automatically discover new objects.

This seminal work uses super-pixel segmentation of the input image through hierarchical segmentation based on boundary features. Segmented regions are then merged to form the hierarchy through the use of color, texture, histogram, boundary strength, and layout agreement features.

Proposed regions are then ranked. The basic idea of ranking is that highly plausible objects should get a large score, while non-objects will receive a low score. This is motivated by the fact that hierarchical segmentation produces many regions that do not correspond to real objects. A ranking function is learned through structured learning, using appearance features and two penalty terms, one corresponding to penalization if the region overlaps one previously ranked proposal, and another penalizing if the region overlaps with multiple highly ranked regions. Features used for ranking consist of boundary characteristics transformed into probabilities, like probability of exterior/interior boundary, occlusion, and background.

This method is evaluated on two datasets: The Berkeley Segmentation Dataset and PASCAL VOC 2008 \cite[-2em]{everingham2010pascal}. As this method outputs both bounding boxes and segmented regions, best segmentation overlap score and recall at $O_t = 0.5$ are evaluated. $80-84$ \% recall is obtained in these datasets, with best segmentation overlap score in the $0.68-0.69$ range. This method works acceptably, but it performs heavy feature engineering, which indicates that the features do generalize but not in order to obtain recall closer to $99$ \%.

Rahtu et al. \cite{rahtu2011learning} use a cascade architecture to learn a category-independent object detector. Their motivation is that a cascade is considerably faster than the usual object detection architectures (like for example, Viola-Jones \cite[1em]{viola2001rapid} for real-time face detection). The authors introduce features that are useful to predict the likelihood that a bounding box contains an object (objectness).

The first step is to generate an initial set of bounding boxes from the input image, where two methods are applied. One is super-pixel segmentation, and the second is to sample 100K bounding boxes from a prior distribution computed from the training set. This prior distribution is parameterized by bounding box width, height, and row/column position in the image.

Each window is the evaluated for objectness and a binary decision is made by a classifier. Three features are used: Super-pixel boundary integral, boundary edge distribution and window symmetry. Non-maxima suppression is applied before outputting the final bounding boxes. Results on the PASCAL VOC 2007 dataset show that $95$ \$ recall can be obtained at IoU threshold of $O_t = 0.5$. This approach works well in terms of recall and is promising about computational performance, but as it has been previously mentioned, their choice of features does not transfer to sonar, specially super-pixel ones due to the noise and lack of clear boundary in sonar images.

Alexe et al. \cite{alexe2012measuring} present an objectness measure, putting a number on how likely is an image window to contain an object of interest, but not belonging to any specific class. The authors define an object with three basic characteristics: a defined closed boundary, a different appearance from its surroundings, and being unique and salient in the image.

The authors use multiple objectness cues that are combined using a Bayesian framework. The cues consist of multi-scale saliency, color contrast, edge density, super-pixel straddling, and location and size. All these cues contain parameters that must be learned from training data.

This method was also evaluated on the PASCAL VOC 2007 dataset, obtaining $91$ \% recall with 1000 bounding boxes per image, taking approximately 4 seconds per image. This objectness measure seems to generalize quite well in unseen objects, but it is quite complex to compute, and does not seem applicable to sonar. This technique relies on other techniques to compute cues, which makes it harder to transfer to sonar images.

Uijlings et al. \cite{uijlings2013selective} introduced selective search as a way to generate bounding boxes on an image that correspond to objects in it. The authors argue that the use of exhaustive search is wasteful due to uniform sampling of the search space. This strategy produces many candidate bounding boxes in the image that do not correspond to plausible objects. Selective Search instead basically guides the sampling process so information in the image is used to generate bounding boxes for plausible objects. This easily builds a class-agnostic object detector, reduces the number of bounding boxes that must be explored, and allows discovery of objects at multiple scales.

Selective search works by performing hierarchical grouping. This is motivated by the fact that the hierarchy produced by grouping allows for natural search of objects at multiple scales. In order to introduce robustness and be able to capture all objects, a diversification strategy is used. This corresponds to the use of different color spaces (RGB, Intensity, Lab, rg + I, HSV, rgb, Hue from HSV), complementary similarity measures for region grouping (color, texture, size, and fill), and by varying the starting regions.

The authors evaluate selective search in the PASCAL VOC 2007 dataset. Average Best Overlap (ABO) score is used to measure how well the produced bounding boxes $B$ fit the VOC ground truth $G$.

\begin{equation}
	\text{ABO}(G, B) = \frac{1}{|G|} \sum_{g \in G} \max_{b \in B} IoU(b, g)
\end{equation}

Varying the diversification strategies produce different ABO scores, with a combination of all similarity measures giving the best ABO. The best color space seems to be HSV. Using a single diversification strategy with only HSV color space and all four similarity measures produces bounding boxes with $0.693$ ABO. The authors define two other selective search configurations: "fast" which uses 8 strategies and produces $0.799$ ABO, and "quality" with 80 strategies and producing $0.878$ ABO.

Due to the high ABO scores produced by selective search, it obtains $98-99$ \% recall on the PASCAL VOC test set. This shows that the produced bounding box are high quality and are usable for object detection/recognition purposes.
Two disadvantage of selective search are the large number of bounding boxes that are required to obtain high recall (over 1000), and the slow computational performance. Selective search in fast mode takes $3.8$ secs, while in quality mode it takes $17.2$ per image. These computation times are prohibitive for robotics applications.

Zitnick and Dollár \cite{zitnick2014edge} present EdgeBoxes, which is a detection proposal algorithm that computes objectness of a candidate bounding box from edge information. The authors make the observation that the number of contours fully enclosed by a bounding box gives information about how likely the box contains an object.

EdgeBoxes works as a sliding window detector. The structured edge detector is first applied to the input image, and then each candidate bounding box has an objectness score assigned. This is computed as the sum of edge strengths of all fully enclosed edges by the bounding box, minus the sum of edge strengths for all edges not fully enclosed (intersecting with the bounding box's boundary). The score is then normalized by the sum of the box width and height. A small threshold is applied to decide when to output a proposal from a candidate bounding box.

EdgeBoxes was evaluated on the PASCAL VOC 2007 dataset, where results show that it is superior to Selective Search, as EdgeBoxes requires to output less bounding boxes to produce the same recall (800 vs 10K). But the most important advantage of EdgeBoxes is that it is much more computationally efficient, with a computation time less than one second.

Edge information is quite unreliable in sonar images, mostly due to noise and non-uniform insonification. This means that an object's edges might have different strengths along its boundary. The edges produced by the shadow cast by the object are also a major problem in sonar images. EdgeBoxes was designed for color information and acoustic images violate many of the assumptions required by EdgeBoxes.

Cheng et al. \cite{cheng2014bing} also propose the use of gradient information to generate detection proposals, but their approach is quite different. This work notes that computing the gradient normalized magnitude of an image (NG), and given any bounding box, if the gradient of the bounding box's contents is down-sampled to a fixed size ($8 \times 8$), then using the raw NG features with a SVM classifier produces a powerful and fast objectness measure.

These features are used with a two-stage cascade of SVMs, first a linear SVM on the NG features, and then a size-specific SVM that produces a final objectness score. The authors also propose a binarized version of the NG features called BING. The advantage of this binary feature is that it is very fast to compute using specialized processor instructions. Their evaluation on the PASCAL VOC shows that they can obtain $96.2$ \% recall with 1000 proposals and $99.5$ \% recall with 5000 proposals.
Computational performance analysis of BING shows that it is very fast, at 3 milliseconds per image.

While BING is very fast, other publications \cite{hosang2016makes} have criticised that its high recall is only available at IoU threshold $O_t = 0.5$, with higher thresholds having considerably lower accuracy. This means that the algorithm is very fast but it cannot localize objects accurately. Its use of gradient information makes it a unlikely good candidate for sonar images.

Kuo et al. \cite{kuo2015deepbox} proposes the learning of an objectness score from a CNN. This work is the most similar to the one presented in this chapter, but big differences still exist. The method presented in this work is called DeepBox and it works by learning to predict a binary objectness decision from cropped and resized bounding boxes produced by EdgeBoxes.
Positive examples are obtained by perturbing ground truth bounding boxes, while negative examples are generated by sampling random bounding boxes and discarding the ones with IoU bigger than $0.5$ with ground truth. Positives and negatives are sampled at the $1:3$ ratio.

The authors used a modified version of AlexNet with less layers, removed in order to make it simpler and avoid overfitting. Evaluating on the PASCAL VOC 2007 dataset shows that DeepBox obtains better recall than EdgeBoxes, requiring less proposals to reach a certain recall target. DeepBox obtains $87$ \% recall at IoU threshold $O_t = 0.7$.

DeepBox does not actually learn an objectness score, but instead it learns a binary classifier to predict which bounding boxes are objects, and which are not. An objectness score could be implicitly present in the probabilities from the softmax activation.

There are two survey papers that cover detection proposals. Hosang et al. \cite{hosang2014good} evaluate the trade-offs made by different detection proposal methods on the PASCAL VOC 2007 dataset. The authors evaluate proposal repeatability, by introducing variations in test images, such as scale, rotation, illumination, and JPEG artifacts. BING and EdgeBoxes are the most repeatable methods, but all algorithms suffer a loss of repeatability as variations are introduced, specially for the most extreme ones.

The authors also point out that recall varies considerably across scales, as large windows are much easier to match and to produce a large recall than smaller windows. This paper also evaluates recall as a function of the number of proposals, where EdgeBoxes and Selective Search are the best methods, requiring less proposals for a given recall target, or achieving a higher overall recall.

Hosang et al. \cite{hosang2016makes} refine their results in a follow-up journal paper. This work repeats the previous experiments but using two new datasets: ImageNet 2013 object detection, and the COCO 2014 object detection dataset. The basic idea of this comparison is to check for overfitting to the PASCAL VOC object categories. But the authors see no general loss of recall performance on other datasets, as methods perform similarly. But on the COCO dataset there are some significant differences, such as EdgeBoxes performing poorly when compared to Selective Search, while other methods improve their performance.

This paper also introduces the average recall (AR) metric in order to predict object detection performance in a class-specific setting, which can be a proxy metric for the more used mean average precision (mAP). AR is computed as the mean recall score as the IoU overlap threshold is varied in the range $O_t \in [0.5, 1.0]$. Better proposals that lead to an improved class-specific object detector will reflect into a higher AR score.

We now cover two methods that use CNNs for detection proposals. The first is MultiBox \cite{erhan2014scalable}, which extends AlexNet to generate detection proposals. The authors propose to use a final layer that outputs a vector of $5K$ values, four of them corresponding to upper-left and lower-right corner coordinates of a bounding box, and a confidence score that represents objectness. The default boxes are obtained by applying K-means to the ground truth bounding boxes, which make them not translation invariant \footnote{This is mentioned in the Faster R-CNN paper as a big downside.} and could hurt generalization.

The assignment between ground truth bounding boxes and predictions by the network is made by solving an optimization problem. The idea is to assign the best ground truth bounding boxes in order to predict a high confidence score. These bounding boxes can then be ranked and less proposals should be needed to achieve high recall.

This network is trained on 10 million positive samples, consisting of patch crops that intersect the ground truth with at least IoU overlap threshold $O_t = 0.5$, and 20 million negative samples obtained from crops with IoU overlap threshold smaller than $O_t = 0.2$. On the PASCAL VOC 2012 dataset, this model can obtain close to $80$ \% recall with 100 proposals per image, which is a considerable improvement over previous work. One important detail is missing from this work, which is if the network is pre-trained on a dataset (likely ImageNet) or trained from scratch. Our experience tells us that training a bounding box regressor from scratch is very hard, so the only explanation for these results is due to the large size of the training set (30 million samples), which make it unusable for our purposes.

The second method is Faster R-CNN \cite{ren2015faster}. While this is a class-specific object detector, it contains a detection proposals sub-network called the Region Proposal Network (RPN). The RPN uses a pre-trained network (typically VGG16) and takes a $3 \times 3$ sliding window over the output feature map, giving this as input to a fully connected network outputting $256-512$ features that are fed to two sibling fully connected layers. One of these produces four coordinates corresponding to a bounding box location, and the other sibling layer produces two softmax scores indicating the presence of an object. This small network can be implemented as a fully convolutional network for efficient evaluation.

The RPN uses the concept of an anchor box. As proposals should cover a wide range of scale and aspect ratio, the RPN produces $k$ bounding box and objectness score predictions, each corresponding to a different anchor box. Then at test time, all anchors are predicted and the objectness score is used to decide final detections. One contribution of the RPN is that the anchors are translation invariant, which is very desirable in order to predict objects at multiple locations.

Training the RPN is not a simple process. Anchors must be labeled as objects or background. Given a set of ground truth bounding boxes, the anchor with the highest IoU overlap is given a positive label ($p_i = 1$), as well as any anchor with IoU larger than $O_t = 0.7$. Anchors with IoU smaller than $O_t = 0.3$ are given a negative label (background, $p_i = 0$). Then the RPN layers are trained using a multi-task loss function:

\begin{equation}
	L = N^{-1} \left(\sum_i CE(p_i, \hat{p}_1) + \lambda \sum_i p_i H(|t_i - \hat{t}_i|) \right)
\end{equation}

Where $p_i$ are the ground truth object label, $t_i$ is the true vector of normalized bounding box coordinates $(t_x, t_y, t_w, t_h)$, $\lambda$ is a trade-off factor used to combine both sub-losses, CE is the cross-entropy loss, and H is the Huber loss with $\delta = 1$, which is also called smooth $L_1$ loss:

\begin{equation}
	H_{\delta}(x) =
	\begin{cases}
		\frac{1}{2} x^2 				& \text{for} |x| < \delta\\
		\delta(|x| - \frac{1}{2}\delta) 	& \text{otherwise}
	\end{cases}
\end{equation}

Bounding box coordinates for regression are normalized by:

\begin{align*}
	t_x = \frac{x - x_a}{w_a} \qquad t_y = \frac{y - y_a}{h_a} \qquad t_w = \frac{w}{w_a} \qquad t_h = \frac{h}{h_a}
\end{align*}

Where the $a$ subscript denotes the anchor coordinates. The RPN is not evaluated as a standalone component, but as part of the whole Faster R-CNN pipeline, including the use of Fast R-CNN \cite{girshick2015fast} for object detection given a set of proposals. Mean average precision (mAP) on PASCAL VOC 2012 improves from $65.7$ \% when using Selective Search proposals to $67.0$ \% with RPN proposals.

Faster R-CNN was a milestone in object detection, being considerably faster than previous iterations (R-CNN and Fast R-CNN), while also being more accurate and introducing the RPN. We have tried to train similar networks performing bounding box regression on our Forward-Looking sonar images, but it fails to converge into a state that produces useful predictions. We believe that this is due to much a smaller training set and failure to learn the "appropriate" features by pre-training the network on a large dataset.
For comparison, the RPN is trained on the PASCAL VOC 07+12 dataset. The 2007 \cite[-3em]{everingham2010pascal} version of this dataset contains 5K training images with 12.5K labeled object instances, while the 2012 \cite[1em]{everingham2015pascal} version contains 11.5K training images with 31.5K object instances. The RPN has also been trained successfully on the COCO dataset \cite[1em]{lin2014microsoft}, with 328K images and 2.5 million labeled object instances. Both datasets vastly surpass our dataset of 2069 images with 2364 labeled object instances.
Additionally, the DSOD (Deeply Supervised Object Detection) \cite[1em]{shen2017dsod} also evaluates the RPN in an end-to-end approach without pre-training the network, and while it works well (and outperforms the RPN and Faster R-CNN) using a proposal-free approach, the authors mention that using the RPN in a proposal-based framework with DSOD failed to converge, which suggests that there are additional issues with the RPN formulation that are not related to the size of the training set. Additional research is needed about understanding how the RPN works and what is required for convergence.

\subsection{Discussion}

Detection proposals have been very successful in color images. The literature is rich in methods, evaluation, and advances. But none of these advances can be directly transferred into sonar images.

Many bottom-up proposal techniques use super-pixel segmentation, which is unreliable in sonar images. As we have previously mentioned in Chapter \ref{chapter:sonar-classification}, segmenting a sonar image is not an easy task and it might be very unreliable for complex objects such as marine debris.

Many other methods rely on manually crafted features, which do not generalize or plain do not work for sonar images. The cues that are typically used for objectness ranking are either unreliable or cannot be applied in sonar, specially features based on color or shape. In theory shape could be used but this requires precise segmentation of the shadow produced by the object.

Most recent methods are quite complex, combining multiple features and cues in order to improve performance, but hurting the ability to generalize to other kinds of images, as methods are too specialized to be usable in sonar. A method that uses feature learning instead is much more promising.

Finally, two large issues are common to most proposal algorithms. One is the large number of proposals that must be generated in order to obtain high recall. Many images in the PASCAL VOC dataset do not contain a comparable number of objects, which means that many generated proposals do not correspond to real objects in the image. This is computationally wasteful. Methods like MultiBox show that it is possible to get high recall with a low number of proposals, showing that other methods are generating unnecessary detections.

The second issue is computational performance. Many detection proposal methods were developed in order to "guide" \cite{hosang2014good} the object detection process, which should reduce the number of bounding boxes that have to be classified with respect to a classic sliding window, but now the added complexity from the detection proposal algorithm is driving the computational budget up. State of the art methods like Faster R-CNN run in real-time, but only if a GPU is used, as the prediction of a dense set of bounding boxes is computationally expensive.

It is clear that CNNs are the most promising method for generating detection proposals in sonar images, but bounding box prediction does not seem to work in a setting where large datasets are not available.

\section{Objectness Prediction with CNNs}

This section describes our detection proposals pipeline and the use of convolutional neural networks for objectness prediction.

\subsection{Objectness Prediction}
\label{sec:objPrediction}

We base our method on the idea that objectness can be estimated from an image window/patch. An objectness regressor can then be trained on such data to learn the relationship between image content and an abstract objectness score concept. This corresponds to a data-driven approach. Many machine learning models can be used for this, and we explore their differences in our evaluation.

First we estimate the ground truth objectness labels with a simple method.  As objectness is not a "measurable" quantity, but an abstract figure, we use a metric that we define as a proxy: The maximum IoU of a sliding window with the ground bounding boxes.

Our reasoning is that the IoU is maximum only when the sliding window perfectly contains the object \footnote{Assuming that labeled bounding boxes do contain the objects of interest.} and is strictly decreasing when the window moves away from the ground truth object. Zero objectness should be predicted from background windows, and this perfectly corresponds to IoU zero as the window does not intersect any ground truth object. As our objective is to train a regressor that predicts an objectness score from an image patch, having a diverse set of ground truth objectness value is desirable. 

But perfectly intersecting bounding boxes are quite unusual, specially for a fixed-size window as we use. In order to train a better regressor, we pass the IoU values through a squashing function in order to increase their range and cover the full range of objectness in $[0, 1]$:

\begin{equation}
	\text{objectness}(\text{iou}) = 
	\begin{cases} 
		1.0 		& \text{if iou} \geq 1.0 - \epsilon \\
		\text{iou}  & \text{if } 1.0 - \epsilon < \text{iou} < \epsilon\\
		0.0      	& \text{if } \text{iou} \leq \epsilon
	\end{cases}
	\label{proposals:iouToObjectness}
\end{equation}

We define $\epsilon$ as the range in objectness value that should be approximated to the closest value (either towards zero or one). This value depends on how well the bounding boxes fit a fixed-size sliding window. For our marine debris dataset we use $\epsilon = 0.2$.

A machine learning model is then trained to predict objectness values from $N \times N$ image patch pixels.

\subsection{Detection Proposals Pipeline}

\begin{marginfigure}[-2cm]
	\centering
	\includegraphics[width = 4cm]{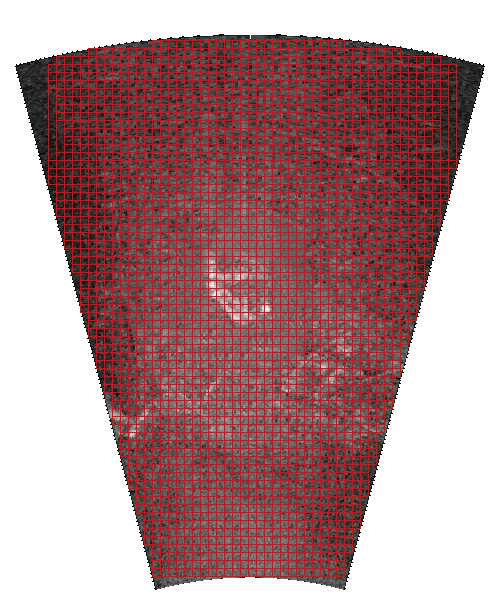}
	\vspace*{0.5cm}
	\caption{Sliding Window example on a polar field of view typical of forward-looking sonars. This grid was generated with $s = 8$. Many sliding windows overlap with each other.}
	\label{proposals:slidingWindowSample}
\end{marginfigure}

Once a objectness prediction model has been obtained, we can use it to generate detection proposals with a very simple method. Given an input sonar image, we slide a $N \times N$ sliding window with a given stride $s$, but only covering the parts of the image corresponding to the polar field of view. This is shown in Figure \ref{proposals:slidingWindowSample}. We use a strided sliding window in order to increase performance, as with $s > 1$ many sliding windows will be skipped, reducing the size of the search space. Using a strided sliding window does not seem to decrease recall unless very tight bounding boxes are required.

We use two methods to decide when a given objectness value produces a detection proposal:

\begin{itemize}
	\item \textbf{Objectness Thresholding}. Any image window having objectness larger or equal to a predefined threshold $T_o$ is output as a detection proposal. The value of $T_o$ must be carefully tuned, as a low value will produce many proposals with high recall, and a large value will generate less proposals with low recall. In general it is desirable to produce the smallest amount of proposals that reaches a given recall target.
	\item \textbf{Objectness Ranking}. Given a predefined number $k$ of detection proposals to output, we obtain all image windows and order them by objectness score in increasing order. The the top-$k$ imag,e windows by objectness score are returned as detection proposals. This method avoid the selection of a minimum objectness value and it is more adaptive to specific conditions in the field, allowing the detection of objects that are not present in the training set.
\end{itemize}

For our specific detection proposals algorithm we use $96 \times 96$ sliding windows on the image, with a window stride of $s = 8$ pixels. The final proposals are produced by applying non-maxima suppression \cite{gonzalezDIP2006}. This consists of greedily obtaining all proposals that intersect more than an IoU overlap threshold $S_t$ and from each pair, only keep the proposal with the highest objectness score. This suppresses duplicated proposals that are produced via the sliding window approach.

\subsection{Network Architectures}

We use two network architectures for objectness prediction. One is a CNN that is similar to ClassicNet, as previously used in Chapter \ref{chapter:sonar-classification}, but with slight modifications. The other is a CNN based on TinyNet, but it is designed to be transformed into a Fully Convolutional Network (FCN) \cite{long2015fully} for efficient evaluation of objectness computation in a full image.

The CNN architecture based on ClassicNet is shown in Figure \ref{proposals:basicArchitecture}. We call this architecture ClassicNet-Objectness. This is a six layer network, with a sigmoid activation at the output, in order to produce an objectness score in the $[0, 1]$ range. The input to this network is a $96 \times 96$ sonar image patch. The idea is that this network can be used in a sliding window fashion to produce an objectness map from a full size sonar image. ReLU activations are used in each layer except at the output, where a sigmoid activation is used instead. For regularization, we use Batch Normalization after each layer except the output layer.

This network is trained with a mean squared error loss for 5 epochs using ADAM. We found out that in order to generalize well, setting the right number of epochs is vital, as if the network is over-trained it will overfit to the objects in the training set. We obtained the number of epochs by monitoring the loss on a validation set, and stopping if it started increasing.

\begin{marginfigure}[-18em]
    \centering
    \begin{tikzpicture}[style={align=center, minimum height=0.5cm, minimum width=2.5cm}]		
    \node[draw] (A) {FC(c)};	
    
    \node[above=1em of A] (output) {Objectness};
    
    \node[above=1 em of output] (dummy) {};
    \node[draw, below=1em of A] (B) {FC(1)};
    \node[draw, below=1em of B] (C) {FC(96)};
    
    \node[draw, below=1em of C] (D) {MaxPool(2, 2)};
    \node[draw, below=1em of D] (E) {Conv($32$, $5 \times 5$)};
    
    \node[draw, below=1em of E] (F) {MaxPool(2, 2)};
    \node[draw, below=1em of F] (G) {Conv($32$, $5 \times 5$)};
    
    \node[below=1em of G] (Input) {$96 \times 96$ Image Patch};	
    
    \draw[-latex] (B) -- (A);
    \draw[-latex] (C) -- (B);
    \draw[-latex] (D) -- (C);
    \draw[-latex] (E) -- (D);
    \draw[-latex] (F) -- (E);
    \draw[-latex] (G) -- (F);
    \draw[-latex] (Input) -- (G);
    \draw[-latex] (A) -- (output);
    \end{tikzpicture}
    \caption{CNN architecture based on ClassicNet for objectness prediction.}
    \label{proposals:basicArchitecture}
\end{marginfigure}

\begin{marginfigure}
	\centering
	\begin{tikzpicture}[style={align=center, minimum height=0.5cm, minimum width=2.5cm}]			
	\node(output) {Objectness};
	
	\node[draw, below=1em of output] (B) {FC(1)};
	
	\node[draw, below=1em of B] (C) {MaxPool(2, 2)};
	\node[draw, below=1em of C] (D) {Conv($24$, $1 \times 1$)};
	\node[draw, below=1em of D] (E) {Conv($24$, $3 \times 3$)};
	
	\node[draw, below=1em of E] (F) {MaxPool(2, 2)};
	\node[draw, below=1em of F] (G) {Conv($24$, $1 \times 1$)};
	\node[draw, below=1em of G] (H) {Conv($24$, $3 \times 3$)};
	
	\node[below=1em of H] (Input) {$96 \times 96$ Image Patch};	
	
	\draw[-latex] (B) -- (output);
	\draw[-latex] (C) -- (B);
	\draw[-latex] (D) -- (C);
	\draw[-latex] (E) -- (D);
	\draw[-latex] (F) -- (E);
	\draw[-latex] (G) -- (F);
	\draw[-latex] (H) -- (G);
	\draw[-latex] (Input) -- (H);
	\end{tikzpicture}
	\caption{TinyNet-Objectness Architecture for objectness prediction.}
	\label{proposals:fcnBaseArchitecture}
\end{marginfigure}

The second architecture based on TinyNet is shown in Figure \ref{proposals:fcnBaseArchitecture}. We call this architecture as TinyNet-Objectness. This represents a base architecture that is later transformed into a fully convolutional network. This network has seven layers, with a combined use of $3 \times 3$ and $1 \times 1$ filters plus Max-Pooling. One difference of this network when compared to TinyNet is that it uses a fully connected layer with a sigmoid activation to predict objectness, while TinyNet uses global average pooling combined with a softmax activation for classification. 

This network is also trained with a mean squared error loss for 15 epochs, using ADAM. We applied the same methodology as before, but the loss converges in more epochs, but does not seem to produce overfitting. This network has a final loss that is slightly higher than the previous one, but still performs adequately. Note that we do not use regularization as part of this network, as using Batch Normalization prevents us into transforming the CNN into a FCN due to the fixed sizes in the Batch Normalization parameters. As mentioned before, this network performs well and does not seem to overfit.

\begin{marginfigure}
    \centering
    \begin{tikzpicture}[style={align=center, minimum height=0.5cm, minimum width=2.5cm}]			
    \node(output) {Objectness Map};
    
    \node[draw, below=1em of output] (B) {Conv($1$, $24 \times 24$)};
    
    \node[draw, below=1em of B] (C) {MaxPool(2, 2)};
    \node[draw, below=1em of C] (D) {Conv($24$, $1 \times 1$)};
    \node[draw, below=1em of D] (E) {Conv($24$, $3 \times 3$)};
    
    \node[draw, below=1em of E] (F) {MaxPool(2, 2)};
    \node[draw, below=1em of F] (G) {Conv($24$, $1 \times 1$)};
    \node[draw, below=1em of G] (H) {Conv($24$, $3 \times 3$)};
    
    \node[below=1em of H] (Input) {Sonar Image};	
    
    \draw[-latex] (B) -- (output);
    \draw[-latex] (C) -- (B);
    \draw[-latex] (D) -- (C);
    \draw[-latex] (E) -- (D);
    \draw[-latex] (F) -- (E);
    \draw[-latex] (G) -- (F);
    \draw[-latex] (H) -- (G);
    \draw[-latex] (Input) -- (H);
    \end{tikzpicture}
    \caption[TinyNet-FCN-Objectness Architecture for objectness prediction]{TinyNet-FCN-Objectness Architecture for objectness prediction. This is a fully convolutional network version of TinyNet-Objectness}
    \label{proposals:fcnArchitecture}
\end{marginfigure}

\begin{marginfigure}[0.5cm]
	\centering
	\stackunder[5pt] {
		\includegraphics[width = 3.8cm]{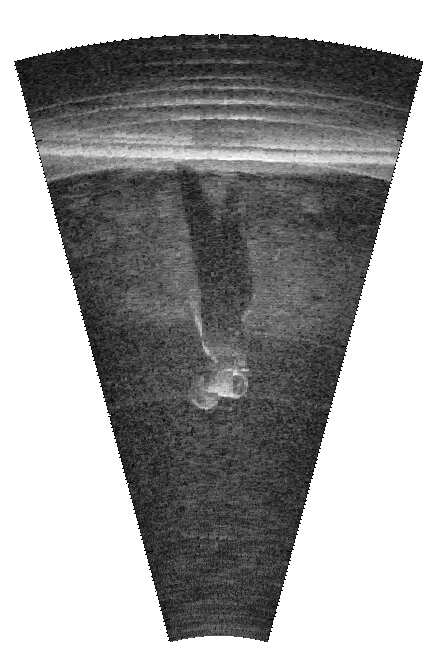}
	}{Sonar Image}
	\stackunder[5pt] {
		\includegraphics[width = 3.8cm]{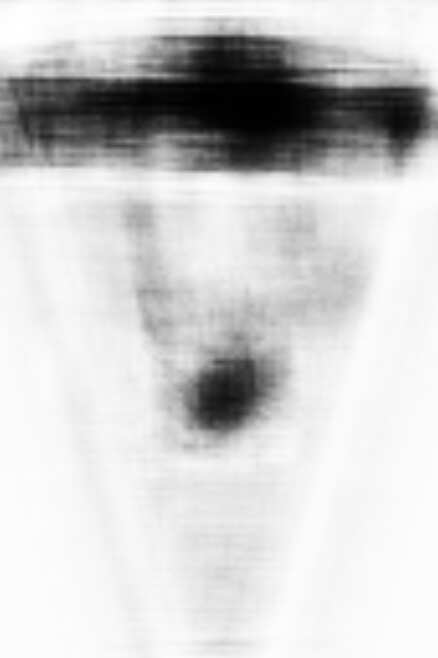}
	}{Objectness Map}
	\vspace*{0.5cm}
	\caption{Objectness Map produced by TinyNet-FCN-Objectness on a given input image}
	\label{proposals:fcnObjectnessSample}
\end{marginfigure}

After training, we convert TinyNet-Objectness into a fully convolutional network by replacing the final fully connected layer with a equivalent convolutional layer. The trained weights have to reshaped to match weights that can be used by a convolutional layer. This is done by creating a convolutional layer with number of filters equals to the output dimensionality of the FC layer, and with filter size equal to the output of the previous layer, which corresponds to the last Max-Pooling layer in our case.

Note that the network is trained on $96 \times 96$ image patches, same as ClassicNet-Objectness, but then it can be evaluated in a fully convolutional way in order to improve performance, as applying a sliding window to ClassicNet-Objectness is computationally very expensive, and does not exploit shared computation in the convolutional layers.

The fully convolutional version of TinyNet-Objectness is called TinyNet-FCN-Objectness. This network can now take variable-sized images as input, producing also a variable-size image as output. The relationship between the input and output dimensionalities depend on the level of down-sampling in the network architecture (mostly by Max-Pooling). As we desire outputs that have the same spatial dimensions as the input, we up-sample the output using bilinear interpolation. Then this objectness map can be used with our sliding window detection proposals pipeline. One example of such up-sampled objectness maps produced by this network is shown in Figure \ref{proposals:fcnObjectnessSample}.

\section{Experimental Evaluation}

In this section we perform an extensive experimental evaluation of our proposed detection proposal algorithms.

\subsection{Training and Datasets}

We have three datasets which are used to train and test this method:

\begin{itemize}
	\item \textbf{Training}: This is a dataset of 51653 $96 \times 96$ sonar image patches, obtained and labeled with the methodology described in Section \ref{sec:objPrediction}. CNN models are trained on this dataset. We perform data augmentation by flipping images left-right and up-down, which increases the amount of data by three times.
	\item \textbf{Validation}: This is a dataset of 22137 $96 \times 96$ sonar image patches, used for validation during network training. We report the mean squared error on this dataset.
	\item \textbf{Testing}: This dataset contains 770 full-size sonar images where we test our detection proposals pipeline. These images are variable sized, as they depend on the specific sonar configuration that was used when capturing them. We report recall, computation time and average best overlap (ABO) on this dataset.
\end{itemize}

\subsection{Evaluation Metrics}

As mentioned in the related work, the main metric used to evaluate detection proposal algorithms is recall.

From the Marine Debris task perspective, the recall metric measures how well marine debris is found by the detector, but false positives might happen when the number of detections is large. For this reason we also wish to minimize the number of detections that the detector produces, as a high quality proposals algorithm will have a high recall with a low number of detections.

We also considered to evaluate precision, as this explicitly accounts for false positives, but it would be inconsistent with the proposals literature, and it could bias the detector to only detect the objects in the training set. Recall is well correlated \cite{hosang2016makes} with object detection performance, and unlike the more common mAP metric used in object detection, it is easier to interpret for a human.

To provide a more comprehensive evaluation, we decided to use three additional metrics. For each image in our test set, we run the proposals algorithm and evaluate all proposed bounding boxes with the following metrics:

\begin{itemize}
	\item \textbf{Recall}: Number of correctly detected objects divided by the total number of ground truth objects in the image. We use a minimum IoU overlap threshold of $O_t = 0.5$, but we also evaluate higher values of $O_t$ for proposal localization quality.
	\item \textbf{Number of Proposals}: We compute the mean and standard deviation of the total number of generated proposals across the images in our test set. A good proposal method should achieve high recall with a minimum number of proposals.
	\item \textbf{Average Best Overlap}: Mean value of the best overlap score for each ground truth object. This metric tells how well the generated proposals match the ground truth bounding boxes.
	\item \textbf{Computation Time}: Mean and standard deviation of computation time across the whole dataset. This was evaluated on the high-power platform mentioned in Section \ref{sic:compPerformanceSection}. A fast detection proposal method is desirable for use in real robot platforms.
\end{itemize}

The use of the computation time metric is motivated by the Marine Debris task, as a typical AUV has limited computational resources and energy, so a fast proposals algorithm is preferable. The Average Best Overlap  metric measures how well the produced bounding boxes match the ground truth, indicating the quality of object localization in the image. High quality object location information is desirable for the sub-task of manipulation and picking the debris object.

As we have two methods to generate detection proposals from objectness prediction, we evaluate both with respect to their tunable parameters: the objectness threshold $T_o$ and the number of ranked proposals $k$. All metrics vary considerably with the tunable parameters, and we can decide their optimal values by evaluating on a validation set. For some experiments, we also vary the non-maxima suppression threshold $S_t$ as it affects recall and the number of proposals. We evaluate each model (CNN or FCN) separately and compare against the template matching baseline.

\subsection{Baseline}

As there is no previous work in detection proposals for sonar images, it is not trivial to define baselines. We produce a baseline by using template matching as a generic objectness score. Cross-correlation Template matching \marginnote{More information about template matching is available in Chapter \ref{chapter:sonar-classification}} is commonly used for image classification and object detection, and a simple modification can transform it into an objectness score.

We randomly select a set of $N$ templates from the training set, and apply cross-correlation as a sliding window between the input image (with size $(W, H)$) and each template. This produces a set of images that correspond to the response of each template, with dimensions $(N, W, H)$. To produce a final objectness score, we take the maximum value across the template dimensions, producing a final image with size $(W, H)$. Taking the maximum makes sense, in order to make sure that only the best matching template produces an objectness score. As cross-correlation takes values in the $[-1, 1]$ range, we produce objectness values in the $[0, 1]$ range by just setting any negative value to zero.

Then we use the produced objectness scores with our detections proposal pipeline, with both objectness thresholding and ranking. We decided to use $N = 100$, as this produces the best recall in our experiments, and there is little variation in recall due to the random selection of templates. Using more templates only degrades computation time without increasing recall.

For comparison with the state of the art detection proposals methods, we selected Selective Search and EdgeBoxes, as both have public implementations in the OpenCV library. For EdgeBoxes the score threshold can be tuned, as well as a number of proposals can be set. We evaluate both parameters by selecting a low score threshold $0.0001$ at a fixed number of 300 proposals, and using a $0.0$ score threshold and varying the number of output proposals.

For Selective Search, we evaluate both the Quality and Fast configurations, with a variable number of output proposals.

\subsection{Proposals from Objectness Thresholding}

We now evaluate our objectness thresholding method, by setting $T_o$ to values $[0.05, 0.10, 0.15, ..., 1.0]$ and computing the previously mentioned metrics for each threshold value. We also vary the non-maxima suppression threshold $S_t$ with values $[0.5, 0.6, 0.7, 0.8, 0.9]$. Higher values for $S_t$ remove less proposals as a more strict overlap threshold must be met, while smaller values for $S_t$ will suppress more proposals and reduce recall slightly.

\begin{figure*}[t]
	\centering
	\begin{tikzpicture}
		\begin{customlegend}[legend columns = 3,legend style = {column sep=1ex}, legend cell align = left,
		legend entries={CNN NMS $S_t = 0.5$, CNN NMS $S_t = 0.6$, CNN NMS $S_t = 0.7$, CNN NMS $S_t = 0.8$, CNN NMS $S_t = 0.9$, CC TM}]
			\addlegendimage{mark=none,red}
			\addlegendimage{mark=none,blue}			
			\addlegendimage{mark=none,green}
			\addlegendimage{mark=none,violet}
			\addlegendimage{mark=none,orange}
			\addlegendimage{mark=none,black}
		\end{customlegend}
	\end{tikzpicture}
	\subfloat[Recall vs Threshold]{
		\begin{tikzpicture}
		\begin{axis}[height = 0.25 \textheight, width = 0.302\textwidth, xlabel={Objectness Threshold ($T_o$)}, ylabel={Recall (\%)}, xmin = 0.0, xmax = 1.0, ymin = 40.0, ymax = 100.0, xtick = {0.0, 0.1, 0.2, 0.3, 0.4, 0.5, 0.6, 0.7, 0.8, 0.9, 1.0}, ytick = {40, 50, 60, 70, 80, 85, 90, 95, 100}, ymajorgrids=true, xmajorgrids=true, grid style=dashed, legend pos = north east, legend style={font=\scriptsize}, tick label style={font=\scriptsize, rotate=90}]
				
		\addplot+[mark = none] table[x  = threshold, y  = meanRecall, col sep = space] {chapters/data/proposals/thresholdVsRecallNMS0.50.csv};
		
		\addplot+[mark = none] table[x  = threshold, y  = meanRecall, col sep = space] {chapters/data/proposals/thresholdVsRecallNMS0.60.csv};
		
		\addplot+[mark = none] table[x  = threshold, y  = meanRecall, col sep = space] {chapters/data/proposals/thresholdVsRecallNMS0.70.csv};
		
		\addplot+[mark = none] table[x  = threshold, y  = meanRecall, col sep = space] {chapters/data/proposals/thresholdVsRecallNMS0.80.csv};
		
		\addplot+[mark = none] table[x  = threshold, y  = meanRecall, col sep = space] {chapters/data/proposals/thresholdVsRecallNMS0.90.csv};
		
		\addplot+[mark = none, black] table[x  = threshold, y  = meanRecall, col sep = space] {chapters/data/proposals/tm/tmProposals-TSPC100-thresholdVsRecallNMS0.00.csv};
		
		\end{axis}        
		\end{tikzpicture}
	}
	\subfloat[Number of Proposals vs Threshold]{
		\begin{tikzpicture}
		\begin{axis}[height = 0.25 \textheight, width = 0.30\textwidth, xlabel={Objectness Threshold ($T_o$)}, ylabel={\# of Proposals}, xmin=0.0, xmax = 1.0, ymin = 0.0, xtick = {0.0, 0.1, 0.2, 0.3, 0.4, 0.5, 0.6, 0.7, 0.8, 0.9, 1.0}, ytick = {10, 50, 100, 200, 300}, ymajorgrids=true, xmajorgrids=true, grid style=dashed, legend pos = north east, legend style={font=\scriptsize}, tick label style={font=\scriptsize, rotate=90}]
		
		\addplot+[mark = none] table[x  = threshold, y  = meanNumProposals, col sep = space] {chapters/data/proposals/thresholdVsRecallNMS0.50.csv};
		
		\addplot+[mark = none] table[x  = threshold, y  = meanNumProposals, col sep = space] {chapters/data/proposals/thresholdVsRecallNMS0.60.csv};
		
		\addplot+[mark = none] table[x  = threshold, y  = meanNumProposals, col sep = space] {chapters/data/proposals/thresholdVsRecallNMS0.70.csv};
		
		\addplot+[mark = none] table[x  = threshold, y  = meanNumProposals, col sep = space] {chapters/data/proposals/thresholdVsRecallNMS0.80.csv};
		
		\addplot+[mark = none] table[x  = threshold, y  = meanNumProposals, col sep = space] {chapters/data/proposals/thresholdVsRecallNMS0.90.csv};
		
		\addplot+[mark = none, black] table[x  = threshold, y  = meanNumProposals, col sep = space] {chapters/data/proposals/tm/tmProposals-TSPC100-thresholdVsRecallNMS0.00.csv};
		\end{axis}        
		\end{tikzpicture}
	}	
	\subfloat[Average Best Overlap vs Threshold]{
		\begin{tikzpicture}
		\begin{axis}[height = 0.25 \textheight, width = 0.30\textwidth, xlabel={Objectness Threshold ($T_o$)}, ylabel={Average Best Overlap (ABO)}, xmin = 0, xmax = 1.0, ymin = 0.0, ymax = 1.0, xtick = {0.0, 0.1, 0.2, 0.3, 0.4, 0.5, 0.6, 0.7, 0.8, 0.9, 1.0}, ytick = {0.0, 0.1, 0.2, 0.3, 0.4, 0.5, 0.6, 0.7, 0.8, 0.9, 1.0}, ymajorgrids=true, xmajorgrids=true, grid style=dashed, legend pos = north east, legend style={font=\scriptsize}, tick label style={font=\scriptsize, rotate=90}]
		
		\addplot+[mark = none] table[x  = threshold, y  = aBO, col sep = space] {chapters/data/proposals/thresholdVsRecallNMS0.50.csv};
		
		\addplot+[mark = none] table[x  = threshold, y  = aBO, col sep = space] {chapters/data/proposals/thresholdVsRecallNMS0.60.csv};
		
		\addplot+[mark = none] table[x  = threshold, y  = aBO, col sep = space] {chapters/data/proposals/thresholdVsRecallNMS0.70.csv};
		
		\addplot+[mark = none] table[x  = threshold, y  = aBO, col sep = space] {chapters/data/proposals/thresholdVsRecallNMS0.80.csv};
		
		\addplot+[mark = none] table[x  = threshold, y  = aBO, col sep = space] {chapters/data/proposals/thresholdVsRecallNMS0.90.csv};
		
		\addplot+[mark = none, black] table[x  = threshold, y  = aBO, col sep = space] {chapters/data/proposals/tm/tmProposals-TSPC100-thresholdVsRecallNMS0.00.csv};
		
		\end{axis}        
		\end{tikzpicture}
	}
	\vspace*{0.5cm}
	\caption{ClassicNet with Objectness Thresholding: Detection proposal results over our dataset.}
	\label{proposals:cnnThresholdingPlotResults}
	\vspace*{0.5cm}
\end{figure*}
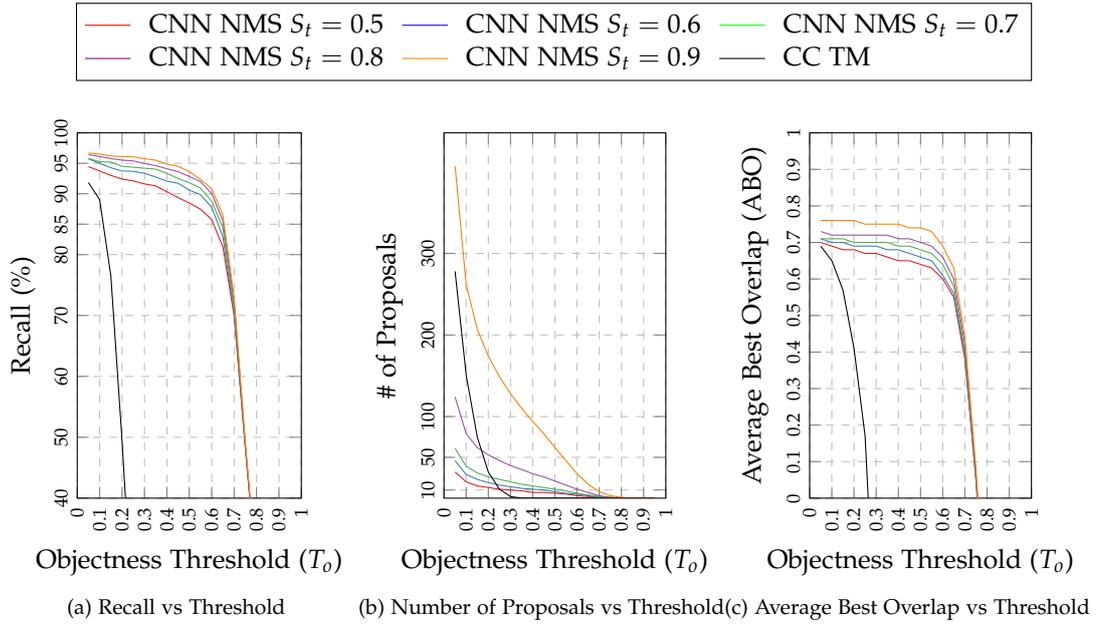

ClassicNet results are shown in Figure \ref{proposals:cnnThresholdingPlotResults}. These figures show that our method can achieve high recall, close to $95$ \% for many combinations of $S_t$. Increasing the NMS threshold decreases recall and the number of proposals significantly, and a good trade-off between high recall and low number of proposals seems to be $S_t = 0.8$ or $S_t = 0.7$. At objectness threshold $T_o = 0.5$ our method produces $94$ \% recall with less than 60 proposals per image. In comparison, up to 1480 candidate bounding boxes are evaluated, showing that the CNN has a high discriminative power between objects and non-objects.

Looking at ABO, all NMS thresholds produce very good scores, considerably far from the overlap threshold $O_t = 0.5$. Increasing the NMS threshold $S_t$ has the effect of reducing the ABO, which explains why recall also decreases.

Computation time for ClassicNet is $12.8 \pm 1.9$ seconds, which is quite far from real-time performance needed for robotics applications. This is expected as sliding a CNN over the image does not share feature maps across the image.

The template matching baseline performs quite poorly. Even as it can reach $91$ \% recall, it requires a very low threshold to obtain such performance, producing over 250 proposals per image. Its recall and number of proposals drop very quickly with increasing values of $T_o$. These results show that template matching cannot be used for reliable detection proposal generation.

\begin{figure*}[t]
	\centering
	\begin{tikzpicture}
	\begin{customlegend}[legend columns = 3,legend style = {column sep=1ex}, legend cell align = left,
	legend entries={FCN NMS $S_t = 0.5$, FCN NMS $S_t = 0.6$, FCN NMS $S_t = 0.7$, FCN NMS $S_t = 0.8$, FCN NMS $S_t = 0.9$, CC TM}]
	\addlegendimage{mark=none,red}
	\addlegendimage{mark=none,blue}			
	\addlegendimage{mark=none,green}
	\addlegendimage{mark=none,violet}
	\addlegendimage{mark=none,orange}
	\addlegendimage{mark=none,black}
	\end{customlegend}
	\end{tikzpicture}
	\subfloat[Recall vs Threshold]{
		\begin{tikzpicture}
		\begin{axis}[height = 0.25 \textheight, width = 0.30\textwidth, xlabel={Objectness Threshold ($T_o$)}, ylabel={Recall (\%)}, xmin = 0.0, xmax = 1.0, ymin = 40.0, ymax = 100.0, xtick = {0.0, 0.1, 0.2, 0.3, 0.4, 0.5, 0.6, 0.7, 0.8, 0.9, 1.0}, ytick = {40, 50, 60, 70, 80, 85, 90, 95, 100}, ymajorgrids=true, xmajorgrids=true, grid style=dashed, legend pos = north east, legend style={font=\scriptsize}, tick label style={font=\scriptsize, rotate=90}]
		
		\addplot+[mark = none] table[x  = threshold, y  = meanRecall, col sep = space] {chapters/data/proposals/fcnProposals-thresholdVsRecallNMS0.50.csv};
		
		\addplot+[mark = none] table[x  = threshold, y  = meanRecall, col sep = space] {chapters/data/proposals/fcnProposals-thresholdVsRecallNMS0.60.csv};
		
		\addplot+[mark = none] table[x  = threshold, y  = meanRecall, col sep = space] {chapters/data/proposals/fcnProposals-thresholdVsRecallNMS0.70.csv};
		
		\addplot+[mark = none] table[x  = threshold, y  = meanRecall, col sep = space] {chapters/data/proposals/fcnProposals-thresholdVsRecallNMS0.80.csv};
		
		\addplot+[mark = none] table[x  = threshold, y  = meanRecall, col sep = space] {chapters/data/proposals/fcnProposals-thresholdVsRecallNMS0.90.csv};
		
		\addplot+[mark = none, black] table[x  = threshold, y  = meanRecall, col sep = space] {chapters/data/proposals/tm/tmProposals-TSPC100-thresholdVsRecallNMS0.00.csv};
		
		\end{axis}        
		\end{tikzpicture}
	}
	\subfloat[Number of Proposals vs Threshold]{
		\begin{tikzpicture}
		\begin{axis}[height = 0.25 \textheight, width = 0.30\textwidth, xlabel={Objectness Threshold ($T_o$)}, ylabel={\# of Proposals}, xmin=0.0, xmax = 1.0, ymin = 0.0, xtick = {0.0, 0.1, 0.2, 0.3, 0.4, 0.5, 0.6, 0.7, 0.8, 0.9, 1.0}, ytick = {50, 100, 200, 300, 400, 500,700,900}, ymajorgrids=true, xmajorgrids=true, grid style=dashed, legend pos = north east, legend style={font=\scriptsize}, tick label style={font=\scriptsize, rotate=90}]
		
		\addplot+[mark = none] table[x  = threshold, y  = meanNumProposals, col sep = space] {chapters/data/proposals/fcnProposals-thresholdVsRecallNMS0.50.csv};
		
		\addplot+[mark = none] table[x  = threshold, y  = meanNumProposals, col sep = space] {chapters/data/proposals/fcnProposals-thresholdVsRecallNMS0.60.csv};
		
		\addplot+[mark = none] table[x  = threshold, y  = meanNumProposals, col sep = space] {chapters/data/proposals/fcnProposals-thresholdVsRecallNMS0.70.csv};
		
		\addplot+[mark = none] table[x  = threshold, y  = meanNumProposals, col sep = space] {chapters/data/proposals/fcnProposals-thresholdVsRecallNMS0.80.csv};
		
		\addplot+[mark = none] table[x  = threshold, y  = meanNumProposals, col sep = space] {chapters/data/proposals/fcnProposals-thresholdVsRecallNMS0.90.csv};
		
		\addplot+[mark = none, black] table[x  = threshold, y  = meanNumProposals, col sep = space] {chapters/data/proposals/tm/tmProposals-TSPC100-thresholdVsRecallNMS0.00.csv};
		\end{axis}        
		\end{tikzpicture}
	}	
	\subfloat[Average Best Overlap vs Threshold]{
		\begin{tikzpicture}
		\begin{axis}[height = 0.25 \textheight, width = 0.30\textwidth, xlabel={Objectness Threshold ($T_o$)}, ylabel={Average Best Overlap (ABO)}, xmin = 0, xmax = 1.0, ymin = 0.0, ymax = 1.0, xtick = {0.0, 0.1, 0.2, 0.3, 0.4, 0.5, 0.6, 0.7, 0.8, 0.9, 1.0}, ytick = {0.0, 0.1, 0.2, 0.3, 0.4, 0.5, 0.6, 0.7, 0.8, 0.9, 1.0}, ymajorgrids=true, xmajorgrids=true, grid style=dashed, legend pos = north east, legend style={font=\scriptsize}, tick label style={font=\scriptsize, rotate=90}]
		
		\addplot+[mark = none] table[x  = threshold, y  = aBO, col sep = space] {chapters/data/proposals/fcnProposals-thresholdVsRecallNMS0.50.csv};
		
		\addplot+[mark = none] table[x  = threshold, y  = aBO, col sep = space] {chapters/data/proposals/fcnProposals-thresholdVsRecallNMS0.60.csv};
		
		\addplot+[mark = none] table[x  = threshold, y  = aBO, col sep = space] {chapters/data/proposals/fcnProposals-thresholdVsRecallNMS0.70.csv};
		
		\addplot+[mark = none] table[x  = threshold, y  = aBO, col sep = space] {chapters/data/proposals/fcnProposals-thresholdVsRecallNMS0.80.csv};
		
		\addplot+[mark = none] table[x  = threshold, y  = aBO, col sep = space] {chapters/data/proposals/fcnProposals-thresholdVsRecallNMS0.90.csv};
		
		\addplot+[mark = none, black] table[x  = threshold, y  = aBO, col sep = space] {chapters/data/proposals/tm/tmProposals-TSPC100-thresholdVsRecallNMS0.00.csv};
		
		\end{axis}        
		\end{tikzpicture}
	}
	\vspace*{0.5cm}
	\caption{TinyNet-FCN with Objectness Thresholding: Detection proposal results over our dataset.}
	\label{proposals:fcnThresholdingPlotResults}	
\end{figure*}
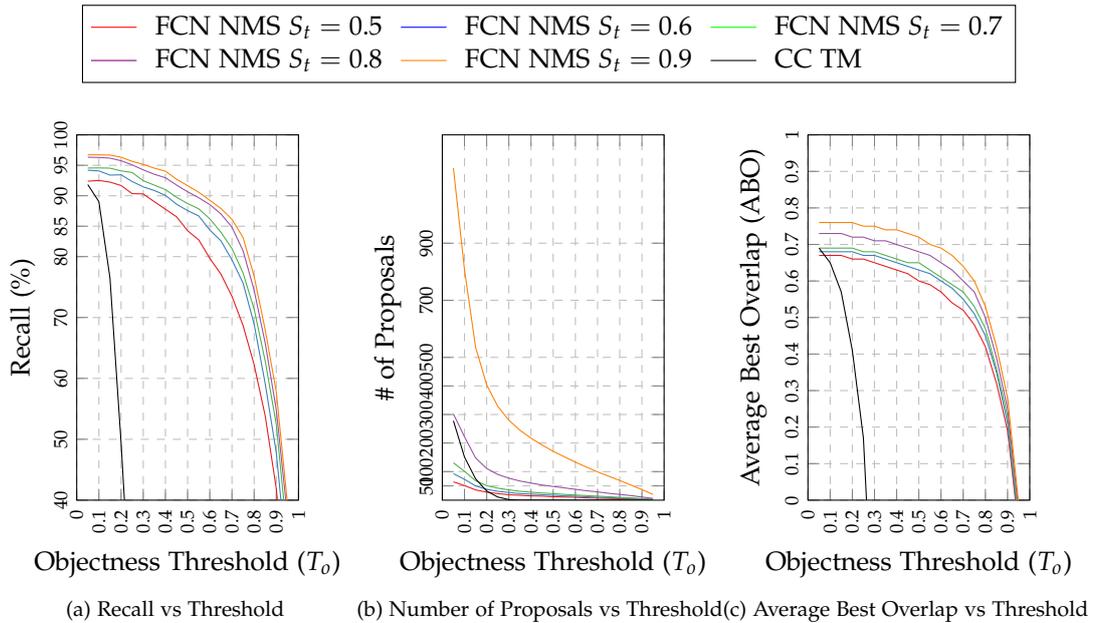

Results for TinyNet-FCN are shown in Figure \ref{proposals:fcnThresholdingPlotResults}. Objectness produced by this network also works quite well, but achieving $95$ \% recall only with a lower threshold $T_o = 0.3$ and 75 proposals per image. One considerable difference with ClassicNet objectness is that the Recall-$T_o$ plot covers the whole objectness threshold range, while ClassicNet drops close to zero for $T_o = 0.8$. This shows that the scores produced by TinyNet-FCN cover the whole objectness range ($[0, 1]$) instead of being clustered to be less than $0.8$ as ClassicNet does.

With the thresholding method, TinyNet-FCN produces more proposals and needs a considerable number of proposals to achieve high recall. This number of proposals can be controller by setting a lower NMS threshold $S_t$, but this reduces recall by up to $5$ \%.

ABO scores in TinyNet-FCN are similar to the ones measured for ClassicNet, showing that proposal localization has a similar quality. The only reason for lower recall are less discriminative objectness scores produced by TinyNet-FCN.

TinyNet-FCN is considerably faster than ClassicNet, at $3.3 \pm 1.0$ seconds per image, corresponding to a speedup of $3.9$. This is easily explained as the FCN computes all feature maps once, on the full-size image, and then use a $24 \times 24$ convolution to produce the final output objectness map. It was expected that this method would be faster, but up-scaling the final objectness map to match the original input image size also contributes significantly to the computation time. In any case, a method that is almost 4 times faster seems to be a good trade-off for $5$ \% less recall, still easily reaching over $90$ \%.

\begin{figure*}[p]
	\centering
    \forceversofloat
	\subfloat[ClassicNet Objectness]{
        \begin{tabular}[b]{c}
		\includegraphics[height=0.22\textheight]{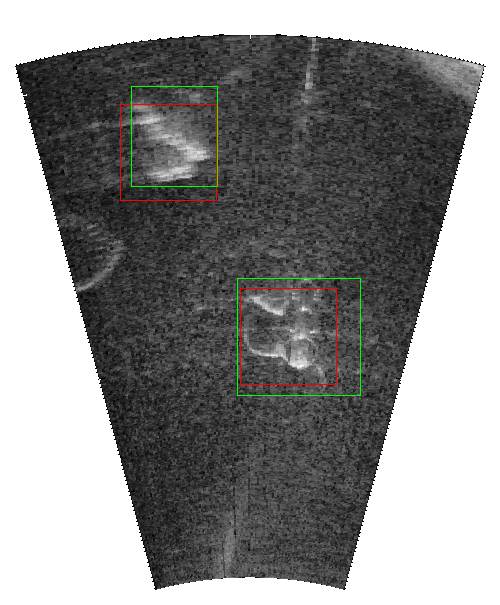}
		\includegraphics[height=0.22\textheight]{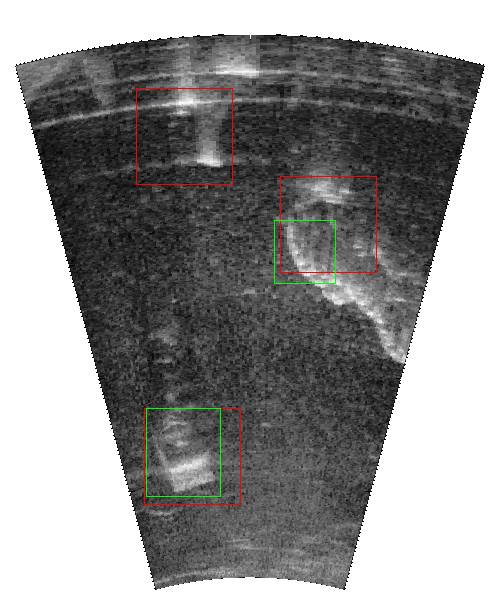}
		\includegraphics[height=0.22\textheight]{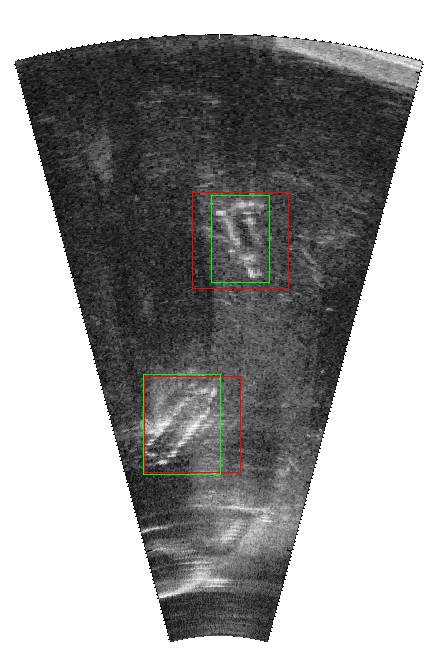}\\
		\includegraphics[height=0.22\textheight]{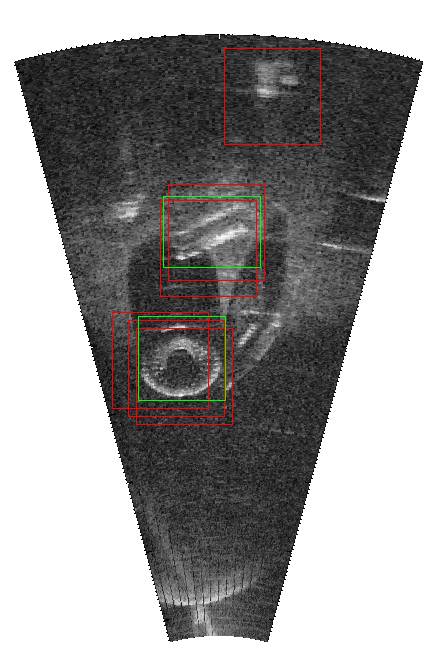}
		\includegraphics[height=0.22\textheight]{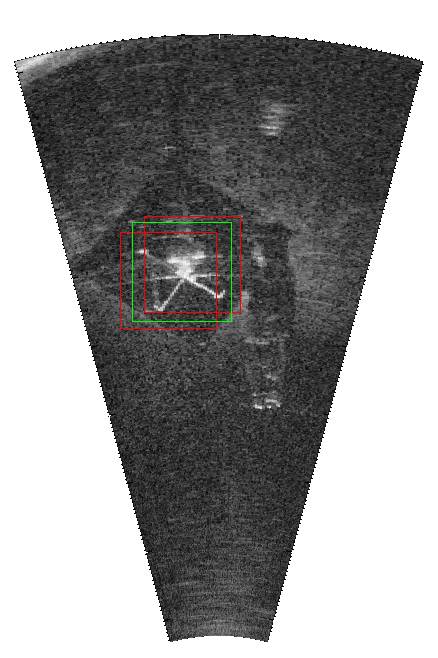}
        \end{tabular}
	}
	
	\subfloat[TinyNet-FCN Objectness]{
        \begin{tabular}[b]{c}
		\includegraphics[height=0.22\textheight]{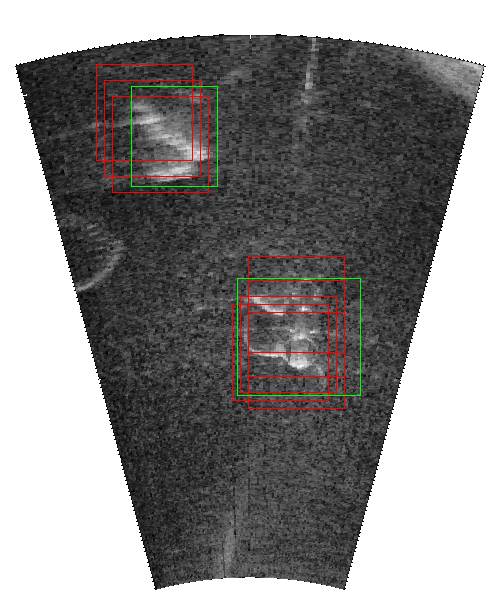}
		\includegraphics[height=0.22\textheight]{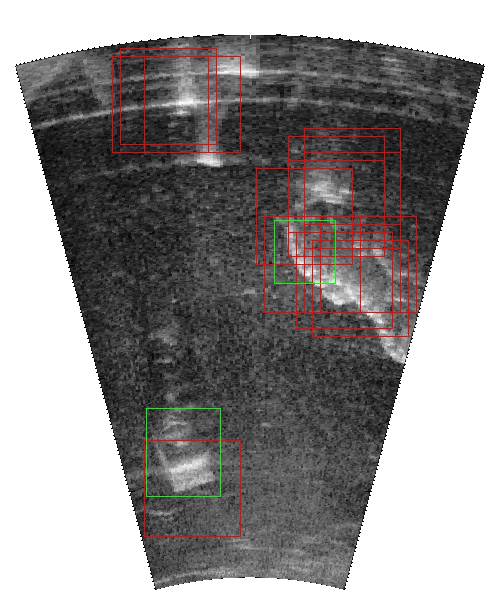}
		\includegraphics[height=0.22\textheight]{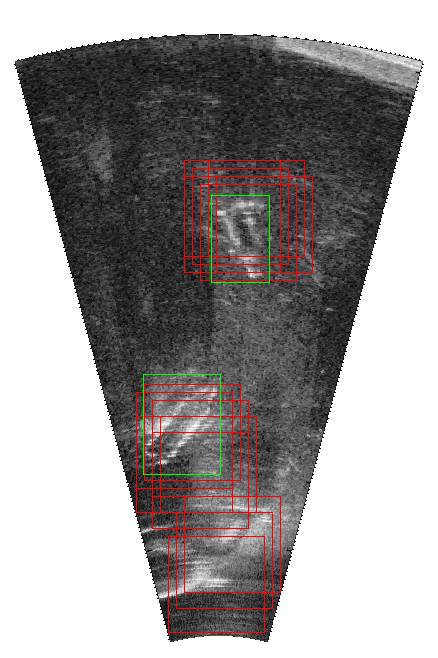}\\        
		\includegraphics[height=0.22\textheight]{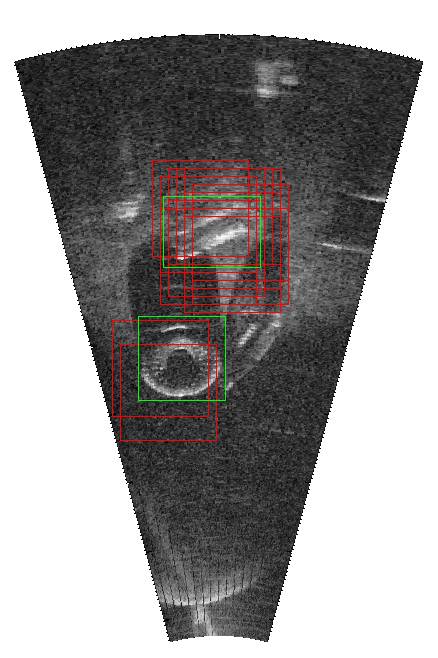}
		\includegraphics[height=0.22\textheight]{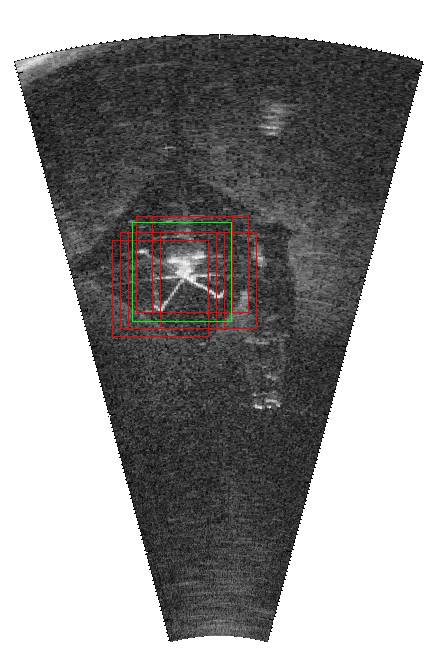}
        \end{tabular}
	}	
	\vspace*{-1.0cm}
	\caption{Sample detections using objectness thresholding with $T_o = 0.6$ and NMS threshold $S_t = 0.7$}
	\label{proposals:thresholdDetectionSamples}
\end{figure*}

Figure \ref{proposals:thresholdDetectionSamples} shows a cherry-picked sample of detections over full sonar images produced ClassicNet and TinyNet-FCN. This example shows how more proposals are generated with objectness computed by TinyNet-FCN, as this network produces scores that are better distributed into the full $[0, 1]$ range. Increasing the threshold $T_o$ will reduce the number of proposals.

\subsection{Proposals from Objectness Ranking}

In this section we evaluate our objectness ranking to produce proposals. We set $K$ to $[1, 2, 3, ..., 10, 20, 30, ... ,100]$ and compute metrics for each value of $K$. We also vary the non-maxima suppression threshold $S_t$ with values $[0.5, 0.6, 0.7, 0.8, 0.9]$. Non-maxima suppression is applied before ranking proposals, in order to prevent it from reducing the number of proposals that are output below a given value of $K$. We do not include the number of proposals as it is implicitly covered by the selected value of $K$.

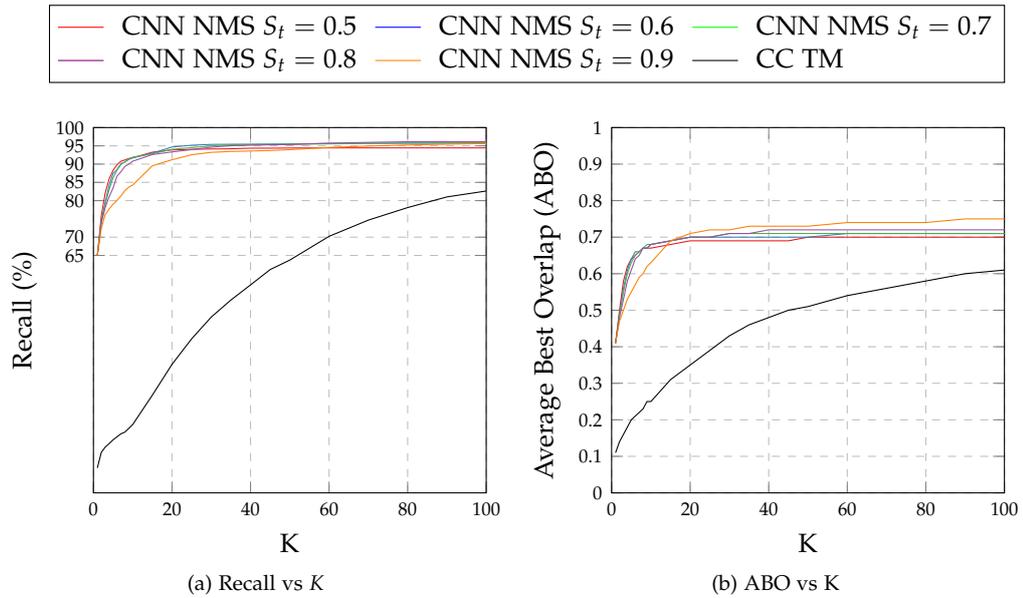
\begin{figure*}[t]
	\centering
	\begin{tikzpicture}
		\begin{customlegend}[legend columns = 3,legend style = {column sep=1ex}, legend cell align = left,
		legend entries={CNN NMS $S_t = 0.5$, CNN NMS $S_t = 0.6$, CNN NMS $S_t = 0.7$, CNN NMS $S_t = 0.8$, CNN NMS $S_t = 0.9$, CC TM}]
			\addlegendimage{mark=none,red}
			\addlegendimage{mark=none,blue}			
			\addlegendimage{mark=none,green}
			\addlegendimage{mark=none,violet}
			\addlegendimage{mark=none,orange}
			\addlegendimage{mark=none,black}
		\end{customlegend}
	\end{tikzpicture}
	\subfloat[Recall vs $K$]{
		\begin{tikzpicture}
		\begin{axis}[height = 0.25 \textheight, width = 0.45\textwidth, xlabel={K}, ylabel={Recall (\%)}, xmin=0, xmax=100, ymin = 0.0, ymax = 100.0, ytick = {65, 70, 80, 85, 90, 95, 100}, ymajorgrids=true, xmajorgrids=true, grid style=dashed, legend pos = south east, legend style={font=\scriptsize}, tick label style={font=\scriptsize}]
		
		\addplot+[mark = none] table[x  = k, y  = meanRecall, col sep = space] {chapters/data/proposals/topKVsRecallAtIoU0.50NMS0.50.csv};
		
		\addplot+[mark = none] table[x  = k, y  = meanRecall, col sep = space]
		{chapters/data/proposals/topKVsRecallAtIoU0.50NMS0.60.csv};
		
		\addplot+[mark = none] table[x  = k, y  = meanRecall, col sep = space]
		{chapters/data/proposals/topKVsRecallAtIoU0.50NMS0.70.csv};
		
		\addplot+[mark = none] table[x  = k, y  = meanRecall, col sep = space]
		{chapters/data/proposals/topKVsRecallAtIoU0.50NMS0.80.csv};
		
		\addplot+[mark = none] table[x  = k, y  = meanRecall, col sep = space]
		{chapters/data/proposals/topKVsRecallAtIoU0.50NMS0.90.csv};
		
		\addplot+[mark = none, black] table[x  = k, y  = meanRecall, col sep = space] {chapters/data/proposals/tm/tmProposals-TSPC100-topKVsRecallAtIoU0.50NMS0.00.csv};
		\end{axis}        
		\end{tikzpicture}
	}
	\subfloat[ABO vs K]{
		\begin{tikzpicture}
		\begin{axis}[height = 0.25 \textheight, width = 0.45\textwidth, xlabel={K}, ylabel={Average Best Overlap (ABO)}, xmin=0, xmax=100, ymin = 0, ymax = 1.0, ytick = {0.0, 0.1, 0.2, 0.3, 0.4, 0.5, 0.6, 0.7, 0.8, 0.9, 1.0}, ymajorgrids=true, xmajorgrids=true, grid style=dashed, legend pos = south east, legend style={font=\scriptsize}, tick label style={font=\scriptsize}]
		
		\addplot+[mark = none] table[x  = k, y  = aBO, col sep = space] {chapters/data/proposals/topKVsRecallAtIoU0.50NMS0.50.csv};
		
		\addplot+[mark = none] table[x  = k, y  = aBO, col sep = space]
		{chapters/data/proposals/topKVsRecallAtIoU0.50NMS0.60.csv};
		
		\addplot+[mark = none] table[x  = k, y  = aBO, col sep = space]
		{chapters/data/proposals/topKVsRecallAtIoU0.50NMS0.70.csv};
		
		\addplot+[mark = none] table[x  = k, y  = aBO, col sep = space]
		{chapters/data/proposals/topKVsRecallAtIoU0.50NMS0.80.csv};
		
		\addplot+[mark = none] table[x  = k, y  = aBO, col sep = space]
		{chapters/data/proposals/topKVsRecallAtIoU0.50NMS0.90.csv};
		
		\addplot+[mark = none, black] table[x  = k, y  = aBO, col sep = space] {chapters/data/proposals/tm/tmProposals-TSPC100-topKVsRecallAtIoU0.50NMS0.00.csv};
		\end{axis}        
		\end{tikzpicture}
	}
	\vspace*{0.5cm}
	\caption[ClassicNet with Objectness Ranking: Detection proposal results over our dataset]{ClassicNet with Objectness Ranking: Detection proposal results over our dataset, including recall and average best overlap (ABO).}
	\label{proposals:cnnTopKPlotResults}
	\vspace*{0.5cm}
\end{figure*}

Results using ClassicNet are shown in Figure \ref{proposals:cnnTopKPlotResults}. Objectness prediction with CNNs works quite well, achieving $95$ \% recall with only 20 proposals per image. In comparison, template matching objectness does not reach a similar recall even with more than 100 proposals per image. Recall increases with $K$ but it does in a much slower way with template matching objectness. This is also reflected in the ABO score, as the $0.5$ threshold is only achieved after $K = 40$.
This definitely shows that template matching objectness is not appropriate for detection proposals, as the template matching scores do not reflect true objectness.

\begin{figure*}[t]
	\centering
	\begin{tikzpicture}
	\begin{customlegend}[legend columns = 3,legend style = {column sep=1ex}, legend cell align = left,
	legend entries={FCN NMS $S_t = 0.5$, FCN NMS $S_t = 0.6$, FCN NMS $S_t = 0.7$, FCN NMS $S_t = 0.8$, FCN NMS $S_t = 0.9$, CC TM}]
	\addlegendimage{mark=none,red}
	\addlegendimage{mark=none,blue}			
	\addlegendimage{mark=none,green}
	\addlegendimage{mark=none,violet}
	\addlegendimage{mark=none,orange}
	\addlegendimage{mark=none,black}
	\end{customlegend}
	\end{tikzpicture}
	\subfloat[Recall vs $K$]{
		\begin{tikzpicture}
		\begin{axis}[height = 0.25 \textheight, width = 0.45\textwidth, xlabel={K}, ylabel={Recall (\%)}, xmin=0, xmax=100, ymin = 0.0, ymax = 100.0, ytick = {65, 70, 80, 85, 90, 95, 100}, ymajorgrids=true, xmajorgrids=true, grid style=dashed, legend pos = south east, legend style={font=\scriptsize}, tick label style={font=\scriptsize}]
		
		\addplot+[mark = none] table[x  = k, y  = meanRecall, col sep = space] {chapters/data/proposals/fcnProposals-topKVsRecallAtIoU0.50NMS0.50.csv};
		
		\addplot+[mark = none] table[x  = k, y  = meanRecall, col sep = space]
		{chapters/data/proposals/fcnProposals-topKVsRecallAtIoU0.50NMS0.60.csv};
		
		\addplot+[mark = none] table[x  = k, y  = meanRecall, col sep = space]
		{chapters/data/proposals/fcnProposals-topKVsRecallAtIoU0.50NMS0.70.csv};
		
		\addplot+[mark = none] table[x  = k, y  = meanRecall, col sep = space]
		{chapters/data/proposals/fcnProposals-topKVsRecallAtIoU0.50NMS0.80.csv};
		
		\addplot+[mark = none] table[x  = k, y  = meanRecall, col sep = space]
		{chapters/data/proposals/fcnProposals-topKVsRecallAtIoU0.50NMS0.90.csv};
		
		\addplot+[mark = none, black] table[x  = k, y  = meanRecall, col sep = space] {chapters/data/proposals/tm/tmProposals-TSPC100-topKVsRecallAtIoU0.50NMS0.00.csv};
		\end{axis}        
		\end{tikzpicture}
	}
	\subfloat[ABO vs K]{
		\begin{tikzpicture}
		\begin{axis}[height = 0.25 \textheight, width = 0.45\textwidth, xlabel={K}, ylabel={Average Best Overlap (ABO)}, xmin=0, xmax=100, ymin = 0, ymax = 1.0, ytick = {0.0, 0.1, 0.2, 0.3, 0.4, 0.5, 0.6, 0.7, 0.8, 0.9, 1.0}, ymajorgrids=true, xmajorgrids=true, grid style=dashed, legend pos = south east, legend style={font=\scriptsize}, tick label style={font=\scriptsize}]
		
		\addplot+[mark = none] table[x  = k, y  = aBO, col sep = space] {chapters/data/proposals/fcnProposals-topKVsRecallAtIoU0.50NMS0.50.csv};
		
		\addplot+[mark = none] table[x  = k, y  = aBO, col sep = space]
		{chapters/data/proposals/fcnProposals-topKVsRecallAtIoU0.50NMS0.60.csv};
		
		\addplot+[mark = none] table[x  = k, y  = aBO, col sep = space]
		{chapters/data/proposals/fcnProposals-topKVsRecallAtIoU0.50NMS0.70.csv};
		
		\addplot+[mark = none] table[x  = k, y  = aBO, col sep = space]
		{chapters/data/proposals/fcnProposals-topKVsRecallAtIoU0.50NMS0.80.csv};
		
		\addplot+[mark = none] table[x  = k, y  = aBO, col sep = space]
		{chapters/data/proposals/fcnProposals-topKVsRecallAtIoU0.50NMS0.90.csv};
		
		\addplot+[mark = none, black] table[x  = k, y  = aBO, col sep = space] {chapters/data/proposals/tm/tmProposals-TSPC100-topKVsRecallAtIoU0.50NMS0.00.csv};
		\end{axis}        
		\end{tikzpicture}
	}
	\vspace*{0.5cm}
	\caption[TinyNet-FCN with Objectness Ranking: Detection proposal results over our dataset]{TinyNet-FCN with Objectness Ranking: Detection proposal results over our dataset, including recall and average best overlap (ABO).}
	\label{proposals:fcnTopKPlotResults}
\end{figure*}
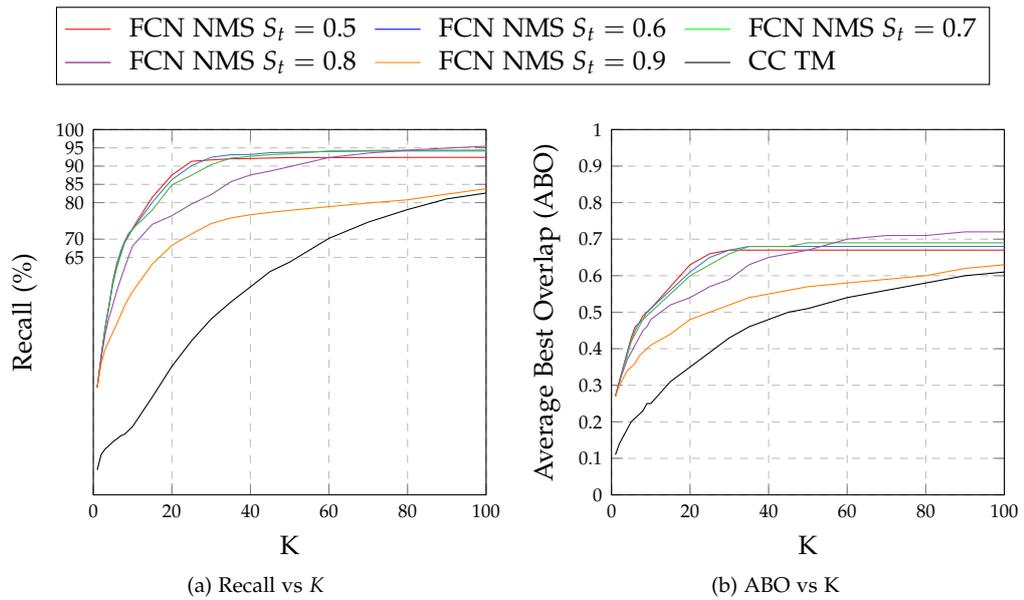

Results for TinyNet-FCN are shown in Figure \ref{proposals:fcnTopKPlotResults}. Objectness produced by this network also works quite well, but achieving $95$ \% recall with 40-60 proposals per image, which is 2-3 times the amount required by ClassicNet. This high recall is easily achieved by methods with aggressive non-maxima suppression ($S_t \in [0.5, 0.7]$), showing that the effect of NMS is also to remove high scoring proposals that do not contribute to high recall.
This effect can clearly be seen in the configuration featuring $S_t = 0.9$, which is almost equivalent to disabling NMS. In this case recall grows slower when compared to the other $S_t$ values, and saturates close to $85$ \%. Removing proposals through NMS has the unexpected effect of increasing recall.

Comparing ClassicNet and TinyNet-FCN with objectness ranking, it is clear that both networks produce high quality objectness scores, and only the methods used to decide proposals account for differences in recall. Using Objectness thresholding seems to be a simple option but does not produce the best results in terms of recall vs number of proposals. Objectness ranking seems to be a superior method as high recall can be achieved with low number of proposals (20 to 40 per image), but only if a low NMS threshold $S_t$ is applied.

\begin{figure*}[p]
	\centering	
	\subfloat[ClassicNet Objectness]{
        \begin{tabular}[b]{c}
		\includegraphics[height=0.22\textheight]{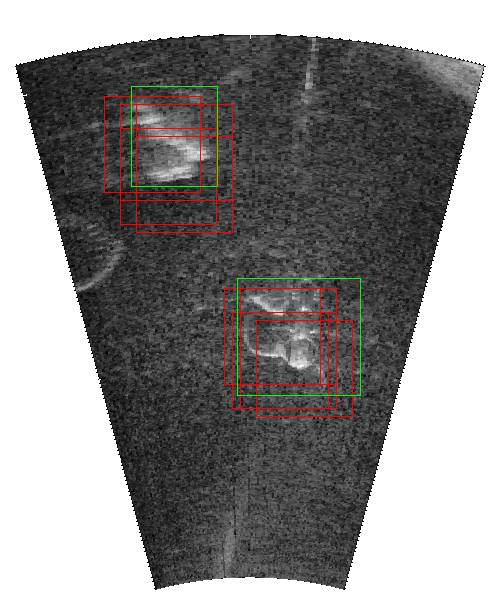}
		\includegraphics[height=0.22\textheight]{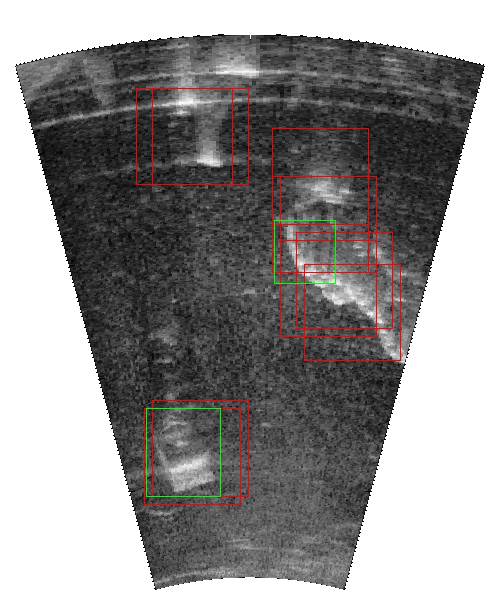}
		\includegraphics[height=0.22\textheight]{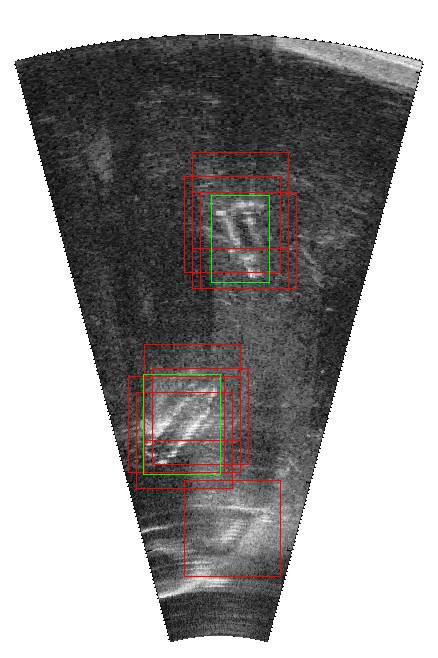}\\
		\includegraphics[height=0.22\textheight]{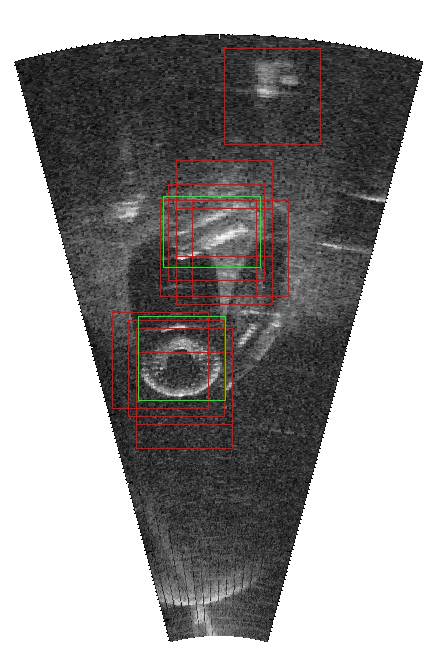}
		\includegraphics[height=0.22\textheight]{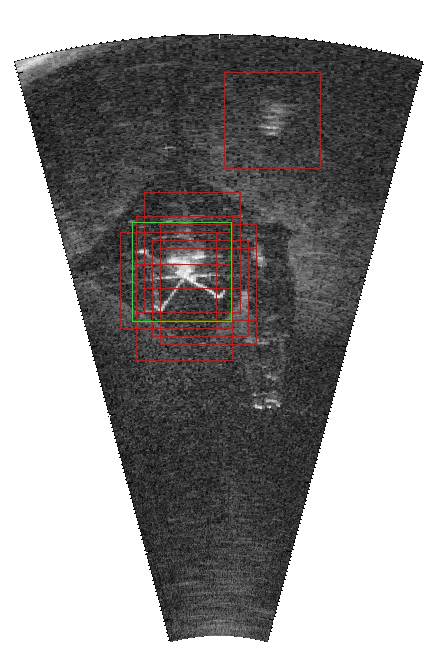}
        \end{tabular}
	}
	
	\subfloat[TinyNet-FCN Objectness]{
        \begin{tabular}[b]{c}
		\includegraphics[height=0.22\textheight]{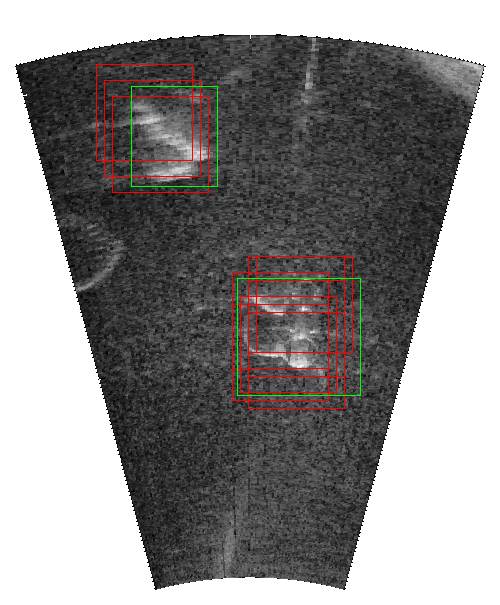}
		\includegraphics[height=0.22\textheight]{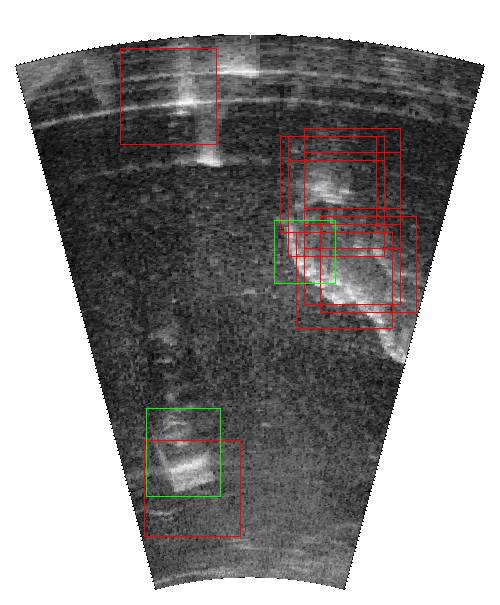}
		\includegraphics[height=0.22\textheight]{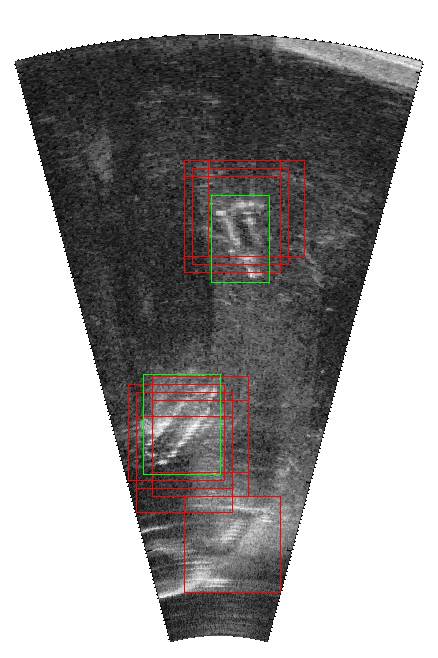}\\
		\includegraphics[height=0.22\textheight]{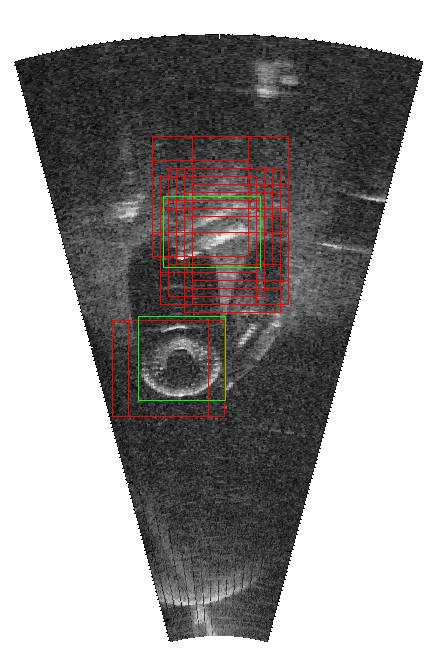}
		\includegraphics[height=0.22\textheight]{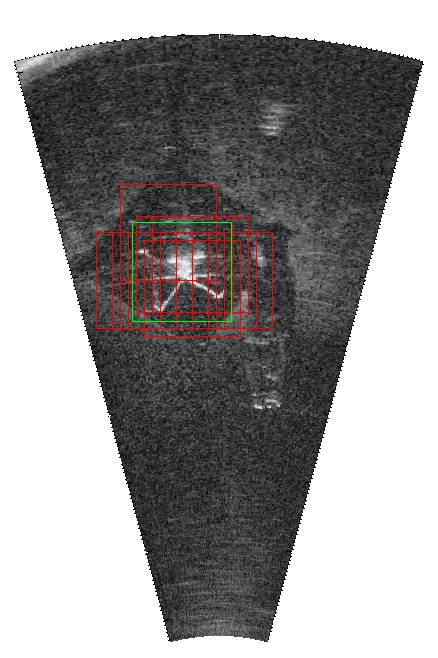}
        \end{tabular}
	}	
	\vspace*{-1.0cm}
	\caption{Sample detections using objectness ranking with $K = 10$ and NMS threshold $S_t = 0.7$}
	\label{proposals:rankingDetectionSamples}
\end{figure*}

Figure \ref{proposals:rankingDetectionSamples} shows a cherry-picked sample of detections over sonar images, computed with $K = 10$. This example shows that the top ranking proposals by objectness cover the objects of interest, including the ground truth objects, and some objects that are present in the dataset but not labeled as such, due to them being blurry and not easily identifiable by the human labeler.

\subsection{Comparison to State of the Art}

In this section we compare our proposed technique with the state of the art. We use three metrics for this, the best (highest) recall on our dataset, the number of proposals required to achieve such recall, and the computation time on the High-Power platform as defined in Chapter \ref{chapter:sonar-classification}. 

\begin{table}[t]
    \centering
    \begin{tabular}{llll}
        \hline 
        Method 						& Best Recall 		& \# of Proposals & Time (s)\\ 
        \hline
        TM CC Threshold				& $91.83$ \% 	& 150 & $10.0 \pm 0.5$\\
        TM CC Ranking				& $88.59$ \% 	& 110 & $10.0 \pm 0.5$\\		
        \hline 
        EdgeBoxes (Thresh)			& $57.01$ \%	& 300	& $0.1$\\ 
        EdgeBoxes (\# Boxes)		& $\mathbf{97.94}$ \%	& 5000	& $0.1$\\
        \hline
        Selective Search Fast		& $84.98$ \%	& 1000	& $1.5 \pm 0.1$\\
        Selective Search Quality	& $\mathbf{95.15}$ \%	& 2000	& $5.4 \pm 0.3$\\
        \hline						 		
        ClassicNet Threshold				& $\mathbf{96.42}$ \%	& \textbf{125}	& $12.4 \pm 2.0$\\
        TinyNet-FCN Threshold				& $\mathbf{96.33}$ \%	& 300	& $3.1 \pm 1.0$\\
        ClassicNet Ranking					& $\mathbf{96.12}$ \%	& \textbf{80}	& $12.4 \pm 2.0$\\
        Tinynet-FCN Ranking					& $\mathbf{95.43}$ \%	& \textbf{100}	& $3.1 \pm 1.0$\\
        \hline 
    \end{tabular}
    \caption[Comparison of detection proposal techniques with state of the art]{Comparison of detection proposal techniques with state of the art. Our proposed methods obtain the highest recall with the lowest number of proposals. Only EdgeBoxes has a higher recall with a considerably larger number of output proposals.}
    \label{proposals:sotaComparison}
\end{table}

Table \ref{proposals:sotaComparison} shows our main results. EdgeBoxes is by far the fastest method at $0.1$ seconds per frame, and it produces the best recall, but doing so requires over 5000 proposals per image, which is not really acceptable for real-world applications. We notice that EdgeBoxes on sonar images produces a large number of very small bounding boxes, which do not really correspond to objects in the sonar image. It also typically produces bounding boxes for parts of the objects, getting confused when acoustic highlights are disconnected.

Selective Search Quality also obtains very good recall but with over 2000 proposals needed for this. While this is lower than what is required by EdgeBoxes, it is still too much for practical purposes. We also notice the same pattern that too many boxes are assigned to just noise in the image. This can be expected as these algorithms are not really designed for sonar images.

Cross-Correlation Template Matching produces the lowest recall we observed on this experiment. Our objectness networks obtain very good recall with a low number of proposals per image. ClassicNet with objectness ranking produces $96$ \% recall with only 80 proposals per image, which is 62 times less than EdgeBoxes with only a $1$ \% absolute loss in recall. TinyNet-FCN with objectness ranking also produces $95 \%$ recall with only 100 proposals per image, at a four times reduced computational cost. Selective Search produces $1$ \% less recall than the best of our methods, but outputting 25 times more proposals.

In terms of computation time, EdgeBoxes is the fastest. FCN objectness is 4 times faster to compute than CNN objectness, due to the fully convolutional network structure, and it only requires a $1$ \% reduction in recall. CC Template Matching is also quite slow, at 10 seconds per image, making it difficult to use in an AUV.

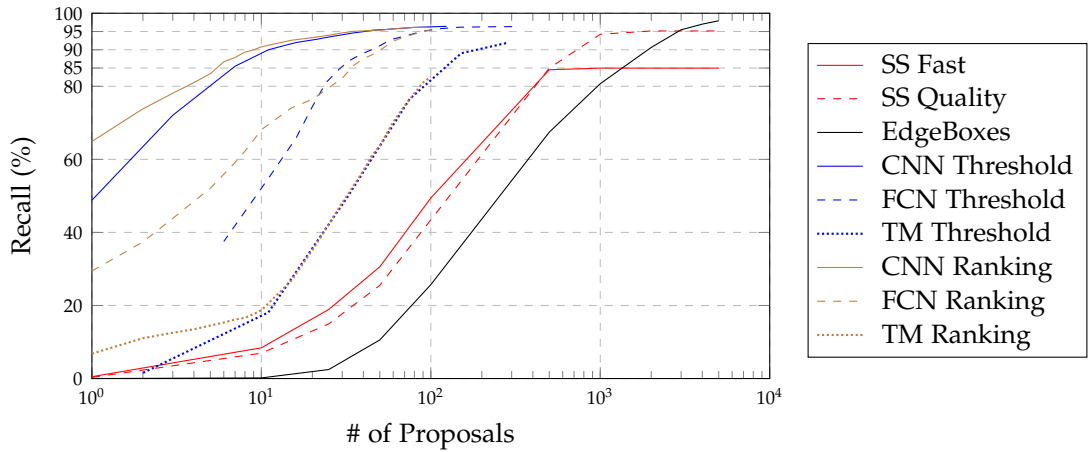
\begin{figure*}[t]
    \centering    
    \begin{minipage}[t]{0.70\textwidth}
    \begin{tikzpicture}
        \begin{axis}[height = 0.25 \textheight, width = 1.0\textwidth, xlabel={\# of Proposals}, ylabel={Recall (\%)}, xmin = 1.0, xmax = 10000, ymin = 0.0, ymax = 100.0, ytick = {0, 20, 40, 60, 80, 85, 90, 95, 100}, ymajorgrids=true, xmajorgrids=true, grid style=dashed, legend pos = north east, legend style={font=\scriptsize}, tick label style={font=\scriptsize}, xmode=log]
        \addplot+[red, solid, mark = none] table [x = k, y = meanRecall] {chapters/data/proposals/selectiveSearchFastTopKVsRecallAtIoU0.50.csv};
        
        \addplot+[red, dashed, mark = none] table [x = k, y = meanRecall] {chapters/data/proposals/selectiveSearchQualityTopKVsRecallAtIoU0.50.csv};
        
        \addplot+[black, solid, mark = none] table [x = k, y = meanRecall] {chapters/data/proposals/edgeBoxes-topKVsRecallAtIoU0.50.csv};
        
        \addplot+[blue, solid, mark = none] table[x  = meanNumProposals, y = meanRecall, col sep = space] {chapters/data/proposals/thresholdVsRecallNMS0.80.csv};
        
        \addplot+[blue, dashed, mark = none] table[x  = meanNumProposals, y = meanRecall, col sep = space] {chapters/data/proposals/fcnProposals-thresholdVsRecallNMS0.80.csv};
        
        \addplot+[blue, densely dotted, thick, mark = none] table[x  = meanNumProposals, y = meanRecall, col sep = space] {chapters/data/proposals/tm/tmProposals-TSPC100-thresholdVsRecallNMS0.00.csv};
        
        \addplot+[brown, solid, mark = none] table[x  = k, y = meanRecall, col sep = space] {chapters/data/proposals/topKVsRecallAtIoU0.50NMS0.80.csv};
        
        \addplot+[brown, dashed, mark = none] table[x  = k, y = meanRecall, col sep = space] {chapters/data/proposals/fcnProposals-topKVsRecallAtIoU0.50NMS0.80.csv};
        
        \addplot+[brown, densely dotted, thick, mark = none] table[x  = k, y = meanRecall, col sep = space] {chapters/data/proposals/tm/tmProposals-TSPC100-topKVsRecallAtIoU0.50NMS0.00.csv};
        \end{axis}		
    \end{tikzpicture}
    \end{minipage}
    \begin{minipage}[t]{0.29\textwidth}
    \vspace*{-5.5cm}
    \begin{tikzpicture}
        \begin{customlegend}[legend columns = 1,legend style = {column sep=1ex}, legend cell align = left,
        legend entries={SS Fast, SS Quality, EdgeBoxes, CNN Threshold, FCN Threshold, TM Threshold, CNN Ranking, FCN Ranking, TM Ranking}]
        \addlegendimage{red,solid}
        \addlegendimage{red,dashed}		
        \addlegendimage{black,solid}
        \addlegendimage{blue,solid}
        \addlegendimage{blue,dashed}
        \addlegendimage{blue,densely dotted,thick}
        \addlegendimage{brown,solid}
        \addlegendimage{brown,dashed}
        \addlegendimage{brown,densely dotted,thick}
        \end{customlegend}
    \end{tikzpicture}
    \end{minipage}
    \forceversofloat
    \vspace*{0.5cm}
    \caption[Number of output proposals versus Recall for different techniques]{Number of output proposals versus Recall for different techniques. State of the art detection proposals methods can achieve high recall but only outputting a large number of proposals. Our proposed methods achieve high recall with orders of magnitude less output proposals. For this plot we use $S_t = 0.8$.}
    \label{proposals:numProposalsVsRecall}
\end{figure*}

Figure \ref{proposals:numProposalsVsRecall} shows a comparison of the selected techniques as the number of output proposals is varied. This provides a more broad overview of how increasing the number of proposals that are output affects recall. The best methods would provide a high recall with a low number of proposals, corresponding to the top left part of the plot, and it can be seen that both ClassicNet and TinyNet-FCN objectness do a better job at predicting the correct objectness values at the right regions in the image, which leads to high recall with less proposals.

Overall we believe that our results show that our CNN-based methods are very competitive with the state of the art, as we can obtain better results than a Template Matching baseline and Selective Search in terms of higher recall and lower number of output proposals, and sightly worse than EdgeBoxes in terms of recall, but still a considerable improvement in the number of required output proposals.

\subsection{Objectness Visualization}

In this section we wish to evaluate a different aspect of our algorithm: how the produced objectness scores map to objects in the image. This corresponds to a qualitative evaluation, in contrast to previous quantitative evaluations.

For this experiment we compute the objectness map with ClassicNet and TinyNet-FCN. For ClassicNet we slide the network over the image, on a grid with a stride of $s = 4$ pixels. Gaps in the output map due to the strided evaluation are filled by performing nearest neighbour filtering. We do not slide the network on parts of the image that are partially or fully outside of the sonar's field of view. This produces a objectness map that is slightly smaller than the input image.
For TinyNet-FCN the complete image is input, and as mentioned previously, the output objectness map is up-scaled with linear interpolation to match the size of the input image.

\begin{figure*}[!htb]
	\centering
    \forcerectofloat
	\subfloat[Bottle and Tire]{
		\includegraphics[width=0.15\textwidth]{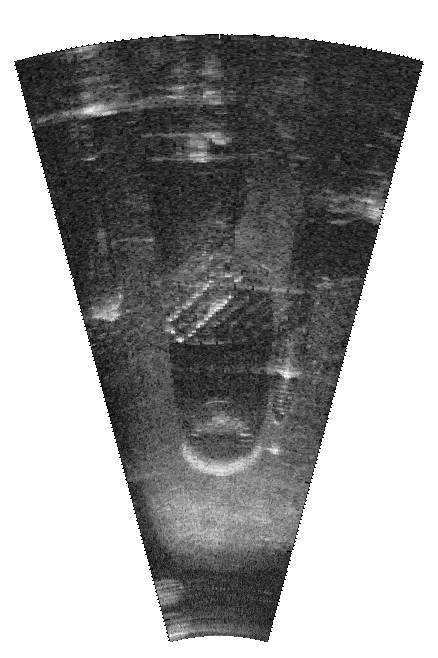}
		\includegraphics[width=0.15\textwidth]{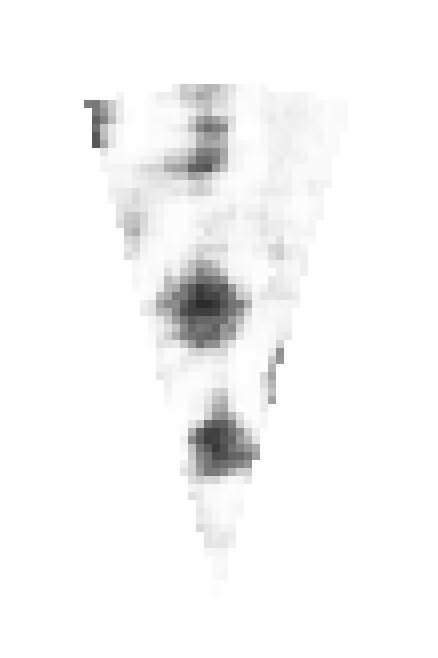}
		\includegraphics[width=0.15\textwidth]{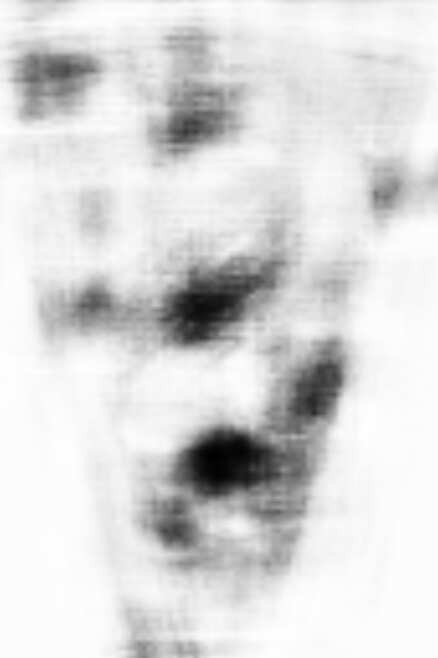}
	}
	\hspace*{0.5cm}
	\subfloat[Bottle \vspace*{0.35cm}]{
		\includegraphics[width=0.15\textwidth]{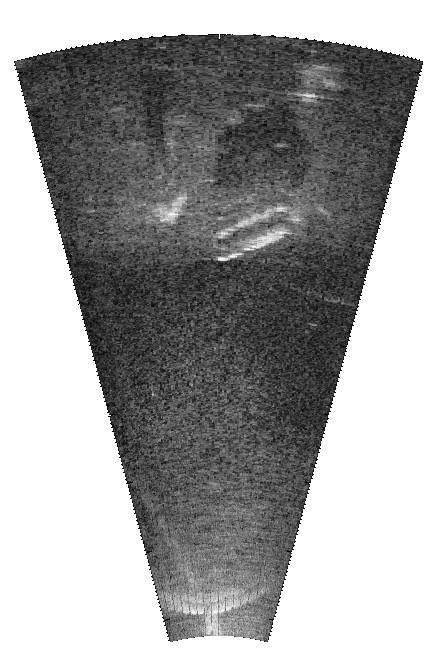}
		\includegraphics[width=0.15\textwidth]{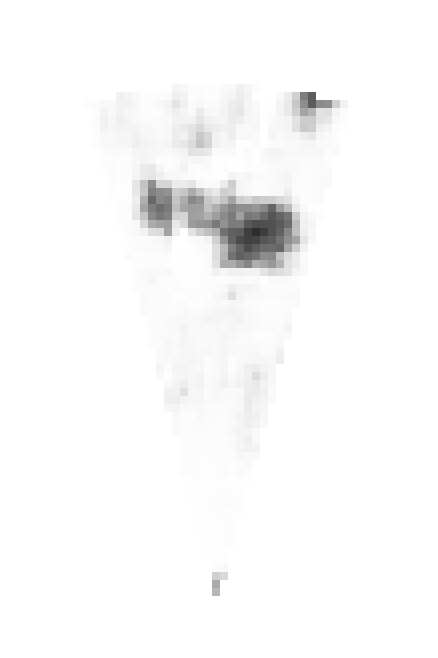}
		\includegraphics[width=0.15\textwidth]{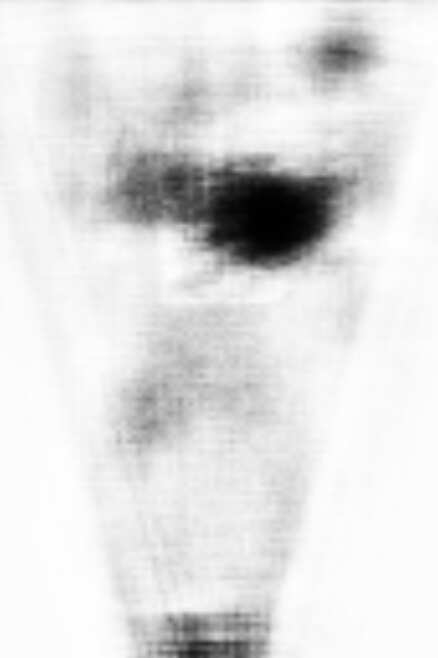}
	}
	
	\subfloat[Tire and Bottle]{
		\includegraphics[width=0.15\textwidth]{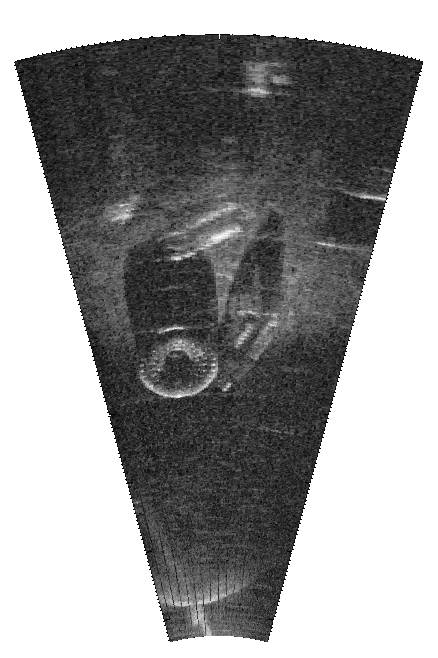}
		\includegraphics[width=0.15\textwidth]{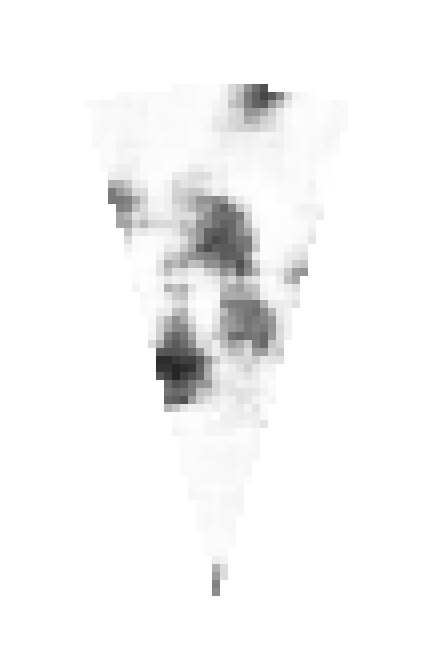}
		\includegraphics[width=0.15\textwidth]{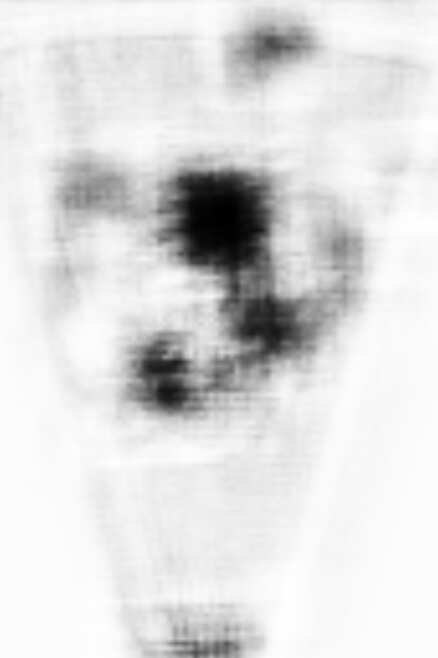}
	}
	\hspace*{0.5cm}
	\subfloat[Propeller and Wall]{
		\includegraphics[width=0.15\textwidth]{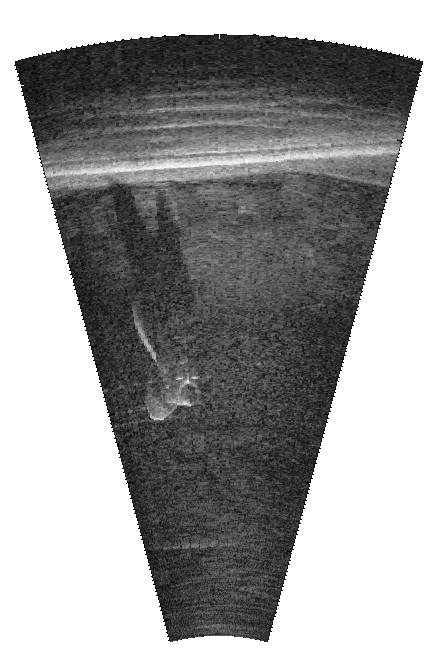}
		\includegraphics[width=0.15\textwidth]{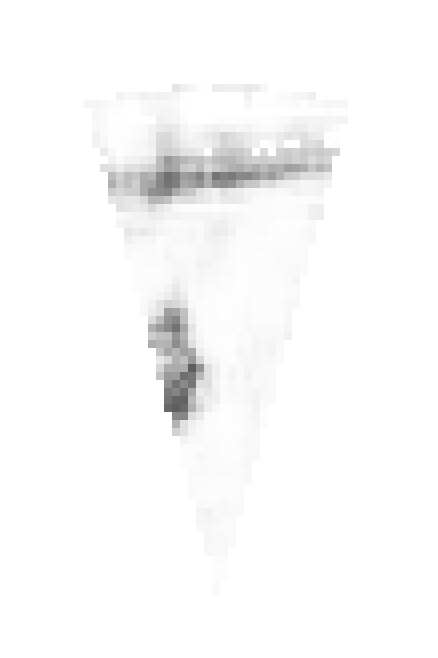}
		\includegraphics[width=0.15\textwidth]{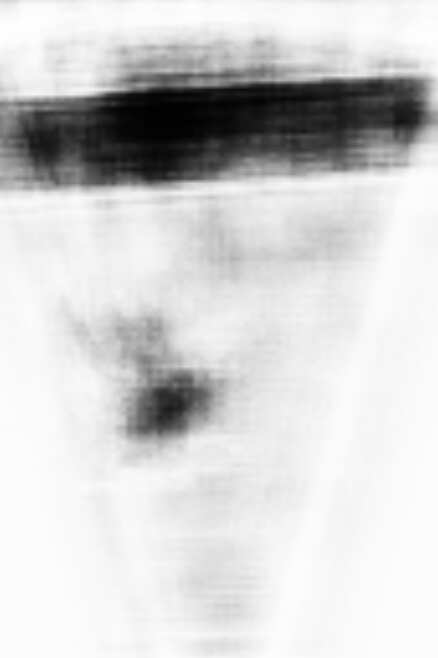}
	}

	\subfloat[Can \vspace*{0.35cm}]{
		\includegraphics[width=0.15\textwidth]{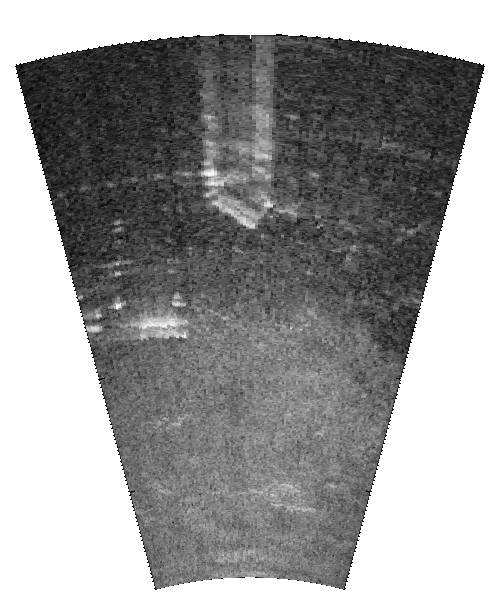}
		\includegraphics[width=0.15\textwidth]{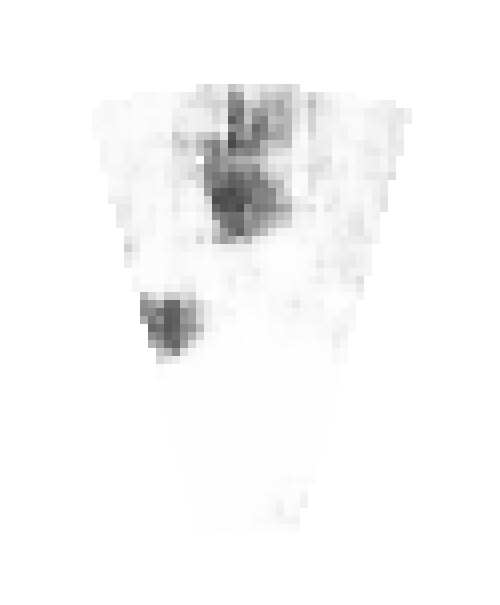}
		\includegraphics[width=0.15\textwidth]{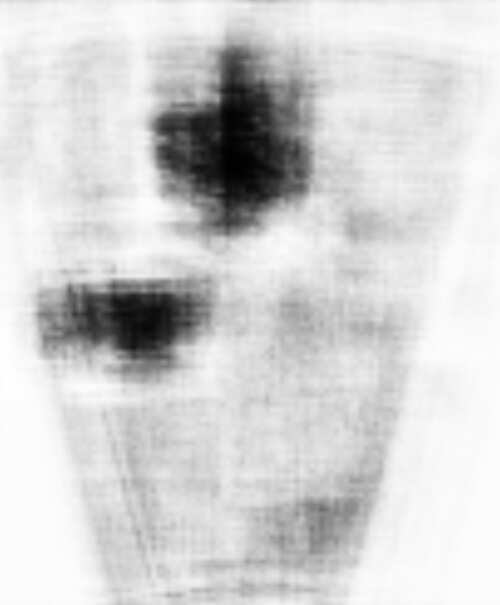}
	}
	\hspace*{0.5cm}
	\subfloat[Hook and Tire  \vspace*{0.35cm}]{
		\includegraphics[width=0.15\textwidth]{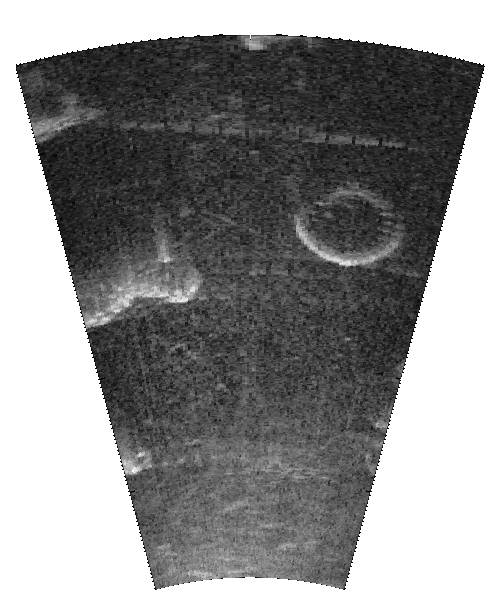}
		\includegraphics[width=0.15\textwidth]{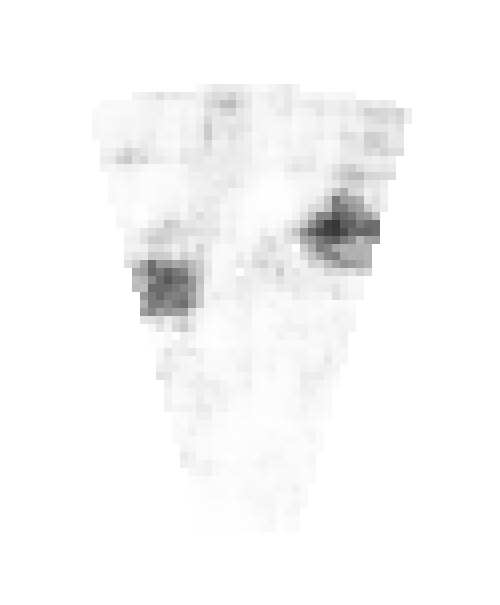}
		\includegraphics[width=0.15\textwidth]{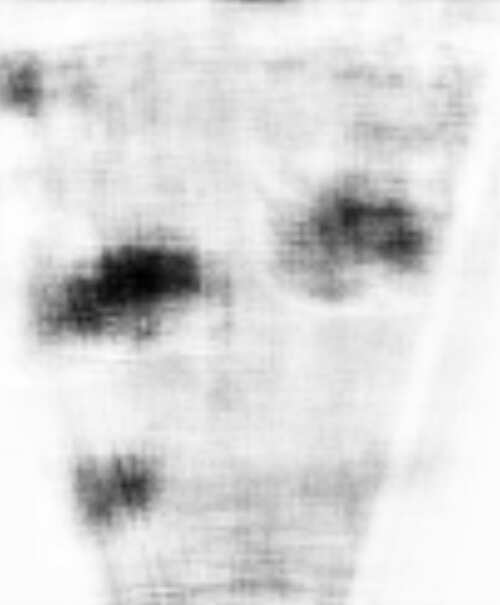}
	}
	\vspace*{0.5cm}
	\caption[Objectness Map Visualization]{Objectness Map Visualization. The first column corresponds to the input image, while the second is objectness produced by ClassicNet, and the third is objectness produced by TinyNet-FCN. In the last two columns light shade represents low objectness, while dark shades represent high objectness.}
	\label{proposals:objectnessVisualization}
\end{figure*}

We selected six representative sonar input images from the test set and computed their objectness map representations. These are shown in Figure \ref{proposals:objectnessVisualization}. Results show that there are large spatial correlations between the presence of an object and a large objectness score.

ClassicNet produces maps that have a sharp discrimination between object and background, with background having close to zero score, and objects having a higher score, but not reaching the maximum value of $1.0$ (full black). TinyNet does produce higher objectness scores in response to objects.

Some objects are of particular interest, such as walls present in Figure \ref{proposals:objectnessVisualization}d. The wall in those images was not part of the training set, but it produces a strong response on the TinyNet-FCN maps, and a weak response on the ClassicNet objectness maps. This indicates that the networks generalize well and are able to produce high objectness for objects that are quite different from the ones in the training set.

ClassicNet has a smaller mean squared error than TinyNet ($0.011$ vs $0.015$) and this difference produces small but non-zero scores for background in maps produced by TinyNet-FCN, while ClassicNet has almost zero scores for the same background. This is not problematic but it does indeed show a difference in the objectness scores produced by both networks. While we use the same metric (loss) to train both networks, it seems that TinyNet-FCN produces better scores that represent well a proper objectness metric.

The objectness maps that we obtained can be related to the concept of complexity maps, which shows the image complexity metric as a map on top of an image. Geilhufe and Midtgaard \cite{geilhufe2014quantifying} showed how a complexity map from a synthetic aperture sonar can be obtained from the variation of the wavelet representation coefficients of a sonar image. Their complexity maps visually look similar to our objectness maps, which suggests that it is possible that the objectness estimator algorithm could be estimating the image complexity indirectly. Additional experimentation is required to assess this hypothesis.

\FloatBarrier
\subsection{Generalization to Unseen Objects}

We have also evaluated our detection proposal techniques in previously unseen objects. For this purpose we used the chain dataset captured by CIRS at the University of Girona (Similar to \cite{hurtos2013automatic}), as well as several objects that we captured in our water tank, including a metal wrench, a large tire \footnote[][1em]{This tire is different from the one in the training set}, a rotating platform, and the wrench mounted on the rotating platform.
We do not have ground truth on these sonar images, so evaluation will only be done qualitatively.

\begin{figure*}[!t]
    \centering	
    \subfloat[Rotating Platform]{
        \includegraphics[height=0.2\textheight]{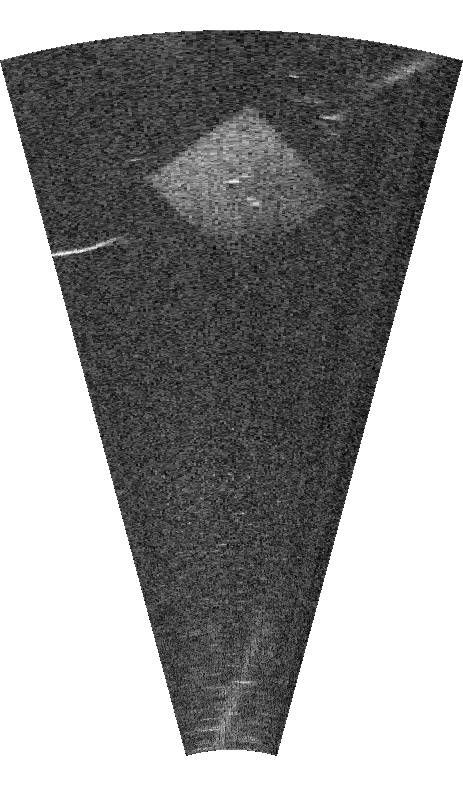}
        \includegraphics[height=0.2\textheight]{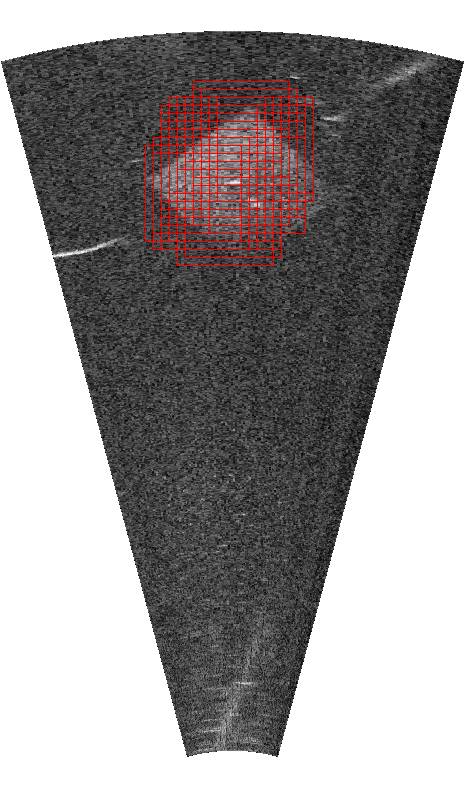}
        \includegraphics[height=0.2\textheight]{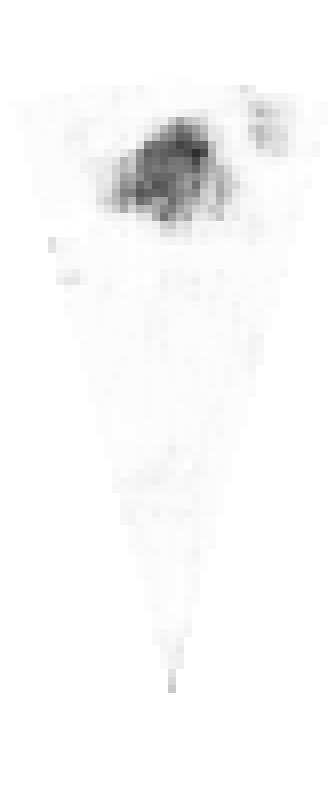}
        \includegraphics[height=0.2\textheight]{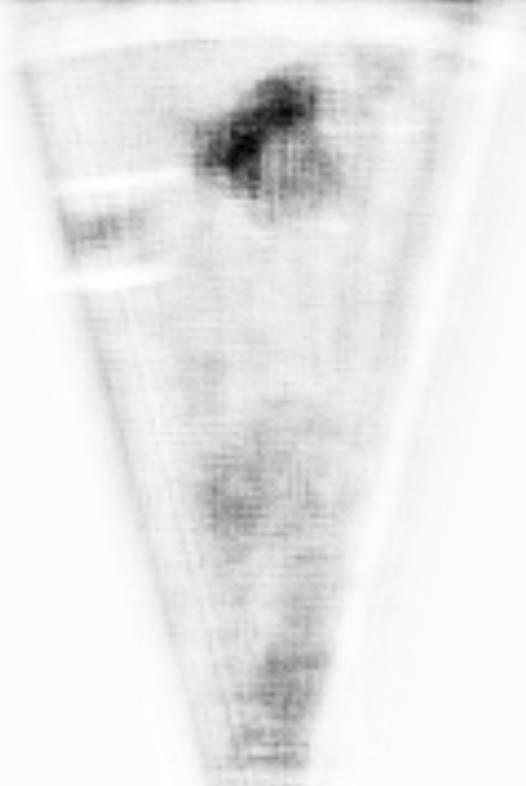}
    }
    
    \subfloat[Rotating Platform + Wrench]{
        \includegraphics[height=0.2\textheight]{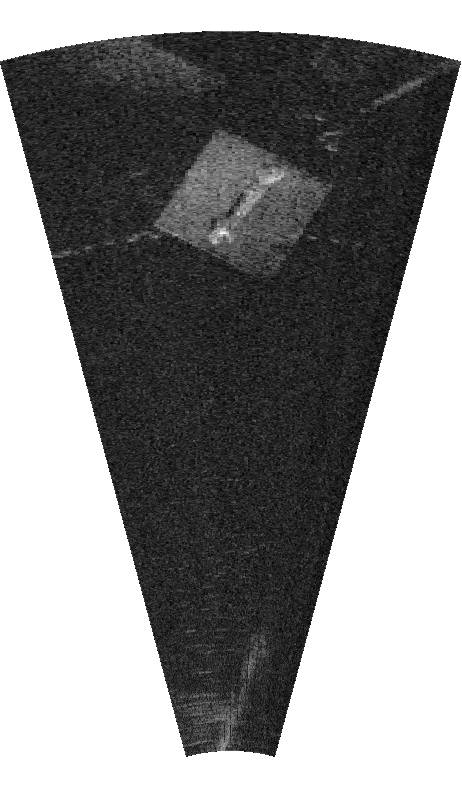}
        \includegraphics[height=0.2\textheight]{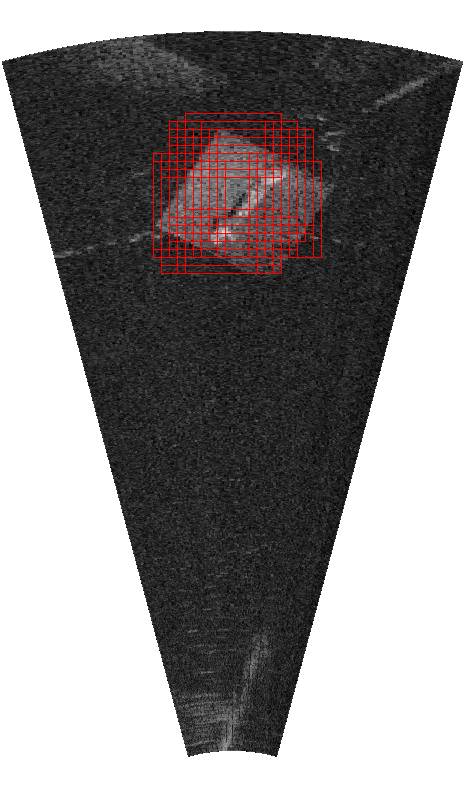}
        \includegraphics[height=0.2\textheight]{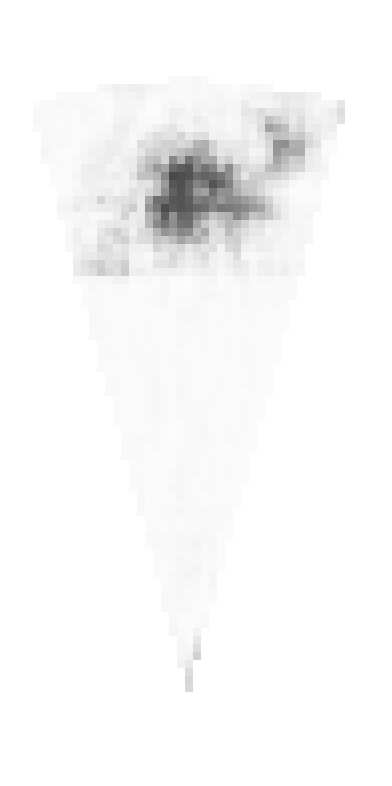}
        \includegraphics[height=0.2\textheight]{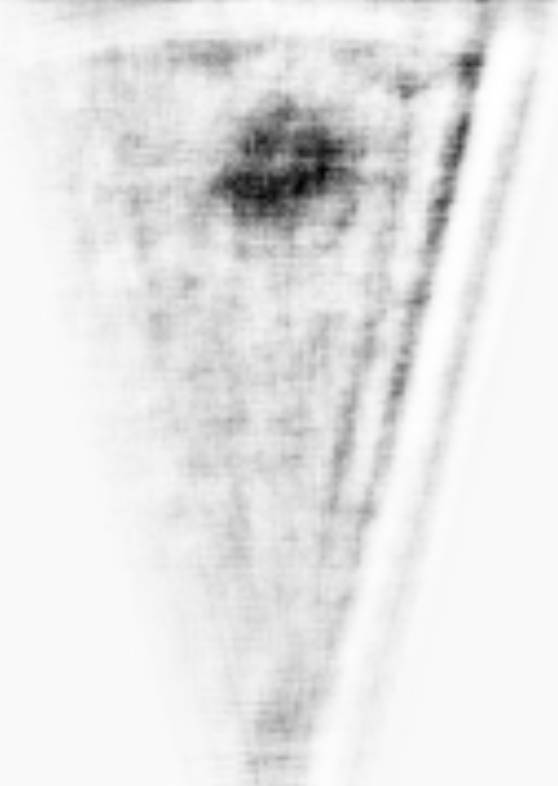}
    }
    \vspace*{0.5cm}
    \forcerectofloat
    \caption[Detection proposals on unseen objects]{Detection proposals on unseen objects. Both images show a rotating platform with and without an object on top. Columns in order show: Input image, detections at $T_o = 0.5$, objectness map produced by ClassicNet, and objectness map produced by TinyNet-FCN.}
    \label{proposals:unseenObjectsTwo}
\end{figure*}

\begin{figure*}[!t]
	\centering	
	\subfloat[Chain]{
		\includegraphics[height=0.17\textheight]{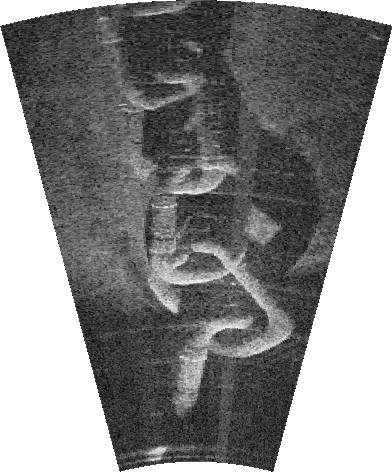}
		\includegraphics[height=0.17\textheight]{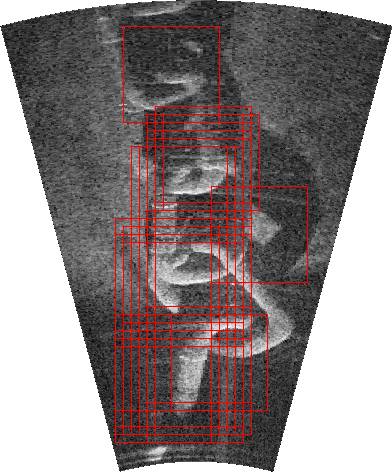}
		\includegraphics[height=0.17\textheight]{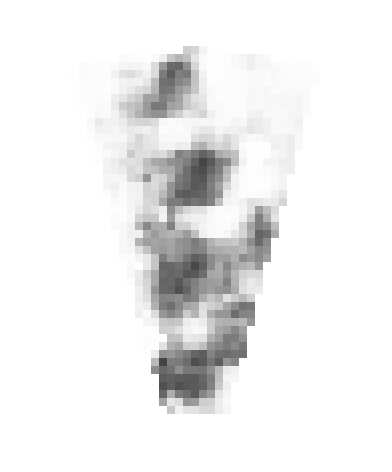}
		\includegraphics[height=0.17\textheight]{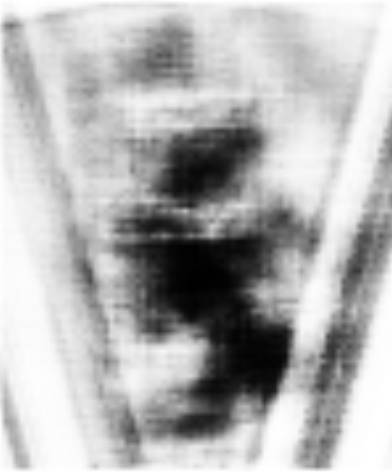}
	}

	\subfloat[Wrench]{
		\includegraphics[height=0.2\textheight]{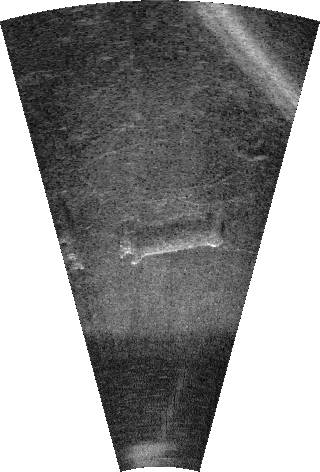}
		\includegraphics[height=0.2\textheight]{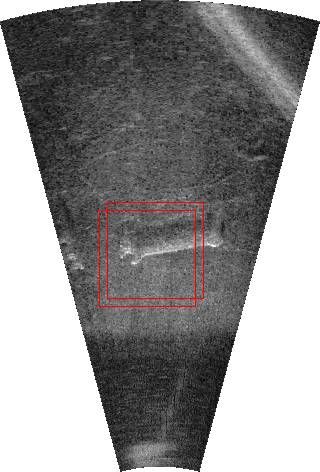}
		\includegraphics[height=0.2\textheight]{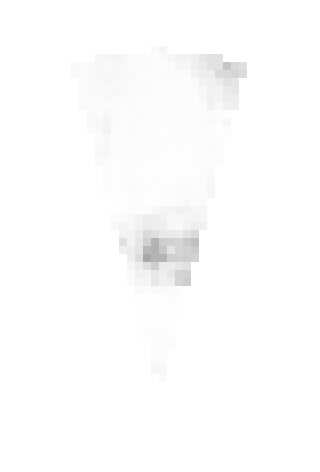}
		\includegraphics[height=0.2\textheight]{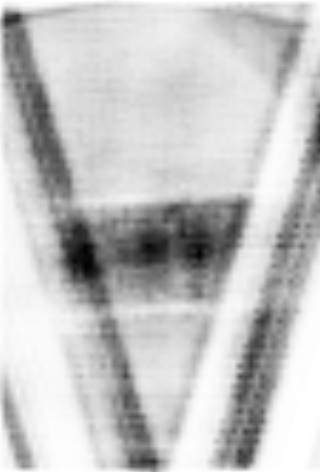}
	}

	\subfloat[Large Tire]{
		\includegraphics[height=0.2\textheight]{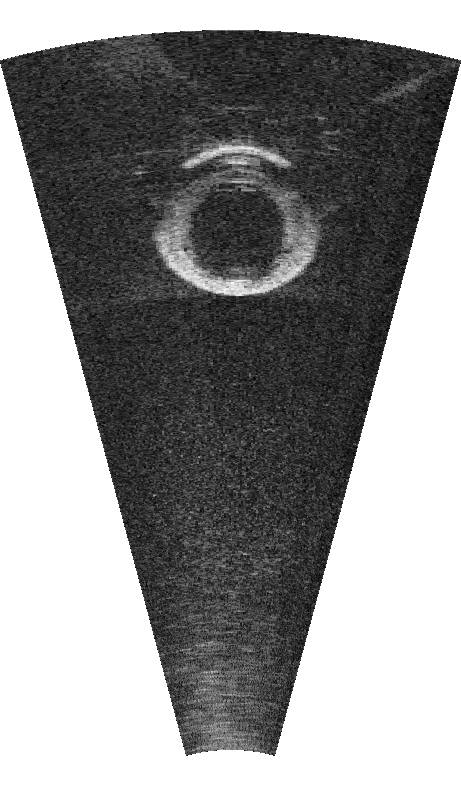}
		\includegraphics[height=0.2\textheight]{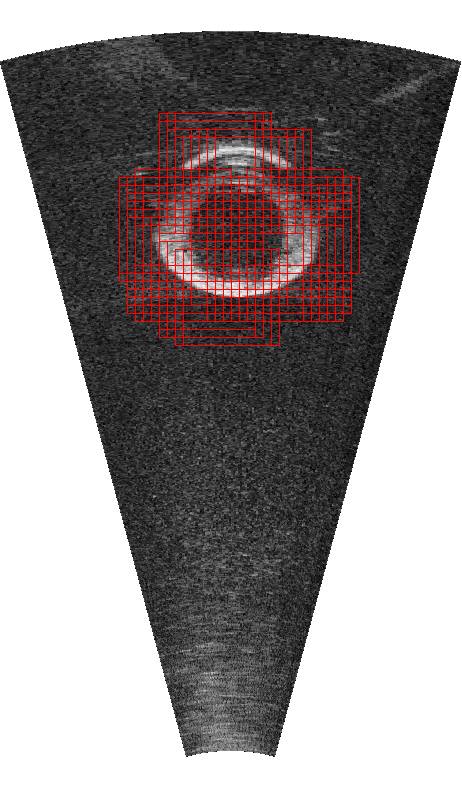}
		\includegraphics[height=0.2\textheight]{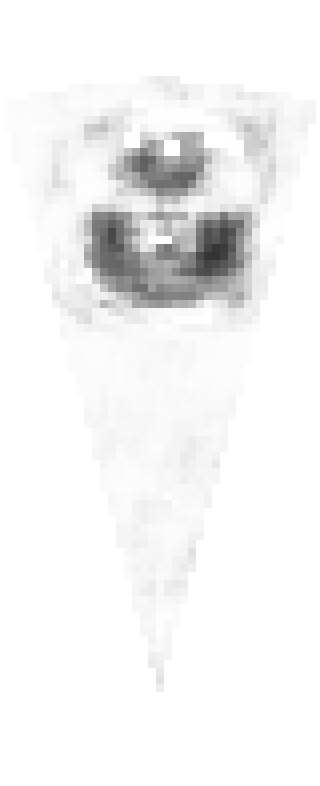}
		\includegraphics[height=0.2\textheight]{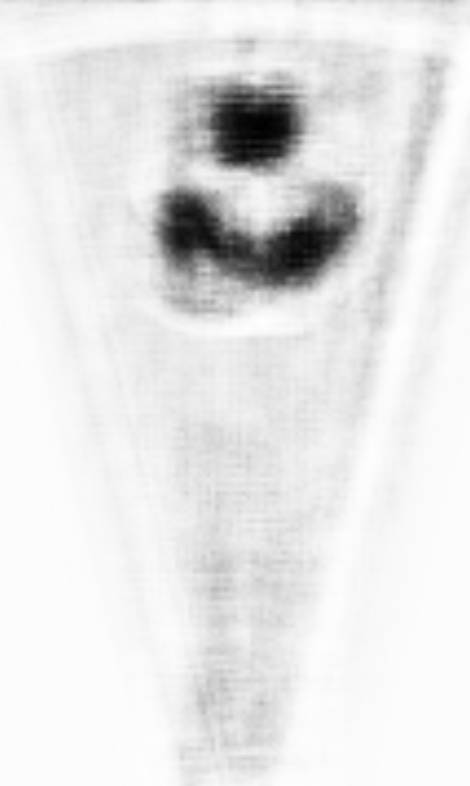}
	}
	\vspace*{0.5cm}
    \forceversofloat
	\caption[Detection proposals on unseen objects (cont)]{Detection proposals on unseen objects. (a) shows a chain, captured by the CIRS (University of Girona, Spain), while (b) shows a wrench, and (c) shows a large tire, both captured by us. In both cases detections can be generated over previously unseen objects, which shows the generalization capability of our proposed system. Columns in order show: Input image, detections at $T_o = 0.5$, objectness map produced by ClassicNet, and objectness map produced by TinyNet-FCN.}
	\label{proposals:unseenObjectsOne}
\end{figure*}

Figures \ref{proposals:unseenObjectsOne} and \ref{proposals:unseenObjectsTwo} contains our results. Fig. \ref{proposals:unseenObjectsOne}a shows detections generated over the chain. Detections completely cover the chain, and the objectness map also shows that there is a good correlation between chain presence and high scores.

Figure \ref{proposals:unseenObjectsOne}b shows detections over a wrench. This wrench is not present in our training set, and it is an object that can be considered radically different from the ones in our training set. This example also shows that unseen objects generally obtain high objectness scores, but ClassicNet produces lower scores than objects in the training set. TinyNet-FCN produces a strong response on unseen objects.

Figure \ref{proposals:unseenObjectsTwo} shows a rotating platform and the wrench placed on top of the rotating platform. These objects also considerably differ from the ones in the training set, and both produce high objectness scores. It should be noted that the scale of these objects is different from the training set, so no bounding box completely covers the objects.

These examples show that our system can generalize to objects that are quite different from the training set. These examples also show that background does not generate a large CNN response and in general objectness scores over background are quite close to zero.

\begin{figure*}[!p]
	\subfloat[ClassicNet Objectness]{
        \begin{tabular}[b]{c}
		\includegraphics[height=0.22\textheight]{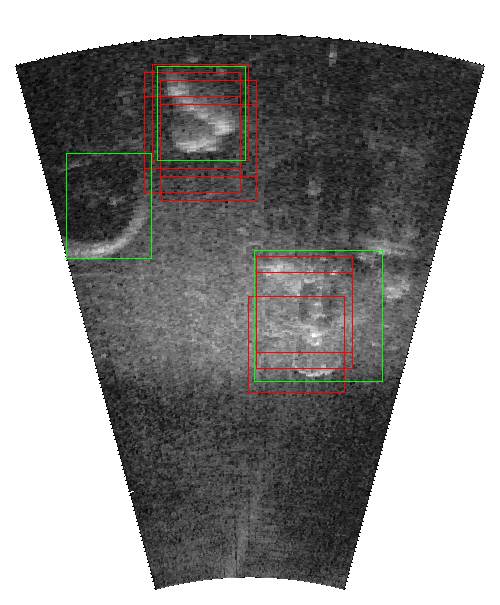}
		\includegraphics[height=0.22\textheight]{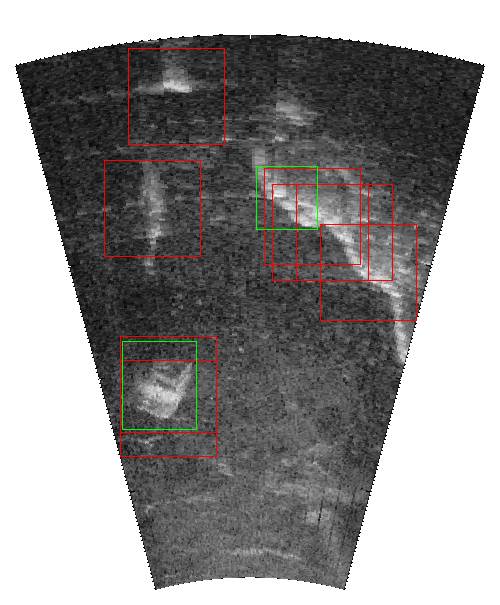}
		\includegraphics[height=0.22\textheight]{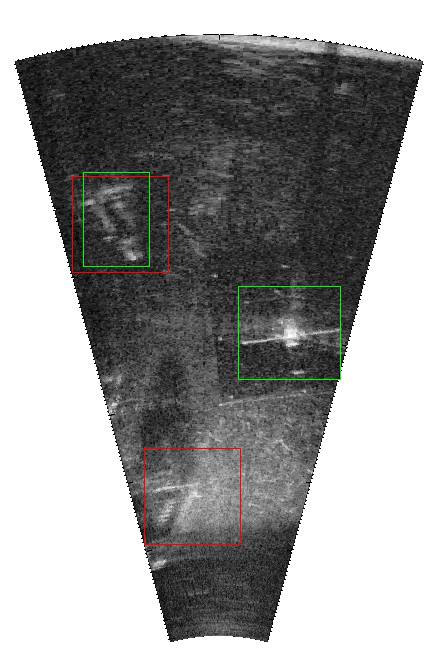}\\
		\includegraphics[height=0.22\textheight]{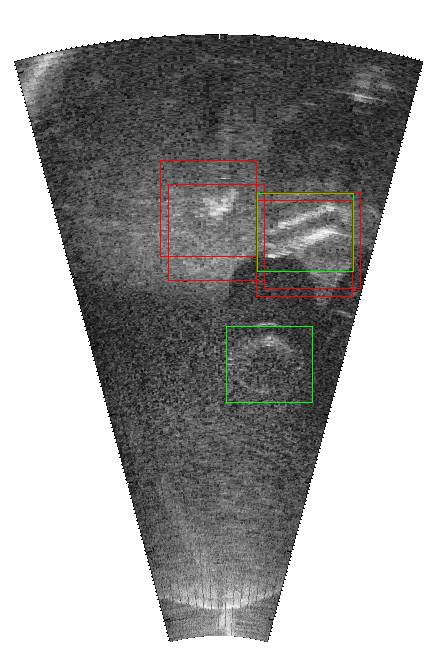}
		\includegraphics[height=0.22\textheight]{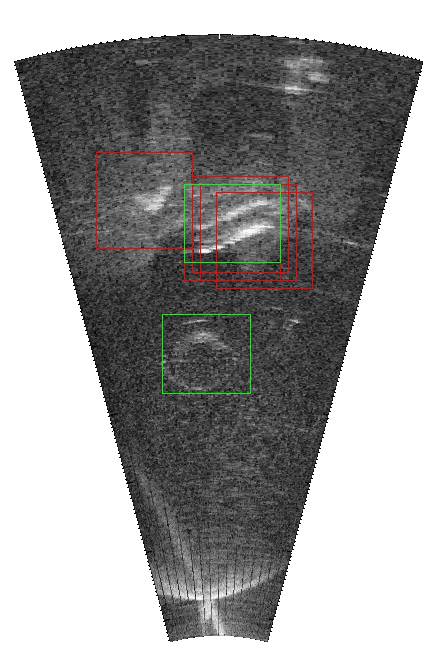}
        \end{tabular}        
	}
	
	\subfloat[TinyNet-FCN Objectness]{
        \begin{tabular}[b]{c}
		\includegraphics[height=0.2\textheight]{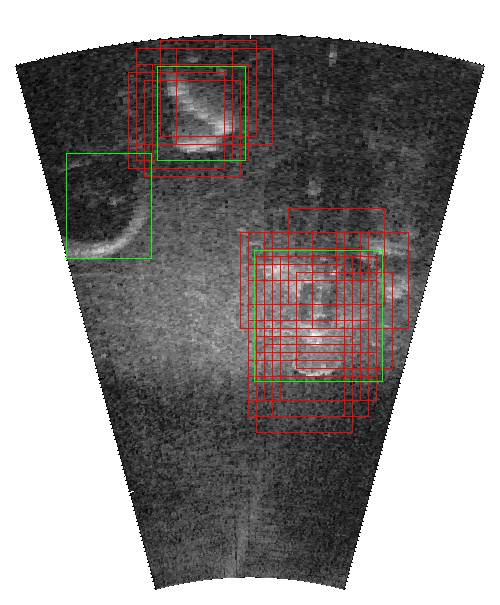}
		\includegraphics[height=0.2\textheight]{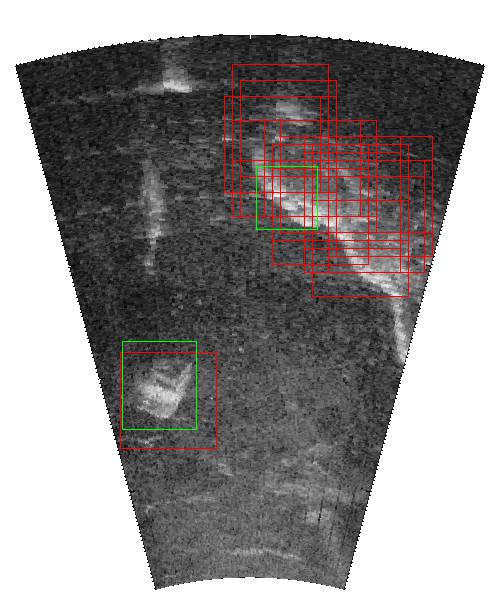}
		\includegraphics[height=0.2\textheight]{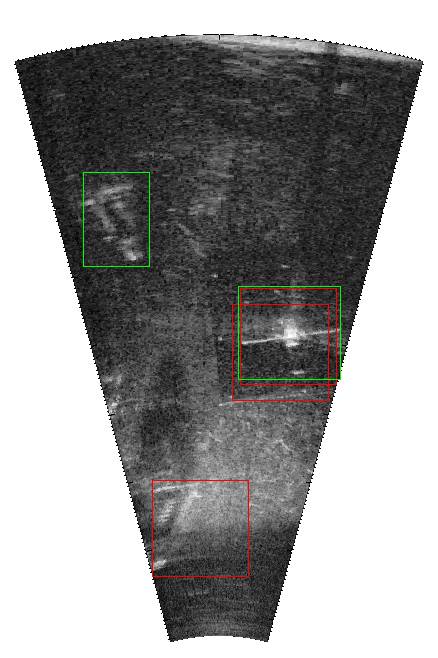}\\
		\includegraphics[height=0.2\textheight]{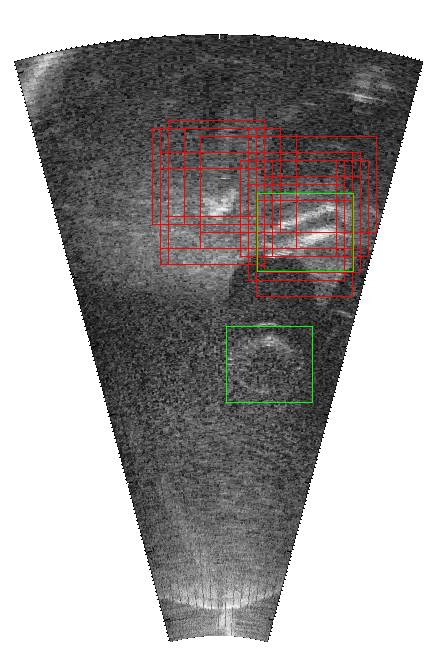}
		\includegraphics[height=0.2\textheight]{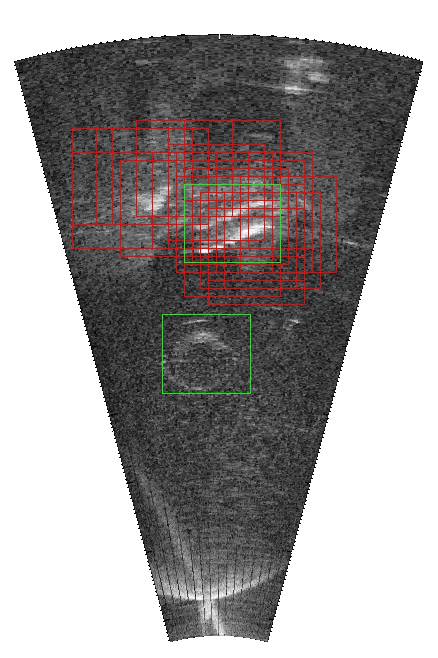}
        \end{tabular}
	}
	
	\vspace*{-1.0cm}
	\caption{Sample missed detections with objectness thresholding at $T_o = 0.6$ and NMS threshold $S_t = 0.7$}
	\label{proposals:errorSamples}
\end{figure*}

\subsection{Missed Detections Analysis}

In this section we provide a small analysis of failure cases in our detection proposals method. As we can reach $95$ \% recall, there is $5$ \% of ground truth objects that are missed and we wish to explain these missed detections.

Figure \ref{proposals:errorSamples} shows a selection of detections that completely missed an object or were considered incorrect. We have identified several kinds of errors that are common in our proposals pipeline:

\begin{itemize}
	\item \textbf{Poor Localization}. This corresponds to bounding boxes with IoU overlap lower than $0.5$, which makes them incorrect detections. This only produced for small bounding boxes, and this is expected as we are using a single scale to produce detections. In many cases the ground truth bounding boxes are too small and this produces a drop in recall. This is shown in the second column of Figure \ref{proposals:errorSamples}.
	\item \textbf{Low Objectness Score}. Some objects receive lower than expected objectness scores, only producing correct detections with very low thresholds, and missing them completely at higher thresholds. This is shown in the third column of Figure \ref{proposals:errorSamples}. This effect could be explained as simple overfitting or failure to generalize, and could improve with more training data.
	\item \textbf{Sonar Pose}. In some cases, like the fourth and fifth column of Figure \ref{proposals:errorSamples}, the sonar pose and field of view produce objects that have a very weak or considerable different highlight. This also produces a low objectness score and a missed detection. Objects like this are quite rare in the dataset but still present, so the objectness regressor could be biased against due to the low number of samples. More training data considering more views of the object and sonar poses could help improve this deficiency.
	\item \textbf{Disconnected Highlights}. In some cases, like the first and second columns of Figure \ref{proposals:errorSamples}, the highlights are very faint and disconnected at the pixel level which causes the objectness map to also be similarly disconnected. This either produces low objectness scores, as the detector seems to be inclined to highly score highlight regions that are connected, and thus a missed detection is produced. We also believe that more training data with increased variability can help solve this deficiency.
\end{itemize}

\subsection{Proposal Localization Quality}

In this section we evaluate the localization quality of the generated proposals. A high quality proposal should closely match the ground truth, while a low quality detection will not have a good match with the ground truth. Quality can be measured by the best match IoU score:

\begin{equation}
	\text{bestMatch}(A, GT) = \max_{R \in GT} \text{IoU}(A, R)
	\label{proposals:bestMatch}
\end{equation}

We evaluate proposal quality by varying the IoU overlap threshold $O_t$ used to decide if a detection is correct. We then compute recall as function of $O_t$. We use $O_t \in [0.5, 0.6, 0.7, 0.8, 0.9]$ and evaluate both objectness thresholding and ranking. We deactivate the use of non-maxima suppression in order to obtain an upper bound on recall as $O_t$ changes.

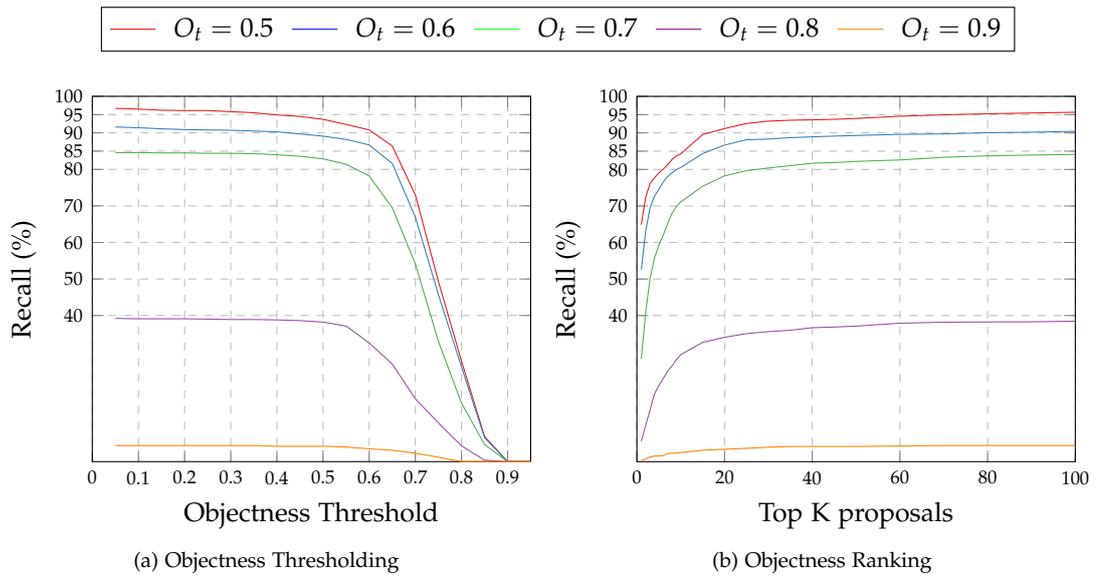
\begin{figure*}[t]
	\centering
    \forcerectofloat
	\begin{tikzpicture}
		\begin{customlegend}[legend columns = 5,legend style = {column sep=1ex}, legend cell align = left,
		legend entries={$O_t = 0.5$, $O_t = 0.6$, $O_t = 0.7$, $O_t = 0.8$, $O_t = 0.9$}]
			\addlegendimage{mark=none,red}
			\addlegendimage{mark=none,blue}			
			\addlegendimage{mark=none,green}
			\addlegendimage{mark=none,violet}
			\addlegendimage{mark=none,orange}
		\end{customlegend}
	\end{tikzpicture}
	\subfloat[Objectness Thresholding]{
		\begin{tikzpicture}
			\begin{axis}[height = 0.25 \textheight, width = 0.49 \textwidth, xlabel={Objectness Threshold}, ylabel={Recall  (\%)}, xmin=0, xmax = 0.95, ymin = 0.0, ymax = 100.0, xtick = {0.0, 0.1, 0.2, 0.3, 0.4, 0.5, 0.6, 0.7, 0.8, 0.9, 1.0}, ytick = {40, 50, 60, 70, 80, 85, 90, 95, 100}, ymajorgrids=true, xmajorgrids=true, grid style=dashed, legend pos = north east, legend style={font=\scriptsize}, tick label style={font=\scriptsize}]
			
				\addplot+[mark = none] table[x  = threshold, y  = meanRecall, col sep = space] {chapters/data/proposals/thresholdVsRecallAtIoU0.50NMS0.90.csv};				
				
				\addplot+[mark = none] table[x  = threshold, y  = meanRecall, col sep = space] {chapters/data/proposals/thresholdVsRecallAtIoU0.60NMS0.90.csv};				
				
				\addplot+[mark = none] table[x  = threshold, y  = meanRecall, col sep = space] {chapters/data/proposals/thresholdVsRecallAtIoU0.70NMS0.90.csv};				
				
				\addplot+[mark = none] table[x  = threshold, y  = meanRecall, col sep = space] {chapters/data/proposals/thresholdVsRecallAtIoU0.80NMS0.90.csv};				
				
				\addplot+[mark = none] table[x  = threshold, y  = meanRecall, col sep = space] {chapters/data/proposals/thresholdVsRecallAtIoU0.90NMS0.90.csv};				
			\end{axis}        
		\end{tikzpicture}
	}
	\subfloat[Objectness Ranking]{
		\begin{tikzpicture}
			\begin{axis}[height = 0.25 \textheight, width = 0.49 \textwidth, xlabel={Top K proposals}, ylabel={Recall (\%)}, xmin=0, xmax = 100, ymin = 0.0, ymax = 100, ytick = {40, 50, 60, 70, 80, 85, 90, 95, 100}, ymajorgrids=true, xmajorgrids=true, grid style=dashed, legend pos = south east, legend style={font=\scriptsize}, tick label style={font=\scriptsize}]        
			
				\addplot+[mark = none] table[x  = k, y  = meanRecall, col sep = space] {chapters/data/proposals/topKVsRecallAtIoU0.50NMS0.90.csv};				
				
				\addplot+[mark = none] table[x  = k, y  = meanRecall, col sep = space] {chapters/data/proposals/topKVsRecallAtIoU0.60NMS0.90.csv};				
				
				\addplot+[mark = none] table[x  = k, y  = meanRecall, col sep = space] {chapters/data/proposals/topKVsRecallAtIoU0.70NMS0.90.csv};				
				
				\addplot+[mark = none] table[x  = k, y  = meanRecall, col sep = space] {chapters/data/proposals/topKVsRecallAtIoU0.80NMS0.90.csv};				
				
				\addplot+[mark = none] table[x  = k, y  = meanRecall, col sep = space] {chapters/data/proposals/topKVsRecallAtIoU0.90NMS0.90.csv};				
			\end{axis}        
		\end{tikzpicture}
	}
	\vspace*{0.5cm}
	\caption[Proposal Quality for ClassicNet objectness]{Proposal Quality for ClassicNet objectness. This plot shows the relationship between Recall and Objectness Threshold $T_o$ and number of proposals $K$ for different values of $O_t$}
	\label{proposals:qualityClassicNet}
\end{figure*}

Our results for ClassicNet objectness are shown in Fig. \ref{proposals:qualityClassicNet}. Objectness thresholding recall slightly decreases when increasing $O_t = 0.7$, but it greatly decreases close to $40\%$ with $O_t = 0.8$. Using $O_t = 0.9$ produces negligible set of correct detections. These results indicate that our generated proposals match well until $O_t = 0.7$, but for higher IoU overlap threshold the match is not good.
In ClassicNet for objectness ranking we observe a similar pattern to objectness thresholding, as recall slightly decreases until $T_d = 0.7$, then greatly decreases for $O_t = 0.8$ and $O_t = 0.9$.

The kind of method used to extract detection proposals from objectness scores does not seem to have a big impact on localization quality. This is probably due to the use of a strided sliding window, as proposals can only be generated up to a precision of $s$ pixels (eight in the case of this thesis).

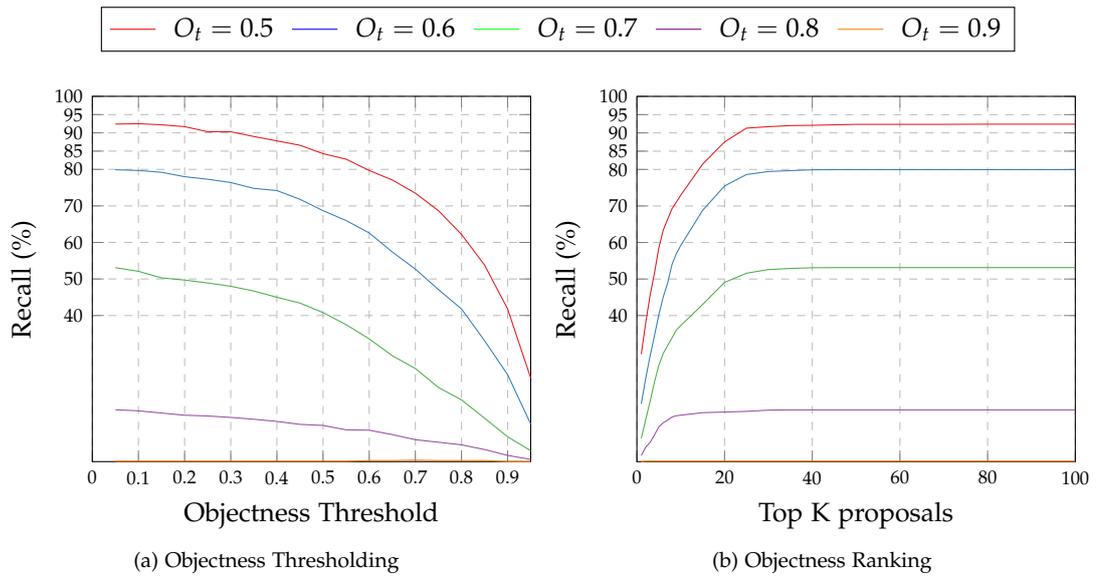
\begin{figure*}[t]
	\centering
    \forceversofloat
	\begin{tikzpicture}
	\begin{customlegend}[legend columns = 5,legend style = {column sep=1ex}, legend cell align = left,
	legend entries={$O_t = 0.5$, $O_t = 0.6$, $O_t = 0.7$, $O_t = 0.8$, $O_t = 0.9$}]
	\addlegendimage{mark=none,red}
	\addlegendimage{mark=none,blue}			
	\addlegendimage{mark=none,green}
	\addlegendimage{mark=none,violet}
	\addlegendimage{mark=none,orange}
	\end{customlegend}
	\end{tikzpicture}
	\subfloat[Objectness Thresholding]{
		\begin{tikzpicture}
		\begin{axis}[height = 0.25 \textheight, width = 0.49 \textwidth, xlabel={Objectness Threshold}, ylabel={Recall  (\%)}, xmin=0, xmax = 0.95, ymin = 0.0, ymax = 100.0, xtick = {0.0, 0.1, 0.2, 0.3, 0.4, 0.5, 0.6, 0.7, 0.8, 0.9, 1.0}, ytick = {40, 50, 60, 70, 80, 85, 90, 95, 100}, ymajorgrids=true, xmajorgrids=true, grid style=dashed, legend pos = north east, legend style={font=\scriptsize}, tick label style={font=\scriptsize}]
		
		\addplot+[mark = none] table[x  = threshold, y  = meanRecall, col sep = space] {chapters/data/proposals/fcnProposals-thresholdVsRecallAtIoU0.50NMS0.50.csv};				
		
		\addplot+[mark = none] table[x  = threshold, y  = meanRecall, col sep = space] {chapters/data/proposals/fcnProposals-thresholdVsRecallAtIoU0.60NMS0.50.csv};				
		
		\addplot+[mark = none] table[x  = threshold, y  = meanRecall, col sep = space] {chapters/data/proposals/fcnProposals-thresholdVsRecallAtIoU0.70NMS0.50.csv};				
		
		\addplot+[mark = none] table[x  = threshold, y  = meanRecall, col sep = space] {chapters/data/proposals/fcnProposals-thresholdVsRecallAtIoU0.80NMS0.50.csv};				
		
		\addplot+[mark = none] table[x  = threshold, y  = meanRecall, col sep = space] {chapters/data/proposals/fcnProposals-thresholdVsRecallAtIoU0.90NMS0.50.csv};				
		\end{axis}        
		\end{tikzpicture}
	}
	\subfloat[Objectness Ranking]{
		\begin{tikzpicture}
		\begin{axis}[height = 0.25 \textheight, width = 0.49 \textwidth, xlabel={Top K proposals}, ylabel={Recall (\%)}, xmin=0, xmax = 100, ymin = 0.0, ymax = 100, ytick = {40, 50, 60, 70, 80, 85, 90, 95, 100}, ymajorgrids=true, xmajorgrids=true, grid style=dashed, legend pos = south east, legend style={font=\scriptsize}, tick label style={font=\scriptsize}]        
		
		\addplot+[mark = none] table[x  = k, y  = meanRecall, col sep = space] {chapters/data/proposals/fcnProposals-topKVsRecallAtIoU0.50NMS0.50.csv};				
		
		\addplot+[mark = none] table[x  = k, y  = meanRecall, col sep = space] {chapters/data/proposals/fcnProposals-topKVsRecallAtIoU0.60NMS0.50.csv};				
		
		\addplot+[mark = none] table[x  = k, y  = meanRecall, col sep = space] {chapters/data/proposals/fcnProposals-topKVsRecallAtIoU0.70NMS0.50.csv};				
		
		\addplot+[mark = none] table[x  = k, y  = meanRecall, col sep = space] {chapters/data/proposals/fcnProposals-topKVsRecallAtIoU0.80NMS0.50.csv};				
		
		\addplot+[mark = none] table[x  = k, y  = meanRecall, col sep = space] {chapters/data/proposals/fcnProposals-topKVsRecallAtIoU0.90NMS0.50.csv};				
		\end{axis}        
		\end{tikzpicture}
	}
	\vspace*{0.5cm}
	\caption[Proposal Quality for TinyNet-FCN objectness]{Proposal Quality for TinyNet-FCN objectness. This plot shows the relationship between Recall and Objectness Threshold $T_o$ and number of proposals $K$ for different values of $O_t$}
	\label{proposals:qualityTinyNet}
\end{figure*}

Results for TinyNet-FCN objectness are shown in Figure \ref{proposals:qualityTinyNet}. A similar pattern to ClassicNet objectness can be observed, but the loss in recall by increasing $O_t$ is greater. For $O_t = 0.6$ recall drops from $95$ \% to $80$ \%, and further degrades close to $50 \%$ for $O_t = 0.7$. This shows that while the objectness scores produced by TinyNet-FCN are better distributed, they produce slightly worse proposal localization than ClassicNet. This is an area that definitely needs improvement.

\subsection{How Much Data is Needed to Generalize?}

In this final experiment we evaluate the effect of training set size on generalization. One big issue compared to similar experiments in previous chapters is that for computational reasons, we cannot run the full proposals algorithm on a test set for each trained network. This would simply take too much time.

As a simplification, we use the area under the ROC curve (AUC) as a proxy metric for detection proposals generalization. This makes sense as we wish to evaluate how well the scores produced by a network generalize to new samples, and how discriminative these scores perform to separate objects from background. 

In order to explore generalization performance, we vary the size of the training set by first generating synthetic class labels, as we framed objectness prediction as a regression and not a classification problem. Any ground truth image patch with objectness less than $0.5$ is labeled as a negative sample, while patches with objectness greater or equal to $0.5$ are labeled as positive samples. Then using this binary class information the dataset can be sub-sampled. We used SPC in range $[1,10000]$.
We evaluate each trained network on the validation set, as it contains image patches where MSE and AUC can be measured easily. We use the same basic methodology as previously mentioned in Section \ref{lim:secNumTrainingSamples} to train multiple networks and present mean and standard deviation of both metrics.

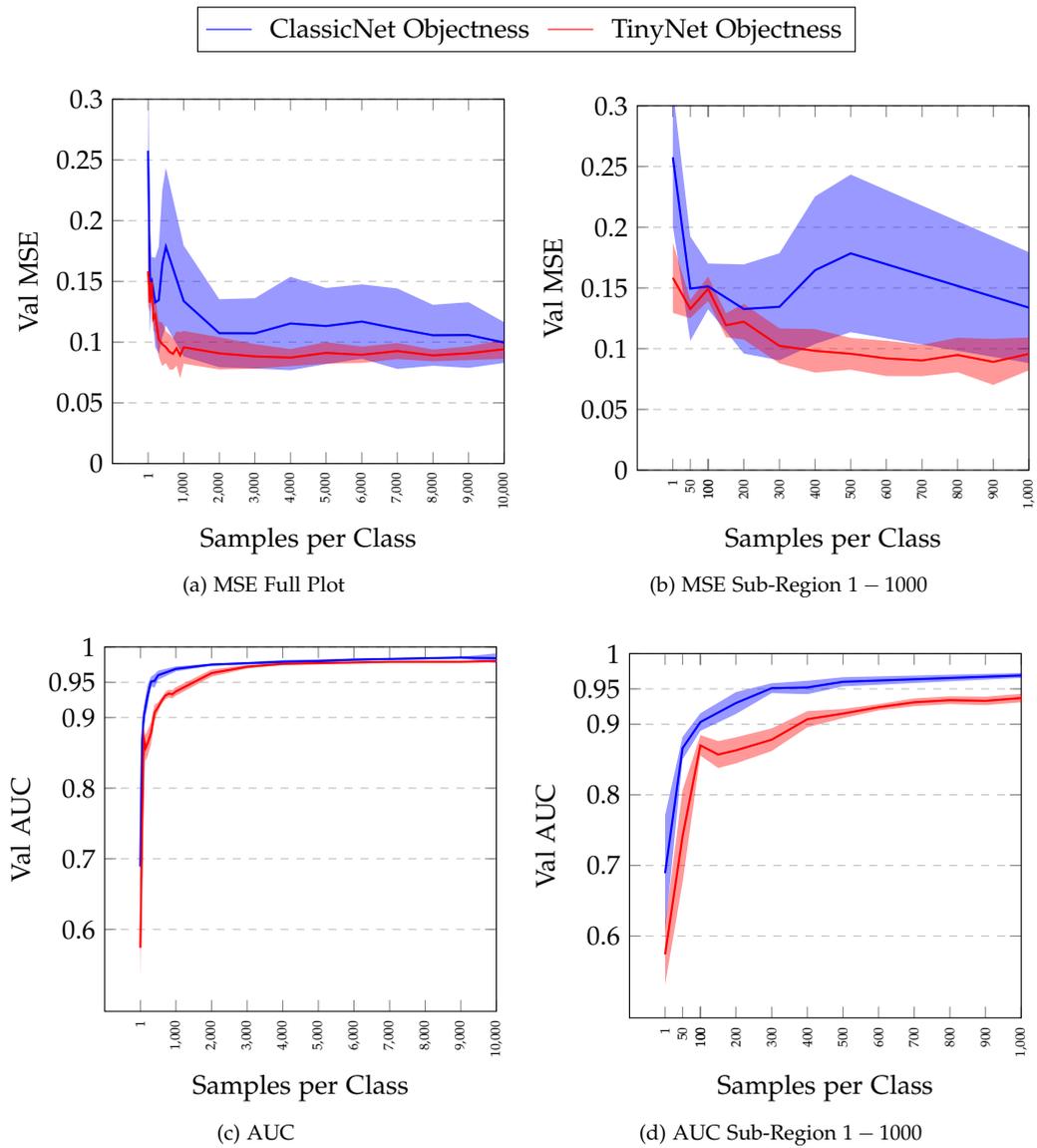
\begin{figure*}[!tb]
	\forcerectofloat
	\centering
	\begin{tikzpicture}
	\begin{customlegend}[legend columns = 4,legend style = {column sep=1ex}, legend cell align = left,
	legend entries={ClassicNet Objectness, TinyNet Objectness}]
	\addlegendimage{mark=none,blue}
	\addlegendimage{mark=none,red}
	
	\addlegendimage{mark=none,green}
	\end{customlegend}
	\end{tikzpicture}
	
	\subfloat[MSE Full Plot]{
		\begin{tikzpicture}
		\begin{axis}[
		xlabel={Samples per Class},
		ylabel={Val MSE},
		xmax = 10000,
		ymin=0.0, ymax = 0.3,
		xtick={1,1000,2000,3000,4000,5000,6000,7000,8000,9000,10000},
		ytick={0.0, 0.05, 0.10, 0.15, 0.20, 0.25, 0.30},
		x tick label style={font=\tiny, rotate=90},
		legend pos=south east,
		ymajorgrids=true,
		grid style=dashed,
		height = 0.25\textheight,
		width = 0.45\textwidth,
		scaled x ticks = false,
		y tick label style={/pgf/number format/fixed}]
		
		\errorband{chapters/data/proposals/proposals-CNNScore-AUCandMSEVsTrainSetSize.csv}{samplesPerClass}{meanMSE}{stdMSE}{blue}{0.4}
		\errorband{chapters/data/proposals/proposals-TinyNetScore-AUCandMSEVsTrainSetSize.csv}{samplesPerClass}{meanMSE}{stdMSE}{red}{0.4}
						
		\end{axis}
		\end{tikzpicture}
	}
	\subfloat[MSE Sub-Region $1-1000$]{
		\begin{tikzpicture}
		\begin{axis}[
		xlabel={Samples per Class},
		ylabel={Val MSE},
		xmax = 1000,
		ymin=0, ymax=0.3,
		xtick={1,50,100,100,200,300,400,500,600,700,800,900,1000},
		ytick={0.0, 0.05, 0.10, 0.15, 0.20, 0.25, 0.30},
		x tick label style={font=\tiny, rotate=90},
		legend pos=south east,
		ymajorgrids=true,
		grid style=dashed,
		height = 0.25\textheight,
		width = 0.45\textwidth,
		y tick label style={/pgf/number format/fixed}]
		
		\errorband{chapters/data/proposals/proposals-CNNScore-AUCandMSEVsTrainSetSize.csv}{samplesPerClass}{meanMSE}{stdMSE}{blue}{0.4}
		\errorband{chapters/data/proposals/proposals-TinyNetScore-AUCandMSEVsTrainSetSize.csv}{samplesPerClass}{meanMSE}{stdMSE}{red}{0.4}
		
		\end{axis}
		\end{tikzpicture}
	}

	\subfloat[AUC]{
		\begin{tikzpicture}
		\begin{axis}[
		xlabel={Samples per Class},
		ylabel={Val AUC},
		xmax = 10000,
		ymax = 1.0,
		xtick={1,1000,2000,3000,4000,5000,6000,7000,8000,9000,10000},
		ytick={0.6, 0.7, 0.8, 0.9, 0.95,1.0},
		x tick label style={font=\tiny, rotate=90},
		legend pos=south east,
		ymajorgrids=true,
		grid style=dashed,
		height = 0.25\textheight,
		width = 0.45\textwidth,
		scaled x ticks = false]
		
		\errorband{chapters/data/proposals/proposals-CNNScore-AUCandMSEVsTrainSetSize.csv}{samplesPerClass}{meanAUC}{stdAUC}{blue}{0.4}
		\errorband{chapters/data/proposals/proposals-TinyNetScore-AUCandMSEVsTrainSetSize.csv}{samplesPerClass}{meanAUC}{stdAUC}{red}{0.4}
		
		\end{axis}
		\end{tikzpicture}
	}
	\subfloat[AUC Sub-Region $1-1000$]{
		\begin{tikzpicture}
		\begin{axis}[
		xlabel={Samples per Class},
		ylabel={Val AUC},
		xmax = 1000,
		ymax = 1.0,
		xtick={1,50,100,100,200,300,400,500,600,700,800,900,1000},
		ytick={0.6, 0.7, 0.8, 0.9, 0.95,1.0},
		x tick label style={font=\tiny, rotate=90},
		legend pos=south east,
		ymajorgrids=true,
		grid style=dashed,
		height = 0.25\textheight,
		width = 0.45\textwidth]
		
		\errorband{chapters/data/proposals/proposals-CNNScore-AUCandMSEVsTrainSetSize.csv}{samplesPerClass}{meanAUC}{stdAUC}{blue}{0.4}
		\errorband{chapters/data/proposals/proposals-TinyNetScore-AUCandMSEVsTrainSetSize.csv}{samplesPerClass}{meanAUC}{stdAUC}{red}{0.4}
		
		\end{axis}
		\end{tikzpicture}
	}
	\vspace*{0.5cm}
	\caption[Training set size evaluation for our proposal networks]{Training set size evaluation for our proposal networks. Samples per Class versus MSE and AUC on the validation set, including error regions.}
	\label{proposals:spcVsAUC}
\end{figure*}

Our results are shown in Figure \ref{proposals:spcVsAUC}. Sub-sampling the datasets produces slightly higher MSE than the networks trained on the fully dataset, as expected. This shows that the experiment is working correctly and it might indicate that the fully training set might not be needed, as the difference in MSE is small.

Looking at Figure \ref{proposals:spcVsAUC}, subplots c and d show AUC as the number of samples per class is varied. With a single sample per class, ClassicNet obtains 0.7 AUC, while TinyNet obtains 0.6 AUC. Both are higher than the random chance limit of 0.5, but still it is not usable as a proposal scoring scheme.

Increasing the number of samples per class has the effect of rapidly increasing the AUC, with only $100$ samples per class required for ClassicNet to reach $0.9$ AUC, with TinyNet closely following at $0.87$ AUC. In order to reach $0.95$ AUC, ClassicNet requires $300$ samples, while TinyNet requires more than $1000$ samples. This is consistent with previous results that show that TinyNet requires more data to be trained properly, even as it has a considerably less number of parameters. As a general pattern, ClassicNet's AUC grows faster then TinyNet.

Analyzing large sample performance, both ClassicNet and TinyNet suffer from diminishing returns as bigger training sets are used. After approximately 3000 samples per class, both of them achieve high AUC, close to $0.97$, but adding more samples does not radically increase AUC, and the curve "flatlines" after this point. This indicates that the full training set might not be needed, and that the model can generalize well with considerably less samples.

We have also found out that measuring plain MSE is not always useful. This can be seen in Figure\ref{proposals:spcVsAUC}a-b as TinyNet has a lower MSE than ClassicNet, but ClassicNet has a higher AUC, specially at low sample sizes. This fact also indicates that the relationship between MSE and the quality of predicted scores is not simple. Only a full evaluation of each network's scores on the full testing set would be able to show this relationship.

\section{Limitations}

In this section we describe some theoretical limitations of our detection proposals approach.

\begin{itemize}
	\item \textbf{Scale invariance}. Our method uses a single scale. This makes sense as a sonar sensor does not have the perspective effect of typical cameras. This means that objects have the same size independent of the distance to the sensor. The only difference with distance is sampling due to the polar field of view. But still our method has problems producing accurate bounding boxes for small objects. Ideally a detection proposal algorithm should be scale invariant and produce variable-sized bounding boxes that fit objects tightly.
	\item \textbf{Computational Performance}. Our method is quite slow. ClassicNet takes more than 13 seconds per image to produce an objectness map and proposals, while TinyNet-FCN is considerably faster at almost 3 seconds per image. While these times are computed on CPU, and we did not evaluate on GPUs, it is still slower than required for real-time and robotics applications. We believe performance can be improved with different techniques, specially exploring a TinyNet architecture with less parameters that produces similar performance. As a quick solution, a GPU can be used where real-time performance is possible, but this is not a good solution for an AUV.
	\item \textbf{Training Data}. Our training set is quite small, only at 50K images. The variability inside the dataset is limited, and more data, including additional views of the object, different sonar poses, and more variation in objects, will definitely help train an objectness regressor with better generalization. 
	\item \textbf{Objectness Peak}. Our method is based on a principle similar to the Hough transform \cite{gonzalezDIP2006}, where we assume that a peak in the objectness map over a window will match with objects in the image. This principle holds in general but in many cases the peak does not produce a precise location for an object. This is problematic and could be improved by post-processing techniques on the objectness map, or by improving the training set and the quality of objectness labels.
	\item \textbf{Bounding Box Regression}. Our method uses a single scale with a sliding window approach. A considerable  better technique is to perform bounding box regression, where the network predicts normalized coordinates of a bounding box. Our experiments trying to make this technique work have failed, probably due to the small size of our dataset and the lack of a big dataset for feature learning and/or pre-training. We leave the implementation of a bounding box regressor as future work.
	\item \textbf{Number of Proposals}. A final issue is that the number of proposals required to reach a high recall is still quite high, specially when considering that our images do not contain more than 5 objects per image. A high quality detection proposals algorithm will produce the minimum amount of proposals for an image, with high localization quality. We believe that as future work our method could be improved to produce better localized proposals in the right quantities.
	\item \textbf{Dataset Bias}. We captured the dataset used to train our detection proposals method in a artificial water tank. This makes the data biased as the background is not the same as in a real environment. We believe that our method can still work in a real environment, but first it must be trained on real data containing seafloor background and objects of interest.
\end{itemize}

\section{Summary of Results}

In this chapter we have presented a method to produce detection proposals in forward-looking sonar images. Based on the concept of objectness prediction with a neural network, we propose two techniques to extract proposals from an objectness map: thresholding and ranking.

We use two neural network architectures, one based on ClassicNet and another based on TinyNet. The latter network has no fully connected layers and can be evaluated in as a fully convolutional network (FCN), improving performance considerably.

In our dataset of marine debris, we show that both methods can reach $95$ \% recall with less than 60 proposals per image. We consider that this is a considerably high number of proposals, but this can be reduced by applying non-maxima suppression. TinyNet-FCN produced better distributed objectness but it is slightly less accurate in terms of recall. Both methods produce proposals that are well localized up to a IoU overlap threshold $O_t = 0.7$, but at this overlap threshold ClassicNet is superior.

We constructed a baseline that uses cross-correlation template matching, which obtains only up to $90$ \% recall, but producing very low objectness scores, which translates as requiring up to 300 proposals per image. We show that template matching does not produce a good objectness measure and our system is superior than this baseline.

We have performed a comprehensive evaluation of our method, showing that it generalizes quite well to unseen objects, even when applied to completely new data. This shows the applicability of our method to a variety of objects and environments, but still generalization to real underwater environments will require re-training on more realistic data.

We also evaluated the effect of training set size on the scores predicted by our neural networks. We found out that a smaller training set would also work well, from 300 to 1000 samples per class. More data does help but performance, as measured as the area under the ROC curve, saturates after approximately 3000 samples per class.

We also have documented the limitations of our techniques, including that it is not scale invariant, it can be computationally prohibitive for real-time applications, it was trained on a relatively small dataset, and it requires a high number of proposals to produce high recall. We believe that these limitations can be addressed in future work in this topic.

We expect that detection proposals will become popular in the sonar image processing and underwater robotics communities.

%% file: chapters/applications.tex
\chapter[Selected Applications Of Detection Proposals]{Selected Applications Of \newline Detection Proposals}
\label{chapter:applications}

In the final chapter of this thesis, we present two applications of our detection proposals method. The first is a end-to-end object detection pipeline that uses detection proposals at its core. The second is a simple tracking system that combines detection proposals and matching.

The purpose of this chapter is not to improve some specific state of the art, but to show that detection proposals can be used for different problems related to underwater robot perception, in the context of marine debris detection. We expect that the ideas presented in this chapter can be further developed to become state of the art in their respective sub-fields, with specific application to tasks related to collecting marine debris from the ocean floor.

\section{End-to-End Object Detection}

\subsection{Introduction}

We have built a object detection pipeline based on detection proposals. While object detection is a well researched subject, many underwater object detection systems suffer from generalization issues that we have mentioned in the past.

There is a rampart use of feature engineering that is problem and object specific, which harms the ability to use such features for different objects and environments. We have previously shown how classic methods for sonar images do not perform adequately for marine debris, and extensions to use deep neural networks are required.

A natural extension of our detection proposal system is to include a classification stage so full object detection can be performed. An additional desirable characteristic of a CNN-based object detector is end-to-end training. This consists of training a single network that perform both tasks (detection and classification) in a way that all intermediate steps are also internally learned by the network. Only input images and labels must be provided (hence end-to-end).

In this section we showcase a CNN architecture that can perform both object detection (through detection proposals) and object recognition (with a integrated classifier) in a single unified network. We do this using multi-task learning, where a CNN architecture is designed to output two sets of values: An objectness score for detection proposals (same as Chapter \ref{chapter:proposals}), and a softmax probability distribution over class labels (like our image classifiers in Chapter \ref{chapter:sonar-classification}).

We show that this architecture can be easily trained with only labeled image crops that contain both objectness and class labels, which effectively makes an end-to-end system with acceptable performance.

\subsection{Proposed Method}

We design a network that takes a single input image patch and produces two outputs, namely:

\begin{itemize}
	\item \textbf{Objectness}. This corresponds as an objectness score in $[0, 1]$ used to decide if an object is present in the image. Its interpretation is the same as with our detection proposals algorithm (in Chapter \ref{chapter:proposals}). The basic idea is to threshold the objectness score or use ranking to produce detections. This is implemented as a sigmoid activation function.
	\item \textbf{Class Decision}. This corresponds to one of $C + 1$ pre-defined classes, which indicate the object class decided by the network. We include an additional class that represents background, meaning no object present in that window. This is implemented as a softmax activation which produces a probability distribution over classes. The output class can then be selected by the one with highest probability.
\end{itemize}

Each output represents a different "task" in a multi-task learning framework \cite{ruder2017overview}. Objectness represents the task of object localization, while class decisions corresponds to the classification task.

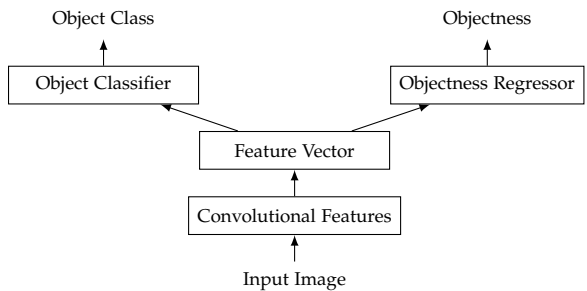
\begin{figure}[!htb]
	\centering
	\begin{tikzpicture}[style={align=center, minimum height=0.5cm, minimum width = 2.5cm}]
	\node[] (dummy) {};
	\node[draw, left=0.0em of dummy] (objClassifier) {{\scriptsize Object Classifier}};
	\node[above=1em of objClassifier] (objClass) {{\scriptsize Object Class}};
	
	\node[draw, right=0.0em of dummy] (objDetector) {{\scriptsize Objectness Regressor}};
	\node[above=1em of objDetector] (obj) {{\scriptsize Objectness}};
	
	\node[draw, below=1em of dummy] (fc) {{\scriptsize Feature Vector}};
	
	\node[draw, below=1em of fc](conv) {\scriptsize{Convolutional Features}};
	\node[below=1em of conv](inputImage) {{\scriptsize Input Image}};
	\draw[-latex] (inputImage) -- (conv);
	\draw[-latex] (conv) -- (fc);
	\draw[-latex] (fc) -- (objDetector);
	\draw[-latex] (fc) -- (objClassifier);
	\draw[-latex] (objClassifier) -- (objClass);
	\draw[-latex] (objDetector) -- (obj);
	\end{tikzpicture}
	\caption[Basic Object Detection Architecture]{Basic Object Detection Architecture. A single image is input to a neural network with two outputs. A shared set of convolutional features are computed, producing a feature vector that is used by both the object classifier and object detector to produce their decisions.}
	\label{apps-detection:basicArchitecture}
\end{figure}

The basic network design that we propose is shown in Figure \ref{apps-detection:basicArchitecture}. A certain number of convolutional feature maps form the "trunk" of the network, which are feed directly from the input image. A 1D feature vector is produced by the network trunk, typically by a fully connected layer receiving input from the convolutional feature maps, but we give freedom to the  network designer to set the right architecture for the problem.

It should be noted that the shared feature vector does not need to be one-dimensional. There is the possibility to keep two-dimensional convolutional feature maps in order to exploit spatial structure in the features. In order to produce a one-dimensional feature vector, flattening is usually applied, and this could lose some spatial information. We mention this option but do not explore it further.

Two output sub-networks are fed by the shared feature vector. One sub-network produces an objectness score, while the other sub-network produces a class decision. Once this network is built and trained, it can be used in a sliding window fashion over the input image to produce detections that also include class information, forming a complete object detection system.

The use of a shared feature vector is based on the idea of information sharing. As objectness and object classification share some similarities, such as the concept of an object and variability between object classes, it makes sense to learn a feature vector that combine/fuses both kinds of information into a single source.  This vector should encode both objectness and class information, allowing for a classifier and objectness regressor to extract useful information for each of the tasks.

Another view of the shared feature vector is that of minimizing the amount of information required for both tasks. As both tasks require common information, a smaller feature vector can be learned instead of learning two feature vectors that might be forced to duplicate information. This should require less data and a simpler network architecture.

There is also evidence \cite{caruana1997multitaskJournalML} that multi-task learning can improve the performance of each individual task . Rich Caruana's PhD thesis \cite{caruana1997multitask} was one of the first to study this topic, and his work found that performance improvements in multi-task learning come from extra information in the training signals of the additional tasks. There are different mechanisms that produce this effect: statistical data amplification, attribute selection, eavesdropping, and representation bias. Caruana also mentions that backpropagation automatically discovers how the tasks are related in a partially unsupervised way.

\newpage
State of the art object detection methods like Faster R-CNN \cite{ren2015faster} and YOLO \cite[1em]{redmon2016you} also use multi-task learning, and we recognize that we use similar ideas to build our object detector.
Our method could be considered a simplification of Faster R-CNN without bounding box regression and only a single anchor box, which allows for one scale only. In comparison, our method is trained end-to end, while Faster R-CNN and YOLO can only be trained successfully with a pre-trained network on the ImageNet dataset.

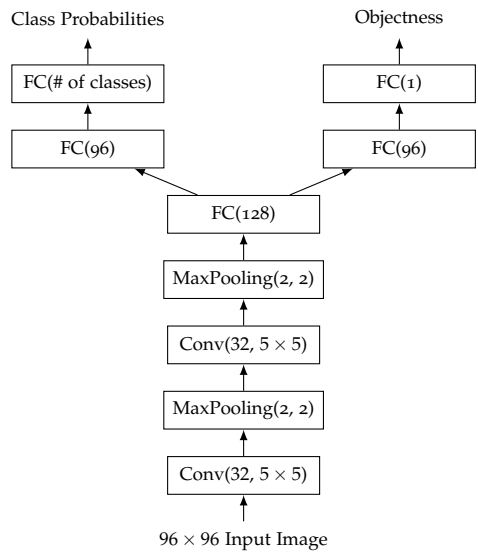
\begin{figure}[!htb]
	\centering
	\begin{tikzpicture}[style={align=center, minimum height=0.5cm, minimum width = 2.0cm}]
	\node[] (dummy) {};
	\node[draw, left=0.1em of dummy] (probsFc1) {{\scriptsize FC(96)}};
	\node[draw, above=1em of probsFc1] (probsFcOut) {{\scriptsize FC(\# of classes)}};
	\node[above=1em of probsFcOut] (probs) {{\scriptsize Class Probabilities}};
	
	\node[draw, right=0.1em of dummy] (objFc1) {{\scriptsize FC(96)}};
	\node[draw, above=1em of objFc1] (objFcOut) {{\scriptsize FC(1)}};
	\node[above=1em of objFcOut] (obj) {{\scriptsize Objectness}};
	
	\node[draw, below=1em of dummy] (fc1) {{\scriptsize FC(128)}};
	
	\node[draw, below=1em of fc1] (mp2) {{\scriptsize MaxPooling(2, 2)}};
	\node[draw, below=1em of mp2] (conv2) {{\scriptsize Conv($32$, $5 \times 5$)}};;
	\node[draw, below=1em of conv2] (mp1) {{\scriptsize MaxPooling(2, 2)}};
	\node[draw, below=1em of mp1](conv1) {{\scriptsize Conv($32$, $5 \times 5$)}};
	\node[below=1em of conv1](inputImage) {{\scriptsize $96 \times 96$ Input Image}};
	\draw[-latex] (inputImage) -- (conv1);
	\draw[-latex] (conv1) -- (mp1);
	\draw[-latex] (mp1) -- (conv2);
	\draw[-latex] (conv2) -- (mp2);
	\draw[-latex] (mp2) -- (fc1);
	\draw[-latex] (fc1) -- (objFc1);
	\draw[-latex] (objFc1) -- (objFcOut);
	\draw[-latex] (fc1) -- (probsFc1);
	\draw[-latex] (probsFc1) -- (probsFcOut);
	\draw[-latex] (probsFcOut) -- (probs);
	\draw[-latex] (objFcOut) -- (obj);
	\end{tikzpicture}
	\caption[Realization of the proposed architecture as a Convolutional Neural Network]{Realization of the proposed architecture as a Convolutional Neural Network. This CNN has 1.8 Million trainable weights.}
	\label{apps-detection:classicNetRealization}
\end{figure}

Figure \ref{apps-detection:classicNetRealization} represents our particular instantiation of the basic architecture fro Figure \ref{apps-detection:basicArchitecture}. This architecture is based on ClassicNet but with two "heads" for classification and object localization. We obtained this model by performing grid search on a validation set with a variation of number of layers, convolutional filters, and shared feature vector size.

The network is trained end-to-end with both tasks of detection and recognition at the same time. This process requires the minimization of a multi-task loss function. In our case we use a linear combination of two loss terms. For classification we use the categorical cross-entropy loss, while for objectness we use the mean squared error:

\begin{equation}
	L(y, \hat{y}) = n^{-1} \sum (y_o - \hat{y}_o)^2 - \gamma \sum_i \sum_c y^c_i \log \hat{y}^c_i
	\label{apps-detection:multiTaskLoss}
\end{equation}

Where the $o$ subscript is used for objectness labels, and the $c$ superscript for class information. The factor $\gamma$ controls the importance of each sub-loss term into the overall loss value. Tuning an appropriate value is key to obtain good performance. We evaluate the effect of this parameter in our experiments.\\

Batch normalization \cite{ioffe2015batch} is used after every trainable layer to prevent overfitting and accelerate training. The network is trained with stochastic gradient descent with a batch size of $b = 128$ images and the ADAM optimizer \cite[1em]{kingma2014adam} with a starting learning rate of $\alpha = 0.01$. We train until the validation loss stops improving, normally after 20 epochs.

\subsection{Experimental Evaluation}

From the perspective of the Marine Debris task, an object detector has two main objectives: to detect debris objects with high recall, and to classify them with high accuracy. We evaluate recall instead of both precision and recall as the object detector has to generalize well outside of its training set, where it might detect novel objects but classify them incorrectly. To evaluate false positives, we measure the number of output proposals, as it correlates with the possibility of a false positive.

We evaluate our object detection method in terms of three metrics: proposal recall, classification accuracy, and number of output proposals. We use proposal recall as defined in Chapter \ref{chapter:proposals}, while we define classification accuracy for this problem as the ratio between correctly classified bounding boxes over the number of ground truth bounding boxes. This kind of evaluation metrics allow for separate assessment of object localization (detection) and classification.

For simplicity we only use objectness thresholding as a way to extract detections from objectness scores, as it was the best performing method when we evaluated it in Chapter \ref{chapter:proposals}. For many point-wise comparisons we use $T_o = 0.5$, as it makes sense since it is the middle value of the valid scale for $T_o$, and in general it produces a good trade-off between number of detection proposals and recall and accuracy.

We have not yet defined a concrete value of the multi-task loss weight $\gamma$. Most papers that use multi-task learning tune this value on a validation set, but we have found out that setting the right value is critical for good performance and balance between the tasks. We take an empiric approach and evaluate a predefined set of values, namely $\gamma \in [0.5, 1, 2, 3, 4]$ and later determine which produces the best result using both recall and accuracy. This is not an exhaustive evaluation, but we found that it produces good results. Recent research by Kendall et al. \cite[-6em]{kendall2017multi} proposes a method to automatically tune multi-task weights using task uncertainty that could be used in the future. This paper also notes that multi-task weights have a great effect on overall system performance.

\begin{figure*}[t]
	\centering
	\forcerectofloat
	\begin{tikzpicture}
		\begin{axis}[height = 0.21\textheight, width = 0.49\textwidth, xlabel={Objectness Threshold ($T_o$)}, ylabel={Detection Recall (\%)}, xmin=0, xmax=1.0, ymin=0, ymax=100.0, ymajorgrids=true, xmajorgrids=true, grid style=dashed, legend pos = south west]        
			\addplot+[mark=none, y filter/.code={\pgfmathparse{\pgfmathresult*100.}\pgfmathresult}] table[x  = threshold, y  = recall, col sep = space] {chapters/data/applications/detection/thresholdVsRecallAndAccuracyAtIoU0.5Lambda0.5.csv};
			\addplot+[mark=none, y filter/.code={\pgfmathparse{\pgfmathresult*100.}\pgfmathresult}] table[x  = threshold, y  = recall, col sep = space] {chapters/data/applications/detection/thresholdVsRecallAndAccuracyAtIoU0.5Lambda1.csv};
			\addplot+[mark=none, y filter/.code={\pgfmathparse{\pgfmathresult*100.}\pgfmathresult}] table[x  = threshold, y  = recall, col sep = space] {chapters/data/applications/detection/thresholdVsRecallAndAccuracyAtIoU0.5Lambda2.csv};
			\addplot+[mark=none, y filter/.code={\pgfmathparse{\pgfmathresult*100.}\pgfmathresult}] table[x  = threshold, y  = recall, col sep = space] {chapters/data/applications/detection/thresholdVsRecallAndAccuracyAtIoU0.5Lambda3.csv};
			\addplot+[mark=none, y filter/.code={\pgfmathparse{\pgfmathresult*100.}\pgfmathresult}] table[x  = threshold, y  = recall, col sep = space] {chapters/data/applications/detection/thresholdVsRecallAndAccuracyAtIoU0.5Lambda4.csv};
		\end{axis}        
	\end{tikzpicture}
	\begin{tikzpicture}
		\begin{axis}[height = 0.21\textheight, width = 0.49\textwidth, xlabel={Objectness Threshold ($T_o$)}, ylabel={Classification Accuracy (\%)}, xmin=0, xmax=1.0, ymin=0, ymax=100.0, ymajorgrids=true, xmajorgrids=true, grid style=dashed, legend pos = south west]        
			\addplot+[mark=none, y filter/.code={\pgfmathparse{\pgfmathresult*100.}\pgfmathresult}] table[x = threshold, y = accuracy, col sep = space] {chapters/data/applications/detection/thresholdVsRecallAndAccuracyAtIoU0.5Lambda0.5.csv};
			\addplot+[mark=none, y filter/.code={\pgfmathparse{\pgfmathresult*100.}\pgfmathresult}] table[x = threshold, y = accuracy, col sep = space] {chapters/data/applications/detection/thresholdVsRecallAndAccuracyAtIoU0.5Lambda1.csv};
			\addplot+[mark=none, y filter/.code={\pgfmathparse{\pgfmathresult*100.}\pgfmathresult}] table[x = threshold, y  = accuracy, col sep = space] {chapters/data/applications/detection/thresholdVsRecallAndAccuracyAtIoU0.5Lambda2.csv};
			\addplot+[mark=none, y filter/.code={\pgfmathparse{\pgfmathresult*100.}\pgfmathresult}] table[x = threshold, y = accuracy, col sep = space] {chapters/data/applications/detection/thresholdVsRecallAndAccuracyAtIoU0.5Lambda3.csv};
			\addplot+[mark=none, y filter/.code={\pgfmathparse{\pgfmathresult*100.}\pgfmathresult}] table[x = threshold, y = accuracy, col sep = space] {chapters/data/applications/detection/thresholdVsRecallAndAccuracyAtIoU0.5Lambda4.csv};
		\end{axis}        
	\end{tikzpicture}
	
	\begin{tikzpicture}
		\begin{axis}[height = 0.21\textheight, width = 0.49\textwidth, xlabel={Objectness Threshold ($T_o$)}, ylabel={Number of Proposals}, xmin=0, xmax=1.0, ymin=0, ymajorgrids=true, xmajorgrids=true, grid style=dashed, legend pos = outer north east]        
			\addplot+[mark=none] table[x  = threshold, y = numberOfProposals, col sep = space] {chapters/data/applications/detection/thresholdVsRecallAndAccuracyAtIoU0.5Lambda0.5.csv};
			\addlegendentry{$\gamma = \frac{1}{2}$}
			\addplot+[mark=none] table[x  = threshold, y = numberOfProposals, col sep = space] {chapters/data/applications/detection/thresholdVsRecallAndAccuracyAtIoU0.5Lambda1.csv};
			\addlegendentry{$\gamma = 1$}
			\addplot+[mark=none] table[x  = threshold, y = numberOfProposals, col sep = space] {chapters/data/applications/detection/thresholdVsRecallAndAccuracyAtIoU0.5Lambda2.csv};
			\addlegendentry{$\gamma = 2$}
			\addplot+[mark=none] table[x  = threshold, y = numberOfProposals, col sep = space] {chapters/data/applications/detection/thresholdVsRecallAndAccuracyAtIoU0.5Lambda3.csv};
			\addlegendentry{$\gamma = 3$}
			\addplot+[mark=none] table[x  = threshold, y = numberOfProposals, col sep = space] {chapters/data/applications/detection/thresholdVsRecallAndAccuracyAtIoU0.5Lambda4.csv};
			\addlegendentry{$\gamma = 4$}
		\end{axis}        
	\end{tikzpicture}
	\caption[Detection Recall, Classification Accuracy and Number of Proposals as a function of Objectness Threshold $T_o$]{Detection Recall, Classification Accuracy and Number of Proposals as a function of Objectness Threshold $T_o$. Different combinations of the multi-task loss weight $\gamma$ are shown.}
	\label{apps-detection:objectnessThreshVsAccuracyRecallAndNumberOfProposals}
\end{figure*}
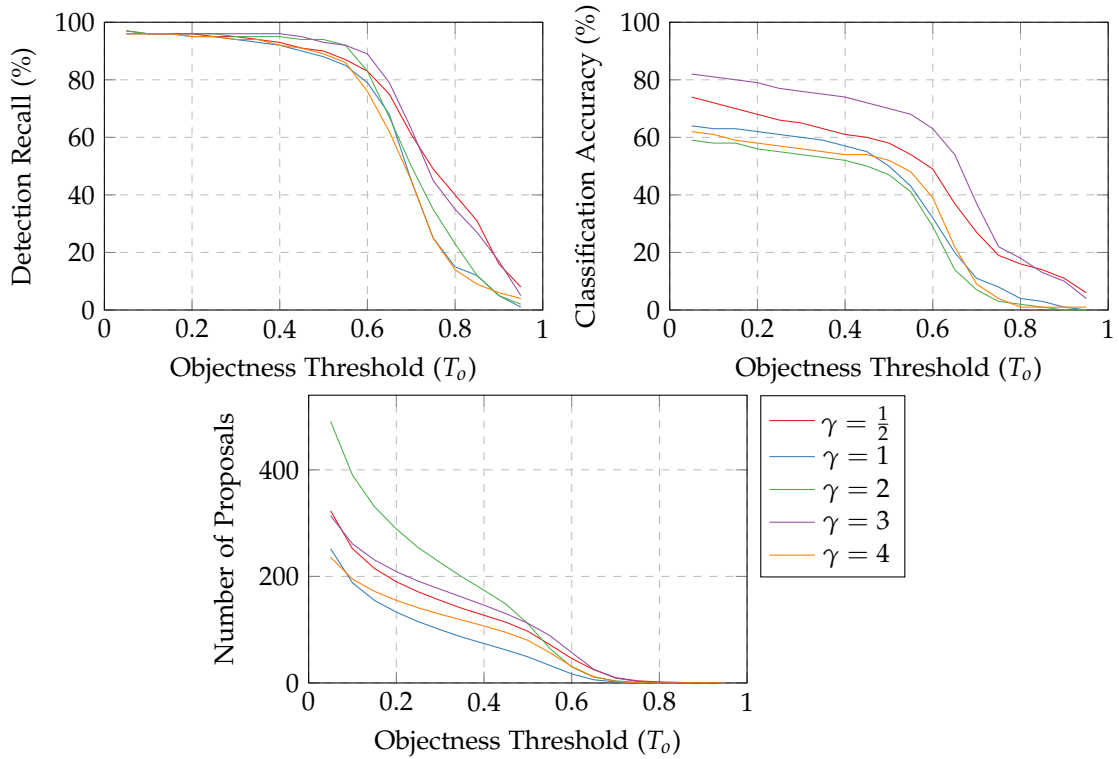

Figure \ref{apps-detection:objectnessThreshVsAccuracyRecallAndNumberOfProposals} shows our principal results as the objectness threshold $T_o$ is varied and we also show the effect of different multi-task loss weight $\gamma$.

Our detection proposals recall is good, close to $95$ \% at $T_o = 0.5$ for many values of $\gamma$, showing little variation between different choices of that parameter. Larger variations in recall are observed for bigger ($> 0.6$) values of $T_o$.

Classification accuracy produced by our system shows a considerable variation across different values of the multi-task loss weight $\gamma$. The best performing value is $\gamma = 3$, but it is not clear why this value is optimal or how other values can be considered (other than trial and error). As mentioned before, setting multi-task loss weights is not trivial and has a large effect on the result. It is also counter-intuitive that a larger value of $gamma$ leads to lower classification performance. At $T_o = 0.5$, the best performing model produces $70$ \% accuracy.

Looking at the number of proposals, again there is a considerable variation between different values of $gamma$, but the number of proposal reduces considerably with an increasing objectness threshold $T_o$. For $T_o = 0.5$ the best performing value $gamma = 3$ produces $112 \pm 62$ proposals per image. This is higher than the number of proposals that our non-classification methods produce (as shown in Chapter \ref{chapter:proposals}).

From a multi-task learning point of view, learning the classification task seems to be considerably harder than the objectness regression task. This is probably due to the different task complexity (modeling class-agnostic objects seems to be easier than class-specific objects), and also we have to consider the limitations from our small training set. Other object detectors like Faster R-CNN are usually trained in much bigger datasets, also containing more object pose variability.

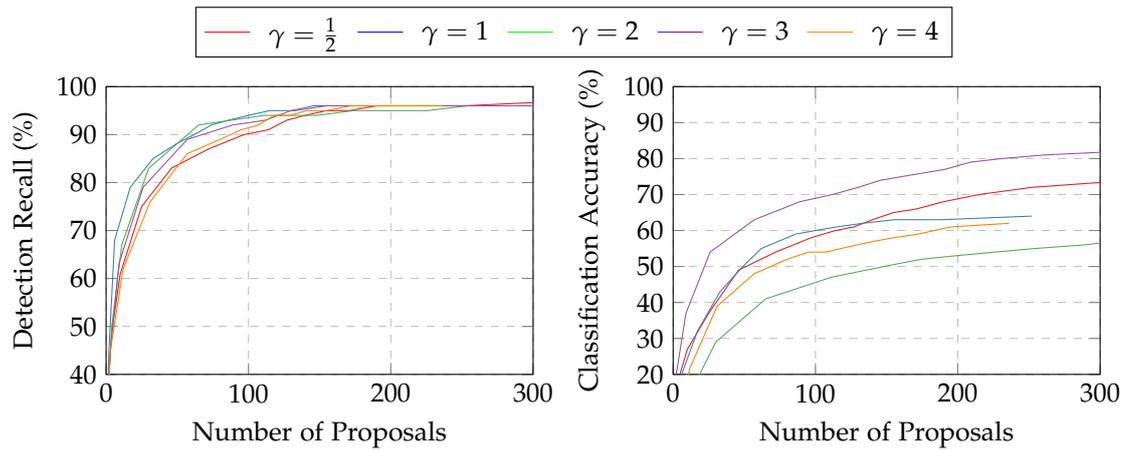
\begin{figure*}[!htb]
	\forceversofloat
	\centering
	\begin{tikzpicture}
		\begin{customlegend}[legend columns = 5,legend style = {column sep=1ex}, legend cell align = left,
	legend entries={$\gamma = \frac{1}{2}$, $\gamma = 1$, $\gamma = 2$, $\gamma = 3$, $\gamma = 4$}]
			\addlegendimage{mark=none,red}
			\addlegendimage{mark=none,blue}			
			\addlegendimage{mark=none,green}
			\addlegendimage{mark=none,violet}
			\addlegendimage{mark=none,orange}
		\end{customlegend}
	\end{tikzpicture}
	
	\begin{tikzpicture}
	\begin{axis}[height = 0.21\textheight, width = 0.48\textwidth, xlabel={Number of Proposals}, ylabel={Detection Recall (\%)}, xmin=0, xmax=300, ymin=40, ymax=100.0, ymajorgrids=true, xmajorgrids=true, grid style=dashed, legend pos = south west, ytick = {40, 50, 60, 70, 80, 90, 100}]        
	\addplot+[mark=none, y filter/.code={\pgfmathparse{#1*100}\pgfmathresult}] table[x  = numberOfProposals, y  = recall, col sep = space] {chapters/data/applications/detection/thresholdVsRecallAndAccuracyAtIoU0.5Lambda0.5.csv};
	\addplot+[mark=none, y filter/.code={\pgfmathparse{#1*100}\pgfmathresult}] table[x  = numberOfProposals, y  = recall, col sep = space] {chapters/data/applications/detection/thresholdVsRecallAndAccuracyAtIoU0.5Lambda1.csv};
	\addplot+[mark=none, y filter/.code={\pgfmathparse{#1*100}\pgfmathresult}] table[x  = numberOfProposals, y  = recall, col sep = space] {chapters/data/applications/detection/thresholdVsRecallAndAccuracyAtIoU0.5Lambda2.csv};
	\addplot+[mark=none, y filter/.code={\pgfmathparse{#1*100}\pgfmathresult}] table[x  = numberOfProposals, y  = recall, col sep = space] {chapters/data/applications/detection/thresholdVsRecallAndAccuracyAtIoU0.5Lambda3.csv};
	\addplot+[mark=none, y filter/.code={\pgfmathparse{#1*100}\pgfmathresult}] table[x  = numberOfProposals, y  = recall, col sep = space] {chapters/data/applications/detection/thresholdVsRecallAndAccuracyAtIoU0.5Lambda4.csv};
	\end{axis}        
	\end{tikzpicture}
	\begin{tikzpicture}
	\begin{axis}[height = 0.21\textheight, width = 0.48\textwidth, xlabel={Number of Proposals}, ylabel={Classification Accuracy (\%)}, xmin=0, xmax=300, ymin=20, ymax=100.0, ymajorgrids=true, xmajorgrids=true, grid style=dashed, legend pos = south west, ytick = {20, 30, 40, 50, 60, 70, 80, 90, 100}]        
	\addplot+[mark=none, y filter/.code={\pgfmathparse{#1*100}\pgfmathresult}] table[x = numberOfProposals, y = accuracy, col sep = space] {chapters/data/applications/detection/thresholdVsRecallAndAccuracyAtIoU0.5Lambda0.5.csv};
	\addplot+[mark=none, y filter/.code={\pgfmathparse{#1*100}\pgfmathresult}] table[x = numberOfProposals, y = accuracy, col sep = space] {chapters/data/applications/detection/thresholdVsRecallAndAccuracyAtIoU0.5Lambda1.csv};
	\addplot+[mark=none, y filter/.code={\pgfmathparse{#1*100}\pgfmathresult}] table[x = numberOfProposals, y  = accuracy, col sep = space] {chapters/data/applications/detection/thresholdVsRecallAndAccuracyAtIoU0.5Lambda2.csv};
	\addplot+[mark=none, y filter/.code={\pgfmathparse{#1*100}\pgfmathresult}] table[x = numberOfProposals, y = accuracy, col sep = space] {chapters/data/applications/detection/thresholdVsRecallAndAccuracyAtIoU0.5Lambda3.csv};
	\addplot+[mark=none, y filter/.code={\pgfmathparse{#1*100}\pgfmathresult}] table[x = numberOfProposals, y = accuracy, col sep = space] {chapters/data/applications/detection/thresholdVsRecallAndAccuracyAtIoU0.5Lambda4.csv};
	\end{axis}        
	\end{tikzpicture}
	\caption[Detection Recall and Classification Accuracy as a function of the Number of Proposals]{Detection Recall and Classification Accuracy as a function of the Number of Proposals. Different combinations of the multi-task loss weight $\gamma$ are shown.}
	\label{apps-detection:accuracyRecallVsNumberOfProposals}
\end{figure*}

Figure \ref{apps-detection:accuracyRecallVsNumberOfProposals} shows the relation between recall/accuracy and the number of detection proposals. For detection proposal recall, our results show that only 100 to 150 proposals per image are required to achieve $95$ \% recall, and using more proposals only marginally increases performance.

For classification accuracy, the pattern is different from proposal recall. As mentioned previously, there is a large variation in classification performance as the multi-task loss weight $\gamma$ is varied, and clearly $\gamma = 3$ performs best. But accuracy increases slowly as the number of proposals is increased, which shows that many proposals are being misclassified, indicating a problem with the classification branch of the network. We expected that classification performance will increase in a similar way as to proposal recall, if the network is performing well at both tasks, but it is likely that our implicit assumption that both tasks are approximately equally hard might not hold.

While our object detector has high proposal recall, the results we obtained in terms of classification are not satisfactory. \newpage 

Inspired by Sharif et al. \cite[5em]{sharif2014cnn}, we evaluated replacing the fully connected layers with a SVM classifier. This has the potential of performing better, as the number of neurons in the fully connected layers used for classification could not be optimal. Nonetheless these layers are required during training, in order to force the shared feature vector to learn a representation that is useful for both classification and objectness regression.

\begin{figure}[t]
	\forcerectofloat
	\centering
	\begin{tikzpicture}
		\begin{customlegend}[legend columns = 5,legend style = {column sep=1ex}, legend cell align = left,
		legend entries={SVM Classifier, FC Classifier}]
			\addlegendimage{mark=none,red}
			\addlegendimage{mark=none,blue}			
		\end{customlegend}
	\end{tikzpicture}
	\begin{tikzpicture}
		\begin{axis}[height = 0.21\textheight, width = 0.85\textwidth, xlabel={Objectness Threshold ($T_o$)}, ylabel={Classification Accuracy (\%)}, xmin=0, xmax=1.0, ymin=30, ymax=100.0, ymajorgrids=true, xmajorgrids=true, grid style=dashed, ytick = {30, 40, 50, 60, 70, 80, 90, 100}]        
			\addplot+[mark=none, y filter/.code={\pgfmathparse{\pgfmathresult*100.}\pgfmathresult}] table[x  = threshold, y  = accuracy, col sep = space] {chapters/data/applications/detection/finetunedThresholdVsAccuracy.csv};
			\addplot+[mark=none, y filter/.code={\pgfmathparse{\pgfmathresult*100.}\pgfmathresult}] table[x  = threshold, y  = accuracy, col sep = space] {chapters/data/applications/detection/thresholdVsRecallAndAccuracyAtIoU0.5Lambda3.csv};
		\end{axis}        
	\end{tikzpicture}
	\caption[Comparison of SVM and FC classifiers using features learned by model $\gamma = 3$]{Comparison of SVM and FC classifiers using features learned by model $\gamma = 3$. The SVM classifier produces a considerable improvement over the FC one. We did not evaluate past $T_o > 0.65$ as the number of available training samples was too low.}
	\label{apps-detection:finetuningResults}
\end{figure}
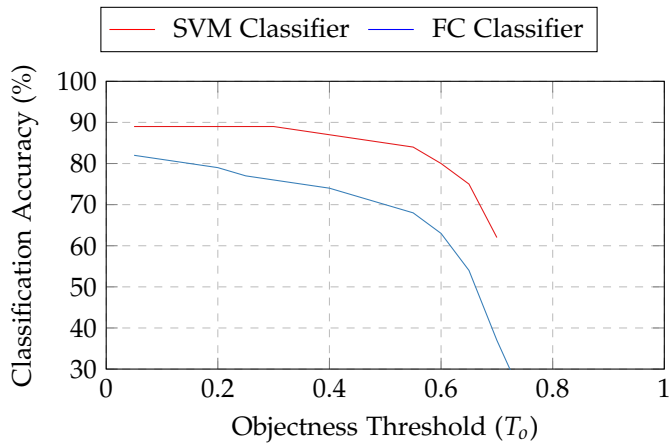

We freeze model weights and compute the shared feature vectors for all images in the training set. Then we train a linear SVM with $C = 1$ (obtained from grid search in a validation set) and one-versus-one scheme for multi-class classification. We then replace the classification branch with this trained classifier. Results are shown in Figure \ref{apps-detection:finetuningResults}.

Our results show that using a SVM classifier outperforms the fully connected one by a large margin. Performance is also more stable with respect to variation in the objectness threshold $T_o$. At $T_o = 0.5$, the SVM classifier produces $85$ \% recall, which is a $15$ \% absolute improvement over using a fully connected layer with a softmax activation. These results are more usable for real world applications, but still there is a large room for improvement.

We believe that our results indicate that a big problem is mismatch between the training set and the test set. During inference, the model has to classify an image that is generated by a sliding window, and depending on the position of the object in the image, classification might be difficult or fail completely. We believe this problem can be alleviated by introducing a more fine sliding window stride $s$ and a larger training set, containing more object variability and object translations inside the image window.

An alternative to the use of a multi-class SVM classifier is to fine-tune the classification layers with a ping-pong approach. First detection proposals are generated on the training set, and matched to correct bounding boxes. Then these matching detection proposals are used to fine-tune the classifier by including them into the classification training set and continue training the classification layers from previous weights (not randomly initialized) for a predefined number of epochs. Then this process is repeated, and feedback between object detection and classification should lead to convergence after a certain number of iterations. This process is costly as it requires a full evaluation cycle on the training set and multiple forward and backward passes to train specific parts of the network. In this thesis we only showcased the simple approach of a SVM classifier.

We have also noticed a problem with the multiple detections that are generated for a single ground truth object. These can be reduced by applying non-maximum suppression, but also there is no guarantee that the highest scoring bounding box is classified as the correct object class. This problem is related to the previously mentioned one, as we found out that the top scoring window was consistently misclassified.

As we do not reach $100$ \% detection recall, then it is not possible to achieve $100$ \% classification accuracy, and we have probably reached the best performance that this simple model can perform. We note that our model uses a single scale, and more complex output configurations could be used. For example, inspired on Faster R-CNN's anchor boxes \cite{ren2015faster}, multiple fixed-size bounding boxes could be output and a classifier may decide which one is the correct one for a given sliding window position. This would allow for multiple scales and/or aspect ratios to be decided by the system, but without introducing full bounding box regression, which he have found to be an issue in small training data scales.

As future work, additional terms to the loss function could be added in order to ensure or force that high objectness scoring windows are more likely to be correctly classified. A more fine tuning of the multi-task loss weight $\gamma$ should be performed, and more training data, including more variability on object pose, should be used.

\FloatBarrier
\newpage
\section{Tracking by Matching}

\subsection{Introduction}

Tracking is the process of first detecting an object of interest in an image, and then continually detect its position in subsequent frames. This operation is typically performed on video frames or equivalently on data that has a temporal dimension. Tracking is challenging due to the possible changes in object appearance as time moves forward.

In general, tracking is performed by finding relevant features in the target object, and try to match them in subsequent frames \cite{yilmaz2006object}. This is called feature tracking. An alternative formulation is tracking by detection \cite[1em]{vcehovin2016visual}, which uses simple object detectors to detect the target in the current and subsequent frames, with additions in order to exploit temporal correlations.

In this section we evaluate a tracker built with our matching function, as to showcase a real underwater robotics application. It should be noted that this section is not intended as a general tracking evaluation in sonar images, but to compare the use of similarity measures and matching for tracking.

\subsection{Proposed Method}

A popular option for tracking is tracking as object detection, where an object is tracked by first running an object detector on the input image, and then matching objects between frames by considering features like overlap, appearance and filter predictions. Our matching networks can also be used for tracking. We evaluate a tracker that is composed of two stages:

\begin{itemize}
	\item \textbf{Detection Stage}. We detect objects in the input image using our detection proposal algorithm presented in Chapter \ref{chapter:proposals}. At tracker initialization, the proposal to be tracked is cropped from the input image and saved as template for matching.
	\item \textbf{Matching Stage}. Each proposal is matched to the template obtained during initialization, and the proposal with the highest matching score is output as the tracked object.
\end{itemize}

While this method is simple, it incorporates all the information that matching and detection proposal algorithms contain. Tracker performance completely depends on matching accuracy, and we will show that our tracker performs appropriately.

\subsection{Experimental Evaluation}

For the Marine Debris task, tracking is required as the AUV might experience underwater currents that result in the target object moving in the sonar image, and during object manipulation, the object must be tracked robustly. One way to measure robustness of the tracker is to measure the number of correctly tracked frames (CTF) \cite{vcehovin2016visual}, which is also human interpretable.

To make a fair comparison, we constructed another tracker that uses a cross-correlation similarity (Eq \ref{ccSimilarityEq}) instead of our CNN matching function. If $S_{CC} > 0.01$, then we declare a match. This is a deliberate low threshold used to evaluate the difficulty of tracking objects in our dataset. The same detection proposal algorithm from Chapter \ref{chapter:proposals} is used. For both trackers we generate the top 10 proposals ranked by objectness score.
\vspace*{1em}
\begin{equation}
S_{CC}(I, T) = \frac{\sum (I - \bar{I}) \sum (T - \bar{T})}{\sqrt{\sum (I - \bar{I})^2 \sum (T - \bar{T})^2}}
\label{ccSimilarityEq}
\end{equation}

We have evaluated 3 sequences with one object each. The first sequence corresponds to 51 frames containing a can, the second sequence corresponds to 55 valve mockup frames, and the last sequence contains a glass bottle over 17 frames. We use network 2-Chan CNN Class as a matching function, as this is the best performing matcher that we have built.

We report the CTF as a function of the overlap score threshold $O_t$ (Intersection over Union,  Eq \ref{iouEquation}) between the predicted bounding box and ground truth.
\vspace*{1em}
\begin{equation}
\text{iou}(A, B) = \frac{\text{area}(A \cap B)}{\text{area}(A \cup B)}
\label{iouEquation}
\end{equation}

The CTF metric captures how well the tracker tracks the object as a function of $O_t$. Typically the CTF is only reported at $O_t = 0.5$, which is the most common value used for object detection. We report the CTF at this value in Table \ref{apps-track:trackingSummary}.

\begin{table*}[t]
	\begin{tabular}{lllll}
		\hline
		Metric / Sequence 						& Method		& Can 			& Valve 	& Glass Bottle\\
		\hline
		\multirow{2}{*}{CTF at IoU $O_t = 0.5$} & Matching CNN	& $\mathbf{74.0}$ \%	& $\mathbf{100.0}$ \%	& $\mathbf{81.3}$ \% \\
					   						    & CC TM			& $6.0$ \%	& $35.2$ \%		& $50.0$ \% \\
		\hline
	\end{tabular}
    \vspace*{0.5cm}
	\caption[Summary of tracking results]{Summary of tracking results. We compare our tracker built using a Matching CNN versus Cross-Correlation Template Matching using the correctly tracked frames ratio (CTF).}
	\label{apps-track:trackingSummary}
\end{table*}

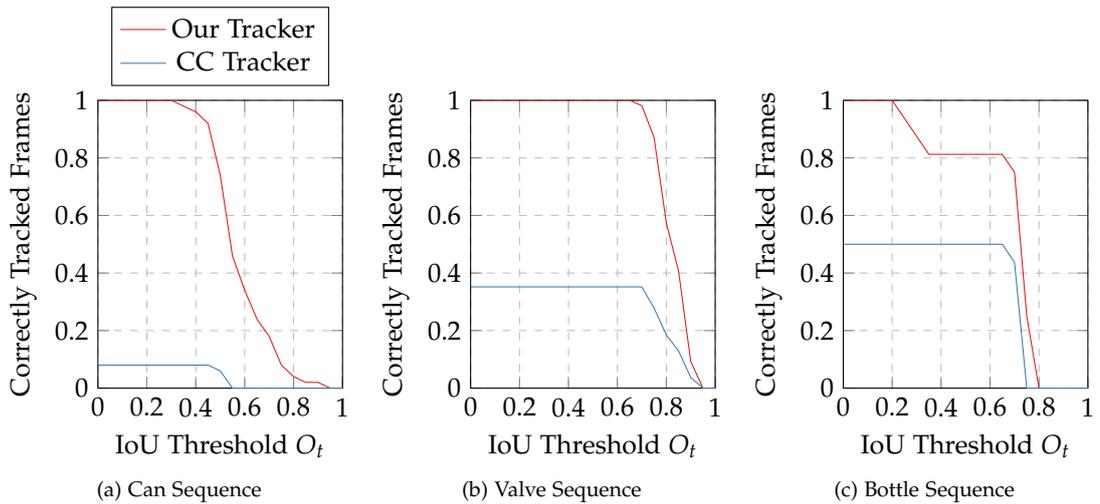
\begin{figure*}[!t]
	\forceversofloat
	\centering
	\subfloat[Can Sequence]{        
		\begin{tikzpicture}
		\begin{axis}[height = 0.21\textheight, width = 0.32\textwidth, xlabel={IoU Threshold $O_t$}, ylabel={Correctly Tracked Frames}, xmin=0, xmax=1.0, ymin=0, ymax=1.0, ymajorgrids=true, xmajorgrids=true, grid style=dashed, legend style={at={(0.5, 1.05)},anchor=south}]        
		\addplot+[mark = none] table[x  = threshold, y  = ctf, col sep = space] {chapters/data/applications/tracking/trackingBinaryMatcherPerformance-can-CTFVsThreshold.csv};
		\addlegendentry{Our Tracker}
		\addplot+[mark = none] table[x  = threshold, y  = ctf, col sep = space] {chapters/data/applications/tracking/trackingTemplateMatcherPerformance-can-CTFVsThreshold.csv};
		\addlegendentry{CC Tracker}        
		\end{axis}        
		\end{tikzpicture}
	}
	\subfloat[Valve Sequence]{
		\begin{tikzpicture}
		\begin{axis}[height = 0.21\textheight, width =  0.32\textwidth, xlabel={IoU Threshold $O_t$}, ylabel={Correctly Tracked Frames}, xmin=0, xmax=1.0, ymin=0, ymax=1.0, ymajorgrids=true, xmajorgrids=true, grid style=dashed, legend pos = south east]        
		\addplot+[mark = none] table[x  = threshold, y  = ctf, col sep = space] {chapters/data/applications/tracking/trackingBinaryMatcherPerformance-valve-CTFVsThreshold.csv};
		\addplot+[mark = none] table[x  = threshold, y  = ctf, col sep = space] {chapters/data/applications/tracking/trackingTemplateMatcherPerformance-valve-CTFVsThreshold.csv};
		\end{axis}        
		\end{tikzpicture}
	}
	\subfloat[Bottle Sequence]{
		\begin{tikzpicture}
		\begin{axis}[height = 0.21\textheight, width = 0.32\textwidth, xlabel={IoU Threshold $O_t$}, ylabel={Correctly Tracked Frames}, xmin=0, xmax=1.0, ymin=0, ymax=1.0, ymajorgrids=true, xmajorgrids=true, grid style=dashed, legend pos = south east]        
		\addplot+[mark = none] table[x  = threshold, y  = ctf, col sep = space] {chapters/data/applications/tracking/trackingBinaryMatcherPerformance-bottle-CTFVsThreshold.csv};
		\addplot+[mark = none] table[x  = threshold, y  = ctf, col sep = space] {chapters/data/applications/tracking/trackingTemplateMatcherPerformance-bottle-CTFVsThreshold.csv};
		\end{axis}        
		\end{tikzpicture}
	}
	\vspace*{0.5cm}
	\caption{Number of correctly tracked frames as a function of the IoU overlap threshold $O_t$ for each sequence.}
	\label{apps-tracking:ctfPlots}
\end{figure*}

Our tracker results are shown in Fig. \ref{apps-tracking:ctfPlots}. In all three sequences the matching CNN makes a better tracker, which can be seen from a higher correctly tracked frames ratio across different $O_t$ values.
This may be due to variations in the background (insonification), reflections and small object rotations. If any of these conditions change, the CC between template and target drops close to zero, producing strong tracker failures. Our matching network is more robust to object rotations than plain CC.  The CC tracker has a high failure rate, specially for the Can Sequence where it achieves the lowest correctly tracker frames ratio. 

The Can Sequence (Fig \ref{apps-tracking:ctfPlots}a) is interesting as our tracker can track the object well ($74.0 \%$ CTF at $O_t = 0.5$), but the CC tracker fails and cannot keep track of the object. This is due to strong reflections generated by the can's lid. Seems that our matching function is partially invariant to these reflections and performs better. This result shows that a CNN provides better generalization for matching than using other classic methods.

It could be considered that initializing the tracker once, without re-initialization at failure, is extreme, but our tracker works well with single initialization. It can successfully track the object even as insonification or orientation changes.

\begin{figure*}[!t]
	\centering
	\subfloat[Can Sequence]{        
		\begin{tikzpicture}
		\begin{axis}[height = 0.21\textheight, width = 0.29\textwidth, xlabel={Normalized Frame Number}, ylabel={IoU}, xmin=0, xmax=1.0, ymin=0, ymax=1.0, ymajorgrids=true, xmajorgrids=true, grid style=dashed, legend style={at={(0.5, 1.05)},anchor=south}]        
		\addplot+[mark = none] table[x  = normFrame, y  = iou, col sep = space] {chapters/data/applications/tracking/trackingBinaryMatcherPerformance-can-K3.csv};
		\addlegendentry{Our Tracker}
		\addplot+[mark = none] table[x  = normFrame, y  = iou, col sep = space] {chapters/data/applications/tracking/trackingTemplateMatcherPerformance-can-K3.csv};
		\addlegendentry{CC Tracker}        
		\end{axis}        
		\end{tikzpicture}
	}
	\subfloat[Valve Sequence]{
		\begin{tikzpicture}
		\begin{axis}[height = 0.21\textheight, width =  0.29\textwidth, xlabel={Normalized Frame Number}, ylabel={IoU}, xmin=0, xmax=1.0, ymin=0, ymax=1.0, ymajorgrids=true, xmajorgrids=true, grid style=dashed, legend pos = south east]        
		\addplot+[mark = none] table[x  = normFrame, y  = iou, col sep = space] {chapters/data/applications/tracking/trackingBinaryMatcherPerformance-valve-K3.csv};
		\addplot+[mark = none] table[x  = normFrame, y  = iou, col sep = space] {chapters/data/applications/tracking/trackingTemplateMatcherPerformance-valve-K3.csv};
		\end{axis}        
		\end{tikzpicture}
	}
	\subfloat[Bottle Sequence]{
		\begin{tikzpicture}
		\begin{axis}[height = 0.21\textheight, width = 0.29\textwidth, xlabel={Normalized Frame Number}, ylabel={IoU}, xmin=0, xmax=1.0, ymin=0, ymax=1.0, ymajorgrids=true, xmajorgrids=true, grid style=dashed, legend pos = south east]        
		\addplot+[mark = none] table[x  = normFrame, y  = iou, col sep = space] {chapters/data/applications/tracking/trackingBinaryMatcherPerformance-bottle-K3.csv};
		\addplot+[mark = none] table[x  = normFrame, y  = iou, col sep = space] {chapters/data/applications/tracking/trackingTemplateMatcherPerformance-bottle-K3.csv};
		\end{axis}        
		\end{tikzpicture}
	}
	\vspace*{0.5cm}
	\caption{IoU score as a function of the frame numbers for each sequence.}
	\label{apps-track:iouPlots}
\end{figure*}
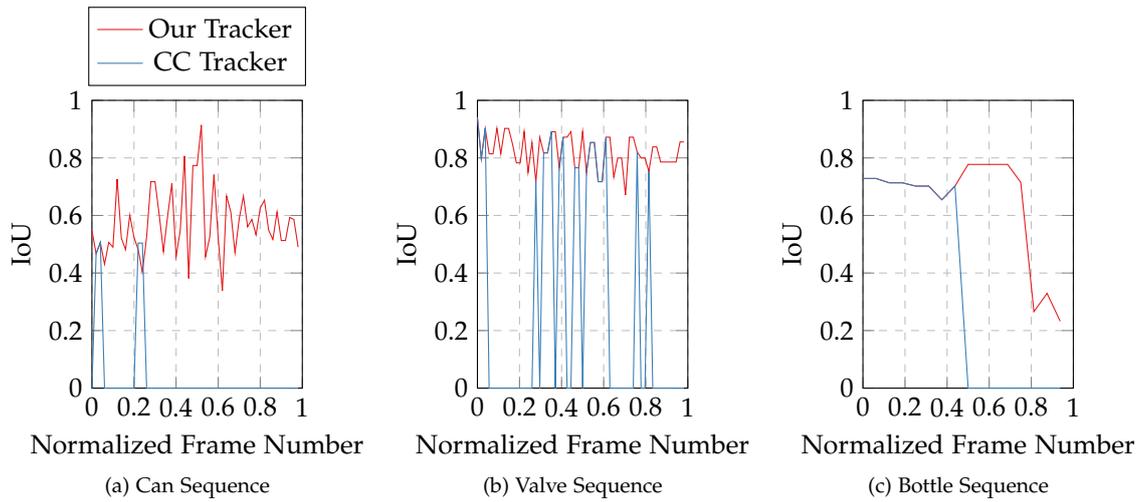

Figure \ref{apps-track:iouPlots} shows a comparison of IoU values as a function of the normalized frame number. These plots show how the cross-correlation tracker fails. For the Can sequence the CC-based tracker can only correctly track the object for a small number of frames, and it constantly confuses the object with background. A similar effect is seen for the Valve sequence, while the Bottle sequence has better tracking performance but completely fails after working for $40$ \% of the frames. CC tracking failures could be due to correlations between the sonar background and template background, which could be larger than the correlations induced by the actual object. This means that a CC-based tracker fails because it learns to track the background instead of the target object. As mentioned in Chapter \ref{chapter:sonar-classification}, many template matching methods first segment the image and the template in order to reduce the effect of background, but doing this requires additional background models.

This result means that the matching network effectively learned to compare object features instead of concentrating on the background, which is highly desirable and could explain the generalization gap with other methods.

\begin{figure*}[!tb]    
    \centering
    \subfloat[Can Sequence]{
        \centering
        \includegraphics[width=0.30\textwidth]{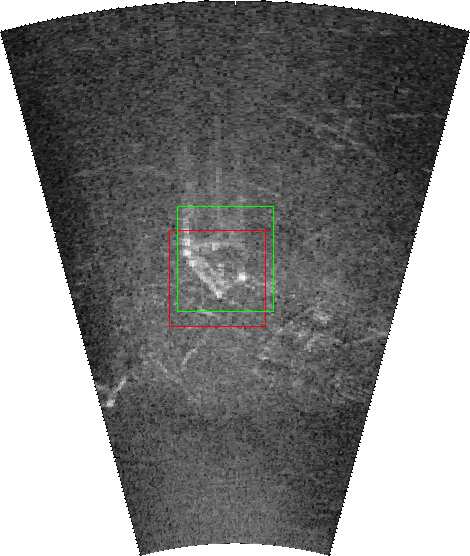}
        \includegraphics[width=0.30 \textwidth]{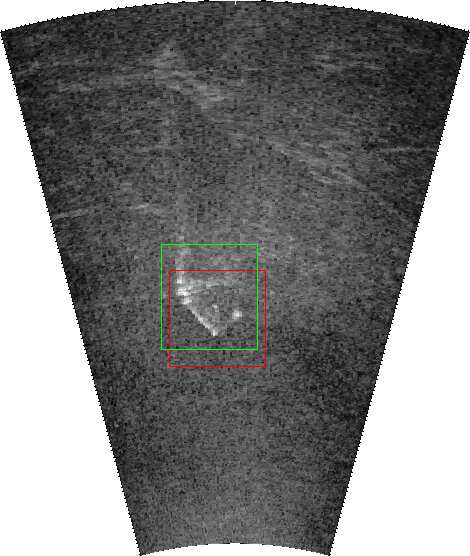}
        \includegraphics[width=0.30 \textwidth]{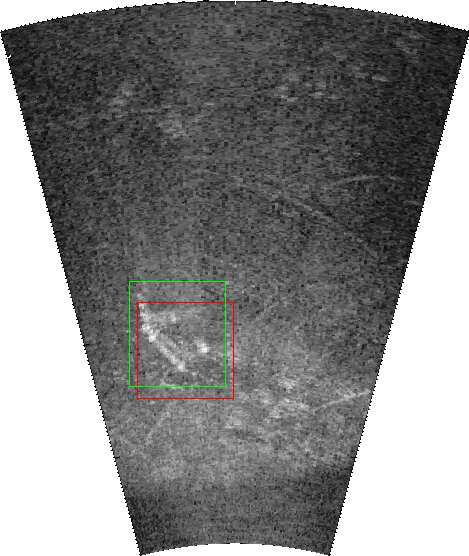}
    }
    
    \subfloat[Valve Sequence]{
        \centering
        \includegraphics[width=0.23\textwidth]{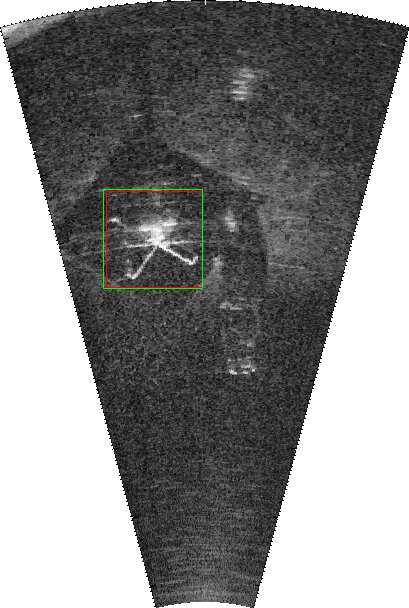}
        \includegraphics[width=0.23\textwidth]{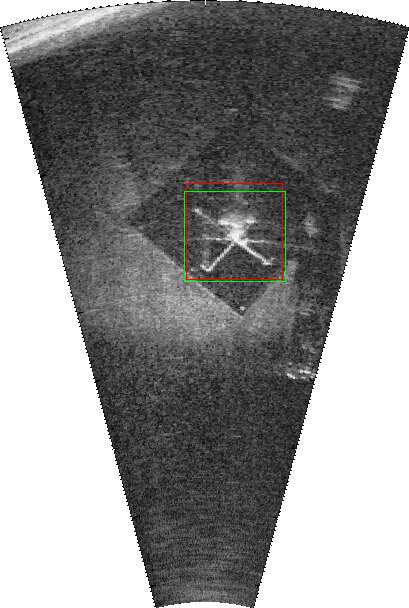}
        \includegraphics[width=0.23\textwidth]{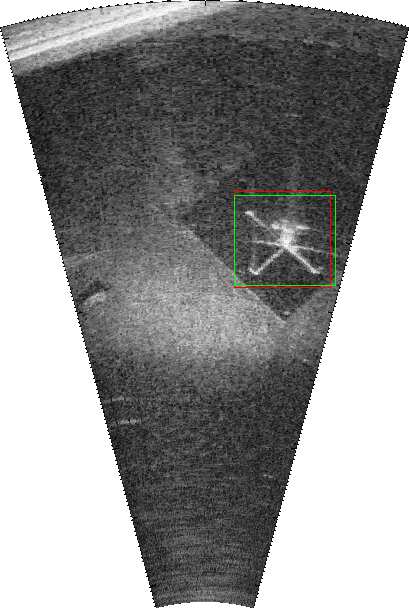}
    }
    
    \subfloat[Bottle Sequence]{
        \centering
        \includegraphics[width=0.23\textwidth]{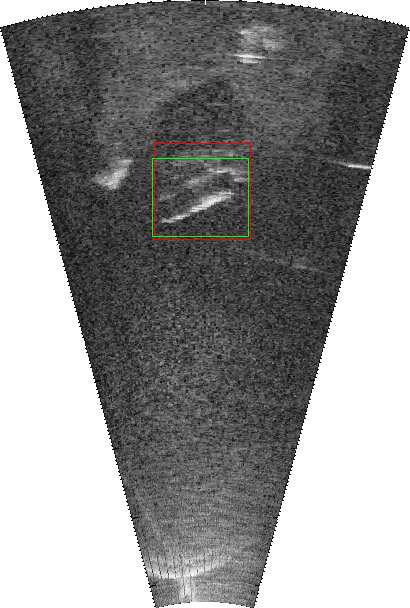}
        \includegraphics[width=0.23\textwidth]{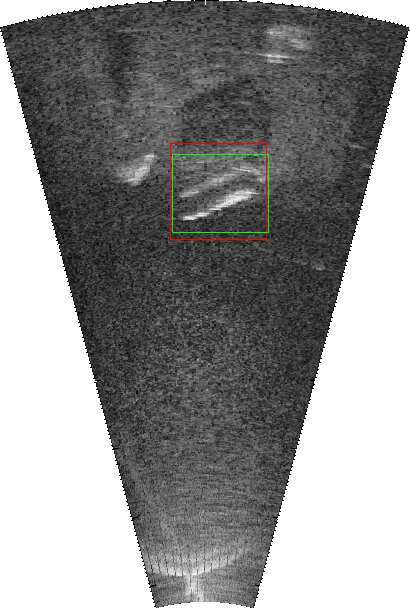}
        \includegraphics[width=0.23\textwidth]{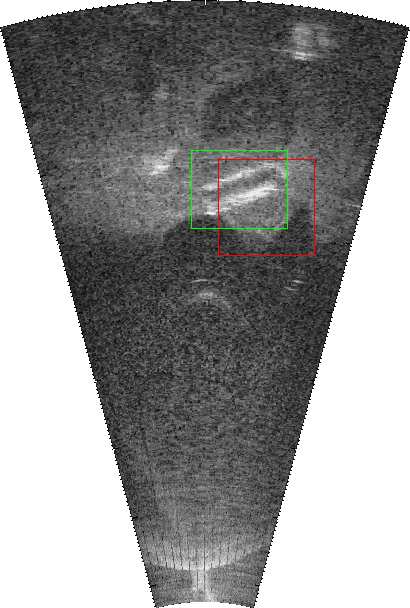}
    }
    \vspace*{0.5cm}
    \caption[Sample tracking results generated by our tracker]{Sample tracking results generated by our tracker. Green is the ground truth, while red is the bounding box generated by our tracker.}
    \label{sampleTrackerFrames}
\end{figure*}

A small sample of tracked objects is shown in  Fig. \ref{sampleTrackerFrames}. These images show the changes in insonification (seen as changes in background around the object) and object appearance that we previously mentioned. This effect is more noticeable in the Bottle sequence. The Can sequence also shows this effect, as the object does not change orientation significantly, but background varies, which confuses a cross-correlation based tracker.

It should be mentioned that using cross-correlation is a commonly used as a simple tracking solution, but it has major issues, like a tendency to lose track of the object by correlating with spurious image features, which make it drift from the real tracked object. We have not observed such behaviour, but our results also show that template matching is not a good solution for patch matching and tracking.

We believe our results show that the matching network can be successfully used for tracking. This is only a limited example with a template matching baseline, but it shows that a simple tracker that can easily be trained from labeled data (including the detection proposal and matching networks) without much effort. Comparing with other state of the art trackers in sonar images, they usually include a heavy probabilistic formulation that is hard to implement and tune.

We do not believe that our tracker can outperform state of the art trackers that are based on CNNs. Other newer approaches are more likely to perform better but require considerable more data. For example, the deep regression networks for tracking by Held et al \cite{held2016learning} . The winners from the Visual Object Tracking (VOT) challenge (held each year) are strong candidates to outperform our method, but usually these methods are quite complex and are trained on considerably larger datasets.

A big challenge in Machine Learning and Computer Vision is to solve tasks without requiring large training sets \cite{sun2017revisiting}. While large datasets are available for many computer vision tasks, their availability for robotics scenarios is smaller. Particularly for underwater environments the availability of public datasets is very scarce. Then our tracking technique is particularly valuable in contexts where large training sets are not available.

Improvements in the matching function, either from better network models or data with more object variability, will immediately transfer into improvements in tracking. As we expect that detection proposals will be used in an AUV to find "interesting" objects, using the matching networks for tracking can be considered as a more streamlined architecture than just one network that performs tracking as a black box separately from object detection.

\section{Summary of Results}

In this chapter we have presented two applications of detection proposals. The first is object detection with an end-to-end pipeline which allows high recall ($95 \%$) and modest accuracy ($80 \%$) without making assumptions on object shape and not requiring pre-training on a classification dataset (thus end-to-end). Learning this way allows features to adapt to the problem, but it has the issue of not providing the best classification performance.
After replacing the fully connected layers that do classification with a multi-class SVM, we see an improvement to $90 \%$ accuracy, which shows that there is room for improvement. We expect that more data and better techniques will further improve performance, specially with respect of reducing false positives.

The second application was object tracking by combining our detection proposals algorithm with matching. By using the first detection as matching target, we can continuously track a desired object. In comparison with a template matching approach, using a neural network for matching provides a much better correctly tracked frames metric in the marine debris sequence dataset, with a higher IoU. The template matching tracker constantly loses track of the object and cannot correctly match it with the original object view. This shows that the matching CNNs are much more robust for real-world applications.

%% file: chapters/conclusions.tex
\chapter{Conclusions and Future Work}
\label{chapter:conclusions}
Marine debris is ubiquitous in the world's oceans, rivers, and lakes, caused by the massive production of disposable products, mostly made out of plastic and glass. Marine debris has been found up to 4000 meters deep in the Mariana Trench.  There is evidence that surface and submerged marine debris \cite{iniguez2016marine} are harmful to marine environments, and efforts to reduce the amount of debris getting into the sea are underway \cite[1em]{mcilgorm2008understanding}. This thesis proposes that Autonomous Underwater Vehicles can be used to survey and detect submerged marine debris with the long term goal of recover and cleaning up.

In this thesis we have developed techniques to detect and recognize marine debris in Forward-Looking sonar images. But more importantly, we have proposed the use of Autonomous Underwater Vehicles to survey and detect submerged marine debris with sonar sensor. We show that detecting marine debris is harder than other objects such as mine like objects (MLOs), because techniques commonly used for MLOs fail to generalize well for debris.

We did this through several research lines, and encouraged by recent advances in Deep Neural Networks (DNNs), we adopt this framework and develop our own neural network models for Forward-Looking Sonar images.

We captured a small dataset of 2069 Forward-Looking Sonar images using a ARIS Explorer 3000 sensor, containing 2364 labeled object instances with bounding box and class information. This dataset was captured from a selected set of household and marine objects, representing non-exhaustive instances of marine debris. The data capture was realized from a water tank, as we had issues with the sensor in a real world underwater environment. We use this dataset for training and evaluation of all machine learning models in this thesis.

The first research line considers Forward-Looking Sonar image classification, as a proxy for posterior performance in object detection. A state of the art template matching classifier obtains $98.1 \%$ accuracy using sum of squared differences similarity, and $93.0 \%$ for cross-correlation similarity, which pales in comparison with a DNN classifier that obtains up to $99.7 \%$ accuracy. Both template matching classifiers require 150 templates per class in order to obtain this result, which implies that the whole training set is memorized. A DNN requires far fewer trainable parameters (up to 4000) to achieve a better accuracy on our dataset.

We have also developed neural network architectures for sonar image classification that generalize well and are able to run in real-time in low power platforms, such as the Raspberry Pi 2. This is done by carefully designing a neural network, based on the Fire module \cite[-7em]{iandola2016squeezenet}, that has a low number of parameters but still having good classification performance, requiring less computational resources. This shows that a DNN can perform well even in the constrained computing environments of an AUV.

As a second research line, we evaluated the practical capabilities of in Forward-Looking sonar image classification DNNs with respect to the amount of training data required for generalization, the input image size, and the effect of transfer learning. We find that when scaling down the training set, classification performance drops considerably, but when using transfer learning to first learn features and use them with an SVM classifier, classification performance increases radically. Even with a training set of one sample per class, it is possible to generalize close to $90 \%$ accuracy when learning features from different objects.
With respect of input image size, we find that classifier architectures with fully connected layers are able to generalize well even if trained with a dataset of $16 \times 16$ images, but this is only possible if using the Adam optimizer and Batch Normalization. Using SGD and Dropout has the effect of producing a linear relation between accuracy and image size.

Our architectures based on the Fire module do not have the same behavior, and suffer from decreased performance with small image sizes, and with smaller datasets. We believe additional research is needed to understand why this happens, as a model with less parameters should be easier to train with less data.

The third research line is the problem of matching two in Forward-Looking Sonar image patches of the same size. This is a difficult problem as multiple viewpoints will produce radically different sonar images. We frame this problem as binary classification. Using two-channel neural networks we obtain up to $0.91$ Area under the ROC Curve (AUC), which is quite high when compared with state of the art keypoint detection techniques such as SIFT, SURF, and AKAZE, whom obtain $0.65-0.80$ AUC. This result holds when a DNN is trained with one set of objects, and tested with a different one, where we obtain $0.89$ AUC.

As fourth research line we explored the use of detection proposal algorithms in order to detect objects in in Forward-Looking sonar images, even when the object shape is unknown or there are no training samples of the object. Our methods based on DNNs can achieve up to $95 \%$ recall, while generalizing well to unseen objects. We even used data captured by the University of Girona, using an older version of the same sonar sensor, and noticed that detections are appropriate for the chain object. We also evaluated the number of output detections/proposals that are required for good generalization, and our methods obtain high recall with less than 100 proposals per image, while a baseline template matching that we built requires at least 300 proposals per image, while only achieving $90 \%$ recall. In comparison, state of the art proposals techniques like EdgeBoxes and Selective Search require thousands of detections to reach comparable recall levels.

As the final technical chapter of this thesis, we showcase two applications of our methods in Forward-Looking Sonar images. The first is performing object detection by classifying proposals, in an end-to-end DNN architecture that learns both to detect proposals through objectness, and classify them with an additional output head. This method obtains $80 \%$ correct classifications, which can be improved to $90 \%$ accuracy by replacing the classifier with an SVM trained on the learned features.

The second application is object tracking in Forward-Looking Sonar images. We build a simple tracker by combining our detection proposals algorithm with the matching system, from where an object is first detected and continually matched to the first detection across time. This method outperforms a cross-correlation based matcher that we used as baseline in the correctly tracked frames metric..

Overall we believe that our results show that using Deep Neural Networks is promising for the task of Marine Debris detection in Forward-Looking Sonar images, and that they can be successfully used in Autonomous Underwater Vehicles, even when computational power is constrained.

We expect that interest on neural networks will increase in the AUV community, and they will be used for detection of other kinds of objects in marine environments, specially as no feature engineering or object shadow/highlight shapes are needed. For example, our detection proposals algorithm has the potential of being used to find \textit{anomalies} in the seafloor, which can be useful to find wrecked ships and airplanes at sea.

\section{Future Work}

There is plenty of work that can be done in the future to extend this thesis.

The dataset that we captured does not fully cover the large variety of marine debris, and only has one environment: the OSL water tank. We believe that a larger scientific effort should be made to capture a ImageNet-scale dataset of marine debris in a variety of real-world environments. This requires the effort of more than just one PhD student. A New dataset should consider a larger set of debris objects, and include many kinds of \textit{distractor} objects such as rocks, marine flora and fauna, with a richer variety of environments, like sand, rocks, mud, etc.

Bounding Box prediction seems to be a complicated issue, as our informal experiments showed that with the data that is available to us, it does not converge into an usable solution. Due to this we used fixed scale bounding boxes for detection proposals, which seem to work well for our objects, but it would not work with larger scale variations. A larger and more varied dataset could make a state of the art object detection method such as SSD or Faster R-CNN work well, from where more advanced detection proposal methods could be built upon.

We only explored the use of the ARIS Explorer 3000 Forward-Looking Sonar to detect marine debris, but other sensors could also be useful. Particularly a civilian Synthetic Aperture Sonar could be used for large scale surveying of the seafloor, in particular to locate regions where debris accumulates, and AUVs can target these regions more thoroughly.
Underwater laser scanners could also prove useful to recognize debris or to perform manipulation and grasping for collection, but these sensors would require new neural network architectures to deal with the highly unstructured outputs that they produce. There are newer advances in neural networks\cite{qi2017pointnet} that can process point clouds produced by laser sensors, but they are computationally expensive.

Another promising approach to detect marine debris is that instead of using object detection methods, which detect based visual appearance on an image, a sensor or algorithm could recognize the materials that compose the object under inspection, from where debris could be identified if the most common material is plastic or metal.

This technique could cover a more broad set of objects and could be generally more useful for other tasks, like finding ship or plane wrecks. This approach would require a radically new sonar sensor design that can use multiple frequencies to insonify an object.  In collaboration with Mariia Dmitrieva we have some research\cite[-5em]{dmitrieva2017object} in this direction.

Last but not least, we did not cover the topic of object manipulation in this thesis. Once objects are detected, it is a must to capture them using a manipulator arm with an appropriate end effector. This would require further research in designing an appropriate end effector that can capture any kind of marine debris, and then perception algorithms to robustly do pose estimation, and finally perform the grasping motion to capture the object.

The biggest research issue in this line is that manipulation has to be performed without making any assumptions in object shape or grasping points. All the grasping information has to be estimated by a robust perception algorithm directly from the perceived object, in order for it to be available at runtime. Any kind of assumption made on the object's structure will limit the kind of debris that can be grasped.

%% file: chapters/appendices.tex
\chapter[Randomly Selected Samples of the\\ Marine Debris Dataset]{Randomly Selected Samples of the Marine Debris Dataset} 

In this appendix we present image crops of each class, in order to showcase the inter and intra class variability.

Due to space in this thesis, we decided to only show a sub-sample of each class, randomly selected. We assigned one page of space to each class, and filled each page with constant width and variable height images, until the page is complete. This produced a variable number of samples that is presented for each class, from 24 to 78 samples. We present sonar image crops based on the labeled bounding boxes, which implies that each crop could have a different size. 

For the background class, we present 78 randomly selected samples that are in our dataset, which are also randomly selected at the time we created the classification dataset. This is because the possible number of background windows is high (possibly infinite). All background class crops have the same size, standardized to $96 \times 96$ pixels.

\begin{figure}
    \includegraphics[width=0.15\textwidth]{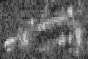}
    \includegraphics[width=0.15\textwidth]{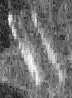}
    \includegraphics[width=0.15\textwidth]{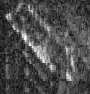}
    \includegraphics[width=0.15\textwidth]{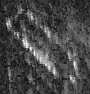}
    \includegraphics[width=0.15\textwidth]{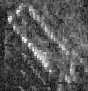}
    \includegraphics[width=0.15\textwidth]{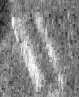}
    
    \includegraphics[width=0.15\textwidth]{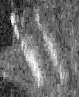}
    \includegraphics[width=0.15\textwidth]{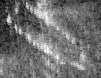}
    \includegraphics[width=0.15\textwidth]{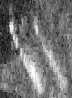}
    \includegraphics[width=0.15\textwidth]{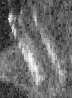}
    \includegraphics[width=0.15\textwidth]{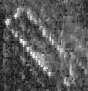}
    \includegraphics[width=0.15\textwidth]{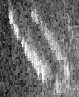}
    
    \includegraphics[width=0.15\textwidth]{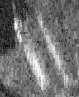}
    \includegraphics[width=0.15\textwidth]{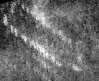}
    \includegraphics[width=0.15\textwidth]{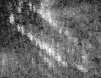}
    \includegraphics[width=0.15\textwidth]{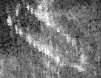}
    \includegraphics[width=0.15\textwidth]{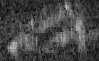}
    \includegraphics[width=0.15\textwidth]{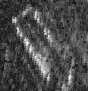}
    
    \includegraphics[width=0.15\textwidth]{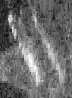}
    \includegraphics[width=0.15\textwidth]{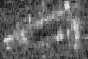}
    \includegraphics[width=0.15\textwidth]{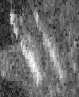}
    \includegraphics[width=0.15\textwidth]{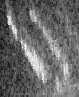}
    \includegraphics[width=0.15\textwidth]{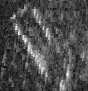}
    \includegraphics[width=0.15\textwidth]{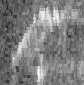}
    
    \includegraphics[width=0.15\textwidth]{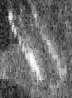}
    \includegraphics[width=0.15\textwidth]{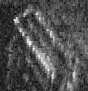}
    \includegraphics[width=0.15\textwidth]{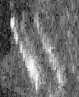}
    \includegraphics[width=0.15\textwidth]{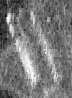}
    \includegraphics[width=0.15\textwidth]{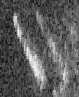}
    \includegraphics[width=0.15\textwidth]{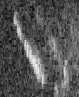}
    
    \includegraphics[width=0.15\textwidth]{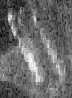}
    \includegraphics[width=0.15\textwidth]{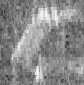}
    \includegraphics[width=0.15\textwidth]{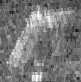}
    \includegraphics[width=0.15\textwidth]{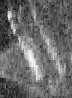}
    \includegraphics[width=0.15\textwidth]{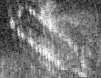}
    \includegraphics[width=0.15\textwidth]{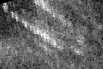}
    
    \includegraphics[width=0.15\textwidth]{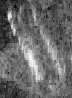}
    \includegraphics[width=0.15\textwidth]{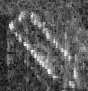}
    \includegraphics[width=0.15\textwidth]{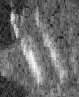}
    \includegraphics[width=0.15\textwidth]{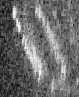}
    \includegraphics[width=0.15\textwidth]{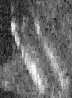}
    \includegraphics[width=0.15\textwidth]{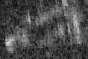}
    
    \includegraphics[width=0.15\textwidth]{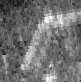}
    \includegraphics[width=0.15\textwidth]{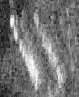}
    \includegraphics[width=0.15\textwidth]{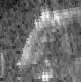}
    \includegraphics[width=0.15\textwidth]{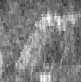}
    \includegraphics[width=0.15\textwidth]{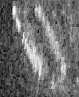}
    \includegraphics[width=0.15\textwidth]{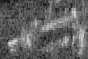}
    
    \includegraphics[width=0.15\textwidth]{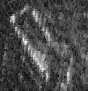}
    \includegraphics[width=0.15\textwidth]{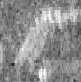}
    \includegraphics[width=0.15\textwidth]{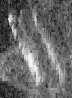}
    \includegraphics[width=0.15\textwidth]{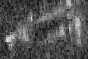}
    \includegraphics[width=0.15\textwidth]{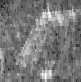}
    \includegraphics[width=0.15\textwidth]{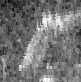}
    
    \includegraphics[width=0.15\textwidth]{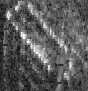}
    \includegraphics[width=0.15\textwidth]{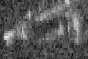}
    \includegraphics[width=0.15\textwidth]{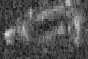}
    \includegraphics[width=0.15\textwidth]{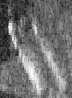}
    \includegraphics[width=0.15\textwidth]{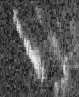}
    \includegraphics[width=0.15\textwidth]{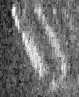}
    
    \includegraphics[width=0.15\textwidth]{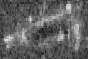}
    \includegraphics[width=0.15\textwidth]{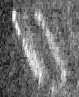}
    \includegraphics[width=0.15\textwidth]{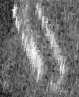}
    \includegraphics[width=0.15\textwidth]{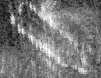}
    \includegraphics[width=0.15\textwidth]{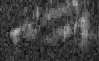}
    \includegraphics[width=0.15\textwidth]{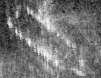}
    
    \caption{Bottle Class. 66 Randomly selected image crops}
    \label{appendix:bottle}
\end{figure}

\begin{figure}
    \centering
    \includegraphics[width=0.15\textwidth]{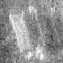}
    \includegraphics[width=0.15\textwidth]{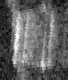}
    \includegraphics[width=0.15\textwidth]{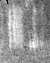}
    \includegraphics[width=0.15\textwidth]{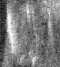}
    \includegraphics[width=0.15\textwidth]{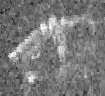}
    \includegraphics[width=0.15\textwidth]{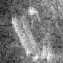}
    
    \includegraphics[width=0.15\textwidth]{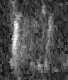}
    \includegraphics[width=0.15\textwidth]{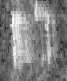}
    \includegraphics[width=0.15\textwidth]{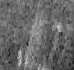}
    \includegraphics[width=0.15\textwidth]{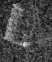}
    \includegraphics[width=0.15\textwidth]{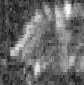}
    \includegraphics[width=0.15\textwidth]{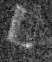}
    
    \includegraphics[width=0.15\textwidth]{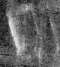}
    \includegraphics[width=0.15\textwidth]{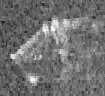}
    \includegraphics[width=0.15\textwidth]{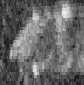}
    \includegraphics[width=0.15\textwidth]{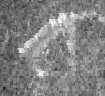}
    \includegraphics[width=0.15\textwidth]{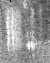}
    \includegraphics[width=0.15\textwidth]{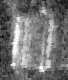}
    
    \includegraphics[width=0.15\textwidth]{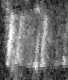}
    \includegraphics[width=0.15\textwidth]{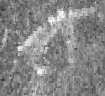}
    \includegraphics[width=0.15\textwidth]{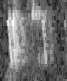}
    \includegraphics[width=0.15\textwidth]{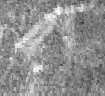}
    \includegraphics[width=0.15\textwidth]{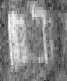}
    \includegraphics[width=0.15\textwidth]{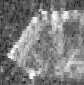}
    
    \includegraphics[width=0.15\textwidth]{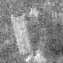}
    \includegraphics[width=0.15\textwidth]{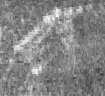}
    \includegraphics[width=0.15\textwidth]{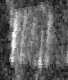}
    \includegraphics[width=0.15\textwidth]{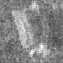}
    \includegraphics[width=0.15\textwidth]{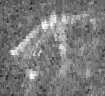}
    \includegraphics[width=0.15\textwidth]{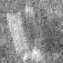}
    
    \includegraphics[width=0.15\textwidth]{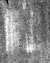}
    \includegraphics[width=0.15\textwidth]{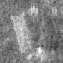}
    \includegraphics[width=0.15\textwidth]{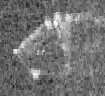}
    \includegraphics[width=0.15\textwidth]{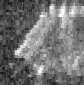}
    \includegraphics[width=0.15\textwidth]{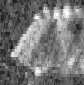}
    \includegraphics[width=0.15\textwidth]{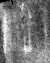}
    
    \includegraphics[width=0.15\textwidth]{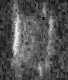}
    \includegraphics[width=0.15\textwidth]{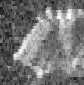}
    \includegraphics[width=0.15\textwidth]{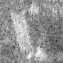}
    \includegraphics[width=0.15\textwidth]{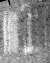}
    \includegraphics[width=0.15\textwidth]{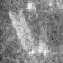}
    \includegraphics[width=0.15\textwidth]{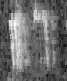}
    
    \includegraphics[width=0.15\textwidth]{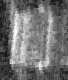}
    \includegraphics[width=0.15\textwidth]{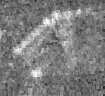}
    \includegraphics[width=0.15\textwidth]{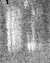}
    \includegraphics[width=0.15\textwidth]{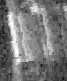}
    \includegraphics[width=0.15\textwidth]{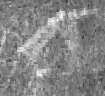}
    \includegraphics[width=0.15\textwidth]{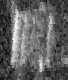}
    
    \includegraphics[width=0.15\textwidth]{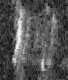}
    \includegraphics[width=0.15\textwidth]{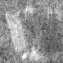}
    \includegraphics[width=0.15\textwidth]{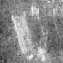}
    \includegraphics[width=0.15\textwidth]{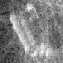}
    \includegraphics[width=0.15\textwidth]{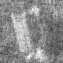}
    \includegraphics[width=0.15\textwidth]{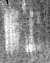}
    
    \includegraphics[width=0.15\textwidth]{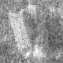}
    \includegraphics[width=0.15\textwidth]{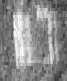}
    \includegraphics[width=0.15\textwidth]{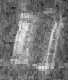}
    \includegraphics[width=0.15\textwidth]{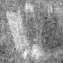}
    \includegraphics[width=0.15\textwidth]{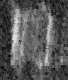}
    \includegraphics[width=0.15\textwidth]{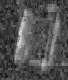}
    
    \includegraphics[width=0.15\textwidth]{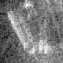}
    \includegraphics[width=0.15\textwidth]{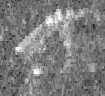}
    \includegraphics[width=0.15\textwidth]{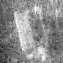}
    \includegraphics[width=0.15\textwidth]{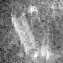}
    \includegraphics[width=0.15\textwidth]{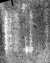}
    \includegraphics[width=0.15\textwidth]{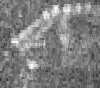}
    
    \includegraphics[width=0.15\textwidth]{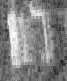}
    \includegraphics[width=0.15\textwidth]{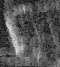}
    \includegraphics[width=0.15\textwidth]{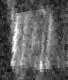}
    \includegraphics[width=0.15\textwidth]{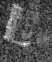}
    \includegraphics[width=0.15\textwidth]{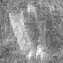}
    \includegraphics[width=0.15\textwidth]{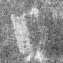}
    
    \caption{Can Class. 72 Randomly selected image crops}
    \label{appendix:can}
\end{figure}

\begin{figure}
    \includegraphics[width=0.15\textwidth]{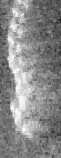}
    \includegraphics[width=0.15\textwidth]{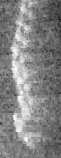}    
    \includegraphics[width=0.15\textwidth]{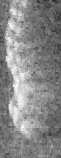}
    \includegraphics[width=0.15\textwidth]{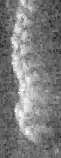}
    \includegraphics[width=0.15\textwidth]{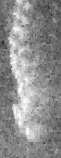}
    \includegraphics[width=0.15\textwidth]{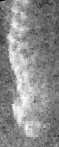}    
    \includegraphics[width=0.15\textwidth]{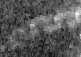}
    \includegraphics[width=0.15\textwidth]{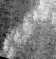}
    \includegraphics[width=0.15\textwidth]{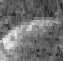}
    \includegraphics[width=0.15\textwidth]{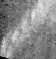}
    \includegraphics[width=0.15\textwidth]{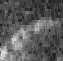}
    \includegraphics[width=0.15\textwidth]{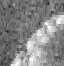}
    \includegraphics[width=0.15\textwidth]{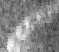}
    \includegraphics[width=0.15\textwidth]{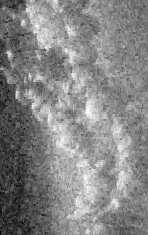}
    \includegraphics[width=0.15\textwidth]{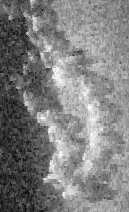}
    \includegraphics[width=0.15\textwidth]{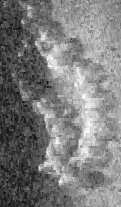}
    \includegraphics[width=0.15\textwidth]{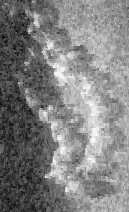}
    \includegraphics[width=0.15\textwidth]{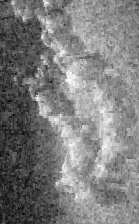}    
    \includegraphics[width=0.15\textwidth]{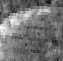}
    \includegraphics[width=0.15\textwidth]{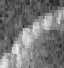}
    \includegraphics[width=0.15\textwidth]{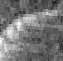}    
    \includegraphics[width=0.15\textwidth]{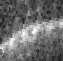}
    \includegraphics[width=0.15\textwidth]{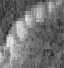}
    \includegraphics[width=0.15\textwidth]{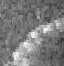}
    \includegraphics[width=0.15\textwidth]{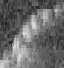}    
    \includegraphics[width=0.15\textwidth]{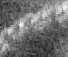}
    \includegraphics[width=0.15\textwidth]{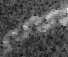}    
    \includegraphics[width=0.15\textwidth]{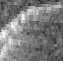}
    \includegraphics[width=0.15\textwidth]{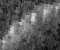}
    \includegraphics[width=0.15\textwidth]{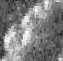}
    \includegraphics[width=0.15\textwidth]{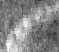}
    \includegraphics[width=0.15\textwidth]{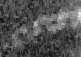}
    \includegraphics[width=0.15\textwidth]{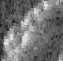}        
    \includegraphics[width=0.15\textwidth]{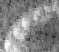}
    \includegraphics[width=0.15\textwidth]{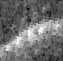}
    \includegraphics[width=0.15\textwidth]{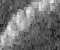}
    \includegraphics[width=0.15\textwidth]{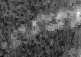}
    \includegraphics[width=0.15\textwidth]{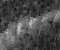}
    \includegraphics[width=0.15\textwidth]{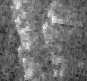}
    \includegraphics[width=0.15\textwidth]{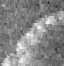}    
    \includegraphics[width=0.15\textwidth]{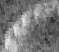}
    \includegraphics[width=0.15\textwidth]{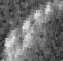} 
       
    \includegraphics[width=0.15\textwidth]{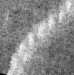}
    \includegraphics[width=0.15\textwidth]{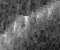}        
    \includegraphics[width=0.15\textwidth]{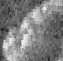}
    \includegraphics[width=0.15\textwidth]{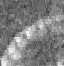}
    \includegraphics[width=0.15\textwidth]{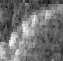}
    \includegraphics[width=0.15\textwidth]{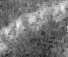}
    
    \includegraphics[width=0.15\textwidth]{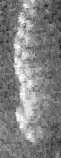}
    \includegraphics[width=0.15\textwidth]{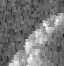}
    \includegraphics[width=0.15\textwidth]{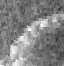}    
    \includegraphics[width=0.15\textwidth]{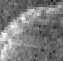}
    \includegraphics[width=0.15\textwidth]{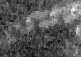}    
    \includegraphics[width=0.15\textwidth]{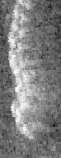}
    
    \caption{Chain Class. 54 Randomly selected image crops}
    \label{appendix:chain}    
\end{figure}

\begin{figure}
    \includegraphics[width=0.15\textwidth]{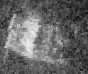}
    \includegraphics[width=0.15\textwidth]{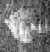}
    \includegraphics[width=0.15\textwidth]{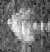}
    \includegraphics[width=0.15\textwidth]{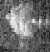}
    \includegraphics[width=0.15\textwidth]{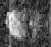}
    \includegraphics[width=0.15\textwidth]{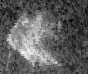}
    
    \includegraphics[width=0.15\textwidth]{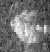}
    \includegraphics[width=0.15\textwidth]{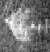}
    \includegraphics[width=0.15\textwidth]{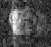}
    \includegraphics[width=0.15\textwidth]{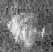}
    \includegraphics[width=0.15\textwidth]{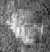}
    \includegraphics[width=0.15\textwidth]{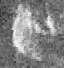}
    
    \includegraphics[width=0.15\textwidth]{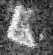}
    \includegraphics[width=0.15\textwidth]{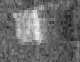}
    \includegraphics[width=0.15\textwidth]{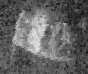}
    \includegraphics[width=0.15\textwidth]{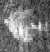}
    \includegraphics[width=0.15\textwidth]{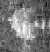}
    \includegraphics[width=0.15\textwidth]{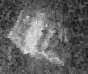}
    
    \includegraphics[width=0.15\textwidth]{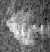}
    \includegraphics[width=0.15\textwidth]{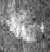}
    \includegraphics[width=0.15\textwidth]{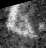}
    \includegraphics[width=0.15\textwidth]{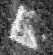}
    \includegraphics[width=0.15\textwidth]{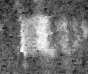}
    \includegraphics[width=0.15\textwidth]{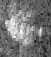}
    
    \includegraphics[width=0.15\textwidth]{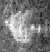}
    \includegraphics[width=0.15\textwidth]{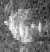}
    \includegraphics[width=0.15\textwidth]{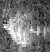}
    \includegraphics[width=0.15\textwidth]{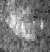}
    \includegraphics[width=0.15\textwidth]{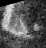}
    \includegraphics[width=0.15\textwidth]{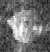}
    
    \includegraphics[width=0.15\textwidth]{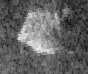}
    \includegraphics[width=0.15\textwidth]{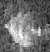}
    \includegraphics[width=0.15\textwidth]{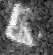}
    \includegraphics[width=0.15\textwidth]{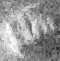}
    \includegraphics[width=0.15\textwidth]{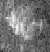}
    \includegraphics[width=0.15\textwidth]{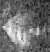}
    
    \includegraphics[width=0.15\textwidth]{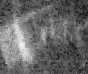}
    \includegraphics[width=0.15\textwidth]{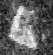}
    \includegraphics[width=0.15\textwidth]{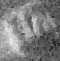}
    \includegraphics[width=0.15\textwidth]{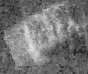}
    \includegraphics[width=0.15\textwidth]{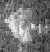}
    \includegraphics[width=0.15\textwidth]{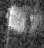}
    
    \includegraphics[width=0.15\textwidth]{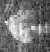}
    \includegraphics[width=0.15\textwidth]{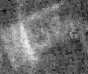}
    \includegraphics[width=0.15\textwidth]{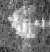}
    \includegraphics[width=0.15\textwidth]{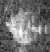}
    \includegraphics[width=0.15\textwidth]{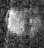}
    \includegraphics[width=0.15\textwidth]{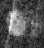}
    
    \includegraphics[width=0.15\textwidth]{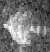}
    \includegraphics[width=0.15\textwidth]{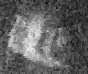}
    \includegraphics[width=0.15\textwidth]{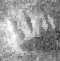}
    \includegraphics[width=0.15\textwidth]{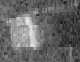}
    \includegraphics[width=0.15\textwidth]{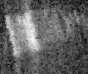}
    \includegraphics[width=0.15\textwidth]{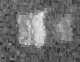}
    
    \includegraphics[width=0.15\textwidth]{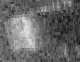}
    \includegraphics[width=0.15\textwidth]{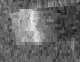}
    \includegraphics[width=0.15\textwidth]{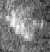}
    \includegraphics[width=0.15\textwidth]{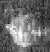}
    \includegraphics[width=0.15\textwidth]{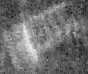}
    \includegraphics[width=0.15\textwidth]{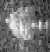}
    
    \includegraphics[width=0.15\textwidth]{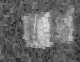}
    \includegraphics[width=0.15\textwidth]{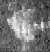}
    \includegraphics[width=0.15\textwidth]{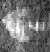}
    \includegraphics[width=0.15\textwidth]{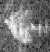}
    \includegraphics[width=0.15\textwidth]{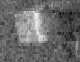}
    \includegraphics[width=0.15\textwidth]{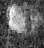}
    
    \includegraphics[width=0.15\textwidth]{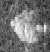}
    \includegraphics[width=0.15\textwidth]{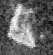}
    \includegraphics[width=0.15\textwidth]{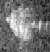}
    \includegraphics[width=0.15\textwidth]{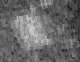}
    \includegraphics[width=0.15\textwidth]{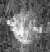}
    \includegraphics[width=0.15\textwidth]{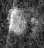}
    
    \includegraphics[width=0.15\textwidth]{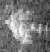}
    \includegraphics[width=0.15\textwidth]{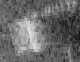}
    \includegraphics[width=0.15\textwidth]{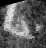}
    \includegraphics[width=0.15\textwidth]{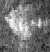}
    \includegraphics[width=0.15\textwidth]{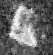}
    \includegraphics[width=0.15\textwidth]{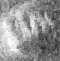}
    
    \caption{Drink Carton Class. 78 Randomly selected image crops}
    \label{appendix:drinkCarton}
\end{figure}

\begin{figure}            
    \includegraphics[width=0.15\textwidth]{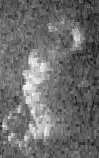}
    \includegraphics[width=0.15\textwidth]{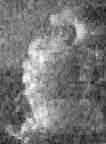}    
    \includegraphics[width=0.15\textwidth]{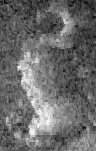}
    \includegraphics[width=0.15\textwidth]{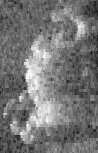}
    \includegraphics[width=0.15\textwidth]{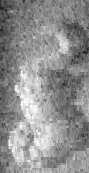}    
    \includegraphics[width=0.15\textwidth]{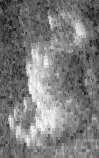}
        
    \includegraphics[width=0.15\textwidth]{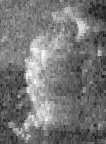}
    \includegraphics[width=0.15\textwidth]{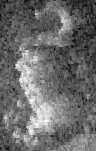}
    \includegraphics[width=0.15\textwidth]{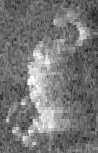}    
    \includegraphics[width=0.15\textwidth]{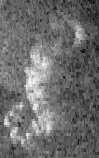}    
    \includegraphics[width=0.15\textwidth]{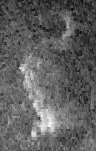}
    \includegraphics[width=0.15\textwidth]{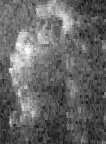}
    
    \includegraphics[width=0.15\textwidth]{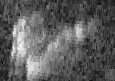}
    \includegraphics[width=0.15\textwidth]{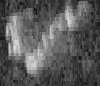}
    \includegraphics[width=0.15\textwidth]{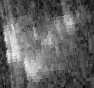}
    \includegraphics[width=0.15\textwidth]{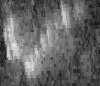}
    \includegraphics[width=0.15\textwidth]{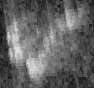}
    \includegraphics[width=0.15\textwidth]{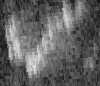}
    
    \includegraphics[width=0.15\textwidth]{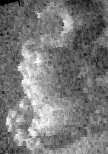}    
    \includegraphics[width=0.15\textwidth]{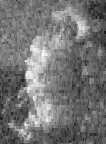}
    \includegraphics[width=0.15\textwidth]{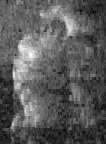}
    \includegraphics[width=0.15\textwidth]{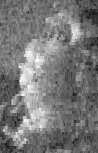}
    \includegraphics[width=0.15\textwidth]{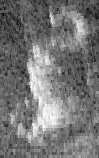}
    \includegraphics[width=0.15\textwidth]{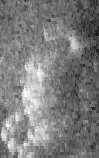}    
    
    \includegraphics[width=0.15\textwidth]{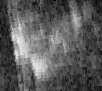}
    \includegraphics[width=0.15\textwidth]{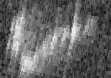}
    \includegraphics[width=0.15\textwidth]{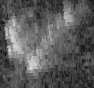}
    \includegraphics[width=0.15\textwidth]{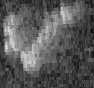}
    \includegraphics[width=0.15\textwidth]{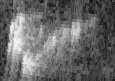}
    \includegraphics[width=0.15\textwidth]{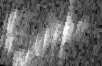}
        
    \includegraphics[width=0.15\textwidth]{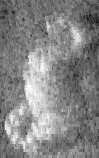}
    \includegraphics[width=0.15\textwidth]{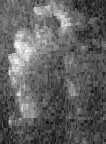}
    \includegraphics[width=0.15\textwidth]{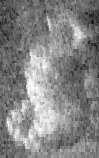}    
    \includegraphics[width=0.15\textwidth]{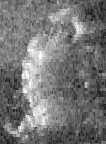}
    \includegraphics[width=0.15\textwidth]{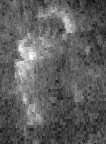}
    \includegraphics[width=0.15\textwidth]{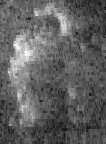}
    
    \includegraphics[width=0.15\textwidth]{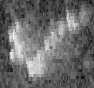}
    \includegraphics[width=0.15\textwidth]{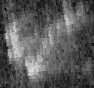}
    \includegraphics[width=0.15\textwidth]{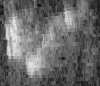}
    \includegraphics[width=0.15\textwidth]{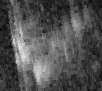}
    \includegraphics[width=0.15\textwidth]{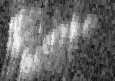}
    \includegraphics[width=0.15\textwidth]{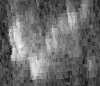}
    
    \includegraphics[width=0.15\textwidth]{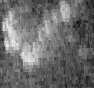}
    \includegraphics[width=0.15\textwidth]{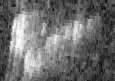}
    \includegraphics[width=0.15\textwidth]{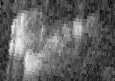}
    \includegraphics[width=0.15\textwidth]{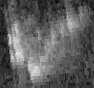}
    \includegraphics[width=0.15\textwidth]{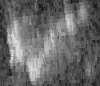}
    \includegraphics[width=0.15\textwidth]{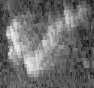}
    
    \includegraphics[width=0.15\textwidth]{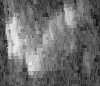}
    \includegraphics[width=0.15\textwidth]{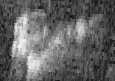}
    \includegraphics[width=0.15\textwidth]{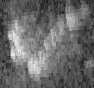}
    \includegraphics[width=0.15\textwidth]{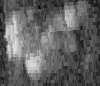}
    \includegraphics[width=0.15\textwidth]{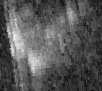}
    \includegraphics[width=0.15\textwidth]{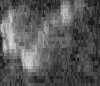}
    
    \includegraphics[width=0.15\textwidth]{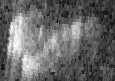}
    \includegraphics[width=0.15\textwidth]{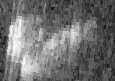}
    \includegraphics[width=0.15\textwidth]{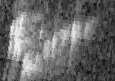}
    \includegraphics[width=0.15\textwidth]{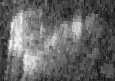}
    \includegraphics[width=0.15\textwidth]{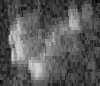}
    \includegraphics[width=0.15\textwidth]{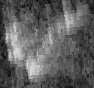}
    
    \includegraphics[width=0.15\textwidth]{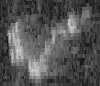}
    \includegraphics[width=0.15\textwidth]{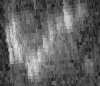}
    \includegraphics[width=0.15\textwidth]{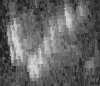}
    \includegraphics[width=0.15\textwidth]{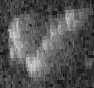}
    \includegraphics[width=0.15\textwidth]{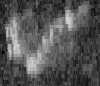}
    \includegraphics[width=0.15\textwidth]{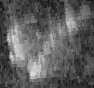}
    
    \includegraphics[width=0.15\textwidth]{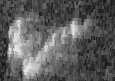}
    \includegraphics[width=0.15\textwidth]{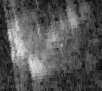}    
    \includegraphics[width=0.15\textwidth]{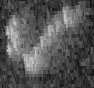}
    \includegraphics[width=0.15\textwidth]{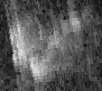}
    \includegraphics[width=0.15\textwidth]{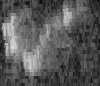}
    \includegraphics[width=0.15\textwidth]{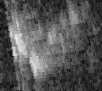}
    
    \includegraphics[width=0.15\textwidth]{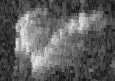}
    \includegraphics[width=0.15\textwidth]{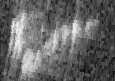}
    \includegraphics[width=0.15\textwidth]{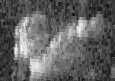}
    \includegraphics[width=0.15\textwidth]{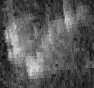}
    \includegraphics[width=0.15\textwidth]{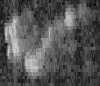}
    \includegraphics[width=0.15\textwidth]{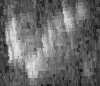}
    
    \caption{Hook Class. 78 Randomly selected image crops}
    \label{appendix:hook}
\end{figure}

\begin{figure}
    \centering
    \includegraphics[width=0.15\textwidth]{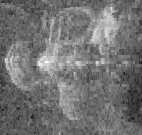}
    \includegraphics[width=0.15\textwidth]{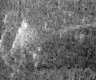}
    \includegraphics[width=0.15\textwidth]{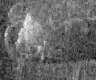}
    \includegraphics[width=0.15\textwidth]{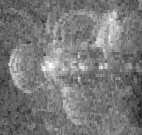}
    \includegraphics[width=0.15\textwidth]{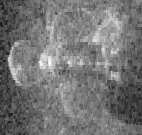}
    \includegraphics[width=0.15\textwidth]{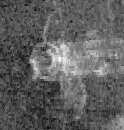}
    
    \includegraphics[width=0.15\textwidth]{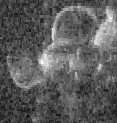}
    \includegraphics[width=0.15\textwidth]{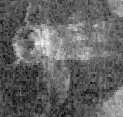}
    \includegraphics[width=0.15\textwidth]{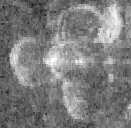}
    \includegraphics[width=0.15\textwidth]{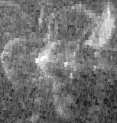}
    \includegraphics[width=0.15\textwidth]{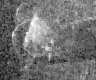}
    \includegraphics[width=0.15\textwidth]{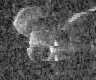}
    
    \includegraphics[width=0.15\textwidth]{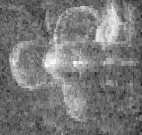}
    \includegraphics[width=0.15\textwidth]{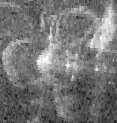}
    \includegraphics[width=0.15\textwidth]{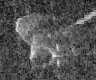}
    \includegraphics[width=0.15\textwidth]{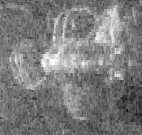}
    \includegraphics[width=0.15\textwidth]{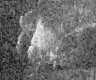}
    \includegraphics[width=0.15\textwidth]{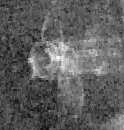}
    
    \includegraphics[width=0.15\textwidth]{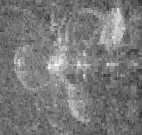}
    \includegraphics[width=0.15\textwidth]{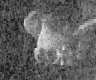}
    \includegraphics[width=0.15\textwidth]{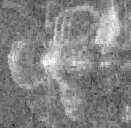}
    \includegraphics[width=0.15\textwidth]{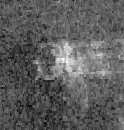}
    \includegraphics[width=0.15\textwidth]{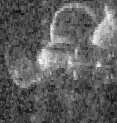}
    \includegraphics[width=0.15\textwidth]{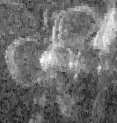}
    
    \includegraphics[width=0.15\textwidth]{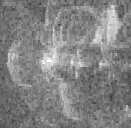}
    \includegraphics[width=0.15\textwidth]{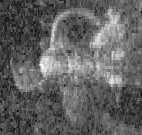}
    \includegraphics[width=0.15\textwidth]{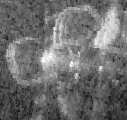}
    \includegraphics[width=0.15\textwidth]{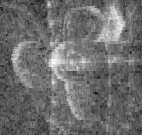}
    \includegraphics[width=0.15\textwidth]{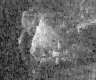}
    \includegraphics[width=0.15\textwidth]{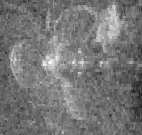}
    
    \includegraphics[width=0.15\textwidth]{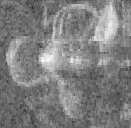}
    \includegraphics[width=0.15\textwidth]{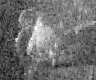}
    \includegraphics[width=0.15\textwidth]{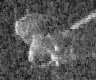}
    \includegraphics[width=0.15\textwidth]{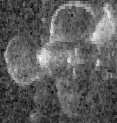}
    \includegraphics[width=0.15\textwidth]{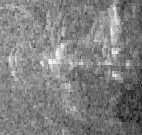}
    \includegraphics[width=0.15\textwidth]{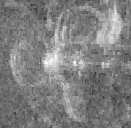}
    
    \includegraphics[width=0.15\textwidth]{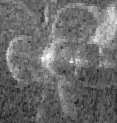}
    \includegraphics[width=0.15\textwidth]{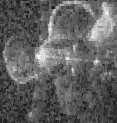}
    \includegraphics[width=0.15\textwidth]{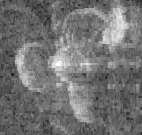}
    \includegraphics[width=0.15\textwidth]{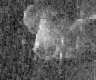}
    \includegraphics[width=0.15\textwidth]{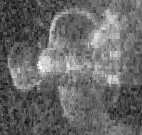}
    \includegraphics[width=0.15\textwidth]{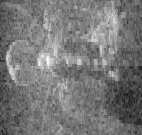}
    
    \includegraphics[width=0.15\textwidth]{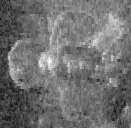}
    \includegraphics[width=0.15\textwidth]{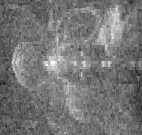}
    \includegraphics[width=0.15\textwidth]{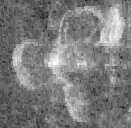}
    \includegraphics[width=0.15\textwidth]{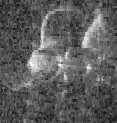}
    \includegraphics[width=0.15\textwidth]{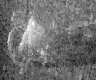}
    \includegraphics[width=0.15\textwidth]{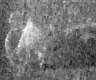}
    
    \includegraphics[width=0.15\textwidth]{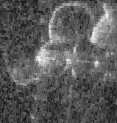}
    \includegraphics[width=0.15\textwidth]{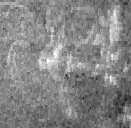}
    \includegraphics[width=0.15\textwidth]{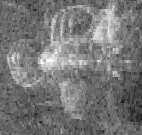}
    \includegraphics[width=0.15\textwidth]{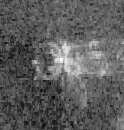}
    \includegraphics[width=0.15\textwidth]{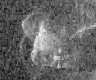}
    \includegraphics[width=0.15\textwidth]{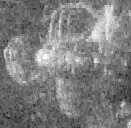}
    
    \includegraphics[width=0.15\textwidth]{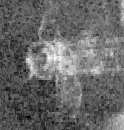}
    \includegraphics[width=0.15\textwidth]{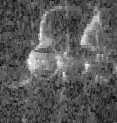}
    \includegraphics[width=0.15\textwidth]{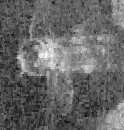}
    \includegraphics[width=0.15\textwidth]{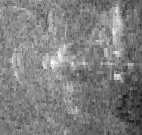}
    \includegraphics[width=0.15\textwidth]{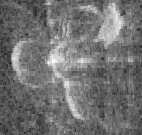}
    \includegraphics[width=0.15\textwidth]{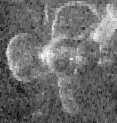}
    
    \includegraphics[width=0.15\textwidth]{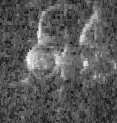}
    \includegraphics[width=0.15\textwidth]{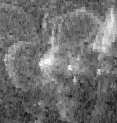}
    \includegraphics[width=0.15\textwidth]{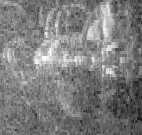}
    \includegraphics[width=0.15\textwidth]{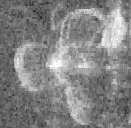}
    \includegraphics[width=0.15\textwidth]{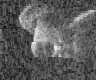}
    \includegraphics[width=0.15\textwidth]{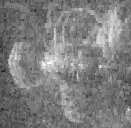}
    
    \includegraphics[width=0.15\textwidth]{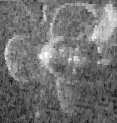}
    \includegraphics[width=0.15\textwidth]{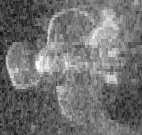}
    \includegraphics[width=0.15\textwidth]{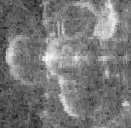}
    \includegraphics[width=0.15\textwidth]{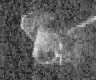}
    \includegraphics[width=0.15\textwidth]{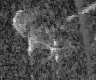}
    \includegraphics[width=0.15\textwidth]{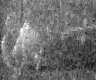}
    
    \includegraphics[width=0.15\textwidth]{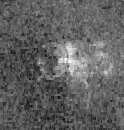}
    \includegraphics[width=0.15\textwidth]{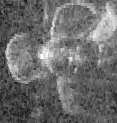}
    \includegraphics[width=0.15\textwidth]{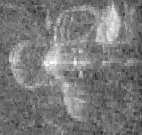}
    \includegraphics[width=0.15\textwidth]{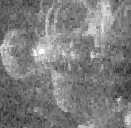}
    \includegraphics[width=0.15\textwidth]{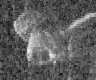}
    \includegraphics[width=0.15\textwidth]{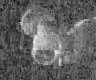}
    
    \caption{Propeller Class. 78 Randomly selected image crops}
    \label{appendix:propeller}
\end{figure}

\begin{figure}
    \centering
    \includegraphics[angle=90,width=0.15\textwidth]{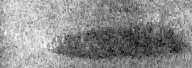}
    \includegraphics[angle=90,width=0.15\textwidth]{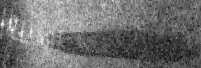}
    \includegraphics[angle=90,width=0.15\textwidth]{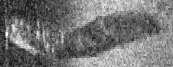}
    \includegraphics[angle=90,width=0.15\textwidth]{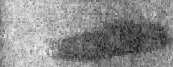}
    \includegraphics[angle=90,width=0.15\textwidth]{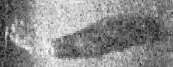}
    \includegraphics[angle=90,width=0.15\textwidth]{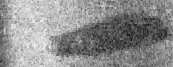}
    
    \includegraphics[angle=90,width=0.15\textwidth]{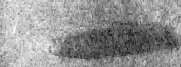}
    \includegraphics[angle=90,width=0.15\textwidth]{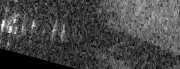}
    \includegraphics[angle=90,width=0.15\textwidth]{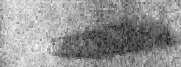}
    \includegraphics[angle=90,width=0.15\textwidth]{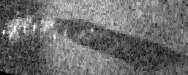}
    \includegraphics[angle=90,width=0.15\textwidth]{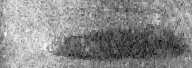}
    \includegraphics[angle=90,width=0.15\textwidth]{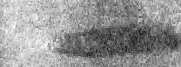}
    
    \includegraphics[angle=90,width=0.15\textwidth]{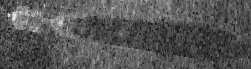}
    \includegraphics[angle=90,width=0.15\textwidth]{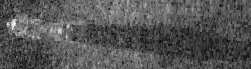}
    \includegraphics[angle=90,width=0.15\textwidth]{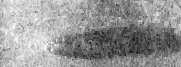}
    \includegraphics[angle=90,width=0.15\textwidth]{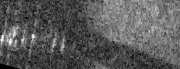}
    \includegraphics[angle=90,width=0.15\textwidth]{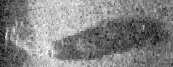}
    \includegraphics[angle=90,width=0.15\textwidth]{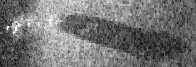}
    
    \includegraphics[angle=90,width=0.15\textwidth]{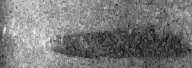}
    \includegraphics[angle=90,width=0.15\textwidth]{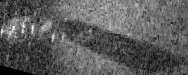}
    \includegraphics[angle=90,width=0.15\textwidth]{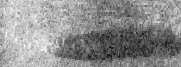}
    \includegraphics[angle=90,width=0.15\textwidth]{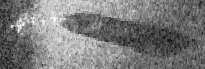}
    \includegraphics[angle=90,width=0.15\textwidth]{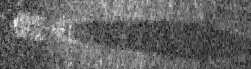}
    \includegraphics[angle=90,width=0.15\textwidth]{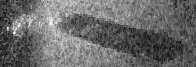}
    
    \includegraphics[angle=90,width=0.15\textwidth]{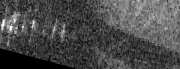}
    \includegraphics[angle=90,width=0.15\textwidth]{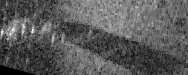}
    \includegraphics[angle=90,width=0.15\textwidth]{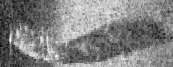}
    \includegraphics[angle=90,width=0.15\textwidth]{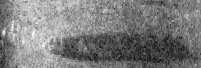}
    \includegraphics[angle=90,width=0.15\textwidth]{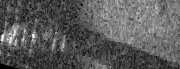}
    \includegraphics[angle=90,width=0.15\textwidth]{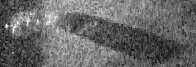}

    \caption{Shampoo Bottle Class. 30 Randomly selected image crops}
    \label{appendix:shampooBottle}    
\end{figure}

\begin{figure}
    \centering
    \includegraphics[angle=90,width=0.15\textwidth]{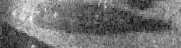}
    \includegraphics[angle=90,width=0.15\textwidth]{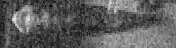}
    \includegraphics[angle=90,width=0.15\textwidth]{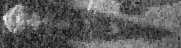}
    \includegraphics[angle=90,width=0.15\textwidth]{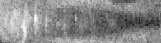}
    \includegraphics[angle=90,width=0.15\textwidth]{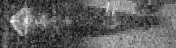}
    \includegraphics[angle=90,width=0.15\textwidth]{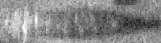}
    
    \includegraphics[angle=90,width=0.15\textwidth]{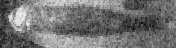}
    \includegraphics[angle=90,width=0.15\textwidth]{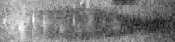}
    \includegraphics[angle=90,width=0.15\textwidth]{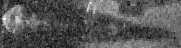}
    \includegraphics[angle=90,width=0.15\textwidth]{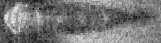}
    \includegraphics[angle=90,width=0.15\textwidth]{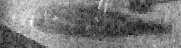}
    \includegraphics[angle=90,width=0.15\textwidth]{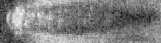}
    
    \includegraphics[angle=90,width=0.15\textwidth]{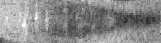}
    \includegraphics[angle=90,width=0.15\textwidth]{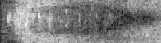}
    \includegraphics[angle=90,width=0.15\textwidth]{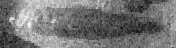}
    \includegraphics[angle=90,width=0.15\textwidth]{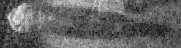}
    \includegraphics[angle=90,width=0.15\textwidth]{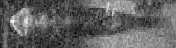}
    \includegraphics[angle=90,width=0.15\textwidth]{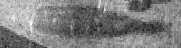}
    
    \includegraphics[angle=90,width=0.15\textwidth]{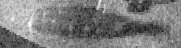}
    \includegraphics[angle=90,width=0.15\textwidth]{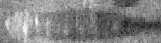}
    \includegraphics[angle=90,width=0.15\textwidth]{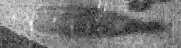}
    \includegraphics[angle=90,width=0.15\textwidth]{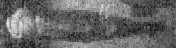}
    \includegraphics[angle=90,width=0.15\textwidth]{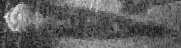}
    \includegraphics[angle=90,width=0.15\textwidth]{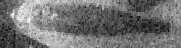}
    \caption{Standing Bottle Class. 24 Randomly selected image crops}
    \label{appendix:standingBottle}    
\end{figure}

\begin{figure}
    \centering
    \includegraphics[width=0.15\textwidth]{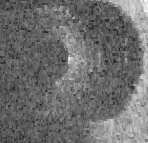}
    \includegraphics[width=0.15\textwidth]{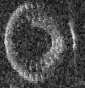}
    \includegraphics[width=0.15\textwidth]{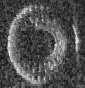}
    \includegraphics[width=0.15\textwidth]{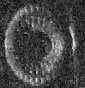}
    \includegraphics[width=0.15\textwidth]{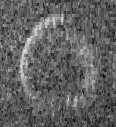}
    \includegraphics[width=0.15\textwidth]{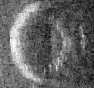}
    
    \includegraphics[width=0.15\textwidth]{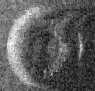}
    \includegraphics[width=0.15\textwidth]{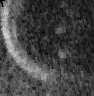}
    \includegraphics[width=0.15\textwidth]{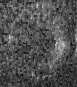}
    \includegraphics[width=0.15\textwidth]{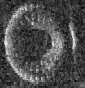}
    \includegraphics[width=0.15\textwidth]{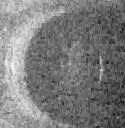}
    \includegraphics[width=0.15\textwidth]{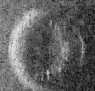}
    
    \includegraphics[width=0.15\textwidth]{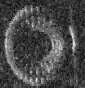}
    \includegraphics[width=0.15\textwidth]{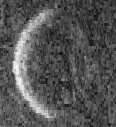}
    \includegraphics[width=0.15\textwidth]{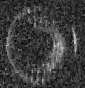}
    \includegraphics[width=0.15\textwidth]{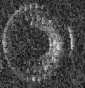}
    \includegraphics[width=0.15\textwidth]{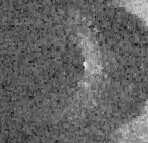}
    \includegraphics[width=0.15\textwidth]{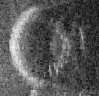}
    
    \includegraphics[width=0.15\textwidth]{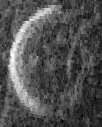}
    \includegraphics[width=0.15\textwidth]{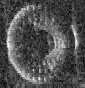}
    \includegraphics[width=0.15\textwidth]{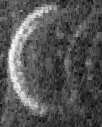}
    \includegraphics[width=0.15\textwidth]{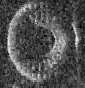}
    \includegraphics[width=0.15\textwidth]{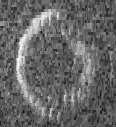}
    \includegraphics[width=0.15\textwidth]{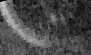}
    
    \includegraphics[width=0.15\textwidth]{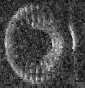}
    \includegraphics[width=0.15\textwidth]{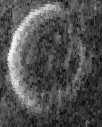}
    \includegraphics[width=0.15\textwidth]{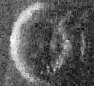}
    \includegraphics[width=0.15\textwidth]{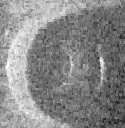}
    \includegraphics[width=0.15\textwidth]{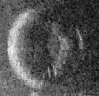}
    \includegraphics[width=0.15\textwidth]{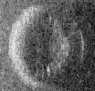}
    
    \includegraphics[width=0.15\textwidth]{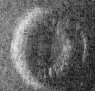}
    \includegraphics[width=0.15\textwidth]{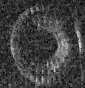}
    \includegraphics[width=0.15\textwidth]{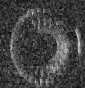}
    \includegraphics[width=0.15\textwidth]{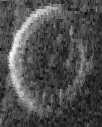}
    \includegraphics[width=0.15\textwidth]{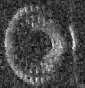}
    \includegraphics[width=0.15\textwidth]{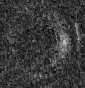}
    
    \includegraphics[width=0.15\textwidth]{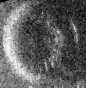}
    \includegraphics[width=0.15\textwidth]{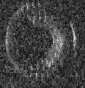}
    \includegraphics[width=0.15\textwidth]{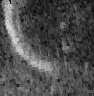}
    \includegraphics[width=0.15\textwidth]{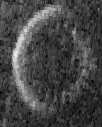}
    \includegraphics[width=0.15\textwidth]{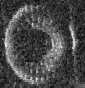}
    \includegraphics[width=0.15\textwidth]{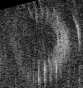}
    
    \includegraphics[width=0.15\textwidth]{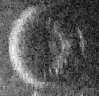}
    \includegraphics[width=0.15\textwidth]{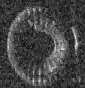}
    \includegraphics[width=0.15\textwidth]{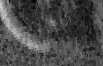}
    \includegraphics[width=0.15\textwidth]{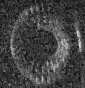}
    \includegraphics[width=0.15\textwidth]{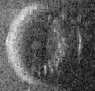}
    \includegraphics[width=0.15\textwidth]{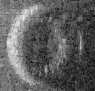}
    
    \includegraphics[width=0.15\textwidth]{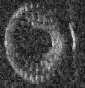}
    \includegraphics[width=0.15\textwidth]{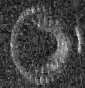}
    \includegraphics[width=0.15\textwidth]{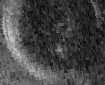}
    \includegraphics[width=0.15\textwidth]{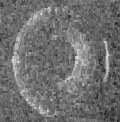}
    \includegraphics[width=0.15\textwidth]{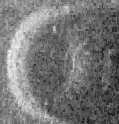}
    \includegraphics[width=0.15\textwidth]{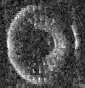}
    
    \includegraphics[width=0.15\textwidth]{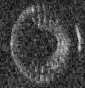}
    \includegraphics[width=0.15\textwidth]{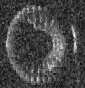}
    \includegraphics[width=0.15\textwidth]{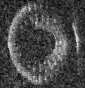}
    \includegraphics[width=0.15\textwidth]{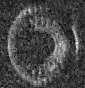}
    \includegraphics[width=0.15\textwidth]{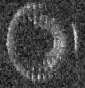}
    \includegraphics[width=0.15\textwidth]{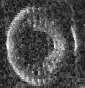}
    
    \includegraphics[width=0.15\textwidth]{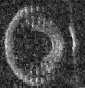}
    \includegraphics[width=0.15\textwidth]{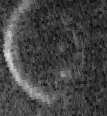}
    \includegraphics[width=0.15\textwidth]{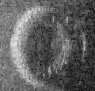}
    \includegraphics[width=0.15\textwidth]{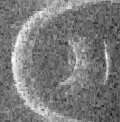}
    \includegraphics[width=0.15\textwidth]{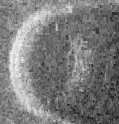}
    \includegraphics[width=0.15\textwidth]{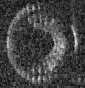}
    
    \includegraphics[width=0.15\textwidth]{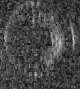}
    \includegraphics[width=0.15\textwidth]{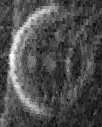}
    \includegraphics[width=0.15\textwidth]{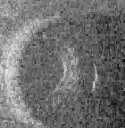}
    \includegraphics[width=0.15\textwidth]{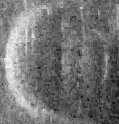}
    \includegraphics[width=0.15\textwidth]{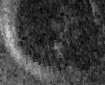}
    \includegraphics[width=0.15\textwidth]{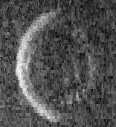}
    
    \includegraphics[width=0.15\textwidth]{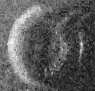}
    \includegraphics[width=0.15\textwidth]{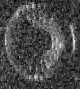}
    \includegraphics[width=0.15\textwidth]{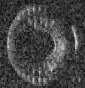}
    \includegraphics[width=0.15\textwidth]{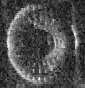}
    \includegraphics[width=0.15\textwidth]{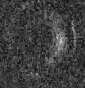}
    \includegraphics[width=0.15\textwidth]{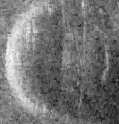}
    
    \caption{Tire Class. 78 Randomly selected image crops}
    \label{appendix:tire}
\end{figure}

\begin{figure}
    \centering
    
    \includegraphics[width=0.15\textwidth]{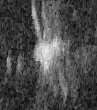}
    \includegraphics[width=0.15\textwidth]{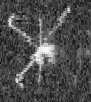}
    \includegraphics[width=0.15\textwidth]{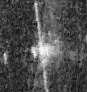}
    \includegraphics[width=0.15\textwidth]{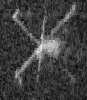}
    \includegraphics[width=0.15\textwidth]{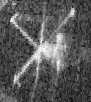}
    \includegraphics[width=0.15\textwidth]{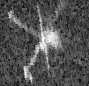}
    
    \includegraphics[width=0.15\textwidth]{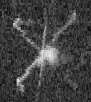}
    \includegraphics[width=0.15\textwidth]{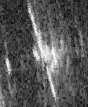}
    \includegraphics[width=0.15\textwidth]{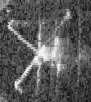}
    \includegraphics[width=0.15\textwidth]{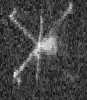}
    \includegraphics[width=0.15\textwidth]{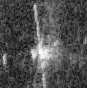}
    \includegraphics[width=0.15\textwidth]{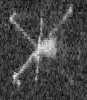}
    
    \includegraphics[width=0.15\textwidth]{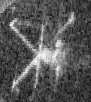}
    \includegraphics[width=0.15\textwidth]{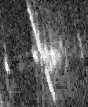}
    \includegraphics[width=0.15\textwidth]{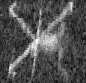}
    \includegraphics[width=0.15\textwidth]{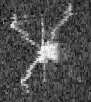}
    \includegraphics[width=0.15\textwidth]{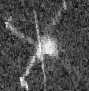}
    \includegraphics[width=0.15\textwidth]{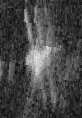}
    
    \includegraphics[width=0.15\textwidth]{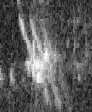}
    \includegraphics[width=0.15\textwidth]{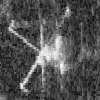}
    \includegraphics[width=0.15\textwidth]{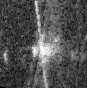}
    \includegraphics[width=0.15\textwidth]{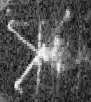}
    \includegraphics[width=0.15\textwidth]{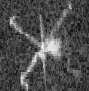}
    \includegraphics[width=0.15\textwidth]{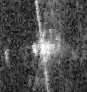}
    
    \includegraphics[width=0.15\textwidth]{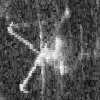}
    \includegraphics[width=0.15\textwidth]{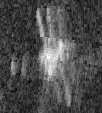}
    \includegraphics[width=0.15\textwidth]{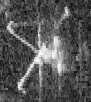}
    \includegraphics[width=0.15\textwidth]{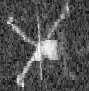}
    \includegraphics[width=0.15\textwidth]{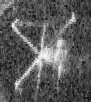}
    \includegraphics[width=0.15\textwidth]{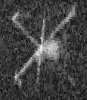}
    
    \includegraphics[width=0.15\textwidth]{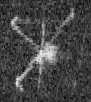}
    \includegraphics[width=0.15\textwidth]{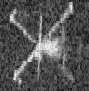}
    \includegraphics[width=0.15\textwidth]{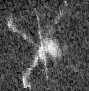}
    \includegraphics[width=0.15\textwidth]{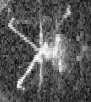}
    \includegraphics[width=0.15\textwidth]{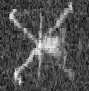}
    \includegraphics[width=0.15\textwidth]{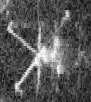}
    
    \includegraphics[width=0.15\textwidth]{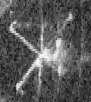}
    \includegraphics[width=0.15\textwidth]{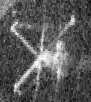}
    \includegraphics[width=0.15\textwidth]{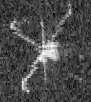}
    \includegraphics[width=0.15\textwidth]{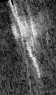}
    \includegraphics[width=0.15\textwidth]{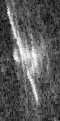}
    \includegraphics[width=0.15\textwidth]{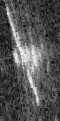}    
    
    \includegraphics[width=0.15\textwidth]{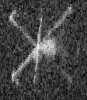}
    \includegraphics[width=0.15\textwidth]{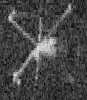}
    \includegraphics[width=0.15\textwidth]{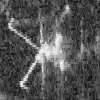}
    \includegraphics[width=0.15\textwidth]{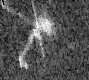}
    \includegraphics[width=0.15\textwidth]{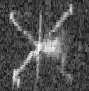}
    \includegraphics[width=0.15\textwidth]{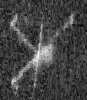}
    
    \includegraphics[width=0.15\textwidth]{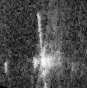}
    \includegraphics[width=0.15\textwidth]{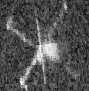}
    \includegraphics[width=0.15\textwidth]{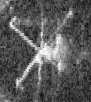}
    \includegraphics[width=0.15\textwidth]{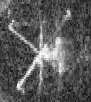}
    \includegraphics[width=0.15\textwidth]{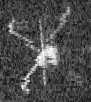}
    \includegraphics[width=0.15\textwidth]{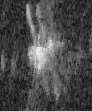}
    
    \includegraphics[width=0.15\textwidth]{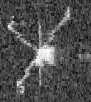}
    \includegraphics[width=0.15\textwidth]{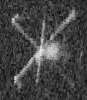}
    \includegraphics[width=0.15\textwidth]{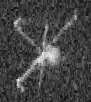}
    \includegraphics[width=0.15\textwidth]{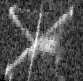}    
    \includegraphics[width=0.15\textwidth]{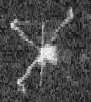}
    \includegraphics[width=0.15\textwidth]{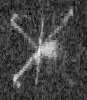}
    
    \includegraphics[width=0.15\textwidth]{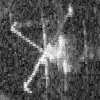}
    \includegraphics[width=0.15\textwidth]{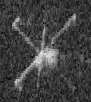}
    \includegraphics[width=0.15\textwidth]{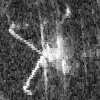}
    \includegraphics[width=0.15\textwidth]{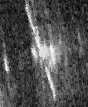}
    \includegraphics[width=0.15\textwidth]{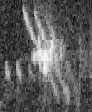}
    \includegraphics[width=0.15\textwidth]{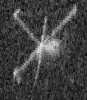}
    
    \includegraphics[width=0.15\textwidth]{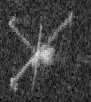}
    \includegraphics[width=0.15\textwidth]{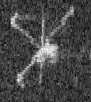}
    \includegraphics[width=0.15\textwidth]{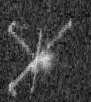}
    \includegraphics[width=0.15\textwidth]{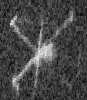}
    \includegraphics[width=0.15\textwidth]{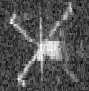}
    \includegraphics[width=0.15\textwidth]{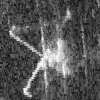}
    
    \caption{Valve Class. 72 Randomly selected image crops}
    \label{appendix:valve}  
\end{figure}

\begin{figure}
    \centering
    \includegraphics[width=0.15\textwidth]{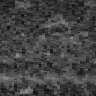}
    \includegraphics[width=0.15\textwidth]{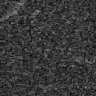}
    \includegraphics[width=0.15\textwidth]{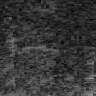}
    \includegraphics[width=0.15\textwidth]{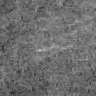}
    \includegraphics[width=0.15\textwidth]{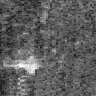}
    \includegraphics[width=0.15\textwidth]{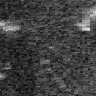}
    
    \includegraphics[width=0.15\textwidth]{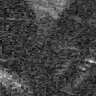}
    \includegraphics[width=0.15\textwidth]{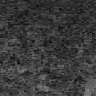}
    \includegraphics[width=0.15\textwidth]{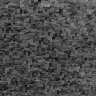}
    \includegraphics[width=0.15\textwidth]{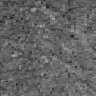}
    \includegraphics[width=0.15\textwidth]{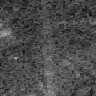}
    \includegraphics[width=0.15\textwidth]{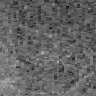}
    
    \includegraphics[width=0.15\textwidth]{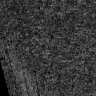}
    \includegraphics[width=0.15\textwidth]{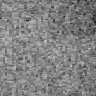}
    \includegraphics[width=0.15\textwidth]{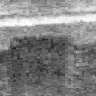}
    \includegraphics[width=0.15\textwidth]{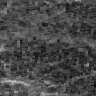}
    \includegraphics[width=0.15\textwidth]{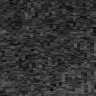}
    \includegraphics[width=0.15\textwidth]{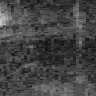}
    
    \includegraphics[width=0.15\textwidth]{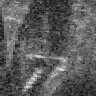}
    \includegraphics[width=0.15\textwidth]{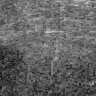}
    \includegraphics[width=0.15\textwidth]{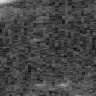}
    \includegraphics[width=0.15\textwidth]{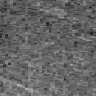}
    \includegraphics[width=0.15\textwidth]{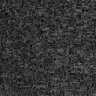}
    \includegraphics[width=0.15\textwidth]{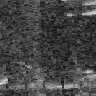}
    
    \includegraphics[width=0.15\textwidth]{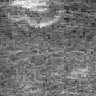}
    \includegraphics[width=0.15\textwidth]{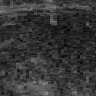}
    \includegraphics[width=0.15\textwidth]{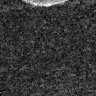}
    \includegraphics[width=0.15\textwidth]{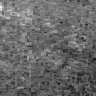}
    \includegraphics[width=0.15\textwidth]{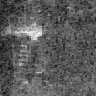}
    \includegraphics[width=0.15\textwidth]{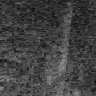}
    
    \includegraphics[width=0.15\textwidth]{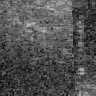}
    \includegraphics[width=0.15\textwidth]{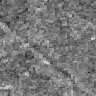}
    \includegraphics[width=0.15\textwidth]{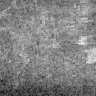}
    \includegraphics[width=0.15\textwidth]{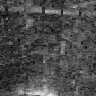}
    \includegraphics[width=0.15\textwidth]{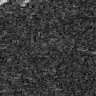}
    \includegraphics[width=0.15\textwidth]{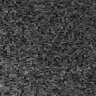}
    
    \includegraphics[width=0.15\textwidth]{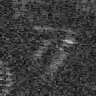}
    \includegraphics[width=0.15\textwidth]{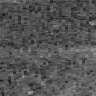}
    \includegraphics[width=0.15\textwidth]{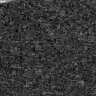}
    \includegraphics[width=0.15\textwidth]{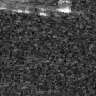}
    \includegraphics[width=0.15\textwidth]{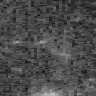}
    \includegraphics[width=0.15\textwidth]{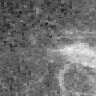}
    
    \includegraphics[width=0.15\textwidth]{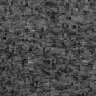}
    \includegraphics[width=0.15\textwidth]{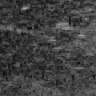}
    \includegraphics[width=0.15\textwidth]{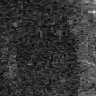}
    \includegraphics[width=0.15\textwidth]{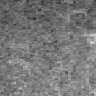}
    \includegraphics[width=0.15\textwidth]{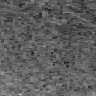}
    \includegraphics[width=0.15\textwidth]{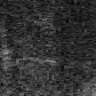}
    
    \includegraphics[width=0.15\textwidth]{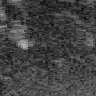}
    \includegraphics[width=0.15\textwidth]{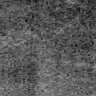}
    \includegraphics[width=0.15\textwidth]{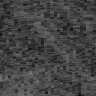}
    \includegraphics[width=0.15\textwidth]{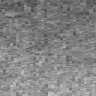}
    \includegraphics[width=0.15\textwidth]{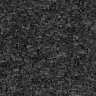}
    \includegraphics[width=0.15\textwidth]{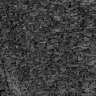}
    
    \includegraphics[width=0.15\textwidth]{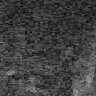}
    \includegraphics[width=0.15\textwidth]{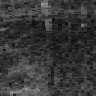}
    \includegraphics[width=0.15\textwidth]{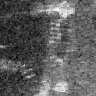}
    \includegraphics[width=0.15\textwidth]{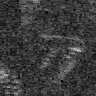}
    \includegraphics[width=0.15\textwidth]{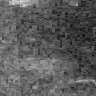}
    \includegraphics[width=0.15\textwidth]{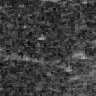}
    
    \includegraphics[width=0.15\textwidth]{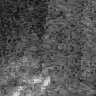}
    \includegraphics[width=0.15\textwidth]{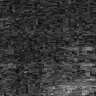}
    \includegraphics[width=0.15\textwidth]{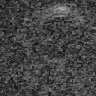}
    \includegraphics[width=0.15\textwidth]{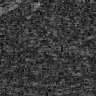}
    \includegraphics[width=0.15\textwidth]{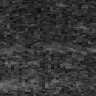}
    \includegraphics[width=0.15\textwidth]{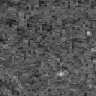}
    
    \includegraphics[width=0.15\textwidth]{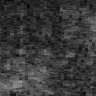}
    \includegraphics[width=0.15\textwidth]{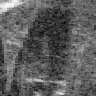}
    \includegraphics[width=0.15\textwidth]{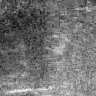}
    \includegraphics[width=0.15\textwidth]{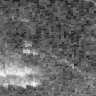}
    \includegraphics[width=0.15\textwidth]{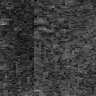}
    \includegraphics[width=0.15\textwidth]{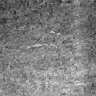}
    
    \includegraphics[width=0.15\textwidth]{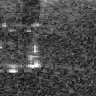}
    \includegraphics[width=0.15\textwidth]{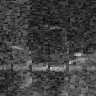}
    \includegraphics[width=0.15\textwidth]{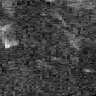}
    \includegraphics[width=0.15\textwidth]{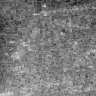}
    \includegraphics[width=0.15\textwidth]{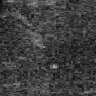}
    \includegraphics[width=0.15\textwidth]{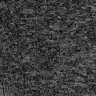}
    
    \caption{Background Class, 78 Randomly selected $96 \times 96$ image crops}
    \label{appendix:background}
\end{figure}

%% file: deep-neural-networks-marine-debris-detection-oneside.bbl
\begin{thebibliography}{100}

\bibitem{aanaes2012interesting}
Henrik Aan{\ae}s, Anders~Lindbjerg Dahl, and Kim~Steenstrup Pedersen.
\newblock Interesting interest points.
\newblock {\em International Journal of Computer Vision}, 97(1):18--35, 2012.

\bibitem{alexe2012measuring}
Bogdan Alexe, Thomas Deselaers, and Vittorio Ferrari.
\newblock Measuring the objectness of image windows.
\newblock {\em Pattern Analysis and Machine Intelligence, IEEE Transactions
  on}, 34(11):2189--2202, 2012.

\bibitem{ba2016layer}
Jimmy~Lei Ba, Jamie~Ryan Kiros, and Geoffrey~E Hinton.
\newblock Layer normalization.
\newblock {\em arXiv preprint arXiv:1607.06450}, 2016.

\bibitem{barngrover2015semisynthetic}
Christopher Barngrover, Ryan Kastner, and Serge Belongie.
\newblock Semisynthetic versus real-world sonar training data for the
  classification of mine-like objects.
\newblock {\em IEEE Journal of Oceanic Engineering}, 40(1):48--56, 2015.

\bibitem{bay2006surf}
Herbert Bay, Tinne Tuytelaars, and Luc Van~Gool.
\newblock Surf: Speeded up robust features.
\newblock In {\em European conference on computer vision}, pages 404--417.
  Springer, 2006.

\bibitem{baydin2017automatic}
Atilim~Gunes Baydin, Barak~A Pearlmutter, Alexey~Andreyevich Radul, and
  Jeffrey~Mark Siskind.
\newblock Automatic differentiation in machine learning: a survey.
\newblock {\em Journal of machine learning research}, 18(153):1--153, 2017.

\bibitem{beduhn1966removal}
Georg~E Beduhn.
\newblock Removal of oil and debris from harbor waters.
\newblock Technical report, DTIC Document, 1966.

\bibitem{belcher2002dual}
Edward Belcher, William Hanot, and Joe Burch.
\newblock Dual-frequency identification sonar (didson).
\newblock In {\em Underwater Technology, 2002. Proceedings of the 2002
  International Symposium on}, pages 187--192. IEEE, 2002.

\bibitem{belcher1999beamforming}
Edward Belcher, Dana Lynn, Hien Dinh, and Thomas Laughlin.
\newblock Beamforming and imaging with acoustic lenses in small, high-frequency
  sonars.
\newblock In {\em OCEANS'99 MTS/IEEE.}, volume~3, pages 1495--1499. IEEE, 1999.

\bibitem{belcher2001object}
Edward Belcher, Brian Matsuyama, and Gary Trimble.
\newblock Object identification with acoustic lenses.
\newblock In {\em OCEANS 2001 MTS/IEEE}, volume~1, pages 6--11. IEEE, 2001.

\bibitem{bergstra2012random}
James Bergstra and Yoshua Bengio.
\newblock Random search for hyper-parameter optimization.
\newblock {\em Journal of Machine Learning Research}, 13(Feb):281--305, 2012.

\bibitem{bishop2006pattern}
Christopher Bishop.
\newblock {\em Pattern Recognition and Machine Learning}.
\newblock Springer, 2006.

\bibitem{bromley1994signature}
Jane Bromley, Isabelle Guyon, Yann LeCun, Eduard S{\"a}ckinger, and Roopak
  Shah.
\newblock Signature verification using a" siamese" time delay neural network.
\newblock In {\em Advances in Neural Information Processing Systems}, pages
  737--744, 1994.

\bibitem{brown2011discriminative}
Matthew Brown, Gang Hua, and Simon Winder.
\newblock Discriminative learning of local image descriptors.
\newblock {\em IEEE transactions on pattern analysis and machine intelligence},
  33(1):43--57, 2011.

\bibitem{buss2018hand}
Matthias Bu{\ss}, Yannik Steiniger, Stephan Benen, Dieter Kraus, Anton Kummert,
  and Dietmar Stiller.
\newblock Hand-crafted feature based classification against convolutional
  neural networks for false alarm reduction on active diver detection sonar
  data.
\newblock In {\em OCEANS 2018 MTS/IEEE Charleston}, pages 1--7. IEEE, 2018.

\bibitem{Cadena16tro-SLAMfuture}
C.~Cadena, L.~Carlone, H.~Carrillo, Y.~Latif, D.~Scaramuzza, J.~Neira, I.~Reid,
  and J.J. Leonard.
\newblock Past, present, and future of simultaneous localization and mapping:
  Towards the robust-perception age.
\newblock {\em {IEEE Transactions on Robotics}}, 32(6):1309–1332, 2016.

\bibitem{caruana1997multitaskJournalML}
Rich Caruana.
\newblock Multitask learning.
\newblock {\em Machine Learning}, 28(1):41--75, Jul 1997.

\bibitem{caruana1997multitask}
Rich Caruana.
\newblock {\em Multitask Learning}.
\newblock PhD thesis, Carnegie Mellon University Pittsburgh, PA, 1997.

\bibitem{vcehovin2016visual}
Luka {\v{C}}ehovin, Ale{\v{s}} Leonardis, and Matej Kristan.
\newblock Visual object tracking performance measures revisited.
\newblock {\em IEEE Transactions on Image Processing}, 25(3):1261--1274, 2016.

\bibitem{chavali2016object}
Neelima Chavali, Harsh Agrawal, Aroma Mahendru, and Dhruv Batra.
\newblock Object-proposal evaluation protocol is' gameable'.
\newblock In {\em Proceedings of the IEEE Conference on Computer Vision and
  Pattern Recognition}, pages 835--844, 2016.

\bibitem{cheng2014bing}
Ming-Ming Cheng, Ziming Zhang, Wen-Yan Lin, and Philip Torr.
\newblock Bing: Binarized normed gradients for objectness estimation at 300fps.
\newblock In {\em Proceedings of the IEEE conference on computer vision and
  pattern recognition}, pages 3286--3293, 2014.

\bibitem{chiba2018human}
Sanae Chiba, Hideaki Saito, Ruth Fletcher, Takayuki Yogi, Makino Kayo, Shin
  Miyagi, Moritaka Ogido, and Katsunori Fujikura.
\newblock Human footprint in the abyss: 30 year records of deep-sea plastic
  debris.
\newblock {\em Marine Policy}, 2018.

\bibitem{dayton1995environmental}
Paul~K Dayton, Simon~F Thrush, M~Tundi Agardy, and Robert~J Hofman.
\newblock Environmental effects of marine fishing.
\newblock {\em Aquatic conservation: marine and freshwater ecosystems},
  5(3):205--232, 1995.

\bibitem{de2011multidimensional}
Jan De~Leeuw and Patrick Mair.
\newblock Multidimensional scaling using majorization: Smacof in r.
\newblock {\em Department of Statistics, UCLA}, 2011.

\bibitem{dmitrieva2017object}
Mariia Dmitrieva, Matias Valdenegro-Toro, Keith Brown, Gary Heald, and David
  Lane.
\newblock {O}bject classification with convolution neural network based on the
  time-frequency representation of their echo.
\newblock In {\em Machine Learning for Signal Processing (MLSP), 2017 IEEE 27th
  International Workshop on}, pages 1--6. IEEE, 2017.

\bibitem{duchi2011adaptive}
John Duchi, Elad Hazan, and Yoram Singer.
\newblock Adaptive subgradient methods for online learning and stochastic
  optimization.
\newblock {\em Journal of Machine Learning Research}, 12(Jul):2121--2159, 2011.

\bibitem{dura2011image}
Esther Dura.
\newblock Image processing techniques for the detection and classification of
  man made objects in side-scan sonar images.
\newblock In {\em Sonar Systems}. InTech, 2011.

\bibitem{endres2010category}
Ian Endres and Derek Hoiem.
\newblock Category independent object proposals.
\newblock In {\em Computer Vision--ECCV 2010}, pages 575--588. Springer, 2010.

\bibitem{engler2012complex}
Richard~E Engler.
\newblock The complex interaction between marine debris and toxic chemicals in
  the ocean.
\newblock {\em Environmental science \& technology}, 46(22):12302--12315, 2012.

\bibitem{erhan2014scalable}
Dumitru Erhan, Christian Szegedy, Alexander Toshev, and Dragomir Anguelov.
\newblock Scalable object detection using deep neural networks.
\newblock In {\em Proceedings of the IEEE Conference on Computer Vision and
  Pattern Recognition}, pages 2147--2154, 2014.

\bibitem{eriksen2014plastic}
Marcus Eriksen, Laurent~CM Lebreton, Henry~S Carson, Martin Thiel, Charles~J
  Moore, Jose~C Borerro, Francois Galgani, Peter~G Ryan, and Julia Reisser.
\newblock Plastic pollution in the world's oceans: more than 5 trillion plastic
  pieces weighing over 250,000 tons afloat at sea.
\newblock {\em PloS one}, 9(12):e111913, 2014.

\bibitem{everingham2015pascal}
Mark Everingham, SM~Ali Eslami, Luc Van~Gool, Christopher~KI Williams, John
  Winn, and Andrew Zisserman.
\newblock The pascal visual object classes challenge: A retrospective.
\newblock {\em International journal of computer vision}, 111(1):98--136, 2015.

\bibitem{everingham2010pascal}
Mark Everingham, Luc Van~Gool, Christopher~KI Williams, John Winn, and Andrew
  Zisserman.
\newblock The pascal visual object classes (voc) challenge.
\newblock {\em International journal of computer vision}, 88(2):303--338, 2010.

\bibitem{fandos2012sparse}
Raquel Fandos, Leyna Sadamori, and Abdelhak~M Zoubir.
\newblock Sparse representation based classification for mine hunting using
  synthetic aperture sonar.
\newblock In {\em Acoustics, Speech and Signal Processing (ICASSP), 2012 IEEE
  International Conference on}, pages 3393--3396. IEEE, 2012.

\bibitem{fandos2011optimal}
Raquel Fandos and Abdelhak~M Zoubir.
\newblock Optimal feature set for automatic detection and classification of
  underwater objects in sas images.
\newblock {\em IEEE Journal of Selected Topics in Signal Processing},
  5(3):454--468, 2011.

\bibitem{fawcett2007computer}
J~Fawcett, A~Crawford, D~Hopkin, M~Couillard, V~Myers, and Benoit Zerr.
\newblock Computer-aided classification of the citadel trial sidescan sonar
  images.
\newblock {\em Defence Research and Development Canada Atlantic TM}, 162:2007,
  2007.

\bibitem{ferreira2014improving}
Fausto Ferreira, Vladimir Djapic, Michele Micheli, and Massimo Caccia.
\newblock Improving automatic target recognition with forward looking sonar
  mosaics.
\newblock {\em IFAC Proceedings Volumes}, 47(3):3382--3387, 2014.

\bibitem{fink1994computer}
Kevin Fink.
\newblock Computer simulation of pressure fields generated by acoustic lens
  beamformers.
\newblock Master's thesis, University of Washington, 1994.

\bibitem{fossen2011handbook}
Thor~I Fossen.
\newblock {\em Handbook of marine craft hydrodynamics and motion control}.
\newblock John Wiley \& Sons, 2011.

\bibitem{gal2016uncertainty}
Yarin Gal.
\newblock {\em Uncertainty in deep learning}.
\newblock PhD thesis, University of Cambridge, 2016.

\bibitem{gal2015dropout}
Yarin Gal and Zoubin Ghahramani.
\newblock Dropout as a bayesian approximation: Representing model uncertainty
  in deep learning.
\newblock {\em arXiv preprint arXiv:1506.02142}, 2, 2015.

\bibitem{gall2015impact}
Sarah~C Gall and Richard~C Thompson.
\newblock The impact of debris on marine life.
\newblock {\em Marine pollution bulletin}, 92(1-2):170--179, 2015.

\bibitem{geilhufe2014quantifying}
M~Geilhufe and {\O}~Midtgaard.
\newblock Quantifying the complexity in sonar images for mcm performance
  estimation.
\newblock In {\em Proc. 2nd International Conference and Exhibition on
  Underwater Acoustics}, pages 1041--1048, 2014.

\bibitem{girshick2015fast}
Ross Girshick.
\newblock Fast r-cnn.
\newblock In {\em Proceedings of the IEEE international conference on computer
  vision}, pages 1440--1448, 2015.

\bibitem{glorot2010understanding}
Xavier Glorot and Yoshua Bengio.
\newblock Understanding the difficulty of training deep feedforward neural
  networks.
\newblock In {\em Proceedings of AISTATS'10}, pages 249--256, 2010.

\bibitem{glorot2011deep}
Xavier Glorot, Antoine Bordes, and Yoshua Bengio.
\newblock Deep sparse rectifier neural networks.
\newblock In {\em Proceedings of AISTATS'11}, volume~15, 2011.

\bibitem{gonzalezDIP2006}
Rafael~C. Gonzalez and Richard~E. Woods.
\newblock {\em Digital Image Processing (3rd Edition)}.
\newblock Prentice-Hall, Inc., Upper Saddle River, NJ, USA, 2006.

\bibitem{Goodfellow2016deep}
Ian Goodfellow, Yoshua Bengio, and Aaron Courville.
\newblock {\em Deep Learning}.
\newblock MIT Press, 2016.
\newblock \url{http://www.deeplearningbook.org}.

\bibitem{hansen2009introduction}
Roy~Edgar Hansen.
\newblock Introduction to sonar.
\newblock {\em Course Material to INF-GEO4310, University of Oslo,(Oct. 7,
  2009)}, 2009.

\bibitem{he2014spatial}
Kaiming He, Xiangyu Zhang, Shaoqing Ren, and Jian Sun.
\newblock Spatial pyramid pooling in deep convolutional networks for visual
  recognition.
\newblock In {\em European Conference on Computer Vision}, pages 346--361.
  Springer, 2014.

\bibitem{he2016deep}
Kaiming He, Xiangyu Zhang, Shaoqing Ren, and Jian Sun.
\newblock Deep residual learning for image recognition.
\newblock In {\em Proceedings of IEEE CVPR}, pages 770--778, 2016.

\bibitem{held2016learning}
David Held, Sebastian Thrun, and Silvio Savarese.
\newblock Learning to track at 100 fps with deep regression networks.
\newblock In {\em European Conference on Computer Vision}, pages 749--765.
  Springer, 2016.

\bibitem{hidalgo2015review}
Franco Hidalgo and Thomas Br{\"a}unl.
\newblock Review of underwater slam techniques.
\newblock In {\em Automation, Robotics and Applications (ICARA), 2015 6th
  International Conference on}, pages 306--311. IEEE, 2015.

\bibitem{hosang2016makes}
Jan Hosang, Rodrigo Benenson, Piotr Doll{\'a}r, and Bernt Schiele.
\newblock What makes for effective detection proposals?
\newblock {\em IEEE transactions on pattern analysis and machine intelligence},
  38(4):814--830, 2016.

\bibitem{hosang2014good}
Jan Hosang, Rodrigo Benenson, and Bernt Schiele.
\newblock How good are detection proposals, really?
\newblock {\em arXiv preprint arXiv:1406.6962}, 2014.

\bibitem{hurtos2014real}
Natalia Hurt{\'o}s, Sharad Nagappa, Narcis Palomeras, and Joaquim Salvi.
\newblock Real-time mosaicing with two-dimensional forward-looking sonar.
\newblock In {\em 2014 IEEE International Conference on Robotics and Automation
  (ICRA)}, pages 601--606. IEEE, 2014.

\bibitem{hurtos2013automatic}
Natalia Hurt{\'o}s, Narcis Palomeras, Sharad Nagappa, and Joaquim Salvi.
\newblock Automatic detection of underwater chain links using a forward-looking
  sonar.
\newblock In {\em OCEANS-Bergen, 2013 MTS/IEEE}, pages 1--7. IEEE, 2013.

\bibitem{iandola2016squeezenet}
Forrest~N Iandola, Song Han, Matthew~W Moskewicz, Khalid Ashraf, William~J
  Dally, and Kurt Keutzer.
\newblock Squeezenet: Alexnet-level accuracy with 50x fewer parameters and< 0.5
  mb model size.
\newblock {\em arXiv preprint arXiv:1602.07360}, 2016.

\bibitem{iniguez2016marine}
Mar{\'\i}a~Esperanza I{\~n}iguez, Juan~A Conesa, and Andres Fullana.
\newblock Marine debris occurrence and treatment: A review.
\newblock {\em Renewable and Sustainable Energy Reviews}, 64:394--402, 2016.

\bibitem{ioffe2015batch}
Sergey Ioffe and Christian Szegedy.
\newblock Batch normalization: Accelerating deep network training by reducing
  internal covariate shift.
\newblock {\em arXiv preprint arXiv:1502.03167}, 2015.

\bibitem{jaderberg2015spatial}
Max Jaderberg, Karen Simonyan, Andrew Zisserman, et~al.
\newblock Spatial transformer networks.
\newblock In {\em Advances in Neural Information Processing Systems}, pages
  2017--2025, 2015.

\bibitem{jambeck2015plastic}
Jenna~R Jambeck, Roland Geyer, Chris Wilcox, Theodore~R Siegler, Miriam
  Perryman, Anthony Andrady, Ramani Narayan, and Kara~Lavender Law.
\newblock Plastic waste inputs from land into the ocean.
\newblock {\em Science}, 347(6223):768--771, 2015.

\bibitem{kader2015design}
ASA Kader, MKM Saleh, MR~Jalal, OO~Sulaiman, and WNW Shamsuri.
\newblock Design of rubbish collecting system for inland waterways.
\newblock {\em Journal of Transport System Engineering}, 2(2):1--13, 2015.

\bibitem{kamgar1998underwater}
B~Kamgar-Parsi, LJ~Rosenblum, and EO~Belcher.
\newblock Underwater imaging with a moving acoustic lens.
\newblock {\em IEEE Transactions on Image Processing}, 7(1):91--99, 1998.

\bibitem{katsanevakis2008marine}
Stelios Katsanevakis.
\newblock Marine debris, a growing problem: Sources, distribution, composition,
  and impacts.
\newblock {\em Marine Pollution: New Research. Nova Science Publishers, New
  York}, pages 53--100, 2008.

\bibitem{kendall2017multi}
Alex Kendall, Yarin Gal, and Roberto Cipolla.
\newblock Multi-task learning using uncertainty to weigh losses for scene
  geometry and semantics.
\newblock {\em arXiv preprint arXiv:1705.07115}, 2017.

\bibitem{kim2005mosaicing}
K~Kim, N~Neretti, and N~Intrator.
\newblock Mosaicing of acoustic camera images.
\newblock {\em IEE Proceedings-Radar, Sonar and Navigation}, 152(4):263--270,
  2005.

\bibitem{kingma2014adam}
Diederik Kingma and Jimmy Ba.
\newblock Adam: A method for stochastic optimization.
\newblock {\em arXiv preprint arXiv:1412.6980}, 2014.

\bibitem{kirkwood2007development}
WJ~Kirkwood.
\newblock Development of the dorado mapping vehicle for multibeam, subbottom,
  and sidescan science missions.
\newblock {\em Journal of Field Robotics}, 24(6):487--495, 2007.

\bibitem{kohntopp2017seafloor}
Daniel K\"ohntopp, Benjamin Lehmann, and Andreas Kraus, Dieter~Birk.
\newblock Seafloor classification for mine countermeasures operations using
  synthetic aperture sonar images.
\newblock In {\em OCEANS-Aberdeen, 2017 MTS/IEEE}. IEEE, 2017.

\bibitem{krizhevsky2012imagenet}
Alex Krizhevsky, Ilya Sutskever, and Geoffrey~E Hinton.
\newblock Imagenet classification with deep convolutional neural networks.
\newblock In {\em Advances in Neural Information Processing Systems}, pages
  1097--1105, 2012.

\bibitem{kuo2015deepbox}
Weicheng Kuo, Bharath Hariharan, and Jitendra Malik.
\newblock Deepbox: Learning objectness with convolutional networks.
\newblock In {\em Proceedings of the IEEE International Conference on Computer
  Vision}, pages 2479--2487, 2015.

\bibitem{laist1987overview}
David~W Laist.
\newblock Overview of the biological effects of lost and discarded plastic
  debris in the marine environment.
\newblock {\em Marine pollution bulletin}, 18(6):319--326, 1987.

\bibitem{laist1997impacts}
David~W Laist.
\newblock Impacts of marine debris: entanglement of marine life in marine
  debris including a comprehensive list of species with entanglement and
  ingestion records.
\newblock In {\em Marine Debris}, pages 99--139. Springer, 1997.

\bibitem{lecun1998gradient}
Yann LeCun, L{\'e}on Bottou, Yoshua Bengio, and Patrick Haffner.
\newblock Gradient-based learning applied to document recognition.
\newblock {\em Proceedings of the IEEE}, 86(11):2278--2324, 1998.

\bibitem{li2016plastic}
WC~Li, HF~Tse, and L~Fok.
\newblock Plastic waste in the marine environment: A review of sources,
  occurrence and effects.
\newblock {\em Science of the Total Environment}, 566:333--349, 2016.

\bibitem{lin2013network}
Min Lin, Qiang Chen, and Shuicheng Yan.
\newblock Network in network.
\newblock {\em arXiv preprint arXiv:1312.4400}, 2013.

\bibitem{lin2014microsoft}
Tsung-Yi Lin, Michael Maire, Serge Belongie, James Hays, Pietro Perona, Deva
  Ramanan, Piotr Doll{\'a}r, and C~Lawrence Zitnick.
\newblock Microsoft {COCO}: Common objects in context.
\newblock In {\em European conference on computer vision}, pages 740--755.
  Springer, 2014.

\bibitem{liu2018darts}
Hanxiao Liu, Karen Simonyan, and Yiming Yang.
\newblock Darts: Differentiable architecture search.
\newblock {\em arXiv preprint arXiv:1806.09055}, 2018.

\bibitem{long2015fully}
Jonathan Long, Evan Shelhamer, and Trevor Darrell.
\newblock Fully convolutional networks for semantic segmentation.
\newblock In {\em Proceedings of the IEEE Conference on Computer Vision and
  Pattern Recognition}, pages 3431--3440, 2015.

\bibitem{lowe2004distinctive}
David~G Lowe.
\newblock Distinctive image features from scale-invariant keypoints.
\newblock {\em International journal of computer vision}, 60(2):91--110, 2004.

\bibitem{luo2018understanding}
Ping Luo, Xinjiang Wang, Wenqi Shao, and Zhanglin Peng.
\newblock Understanding regularization in batch normalization.
\newblock {\em arXiv preprint arXiv:1809.00846}, 2018.

\bibitem{maaten2008visualizing}
Laurens van~der Maaten and Geoffrey Hinton.
\newblock Visualizing data using t-sne.
\newblock {\em Journal of Machine Learning Research}, 9(Nov):2579--2605, 2008.

\bibitem{mackay2003information}
David~JC MacKay.
\newblock {\em Information theory, inference and learning algorithms}.
\newblock Cambridge university press, 2003.

\bibitem{mcilgorm2008understanding}
A~McIlgorm, HF~Campbell, and MJ~Rule.
\newblock Understanding the economic benefits and costs of controlling marine
  debris in the apec region (mrc 02/2007).
\newblock {\em A report to the Asia-Pacific Economic Cooperation Marine
  Resource Conservation Working Group by the National Marine Science Centre
  (University of New England and Southern Cross University), Coffs Harbour,
  NSW, Australia, December}, 2008.

\bibitem{mishkin2016systematic}
Dmytro Mishkin, Nikolay Sergievskiy, and Jiri Matas.
\newblock Systematic evaluation of cnn advances on the imagenet.
\newblock {\em arXiv preprint arXiv:1606.02228}, 2016.

\bibitem{mori2011survey}
Nobuhito Mori, Tomoyuki Takahashi, Tomohiro Yasuda, and Hideaki Yanagisawa.
\newblock Survey of 2011 tohoku earthquake tsunami inundation and run-up.
\newblock {\em Geophysical research letters}, 38(7), 2011.

\bibitem{murphy2012machine}
Kevin~P Murphy.
\newblock {\em Machine learning: a probabilistic perspective}.
\newblock MIT press, 2012.

\bibitem{myers2010template}
Vincent Myers and John Fawcett.
\newblock A template matching procedure for automatic target recognition in
  synthetic aperture sonar imagery.
\newblock {\em Signal Processing Letters, IEEE}, 17(7):683--686, 2010.

\bibitem{negahdaripour2011dynamic}
Shahriar Negahdaripour, MD~Aykin, and Shayanth Sinnarajah.
\newblock Dynamic scene analysis and mosaicing of benthic habitats by fs sonar
  imaging-issues and complexities.
\newblock In {\em OCEANS'11 MTS/IEEE}, pages 1--7. IEEE, 2011.

\bibitem{nelms2017marine}
SE~Nelms, C~Coombes, LC~Foster, TS~Galloway, BJ~Godley, PK~Lindeque, and
  MJ~Witt.
\newblock Marine anthropogenic litter on british beaches: a 10-year nationwide
  assessment using citizen science data.
\newblock {\em Science of the Total Environment}, 579:1399--1409, 2017.

\bibitem{canyon1978international}
International~Maritime Organization.
\newblock International convention for the prevention of pollution from ships
  (marpol 73/78).
\newblock 1978.

\bibitem{pailhas2010high}
Yan Pailhas, Yvan Petillot, and Chris Capus.
\newblock High-resolution sonars: what resolution do we need for target
  recognition?
\newblock {\em EURASIP Journal on Advances in Signal Processing},
  2010(1):205095, 2010.

\bibitem{paleczny2015population}
Michelle Paleczny, Edd Hammill, Vasiliki Karpouzi, and Daniel Pauly.
\newblock Population trend of the world’s monitored seabirds, 1950-2010.
\newblock {\em PLoS One}, 10(6):e0129342, 2015.

\bibitem{pham2013guided}
Minh~T{\^a}n Pham and Didier Gu{\'e}riot.
\newblock Guided block-matching for sonar image registration using unsupervised
  kohonen neural networks.
\newblock In {\em 2013 OCEANS San Diego}, pages 1--5. IEEE, 2013.

\bibitem{qi2017pointnet}
Charles~R Qi, Hao Su, Kaichun Mo, and Leonidas~J Guibas.
\newblock Pointnet: Deep learning on point sets for 3d classification and
  segmentation.
\newblock {\em Proc. Computer Vision and Pattern Recognition (CVPR), IEEE},
  1(2):4, 2017.

\bibitem{quanming2018taking}
Yao Quanming, Wang Mengshuo, Jair~Escalante Hugo, Guyon Isabelle, Hu~Yi-Qi,
  Li~Yu-Feng, Tu~Wei-Wei, Yang Qiang, and Yu~Yang.
\newblock Taking human out of learning applications: A survey on automated
  machine learning.
\newblock {\em arXiv preprint arXiv:1810.13306}, 2018.

\bibitem{rahtu2011learning}
Esa Rahtu, Juho Kannala, and Matthew Blaschko.
\newblock Learning a category independent object detection cascade.
\newblock In {\em Computer Vision (ICCV), 2011 IEEE International Conference
  on}, pages 1052--1059. IEEE, 2011.

\bibitem{reddi2018convergence}
Sashank~J Reddi, Satyen Kale, and Sanjiv Kumar.
\newblock On the convergence of adam and beyond.
\newblock In {\em Proceedings of ICLR'18}, 2018.

\bibitem{redmon2016you}
Joseph Redmon, Santosh Divvala, Ross Girshick, and Ali Farhadi.
\newblock You only look once: Unified, real-time object detection.
\newblock In {\em Proceedings of the IEEE Conference on Computer Vision and
  Pattern Recognition}, pages 779--788, 2016.

\bibitem{reed2004automated}
Scott Reed, Yvan Petillot, and J~Bell.
\newblock Automated approach to classification of mine-like objects in sidescan
  sonar using highlight and shadow information.
\newblock {\em IEE Proceedings-Radar, Sonar and Navigation}, 151(1):48--56,
  2004.

\bibitem{ren2015faster}
Shaoqing Ren, Kaiming He, Ross Girshick, and Jian Sun.
\newblock Faster r-cnn: Towards real-time object detection with region proposal
  networks.
\newblock In {\em Advances in Neural Information Processing Systems}, pages
  91--99, 2015.

\bibitem{ridao2014intervention}
Pere Ridao, Marc Carreras, David Ribas, Pedro~J Sanz, and Gabriel Oliver.
\newblock Intervention auvs: the next challenge.
\newblock {\em IFAC Proceedings Volumes}, 47(3):12146--12159, 2014.

\bibitem{rochman2016strategies}
Chelsea~M Rochman.
\newblock Strategies for reducing ocean plastic debris should be diverse and
  guided by science.
\newblock {\em Environmental Research Letters}, 11(4):041001, 2016.

\bibitem{rublee2011orb}
Ethan Rublee, Vincent Rabaud, Kurt Konolige, and Gary Bradski.
\newblock Orb: An efficient alternative to sift or surf.
\newblock In {\em 2011 International conference on computer vision}, pages
  2564--2571. IEEE, 2011.

\bibitem{ruder2016overview}
Sebastian Ruder.
\newblock An overview of gradient descent optimization algorithms.
\newblock {\em arXiv preprint arXiv:1609.04747}, 2016.

\bibitem{ruder2017overview}
Sebastian Ruder.
\newblock An overview of multi-task learning in deep neural networks.
\newblock {\em arXiv preprint arXiv:1706.05098}, 2017.

\bibitem{russakovsky2015imagenet}
Olga Russakovsky, Jia Deng, Hao Su, Jonathan Krause, Sanjeev Satheesh, Sean Ma,
  Zhiheng Huang, Andrej Karpathy, Aditya Khosla, Michael Bernstein, et~al.
\newblock Imagenet large scale visual recognition challenge.
\newblock {\em International Journal of Computer Vision}, 115(3):211--252,
  2015.

\bibitem{ryan2014litter}
Peter~G Ryan.
\newblock Litter survey detects the south atlantic ‘garbage patch’.
\newblock {\em Marine Pollution Bulletin}, 79(1-2):220--224, 2014.

\bibitem{sammut2011encyclopedia}
Claude Sammut and Geoffrey~I Webb.
\newblock {\em Encyclopedia of machine learning}.
\newblock Springer Science \& Business Media, 2011.

\bibitem{santurkar2018does}
Shibani Santurkar, Dimitris Tsipras, Andrew Ilyas, and Aleksander Madry.
\newblock How does batch normalization help optimization?(no, it is not about
  internal covariate shift).
\newblock {\em arXiv preprint arXiv:1805.11604}, 2018.

\bibitem{sawas2012cascade}
Jamil Sawas and Yvan Petillot.
\newblock Cascade of boosted classifiers for automatic target recognition in
  synthetic aperture sonar imagery.
\newblock In {\em Proceedings of Meetings on Acoustics ECUA2012}. ASA, 2012.

\bibitem{sawas2010cascade}
Jamil Sawas, Yvan Petillot, and Yan Pailhas.
\newblock Cascade of boosted classifiers for rapid detection of underwater
  objects.
\newblock In {\em Proceedings of the European Conference on Underwater
  Acoustics}, 2010.

\bibitem{saxe2013exact}
Andrew~M Saxe, James~L McClelland, and Surya Ganguli.
\newblock Exact solutions to the nonlinear dynamics of learning in deep linear
  neural networks.
\newblock {\em arXiv preprint arXiv:1312.6120}, 2013.

\bibitem{schlining2013debris}
Kyra Schlining, Susan Von~Thun, Linda Kuhnz, Brian Schlining, Lonny Lundsten,
  Nancy~Jacobsen Stout, Lori Chaney, and Judith Connor.
\newblock Debris in the deep: Using a 22-year video annotation database to
  survey marine litter in monterey canyon, central california, usa.
\newblock {\em Deep Sea Research Part I: Oceanographic Research Papers},
  79:96--105, 2013.

\bibitem{Schroff_2015_CVPR}
Florian Schroff, Dmitry Kalenichenko, and James Philbin.
\newblock Facenet: A unified embedding for face recognition and clustering.
\newblock In {\em The IEEE Conference on Computer Vision and Pattern
  Recognition (CVPR)}, June 2015.

\bibitem{sermanet2013overfeat}
Pierre Sermanet, David Eigen, Xiang Zhang, Micha{\"e}l Mathieu, Rob Fergus, and
  Yann LeCun.
\newblock Overfeat: Integrated recognition, localization and detection using
  convolutional networks.
\newblock {\em arXiv preprint arXiv:1312.6229}, 2013.

\bibitem{sharif2014cnn}
Ali Sharif~Razavian, Hossein Azizpour, Josephine Sullivan, and Stefan Carlsson.
\newblock Cnn features off-the-shelf: an astounding baseline for recognition.
\newblock In {\em Proceedings of IEEE CVPR workshops}, pages 806--813, 2014.

\bibitem{sheavly2007marine}
SB~Sheavly and KM~Register.
\newblock Marine debris \& plastics: environmental concerns, sources, impacts
  and solutions.
\newblock {\em Journal of Polymers and the Environment}, 15(4):301--305, 2007.

\bibitem{shen2017dsod}
Zhiqiang Shen, Zhuang Liu, Jianguo Li, Yu-Gang Jiang, Yurong Chen, and
  Xiangyang Xue.
\newblock Dsod: Learning deeply supervised object detectors from scratch.
\newblock In {\em The IEEE International Conference on Computer Vision (ICCV)},
  volume~3, page~7, 2017.

\bibitem{simonyan2014very}
Karen Simonyan and Andrew Zisserman.
\newblock Very deep convolutional networks for large-scale image recognition.
\newblock {\em arXiv preprint arXiv:1409.1556}, 2014.

\bibitem{smith1997marine}
V~Kerry Smith, Xiaolong Zhang, and Raymond~B Palmquist.
\newblock Marine debris, beach quality, and non-market values.
\newblock {\em Environmental and Resource Economics}, 10(3):223--247, 1997.

\bibitem{arisExplorer3K}
SoundMetrics.
\newblock {ARIS} {E}xplorer 3000: {S}ee what others can't, 2018.
\newblock Accessed 1-9-2018. Available at \url{
  http://www.soundmetrics.com/products/aris-sonars/ARIS-Explorer-3000/015335\_Rev
  C \_ARIS-Explorer-3000_Brochure}.

\bibitem{spengler2008methods}
Angela Spengler and Monica~F Costa.
\newblock Methods applied in studies of benthic marine debris.
\newblock {\em Marine Pollution Bulletin}, 56(2):226--230, 2008.

\bibitem{srivastava2014dropout}
Nitish Srivastava, Geoffrey Hinton, Alex Krizhevsky, Ilya Sutskever, and Ruslan
  Salakhutdinov.
\newblock Dropout: A simple way to prevent neural networks from overfitting.
\newblock {\em The Journal of Machine Learning Research}, 15(1):1929--1958,
  2014.

\bibitem{stanley2002evolving}
Kenneth~O Stanley and Risto Miikkulainen.
\newblock Evolving neural networks through augmenting topologies.
\newblock {\em Evolutionary computation}, 10(2):99--127, 2002.

\bibitem{stefatos1999marine}
A~Stefatos, M~Charalampakis, G~Papatheodorou, and G~Ferentinos.
\newblock Marine debris on the seafloor of the mediterranean sea: examples from
  two enclosed gulfs in western greece.
\newblock {\em Marine Pollution Bulletin}, 38(5):389--393, 1999.

\bibitem{sun2017revisiting}
Chen Sun, Abhinav Shrivastava, Saurabh Singh, and Abhinav Gupta.
\newblock Revisiting unreasonable effectiveness of data in deep learning era.
\newblock In {\em Proceedings of the IEEE Conference on Computer Vision and
  Pattern Recognition}, pages 843--852, 2017.

\bibitem{szegedy2015going}
Christian Szegedy, Wei Liu, Yangqing Jia, Pierre Sermanet, Scott Reed, Dragomir
  Anguelov, Dumitru Erhan, Vincent Vanhoucke, and Andrew Rabinovich.
\newblock Going deeper with convolutions.
\newblock In {\em Proceedings of IEEE CVPR}, pages 1--9, 2015.

\bibitem{szeliski2010computer}
Richard Szeliski.
\newblock {\em Computer vision: algorithms and applications}.
\newblock Springer Science \& Business Media, 2010.

\bibitem{tieleman2012lecture}
Tijmen Tieleman and Geoffrey Hinton.
\newblock Lecture 6.5-rmsprop: Divide the gradient by a running average of its
  recent magnitude.
\newblock {\em COURSERA: Neural networks for machine learning}, 4(2):26--31,
  2012.

\bibitem{tola2008fast}
Engin Tola, Vincent Lepetit, and Pascal Fua.
\newblock A fast local descriptor for dense matching.
\newblock In {\em Computer Vision and Pattern Recognition, 2008. CVPR 2008.
  IEEE Conference on}, pages 1--8. IEEE, 2008.

\bibitem{uijlings2013selective}
Jasper~RR Uijlings, Koen~EA van~de Sande, Theo Gevers, and Arnold~WM Smeulders.
\newblock Selective search for object recognition.
\newblock {\em International journal of computer vision}, 104(2):154--171,
  2013.

\bibitem{valdenegro2016object}
Matias Valdenegro-Toro.
\newblock Object recognition in forward-looking sonar images with convolutional
  neural networks.
\newblock In {\em OCEANS 2016 MTS/IEEE Monterey}, pages 1--6. IEEE, 2016.

\bibitem{valdenegro2017limits}
Matias Valdenegro-Toro.
\newblock {B}est {P}ractices in {C}onvolutional {N}etworks for
  {F}orward-{L}ooking {S}onar {I}mage {R}ecognition.
\newblock In {\em OCEANS 2017 MTS/IEEE Aberdeen}. IEEE, 2017.

\bibitem{valdenegro2017rtcnns}
Matias Valdenegro-Toro.
\newblock {R}eal-time convolutional networks for sonar image classification in
  low-power embedded systems.
\newblock In {\em European Symposium on Artificial Neural Networks,
  Computational Intelligence and Machine Learning (ESANN)}, 2017.

\bibitem{vandrish2011side}
Peter Vandrish, Andrew Vardy, Dan Walker, and OA~Dobre.
\newblock Side-scan sonar image registration for auv navigation.
\newblock In {\em Underwater Technology (UT), 2011 IEEE Symposium on and 2011
  Workshop on Scientific Use of Submarine Cables and Related Technologies
  (SSC)}, pages 1--7. IEEE, 2011.

\bibitem{viola2001rapid}
Paul Viola and Michael Jones.
\newblock Rapid object detection using a boosted cascade of simple features.
\newblock In {\em Computer Vision and Pattern Recognition, 2001. CVPR 2001.
  Proceedings of the 2001 IEEE Computer Society Conference on}, volume~1, pages
  I--I. IEEE, 2001.

\bibitem{weis2015marine}
Judith~S Weis.
\newblock {\em Marine pollution: what everyone needs to know}.
\newblock Oxford University Press, 2015.

\bibitem{wilcox2015threat}
Chris Wilcox, Erik Van~Sebille, and Britta~Denise Hardesty.
\newblock Threat of plastic pollution to seabirds is global, pervasive, and
  increasing.
\newblock {\em Proceedings of the National Academy of Sciences},
  112(38):11899--11904, 2015.

\bibitem{williams2016underwater}
David~P Williams.
\newblock Underwater target classification in synthetic aperture sonar imagery
  using deep convolutional neural networks.
\newblock In {\em Pattern Recognition (ICPR), 2016 23rd International
  Conference on}, pages 2497--2502. IEEE, 2016.

\bibitem{williams2018underwater}
David~P Williams.
\newblock Convolutional neural network transfer learning for underwater object
  classification.
\newblock In {\em Proceedings of the Institute of Acoustic}, 2018.

\bibitem{willis2017differentiating}
Kathryn Willis, Britta~Denise Hardesty, Lorne Kriwoken, and Chris Wilcox.
\newblock Differentiating littering, urban runoff and marine transport as
  sources of marine debris in coastal and estuarine environments.
\newblock {\em Scientific Reports}, 7:44479, 2017.

\bibitem{wu2010beam}
Ju~Wu and Hongyu Bian.
\newblock Beam-forming and imaging using acoustic lenses: Some simulation and
  experimental results.
\newblock In {\em Signal Processing Systems (ICSPS), 2010 2nd International
  Conference on}, volume~2, pages V2--764. IEEE, 2010.

\bibitem{wu2018group}
Yuxin Wu and Kaiming He.
\newblock Group normalization.
\newblock {\em arXiv preprint arXiv:1803.08494}, 2018.

\bibitem{yi2016lift}
Kwang~Moo Yi, Eduard Trulls, Vincent Lepetit, and Pascal Fua.
\newblock Lift: Learned invariant feature transform.
\newblock In {\em European Conference on Computer Vision}, pages 467--483.
  Springer, 2016.

\bibitem{yilmaz2006object}
Alper Yilmaz, Omar Javed, and Mubarak Shah.
\newblock Object tracking: A survey.
\newblock {\em Acm computing surveys (CSUR)}, 38(4):13, 2006.

\bibitem{zagoruyko2015learning}
Sergey Zagoruyko and Nikos Komodakis.
\newblock Learning to compare image patches via convolutional neural networks.
\newblock In {\em Proceedings of the IEEE Conference on Computer Vision and
  Pattern Recognition}, pages 4353--4361, 2015.

\bibitem{zbontar2016stereo}
Jure Zbontar and Yann LeCun.
\newblock Stereo matching by training a convolutional neural network to compare
  image patches.
\newblock {\em Journal of Machine Learning Research}, 17:1--32, 2016.

\bibitem{zeiler2014visualizing}
Matthew~D Zeiler and Rob Fergus.
\newblock Visualizing and understanding convolutional networks.
\newblock In {\em European conference on computer vision}, pages 818--833.
  Springer, 2014.

\bibitem{zhang2016understanding}
Chiyuan Zhang, Samy Bengio, Moritz Hardt, Benjamin Recht, and Oriol Vinyals.
\newblock Understanding deep learning requires rethinking generalization.
\newblock {\em arXiv preprint arXiv:1611.03530}, 2016.

\bibitem{zhang1995robust}
Zhengyou Zhang, Rachid Deriche, Olivier Faugeras, and Quang-Tuan Luong.
\newblock A robust technique for matching two uncalibrated images through the
  recovery of the unknown epipolar geometry.
\newblock {\em Artificial intelligence}, 78(1-2):87--119, 1995.

\bibitem{zhu2017deeplearning}
P.~Zhu, J.~Isaacs, B.~Fu, and S.~Ferrari.
\newblock Deep learning feature extraction for target recognition and
  classification in underwater sonar images.
\newblock In {\em 2017 IEEE 56th Annual Conference on Decision and Control
  (CDC)}, pages 2724--2731, Dec 2017.

\bibitem{zitnick2014edge}
C~Lawrence Zitnick and Piotr Doll{\'a}r.
\newblock Edge boxes: Locating object proposals from edges.
\newblock In {\em Computer Vision--ECCV 2014}, pages 391--405. Springer, 2014.

\bibitem{zoph2018learning}
Barret Zoph, Vijay Vasudevan, Jonathon Shlens, and Quoc~V Le.
\newblock Learning transferable architectures for scalable image recognition.
\newblock In {\em Proceedings of the IEEE conference on computer vision and
  pattern recognition}, pages 8697--8710, 2018.

\end{thebibliography}
